\documentclass{article}

\usepackage{styles/debstyle}   %
\usepackage{styles/mistill}    %

\usepackage{styles/iclr2027_conference,times}

\usepackage[utf8]{inputenc}
\usepackage[T1]{fontenc}
\usepackage{microtype}
\usepackage{hyperref}
\usepackage{url}
\usepackage{booktabs}
\usepackage{amsfonts}
\usepackage{nicefrac}
\usepackage{xcolor}

\usepackage[hang,flushmargin]{footmisc}

\definecolor{darkblue}{rgb}{0, 0, 0.5}
\hypersetup{
  colorlinks=true,
  citecolor=darkblue,
  linkcolor=darkblue,
  urlcolor=darkblue,
  pdfborder={0 0 0}   %
}
\hypersetup{pdftitle={Characterize Then Distill: Mechanistic Reasoning in Large Output Spaces},pdfauthor={Debjyoti Saha Roy}}

\title{Characterize Then Distill:\\
Mechanistic Reasoning in Large Output Spaces}

\author{%
Debjyoti Saha Roy \\
Khoury College of Computer Sciences,
Northeastern University \\
\texttt{saharoy.d@northeastern.edu}
}

\draftmodefalse
\wikifalse
\ifdefined\FINALBUILD\draftmodefalse\fi
\ifdefined\WIKIBUILD\wikitrue\fi

\providecommand{\gensd}[2]{\begin{tabular}[t]{@{}c@{}}#1\\[-2pt]{\scriptsize$\pm$#2}\end{tabular}}
\providecommand{\genstack}[2]{\begin{tabular}[t]{@{}c@{}}#1\\[-2pt]{\scriptsize #2}\end{tabular}}
\providecommand{\genstackl}[2]{\begin{tabular}[t]{@{}l@{}}#1\\[-2pt]{\scriptsize #2}\end{tabular}}
\providecommand{\genhead}[2]{\begin{tabular}[b]{@{}c@{}}#1\\#2\end{tabular}}
\providecommand{\genstackiii}[3]{\begin{tabular}[t]{@{}c@{}}#1\\[-2pt]{\scriptsize #2}\\[-3pt]{\scriptsize #3}\end{tabular}}

\newcommand{\NEvPOneCount}{2{,}064}
\newcommand{\NEvPOneMargin}{+6.02}

\newcommand{\NEvPTwoCount}{7{,}688}
\newcommand{\NEvPTwoMargin}{+2.76}

\newcommand{\NEvPTwoCCount}{3{,}855}
\newcommand{\NEvPTwoCMargin}{+3.38}

\newcommand{\NEvNullCount}{2{,}000}
\newcommand{\NEvNullMargin}{\ensuremath{-}4.95}

\newcommand{\NFaithPOneNecKThreeGlobal}{0.24}

\newcommand{\NFaithPOneNecKThreeOwn}{0.30}
\newcommand{\NFaithPOneNecKThreeWins}{92}
\newcommand{\NFaithPTwoNecKThreeGlobal}{0.41}

\newcommand{\NFaithPTwoNecKThreeOwn}{0.55}

\newcommand{\NFaithPTwoCNecKThreeGlobal}{0.26}

\newcommand{\NFaithPTwoCNecKThreeOwn}{0.45}

\newcommand{\NFaithPOneNecKEightGlobal}{0.62}

\newcommand{\NFaithPOneNecKEightBottom}{\ensuremath{-}0.36}
\newcommand{\NFaithPOneNecKEightOwn}{0.63}

\newcommand{\NFaithPTwoNecKEightGlobal}{0.78}

\newcommand{\NFaithPTwoNecKEightOwn}{1.10}

\newcommand{\NFaithPTwoCNecKEightGlobal}{0.63}

\newcommand{\NFaithPTwoCNecKEightOwn}{0.75}
\newcommand{\NFaithPTwoCNecKEightWins}{90}
\newcommand{\NFaithPOneNecKSixteenGlobal}{0.88}

\newcommand{\NFaithPOneNecKSixteenOwn}{0.91}

\newcommand{\NFaithPTwoNecKSixteenGlobal}{1.10}

\newcommand{\NFaithPTwoNecKSixteenOwn}{1.38}

\newcommand{\NFaithPTwoCNecKSixteenGlobal}{0.72}

\newcommand{\NFaithPTwoCNecKSixteenOwn}{0.84}

\newcommand{\NFaithPOneNecKFortyGlobal}{1.24}

\newcommand{\NFaithPOneNecKFortyBottom}{0.89}
\newcommand{\NFaithPOneNecKFortyOwn}{1.26}
\newcommand{\NFaithPOneNecKFortyWins}{100}
\newcommand{\NFaithPTwoNecKFortyGlobal}{1.61}

\newcommand{\NFaithPTwoNecKFortyOwn}{1.81}

\newcommand{\NFaithPTwoCNecKFortyGlobal}{0.86}

\newcommand{\NFaithPTwoCNecKFortyOwn}{0.97}

\newcommand{\NFaithPOneNecKOneHundredGlobal}{1.60}

\newcommand{\NFaithPOneNecKOneHundredBottom}{1.57}
\newcommand{\NFaithPOneNecKOneHundredOwn}{1.45}

\newcommand{\NFaithPTwoNecKOneHundredGlobal}{2.21}

\newcommand{\NFaithPTwoNecKOneHundredOwn}{2.14}

\newcommand{\NFaithPTwoCNecKOneHundredGlobal}{0.89}

\newcommand{\NFaithPTwoCNecKOneHundredOwn}{1.00}

\newcommand{\NFaithPOneKiKThreeGlobal}{0.16}

\newcommand{\NFaithPOneKiKThreeBottom}{\ensuremath{-}0.16}
\newcommand{\NFaithPOneKiKThreeOwn}{0.18}

\newcommand{\NFaithPTwoKiKThreeGlobal}{0.24}

\newcommand{\NFaithPTwoKiKThreeOwn}{0.33}

\newcommand{\NFaithPTwoCKiKThreeGlobal}{0.05}

\newcommand{\NFaithPTwoCKiKThreeOwn}{0.35}
\newcommand{\NFaithPTwoCKiKThreeWins}{80}
\newcommand{\NFaithPOneKiKEightGlobal}{0.67}

\newcommand{\NFaithPOneKiKEightOwn}{0.45}
\newcommand{\NFaithPOneKiKEightWins}{100}
\newcommand{\NFaithPTwoKiKEightGlobal}{0.48}

\newcommand{\NFaithPTwoKiKEightOwn}{0.59}

\newcommand{\NFaithPTwoCKiKEightGlobal}{0.69}

\newcommand{\NFaithPTwoCKiKEightOwn}{0.63}
\newcommand{\NFaithPTwoCKiKEightWins}{92}
\newcommand{\NFaithPOneKiKSixteenGlobal}{0.76}

\newcommand{\NFaithPOneKiKSixteenOwn}{0.61}

\newcommand{\NFaithPTwoKiKSixteenGlobal}{0.64}

\newcommand{\NFaithPTwoKiKSixteenOwn}{0.69}

\newcommand{\NFaithPTwoCKiKSixteenGlobal}{0.73}

\newcommand{\NFaithPTwoCKiKSixteenOwn}{0.76}

\newcommand{\NFaithPOneKiKFortyGlobal}{0.86}
\newcommand{\NFaithPOneKiKFortyRandom}{0.04}

\newcommand{\NFaithPOneKiKFortyOwn}{0.78}
\newcommand{\NFaithPOneKiKFortyWins}{100}
\newcommand{\NFaithPTwoKiKFortyGlobal}{0.79}
\newcommand{\NFaithPTwoKiKFortyRandom}{0.02}
\newcommand{\NFaithPTwoKiKFortyBottom}{\ensuremath{-}0.81}
\newcommand{\NFaithPTwoKiKFortyOwn}{0.82}
\newcommand{\NFaithPTwoKiKFortyWins}{100}
\newcommand{\NFaithPTwoCKiKFortyGlobal}{0.85}
\newcommand{\NFaithPTwoCKiKFortyRandom}{0.02}

\newcommand{\NFaithPTwoCKiKFortyOwn}{0.87}
\newcommand{\NFaithPTwoCKiKFortyWins}{92}
\newcommand{\NFaithPOneKiKOneHundredGlobal}{0.96}
\newcommand{\NFaithPOneKiKOneHundredRandom}{0.09}

\newcommand{\NFaithPOneKiKOneHundredOwn}{0.89}

\newcommand{\NFaithPTwoKiKOneHundredGlobal}{0.89}

\newcommand{\NFaithPTwoKiKOneHundredOwn}{0.91}

\newcommand{\NFaithPTwoCKiKOneHundredGlobal}{0.94}

\newcommand{\NFaithPTwoCKiKOneHundredOwn}{0.98}

\newcommand{\NFaithPOneCorrMargin}{8.7}
\newcommand{\NFaithPTwoCorrMargin}{3.6}
\newcommand{\NFaithPTwoCCorrMargin}{0.4}

\newcommand{\NHeadLThirtyNineHThirtyEightPOneIc}{0.91}

\newcommand{\NHeadLThirtyNineHThirtyEightPTwoIc}{0.97}

\newcommand{\NHeadLThirtyNineHThirtyEightPTwoCIc}{0.62}
\newcommand{\NHeadLThirtyNineHThirtyEightPTwoCRank}{1}
\newcommand{\NHeadLThirtyNineHThirtyEightPTwoCExact}{0.22}

\newcommand{\NHeadLTwentySixHThirtySixPOneIc}{0.43}

\newcommand{\NHeadLTwentySixHThirtySevenPOneIc}{0.29}

\newcommand{\NHeadLTwentySevenHElevenPOneIc}{0.24}

\newcommand{\NHeadLTwentySevenHThirteenPOneIc}{0.21}

\newcommand{\NHeadLThirtyNineHTenPOneIc}{0.18}

\newcommand{\NHeadLThirtyNineHTenPTwoCRank}{7}
\newcommand{\NHeadLThirtyNineHTenPTwoCExact}{0.36}

\newcommand{\NHeadLThirtyNineHTwelvePTwoCRank}{6}
\newcommand{\NHeadLThirtyNineHTwelvePTwoCExact}{0.64}

\newcommand{\NHeadLThirtyEightHTwentyTwoPTwoCIc}{0.22}

\newcommand{\NHeadLThirtyHElevenPOneRank}{1{,}500}

\newcommand{\NHeadLThirtyHElevenPTwoCIc}{0.51}
\newcommand{\NHeadLThirtyHElevenPTwoCRank}{2}
\newcommand{\NHeadLThirtyHElevenPTwoCExact}{0.28}

\newcommand{\NHeadLThirtyEightHThirtyNinePTwoCIc}{0.19}
\newcommand{\NHeadLThirtyEightHThirtyNinePTwoCRank}{9}
\newcommand{\NHeadLThirtyEightHThirtyNinePTwoCExact}{0.12}

\newcommand{\NHeadPOneTopFortyLayersZeroToElevenCount}{0}

\newcommand{\NHeadPOneTopFortyLayersTwelveToTwentyThreeCount}{2}

\newcommand{\NHeadPOneTopFortyLayersTwentyFourToThirtyOneCount}{24}
\newcommand{\NHeadPOneTopFortyLayersTwentyFourToThirtyOneShare}{53\%}
\newcommand{\NHeadPOneTopFortyLayersThirtyTwoToThirtyNineCount}{14}
\newcommand{\NHeadPOneTopFortyLayersThirtyTwoToThirtyNineShare}{45\%}
\newcommand{\NHeadCorrPOnePTwo}{0.75}
\newcommand{\NHeadOverlapFortyPOnePTwo}{25}
\newcommand{\NHeadCorrPOnePTwoC}{0.63}
\newcommand{\NHeadOverlapFortyPOnePTwoC}{24}
\newcommand{\NHeadCorrPTwoPTwoC}{0.80}
\newcommand{\NHeadOverlapFortyPTwoPTwoC}{26}
\newcommand{\NHeadPOneBhPos}{429}
\newcommand{\NHeadPTwoBhPos}{543}
\newcommand{\NHeadPTwoCBhPos}{340}

\newcommand{\NReadPOneTopLift}{3.54}
\newcommand{\NReadPOneRandLift}{0.87}
\newcommand{\NReadPOneBottomLift}{3.11}
\newcommand{\NReadPOneWins}{100}
\newcommand{\NReadPTwoTopLift}{4.08}
\newcommand{\NReadPTwoRandLift}{1.32}
\newcommand{\NReadPTwoWins}{100}
\newcommand{\NReadPTwoCTopLift}{3.94}
\newcommand{\NReadPTwoCRandLift}{1.89}
\newcommand{\NReadPTwoCBottomLift}{3.61}
\newcommand{\NReadPTwoCWins}{92}
\newcommand{\NReadLTwentySixHThirtySixRegion}{0.47}
\newcommand{\NReadLTwentySixHThirtySixProse}{0.03}
\newcommand{\NReadLTwentySixHThirtySixLift}{9.4}
\newcommand{\NReadLTwentySixHThirtySixR}{+0.44}
\newcommand{\NReadLTwentyEightHTwentyOneRegion}{0.45}
\newcommand{\NReadLTwentyEightHTwentyOneProse}{0.01}
\newcommand{\NReadLTwentyEightHTwentyOneLift}{9.1}
\newcommand{\NReadLTwentyEightHTwentyOneR}{+0.35}
\newcommand{\NReadLTwentySixHThirtySevenRegion}{0.17}
\newcommand{\NReadLTwentySixHThirtySevenProse}{0.34}
\newcommand{\NReadLTwentySixHThirtySevenLift}{8.7}
\newcommand{\NReadLTwentySixHThirtySevenR}{+0.29}
\newcommand{\NReadLThirtyNineHThirtyEightRegion}{0.00}
\newcommand{\NReadLThirtyNineHThirtyEightProse}{0.84}
\newcommand{\NReadLThirtyNineHThirtyEightLift}{1.1}
\newcommand{\NReadLThirtyNineHThirtyEightR}{+0.06}
\newcommand{\NReadLThirtyNineHThirteenRegion}{0.00}
\newcommand{\NReadLThirtyNineHThirteenProse}{0.73}
\newcommand{\NReadLThirtyNineHThirteenLift}{1.2}
\newcommand{\NReadLThirtyNineHThirteenR}{\ensuremath{-}0.10}
\newcommand{\NReadLThirtyEightHTwentyFourRegion}{0.01}
\newcommand{\NReadLThirtyEightHTwentyFourProse}{0.79}
\newcommand{\NReadLThirtyEightHTwentyFourLift}{1.0}
\newcommand{\NReadLThirtyEightHTwentyFourR}{\ensuremath{-}0.12}
\newcommand{\NReadLTwentySevenHElevenRegion}{0.00}
\newcommand{\NReadLTwentySevenHElevenProse}{0.78}
\newcommand{\NReadLTwentySevenHElevenLift}{2.5}
\newcommand{\NReadLTwentySevenHElevenR}{\ensuremath{-}0.01}
\newcommand{\NReadLThirtyHElevenSibling}{0.33}
\newcommand{\NReadLThirtyHElevenOwn}{0.13}
\newcommand{\NReadLThirtyHElevenPTwoCR}{+0.30}
\newcommand{\NReadLThirtyEightHThirtyNineSibling}{0.14}
\newcommand{\NReadLThirtyEightHThirtyNineOwn}{0.02}

\newcommand{\NReadLThirtyEightHTwentyTwoSibling}{0.21}
\newcommand{\NReadLThirtyEightHTwentyTwoOwn}{0.15}

\newcommand{\NReadPTwoCRandSibling}{0.02}

\newcommand{\NLocusWindowPct}{36\%}

\newcommand{\NLocusInheritedPct}{29\%}

\newcommand{\NLocusProbeFailPct}{32\%}

\newcommand{\NLocusAtQPct}{2\%}

\newcommand{\NLocusDMedian}{7}
\newcommand{\NLocusFirstCodeD}{23}
\newcommand{\NLocusJump}{8.3}

\newcommand{\NLocusOwnDescPct}{32\%}
\newcommand{\NLocusOwnDescLift}{14}
\newcommand{\NLocusOverlapEmissionKEight}{0}
\newcommand{\NLocusOverlapEmissionKSixteen}{1}
\newcommand{\NLocusOverlapEmissionKForty}{9}
\newcommand{\NLocusTopFortyLayersZeroToEleven}{0}
\newcommand{\NLocusTopFortyLayersTwelveToTwentyThree}{31}
\newcommand{\NLocusTopFortyLayersTwentyFourToThirtyOne}{9}
\newcommand{\NLocusTopFortyLayersThirtyTwoToThirtyNine}{0}
\newcommand{\NLocusProseWinKEight}{1.00}
\newcommand{\NLocusRandWinKEight}{0.02}
\newcommand{\NLocusProseWinKEightWins}{100}
\newcommand{\NLocusProseWinKSixteen}{1.32}
\newcommand{\NLocusRandWinKSixteen}{0.01}
\newcommand{\NLocusProseWinKSixteenWins}{100}
\newcommand{\NLocusProseWinKForty}{1.69}
\newcommand{\NLocusRandWinKForty}{0.07}
\newcommand{\NLocusProseWinKFortyWins}{99}
\newcommand{\NLocusOwnSetWinKSixteen}{1.08}
\newcommand{\NLocusEmissionWinKSixteen}{0.16}
\newcommand{\NLocusReadProseNote}{0.25}
\newcommand{\NLocusReadProseRegion}{0.08}
\newcommand{\NLocusReadProseProseText}{0.42}
\newcommand{\NLocusReadEmissionNote}{0.20}
\newcommand{\NLocusReadEmissionRegion}{0.06}
\newcommand{\NLocusReadEmissionProseText}{0.36}
\newcommand{\NLocusReadRandomNote}{0.08}
\newcommand{\NLocusReadRandomRegion}{0.05}
\newcommand{\NLocusReadRandomProseText}{0.27}
\newcommand{\NLocusTopHeadOne}{\head{24}{19}}
\newcommand{\NLocusTopHeadTwo}{\head{22}{16}}
\newcommand{\NLocusTopHeadThree}{\head{24}{1}}

\newcommand{\NShortPOneSpearman}{0.55}
\newcommand{\NShortPTwoSpearman}{0.63}
\newcommand{\NShortPTwoCSpearman}{0.55}
\newcommand{\NShortPOneOverlapForty}{31}
\newcommand{\NShortPTwoOverlapForty}{31}
\newcommand{\NShortPTwoCOverlapForty}{32}
\newcommand{\NShortPOneOverlapEight}{5}
\newcommand{\NShortPTwoOverlapEight}{6}
\newcommand{\NShortPTwoCOverlapEight}{7}
\newcommand{\NShortPOneKiKEightFull}{0.67}
\newcommand{\NShortPOneKiKEightShort}{0.47}

\newcommand{\NShortPOneKiKFortyShort}{0.92}

\newcommand{\NShortPTwoKiKFortyShort}{0.82}

\newcommand{\NShortPTwoCKiKFortyShort}{0.85}

\newcommand{\NShortPOneKiKFortyRandShort}{0.04}

\newcommand{\NShortPTwoKiKFortyRandShort}{0.03}

\newcommand{\NShortPTwoCKiKFortyRandShort}{0.02}

\newcommand{\NShortReadPOneRegionFull}{0.10}
\newcommand{\NShortReadPOneRegionShort}{0.14}
\newcommand{\NShortReadPOneLiftFull}{3.54}
\newcommand{\NShortReadPOneLiftShort}{3.04}
\newcommand{\NShortReadPTwoCNearMissFull}{0.074}

\newcommand{\NShortLocusDMedianFull}{7}
\newcommand{\NShortLocusDMedianShort}{9}
\newcommand{\NShortLocusWindowFull}{0.36}
\newcommand{\NShortLocusWindowShort}{0.27}
\newcommand{\NShortLocusInheritedFull}{0.29}
\newcommand{\NShortLocusInheritedShort}{0.35}
\newcommand{\NShortLocusProbeFailFull}{0.32}
\newcommand{\NShortLocusProbeFailShort}{0.37}
\newcommand{\NShortLocusProseSpearman}{0.79}
\newcommand{\NShortLocusProseOverlapForty}{31}

\newcommand{\NShortLocusLocusWinSixteenShort}{1.04}

\newcommand{\NShortLocusEmissionWinSixteenShort}{0.19}
\newcommand{\NShortLocusProseWinEightFull}{1.00}
\newcommand{\NShortLocusProseWinEightShort}{0.73}

\newcommand{\NQwenBaseSZeroMicroF}{0.104}
\newcommand{\NQwenBaseSZeroMicroP}{0.180}
\newcommand{\NQwenBaseSZeroMicroR}{0.073}

\newcommand{\NQwenBaseSZeroValid}{0.987}
\newcommand{\NQwenBaseSZeroHitCap}{0.013}
\newcommand{\NQwenBaseSZeroGenTokens}{4{,}329}

\newcommand{\NQwenBaseSZeroLAS}{0.093}
\newcommand{\NQwenBaseSZeroMOnePOne}{0.813}
\newcommand{\NQwenBaseSZeroMOnePTwo}{0.660}
\newcommand{\NQwenBaseSZeroMOnePTwoC}{0.759}

\newcommand{\NQwenBaseMicroF}{0.104}

\newcommand{\NQwenBaseValid}{0.987}

\newcommand{\NQwenVanillaSZeroMicroF}{0.193}
\newcommand{\NQwenVanillaSZeroMicroP}{0.303}
\newcommand{\NQwenVanillaSZeroMicroR}{0.141}

\newcommand{\NQwenVanillaSZeroValid}{0.847}
\newcommand{\NQwenVanillaSZeroHitCap}{0.153}
\newcommand{\NQwenVanillaSZeroGenTokens}{7{,}771}

\newcommand{\NQwenVanillaSZeroLAS}{0.271}
\newcommand{\NQwenVanillaSZeroMOnePOne}{0.888}
\newcommand{\NQwenVanillaSZeroMOnePTwo}{0.746}
\newcommand{\NQwenVanillaSZeroMOnePTwoC}{0.794}
\newcommand{\NQwenVanillaSOneMicroF}{0.189}
\newcommand{\NQwenVanillaSOneMicroP}{0.306}
\newcommand{\NQwenVanillaSOneMicroR}{0.137}

\newcommand{\NQwenVanillaSOneValid}{0.793}
\newcommand{\NQwenVanillaSOneHitCap}{0.207}
\newcommand{\NQwenVanillaSOneGenTokens}{8{,}149}

\newcommand{\NQwenVanillaSOneLAS}{0.264}
\newcommand{\NQwenVanillaSOneMOnePOne}{0.887}
\newcommand{\NQwenVanillaSOneMOnePTwo}{0.746}
\newcommand{\NQwenVanillaSOneMOnePTwoC}{0.791}
\newcommand{\NQwenVanillaSTwoMicroF}{0.180}
\newcommand{\NQwenVanillaSTwoMicroP}{0.280}
\newcommand{\NQwenVanillaSTwoMicroR}{0.132}

\newcommand{\NQwenVanillaSTwoValid}{0.833}
\newcommand{\NQwenVanillaSTwoHitCap}{0.167}
\newcommand{\NQwenVanillaSTwoGenTokens}{7{,}634}

\newcommand{\NQwenVanillaSTwoLAS}{0.266}
\newcommand{\NQwenVanillaSTwoMOnePOne}{0.888}
\newcommand{\NQwenVanillaSTwoMOnePTwo}{0.745}
\newcommand{\NQwenVanillaSTwoMOnePTwoC}{0.790}

\newcommand{\NQwenVanillaMicroF}{0.187}
\newcommand{\NQwenVanillaMicroFSd}{0.007}
\newcommand{\NQwenVanillaMicroP}{0.296}

\newcommand{\NQwenVanillaMicroR}{0.137}

\newcommand{\NQwenVanillaTailF}{0.059}

\newcommand{\NQwenVanillaValid}{0.824}

\newcommand{\NQwenVanillaPPL}{1.516}

\newcommand{\NQwenVanillaBLEU}{0.173}

\newcommand{\NQwenVanillaLAS}{0.267}
\newcommand{\NQwenVanillaLASSd}{0.004}
\newcommand{\NQwenVanillaMOnePOne}{0.888}

\newcommand{\NQwenVanillaMOnePTwoC}{0.791}

\newcommand{\NQwenVanillaMicroFVsVanillaBfDelta}{+0.041}

\newcommand{\NQwenVanillaMicroFVsMistillDelta}{+0.034}
\newcommand{\NQwenVanillaMicroFVsMistillLo}{+0.018}
\newcommand{\NQwenVanillaMicroFVsMistillHi}{+0.050}

\newcommand{\NQwenVanillaMicroFVsBaseDelta}{+0.081}
\newcommand{\NQwenVanillaMicroFVsBaseLo}{+0.063}
\newcommand{\NQwenVanillaMicroFVsBaseHi}{+0.101}

\newcommand{\NQwenVanillaBfSZeroMicroF}{0.136}
\newcommand{\NQwenVanillaBfSZeroMicroP}{0.227}
\newcommand{\NQwenVanillaBfSZeroMicroR}{0.097}

\newcommand{\NQwenVanillaBfSZeroValid}{0.900}
\newcommand{\NQwenVanillaBfSZeroHitCap}{0.100}
\newcommand{\NQwenVanillaBfSZeroGenTokens}{6{,}876}

\newcommand{\NQwenVanillaBfSZeroLAS}{0.179}
\newcommand{\NQwenVanillaBfSZeroMOnePOne}{0.864}
\newcommand{\NQwenVanillaBfSZeroMOnePTwo}{0.713}
\newcommand{\NQwenVanillaBfSZeroMOnePTwoC}{0.773}
\newcommand{\NQwenVanillaBfSOneMicroF}{0.148}
\newcommand{\NQwenVanillaBfSOneMicroP}{0.251}
\newcommand{\NQwenVanillaBfSOneMicroR}{0.105}

\newcommand{\NQwenVanillaBfSOneValid}{0.887}
\newcommand{\NQwenVanillaBfSOneHitCap}{0.113}
\newcommand{\NQwenVanillaBfSOneGenTokens}{7{,}383}

\newcommand{\NQwenVanillaBfSOneLAS}{0.177}
\newcommand{\NQwenVanillaBfSOneMOnePOne}{0.862}
\newcommand{\NQwenVanillaBfSOneMOnePTwo}{0.713}
\newcommand{\NQwenVanillaBfSOneMOnePTwoC}{0.774}
\newcommand{\NQwenVanillaBfSTwoMicroF}{0.147}
\newcommand{\NQwenVanillaBfSTwoMicroP}{0.239}
\newcommand{\NQwenVanillaBfSTwoMicroR}{0.106}

\newcommand{\NQwenVanillaBfSTwoValid}{0.907}
\newcommand{\NQwenVanillaBfSTwoHitCap}{0.093}
\newcommand{\NQwenVanillaBfSTwoGenTokens}{7{,}191}

\newcommand{\NQwenVanillaBfSTwoLAS}{0.192}
\newcommand{\NQwenVanillaBfSTwoMOnePOne}{0.862}
\newcommand{\NQwenVanillaBfSTwoMOnePTwo}{0.713}
\newcommand{\NQwenVanillaBfSTwoMOnePTwoC}{0.773}

\newcommand{\NQwenVanillaBfMicroF}{0.143}
\newcommand{\NQwenVanillaBfMicroFSd}{0.007}
\newcommand{\NQwenVanillaBfMicroP}{0.239}

\newcommand{\NQwenVanillaBfMicroR}{0.102}

\newcommand{\NQwenVanillaBfValid}{0.898}

\newcommand{\NQwenVanillaBfLAS}{0.183}
\newcommand{\NQwenVanillaBfLASSd}{0.008}
\newcommand{\NQwenVanillaBfMOnePOne}{0.863}

\newcommand{\NQwenVanillaBfMOnePTwoC}{0.773}

\newcommand{\NQwenVanillaBfMicroFVsVanillaDelta}{\ensuremath{-}0.041}
\newcommand{\NQwenVanillaBfMicroFVsVanillaLo}{\ensuremath{-}0.056}
\newcommand{\NQwenVanillaBfMicroFVsVanillaHi}{\ensuremath{-}0.026}

\newcommand{\NQwenVanillaBfMicroFVsMistillDelta}{\ensuremath{-}0.007}
\newcommand{\NQwenVanillaBfMicroFVsMistillLo}{\ensuremath{-}0.019}
\newcommand{\NQwenVanillaBfMicroFVsMistillHi}{+0.005}

\newcommand{\NQwenFeatKDSZeroMicroF}{0.142}
\newcommand{\NQwenFeatKDSZeroMicroP}{0.266}
\newcommand{\NQwenFeatKDSZeroMicroR}{0.097}

\newcommand{\NQwenFeatKDSZeroValid}{0.753}
\newcommand{\NQwenFeatKDSZeroHitCap}{0.247}
\newcommand{\NQwenFeatKDSZeroGenTokens}{9{,}017}

\newcommand{\NQwenFeatKDSZeroLAS}{0.194}
\newcommand{\NQwenFeatKDSZeroMOnePOne}{0.873}
\newcommand{\NQwenFeatKDSZeroMOnePTwo}{0.727}
\newcommand{\NQwenFeatKDSZeroMOnePTwoC}{0.772}
\newcommand{\NQwenFeatKDSOneMicroF}{0.163}
\newcommand{\NQwenFeatKDSOneMicroP}{0.281}
\newcommand{\NQwenFeatKDSOneMicroR}{0.115}

\newcommand{\NQwenFeatKDSOneValid}{0.833}
\newcommand{\NQwenFeatKDSOneHitCap}{0.167}
\newcommand{\NQwenFeatKDSOneGenTokens}{8{,}062}

\newcommand{\NQwenFeatKDSOneLAS}{0.217}
\newcommand{\NQwenFeatKDSOneMOnePOne}{0.871}
\newcommand{\NQwenFeatKDSOneMOnePTwo}{0.727}
\newcommand{\NQwenFeatKDSOneMOnePTwoC}{0.773}
\newcommand{\NQwenFeatKDSTwoMicroF}{0.152}
\newcommand{\NQwenFeatKDSTwoMicroP}{0.281}
\newcommand{\NQwenFeatKDSTwoMicroR}{0.104}

\newcommand{\NQwenFeatKDSTwoValid}{0.753}
\newcommand{\NQwenFeatKDSTwoHitCap}{0.247}
\newcommand{\NQwenFeatKDSTwoGenTokens}{8{,}424}

\newcommand{\NQwenFeatKDSTwoLAS}{0.227}
\newcommand{\NQwenFeatKDSTwoMOnePOne}{0.874}
\newcommand{\NQwenFeatKDSTwoMOnePTwo}{0.724}
\newcommand{\NQwenFeatKDSTwoMOnePTwoC}{0.768}

\newcommand{\NQwenFeatKDMicroF}{0.153}
\newcommand{\NQwenFeatKDMicroFSd}{0.010}

\newcommand{\NQwenFeatKDMicroFVsVanillaDelta}{\ensuremath{-}0.035}
\newcommand{\NQwenFeatKDMicroFVsVanillaLo}{\ensuremath{-}0.049}
\newcommand{\NQwenFeatKDMicroFVsVanillaHi}{\ensuremath{-}0.021}

\newcommand{\NQwenFeatKDLASVsVanillaDelta}{\ensuremath{-}0.054}

\newcommand{\NQwenDistillmSZeroMicroF}{0.187}
\newcommand{\NQwenDistillmSZeroMicroP}{0.287}
\newcommand{\NQwenDistillmSZeroMicroR}{0.139}

\newcommand{\NQwenDistillmSZeroValid}{0.880}
\newcommand{\NQwenDistillmSZeroHitCap}{0.120}
\newcommand{\NQwenDistillmSZeroGenTokens}{7{,}285}

\newcommand{\NQwenDistillmSZeroLAS}{0.271}
\newcommand{\NQwenDistillmSZeroMOnePOne}{0.891}
\newcommand{\NQwenDistillmSZeroMOnePTwo}{0.736}
\newcommand{\NQwenDistillmSZeroMOnePTwoC}{0.788}

\newcommand{\NQwenDistillmMicroF}{0.187}

\newcommand{\NQwenDistillmMicroFVsVanillaDelta}{+0.003}

\newcommand{\NQwenDistillmMicroFVsVanillaP}{0.57}

\newcommand{\NQwenOPDSZeroMicroF}{0.179}
\newcommand{\NQwenOPDSZeroMicroP}{0.275}
\newcommand{\NQwenOPDSZeroMicroR}{0.132}

\newcommand{\NQwenOPDSZeroValid}{0.973}
\newcommand{\NQwenOPDSZeroHitCap}{0.027}
\newcommand{\NQwenOPDSZeroGenTokens}{6{,}991}

\newcommand{\NQwenOPDSZeroLAS}{0.229}
\newcommand{\NQwenOPDSZeroMOnePOne}{0.865}
\newcommand{\NQwenOPDSZeroMOnePTwo}{0.718}
\newcommand{\NQwenOPDSZeroMOnePTwoC}{0.770}

\newcommand{\NQwenOPDMicroF}{0.179}

\newcommand{\NQwenOPDTailF}{0.063}

\newcommand{\NQwenOPDValid}{0.973}
\newcommand{\NQwenOPDHitCap}{0.027}

\newcommand{\NQwenOPDPPL}{1.638}
\newcommand{\NQwenOPDBLEU}{0.200}
\newcommand{\NQwenOPDLAS}{0.229}
\newcommand{\NQwenOPDMOnePOne}{0.865}

\newcommand{\NQwenOPDMOnePTwoC}{0.770}
\newcommand{\NQwenOPDMicroFVsVanillaDelta}{\ensuremath{-}0.005}
\newcommand{\NQwenOPDMicroFVsVanillaLo}{\ensuremath{-}0.025}
\newcommand{\NQwenOPDMicroFVsVanillaHi}{+0.015}
\newcommand{\NQwenOPDMicroFVsVanillaP}{0.67}

\newcommand{\NQwenOPDLASVsVanillaDelta}{\ensuremath{-}0.038}
\newcommand{\NQwenOPDLASVsVanillaLo}{\ensuremath{-}0.065}
\newcommand{\NQwenOPDLASVsVanillaHi}{\ensuremath{-}0.011}
\newcommand{\NQwenOPDLASVsVanillaP}{0.006}

\newcommand{\NQwenOPDBLEUVsVanillaBfDelta}{+0.031}
\newcommand{\NQwenOPDBLEUVsVanillaBfLo}{+0.022}
\newcommand{\NQwenOPDBLEUVsVanillaBfHi}{+0.040}

\newcommand{\NQwenOPDLASVsVanillaBfDelta}{+0.046}
\newcommand{\NQwenOPDLASVsVanillaBfLo}{+0.021}
\newcommand{\NQwenOPDLASVsVanillaBfHi}{+0.073}

\newcommand{\NQwenMistillSZeroMicroF}{0.151}
\newcommand{\NQwenMistillSZeroMicroP}{0.235}
\newcommand{\NQwenMistillSZeroMicroR}{0.111}

\newcommand{\NQwenMistillSZeroValid}{0.907}
\newcommand{\NQwenMistillSZeroHitCap}{0.093}
\newcommand{\NQwenMistillSZeroGenTokens}{6{,}886}

\newcommand{\NQwenMistillSZeroLAS}{0.201}
\newcommand{\NQwenMistillSZeroMOnePOne}{0.857}
\newcommand{\NQwenMistillSZeroMOnePTwo}{0.712}
\newcommand{\NQwenMistillSZeroMOnePTwoC}{0.802}
\newcommand{\NQwenMistillSOneMicroF}{0.147}
\newcommand{\NQwenMistillSOneMicroP}{0.247}
\newcommand{\NQwenMistillSOneMicroR}{0.105}

\newcommand{\NQwenMistillSOneValid}{0.887}
\newcommand{\NQwenMistillSOneHitCap}{0.113}
\newcommand{\NQwenMistillSOneGenTokens}{7{,}121}

\newcommand{\NQwenMistillSOneLAS}{0.191}
\newcommand{\NQwenMistillSOneMOnePOne}{0.857}
\newcommand{\NQwenMistillSOneMOnePTwo}{0.712}
\newcommand{\NQwenMistillSOneMOnePTwoC}{0.804}
\newcommand{\NQwenMistillSTwoMicroF}{0.150}
\newcommand{\NQwenMistillSTwoMicroP}{0.247}
\newcommand{\NQwenMistillSTwoMicroR}{0.108}

\newcommand{\NQwenMistillSTwoValid}{0.887}
\newcommand{\NQwenMistillSTwoHitCap}{0.113}
\newcommand{\NQwenMistillSTwoGenTokens}{7{,}128}

\newcommand{\NQwenMistillSTwoLAS}{0.185}
\newcommand{\NQwenMistillSTwoMOnePOne}{0.859}
\newcommand{\NQwenMistillSTwoMOnePTwo}{0.715}
\newcommand{\NQwenMistillSTwoMOnePTwoC}{0.804}

\newcommand{\NQwenMistillMicroF}{0.149}
\newcommand{\NQwenMistillMicroFSd}{0.002}
\newcommand{\NQwenMistillMicroP}{0.243}

\newcommand{\NQwenMistillMicroR}{0.108}

\newcommand{\NQwenMistillValid}{0.893}

\newcommand{\NQwenMistillLAS}{0.192}
\newcommand{\NQwenMistillLASSd}{0.008}

\newcommand{\NQwenMistillMOnePTwoC}{0.803}

\newcommand{\NQwenMistillMicroFVsVanillaDelta}{\ensuremath{-}0.034}
\newcommand{\NQwenMistillMicroFVsVanillaLo}{\ensuremath{-}0.050}
\newcommand{\NQwenMistillMicroFVsVanillaHi}{\ensuremath{-}0.018}

\newcommand{\NQwenMistillMicroFVsVanillaBfDelta}{+0.007}
\newcommand{\NQwenMistillMicroFVsVanillaBfLo}{\ensuremath{-}0.005}
\newcommand{\NQwenMistillMicroFVsVanillaBfHi}{+0.019}

\newcommand{\NQwenMistillFpSZeroMicroF}{0.161}
\newcommand{\NQwenMistillFpSZeroMicroP}{0.249}
\newcommand{\NQwenMistillFpSZeroMicroR}{0.119}

\newcommand{\NQwenMistillFpSZeroValid}{0.887}
\newcommand{\NQwenMistillFpSZeroHitCap}{0.113}
\newcommand{\NQwenMistillFpSZeroGenTokens}{7{,}715}

\newcommand{\NQwenMistillFpSZeroLAS}{0.212}
\newcommand{\NQwenMistillFpSZeroMOnePOne}{0.883}
\newcommand{\NQwenMistillFpSZeroMOnePTwo}{0.718}
\newcommand{\NQwenMistillFpSZeroMOnePTwoC}{0.778}
\newcommand{\NQwenMistillFpSOneMicroF}{0.165}
\newcommand{\NQwenMistillFpSOneMicroP}{0.282}
\newcommand{\NQwenMistillFpSOneMicroR}{0.117}

\newcommand{\NQwenMistillFpSOneValid}{0.787}
\newcommand{\NQwenMistillFpSOneHitCap}{0.213}
\newcommand{\NQwenMistillFpSOneGenTokens}{8{,}422}

\newcommand{\NQwenMistillFpSOneLAS}{0.239}
\newcommand{\NQwenMistillFpSOneMOnePOne}{0.880}
\newcommand{\NQwenMistillFpSOneMOnePTwo}{0.718}
\newcommand{\NQwenMistillFpSOneMOnePTwoC}{0.780}
\newcommand{\NQwenMistillFpSTwoMicroF}{0.169}
\newcommand{\NQwenMistillFpSTwoMicroP}{0.282}
\newcommand{\NQwenMistillFpSTwoMicroR}{0.121}

\newcommand{\NQwenMistillFpSTwoValid}{0.800}
\newcommand{\NQwenMistillFpSTwoHitCap}{0.200}
\newcommand{\NQwenMistillFpSTwoGenTokens}{8{,}478}

\newcommand{\NQwenMistillFpSTwoLAS}{0.232}
\newcommand{\NQwenMistillFpSTwoMOnePOne}{0.883}
\newcommand{\NQwenMistillFpSTwoMOnePTwo}{0.719}
\newcommand{\NQwenMistillFpSTwoMOnePTwoC}{0.779}
\newcommand{\NQwenMistillFpSeeds}{3}
\newcommand{\NQwenMistillFpMicroF}{0.165}
\newcommand{\NQwenMistillFpMicroFSd}{0.004}

\newcommand{\NQwenMistillFpValid}{0.824}

\newcommand{\NQwenMistillFpLAS}{0.228}
\newcommand{\NQwenMistillFpLASSd}{0.014}

\newcommand{\NQwenMistillFpMOnePTwoC}{0.779}

\newcommand{\NQwenMistillFpMicroFVsVanillaDelta}{\ensuremath{-}0.024}
\newcommand{\NQwenMistillFpMicroFVsVanillaLo}{\ensuremath{-}0.039}
\newcommand{\NQwenMistillFpMicroFVsVanillaHi}{\ensuremath{-}0.009}

\newcommand{\NQwenMistillFpLASVsVanillaDelta}{\ensuremath{-}0.039}

\newcommand{\NQwenMistillFpMicroFVsMistillDelta}{+0.010}
\newcommand{\NQwenMistillFpMicroFVsMistillLo}{\ensuremath{-}0.005}
\newcommand{\NQwenMistillFpMicroFVsMistillHi}{+0.025}

\newcommand{\NQwenMistillFpSelSZeroMicroF}{0.177}
\newcommand{\NQwenMistillFpSelSZeroMicroP}{0.288}
\newcommand{\NQwenMistillFpSelSZeroMicroR}{0.128}

\newcommand{\NQwenMistillFpSelSZeroValid}{0.820}
\newcommand{\NQwenMistillFpSelSZeroHitCap}{0.180}
\newcommand{\NQwenMistillFpSelSZeroGenTokens}{7{,}935}

\newcommand{\NQwenMistillFpSelSZeroLAS}{0.238}
\newcommand{\NQwenMistillFpSelSZeroMOnePOne}{0.888}
\newcommand{\NQwenMistillFpSelSZeroMOnePTwo}{0.747}
\newcommand{\NQwenMistillFpSelSZeroMOnePTwoC}{0.800}

\newcommand{\NQwenMistillFpSelMicroF}{0.176}

\newcommand{\NQwenMistillFpSelMicroFVsVanillaDelta}{\ensuremath{-}0.013}
\newcommand{\NQwenMistillFpSelMicroFVsVanillaLo}{\ensuremath{-}0.028}
\newcommand{\NQwenMistillFpSelMicroFVsVanillaHi}{+0.002}

\newcommand{\NSmolBaseSZeroMicroF}{0.003}
\newcommand{\NSmolBaseSZeroMicroP}{0.043}
\newcommand{\NSmolBaseSZeroMicroR}{0.002}

\newcommand{\NSmolBaseSZeroValid}{0.153}
\newcommand{\NSmolBaseSZeroHitCap}{0.847}
\newcommand{\NSmolBaseSZeroGenTokens}{14{,}147}

\newcommand{\NSmolBaseSZeroLAS}{\ensuremath{-}0.019}
\newcommand{\NSmolBaseSZeroMOnePOne}{0.731}
\newcommand{\NSmolBaseSZeroMOnePTwo}{0.379}
\newcommand{\NSmolBaseSZeroMOnePTwoC}{0.721}

\newcommand{\NSmolBaseMicroF}{0.003}

\newcommand{\NSmolBaseValid}{0.153}

\newcommand{\NSmolBaseMOnePOne}{0.731}
\newcommand{\NSmolBaseMOnePTwo}{0.379}
\newcommand{\NSmolBaseMOnePTwoC}{0.721}

\newcommand{\NSmolVanillaSZeroMicroF}{0.148}
\newcommand{\NSmolVanillaSZeroMicroP}{0.306}
\newcommand{\NSmolVanillaSZeroMicroR}{0.097}

\newcommand{\NSmolVanillaSZeroValid}{0.593}
\newcommand{\NSmolVanillaSZeroHitCap}{0.407}
\newcommand{\NSmolVanillaSZeroGenTokens}{10{,}232}

\newcommand{\NSmolVanillaSZeroLAS}{0.237}
\newcommand{\NSmolVanillaSZeroMOnePOne}{0.880}
\newcommand{\NSmolVanillaSZeroMOnePTwo}{0.592}
\newcommand{\NSmolVanillaSZeroMOnePTwoC}{0.789}
\newcommand{\NSmolVanillaSOneMicroF}{0.113}
\newcommand{\NSmolVanillaSOneMicroP}{0.285}
\newcommand{\NSmolVanillaSOneMicroR}{0.070}

\newcommand{\NSmolVanillaSOneValid}{0.500}
\newcommand{\NSmolVanillaSOneHitCap}{0.500}
\newcommand{\NSmolVanillaSOneGenTokens}{11{,}201}

\newcommand{\NSmolVanillaSOneLAS}{0.214}
\newcommand{\NSmolVanillaSOneMOnePOne}{0.879}
\newcommand{\NSmolVanillaSOneMOnePTwo}{0.591}
\newcommand{\NSmolVanillaSOneMOnePTwoC}{0.786}

\newcommand{\NSmolVanillaMicroF}{0.130}
\newcommand{\NSmolVanillaMicroFSd}{0.025}

\newcommand{\NSmolVanillaValid}{0.547}

\newcommand{\NSmolVanillaLAS}{0.226}
\newcommand{\NSmolVanillaLASSd}{0.016}

\newcommand{\NSmolVanillaMOnePTwoC}{0.787}

\newcommand{\NSmolVanillaMicroFVsMistillDelta}{+0.075}
\newcommand{\NSmolVanillaMicroFVsMistillLo}{+0.058}
\newcommand{\NSmolVanillaMicroFVsMistillHi}{+0.092}

\newcommand{\NSmolVanillaBfSZeroMicroF}{0.055}
\newcommand{\NSmolVanillaBfSZeroMicroP}{0.222}
\newcommand{\NSmolVanillaBfSZeroMicroR}{0.031}

\newcommand{\NSmolVanillaBfSZeroValid}{0.380}
\newcommand{\NSmolVanillaBfSZeroHitCap}{0.620}
\newcommand{\NSmolVanillaBfSZeroGenTokens}{12{,}312}

\newcommand{\NSmolVanillaBfSZeroLAS}{0.081}
\newcommand{\NSmolVanillaBfSZeroMOnePOne}{0.844}
\newcommand{\NSmolVanillaBfSZeroMOnePTwo}{0.535}
\newcommand{\NSmolVanillaBfSZeroMOnePTwoC}{0.756}

\newcommand{\NSmolVanillaBfMicroF}{0.055}

\newcommand{\NSmolVanillaBfValid}{0.380}

\newcommand{\NSmolVanillaBfLAS}{0.081}

\newcommand{\NSmolVanillaBfMOnePTwoC}{0.756}
\newcommand{\NSmolVanillaBfMicroFVsVanillaDelta}{\ensuremath{-}0.069}
\newcommand{\NSmolVanillaBfMicroFVsVanillaLo}{\ensuremath{-}0.092}
\newcommand{\NSmolVanillaBfMicroFVsVanillaHi}{\ensuremath{-}0.047}

\newcommand{\NSmolVanillaBfMicroFVsMistillDelta}{+0.006}
\newcommand{\NSmolVanillaBfMicroFVsMistillLo}{\ensuremath{-}0.011}
\newcommand{\NSmolVanillaBfMicroFVsMistillHi}{+0.021}

\newcommand{\NSmolFeatKDSZeroMicroF}{0.062}
\newcommand{\NSmolFeatKDSZeroMicroP}{0.245}
\newcommand{\NSmolFeatKDSZeroMicroR}{0.036}

\newcommand{\NSmolFeatKDSZeroValid}{0.360}
\newcommand{\NSmolFeatKDSZeroHitCap}{0.640}
\newcommand{\NSmolFeatKDSZeroGenTokens}{12{,}471}

\newcommand{\NSmolFeatKDSZeroLAS}{0.117}
\newcommand{\NSmolFeatKDSZeroMOnePOne}{0.861}
\newcommand{\NSmolFeatKDSZeroMOnePTwo}{0.560}
\newcommand{\NSmolFeatKDSZeroMOnePTwoC}{0.746}

\newcommand{\NSmolFeatKDMicroF}{0.062}

\newcommand{\NSmolFeatKDValid}{0.360}

\newcommand{\NSmolFeatKDMicroFVsVanillaDelta}{\ensuremath{-}0.060}
\newcommand{\NSmolFeatKDMicroFVsVanillaLo}{\ensuremath{-}0.080}
\newcommand{\NSmolFeatKDMicroFVsVanillaHi}{\ensuremath{-}0.038}

\newcommand{\NSmolGOLDSZeroMicroF}{0.000}
\newcommand{\NSmolGOLDSZeroMicroP}{0.000}
\newcommand{\NSmolGOLDSZeroMicroR}{0.000}

\newcommand{\NSmolGOLDSZeroValid}{0.613}
\newcommand{\NSmolGOLDSZeroHitCap}{0.387}
\newcommand{\NSmolGOLDSZeroGenTokens}{10{,}794}

\newcommand{\NSmolGOLDSZeroLAS}{\ensuremath{-}0.051}
\newcommand{\NSmolGOLDSZeroMOnePOne}{0.832}
\newcommand{\NSmolGOLDSZeroMOnePTwo}{0.355}
\newcommand{\NSmolGOLDSZeroMOnePTwoC}{0.732}

\newcommand{\NSmolGOLDMicroF}{0.000}

\newcommand{\NSmolGOLDValid}{0.613}

\newcommand{\NSmolGOLDPPL}{2.132}
\newcommand{\NSmolGOLDBLEU}{0.116}

\newcommand{\NSmolGOLDMOnePOne}{0.832}
\newcommand{\NSmolGOLDMOnePTwo}{0.355}
\newcommand{\NSmolGOLDMOnePTwoC}{0.732}

\newcommand{\NSmolMistillSZeroMicroF}{0.060}
\newcommand{\NSmolMistillSZeroMicroP}{0.246}
\newcommand{\NSmolMistillSZeroMicroR}{0.034}

\newcommand{\NSmolMistillSZeroValid}{0.320}
\newcommand{\NSmolMistillSZeroHitCap}{0.680}
\newcommand{\NSmolMistillSZeroGenTokens}{13{,}044}

\newcommand{\NSmolMistillSZeroLAS}{0.081}
\newcommand{\NSmolMistillSZeroMOnePOne}{0.845}
\newcommand{\NSmolMistillSZeroMOnePTwo}{0.538}
\newcommand{\NSmolMistillSZeroMOnePTwoC}{0.741}
\newcommand{\NSmolMistillSOneMicroF}{0.042}
\newcommand{\NSmolMistillSOneMicroP}{0.213}
\newcommand{\NSmolMistillSOneMicroR}{0.023}

\newcommand{\NSmolMistillSOneValid}{0.273}
\newcommand{\NSmolMistillSOneHitCap}{0.727}
\newcommand{\NSmolMistillSOneGenTokens}{13{,}434}

\newcommand{\NSmolMistillSOneLAS}{0.080}
\newcommand{\NSmolMistillSOneMOnePOne}{0.845}
\newcommand{\NSmolMistillSOneMOnePTwo}{0.537}
\newcommand{\NSmolMistillSOneMOnePTwoC}{0.741}
\newcommand{\NSmolMistillSTwoMicroF}{0.036}
\newcommand{\NSmolMistillSTwoMicroP}{0.180}
\newcommand{\NSmolMistillSTwoMicroR}{0.020}

\newcommand{\NSmolMistillSTwoValid}{0.273}
\newcommand{\NSmolMistillSTwoHitCap}{0.727}
\newcommand{\NSmolMistillSTwoGenTokens}{13{,}464}

\newcommand{\NSmolMistillSTwoLAS}{0.063}
\newcommand{\NSmolMistillSTwoMOnePOne}{0.845}
\newcommand{\NSmolMistillSTwoMOnePTwo}{0.537}
\newcommand{\NSmolMistillSTwoMOnePTwoC}{0.740}

\newcommand{\NSmolMistillMicroF}{0.046}
\newcommand{\NSmolMistillMicroFSd}{0.013}

\newcommand{\NSmolMistillValid}{0.289}

\newcommand{\NSmolMistillLAS}{0.075}
\newcommand{\NSmolMistillLASSd}{0.010}

\newcommand{\NSmolMistillMOnePTwoC}{0.740}

\newcommand{\NSmolMistillMicroFVsVanillaDelta}{\ensuremath{-}0.075}
\newcommand{\NSmolMistillMicroFVsVanillaLo}{\ensuremath{-}0.092}
\newcommand{\NSmolMistillMicroFVsVanillaHi}{\ensuremath{-}0.058}

\newcommand{\NQwenAttnOnlyMicroF}{0.130}

\newcommand{\NQwenAttnOnlyMicroFVsMistillDelta}{\ensuremath{-}0.019}
\newcommand{\NQwenAttnOnlyMicroFVsMistillLo}{\ensuremath{-}0.038}
\newcommand{\NQwenAttnOnlyMicroFVsMistillHi}{\ensuremath{-}0.002}

\newcommand{\NQwenBottomMicroF}{0.114}

\newcommand{\NQwenInterOnlyMicroF}{0.157}

\newcommand{\NQwenInterOnlyMicroFVsMistillDelta}{+0.010}
\newcommand{\NQwenInterOnlyMicroFVsMistillLo}{\ensuremath{-}0.008}
\newcommand{\NQwenInterOnlyMicroFVsMistillHi}{+0.027}

\newcommand{\NQwenLowDoseMicroF}{0.150}

\newcommand{\NQwenLowDoseMicroFVsMistillDelta}{\ensuremath{-}0.002}
\newcommand{\NQwenLowDoseMicroFVsMistillLo}{\ensuremath{-}0.021}
\newcommand{\NQwenLowDoseMicroFVsMistillHi}{+0.017}

\newcommand{\NQwenLowPOneDoseMicroF}{0.152}

\newcommand{\NQwenLowPOneDoseMOnePOne}{0.860}
\newcommand{\NQwenLowPOneDoseMOnePTwo}{0.714}
\newcommand{\NQwenLowPOneDoseMOnePTwoC}{0.804}

\newcommand{\NQwenLowPOneDoseMicroFVsMistillDelta}{+0.001}
\newcommand{\NQwenLowPOneDoseMicroFVsMistillLo}{\ensuremath{-}0.015}
\newcommand{\NQwenLowPOneDoseMicroFVsMistillHi}{+0.017}

\newcommand{\NQwenNoWindowMicroF}{0.147}

\newcommand{\NQwenPOneOnlyMicroF}{0.131}

\newcommand{\NQwenPOneOnlyMicroFVsMistillDelta}{\ensuremath{-}0.022}
\newcommand{\NQwenPOneOnlyMicroFVsMistillLo}{\ensuremath{-}0.040}
\newcommand{\NQwenPOneOnlyMicroFVsMistillHi}{\ensuremath{-}0.001}

\newcommand{\NQwenPTwoOnlyMicroF}{0.154}

\newcommand{\NQwenPTwoOnlyMicroFVsMistillDelta}{+0.004}
\newcommand{\NQwenPTwoOnlyMicroFVsMistillLo}{\ensuremath{-}0.011}
\newcommand{\NQwenPTwoOnlyMicroFVsMistillHi}{+0.020}

\newcommand{\NQwenRandHeadsSZeroMicroF}{0.133}

\newcommand{\NQwenRandHeadsSOneMicroF}{0.117}

\newcommand{\NQwenUniformMicroF}{0.121}

\newcommand{\NQwenUniformHitCap}{0.187}

\newcommand{\NQwenUniformMOnePTwoC}{0.766}

\newcommand{\NQwenEventWeightedMicroF}{0.141}

\newcommand{\NQwenEventWeightedHitCap}{0.200}

\newcommand{\NQwenEventWeightedMOnePTwoC}{0.772}

\newcommand{\NSmolAttnTermMicroF}{0.036}

\newcommand{\NSmolAttnTermValid}{0.300}
\newcommand{\NSmolAttnTermHitCap}{0.700}

\newcommand{\NSmolAttnTermLAS}{0.082}
\newcommand{\NSmolAttnTermMOnePOne}{0.844}
\newcommand{\NSmolAttnTermMOnePTwo}{0.535}
\newcommand{\NSmolAttnTermMOnePTwoC}{0.736}

\newcommand{\NSmolBottomMicroF}{0.048}

\newcommand{\NSmolBottomMicroFVsMistillDelta}{\ensuremath{-}0.001}
\newcommand{\NSmolBottomMicroFVsMistillLo}{\ensuremath{-}0.014}
\newcommand{\NSmolBottomMicroFVsMistillHi}{+0.012}

\newcommand{\NSmolKEightMicroF}{0.047}

\newcommand{\NSmolKEightMicroFVsMistillDelta}{\ensuremath{-}0.001}
\newcommand{\NSmolKEightMicroFVsMistillLo}{\ensuremath{-}0.016}
\newcommand{\NSmolKEightMicroFVsMistillHi}{+0.015}

\newcommand{\NSmolRandPosMicroF}{0.063}

\newcommand{\NOwnTeacherPTwoCKiForty}{0.847}

\newcommand{\NOwnTeacherPOneKiForty}{0.915}

\newcommand{\NOwnTeacherMTwoPTwoCLift}{3.38}
\newcommand{\NOwnTeacherMTwoPTwoCLiftRand}{1.95}
\newcommand{\NOwnTeacherPrefix}{0.260}

\newcommand{\NOwnQwenBasePOneKiForty}{0.946}
\newcommand{\NOwnQwenBasePOneKiFortyRand}{0.043}

\newcommand{\NOwnQwenBasePTwoKiForty}{0.815}

\newcommand{\NOwnQwenBasePTwoCKiForty}{0.788}
\newcommand{\NOwnQwenBasePTwoCKiFortyRand}{0.030}

\newcommand{\NOwnQwenBaseMTwoPTwoCLift}{2.56}

\newcommand{\NOwnQwenBaseMFiveLocus}{8}
\newcommand{\NOwnQwenBasePrefix}{0.185}

\newcommand{\NOwnQwenVanillaSZeroPTwoCKiForty}{0.840}

\newcommand{\NOwnQwenVanillaSZeroMFiveLocus}{7}

\newcommand{\NOwnQwenVanillaSOnePTwoCKiForty}{0.848}

\newcommand{\NOwnQwenVanillaSTwoPTwoCKiForty}{0.891}

\newcommand{\NOwnQwenVanillaBfSZeroPTwoCKiEight}{0.467}

\newcommand{\NOwnQwenVanillaBfSZeroPTwoCKiForty}{0.834}

\newcommand{\NOwnQwenVanillaBfSOnePTwoCKiForty}{0.835}

\newcommand{\NOwnQwenVanillaBfSTwoPTwoCKiForty}{0.847}

\newcommand{\NOwnQwenFeatKDSZeroPTwoCKiForty}{0.844}

\newcommand{\NOwnQwenMistillSZeroPOneKiForty}{0.956}

\newcommand{\NOwnQwenMistillSZeroPTwoKiForty}{0.736}

\newcommand{\NOwnQwenMistillSZeroPTwoCKiForty}{0.873}

\newcommand{\NOwnQwenMistillSZeroMTwoPOneLift}{2.26}

\newcommand{\NOwnQwenMistillSZeroMTwoPTwoCLift}{3.65}

\newcommand{\NOwnQwenMistillSZeroMFiveLocus}{7}

\newcommand{\NOwnQwenMistillSOnePTwoCKiForty}{0.870}

\newcommand{\NOwnQwenMistillSTwoPTwoCKiForty}{0.874}

\newcommand{\NOwnQwenMistillFpSZeroPTwoCKiEight}{0.186}

\newcommand{\NOwnQwenMistillFpSZeroPTwoCKiForty}{0.870}
\newcommand{\NOwnQwenMistillFpSZeroPTwoCKiFortyRand}{0.022}

\newcommand{\NOwnQwenMistillFpSOnePTwoCKiForty}{0.836}

\newcommand{\NOwnQwenMistillFpSTwoPTwoCKiForty}{0.850}

\newcommand{\NOwnQwenMistillFpSelSZeroPTwoCKiEight}{0.133}

\newcommand{\NOwnQwenMistillFpSelSZeroPTwoCKiForty}{0.892}

\newcommand{\NOwnQwenMistillFpSelSOnePTwoCKiForty}{0.897}

\newcommand{\NOwnQwenMistillFpSelSTwoPTwoCKiForty}{0.892}

\newcommand{\NOwnQwenRandHeadsSZeroPTwoKiForty}{0.880}

\newcommand{\NOwnQwenRandHeadsSZeroPTwoCKiForty}{0.869}

\newcommand{\NOwnQwenRandPosSZeroPTwoCKiForty}{0.631}

\newcommand{\NOwnQwenVanillaSeedSpreadPTwoC}{0.052}
\newcommand{\NOwnQwenVanillaBfSeedSpreadPTwoC}{0.013}
\newcommand{\NOwnQwenMistillSeedSpreadPTwoC}{0.004}

\newcommand{\NPoolQwenMistillVsVanillaPTwoCClusterDelta}{+0.011}

\newcommand{\NPoolQwenMistillVsVanillaPTwoCClusterHigher}{84}
\newcommand{\NPoolQwenMistillVsVanillaPTwoCClusterLower}{66}

\newcommand{\NPoolQwenMistillVsVanillaPTwoCClusterP}{0.16}

\newcommand{\NPoolQwenMistillVsVanillaPTwoClusterDelta}{+0.019}

\newcommand{\NPoolQwenMistillVsVanillaMTwoPTwoCDelta}{+0.113}

\newcommand{\NPoolQwenMistillVsVanillaBfPTwoCClusterDelta}{+0.027}

\newcommand{\NPoolQwenMistillVsVanillaBfPTwoCClusterHigher}{107}
\newcommand{\NPoolQwenMistillVsVanillaBfPTwoCClusterLower}{43}

\newcommand{\NPoolQwenMistillVsVanillaBfPTwoCClusterP}{\ensuremath{2{\times}10^{-7}}}
\newcommand{\NPoolQwenMistillVsVanillaBfPTwoCClusterWilcoxonP}{\ensuremath{10^{-6}}}

\newcommand{\NPoolQwenMistillVsVanillaBfPTwoClusterDelta}{\ensuremath{-}0.020}

\newcommand{\NPoolQwenMistillVsVanillaBfPTwoClusterLower}{108}
\newcommand{\NPoolQwenMistillVsVanillaBfPTwoClusterN}{150}

\newcommand{\NPoolQwenMistillFpVsVanillaPTwoCClusterDelta}{\ensuremath{-}0.005}

\newcommand{\NPoolQwenMistillFpVsVanillaPTwoCClusterP}{0.57}

\newcommand{\NPoolQwenMistillFpVsVanillaBfPTwoCClusterDelta}{+0.004}

\newcommand{\NPoolQwenMistillFpVsVanillaBfPTwoCClusterP}{0.46}

\newcommand{\NPoolQwenMistillFpVsVanillaBfMTwoPTwoCClusterDelta}{\ensuremath{-}0.308}

\newcommand{\NPoolQwenMistillFpVsVanillaBfMTwoPOneClusterDelta}{+0.744}

\newcommand{\NPoolQwenMistillFpVsMistillPTwoCClusterDelta}{\ensuremath{-}0.014}

\newcommand{\NPoolQwenMistillFpVsMistillPTwoCClusterP}{0.007}

\newcommand{\NOwnSmolBasePOneKiForty}{0.985}
\newcommand{\NOwnSmolBasePOneKiFortyRand}{0.134}

\newcommand{\NOwnSmolBasePTwoKiForty}{0.809}

\newcommand{\NOwnSmolBasePTwoCKiEight}{0.174}

\newcommand{\NOwnSmolBasePTwoCKiForty}{0.416}
\newcommand{\NOwnSmolBasePTwoCKiFortyRand}{0.107}

\newcommand{\NOwnSmolBaseMThreeNearMiss}{0.081}

\newcommand{\NOwnSmolVanillaSZeroPTwoCKiForty}{0.541}

\newcommand{\NOwnSmolVanillaSZeroMThreeNearMiss}{0.058}

\newcommand{\NOwnSmolVanillaSOnePTwoCKiForty}{0.523}

\newcommand{\NOwnSmolVanillaSOneMThreeNearMiss}{0.061}

\newcommand{\NOwnSmolVanillaBfSZeroPTwoKiForty}{0.867}

\newcommand{\NOwnSmolVanillaBfSZeroPTwoCKiEight}{0.217}

\newcommand{\NOwnSmolVanillaBfSZeroPTwoCKiForty}{0.476}

\newcommand{\NOwnSmolFeatKDSZeroPTwoCKiForty}{0.628}

\newcommand{\NOwnSmolMistillSZeroPTwoKiForty}{0.866}

\newcommand{\NOwnSmolMistillSZeroPTwoCKiEight}{0.213}

\newcommand{\NOwnSmolMistillSZeroPTwoCKiForty}{0.789}

\newcommand{\NOwnSmolMistillSZeroMTwoPTwoCLift}{2.65}

\newcommand{\NOwnSmolMistillSZeroMThreeNearMiss}{0.025}

\newcommand{\NOwnSmolMistillSZeroPrefix}{0.439}

\newcommand{\NOwnSmolMistillSOnePTwoKiForty}{0.868}

\newcommand{\NOwnSmolMistillSOnePTwoCKiEight}{0.199}

\newcommand{\NOwnSmolMistillSOnePTwoCKiForty}{0.774}

\newcommand{\NOwnSmolMistillSTwoPTwoKiForty}{0.857}

\newcommand{\NOwnSmolMistillSTwoPTwoCKiEight}{0.209}

\newcommand{\NOwnSmolMistillSTwoPTwoCKiForty}{0.750}

\newcommand{\NOwnSmolRandHeadsSZeroPTwoKiForty}{0.725}

\newcommand{\NOwnSmolRandHeadsSZeroPTwoCKiEight}{0.208}

\newcommand{\NOwnSmolRandHeadsSZeroPTwoCKiForty}{0.585}

\newcommand{\NOwnSmolRandHeadsSZeroPrefix}{0.284}

\newcommand{\NOwnSmolRandPosSZeroPTwoCKiForty}{0.304}

\newcommand{\NOwnSmolRandPosSZeroMTwoPTwoCLift}{2.52}

\newcommand{\NPoolSmolMistillVsVanillaPTwoCClusterDelta}{+0.233}

\newcommand{\NPoolSmolMistillVsVanillaPTwoCClusterHigher}{132}

\newcommand{\NPoolSmolMistillVsVanillaBfPTwoCClusterDelta}{+0.310}

\newcommand{\NPoolSmolMistillVsVanillaBfPTwoCClusterHigher}{130}

\newcommand{\NPoolSmolMistillVsVanillaBfPTwoCClusterP}{\ensuremath{6{\times}10^{-21}}}

\newcommand{\NOwnQwenVanillaSeedMeanPTwoCKiForty}{0.859}

\newcommand{\NOwnQwenMistillFpSelSeedMeanPTwoCKiForty}{0.894}

\newcommand{\NOwnSmolVanillaSeedMeanPTwoCKiForty}{0.532}

\newcommand{\NOwnSmolMistillSeedMeanPTwoCKiForty}{0.771}

\newcommand{\NOwnQwenRandBothSZeroPTwoCKiForty}{0.863}

\newcommand{\NOwnQwenRandBothSZeroMTwoPTwoCLift}{2.69}

\newcommand{\NOwnQwenRandBothSZeroMThreeNearMiss}{0.040}

\newcommand{\NOwnQwenRandBothSZeroPrefix}{0.139}

\newcommand{\NOwnQwenBottomSZeroPTwoKiForty}{0.851}

\newcommand{\NOwnQwenBottomSZeroPTwoCKiForty}{0.717}

\newcommand{\NOwnQwenBottomSZeroMTwoPTwoCLift}{3.68}

\newcommand{\NOwnQwenBottomSZeroMThreeNearMiss}{0.082}

\newcommand{\NPairQwenBottomSZeroVsBasePTwoCLower}{113}

\newcommand{\NPairQwenBottomSZeroVsMistillSZeroPTwoCKiEightLower}{150}

\newcommand{\NPairQwenBottomSZeroVsMistillSZeroPTwoCKiEightMeanA}{0.165}
\newcommand{\NPairQwenBottomSZeroVsMistillSZeroPTwoCKiEightMeanB}{0.605}

\newcommand{\NOwnQwenPOneOnlySZeroPTwoCKiForty}{0.802}

\newcommand{\NOwnQwenPTwoOnlySZeroPTwoCKiForty}{0.864}

\newcommand{\NPairQwenPTwoOnlySZeroVsMistillSZeroPTwoCDelta}{\ensuremath{-}0.006}

\newcommand{\NPairQwenPTwoOnlySZeroVsBasePTwoCDelta}{+0.073}

\newcommand{\NOwnQwenLowDoseSZeroPTwoCKiForty}{0.880}

\newcommand{\NOwnQwenAttnOnlySZeroPTwoCKiEight}{0.500}

\newcommand{\NOwnQwenAttnOnlySZeroPTwoCKiForty}{0.861}

\newcommand{\NPairQwenAttnOnlySZeroVsBasePTwoCDelta}{+0.078}

\newcommand{\NOwnQwenInterOnlySZeroPTwoCKiForty}{0.866}

\newcommand{\NPairQwenInterOnlySZeroVsBasePTwoCDelta}{+0.064}

\newcommand{\NOwnQwenRandHeadsPoolSOnePTwoCKiForty}{0.821}

\newcommand{\NOwnQwenKEightSZeroPTwoCKiForty}{0.881}

\newcommand{\NOwnQwenKHundredSZeroPTwoKiForty}{0.861}

\newcommand{\NOwnQwenKHundredSZeroPTwoCKiForty}{0.875}

\newcommand{\NPairQwenKHundredSZeroVsMistillSZeroPTwoHigher}{150}

\newcommand{\NOwnQwenEventWeightedSZeroPTwoCKiEight}{0.467}

\newcommand{\NOwnQwenEventWeightedSZeroPTwoCKiForty}{0.828}

\newcommand{\NPairQwenEventWeightedSZeroVsMistillSZeroPTwoCDelta}{\ensuremath{-}0.042}

\newcommand{\NPairQwenEventWeightedSZeroVsMistillSZeroPTwoCLower}{102}

\newcommand{\NPairQwenEventWeightedSZeroVsMistillSOnePTwoCLower}{100}

\newcommand{\NPairQwenEventWeightedSZeroVsMistillSZeroPTwoCKiEightLower}{140}

\newcommand{\NOwnQwenUniformSZeroPTwoCKiEight}{0.631}

\newcommand{\NOwnQwenUniformSZeroPTwoCKiForty}{0.915}

\newcommand{\NOwnQwenUniformSZeroMTwoPTwoCLift}{3.68}

\newcommand{\NPairQwenUniformSZeroVsMistillSZeroPTwoCDelta}{+0.047}
\newcommand{\NPairQwenUniformSZeroVsMistillSZeroPTwoCHigher}{123}

\newcommand{\NOwnQwenNoWindowSZeroPOneKiForty}{0.960}

\newcommand{\NOwnQwenNoWindowSZeroPTwoCKiForty}{0.870}

\newcommand{\NOwnQwenNoWindowSZeroMTwoPOneLift}{2.49}

\newcommand{\NPairQwenNoWindowSZeroVsMistillSZeroPTwoCDelta}{\ensuremath{-}0.008}

\newcommand{\NPairQwenNoWindowSZeroVsMistillSZeroPTwoCP}{\ensuremath{10^{-9}}}

\newcommand{\NPairQwenRandHeadsSOneVsRandHeadsSZeroPTwoCDelta}{\ensuremath{-}0.033}

\newcommand{\NPairQwenRandHeadsSOneVsRandHeadsSZeroPTwoCLower}{111}
\newcommand{\NPairQwenRandHeadsSOneVsRandHeadsSZeroPTwoCP}{\ensuremath{3{\times}10^{-9}}}

\newcommand{\NPairQwenRandHeadsSOneVsRandHeadsSZeroPTwoDelta}{\ensuremath{-}0.089}

\newcommand{\NPairQwenRandHeadsSOneVsRandHeadsSZeroPTwoLower}{147}

\newcommand{\NPairQwenMistillSOneVsMistillSZeroPTwoCDelta}{+0.000}
\newcommand{\NPairQwenMistillSOneVsMistillSZeroPTwoCHigher}{75}
\newcommand{\NPairQwenMistillSOneVsMistillSZeroPTwoCLower}{75}

\newcommand{\NPairQwenAttnOnlySZeroVsInterOnlySZeroPTwoCP}{0.37}

\newcommand{\NPairQwenPTwoOnlySZeroVsPOneOnlySZeroPTwoCDelta}{+0.046}
\newcommand{\NPairQwenPTwoOnlySZeroVsPOneOnlySZeroPTwoCHigher}{109}

\newcommand{\NPairQwenRandBothSZeroVsRandPosSZeroPTwoCDelta}{+0.221}
\newcommand{\NPairQwenRandBothSZeroVsRandPosSZeroPTwoCHigher}{140}

\newcommand{\NPairQwenRandBothSZeroVsRandHeadsSZeroPTwoCDelta}{\ensuremath{-}0.006}

\newcommand{\NPairQwenRandBothSZeroVsRandHeadsSZeroPTwoCP}{0.46}

\newcommand{\NPairQwenVanillaSTwoVsVanillaSOnePTwoCDelta}{+0.040}
\newcommand{\NPairQwenVanillaSTwoVsVanillaSOnePTwoCHigher}{112}

\newcommand{\NPairQwenVanillaSTwoVsVanillaSOnePTwoCP}{\ensuremath{10^{-9}}}

\newcommand{\NOwnSmolBottomSZeroPTwoKiForty}{0.857}

\newcommand{\NOwnSmolBottomSZeroPTwoCKiForty}{0.841}

\newcommand{\NOwnSmolBottomSZeroMTwoPTwoCLift}{2.45}

\newcommand{\NOwnSmolBottomSZeroMThreeNearMiss}{0.042}

\newcommand{\NOwnSmolBottomSZeroPrefix}{0.279}

\newcommand{\NPairSmolBottomSZeroVsMistillSZeroPTwoCKiEightMeanA}{0.342}
\newcommand{\NPairSmolBottomSZeroVsMistillSZeroPTwoCKiEightMeanB}{0.232}

\newcommand{\NPairSmolBottomSZeroVsMistillSOnePTwoCKiEightMeanB}{0.228}
\newcommand{\NPairSmolBottomSZeroVsRandHeadsSZeroPTwoCDelta}{+0.278}

\newcommand{\NOwnSmolBottomSOnePTwoKiForty}{0.862}

\newcommand{\NOwnSmolBottomSOnePTwoCKiForty}{0.835}

\newcommand{\NPairSmolBottomSOneVsMistillSZeroPTwoCKiEightMeanA}{0.334}

\newcommand{\NPairSmolBottomSOneVsRandHeadsSZeroPTwoCDelta}{+0.262}

\newcommand{\NOwnSmolKEightSZeroPTwoKiForty}{0.860}

\newcommand{\NOwnSmolKEightSZeroPTwoCKiForty}{0.671}

\newcommand{\NOwnSmolKEightSZeroMTwoPTwoCLift}{3.11}

\newcommand{\NOwnSmolKEightSZeroMThreeNearMiss}{0.037}

\newcommand{\NOwnSmolKEightSZeroPrefix}{0.388}

\newcommand{\NPairSmolKEightSZeroVsMistillSZeroPTwoCKiEightMeanA}{0.170}

\newcommand{\NPairSmolKEightSZeroVsRandHeadsSZeroPTwoCDelta}{+0.114}
\newcommand{\NPairSmolKEightSZeroVsRandHeadsSZeroPTwoCHigher}{104}

\newcommand{\NPairSmolMistillSOneVsMistillSZeroPTwoCP}{0.93}

\newcommand{\NPairSmolMistillSOneVsMistillSZeroPTwoCMeanA}{0.788}
\newcommand{\NPairSmolMistillSOneVsMistillSZeroPTwoCMeanB}{0.786}

\newcommand{\NPairSmolVanillaSOneVsVanillaSZeroPTwoCMeanA}{0.528}
\newcommand{\NPairSmolVanillaSOneVsVanillaSZeroPTwoCMeanB}{0.550}

\newcommand{\NPairSmolKEightSZeroVsBottomSZeroPTwoCDelta}{\ensuremath{-}0.127}

\newcommand{\NPairSmolKEightSZeroVsBottomSZeroPTwoDelta}{+0.008}

\newcommand{\NPairSmolKEightSZeroVsBottomSZeroPTwoP}{0.57}

\newcommand{\NPairSmolBottomSOneVsBottomSZeroPTwoCDelta}{\ensuremath{-}0.023}

\newcommand{\NPairSmolBottomSOneVsBottomSZeroPTwoCLower}{93}
\newcommand{\NPairSmolBottomSOneVsBottomSZeroPTwoCP}{0.004}

\newcommand{\NPairSmolBottomSOneVsBottomSZeroPTwoDelta}{+0.002}

\newcommand{\NPairSmolBottomSOneVsBottomSZeroPTwoP}{0.57}

\newcommand{\NPairSmolKEightSZeroVsBottomSOnePTwoCDelta}{\ensuremath{-}0.111}

\newcommand{\NCurveVanillaFullSZeroMicroF}{0.193}

\newcommand{\NCurveMistillFullSZeroMicroF}{0.151}

\newcommand{\NCurvePairOneHundredMistillVsVanillaTaskHigher}{58}
\newcommand{\NCurvePairOneHundredMistillVsVanillaTaskLower}{59}
\newcommand{\NCurvePairOneHundredMistillVsVanillaTaskP}{1.00}

\newcommand{\NCurvePairOneHundredMistillVsVanillaMOnePTwoCMean}{+0.024}
\newcommand{\NCurvePairTwoFiftyMistillVsVanillaTaskHigher}{67}
\newcommand{\NCurvePairTwoFiftyMistillVsVanillaTaskLower}{54}
\newcommand{\NCurvePairTwoFiftyMistillVsVanillaTaskP}{0.28}

\newcommand{\NCurvePairTwoFiftyMistillVsVanillaMOnePTwoCMean}{+0.032}

\newcommand{\NCurvePairFullMistillVsVanillaMOnePTwoCMean}{+0.012}
\newcommand{\NCurvePairFullMistillVsVanillaBfTaskHigher}{205}
\newcommand{\NCurvePairFullMistillVsVanillaBfTaskLower}{184}
\newcommand{\NCurvePairFullMistillVsVanillaBfTaskP}{0.31}

\newcommand{\NCurvePairFullMistillVsVanillaBfMOnePTwoCHigher}{348}
\newcommand{\NCurvePairFullMistillVsVanillaBfMOnePTwoCLower}{29}
\newcommand{\NCurvePairFullMistillVsVanillaBfMOnePTwoCP}{\ensuremath{<10^{-4}}}
\newcommand{\NCurvePairFullMistillVsVanillaBfMOnePTwoCMean}{+0.035}

\newcommand{\NPoolQwenMistillFpVsVanillaBfPTwoCKiEightClusterDelta}{\ensuremath{-}0.281}

\newcommand{\NCorpusQwenMistillVsVanillaBfDelta}{+0.006}
\newcommand{\NCorpusQwenMistillVsVanillaBfLo}{\ensuremath{-}0.007}
\newcommand{\NCorpusQwenMistillVsVanillaBfHi}{+0.019}

\newcommand{\NCorpusQwenMistillFpSelVsVanillaDelta}{\ensuremath{-}0.012}
\newcommand{\NCorpusQwenMistillFpSelVsVanillaLo}{\ensuremath{-}0.026}
\newcommand{\NCorpusQwenMistillFpSelVsVanillaHi}{+0.003}

\newcommand{\NCorpusMaxGapToPerNote}{0.012}

\newcommand{\NSelPairClusterTaskMean}{+0.012}
\newcommand{\NSelPairClusterTaskLo}{\ensuremath{-}0.001}
\newcommand{\NSelPairClusterTaskHi}{+0.025}
\newcommand{\NSelPairClusterTaskHigher}{54}
\newcommand{\NSelPairClusterTaskLower}{37}
\newcommand{\NSelPairClusterTaskTies}{9}
\newcommand{\NSelPairClusterTaskN}{100}

\newcommand{\NSelPairClusterTaskP}{0.09}

\newcommand{\NSelPairClusterMOnePTwoCMean}{+0.024}
\newcommand{\NSelPairClusterMOnePTwoCLo}{+0.017}
\newcommand{\NSelPairClusterMOnePTwoCHi}{+0.031}
\newcommand{\NSelPairClusterMOnePTwoCHigher}{79}
\newcommand{\NSelPairClusterMOnePTwoCLower}{12}
\newcommand{\NSelPairClusterMOnePTwoCTies}{8}
\newcommand{\NSelPairClusterMOnePTwoCN}{99}

\newcommand{\NSelPairClusterMOnePTwoCP}{\ensuremath{3{\times}10^{-13}}}
\newcommand{\NSelPairClusterMOnePOneMean}{\ensuremath{-}0.007}
\newcommand{\NSelPairClusterMOnePOneLo}{\ensuremath{-}0.010}
\newcommand{\NSelPairClusterMOnePOneHi}{\ensuremath{-}0.004}
\newcommand{\NSelPairClusterMOnePOneHigher}{21}
\newcommand{\NSelPairClusterMOnePOneLower}{55}
\newcommand{\NSelPairClusterMOnePOneTies}{23}
\newcommand{\NSelPairClusterMOnePOneN}{99}

\newcommand{\NSelPairClusterMOnePOneP}{\ensuremath{10^{-4}}}
\newcommand{\NSelPairClusterMOnePTwoMean}{\ensuremath{-}0.003}
\newcommand{\NSelPairClusterMOnePTwoLo}{\ensuremath{-}0.007}
\newcommand{\NSelPairClusterMOnePTwoHi}{+0.001}
\newcommand{\NSelPairClusterMOnePTwoHigher}{39}
\newcommand{\NSelPairClusterMOnePTwoLower}{58}
\newcommand{\NSelPairClusterMOnePTwoTies}{2}

\newcommand{\NSelPairClusterMOnePTwoP}{0.07}

\newcommand{\NFaithPatients}{94}
\newcommand{\NFaithPairedPatients}{6}

\newcommand{\NFaithPTwoNecKEightGlobalPatients}{0.82}

\newcommand{\NFaithPatientsMaxKiMove}{0.008}
\newcommand{\NFaithPatientsMaxNecMoveNotPTwo}{0.007}
\newcommand{\NFaithPatientsPTwoNecMoveMin}{0.02}
\newcommand{\NFaithPatientsPTwoNecMoveMax}{0.06}
\newcommand{\NFaithPatientsMaxMTwoMove}{0.04}

\newcommand{\NCovSmolPOneKeptPct}{100\%}

\newcommand{\NCovSmolPTwoKeptPct}{53\%}

\newcommand{\NCovQwenPTwoCTeacher}{57{,}584}

\newcommand{\NCovSmolPTwoCKept}{41{,}824}
\newcommand{\NCovSmolPTwoCKeptPct}{73\%}
\newcommand{\NCovSmolPTwoCPerNoteMedian}{0.74}

\newcommand{\NGateTeacherFullPTwoC}{0.71}

\newcommand{\NGateQwenMistillPTwoC}{0.68}

\newcommand{\NGateSmolBasePTwoC}{0.56}

\newcommand{\NAlignedSmolMistillVsVanillaPTwoCClusterDelta}{+0.224}

\newcommand{\NAlignedSmolMistillVsVanillaBfPTwoCClusterDelta}{+0.334}

\newcommand{\NAlignedQwenMistillVsVanillaBfPTwoCClusterDelta}{+0.065}

\newcommand{\NProxyEvents}{1{,}959}

\newcommand{\NProxyHeads}{395}
\newcommand{\NProxyCSpearmanEvent}{+0.40}

\newcommand{\NProxyCSpearmanPooled}{+0.34}
\newcommand{\NProxyCPearsonEvent}{+0.52}

\newcommand{\NProxyMCSpearmanEvent}{+0.35}

\newcommand{\NProxyMCSpearmanPooled}{+0.26}

\newcommand{\NProxyASpearmanEvent}{+0.69}

\newcommand{\NProxyASpearmanPooled}{+0.77}

\newcommand{\NProxyAbsRatio}{0.26}
\newcommand{\NProxyHeadLThirtyNineHTwelveI}{0.63}
\newcommand{\NProxyHeadLThirtyNineHTwelveC}{0.33}

\newcommand{\NProxyHeadLThirtyNineHTenI}{0.31}
\newcommand{\NProxyHeadLThirtyNineHTenC}{0.20}

\newcommand{\NProxyHeadLThirtyHElevenI}{0.27}
\newcommand{\NProxyHeadLThirtyHElevenC}{0.05}

\newcommand{\NRetrieverTopTenMicroF}{0.179}

\newcommand{\NRetrieverTopTwentyMicroF}{0.217}
\newcommand{\NRetrieverTopTwentyMicroP}{0.191}
\newcommand{\NRetrieverTopTwentyMicroR}{0.249}

\newcommand{\NRetrieverShortlistReproduced}{53}
\newcommand{\NRetrieverShortlistJaccardMedian}{0.9993}

\newcommand{\NSplitHalfHDTwoPTwoCSpearman}{0.63}
\newcommand{\NSplitHalfHDTwoPTwoCTopEight}{8}

\newcommand{\NSplitHalfHDTwoPTwoCTopForty}{36}

\newcommand{\NSplitHalfHDOneSplits}{200}

\newcommand{\NSplitHalfHDOnePTwoSpearman}{0.94}

\newcommand{\NSplitHalfHDOnePTwoTopForty}{37}

\newcommand{\NSplitHalfHDOnePTwoCSpearman}{0.82}

\newcommand{\NSplitHalfExactCrossfitContrastEight}{0.83}
\newcommand{\NSplitHalfExactCrossfitExactEight}{1.00}

\newcommand{\NSweepLPointZeroOneValMicroF}{0.183}

\newcommand{\NSweepLPointZeroOneTestMicroF}{0.177}

\newcommand{\NSweepLPointZeroThreeValMicroF}{0.181}

\newcommand{\NSweepLPointZeroThreeTestMicroF}{0.203}

\newcommand{\NSweepLPointOneTestMicroF}{0.161}

\newcommand{\NSweepSelected}{0.01}
\newcommand{\NSweepLPointZeroOnePTwoCVsVanillaClusterDelta}{+0.032}

\newcommand{\NSweepLPointZeroOnePTwoCVsVanillaClusterHigher}{113}

\newcommand{\NSweepLPointZeroOnePTwoCVsVanillaClusterP}{\ensuremath{4{\times}10^{-10}}}

\newcommand{\NSweepLPointZeroOnePTwoCVsMistillSZeroDelta}{+0.025}

\newcommand{\NOwnQwenMistillFpLPointThreeSZeroPTwoCKiForty}{0.643}

\newcommand{\NSweepLPointZeroOneProbeKiEight}{0.572}
\newcommand{\NSweepLPointZeroOneProbeKiEightVsVanillaClusterDelta}{+0.048}

\newcommand{\NSweepLPointZeroOneProbeKiEightVsVanillaClusterHigher}{115}

\newcommand{\NSweepLPointOneProbeKiEight}{0.425}

\newcommand{\NSweepLPointZeroOneRankLThirtyFiveHTwentySix}{12}

\newcommand{\NSweepLPointOneRankLThirtyFiveHTwentySix}{11}

\newcommand{\NSweepMistillSZeroRankLThirtyFiveHTwentySix}{3}

\newcommand{\NSweepLPointZeroOneTaskVsVanillaSZeroNoteDelta}{\ensuremath{-}0.022}

\newcommand{\NSweepLPointZeroOneTaskVsVanillaSOneNoteDelta}{\ensuremath{-}0.015}

\newcommand{\NSweepLPointZeroOneTaskVsVanillaSTwoNoteDelta}{\ensuremath{-}0.007}

\newcommand{\NPairQwenEventWeightedSZeroVsVanillaBfSZeroPTwoCDelta}{\ensuremath{-}0.005}

\newcommand{\NPairQwenEventWeightedSZeroVsVanillaBfSZeroPTwoCP}{0.12}

\newcommand{\NPairQwenEventWeightedSZeroVsRandHeadsSZeroPTwoCDelta}{\ensuremath{-}0.042}

\newcommand{\NPairQwenNoWindowSZeroVsVanillaBfSZeroPTwoCKiEightDelta}{+0.113}
\newcommand{\NPairQwenNoWindowSZeroVsVanillaBfSZeroPTwoCKiEightHigher}{140}

\newcommand{\NPairQwenRandHeadsPoolSTwoVsRandHeadsPoolSOnePTwoCDelta}{+0.021}
\newcommand{\NPairQwenRandHeadsPoolSTwoVsRandHeadsPoolSOnePTwoCHigher}{108}

\newcommand{\NPairQwenRandHeadsPoolSOneVsRandHeadsSZeroPTwoCDelta}{\ensuremath{-}0.058}

\newcommand{\NPoolQwenMistillVsRandHeadsAllPTwoCClusterDelta}{+0.026}

\newcommand{\NPoolQwenMistillVsRandHeadsAllPTwoCClusterHigher}{107}

\newcommand{\NPoolQwenMistillVsRandHeadsAllPTwoCClusterP}{\ensuremath{2{\times}10^{-7}}}

\newcommand{\NPairQwenRandHeadsPoolSTwoVsRandHeadsPoolSOnePTwoCKiEightDelta}{+0.013}

\newcommand{\NPairQwenRandHeadsPoolSOneVsRandHeadsSZeroPTwoCKiEightDelta}{\ensuremath{-}0.093}

\newcommand{\NPoolQwenMistillVsRandHeadsAllPTwoCKiEightClusterDelta}{+0.095}

\newcommand{\NPoolQwenMistillVsRandHeadsAllPTwoCKiEightClusterHigher}{139}

\newcommand{\NOwnSmolMistillThreeSeedPTwoCKiForty}{0.782}
\newcommand{\NOwnSmolMistillThreeSeedPTwoCKiFortyLo}{0.747}
\newcommand{\NOwnSmolMistillThreeSeedPTwoCKiFortyHi}{0.819}

\newcommand{\NOwnSmolBasePTwoCKiFortyMean}{0.404}
\newcommand{\NOwnSmolMistillThreeSeedRatioBase}{1.94}
\newcommand{\NOwnSmolMistillSeedMeanRatioBase}{1.85}

\newcommand{\NAuditN}{181}

\newcommand{\NAuditRejectedPct}{35\%}

\newcommand{\NAuditRejectedWeighted}{30\%}

\newcommand{\NAuditEnumerationPct}{55\%}

\newcommand{\NAuditEnumerationWeighted}{62\%}

\newcommand{\NAuditBothPct}{3\%}

\newcommand{\NAuditOtherPct}{7\%}

\newcommand{\NAuditMeaningfulYesPct}{87\%}

\newcommand{\NAuditRejectedInAnswerPct}{60\%}

\newcommand{\NAuditRejectedNotInAnswerPct}{25\%}
\newcommand{\NAuditQwenMistillSZeroMOneRejected}{0.794}
\newcommand{\NAuditQwenMistillSZeroMOneOther}{0.914}
\newcommand{\NAuditMOneRejectedN}{63}
\newcommand{\NAuditMOneOtherN}{105}

\newcommand{\NAuditMFourRejectedN}{21}
\newcommand{\NAuditMFourOtherN}{17}

\newcommand{\NNearMissTeacherNeutralN}{1{,}365}
\newcommand{\NNearMissTeacherNeutralMean}{+1.02}
\newcommand{\NNearMissTeacherNeutralMedian}{+0.31}
\newcommand{\NNearMissTeacherNeutralHigher}{796}

\newcommand{\NNearMissTeacherNeutralP}{\ensuremath{3{\times}10^{-16}}}
\newcommand{\NNearMissTeacherNeutralMediated}{0.09}

\newcommand{\NNearMissTeacherNeutralMediatedMeanFrac}{0.36}

\newcommand{\NNearMissTeacherHeadsShareClean}{13\%}
\newcommand{\NNearMissTeacherNeutralMinusUnrelatedMedian}{+0.000}

\newcommand{\NNearMissTeacherNeutralMinusUnrelatedP}{0.07}
\newcommand{\NNearMissTeacherNeutralMinusOthercatMedian}{+0.000}

\newcommand{\NNearMissTeacherNeutralMinusOthercatP}{0.23}

\newcommand{\NNearMissQwenMistillSZeroMediated}{0.00}

\newcommand{\NNearMissQwenMistillVsVanillaBfPooledDelta}{+0.29}
\newcommand{\NNearMissQwenMistillVsVanillaBfPooledLo}{+0.26}
\newcommand{\NNearMissQwenMistillVsVanillaBfPooledHi}{+0.31}
\newcommand{\NNearMissQwenMistillVsVanillaBfPooledHigher}{147}

\newcommand{\NNearMissQwenMistillFpVsVanillaPooledDelta}{\ensuremath{-}0.22}

\newcommand{\NNearMissQwenMistillFpSelVsVanillaPooledDelta}{+0.06}

\newcommand{\NNearMissQwenMistillFpSelVsVanillaPooledHigher}{115}

\newcommand{\NNearMissSmolMistillVsVanillaBfPooledDelta}{\ensuremath{-}0.16}

\newcommand{\NNearMissSmolMistillVsVanillaPooledDelta}{\ensuremath{-}0.51}

\newcommand{\NEvidenceOk}{428}

\newcommand{\NEvidenceOkPct}{31\%}

\newcommand{\NEvidenceEvidenceMinusControlNoteMean}{+0.055}
\newcommand{\NEvidenceEvidenceMinusControlNoteLo}{\ensuremath{-}0.003}
\newcommand{\NEvidenceEvidenceMinusControlNoteHi}{+0.123}
\newcommand{\NEvidenceEvidenceMinusControlNoteHigher}{42}

\newcommand{\NEvidenceEvidenceMinusControlNoteN}{75}
\newcommand{\NEvidenceEvidenceMinusControlNoteP}{0.07}

\newcommand{\NEvidenceEvidenceMinusControlNdNoteMean}{+0.053}
\newcommand{\NEvidenceEvidenceMinusControlNdNoteLo}{\ensuremath{-}0.007}
\newcommand{\NEvidenceEvidenceMinusControlNdNoteHi}{+0.125}

\newcommand{\NFollowMentions}{750}

\newcommand{\NFollowDriftMedian}{0.11}

\newcommand{\NFollowEmissionAtCodeDrop}{4.73}

\newcommand{\NFollowProseSixteenDF}{+4.46}
\newcommand{\NFollowProseSixteenDNWin}{+0.84}
\newcommand{\NFollowProseSixteenDNWinMean}{+1.37}
\newcommand{\NFollowProseSixteenDNAll}{+4.50}
\newcommand{\NFollowProseSixteenRatio}{0.23}
\newcommand{\NFollowProseSixteenNBig}{686}
\newcommand{\NFollowProseSixteenFlipped}{4.0\%}

\newcommand{\NFollowProseVsRandomNoteMean}{+1.54}
\newcommand{\NFollowProseVsRandomNoteLo}{+1.33}
\newcommand{\NFollowProseVsRandomNoteHi}{+1.79}
\newcommand{\NFollowProseVsRandomNoteHigher}{99}

\newcommand{\NFollowProseVsRandomNoteP}{\ensuremath{2{\times}10^{-28}}}

\newcommand{\NPTwoIntactNotes}{50}
\newcommand{\NPTwoIntactPTwoSpearman}{0.68}
\newcommand{\NPTwoIntactPTwoTopEight}{5}

\newcommand{\NPTwoIntactPTwoTopForty}{30}

\newcommand{\NPTwoIntactPOneTopForty}{33}

\newcommand{\NPTwoIntactPTwoCTopForty}{31}
\newcommand{\NPTwoIntactFirstOrderVsExactSpearman}{0.98}
\newcommand{\NPTwoIntactFirstOrderVsExactTopForty}{38}

\newcommand{\NPTwoIntactHeadLThirtyNineHTenIntactRank}{2}
\newcommand{\NPTwoIntactHeadLThirtyNineHTenCutRank}{2}
\newcommand{\NPTwoIntactHeadLThirtyNineHTwentyNineIntactRank}{4}
\newcommand{\NPTwoIntactHeadLThirtyNineHTwentyNineCutRank}{3}
\newcommand{\NPTwoIntactHeadLThirtyNineHTwelveIntactRank}{3}
\newcommand{\NPTwoIntactHeadLThirtyNineHTwelveCutRank}{9}
\newcommand{\NPTwoIntactHeadLThirtySevenHSixIntactRank}{9}
\newcommand{\NPTwoIntactHeadLThirtySevenHSixCutRank}{8}

\newcommand{\NOwnTraceQwenMistillSZeroKiForty}{0.863}

\newcommand{\NOwnTraceQwenMistillSZeroKiFortyTF}{0.863}

\newcommand{\NOwnTraceQwenMistillSOneKiEightDelta}{+0.068}

\newcommand{\NOwnTraceQwenVanillaSOneNotes}{119}

\newcommand{\NOwnTraceQwenVanillaSOneKiFortyDelta}{+0.031}

\newcommand{\NOwnTraceQwenMistillFpSelSZeroNotes}{123}

\newcommand{\NOwnTraceQwenMistillFpSelSZeroKiForty}{0.914}

\newcommand{\NOwnTraceQwenMistillFpSelSZeroKiFortyTF}{0.885}

\newcommand{\NOwnTraceQwenMistillFpSelSOneNotes}{125}

\newcommand{\NOwnTraceQwenMistillFpSelSOneKiForty}{0.902}

\newcommand{\NOwnTraceQwenMistillFpSelSOneKiFortyTF}{0.888}

\newcommand{\NOwnTraceQwenMistillFpSelSTwoNotes}{124}

\newcommand{\NOwnTraceQwenMistillFpSelSTwoKiForty}{0.885}

\newcommand{\NOwnTraceQwenMistillFpSelSTwoKiFortyTF}{0.876}

\newcommand{\NOwnTraceQwenBaseNotes}{138}

\newcommand{\NOwnTraceQwenBaseKiFortyDelta}{+0.080}

\newcommand{\NOwnTraceSmolMistillSZeroNotes}{48}

\newcommand{\NOwnTraceSmolMistillSZeroKiForty}{0.880}

\newcommand{\NOwnTraceSmolMistillSZeroKiFortyDelta}{+0.105}

\newcommand{\NOwnTraceSmolMistillSOneNotes}{41}

\newcommand{\NOwnTraceSmolMistillSOneKiForty}{0.897}

\newcommand{\NOwnTraceSmolMistillSOneKiFortyDelta}{+0.126}

\newcommand{\NOwnTraceSmolMistillSTwoNotes}{40}
\newcommand{\NOwnTraceSmolMistillSTwoComplete}{41}
\newcommand{\NOwnTraceSmolMistillSTwoKiForty}{0.904}

\newcommand{\NOwnTraceSmolMistillSTwoKiFortyTF}{0.724}
\newcommand{\NOwnTraceSmolMistillSTwoKiFortyDelta}{+0.181}
\newcommand{\NOwnTraceSmolMistillSTwoKiFortyHigher}{25}
\newcommand{\NOwnTraceSmolMistillSTwoKiFortyN}{40}

\newcommand{\NOwnTraceSmolMistillSTwoKiEightDelta}{+0.032}

\newcommand{\NOwnTraceSmolVanillaBfSZeroKiForty}{0.499}

\newcommand{\NOwnTraceSmolVanillaBfSZeroKiFortyDelta}{\ensuremath{-}0.026}

\newcommand{\NOwnTraceSmolVanillaSOneNotes}{75}

\newcommand{\NOwnTraceSmolVanillaSOneKiForty}{0.579}

\newcommand{\NOwnTraceSmolVanillaSOneKiFortyDelta}{+0.053}

\newcommand{\NOwnTraceQwenMistillVsVanillaBfPooledDelta}{+0.041}
\newcommand{\NOwnTraceQwenMistillVsVanillaBfPooledLo}{+0.020}
\newcommand{\NOwnTraceQwenMistillVsVanillaBfPooledHi}{+0.062}
\newcommand{\NOwnTraceQwenMistillVsVanillaBfPooledHigher}{102}

\newcommand{\NOwnTraceQwenMistillVsVanillaBfPooledNotes}{148}

\newcommand{\NOwnTraceQwenMistillVsVanillaBfKiEightPooledDelta}{+0.166}

\newcommand{\NOwnTraceQwenMistillVsVanillaPooledDelta}{\ensuremath{-}0.001}
\newcommand{\NOwnTraceQwenMistillVsVanillaPooledLo}{\ensuremath{-}0.026}
\newcommand{\NOwnTraceQwenMistillVsVanillaPooledHi}{+0.023}

\newcommand{\NOwnTraceQwenMistillVsVanillaKiEightPooledDelta}{+0.094}

\newcommand{\NOwnTraceQwenMistillFpSelVsVanillaSZeroN}{105}

\newcommand{\NOwnTraceQwenMistillFpSelVsVanillaSOneN}{103}

\newcommand{\NOwnTraceQwenMistillFpSelVsVanillaSTwoN}{104}

\newcommand{\NOwnTraceQwenMistillFpSelVsVanillaPooledDelta}{+0.029}
\newcommand{\NOwnTraceQwenMistillFpSelVsVanillaPooledLo}{+0.006}
\newcommand{\NOwnTraceQwenMistillFpSelVsVanillaPooledHi}{+0.051}

\newcommand{\NOwnTraceQwenMistillFpSelVsVanillaPooledNotes}{137}

\newcommand{\NOwnTraceQwenMistillFpSelVsVanillaKiEightPooledDelta}{\ensuremath{-}0.258}

\newcommand{\NNeutralNotes}{30}
\newcommand{\NNeutralPatients}{29}
\newcommand{\NNeutralNeutralThinkClosed}{29}

\newcommand{\NNeutralNeutralGenTokens}{3{,}880}

\newcommand{\NNeutralOrigGenTokens}{3{,}744}

\newcommand{\NNeutralOrigPOneKiKEightWins}{30/30}

\newcommand{\NNeutralOrigPTwoCKiKEightWins}{28/28}

\newcommand{\NNeutralOverlapPOneForty}{33}

\newcommand{\NNeutralOverlapPTwoCForty}{31}
\newcommand{\NNeutralOverlapPTwoCEight}{8}

\newcommand{\NNeutralNeutralLocusWindowPct}{47\%}

\newcommand{\NNeutralViewLocusWindowPct}{38\%}

\newcommand{\NNeutralProseEmissionShared}{0}
\newcommand{\NNeutralProseTopEightSharedHDOne}{6}

\newcommand{\NKeyAblMinEffect}{0.1}

\newcommand{\NKeyAblLTwentySixHThirtySixShare}{0.33}

\newcommand{\NKeyAblLTwentyEightHTwentyOneShare}{0.70}

\newcommand{\NKeyAblLTwentyEightHTwentyOneNoteHigher}{100}

\newcommand{\NKeyAblLThirtyHElevenEvents}{1{,}365}

\newcommand{\NKeyAblLThirtyHElevenDNamedMedian}{+0.25}

\newcommand{\NKeyAblLThirtyHElevenDCtrlMedian}{+0.00}

\newcommand{\NKeyAblLThirtyHElevenShare}{1.00}

\newcommand{\NKeyAblLThirtyHElevenNoteHigher}{92}

\newcommand{\NKeyAblLThirtyHElevenNoteN}{92}

\newcommand{\NKeyAblSetLineShare}{0.00}

\newcommand{\NResidMentions}{746}
\newcommand{\NResidNotes}{96}
\newcommand{\NResidLStar}{4}
\newcommand{\NResidFlipLayer}{24}
\newcommand{\NResidRatio}{0.48}
\newcommand{\NResidNBig}{720}
\newcommand{\NResidForcedDropLStar}{11.1}
\newcommand{\NResidNaturalDropLStar}{3.8}
\newcommand{\NResidDonorRiseLStar}{4.2}
\newcommand{\NResidRatioMediansLTwelve}{0.26}

\newcommand{\NResidRatioMediansLTwenty}{0.22}

\newcommand{\NResidCodeLTwentyFourDrop}{9.1}

\newcommand{\NResidCodeLTwentyEightDrop}{16.7}

\newcommand{\NResidNaturalDropNoteMean}{+6.52}
\newcommand{\NResidNaturalDropNoteLo}{+5.94}
\newcommand{\NResidNaturalDropNoteHi}{+7.12}
\newcommand{\NResidNaturalDropNoteHigher}{96}

\newcommand{\NSecondLayers}{64}
\newcommand{\NSecondHeads}{40}

\newcommand{\NSecondPOneKiKForty}{0.90}

\newcommand{\NSecondPTwoKiKEight}{0.17}

\newcommand{\NSecondPTwoKiKForty}{0.70}

\newcommand{\NSecondPTwoCKiKForty}{0.72}
\newcommand{\NSecondPTwoCKiKFortyRand}{0.00}
\newcommand{\NSecondPTwoCKiKFortyWins}{100/100}

\newcommand{\NSecondExactTopHead}{\head{63}{32}}
\newcommand{\NSecondExactTopI}{+0.66}
\newcommand{\NSecondExactTopLo}{+0.58}
\newcommand{\NSecondExactTopHi}{+0.76}
\newcommand{\NSecondExactSecondHead}{\head{52}{33}}
\newcommand{\NSecondExactSecondI}{+0.15}
\newcommand{\NSecondExactPearsonMedian}{0.87}

\newcommand{\NSecondOverlapPOnePTwoCEight}{3}
\newcommand{\NTeacherOverlapPOnePTwoCEight}{2}
\newcommand{\NSecondOverlapPOnePTwoCForty}{21}
\newcommand{\NTeacherOverlapPOnePTwoCForty}{24}

\newcommand{\NSecondPTwoCTopFortyLayerMin}{43}
\newcommand{\NSecondPTwoCTopFortyLayerMax}{63}
\newcommand{\NSecondPTwoCTopFortyLastTwoLayers}{15}

\newcommand{\NSecondLocusWindowPct}{34.8\%}

\newcommand{\NSecondLocusDMedian}{10}

\newcommand{\NSecondLocusOverlapEmissionKEight}{1}

\newcommand{\NSecondLocusOverlapEmissionKForty}{3}
\newcommand{\NSecondLocusLayerMedian}{41}

\newcommand{\NLocusTopFortyLayerMedian}{21}

\newcommand{\NSecondLocusAEmisTopSixteenWin}{+0.08}
\newcommand{\NSecondLocusAEmisTopSixteenSuf}{+0.90}

\newcommand{\NSecondLocusBKSixteenWin}{+0.55}
\newcommand{\NSecondLocusBKSixteenSuf}{+0.67}

\newcommand{\NSecondLocusBKSixteenRandWin}{+0.01}
\newcommand{\NLocusProseSufKSixteen}{+0.48}

\newcommand{\NSecondReadProseNote}{0.20}

\newcommand{\NSecondReadRandomNote}{0.03}

\newcommand{\NSecondTraceGenTokensMedian}{8{,}033}
\newcommand{\NSecondTraceHitCap}{1}
\newcommand{\NSecondTraceN}{100}
\newcommand{\NSecondTraceTeacherGenTokensMedian}{3{,}280}

\newcommand{\NSecondAllEventsPTwoC}{7{,}501}
\newcommand{\NSecondAllEventsPTwoCTeacher}{3{,}855}

\newcommand{\NLocusSuffixNotes}{90}
\newcommand{\NLocusSuffixBMentions}{1{,}861}
\newcommand{\NLocusSuffixBWindowPct}{36.1\%}

\newcommand{\NLocusSuffixBDMedian}{7}

\newcommand{\NLocusSuffixDWindowPct}{43.2\%}

\newcommand{\NLocusSuffixDDMedian}{8}

\newcommand{\NLocusSuffixDAgree}{80.1\%}

\newcommand{\NLocusSuffixDDDiffMedian}{1}
\newcommand{\NLocusSuffixDDWithinEight}{85\%}

\newcommand{\NLocusSuffixAWindowPct}{43.6\%}

\newcommand{\NLocusSuffixADMedian}{7}

\newcommand{\NLocusSuffixAAgree}{79.4\%}

\newcommand{\NRepLPointZeroOnePTwoCSZeroMean}{+0.041}
\newcommand{\NRepLPointZeroOnePTwoCSZeroLo}{+0.032}
\newcommand{\NRepLPointZeroOnePTwoCSZeroHi}{+0.051}

\newcommand{\NRepLPointZeroOnePTwoCSOneMean}{+0.047}
\newcommand{\NRepLPointZeroOnePTwoCSOneLo}{+0.037}
\newcommand{\NRepLPointZeroOnePTwoCSOneHi}{+0.057}

\newcommand{\NRepLPointZeroOnePTwoCSTwoMean}{+0.002}
\newcommand{\NRepLPointZeroOnePTwoCSTwoLo}{\ensuremath{-}0.007}
\newcommand{\NRepLPointZeroOnePTwoCSTwoHi}{+0.011}

\newcommand{\NRepLPointZeroOnePTwoCClusterMean}{+0.030}
\newcommand{\NRepLPointZeroOnePTwoCClusterLo}{+0.023}
\newcommand{\NRepLPointZeroOnePTwoCClusterHi}{+0.037}
\newcommand{\NRepLPointZeroOnePTwoCClusterHigher}{118}

\newcommand{\NRepLPointZeroOnePTwoCKiEightSZeroMean}{\ensuremath{-}0.384}

\newcommand{\NRepLPointZeroOnePTwoCKiEightSOneMean}{+0.079}

\newcommand{\NRepLPointZeroOnePTwoCKiEightSTwoMean}{\ensuremath{-}0.469}

\newcommand{\NRepLPointZeroOnePTwoCKiEightClusterMean}{\ensuremath{-}0.258}
\newcommand{\NRepLPointZeroOnePTwoCKiEightClusterLo}{\ensuremath{-}0.270}
\newcommand{\NRepLPointZeroOnePTwoCKiEightClusterHi}{\ensuremath{-}0.247}

\newcommand{\NRepLPointZeroOneMTwoPTwoCClusterMean}{\ensuremath{-}0.262}
\newcommand{\NRepLPointZeroOneMTwoPTwoCClusterLo}{\ensuremath{-}0.366}
\newcommand{\NRepLPointZeroOneMTwoPTwoCClusterHi}{\ensuremath{-}0.170}

\newcommand{\NRepLPointZeroOneEditPooledMean}{+0.053}
\newcommand{\NRepLPointZeroOneEditPooledLo}{+0.042}
\newcommand{\NRepLPointZeroOneEditPooledHi}{+0.064}
\newcommand{\NRepLPointZeroOneEditPooledNotes}{150}

\newcommand{\NMatchedSelectMistillPick}{0.03}

\newcommand{\NMatchedSelectMistillGapFull}{0.0040204002}

\newcommand{\NMatchedSelectMistillPickFive}{0.1}

\newcommand{\NMatchedSelectRandHeadsPick}{0.3}

\newcommand{\NMatchedSelectRandHeadsGapFull}{0.0041750956}

\newcommand{\NMatchedSelectRandHeadsPickFive}{0.03}

\newcommand{\NMatchedMistillSeedMeanMicroF}{0.155}

\newcommand{\NMatchedMistillSeedMeanPTwoCKiForty}{0.880}

\newcommand{\NMatchedRandHeadsSeedMeanMicroF}{0.124}

\newcommand{\NMatchedRandHeadsSeedMeanPTwoCKiForty}{0.821}

\newcommand{\NMatchedPTwoCSZeroMean}{+0.062}

\newcommand{\NMatchedPTwoCSOneMean}{+0.070}

\newcommand{\NMatchedPTwoCSTwoMean}{+0.063}

\newcommand{\NMatchedPTwoCClusterMean}{+0.065}
\newcommand{\NMatchedPTwoCClusterLo}{+0.053}
\newcommand{\NMatchedPTwoCClusterHi}{+0.077}
\newcommand{\NMatchedPTwoCClusterHigher}{125}

\newcommand{\NMatchedPTwoCKiEightClusterMean}{+0.114}
\newcommand{\NMatchedPTwoCKiEightClusterLo}{+0.104}
\newcommand{\NMatchedPTwoCKiEightClusterHi}{+0.124}

\newcommand{\NMatchedMTwoPTwoCClusterMean}{\ensuremath{-}0.426}
\newcommand{\NMatchedMTwoPTwoCClusterLo}{\ensuremath{-}0.650}
\newcommand{\NMatchedMTwoPTwoCClusterHi}{\ensuremath{-}0.261}

\newcommand{\NMatchedPOneClusterMean}{\ensuremath{-}0.011}

\newcommand{\NMatchedTaskDelta}{+0.032}
\newcommand{\NMatchedTaskLo}{+0.019}
\newcommand{\NMatchedTaskHi}{+0.044}

\newcommand{\NMatchedTaskSZeroDelta}{+0.043}

\newcommand{\NMatchedTaskSOneDelta}{+0.029}

\newcommand{\NMatchedTaskSTwoDelta}{+0.023}

\newcommand{\NEFDelta}{0.10}
\newcommand{\NEFCases}{33}
\newcommand{\NEFPatients}{33}
\newcommand{\NEFTeacherD}{+0.511}
\newcommand{\NEFTeacherDLo}{+0.398}
\newcommand{\NEFTeacherDHi}{+0.621}

\newcommand{\NEFTeacherFollowFlip}{0.447}
\newcommand{\NEFTeacherPaired}{0.333}

\newcommand{\NEFMistillSZeroD}{+0.242}

\newcommand{\NEFMistillSZeroFollowFlip}{0.083}
\newcommand{\NEFMistillSZeroPaired}{0.053}

\newcommand{\NEFMistillSOneD}{+0.223}

\newcommand{\NEFMistillSOneFollowFlip}{0.098}

\newcommand{\NEFMistillSTwoD}{+0.178}

\newcommand{\NEFVanillaBfSZeroD}{+0.205}

\newcommand{\NEFVanillaBfSOneD}{+0.174}

\newcommand{\NEFVanillaBfSTwoD}{+0.208}

\newcommand{\NEFCtrlMaxAgreeShift}{0.014}
\newcommand{\NEFCtrlMaxExpShift}{0.020}
\newcommand{\NEFCtrlN}{32}
\newcommand{\NEFPrimaryN}{33}
\newcommand{\NEFPrimaryMean}{+0.019}
\newcommand{\NEFPrimaryLo}{\ensuremath{-}0.023}
\newcommand{\NEFPrimaryHi}{+0.061}

\newcommand{\NEFPrimaryHigher}{15}
\newcommand{\NEFPrimaryLower}{13}
\newcommand{\NEFPrimaryTies}{5}
\newcommand{\NEFPrimaryP}{0.85}

\newcommand{\NEFPrimarySD}{0.125}
\newcommand{\NEFPrimaryDetectable}{0.061}

\newcommand{\NEFSecondaryMean}{\ensuremath{-}0.006}
\newcommand{\NEFSecondaryLo}{\ensuremath{-}0.057}
\newcommand{\NEFSecondaryHi}{+0.042}

\newcommand{\NEFCausalTeacherCleanMean}{+0.413}

\newcommand{\NEFCausalTeacherNamedSelMean}{\ensuremath{-}0.023}
\newcommand{\NEFCausalTeacherNamedSelLo}{\ensuremath{-}0.091}
\newcommand{\NEFCausalTeacherNamedSelHi}{+0.045}

\newcommand{\NEFCausalTeacherTopFortySelMean}{+0.030}
\newcommand{\NEFCausalTeacherTopFortySelLo}{\ensuremath{-}0.049}
\newcommand{\NEFCausalTeacherTopFortySelHi}{+0.106}

\newcommand{\NEFCausalMistillSZeroTopFortyShamMean}{+0.095}

\newcommand{\NEFCausalMistillSZeroTopFortyTargetMean}{+0.098}

\newcommand{\NEFCausalMistillSZeroTopFortySelMean}{+0.004}
\newcommand{\NEFCausalMistillSZeroTopFortySelLo}{\ensuremath{-}0.068}
\newcommand{\NEFCausalMistillSZeroTopFortySelHi}{+0.076}

\newcommand{\NEFCausalVanillaBfSZeroTopFortySelMean}{+0.042}
\newcommand{\NEFCausalVanillaBfSZeroTopFortySelLo}{\ensuremath{-}0.011}
\newcommand{\NEFCausalVanillaBfSZeroTopFortySelHi}{+0.095}

\newcommand{\NEFLMean}{\ensuremath{-}0.038}
\newcommand{\NEFLLo}{\ensuremath{-}0.121}
\newcommand{\NEFLHi}{+0.045}

\newcommand{\NEFLN}{33}

\newcommand{\NEFSensLMean}{\ensuremath{-}0.034}
\newcommand{\NEFSensLLo}{\ensuremath{-}0.129}
\newcommand{\NEFSensLHi}{+0.060}

\newcommand{\NEFSensN}{29}

\newcommand{\NEFBackendTeacherMean}{+0.098}
\newcommand{\NEFBackendTeacherLo}{+0.019}
\newcommand{\NEFBackendTeacherHi}{+0.174}

\newcommand{\NEFBackendMistillSZeroMean}{\ensuremath{-}0.011}
\newcommand{\NEFBackendMistillSZeroLo}{\ensuremath{-}0.102}
\newcommand{\NEFBackendMistillSZeroHi}{+0.083}

\newcommand{\NEFBackendVanillaBfSZeroMean}{+0.004}
\newcommand{\NEFBackendVanillaBfSZeroLo}{\ensuremath{-}0.068}
\newcommand{\NEFBackendVanillaBfSZeroHi}{+0.080}

\newcommand{\NEFPrecisionLSD}{0.251}
\newcommand{\NEFPrecisionLDetectable}{0.122}

\newcommand{\NEFPrecisionTargetShamTeacherDetectable}{0.114}

\newcommand{\NEFPrecisionTargetShamCotkdDetectable}{0.077}
\newcommand{\NEFCohortDrawn}{300}
\newcommand{\NEFCohortProposals}{232}
\newcommand{\NEFCohortReviewed}{80}
\newcommand{\NEFCohortAccepted}{33}
\newcommand{\NEFCohortExcluded}{47}
\newcommand{\NEFCohortNotReached}{152}

\newcommand{\NEFCohortCandidateAttempts}{406}
\newcommand{\NEFCohortPropRejected}{174}
\newcommand{\NEFCohortNoProposal}{68}
\newcommand{\NEFCohortPropNoAEvidence}{118}
\newcommand{\NEFCohortPropNoPartner}{53}
\newcommand{\NEFCohortPropNoBWords}{3}
\newcommand{\NEFCohortRuleOne}{5}
\newcommand{\NEFCohortRuleTwo}{2}
\newcommand{\NEFCohortRuleThree}{40}

\newcommand{\NMQTrain}{2{,}000}
\newcommand{\NMQTrainGroups}{461}
\newcommand{\NMQVal}{1{,}203}
\newcommand{\NMQEval}{1{,}610}
\newcommand{\NMQEvalGroups}{398}
\newcommand{\NMQRankQ}{102}
\newcommand{\NMQRankGroups}{45}
\newcommand{\NMQPairs}{16}
\newcommand{\NMQPairGroups}{11}
\newcommand{\NMQCells}{15}
\newcommand{\NMQNoEvents}{81}
\newcommand{\NMQLambda}{0.01}
\newcommand{\NMQRecMean}{\ensuremath{-}0.002}
\newcommand{\NMQRecLo}{\ensuremath{-}0.025}
\newcommand{\NMQRecHi}{+0.019}
\newcommand{\NMQRecN}{73}

\newcommand{\NMQRecSZeroMean}{+0.007}

\newcommand{\NMQRecSZeroMeanA}{0.425}
\newcommand{\NMQRecSZeroMeanB}{0.418}
\newcommand{\NMQRecSOneMean}{\ensuremath{-}0.010}

\newcommand{\NMQRecSOneMeanA}{0.442}
\newcommand{\NMQRecSOneMeanB}{0.452}
\newcommand{\NMQRecSTwoMean}{\ensuremath{-}0.002}

\newcommand{\NMQRecSTwoMeanA}{0.444}
\newcommand{\NMQRecSTwoMeanB}{0.446}
\newcommand{\NMQRecRandMean}{\ensuremath{-}0.068}
\newcommand{\NMQRecRandLo}{\ensuremath{-}0.103}
\newcommand{\NMQRecRandHi}{\ensuremath{-}0.035}
\newcommand{\NMQRecRandN}{71}

\newcommand{\NMQRecRandSZeroMeanB}{0.523}

\newcommand{\NMQRecRandSOneMeanB}{0.508}

\newcommand{\NMQRecRandSTwoMeanB}{0.515}
\newcommand{\NMQRecGroups}{41}
\newcommand{\NMQRecTrainedMin}{0.418}
\newcommand{\NMQRecTrainedMax}{0.523}
\newcommand{\NMQRecBaseKi}{0.28}
\newcommand{\NMQTaskMean}{\ensuremath{-}0.0018}
\newcommand{\NMQTaskLo}{\ensuremath{-}0.0114}
\newcommand{\NMQTaskHi}{+0.0073}

\newcommand{\NMQTaskRandMean}{+0.0040}
\newcommand{\NMQTaskRandLo}{\ensuremath{-}0.0053}
\newcommand{\NMQTaskRandHi}{+0.0138}

\newcommand{\NMQTaskRtxMean}{\ensuremath{-}0.0015}
\newcommand{\NMQTaskRtxLo}{\ensuremath{-}0.0103}
\newcommand{\NMQTaskRtxHi}{+0.0075}

\newcommand{\NMQTaskMargin}{0.02}
\newcommand{\NMQFBase}{0.547}
\newcommand{\NMQValidBase}{0.989}
\newcommand{\NMQFCotKdSZero}{0.550}
\newcommand{\NMQValidCotKdSZero}{0.998}
\newcommand{\NMQFCotKdSOne}{0.536}
\newcommand{\NMQValidCotKdSOne}{0.998}
\newcommand{\NMQFCotKdSTwo}{0.531}
\newcommand{\NMQValidCotKdSTwo}{0.999}
\newcommand{\NMQFMistillSZero}{0.541}
\newcommand{\NMQValidMistillSZero}{0.997}
\newcommand{\NMQFMistillSOne}{0.526}
\newcommand{\NMQValidMistillSOne}{0.998}
\newcommand{\NMQFMistillSTwo}{0.545}
\newcommand{\NMQValidMistillSTwo}{0.998}
\newcommand{\NMQFRandHeadsSZero}{0.533}
\newcommand{\NMQValidRandHeadsSZero}{0.994}
\newcommand{\NMQFRandHeadsSOne}{0.526}
\newcommand{\NMQValidRandHeadsSOne}{0.993}
\newcommand{\NMQFRandHeadsSTwo}{0.540}
\newcommand{\NMQValidRandHeadsSTwo}{0.993}
\newcommand{\NMQBehMean}{+0.026}
\newcommand{\NMQBehLo}{\ensuremath{-}0.013}
\newcommand{\NMQBehHi}{+0.050}
\newcommand{\NMQBehSZeroDelta}{+0.031}

\newcommand{\NMQBehSOneDelta}{\ensuremath{-}0.016}

\newcommand{\NMQBehSTwoDelta}{+0.062}

\newcommand{\NMQBehRandMean}{+0.026}
\newcommand{\NMQBehRandLo}{\ensuremath{-}0.021}
\newcommand{\NMQBehRandHi}{+0.071}

\newcommand{\NMQPBase}{0.781}

\newcommand{\NMQPCotKdSZero}{0.672}

\newcommand{\NMQPCotKdSOne}{0.719}

\newcommand{\NMQPCotKdSTwo}{0.656}

\newcommand{\NMQPMistillSZero}{0.703}

\newcommand{\NMQPMistillSOne}{0.703}

\newcommand{\NMQPMistillSTwo}{0.719}

\newcommand{\NMQPRandHeadsSZero}{0.672}

\newcommand{\NMQPRandHeadsSOne}{0.703}

\newcommand{\NMQPRandHeadsSTwo}{0.672}

\newcommand{\NMQGateCleanPMean}{0.917}
\newcommand{\NMQGateCleanPLo}{0.826}
\newcommand{\NMQGateCleanPHi}{1.000}
\newcommand{\NMQGateCleanPPair}{0.891}
\newcommand{\NMQGateEPMean}{+0.110}
\newcommand{\NMQGateEPLo}{+0.000}
\newcommand{\NMQGateEPHi}{+0.273}
\newcommand{\NMQGateEPDetectDelta}{0.195}
\newcommand{\NMQGateEPRandMean}{+0.087}
\newcommand{\NMQGateEPRandLo}{\ensuremath{-}0.025}
\newcommand{\NMQGateEPRandHi}{+0.253}
\newcommand{\NMQGateEPRandDetectDelta}{0.199}
\newcommand{\NMQGatePairEPMean}{+0.109}
\newcommand{\NMQGatePairEPLo}{+0.000}
\newcommand{\NMQGatePairEPHi}{+0.250}
\newcommand{\NMQGatePairEPRandMean}{+0.073}
\newcommand{\NMQGatePairEPRandLo}{\ensuremath{-}0.032}
\newcommand{\NMQGatePairEPRandHi}{+0.229}
\newcommand{\NMQGateDeltaZero}{0.10}
\newcommand{\NMQGateSlots}{1{,}536}

\newcommand{\NXTEurDevScreened}{80}
\newcommand{\NXTEurDevWithEvents}{78}
\newcommand{\NXTEurDevEvaluable}{76}
\newcommand{\NXTEurDevEvents}{580}
\newcommand{\NXTEurDevEventsEvaluable}{476}

\newcommand{\NXTEurDevRecMean}{0.401}
\newcommand{\NXTEurDevRecLo}{0.323}
\newcommand{\NXTEurDevRecHi}{0.483}

\newcommand{\NXTEurDevDeltaMean}{+0.390}
\newcommand{\NXTEurDevDeltaLo}{+0.304}
\newcommand{\NXTEurDevDeltaHi}{+0.479}
\newcommand{\NXTEurDevAbove}{69}

\newcommand{\NXTEurDevNecMean}{1.011}

\newcommand{\NXTEurDevNecRandMean}{+0.038}

\newcommand{\NXTEurDevRbRandMean}{+0.044}
\newcommand{\NXTEurDevRbRandLo}{+0.038}
\newcommand{\NXTEurDevRbRandHi}{+0.050}
\newcommand{\NXTEurDevRbRandHigher}{78}
\newcommand{\NXTEurDevRbRandN}{78}
\newcommand{\NXTEurDevRbBottomMean}{+0.032}
\newcommand{\NXTEurDevRbBottomLo}{+0.027}
\newcommand{\NXTEurDevRbBottomHi}{+0.037}
\newcommand{\NXTEurDevRbBottomHigher}{77}
\newcommand{\NXTEurDevRbBottomN}{78}

\newcommand{\NXTEurConfScreened}{100}
\newcommand{\NXTEurConfWithEvents}{97}
\newcommand{\NXTEurConfEvaluable}{94}
\newcommand{\NXTEurConfEvents}{668}
\newcommand{\NXTEurConfEventsEvaluable}{553}

\newcommand{\NXTEurConfRecMean}{0.452}
\newcommand{\NXTEurConfRecLo}{0.336}
\newcommand{\NXTEurConfRecHi}{0.586}
\newcommand{\NXTEurConfRecRandMean}{0.023}

\newcommand{\NXTEurConfDeltaMean}{+0.429}
\newcommand{\NXTEurConfDeltaLo}{+0.295}
\newcommand{\NXTEurConfDeltaHi}{+0.579}
\newcommand{\NXTEurConfAbove}{80}

\newcommand{\NXTEurConfNecMean}{0.982}

\newcommand{\NXTEurConfNecRandMean}{+0.095}

\newcommand{\NXTEurConfRecHundredMean}{0.700}

\newcommand{\NXTEurConfVerdict}{CONFIRMED}
\newcommand{\NXTEurConfRbRandMean}{+0.050}
\newcommand{\NXTEurConfRbRandLo}{+0.044}
\newcommand{\NXTEurConfRbRandHi}{+0.056}
\newcommand{\NXTEurConfRbRandHigher}{97}
\newcommand{\NXTEurConfRbRandN}{97}
\newcommand{\NXTEurConfRbBottomMean}{+0.034}
\newcommand{\NXTEurConfRbBottomLo}{+0.030}
\newcommand{\NXTEurConfRbBottomHi}{+0.039}
\newcommand{\NXTEurConfRbBottomHigher}{96}
\newcommand{\NXTEurConfRbBottomN}{97}

\newcommand{\NXTBcbDevScreened}{80}
\newcommand{\NXTBcbDevWithEvents}{42}
\newcommand{\NXTBcbDevEvaluable}{41}
\newcommand{\NXTBcbDevEvents}{196}
\newcommand{\NXTBcbDevEventsEvaluable}{193}

\newcommand{\NXTBcbDevRecMean}{0.323}
\newcommand{\NXTBcbDevRecLo}{0.281}
\newcommand{\NXTBcbDevRecHi}{0.367}

\newcommand{\NXTBcbDevDeltaMean}{+0.315}
\newcommand{\NXTBcbDevDeltaLo}{+0.272}
\newcommand{\NXTBcbDevDeltaHi}{+0.360}
\newcommand{\NXTBcbDevAbove}{40}

\newcommand{\NXTBcbDevNecMean}{0.442}

\newcommand{\NXTBcbDevNecRandMean}{+0.019}

\newcommand{\NXTBcbDevRbRandMean}{+0.013}
\newcommand{\NXTBcbDevRbRandLo}{+0.008}
\newcommand{\NXTBcbDevRbRandHi}{+0.018}
\newcommand{\NXTBcbDevRbRandHigher}{36}
\newcommand{\NXTBcbDevRbRandN}{42}
\newcommand{\NXTBcbDevRbBottomMean}{+0.003}
\newcommand{\NXTBcbDevRbBottomLo}{\ensuremath{-}0.001}
\newcommand{\NXTBcbDevRbBottomHi}{+0.007}
\newcommand{\NXTBcbDevRbBottomHigher}{24}
\newcommand{\NXTBcbDevRbBottomN}{42}

\newcommand{\NXTBcbConfScreened}{100}
\newcommand{\NXTBcbConfWithEvents}{57}
\newcommand{\NXTBcbConfEvaluable}{56}
\newcommand{\NXTBcbConfEvents}{311}
\newcommand{\NXTBcbConfEventsEvaluable}{280}

\newcommand{\NXTBcbConfRecMean}{0.368}
\newcommand{\NXTBcbConfRecLo}{0.283}
\newcommand{\NXTBcbConfRecHi}{0.474}
\newcommand{\NXTBcbConfRecRandMean}{0.016}

\newcommand{\NXTBcbConfDeltaMean}{+0.352}
\newcommand{\NXTBcbConfDeltaLo}{+0.264}
\newcommand{\NXTBcbConfDeltaHi}{+0.467}
\newcommand{\NXTBcbConfAbove}{53}

\newcommand{\NXTBcbConfNecMean}{0.865}

\newcommand{\NXTBcbConfNecRandMean}{\ensuremath{-}0.075}

\newcommand{\NXTBcbConfRecHundredMean}{0.563}

\newcommand{\NXTBcbConfVerdict}{CONFIRMED}
\newcommand{\NXTBcbConfRbRandMean}{+0.016}
\newcommand{\NXTBcbConfRbRandLo}{+0.010}
\newcommand{\NXTBcbConfRbRandHi}{+0.023}
\newcommand{\NXTBcbConfRbRandHigher}{48}
\newcommand{\NXTBcbConfRbRandN}{57}
\newcommand{\NXTBcbConfRbBottomMean}{+0.003}
\newcommand{\NXTBcbConfRbBottomLo}{\ensuremath{-}0.002}
\newcommand{\NXTBcbConfRbBottomHi}{+0.009}
\newcommand{\NXTBcbConfRbBottomHigher}{32}
\newcommand{\NXTBcbConfRbBottomN}{57}

\newcommand{\NXTEurNativeTeacherMean}{0.270}
\newcommand{\NXTEurNativeTeacherLo}{0.239}
\newcommand{\NXTEurNativeTeacherHi}{0.303}
\newcommand{\NXTEurNativeRefMean}{0.671}
\newcommand{\NXTEurNativeRefLo}{0.628}
\newcommand{\NXTEurNativeRefHi}{0.711}

\newcommand{\NXTEurNativeRefN}{5}

\newcommand{\NXTBcbNativePass}{0.57}
\newcommand{\NXTBcbNativePassLo}{0.47}
\newcommand{\NXTBcbNativePassHi}{0.66}
\newcommand{\NXTBcbNativePassCount}{57}
\newcommand{\NXTBcbNativeFailCount}{41}
\newcommand{\NXTBcbNativeNoCodeCount}{2}

\newcommand{\NXTEurAuditRows}{40}
\newcommand{\NXTEurAuditInputs}{33}
\newcommand{\NXTEurAuditExtraction}{40}
\newcommand{\NXTEurAuditVerifiedRej}{0}
\newcommand{\NXTBcbAuditRows}{40}
\newcommand{\NXTBcbAuditInputs}{30}
\newcommand{\NXTBcbAuditExtraction}{39}
\newcommand{\NXTBcbAuditVerifiedRej}{3}
\newcommand{\NXTBcbResidualAonrNum}{4}
\newcommand{\NXTBcbResidualAonrDen}{219}
\newcommand{\NXTBcbResidualBzeroNum}{3}
\newcommand{\NXTBcbResidualBzeroDen}{79}
\newcommand{\NXTBcbResidualBoneNum}{1}
\newcommand{\NXTBcbResidualBoneDen}{196}
\newcommand{\NXTWikiCovOrig}{25}
\newcommand{\NXTWikiCovOrigN}{50}
\newcommand{\NXTWikiCovRev}{17}
\newcommand{\NXTWikiCovRevN}{50}
\newcommand{\NXTWikiCovMin}{20}
\newcommand{\NXTPilotPtwocKfortyMcMed}{\ensuremath{-}0.023}
\newcommand{\NXTPilotPtwocKfortyMcMean}{\ensuremath{-}0.028}
\newcommand{\NXTPilotPtwocKfortyMcLo}{\ensuremath{-}0.044}
\newcommand{\NXTPilotPtwocKfortyMcHi}{\ensuremath{-}0.015}
\newcommand{\NXTPilotPtwocKfortyMcN}{40}
\newcommand{\NXTPilotPtwocKfortyMcAbove}{11}

\newcommand{\NXTPilotPtwocKfortyMcSZeroMed}{\ensuremath{-}0.0230}

\newcommand{\NXTPilotPtwocKfortyMcSOneMed}{\ensuremath{-}0.0269}

\newcommand{\NXTPilotPtwocKfortyMrMed}{\ensuremath{-}0.006}
\newcommand{\NXTPilotPtwocKfortyMrMean}{\ensuremath{-}0.027}
\newcommand{\NXTPilotPtwocKfortyMrLo}{\ensuremath{-}0.056}
\newcommand{\NXTPilotPtwocKfortyMrHi}{\ensuremath{-}0.006}
\newcommand{\NXTPilotPtwocKfortyMrN}{40}
\newcommand{\NXTPilotPtwocKfortyMrAbove}{15}

\newcommand{\NXTPilotPtwocKfortyMrSZeroMed}{\ensuremath{-}0.0083}

\newcommand{\NXTPilotPtwocKfortyMrSOneMed}{\ensuremath{-}0.0068}

\newcommand{\NXTPilotPtwocKfortyRcMed}{\ensuremath{-}0.001}
\newcommand{\NXTPilotPtwocKfortyRcMean}{\ensuremath{-}0.002}
\newcommand{\NXTPilotPtwocKfortyRcLo}{\ensuremath{-}0.018}
\newcommand{\NXTPilotPtwocKfortyRcHi}{+0.017}
\newcommand{\NXTPilotPtwocKfortyRcN}{40}
\newcommand{\NXTPilotPtwocKfortyRcAbove}{18}

\newcommand{\NXTPilotPtwocKfortyRcSZeroMed}{\ensuremath{-}0.0009}

\newcommand{\NXTPilotPtwocKfortyRcSOneMed}{\ensuremath{-}0.0039}

\newcommand{\NXTPilotPtwocKtwonineMcMed}{+0.002}
\newcommand{\NXTPilotPtwocKtwonineMcMean}{\ensuremath{-}0.005}
\newcommand{\NXTPilotPtwocKtwonineMcLo}{\ensuremath{-}0.021}
\newcommand{\NXTPilotPtwocKtwonineMcHi}{+0.009}
\newcommand{\NXTPilotPtwocKtwonineMcN}{40}
\newcommand{\NXTPilotPtwocKtwonineMcAbove}{21}

\newcommand{\NXTPilotPtwocKtwonineMcSZeroMed}{\ensuremath{-}0.0005}

\newcommand{\NXTPilotPtwocKtwonineMcSOneMed}{+0.0020}

\newcommand{\NXTPilotPtwocKtwonineMrMed}{+0.023}
\newcommand{\NXTPilotPtwocKtwonineMrMean}{+0.019}
\newcommand{\NXTPilotPtwocKtwonineMrLo}{+0.006}
\newcommand{\NXTPilotPtwocKtwonineMrHi}{+0.031}
\newcommand{\NXTPilotPtwocKtwonineMrN}{40}
\newcommand{\NXTPilotPtwocKtwonineMrAbove}{32}

\newcommand{\NXTPilotPtwocKtwonineMrSZeroMed}{+0.0262}

\newcommand{\NXTPilotPtwocKtwonineMrSOneMed}{+0.0198}

\newcommand{\NXTPilotPtwocKtwonineRcMed}{\ensuremath{-}0.022}
\newcommand{\NXTPilotPtwocKtwonineRcMean}{\ensuremath{-}0.024}
\newcommand{\NXTPilotPtwocKtwonineRcLo}{\ensuremath{-}0.038}
\newcommand{\NXTPilotPtwocKtwonineRcHi}{\ensuremath{-}0.011}
\newcommand{\NXTPilotPtwocKtwonineRcN}{40}
\newcommand{\NXTPilotPtwocKtwonineRcAbove}{10}

\newcommand{\NXTPilotPtwocKtwonineRcSZeroMed}{\ensuremath{-}0.0309}

\newcommand{\NXTPilotPtwocKtwonineRcSOneMed}{\ensuremath{-}0.0180}

\newcommand{\NXTPilotRawKfortyMcMed}{\ensuremath{-}0.32}
\newcommand{\NXTPilotRawKfortyMcMean}{\ensuremath{-}0.35}
\newcommand{\NXTPilotRawKfortyMcLo}{\ensuremath{-}0.53}
\newcommand{\NXTPilotRawKfortyMcHi}{\ensuremath{-}0.18}
\newcommand{\NXTPilotRawKfortyMcSZeroMed}{\ensuremath{-}0.38}
\newcommand{\NXTPilotRawKfortyMcSOneMed}{\ensuremath{-}0.33}

\newcommand{\NXTPilotPassOneCotKdSZero}{0.494}
\newcommand{\NXTPilotPassOneCotKdSOne}{0.483}
\newcommand{\NXTPilotPassOneMistillSZero}{0.511}
\newcommand{\NXTPilotPassOneMistillSOne}{0.478}
\newcommand{\NXTPilotPassOneRandHeadsSZero}{0.522}
\newcommand{\NXTPilotPassOneRandHeadsSOne}{0.489}
\newcommand{\NXTPilotPassMcMean}{+0.006}
\newcommand{\NXTPilotPassMcLo}{\ensuremath{-}0.036}
\newcommand{\NXTPilotPassMcHi}{+0.050}

\newcommand{\NXTPilotPassMcN}{180}
\newcommand{\NXTPilotPassMcSZeroMean}{+0.017}
\newcommand{\NXTPilotPassMcSOneMean}{\ensuremath{-}0.006}
\newcommand{\NXTPilotPassMrMean}{\ensuremath{-}0.011}
\newcommand{\NXTPilotPassMrLo}{\ensuremath{-}0.056}
\newcommand{\NXTPilotPassMrHi}{+0.031}

\newcommand{\NXTPilotPassMrSZeroMean}{\ensuremath{-}0.011}
\newcommand{\NXTPilotPassMrSOneMean}{\ensuremath{-}0.011}

\newcommand{\NXTPilotPrimaryMin}{+0.030}
\newcommand{\NXTPilotVsRandomMin}{+0.015}
\newcommand{\NXTPilotPassFloor}{\ensuremath{-}0.030}
\newcommand{\NXTPilotThresholdsAt}{2026-10-28 13:37}
\newcommand{\NXTPilotLockAt}{2026-10-28 08:14}
\newcommand{\NXTPilotMeaningful}{not met}
\newcommand{\NXTPilotAcceptable}{met}

\newcommand{\NXTPilotGapKfortyMean}{\ensuremath{-}0.029}
\newcommand{\NXTPilotGapKfortyLo}{\ensuremath{-}0.108}
\newcommand{\NXTPilotGapKfortyHi}{+0.046}
\newcommand{\NXTPilotGapKfortyN}{40}

\newcommand{\NXTPilotGapKtwonineMean}{\ensuremath{-}0.118}
\newcommand{\NXTPilotGapKtwonineLo}{\ensuremath{-}0.193}
\newcommand{\NXTPilotGapKtwonineHi}{\ensuremath{-}0.046}

\newcommand{\NXTPilotTrainSplit}{2{,}000}
\newcommand{\NXTPilotTrainUsable}{1{,}991}
\newcommand{\NXTPilotNoEvents}{233}

\newcommand{\NXTPilotPoolPtwo}{40}
\newcommand{\NXTPilotPoolOverlapPtwo}{0}
\newcommand{\NXTPilotPoolPtwoc}{40}
\newcommand{\NXTPilotPoolOverlapPtwoc}{2}
\newcommand{\NXTPilotRecGroups}{40}
\newcommand{\NXTPilotPassGroups}{180}

\newcommand{\NZsDocsBzero}{50}
\newcommand{\NZsDocsBone}{80}
\newcommand{\NZsDocsConf}{100}
\newcommand{\NZsDocsEn}{100}
\newcommand{\NZsMqQuestions}{102}
\newcommand{\NZsBoneMcMean}{+0.063}
\newcommand{\NZsBoneMcLo}{+0.037}
\newcommand{\NZsBoneMcHi}{+0.089}

\newcommand{\NZsBoneMcAbove}{56}
\newcommand{\NZsBoneMcBelow}{18}
\newcommand{\NZsBoneMcN}{74}

\newcommand{\NZsBoneMrMean}{+0.048}
\newcommand{\NZsBoneMrLo}{+0.014}
\newcommand{\NZsBoneMrHi}{+0.082}

\newcommand{\NZsBoneMrN}{74}

\newcommand{\NZsBoneMedMistillRange}{0.88--0.91}

\newcommand{\NZsBoneMedCotKdRange}{0.80--0.82}

\newcommand{\NZsBoneMedRandHeadsRange}{0.79--0.83}
\newcommand{\NZsBoneMedBase}{0.73}
\newcommand{\NZsConfMcMean}{+0.069}
\newcommand{\NZsConfMcLo}{+0.050}
\newcommand{\NZsConfMcHi}{+0.087}

\newcommand{\NZsConfMcAbove}{76}
\newcommand{\NZsConfMcBelow}{15}
\newcommand{\NZsConfMcN}{91}

\newcommand{\NZsConfMrMean}{+0.064}
\newcommand{\NZsConfMrLo}{+0.037}
\newcommand{\NZsConfMrHi}{+0.093}

\newcommand{\NZsConfMrN}{91}

\newcommand{\NZsConfMedMistillRange}{0.89--0.92}

\newcommand{\NZsConfMedCotKdRange}{0.82--0.85}

\newcommand{\NZsConfMedRandHeadsRange}{0.83--0.84}
\newcommand{\NZsConfMedBase}{0.74}
\newcommand{\NZsEnMcMean}{+0.071}
\newcommand{\NZsEnMcLo}{+0.044}
\newcommand{\NZsEnMcHi}{+0.099}

\newcommand{\NZsEnMcAbove}{68}
\newcommand{\NZsEnMcBelow}{24}
\newcommand{\NZsEnMcN}{92}

\newcommand{\NZsEnMrMean}{+0.045}
\newcommand{\NZsEnMrLo}{+0.016}
\newcommand{\NZsEnMrHi}{+0.075}

\newcommand{\NZsEnMrN}{92}

\newcommand{\NZsEnMedMistillRange}{0.89--0.91}

\newcommand{\NZsEnMedCotKdRange}{0.80--0.84}

\newcommand{\NZsEnMedRandHeadsRange}{0.85--0.89}
\newcommand{\NZsEnMedBase}{0.72}
\newcommand{\NZsBoneKeightMcMean}{+0.099}
\newcommand{\NZsBoneKeightMcLo}{+0.074}
\newcommand{\NZsBoneKeightMcHi}{+0.123}

\newcommand{\NZsBoneKeightMcAbove}{63}
\newcommand{\NZsBoneKeightMcBelow}{11}
\newcommand{\NZsBoneKeightMcN}{74}

\newcommand{\NZsBoneKeightMrMean}{+0.102}
\newcommand{\NZsBoneKeightMrLo}{+0.077}
\newcommand{\NZsBoneKeightMrHi}{+0.126}

\newcommand{\NZsBoneKeightMrN}{74}

\newcommand{\NZsBoneKeightMedMistillRange}{0.71--0.73}

\newcommand{\NZsBoneKeightMedCotKdRange}{0.59--0.62}

\newcommand{\NZsBoneKeightMedRandHeadsRange}{0.60}
\newcommand{\NZsBoneKeightMedBase}{0.56}
\newcommand{\NZsConfKeightMcMean}{+0.117}
\newcommand{\NZsConfKeightMcLo}{+0.096}
\newcommand{\NZsConfKeightMcHi}{+0.138}

\newcommand{\NZsConfKeightMcAbove}{82}
\newcommand{\NZsConfKeightMcBelow}{9}
\newcommand{\NZsConfKeightMcN}{91}

\newcommand{\NZsConfKeightMrMean}{+0.127}
\newcommand{\NZsConfKeightMrLo}{+0.099}
\newcommand{\NZsConfKeightMrHi}{+0.157}

\newcommand{\NZsConfKeightMrN}{91}

\newcommand{\NZsConfKeightMedMistillRange}{0.74}

\newcommand{\NZsConfKeightMedCotKdRange}{0.62}

\newcommand{\NZsConfKeightMedRandHeadsRange}{0.58--0.59}
\newcommand{\NZsConfKeightMedBase}{0.59}
\newcommand{\NZsEnKeightMcMean}{+0.100}
\newcommand{\NZsEnKeightMcLo}{+0.074}
\newcommand{\NZsEnKeightMcHi}{+0.124}

\newcommand{\NZsEnKeightMcAbove}{73}
\newcommand{\NZsEnKeightMcBelow}{19}
\newcommand{\NZsEnKeightMcN}{92}

\newcommand{\NZsEnKeightMrMean}{+0.100}
\newcommand{\NZsEnKeightMrLo}{+0.073}
\newcommand{\NZsEnKeightMrHi}{+0.126}

\newcommand{\NZsEnKeightMrN}{92}

\newcommand{\NZsEnKeightMedMistillRange}{0.71}

\newcommand{\NZsEnKeightMedCotKdRange}{0.59}

\newcommand{\NZsEnKeightMedRandHeadsRange}{0.61--0.62}
\newcommand{\NZsEnKeightMedBase}{0.53}
\newcommand{\NZsMqMcMean}{+0.033}
\newcommand{\NZsMqMcLo}{+0.017}
\newcommand{\NZsMqMcHi}{+0.049}

\newcommand{\NZsMqMcAbove}{51}
\newcommand{\NZsMqMcBelow}{15}
\newcommand{\NZsMqMcN}{66}

\newcommand{\NZsMqMrMean}{+0.039}
\newcommand{\NZsMqMrLo}{+0.014}
\newcommand{\NZsMqMrHi}{+0.067}

\newcommand{\NZsMqMrN}{66}

\newcommand{\NZsMqMedMistillRange}{0.36--0.37}

\newcommand{\NZsMqMedCotKdRange}{0.30--0.32}

\newcommand{\NZsMqMedRandHeadsRange}{0.32--0.33}
\newcommand{\NZsMqMedBase}{0.28}
\newcommand{\NZsSmolBoneMcMean}{+0.253}
\newcommand{\NZsSmolBoneMcLo}{+0.167}
\newcommand{\NZsSmolBoneMcHi}{+0.341}

\newcommand{\NZsSmolBoneMcAbove}{60}
\newcommand{\NZsSmolBoneMcBelow}{16}
\newcommand{\NZsSmolBoneMcN}{76}

\newcommand{\NZsSmolBoneMfMean}{+0.254}
\newcommand{\NZsSmolBoneMfLo}{+0.185}
\newcommand{\NZsSmolBoneMfHi}{+0.321}

\newcommand{\NZsSmolBoneMrMean}{+0.293}
\newcommand{\NZsSmolBoneMrLo}{+0.195}
\newcommand{\NZsSmolBoneMrHi}{+0.394}

\newcommand{\NZsSmolBoneMrN}{75}

\newcommand{\NZsSmolBoneMedMistillLo}{0.75}
\newcommand{\NZsSmolBoneMedMistillHi}{0.80}
\newcommand{\NZsSmolBoneMedMistillRange}{0.75--0.80}
\newcommand{\NZsSmolBoneMedCotKd}{0.54}

\newcommand{\NZsSmolBoneMedRandHeads}{0.52}
\newcommand{\NZsSmolBoneMedBase}{0.44}
\newcommand{\NZsSmolConfMcMean}{+0.245}
\newcommand{\NZsSmolConfMcLo}{+0.162}
\newcommand{\NZsSmolConfMcHi}{+0.323}

\newcommand{\NZsSmolConfMcAbove}{73}
\newcommand{\NZsSmolConfMcBelow}{18}
\newcommand{\NZsSmolConfMcN}{91}

\newcommand{\NZsSmolConfMrMean}{+0.147}
\newcommand{\NZsSmolConfMrLo}{+0.061}
\newcommand{\NZsSmolConfMrHi}{+0.229}

\newcommand{\NZsSmolConfMrN}{92}

\newcommand{\NZsSmolConfMedMistillRange}{0.80--0.81}
\newcommand{\NZsSmolConfMedCotKd}{0.51}

\newcommand{\NZsSmolConfMedRandHeads}{0.55}
\newcommand{\NZsSmolConfMedBase}{0.49}

\newcommand{\NEaCxDocs}{250}
\newcommand{\NEmCharts}{302}
\newcommand{\NEmInputs}{299}
\newcommand{\NEmNoGold}{3}
\newcommand{\NEmGold}{2{,}177}
\newcommand{\NEmGoldMedian}{7}

\newcommand{\NEmSpans}{3{,}223}
\newcommand{\NEmSpansKept}{2{,}311}
\newcommand{\NEmSpansCut}{598}
\newcommand{\NEmSpansOut}{314}
\newcommand{\NEmWordsMedian}{1{,}360}

\newcommand{\NEmTraces}{299}

\newcommand{\NEaCxTeach}{4.62}
\newcommand{\NEaCxTeachLo}{4.29}
\newcommand{\NEaCxTeachHi}{4.95}

\newcommand{\NEaCxLiftMcMean}{+0.54}
\newcommand{\NEaCxLiftMcLo}{+0.49}
\newcommand{\NEaCxLiftMcHi}{+0.60}

\newcommand{\NEaCxLiftMcAbove}{220}
\newcommand{\NEaCxLiftMcBelow}{8}
\newcommand{\NEaCxLiftMcN}{228}

\newcommand{\NEaCxLiftMrMean}{\ensuremath{-}0.04}
\newcommand{\NEaCxLiftMrLo}{\ensuremath{-}0.07}
\newcommand{\NEaCxLiftMrHi}{\ensuremath{-}0.02}

\newcommand{\NEaCxLiftMrAbove}{84}
\newcommand{\NEaCxLiftMrBelow}{144}
\newcommand{\NEaCxLiftMrN}{228}

\newcommand{\NEaCxLiftMfMean}{+0.82}
\newcommand{\NEaCxLiftMfLo}{+0.75}
\newcommand{\NEaCxLiftMfHi}{+0.90}

\newcommand{\NEaCxLiftMfAbove}{220}
\newcommand{\NEaCxLiftMfBelow}{8}
\newcommand{\NEaCxLiftMfN}{228}

\newcommand{\NEaCxLiftMwMean}{+0.11}
\newcommand{\NEaCxLiftMwLo}{+0.09}
\newcommand{\NEaCxLiftMwHi}{+0.13}

\newcommand{\NEaCxLiftMwAbove}{197}
\newcommand{\NEaCxLiftMwBelow}{31}
\newcommand{\NEaCxLiftMwN}{228}

\newcommand{\NEaCxLiftLvlMistillRange}{3.30--3.35}

\newcommand{\NEaCxLiftLvlCotKdRange}{2.74--2.80}

\newcommand{\NEaCxLiftLvlRandHeadsRange}{3.35--3.38}

\newcommand{\NEaCxLiftLvlMistillFpRange}{3.56--3.61}

\newcommand{\NEaCxLiftLvlCotKdFpRange}{2.71--2.80}

\newcommand{\NEaCxLiftLvlNoWin}{3.20}
\newcommand{\NEaCxLiftLvlBase}{2.82}
\newcommand{\NEaCxAbsMcMean}{+0.0188}
\newcommand{\NEaCxAbsMcLo}{+0.0176}
\newcommand{\NEaCxAbsMcHi}{+0.0201}

\newcommand{\NEaCxAbsMcAbove}{227}
\newcommand{\NEaCxAbsMcBelow}{1}
\newcommand{\NEaCxAbsMcN}{228}

\newcommand{\NEaCxAbsMrMean}{+0.0075}
\newcommand{\NEaCxAbsMrLo}{+0.0069}
\newcommand{\NEaCxAbsMrHi}{+0.0080}

\newcommand{\NEaCxAbsMrAbove}{225}
\newcommand{\NEaCxAbsMrBelow}{3}
\newcommand{\NEaCxAbsMrN}{228}

\newcommand{\NEaCxAbsMwMean}{+0.0122}
\newcommand{\NEaCxAbsMwLo}{+0.0114}
\newcommand{\NEaCxAbsMwHi}{+0.0129}

\newcommand{\NEaCxAbsMwAbove}{227}
\newcommand{\NEaCxAbsMwBelow}{1}
\newcommand{\NEaCxAbsMwN}{228}

\newcommand{\NEaCxAbsLvlMistillRange}{0.039--0.040}
\newcommand{\NEaCxAbsLvlMistillAvg}{0.039}

\newcommand{\NEaCxAbsLvlCotKdRange}{0.020--0.021}
\newcommand{\NEaCxAbsLvlCotKdAvg}{0.020}

\newcommand{\NEaCxAbsLvlRandHeadsRange}{0.032}

\newcommand{\NEaCxAbsLvlMistillFpRange}{0.042--0.044}

\newcommand{\NEaCxAbsLvlCotKdFpRange}{0.019--0.020}

\newcommand{\NEaCxAbsLvlNoWin}{0.027}
\newcommand{\NEaCxAbsLvlBase}{0.021}
\newcommand{\NEaCxNoteMcMean}{+0.036}
\newcommand{\NEaCxNoteMcLo}{+0.034}
\newcommand{\NEaCxNoteMcHi}{+0.038}

\newcommand{\NEaCxNoteMcAbove}{226}
\newcommand{\NEaCxNoteMcBelow}{2}
\newcommand{\NEaCxNoteMcN}{228}

\newcommand{\NEaCxNoteMrMean}{+0.018}
\newcommand{\NEaCxNoteMrLo}{+0.017}
\newcommand{\NEaCxNoteMrHi}{+0.019}

\newcommand{\NEaCxNoteMrAbove}{225}
\newcommand{\NEaCxNoteMrBelow}{3}
\newcommand{\NEaCxNoteMrN}{228}

\newcommand{\NEaCxPcMcMean}{+0.32}
\newcommand{\NEaCxPcMcLo}{+0.23}
\newcommand{\NEaCxPcMcHi}{+0.41}

\newcommand{\NEaCxPcMcAbove}{137}
\newcommand{\NEaCxPcMcBelow}{38}
\newcommand{\NEaCxPcMcN}{175}

\newcommand{\NEaCxPcMrMean}{+0.40}
\newcommand{\NEaCxPcMrLo}{+0.29}
\newcommand{\NEaCxPcMrHi}{+0.51}

\newcommand{\NEaCxPcMrAbove}{130}
\newcommand{\NEaCxPcMrBelow}{45}
\newcommand{\NEaCxPcMrN}{175}

\newcommand{\NEaCxSupMcMean}{\ensuremath{-}0.24}
\newcommand{\NEaCxSupMcLo}{\ensuremath{-}0.27}
\newcommand{\NEaCxSupMcHi}{\ensuremath{-}0.21}

\newcommand{\NEaCxSupMcAbove}{23}
\newcommand{\NEaCxSupMcBelow}{205}
\newcommand{\NEaCxSupMcN}{228}

\newcommand{\NEaCxSupNoteMcMean}{+0.027}
\newcommand{\NEaCxSupNoteMcLo}{+0.026}
\newcommand{\NEaCxSupNoteMcHi}{+0.029}

\newcommand{\NEaCxSupNoteMcAbove}{225}
\newcommand{\NEaCxSupNoteMcBelow}{3}
\newcommand{\NEaCxSupNoteMcN}{228}

\newcommand{\NEaCxLvlTeacherLift}{4.62}
\newcommand{\NEaCxLvlTeacherAbs}{0.082}

\newcommand{\NEaMdTeach}{15.77}
\newcommand{\NEaMdTeachLo}{14.73}
\newcommand{\NEaMdTeachHi}{16.88}

\newcommand{\NEaMdLiftMcMean}{+2.04}
\newcommand{\NEaMdLiftMcLo}{+1.88}
\newcommand{\NEaMdLiftMcHi}{+2.19}

\newcommand{\NEaMdLiftMcAbove}{260}
\newcommand{\NEaMdLiftMcBelow}{4}
\newcommand{\NEaMdLiftMcN}{264}

\newcommand{\NEaMdLiftMrMean}{+0.05}
\newcommand{\NEaMdLiftMrLo}{\ensuremath{-}0.05}
\newcommand{\NEaMdLiftMrHi}{+0.15}

\newcommand{\NEaMdLiftMrAbove}{141}
\newcommand{\NEaMdLiftMrBelow}{123}
\newcommand{\NEaMdLiftMrN}{264}

\newcommand{\NEaMdLiftMfMean}{+4.03}
\newcommand{\NEaMdLiftMfLo}{+3.75}
\newcommand{\NEaMdLiftMfHi}{+4.32}

\newcommand{\NEaMdLiftMfAbove}{261}
\newcommand{\NEaMdLiftMfBelow}{3}
\newcommand{\NEaMdLiftMfN}{264}

\newcommand{\NEaMdLiftMwMean}{+1.00}
\newcommand{\NEaMdLiftMwLo}{+0.91}
\newcommand{\NEaMdLiftMwHi}{+1.09}

\newcommand{\NEaMdLiftMwAbove}{255}
\newcommand{\NEaMdLiftMwBelow}{9}
\newcommand{\NEaMdLiftMwN}{264}

\newcommand{\NEaMdLiftLvlMistillRange}{10.39--10.64}

\newcommand{\NEaMdLiftLvlCotKdRange}{8.46--8.49}

\newcommand{\NEaMdLiftLvlRandHeadsRange}{10.43--10.49}

\newcommand{\NEaMdLiftLvlMistillFpRange}{11.49--11.76}

\newcommand{\NEaMdLiftLvlCotKdFpRange}{7.47--7.74}

\newcommand{\NEaMdLiftLvlNoWin}{9.39}
\newcommand{\NEaMdLiftLvlBase}{6.03}
\newcommand{\NEaMdAbsMcMean}{+0.0103}
\newcommand{\NEaMdAbsMcLo}{+0.0096}
\newcommand{\NEaMdAbsMcHi}{+0.0110}

\newcommand{\NEaMdAbsMcAbove}{264}
\newcommand{\NEaMdAbsMcBelow}{0}
\newcommand{\NEaMdAbsMcN}{264}

\newcommand{\NEaMdAbsMrMean}{+0.0041}
\newcommand{\NEaMdAbsMrLo}{+0.0037}
\newcommand{\NEaMdAbsMrHi}{+0.0045}

\newcommand{\NEaMdAbsMrAbove}{249}
\newcommand{\NEaMdAbsMrBelow}{15}
\newcommand{\NEaMdAbsMrN}{264}

\newcommand{\NEaMdAbsMwMean}{+0.0079}
\newcommand{\NEaMdAbsMwLo}{+0.0073}
\newcommand{\NEaMdAbsMwHi}{+0.0084}

\newcommand{\NEaMdAbsMwAbove}{264}
\newcommand{\NEaMdAbsMwBelow}{0}
\newcommand{\NEaMdAbsMwN}{264}

\newcommand{\NEaMdAbsLvlMistillRange}{0.027}
\newcommand{\NEaMdAbsLvlMistillAvg}{0.027}

\newcommand{\NEaMdAbsLvlCotKdRange}{0.017}
\newcommand{\NEaMdAbsLvlCotKdAvg}{0.017}

\newcommand{\NEaMdAbsLvlRandHeadsRange}{0.023}

\newcommand{\NEaMdAbsLvlMistillFpRange}{0.030}

\newcommand{\NEaMdAbsLvlCotKdFpRange}{0.014}

\newcommand{\NEaMdAbsLvlNoWin}{0.019}
\newcommand{\NEaMdAbsLvlBase}{0.011}
\newcommand{\NEaMdNoteMcMean}{+0.026}
\newcommand{\NEaMdNoteMcLo}{+0.025}
\newcommand{\NEaMdNoteMcHi}{+0.028}

\newcommand{\NEaMdNoteMcAbove}{257}
\newcommand{\NEaMdNoteMcBelow}{7}
\newcommand{\NEaMdNoteMcN}{264}

\newcommand{\NEaMdNoteMrMean}{+0.015}
\newcommand{\NEaMdNoteMrLo}{+0.014}
\newcommand{\NEaMdNoteMrHi}{+0.016}

\newcommand{\NEaMdNoteMrAbove}{255}
\newcommand{\NEaMdNoteMrBelow}{9}
\newcommand{\NEaMdNoteMrN}{264}

\newcommand{\NEaMdPcMcMean}{+1.17}
\newcommand{\NEaMdPcMcLo}{+0.97}
\newcommand{\NEaMdPcMcHi}{+1.38}

\newcommand{\NEaMdPcMcAbove}{174}
\newcommand{\NEaMdPcMcBelow}{25}
\newcommand{\NEaMdPcMcN}{199}

\newcommand{\NEaMdPcMrMean}{+0.30}
\newcommand{\NEaMdPcMrLo}{+0.06}
\newcommand{\NEaMdPcMrHi}{+0.54}

\newcommand{\NEaMdPcMrAbove}{121}
\newcommand{\NEaMdPcMrBelow}{78}
\newcommand{\NEaMdPcMrN}{199}

\newcommand{\NEaMdSupMcMean}{\ensuremath{-}0.83}
\newcommand{\NEaMdSupMcLo}{\ensuremath{-}0.94}
\newcommand{\NEaMdSupMcHi}{\ensuremath{-}0.72}

\newcommand{\NEaMdSupMcAbove}{46}
\newcommand{\NEaMdSupMcBelow}{218}
\newcommand{\NEaMdSupMcN}{264}

\newcommand{\NEaMdSupNoteMcMean}{+0.062}
\newcommand{\NEaMdSupNoteMcLo}{+0.060}
\newcommand{\NEaMdSupNoteMcHi}{+0.065}

\newcommand{\NEaMdSupNoteMcAbove}{264}
\newcommand{\NEaMdSupNoteMcBelow}{0}
\newcommand{\NEaMdSupNoteMcN}{264}

\newcommand{\NEaMdCxhMcMean}{+1.86}
\newcommand{\NEaMdCxhMcLo}{+1.72}
\newcommand{\NEaMdCxhMcHi}{+2.00}

\newcommand{\NEaMdCxhMcAbove}{261}
\newcommand{\NEaMdCxhMcBelow}{3}
\newcommand{\NEaMdCxhMcN}{264}

\newcommand{\NEaMdLvlTeacherLift}{15.77}
\newcommand{\NEaMdLvlTeacherAbs}{0.065}

\newcommand{\NEbConfSupMcMean}{\ensuremath{-}0.22}
\newcommand{\NEbConfSupMcLo}{\ensuremath{-}0.29}
\newcommand{\NEbConfSupMcHi}{\ensuremath{-}0.16}

\newcommand{\NEbConfSupNoteMcMean}{+0.027}

\newcommand{\NEbConfLiftMcMean}{+0.52}
\newcommand{\NEbConfLiftMcLo}{+0.43}
\newcommand{\NEbConfLiftMcHi}{+0.62}

\newcommand{\NEbConfSupAbsMcMean}{+0.0059}

\newcommand{\NEbBoneSupMcMean}{\ensuremath{-}0.23}
\newcommand{\NEbBoneSupMcLo}{\ensuremath{-}0.28}
\newcommand{\NEbBoneSupMcHi}{\ensuremath{-}0.18}

\newcommand{\NEbBoneLiftMcMean}{+0.52}
\newcommand{\NEbBoneLiftMcLo}{+0.44}
\newcommand{\NEbBoneLiftMcHi}{+0.61}

\newcommand{\NEaPhRead}{43}

\newcommand{\NEaPhUp}{41}

\newcommand{\NEaPhUpCi}{38}
\newcommand{\NEaPhDownCi}{0}

\newcommand{\NEaPhWsup}{+0.066}

\newcommand{\NEaPhTopHead}{\head{35}{25}}
\newcommand{\NEaPhTopW}{0.74}

\newcommand{\NEdThrMin}{+0.03}
\newcommand{\NEdCxEvents}{2{,}061}
\newcommand{\NEdCxTeach}{+0.079}
\newcommand{\NEdCxTeachLo}{+0.042}
\newcommand{\NEdCxTeachHi}{+0.114}

\newcommand{\NEdCxDeltaMcMean}{+0.014}
\newcommand{\NEdCxDeltaMcLo}{+0.004}
\newcommand{\NEdCxDeltaMcHi}{+0.026}

\newcommand{\NEdCxDeltaMcAbove}{115}
\newcommand{\NEdCxDeltaMcBelow}{113}
\newcommand{\NEdCxDeltaMcN}{228}

\newcommand{\NEdCxDeltaMrMean}{+0.002}
\newcommand{\NEdCxDeltaMrLo}{\ensuremath{-}0.013}
\newcommand{\NEdCxDeltaMrHi}{+0.017}

\newcommand{\NEdCxDeltaMrAbove}{122}
\newcommand{\NEdCxDeltaMrBelow}{106}
\newcommand{\NEdCxDeltaMrN}{228}

\newcommand{\NEdCxDeltaMfMean}{+0.003}
\newcommand{\NEdCxDeltaMfLo}{\ensuremath{-}0.003}
\newcommand{\NEdCxDeltaMfHi}{+0.008}

\newcommand{\NEdCxDeltaMfAbove}{122}
\newcommand{\NEdCxDeltaMfBelow}{106}
\newcommand{\NEdCxDeltaMfN}{228}

\newcommand{\NEdCxDeltaMwMean}{\ensuremath{-}0.002}
\newcommand{\NEdCxDeltaMwLo}{\ensuremath{-}0.014}
\newcommand{\NEdCxDeltaMwHi}{+0.010}

\newcommand{\NEdCxDeltaMwAbove}{103}
\newcommand{\NEdCxDeltaMwBelow}{125}
\newcommand{\NEdCxDeltaMwN}{228}

\newcommand{\NEdCxDeltaLvlMistillRange}{0.063--0.076}

\newcommand{\NEdCxDeltaLvlCotKdRange}{0.051--0.061}

\newcommand{\NEdCxDeltaLvlRandHeadsRange}{0.065--0.069}

\newcommand{\NEdCxDeltaLvlMistillFpRange}{0.041--0.050}

\newcommand{\NEdCxDeltaLvlCotKdFpRange}{0.036--0.051}

\newcommand{\NEdCxDeltaLvlNoWin}{0.078}
\newcommand{\NEdCxDeltaLvlBase}{\ensuremath{-}0.002}
\newcommand{\NEdCxDeltaLvlTeacher}{0.079}

\newcommand{\NEdCxMediatedMcMean}{+0.015}
\newcommand{\NEdCxMediatedMcLo}{+0.003}
\newcommand{\NEdCxMediatedMcHi}{+0.028}

\newcommand{\NEdCxMediatedMcAbove}{117}
\newcommand{\NEdCxMediatedMcBelow}{111}
\newcommand{\NEdCxMediatedMcN}{228}

\newcommand{\NEdCxAblLvlMistillRange}{0.000--0.008}

\newcommand{\NEdCxAblLvlCotKdRange}{\ensuremath{-}0.001--0.012}

\newcommand{\NEdCxAblLvlTeacher}{0.083}
\newcommand{\NEdCxPcDeltaMcMean}{\ensuremath{-}0.006}
\newcommand{\NEdCxPcDeltaMcLo}{\ensuremath{-}0.016}
\newcommand{\NEdCxPcDeltaMcHi}{+0.002}

\newcommand{\NEdCxPcDeltaMcAbove}{76}
\newcommand{\NEdCxPcDeltaMcBelow}{99}
\newcommand{\NEdCxPcDeltaMcN}{175}

\newcommand{\NEdCxDeltaMcPairZero}{+0.003}
\newcommand{\NEdCxDeltaMcPairOne}{+0.002}
\newcommand{\NEdCxDeltaMcPairTwo}{\ensuremath{-}0.001}
\newcommand{\NEdMdEvents}{2{,}282}
\newcommand{\NEdMdTeach}{+0.085}
\newcommand{\NEdMdTeachLo}{+0.022}
\newcommand{\NEdMdTeachHi}{+0.142}

\newcommand{\NEdMdDeltaMcMean}{+0.003}
\newcommand{\NEdMdDeltaMcLo}{\ensuremath{-}0.005}
\newcommand{\NEdMdDeltaMcHi}{+0.010}

\newcommand{\NEdMdDeltaMcAbove}{146}
\newcommand{\NEdMdDeltaMcBelow}{118}
\newcommand{\NEdMdDeltaMcN}{264}

\newcommand{\NEdMdDeltaMrMean}{+0.018}
\newcommand{\NEdMdDeltaMrLo}{+0.005}
\newcommand{\NEdMdDeltaMrHi}{+0.030}

\newcommand{\NEdMdDeltaMrAbove}{161}
\newcommand{\NEdMdDeltaMrBelow}{103}
\newcommand{\NEdMdDeltaMrN}{264}

\newcommand{\NEdMdDeltaMfMean}{+0.004}
\newcommand{\NEdMdDeltaMfLo}{\ensuremath{-}0.001}
\newcommand{\NEdMdDeltaMfHi}{+0.009}

\newcommand{\NEdMdDeltaMfAbove}{162}
\newcommand{\NEdMdDeltaMfBelow}{102}
\newcommand{\NEdMdDeltaMfN}{264}

\newcommand{\NEdMdDeltaMwMean}{+0.002}
\newcommand{\NEdMdDeltaMwLo}{\ensuremath{-}0.005}
\newcommand{\NEdMdDeltaMwHi}{+0.009}

\newcommand{\NEdMdDeltaMwAbove}{130}
\newcommand{\NEdMdDeltaMwBelow}{134}
\newcommand{\NEdMdDeltaMwN}{264}

\newcommand{\NEdMdDeltaLvlMistillRange}{0.047--0.050}

\newcommand{\NEdMdDeltaLvlCotKdRange}{0.044--0.048}

\newcommand{\NEdMdDeltaLvlRandHeadsRange}{0.028--0.032}

\newcommand{\NEdMdDeltaLvlMistillFpRange}{0.039--0.053}

\newcommand{\NEdMdDeltaLvlCotKdFpRange}{0.035--0.045}

\newcommand{\NEdMdDeltaLvlNoWin}{0.047}
\newcommand{\NEdMdDeltaLvlBase}{\ensuremath{-}0.003}
\newcommand{\NEdMdDeltaLvlTeacher}{0.085}

\newcommand{\NEdMdMediatedMcMean}{\ensuremath{-}0.012}
\newcommand{\NEdMdMediatedMcLo}{\ensuremath{-}0.021}
\newcommand{\NEdMdMediatedMcHi}{\ensuremath{-}0.002}

\newcommand{\NEdMdMediatedMcAbove}{108}
\newcommand{\NEdMdMediatedMcBelow}{156}
\newcommand{\NEdMdMediatedMcN}{264}

\newcommand{\NEdMdAblLvlMistillRange}{0.042--0.066}

\newcommand{\NEdMdAblLvlCotKdRange}{0.040--0.047}

\newcommand{\NEdMdAblLvlTeacher}{0.151}
\newcommand{\NEdMdPcDeltaMcMean}{+0.000}
\newcommand{\NEdMdPcDeltaMcLo}{\ensuremath{-}0.005}
\newcommand{\NEdMdPcDeltaMcHi}{+0.005}

\newcommand{\NEdMdPcDeltaMcAbove}{92}
\newcommand{\NEdMdPcDeltaMcBelow}{107}
\newcommand{\NEdMdPcDeltaMcN}{199}

\newcommand{\NEdMdDeltaMcPairZero}{+0.003}
\newcommand{\NEdMdDeltaMcPairOne}{+0.005}
\newcommand{\NEdMdDeltaMcPairTwo}{+0.002}
\newcommand{\NEsConfQuoteMcMean}{+0.021}
\newcommand{\NEsConfQuoteMcLo}{\ensuremath{-}0.079}
\newcommand{\NEsConfQuoteMcHi}{+0.120}

\newcommand{\NEsTestQuoteMcMean}{+0.020}
\newcommand{\NEsTestQuoteMcLo}{\ensuremath{-}0.039}
\newcommand{\NEsTestQuoteMcHi}{+0.081}

\newcommand{\NEsTestQuoteLvlMistillRange}{0.49--0.58}

\newcommand{\NEsTestQuoteLvlCotKdRange}{0.48--0.51}

\newcommand{\NEsTestQuoteLvlBase}{0.51}

\newcommand{\NTpRbSZeroMed}{0.385}
\newcommand{\NTpRbSZeroLowerSZero}{140}
\newcommand{\NTpRbSZeroLowerSOne}{138}
\newcommand{\NTpRbSZeroLowerSTwo}{135}
\newcommand{\NTpRbSZeroBfDelta}{\ensuremath{-}0.097}
\newcommand{\NTpRbSZeroBfP}{\ensuremath{6{\times}10^{-6}}}

\newcommand{\NTpRbSZeroBaseP}{0.29}
\newcommand{\NTpRbSZeroKEight}{0.040}
\newcommand{\NTpRbSZeroPTwo}{0.851}
\newcommand{\NTpRbSZeroTaskDelta}{+0.006}
\newcommand{\NTpRbSZeroTaskLo}{\ensuremath{-}0.009}
\newcommand{\NTpRbSZeroTaskHi}{+0.023}
\newcommand{\NTpRbSOneMed}{0.394}
\newcommand{\NTpRbSOneLowerSZero}{137}
\newcommand{\NTpRbSOneLowerSOne}{138}
\newcommand{\NTpRbSOneLowerSTwo}{137}
\newcommand{\NTpRbSOneBfDelta}{\ensuremath{-}0.100}
\newcommand{\NTpRbSOneBfP}{\ensuremath{3{\times}10^{-5}}}

\newcommand{\NTpRbSOneBaseP}{0.68}
\newcommand{\NTpRbSOneKEight}{0.169}
\newcommand{\NTpRbSOnePTwo}{0.866}
\newcommand{\NTpRbSOneTaskDelta}{+0.007}
\newcommand{\NTpRbSOneTaskLo}{\ensuremath{-}0.009}
\newcommand{\NTpRbSOneTaskHi}{+0.023}
\newcommand{\NTpTuASZeroMed}{0.464}
\newcommand{\NTpTuASZeroLowerSZero}{137}
\newcommand{\NTpTuASZeroLowerSOne}{135}
\newcommand{\NTpTuASZeroLowerSTwo}{134}
\newcommand{\NTpTuASZeroBfDelta}{+0.017}
\newcommand{\NTpTuASZeroBfP}{0.68}
\newcommand{\NTpTuASZeroBaseDelta}{+0.104}

\newcommand{\NTpTuASZeroKEight}{\ensuremath{-}0.027}
\newcommand{\NTpTuASZeroPTwo}{0.876}
\newcommand{\NTpTuASZeroTaskDelta}{\ensuremath{-}0.004}
\newcommand{\NTpTuASZeroTaskLo}{\ensuremath{-}0.019}
\newcommand{\NTpTuASZeroTaskHi}{+0.013}
\newcommand{\NTpTuASOneMed}{0.460}
\newcommand{\NTpTuASOneLowerSZero}{136}
\newcommand{\NTpTuASOneLowerSOne}{136}
\newcommand{\NTpTuASOneLowerSTwo}{136}
\newcommand{\NTpTuASOneBfDelta}{+0.018}
\newcommand{\NTpTuASOneBfP}{0.57}

\newcommand{\NTpTuASOneKEight}{\ensuremath{-}0.039}
\newcommand{\NTpTuASOnePTwo}{0.869}
\newcommand{\NTpTuASOneTaskDelta}{\ensuremath{-}0.005}
\newcommand{\NTpTuASOneTaskLo}{\ensuremath{-}0.019}
\newcommand{\NTpTuASOneTaskHi}{+0.012}
\newcommand{\NTpTuFMed}{0.463}
\newcommand{\NTpTuFLowerSZero}{136}
\newcommand{\NTpTuFLowerSOne}{134}
\newcommand{\NTpTuFLowerSTwo}{136}
\newcommand{\NTpTuFBfDelta}{+0.014}
\newcommand{\NTpTuFBfP}{0.68}

\newcommand{\NTpTuFKEight}{\ensuremath{-}0.019}
\newcommand{\NTpTuFPTwo}{0.880}
\newcommand{\NTpTuFTaskDelta}{\ensuremath{-}0.011}
\newcommand{\NTpTuFTaskLo}{\ensuremath{-}0.026}
\newcommand{\NTpTuFTaskHi}{+0.004}
\newcommand{\NTpTrZeroAMed}{0.415}
\newcommand{\NTpTrZeroALowerSZero}{140}
\newcommand{\NTpTrZeroALowerSOne}{138}
\newcommand{\NTpTrZeroALowerSTwo}{139}
\newcommand{\NTpTrZeroABfDelta}{\ensuremath{-}0.027}
\newcommand{\NTpTrZeroABfP}{0.22}
\newcommand{\NTpTrZeroABaseDelta}{+0.006}

\newcommand{\NTpTrZeroAKEight}{\ensuremath{-}0.028}
\newcommand{\NTpTrZeroAPTwo}{0.893}
\newcommand{\NTpTrZeroATaskDelta}{\ensuremath{-}0.010}
\newcommand{\NTpTrZeroATaskLo}{\ensuremath{-}0.025}
\newcommand{\NTpTrZeroATaskHi}{+0.006}
\newcommand{\NTpTrZeroFMed}{0.407}
\newcommand{\NTpTrZeroFLowerSZero}{142}
\newcommand{\NTpTrZeroFLowerSOne}{142}
\newcommand{\NTpTrZeroFLowerSTwo}{142}
\newcommand{\NTpTrZeroFBfDelta}{\ensuremath{-}0.038}
\newcommand{\NTpTrZeroFBfP}{0.09}

\newcommand{\NTpTrZeroFKEight}{\ensuremath{-}0.020}
\newcommand{\NTpTrZeroFPTwo}{0.899}
\newcommand{\NTpTrZeroFTaskDelta}{\ensuremath{-}0.010}
\newcommand{\NTpTrZeroFTaskLo}{\ensuremath{-}0.024}
\newcommand{\NTpTrZeroFTaskHi}{+0.002}
\newcommand{\NTpTrOneAMed}{0.414}
\newcommand{\NTpTrOneALowerSZero}{142}
\newcommand{\NTpTrOneALowerSOne}{140}
\newcommand{\NTpTrOneALowerSTwo}{134}
\newcommand{\NTpTrOneABfDelta}{\ensuremath{-}0.044}
\newcommand{\NTpTrOneABfP}{0.22}

\newcommand{\NTpTrOneAKEight}{\ensuremath{-}0.020}
\newcommand{\NTpTrOneAPTwo}{0.881}
\newcommand{\NTpTrOneATaskDelta}{\ensuremath{-}0.007}
\newcommand{\NTpTrOneATaskLo}{\ensuremath{-}0.021}
\newcommand{\NTpTrOneATaskHi}{+0.006}
\newcommand{\NTpTrOneFMed}{0.441}
\newcommand{\NTpTrOneFLowerSZero}{139}
\newcommand{\NTpTrOneFLowerSOne}{138}
\newcommand{\NTpTrOneFLowerSTwo}{135}
\newcommand{\NTpTrOneFBfDelta}{\ensuremath{-}0.030}
\newcommand{\NTpTrOneFBfP}{0.29}

\newcommand{\NTpTrOneFKEight}{\ensuremath{-}0.017}
\newcommand{\NTpTrOneFPTwo}{0.871}
\newcommand{\NTpTrOneFTaskDelta}{+0.001}
\newcommand{\NTpTrOneFTaskLo}{\ensuremath{-}0.016}
\newcommand{\NTpTrOneFTaskHi}{+0.018}
\newcommand{\NTpPcRMed}{0.778}
\newcommand{\NTpPcRLowerSZero}{97}
\newcommand{\NTpPcRLowerSOne}{87}
\newcommand{\NTpPcRLowerSTwo}{78}
\newcommand{\NTpPcRKEight}{0.206}
\newcommand{\NTpPcRPTwo}{0.863}
\newcommand{\NTpPcTMed}{0.793}
\newcommand{\NTpPcTLowerSZero}{83}
\newcommand{\NTpPcTLowerSOne}{77}
\newcommand{\NTpPcTLowerSTwo}{72}
\newcommand{\NTpPcTKEight}{0.209}
\newcommand{\NTpPcTPTwo}{0.870}
\newcommand{\NTpPcTTaskDelta}{+0.008}
\newcommand{\NTpPcTTaskLo}{\ensuremath{-}0.004}
\newcommand{\NTpPcTTaskHi}{+0.020}
\newcommand{\NTpRbSZeroLowerPcR}{139}
\newcommand{\NTpRbSZeroLowerPcT}{141}
\newcommand{\NTpRbSOneLowerPcR}{135}
\newcommand{\NTpRbSOneLowerPcT}{136}
\newcommand{\NTpTuASZeroLowerPcR}{134}
\newcommand{\NTpTuASZeroLowerPcT}{135}
\newcommand{\NTpTuASOneLowerPcR}{135}
\newcommand{\NTpTuASOneLowerPcT}{135}
\newcommand{\NTpTuFLowerPcR}{134}
\newcommand{\NTpTuFLowerPcT}{135}
\newcommand{\NTpTrZeroALowerPcR}{137}
\newcommand{\NTpTrZeroALowerPcT}{141}
\newcommand{\NTpTrZeroFLowerPcR}{139}
\newcommand{\NTpTrZeroFLowerPcT}{144}
\newcommand{\NTpTrOneALowerPcR}{138}
\newcommand{\NTpTrOneALowerPcT}{140}
\newcommand{\NTpTrOneFLowerPcR}{138}
\newcommand{\NTpTrOneFLowerPcT}{139}
\newcommand{\NTpCtlMin}{0.407}
\newcommand{\NTpCtlMax}{0.464}
\newcommand{\NTpCtlMinLower}{134}
\newcommand{\NTpRbMinLower}{135}
\newcommand{\NTpRbSZeroVsRandHeadsDelta}{\ensuremath{-}0.221}
\newcommand{\NTpRbSZeroVsRandHeadsLower}{114}
\newcommand{\NTpRbSOneVsRandHeadsDelta}{\ensuremath{-}0.184}
\newcommand{\NTpRbSOneVsRandHeadsLower}{117}
\newcommand{\NTpRbSeedsP}{0.81}
\newcommand{\NTpRegionRanked}{0.457}
\newcommand{\NTpRegionUni}{0.109}
\newcommand{\NTpRegionRandZero}{0.030}
\newcommand{\NTpRegionRandOne}{0.080}
\newcommand{\NTpWriteRanked}{1.611}
\newcommand{\NTpWriteCtlMin}{0.003}
\newcommand{\NTpWriteCtlMax}{0.016}
\newcommand{\NTpKlUni}{0.528}
\newcommand{\NTpKlRandZero}{1.417}
\newcommand{\NTpKlRandOne}{0.663}
\newcommand{\NTpScopeMaxDiff}{0.012}
\newcommand{\NTpScopeCorrMin}{0.93}
\newcommand{\NTpScopeCorrMax}{0.97}
\newcommand{\NTpUniOverRandMin}{+0.032}
\newcommand{\NTpUniOverRandMax}{+0.053}
\newcommand{\NTpFitMistillStart}{0.986}
\newcommand{\NTpFitMistillEnd}{0.244}
\newcommand{\NTpFitUniStart}{1.021}
\newcommand{\NTpFitUniEnd}{0.478}
\newcommand{\NTpFitRandZeroStart}{1.484}
\newcommand{\NTpFitRandZeroEnd}{0.712}
\newcommand{\NTpFitRandOneStart}{1.040}
\newcommand{\NTpFitRandOneEnd}{0.497}
\newcommand{\NTpTaskDeltaMin}{\ensuremath{-}0.011}
\newcommand{\NTpTaskDeltaMax}{+0.001}
\newcommand{\NTpNegShareMin}{56}
\newcommand{\NTpNegShareMax}{67}
\newcommand{\NTpNegShareMistill}{3}
\newcommand{\NTpPcLowerMin}{134}
\newcommand{\NTpPcLowerMax}{144}
\newcommand{\NTpPcRVsSZeroDelta}{\ensuremath{-}0.016}
\newcommand{\NTpPcTVsSZeroDelta}{\ensuremath{-}0.005}

\newcommand{\NTpPcCorr}{0.980}
\newcommand{\NTpPcTrainMaxRel}{0.15}
\newcommand{\NTpPcTMicroF}{0.056}

\newcommand{\NMTeacherMicroFFullInventory}{0.168}
\newcommand{\NMAttrPearsonAllDetached}{0.36}
\newcommand{\NMAttrPearsonAboveNoiseDetached}{0.47}
\newcommand{\NMAttrPearsonAllIntact}{0.84}
\newcommand{\NMAttrPearsonAboveNoiseIntact}{0.85}
\newcommand{\NMNoiseFloorNats}{0.046}
\newcommand{\NMKnockinCeilingDiff}{0.000}
\newcommand{\NMNullRawBhPos}{309}
\newcommand{\NMNullRawBhNeg}{282}
\newcommand{\NMNullSplitBhDiscoveries}{0}

\newcommand{\NMKiOwnGapTopForty}{0.04}
\newcommand{\NMLThirtyHElevenSiblingOverOwnNotes}{91}
\newcommand{\NMLocusEmissionSufKSixteen}{1.01}
\newcommand{\NMLocusProseSufKSixteen}{0.48}

\newcommand{\NMRetrieverRecallVal}{0.903}
\newcommand{\NMRetrieverRecallTest}{0.936}
\newcommand{\NMTeacherShortlistTestMicroF}{0.227}

\newcommand{\NMPoolQwenMOnePTwoCVsVanillaBfHigher}{348}
\newcommand{\NMPoolQwenMOnePTwoCVsVanillaBfN}{441}
\newcommand{\NMQwenBasePctTeacher}{93\%}
\newcommand{\NMQwenMistillSeedSpreadPTwoC}{0.004}
\newcommand{\NMCarrierOverlapRange}{27--38}

\newcommand{\NMCZeroShotEncoderRecallAtFiveHundredVal}{0.45}
\newcommand{\NMCRetrieverRecallKThousandVal}{0.89}
\newcommand{\NMCRetrieverRecallKThousandCorpus}{0.92}
\newcommand{\NMCRetrieverRecallKFiveHundredVal}{0.80}
\newcommand{\NMCRetrieverRecallKFiveHundredCorpus}{0.84}
\newcommand{\NMCTeacherMicroFShortlistVal}{0.199}

\newcommand{\NMCGoldChaptersPerNoteMedian}{8}
\newcommand{\NMCAnchorCentroidNotesPassing}{6}
\newcommand{\NMCAnchorPerConceptNotesPassing}{19}
\newcommand{\NMCAnchorPerConceptShiftedHeads}{209}
\newcommand{\NMCAnchorNullNonemptyTrials}{800}
\newcommand{\NMCAnchorNullSignConsistency}{81.5\%}
\newcommand{\NMCAnchorPositiveControlNotesTwoHundredDraws}{44}
\newcommand{\NMCAnchorPositiveControlNotesTwentyDraws}{73}
\newcommand{\NMCLTwentySevenHSeventeenRecurrenceNotes}{42}
\newcommand{\NMCLTwentySevenHSeventeenRankPOne}{128}
\newcommand{\NMCLElevenHTenRankPOne}{364}
\newcommand{\NMCPerNoteSplitHalfJaccard}{0.000}
\newcommand{\NMCPerNoteBHPassNotesLTwentySevenHSeventeen}{2}

\newcommand{\NMCFirstMentionsSiblingPct}{30\%}
\newcommand{\NMCPOneMarginPTen}{+0.74}
\newcommand{\NMCPOneMarginPNinety}{+10.85}
\newcommand{\NMCPOneEntropyMedian}{0.02}
\newcommand{\NMCPOneMarginBelowFivePct}{42\%}
\newcommand{\NMCPOneMarginNegativePct}{6\%}
\newcommand{\NMCPTwoMarginPTen}{\ensuremath{-}0.49}
\newcommand{\NMCPTwoMarginPNinety}{+9.19}
\newcommand{\NMCPTwoEntropyMedian}{0.26}
\newcommand{\NMCPTwoMarginBelowFivePct}{68\%}
\newcommand{\NMCPTwoMarginNegativePct}{15\%}
\newcommand{\NMCPTwoCMarginPTen}{0.00}
\newcommand{\NMCPTwoCMarginPNinety}{+8.00}
\newcommand{\NMCPTwoCEntropyMedian}{0.15}
\newcommand{\NMCPTwoCMarginBelowFivePct}{66\%}
\newcommand{\NMCPTwoCMarginNegativePct}{10\%}
\newcommand{\NMCNullMarginPTen}{\ensuremath{-}10.44}
\newcommand{\NMCNullMarginPNinety}{\ensuremath{-}2.68}
\newcommand{\NMCNullEntropyMedian}{1.80}
\newcommand{\NMCNullMarginBelowFivePct}{100\%}
\newcommand{\NMCNullMarginNegativePct}{99\%}
\newcommand{\NMCAttrTopTwentyRecoveredDetached}{48\%}
\newcommand{\NMCAttrSpearmanIntact}{0.51}
\newcommand{\NMCAttrTopTwentyOverlapIntact}{16}
\newcommand{\NMCAttrTopTwentyRecoveredIntact}{92\%}
\newcommand{\NMCIGPearsonIntact}{0.63}
\newcommand{\NMCDLAVsFirstOrderPearsonMedian}{\ensuremath{-}0.32}
\newcommand{\NMCPTwoCFirstOrderExactPearsonMedian}{0.72}
\newcommand{\NMCIGPearsonAllDetached}{0.22}
\newcommand{\NMCDLAPearsonSelftestMin}{\ensuremath{-}0.11}
\newcommand{\NMCDLAPearsonSelftestMax}{+0.07}
\newcommand{\NMCIGPearsonAboveNoiseDetached}{0.25}
\newcommand{\NMCIGTopTwentyRecoveredDetached}{29\%}
\newcommand{\NMCNoiseFloorRange}{0.09}
\newcommand{\NMCNoiseFloorThreeSd}{0.14}
\newcommand{\NMCHeadsAboveFloorPerEventMin}{71}
\newcommand{\NMCHeadsAboveFloorPerEventMax}{397}
\newcommand{\NMCDonorTokenMatchPOne}{55\%}
\newcommand{\NMCDonorTokenMatchPTwo}{75\%}
\newcommand{\NMCDonorTokenMatchPTwoC}{95\%}
\newcommand{\NMCDonorDepthMatchPOne}{100\%}
\newcommand{\NMCDonorDepthMatchPTwo}{87\%}
\newcommand{\NMCDonorDepthMatchPTwoC}{85\%}
\newcommand{\NMCKnockinCeilingPTen}{0.00}
\newcommand{\NMCKnockinCeilingPNinety}{0.00}
\newcommand{\NMCKiExcludedPOne}{0.5\%}
\newcommand{\NMCKiExcludedPTwo}{9\%}
\newcommand{\NMCKiExcludedPTwoC}{29\%}
\newcommand{\NMCCorruptedNegativePctPOne}{99\%}
\newcommand{\NMCCorruptedNegativePctPTwo}{87\%}
\newcommand{\NMCCorruptedNegativePctPTwoC}{54\%}
\newcommand{\NMCKiRangePTwoCMedian}{3.1}
\newcommand{\NMCNullTopSixteenMeanEffect}{0.16}
\newcommand{\NMCPOneTopSixteenMeanEffect}{0.26}
\newcommand{\NMCRawNullSpearman}{\ensuremath{-}0.24}
\newcommand{\NMCRawNullOverlapTopSixteen}{0}
\newcommand{\NMCRawNullOverlapTopForty}{1}
\newcommand{\NMCPOneBHNeg}{546}
\newcommand{\NMCBootstrapPvalueSyntheticDiscoveries}{13}

\newcommand{\NMCIcPOneLThirtyNineHThirteen}{0.47}
\newcommand{\NMCRankPOneLThirtyNineHThirteen}{2}
\newcommand{\NMCIcPTwoLThirtyNineHThirteen}{0.12}
\newcommand{\NMCRankPTwoLThirtyNineHThirteen}{17}
\newcommand{\NMCIcPTwoCLThirtyNineHThirteen}{0.02}
\newcommand{\NMCRankPTwoCLThirtyNineHThirteen}{68}
\newcommand{\NMCIcPOneLThirtyEightHTwentyFour}{0.39}
\newcommand{\NMCRankPOneLThirtyEightHTwentyFour}{4}
\newcommand{\NMCIcPTwoLThirtyEightHTwentyFour}{0.27}
\newcommand{\NMCRankPTwoLThirtyEightHTwentyFour}{7}
\newcommand{\NMCIcPTwoCLThirtyEightHTwentyFour}{0.27}
\newcommand{\NMCRankPTwoCLThirtyEightHTwentyFour}{5}
\newcommand{\NMCIcPOneLTwentyEightHTwentyOne}{0.36}
\newcommand{\NMCRankPOneLTwentyEightHTwentyOne}{5}
\newcommand{\NMCIcPTwoLTwentyEightHTwentyOne}{0.27}
\newcommand{\NMCRankPTwoLTwentyEightHTwentyOne}{6}
\newcommand{\NMCIcPTwoCLTwentyEightHTwentyOne}{0.11}
\newcommand{\NMCRankPTwoCLTwentyEightHTwentyOne}{14}
\newcommand{\NMCIcPOneLThirtyFiveHEighteen}{0.21}
\newcommand{\NMCRankPOneLThirtyFiveHEighteen}{9}
\newcommand{\NMCIcPTwoLThirtyFiveHEighteen}{0.13}
\newcommand{\NMCRankPTwoLThirtyFiveHEighteen}{16}
\newcommand{\NMCIcPTwoCLThirtyFiveHEighteen}{0.14}
\newcommand{\NMCRankPTwoCLThirtyFiveHEighteen}{11}
\newcommand{\NMCIcPOneLThirtyNineHTwentyNine}{0.19}
\newcommand{\NMCRankPOneLThirtyNineHTwentyNine}{10}
\newcommand{\NMCIcPTwoLThirtyNineHTwentyNine}{0.31}
\newcommand{\NMCRankPTwoLThirtyNineHTwentyNine}{3}
\newcommand{\NMCIcPTwoCLThirtyNineHTwentyNine}{0.31}
\newcommand{\NMCRankPTwoCLThirtyNineHTwentyNine}{3}
\newcommand{\NMCIcPOneLThirtySevenHSix}{0.19}
\newcommand{\NMCRankPOneLThirtySevenHSix}{11}
\newcommand{\NMCIcPTwoLThirtySevenHSix}{0.28}
\newcommand{\NMCRankPTwoLThirtySevenHSix}{5}
\newcommand{\NMCIcPTwoCLThirtySevenHSix}{0.28}
\newcommand{\NMCRankPTwoCLThirtySevenHSix}{4}
\newcommand{\NMCIcPOneLTwentySixHEighteen}{0.18}
\newcommand{\NMCRankPOneLTwentySixHEighteen}{12}
\newcommand{\NMCIcPTwoLTwentySixHEighteen}{0.06}
\newcommand{\NMCRankPTwoLTwentySixHEighteen}{47}
\newcommand{\NMCIcPTwoCLTwentySixHEighteen}{0.02}
\newcommand{\NMCRankPTwoCLTwentySixHEighteen}{74}
\newcommand{\NMCPTwoCExactFirstOrderOverlapTopForty}{33}
\newcommand{\NMCPTwoCExactBHPos}{313}
\newcommand{\NMCPTwoCExactCiLoLThirtyNineHTwelve}{0.54}
\newcommand{\NMCPTwoCExactCiHiLThirtyNineHTwelve}{0.75}
\newcommand{\NMCPTwoCFirstOrderILThirtyNineHTwelve}{0.19}
\newcommand{\NMCPTwoCExactCiLoLThirtyNineHTen}{0.29}
\newcommand{\NMCPTwoCExactCiHiLThirtyNineHTen}{0.42}
\newcommand{\NMCPTwoCFirstOrderILThirtyNineHTen}{0.23}
\newcommand{\NMCPTwoCExactCiLoLThirtyHEleven}{0.25}
\newcommand{\NMCPTwoCExactCiHiLThirtyHEleven}{0.32}
\newcommand{\NMCPTwoCFirstOrderILThirtyHEleven}{0.53}
\newcommand{\NMCPTwoCExactCiLoLThirtyNineHThirtyEight}{0.20}
\newcommand{\NMCPTwoCExactCiHiLThirtyNineHThirtyEight}{0.25}
\newcommand{\NMCPTwoCFirstOrderILThirtyNineHThirtyEight}{0.25}
\newcommand{\NMCPTwoCExactLThirtyNineHTwentyNine}{0.16}
\newcommand{\NMCPTwoCExactCiLoLThirtyNineHTwentyNine}{0.14}
\newcommand{\NMCPTwoCExactCiHiLThirtyNineHTwentyNine}{0.19}
\newcommand{\NMCPTwoCFirstOrderILThirtyNineHTwentyNine}{0.16}
\newcommand{\NMCPTwoCExactLThirtyNineHTwentyFive}{0.14}
\newcommand{\NMCPTwoCExactCiLoLThirtyNineHTwentyFive}{0.13}
\newcommand{\NMCPTwoCExactCiHiLThirtyNineHTwentyFive}{0.16}
\newcommand{\NMCPTwoCFirstOrderILThirtyNineHTwentyFive}{0.14}
\newcommand{\NMCRankPTwoCLThirtyNineHTwentyFive}{10}
\newcommand{\NMCPTwoCExactCiLoLThirtyEightHThirtyNine}{0.11}
\newcommand{\NMCPTwoCExactCiHiLThirtyEightHThirtyNine}{0.14}
\newcommand{\NMCPTwoCFirstOrderILThirtyEightHThirtyNine}{0.21}
\newcommand{\NMCPTwoCExactLTwentyEightHTwentyOne}{0.05}
\newcommand{\NMCPTwoCExactCiLoLTwentyEightHTwentyOne}{0.04}
\newcommand{\NMCPTwoCExactCiHiLTwentyEightHTwentyOne}{0.06}
\newcommand{\NMCPTwoCFirstOrderILTwentyEightHTwentyOne}{0.06}
\newcommand{\NMCWeightLeaderKThreePOne}{0.81}
\newcommand{\NMCWeightLeaderKThreePTwoC}{0.66}
\newcommand{\NMCWeightLThirtyNineHThirtyEightTopFortyPOne}{0.742}
\newcommand{\NMCWeightLThirtyNineHThirtyEightTopFortyPTwo}{0.813}
\newcommand{\NMCWeightLThirtyNineHThirtyEightTopFortyPTwoC}{0.444}
\newcommand{\NMCShortWeightLeaderTopFortyPOne}{0.254}
\newcommand{\NMCShortWeightLeaderTopFortyPTwo}{0.303}
\newcommand{\NMCShortWeightLeaderTopFortyPTwoC}{0.215}
\newcommand{\NMCPOneNecGlobalGeOwnNotesMin}{36}
\newcommand{\NMCPOneNecGlobalGeOwnNotesMax}{52}
\newcommand{\NMCPOneNecGlobalGeOwnNotesKOneHundred}{72}
\newcommand{\NMCPOneNecTopEightPTen}{0.30}
\newcommand{\NMCPOneNecTopEightPNinety}{1.79}
\newcommand{\NMCPOneKiTopEightPTen}{0.56}
\newcommand{\NMCPOneKiTopEightPNinety}{0.74}
\newcommand{\NMCPOneKiTopEightAboveHalfNotes}{99}
\newcommand{\NMCIsolationMarginRetained}{\ensuremath{-}0.6}
\newcommand{\NMCAttrPassOwnNecKThree}{0.29}
\newcommand{\NMCAttrPassOwnNecKEight}{0.62}
\newcommand{\NMCAttrPassOwnNecKSixteen}{0.88}
\newcommand{\NMCAttrPassOwnNecKForty}{1.22}
\newcommand{\NMCAttrPassOwnNecKOneHundred}{1.42}
\newcommand{\NMCAttrPassOwnNecWinsMin}{96}
\newcommand{\NMCAttrPassOwnNecWinsMax}{100}
\newcommand{\NMCRecurrentHeadsCount}{41}
\newcommand{\NMCRecurrenceLThirtyNineHThirtyEight}{100}
\newcommand{\NMCRecurrenceLThirtyNineHThirteen}{100}
\newcommand{\NMCRecurrenceLTwentySixHThirtySix}{95}
\newcommand{\NMCRecurrenceLTwentySixHThirtySeven}{94}
\newcommand{\NMCRecurrenceLTwentyEightHTwentyOne}{84}
\newcommand{\NMCRecurrenceLTwentySixHEighteen}{78}
\newcommand{\NMCRecurrenceLTwentySevenHThirteen}{75}
\newcommand{\NMCRecurrenceLTwentySevenHEleven}{74}
\newcommand{\NMCRecurrenceLTwentyFourHOne}{71}
\newcommand{\NMCRecurrenceLThirtyNineHTen}{68}
\newcommand{\NMCNullFloorNats}{0.004}
\newcommand{\NMCPOneHeadsAboveFloor}{289}
\newcommand{\NMCPOneHeadsForEightyPct}{92}
\newcommand{\NMCPerNoteHeadsForEightyPctMedian}{64}

\newcommand{\NMCPOneTopSixteenSpearmanLiftIc}{0.00}
\newcommand{\NMCPOneTopSixteenLiftPTen}{2.7}
\newcommand{\NMCPOneTopSixteenLiftPNinety}{4.9}
\newcommand{\NMCPOneTopSixteenOnSink}{0.10}
\newcommand{\NMCPOneTopSixteenOnInventoryLines}{0.28}
\newcommand{\NMCPOneTopSixteenOnNote}{0.07}
\newcommand{\NMCPOneTopSixteenOnProse}{0.38}
\newcommand{\NMCPTwoTopSixteenOnSink}{0.11}
\newcommand{\NMCPTwoTopSixteenOnInventoryLines}{0.37}
\newcommand{\NMCPTwoTopSixteenOnNote}{0.02}
\newcommand{\NMCPTwoTopSixteenOnProse}{0.16}
\newcommand{\NMCPTwoTopSixteenOnPrefix}{0.251}
\newcommand{\NMCPTwoRandomSixteenOnPrefix}{0.135}
\newcommand{\NMCPTwoCTopSixteenOnSink}{0.07}
\newcommand{\NMCPTwoCTopSixteenOnInventoryLines}{0.32}
\newcommand{\NMCPTwoCTopSixteenOnNote}{0.02}
\newcommand{\NMCPTwoCTopSixteenOnProse}{0.14}
\newcommand{\NMCPTwoCTopSixteenOnPrefix}{0.254}
\newcommand{\NMCPTwoCRandomSixteenOnPrefix}{0.151}
\newcommand{\NMCPOneRandomSixteenOnRegion}{0.03}
\newcommand{\NMCLTwentySixHEighteenRPOne}{+0.22}
\newcommand{\NMCLTwentySixHThirtySixOnSinkPOne}{0.00}
\newcommand{\NMCLTwentySixHThirtySixNeffPOne}{23}
\newcommand{\NMCLTwentyEightHTwentyOneOnSinkPOne}{0.00}
\newcommand{\NMCLTwentyEightHTwentyOneNeffPOne}{44}
\newcommand{\NMCLTwentySixHThirtySevenOnSinkPOne}{0.00}
\newcommand{\NMCLTwentySixHThirtySevenNeffPOne}{56}
\newcommand{\NMCLTwentySixHThirtySevenOnNotePOne}{0.23}
\newcommand{\NMCLTwentySixHEighteenOnRegionPOne}{0.06}
\newcommand{\NMCLTwentySixHEighteenOnProsePOne}{0.48}
\newcommand{\NMCLTwentySixHEighteenOnSinkPOne}{0.00}
\newcommand{\NMCLTwentySixHEighteenLiftPOne}{6.1}
\newcommand{\NMCLTwentySixHEighteenNeffPOne}{60}
\newcommand{\NMCLTwentySixHEighteenOnNotePOne}{0.25}
\newcommand{\NMCIcPOneLTwentyFourHOne}{0.17}
\newcommand{\NMCLTwentyFourHOneOnRegionPOne}{0.16}
\newcommand{\NMCLTwentyFourHOneOnProsePOne}{0.30}
\newcommand{\NMCLTwentyFourHOneOnSinkPOne}{0.00}
\newcommand{\NMCLTwentyFourHOneLiftPOne}{5.0}
\newcommand{\NMCLTwentyFourHOneNeffPOne}{38}
\newcommand{\NMCLTwentyFourHOneRPOne}{+0.11}
\newcommand{\NMCIcPOneLTwentyFiveHTwentyEight}{0.17}
\newcommand{\NMCLTwentyFiveHTwentyEightOnRegionPOne}{0.09}
\newcommand{\NMCLTwentyFiveHTwentyEightOnProsePOne}{0.29}
\newcommand{\NMCLTwentyFiveHTwentyEightOnSinkPOne}{0.05}
\newcommand{\NMCLTwentyFiveHTwentyEightLiftPOne}{3.2}
\newcommand{\NMCLTwentyFiveHTwentyEightNeffPOne}{70}
\newcommand{\NMCLTwentyFiveHTwentyEightRPOne}{+0.32}
\newcommand{\NMCLThirtyNineHThirtyEightOnSinkPOne}{0.00}
\newcommand{\NMCLThirtyNineHThirtyEightNeffPOne}{727}
\newcommand{\NMCLThirtyNineHThirteenOnSinkPOne}{0.17}
\newcommand{\NMCLThirtyNineHThirteenNeffPOne}{1{,}949}
\newcommand{\NMCLThirtyEightHTwentyFourOnSinkPOne}{0.02}
\newcommand{\NMCLThirtyEightHTwentyFourNeffPOne}{32}
\newcommand{\NMCLTwentySevenHElevenOnSinkPOne}{0.00}
\newcommand{\NMCLTwentySevenHElevenNeffPOne}{54}
\newcommand{\NMCLTwentySevenHThirteenOnRegionPOne}{0.02}
\newcommand{\NMCLTwentySevenHThirteenOnProsePOne}{0.52}
\newcommand{\NMCLTwentySevenHThirteenOnSinkPOne}{0.03}
\newcommand{\NMCLTwentySevenHThirteenLiftPOne}{0.5}
\newcommand{\NMCLTwentySevenHThirteenNeffPOne}{24}
\newcommand{\NMCLTwentySevenHThirteenRPOne}{\ensuremath{-}0.08}
\newcommand{\NMCLTwentySevenHThirteenOnOtherChaptersPOne}{0.32}
\newcommand{\NMCLThirtyNineHTwentyNineOnRegionPOne}{0.01}
\newcommand{\NMCLThirtyNineHTwentyNineOnProsePOne}{0.52}
\newcommand{\NMCLThirtyNineHTwentyNineOnSinkPOne}{0.35}
\newcommand{\NMCLThirtyNineHTwentyNineLiftPOne}{0.9}
\newcommand{\NMCLThirtyNineHTwentyNineNeffPOne}{355}
\newcommand{\NMCLThirtyNineHTwentyNineRPOne}{\ensuremath{-}0.03}
\newcommand{\NMCLThirtyFiveHEighteenOnRegionPOne}{0.08}
\newcommand{\NMCLThirtyFiveHEighteenOnProsePOne}{0.05}
\newcommand{\NMCLThirtyFiveHEighteenOnSinkPOne}{0.17}
\newcommand{\NMCLThirtyFiveHEighteenLiftPOne}{2.2}
\newcommand{\NMCLThirtyFiveHEighteenNeffPOne}{33}
\newcommand{\NMCLThirtyFiveHEighteenRPOne}{+0.11}
\newcommand{\NMCLThirtyFiveHEighteenOnOtherChaptersPOne}{0.56}
\newcommand{\NMCLThirtySevenHSixOnRegionPOne}{0.07}
\newcommand{\NMCLThirtySevenHSixOnProsePOne}{0.05}
\newcommand{\NMCLThirtySevenHSixOnSinkPOne}{0.14}
\newcommand{\NMCLThirtySevenHSixLiftPOne}{1.9}
\newcommand{\NMCLThirtySevenHSixNeffPOne}{107}
\newcommand{\NMCLThirtySevenHSixRPOne}{+0.05}
\newcommand{\NMCLThirtyNineHTenOnRegionPOne}{0.01}
\newcommand{\NMCLThirtyNineHTenOnProsePOne}{0.20}
\newcommand{\NMCLThirtyNineHTenOnSinkPOne}{0.60}
\newcommand{\NMCLThirtyNineHTenLiftPOne}{1.1}
\newcommand{\NMCLThirtyNineHTenNeffPOne}{1{,}468}
\newcommand{\NMCLThirtyNineHTenRPOne}{\ensuremath{-}0.09}
\newcommand{\NMCIcPOneLTwentySixHThirtyThree}{0.17}
\newcommand{\NMCLTwentySixHThirtyThreeOnRegionPOne}{0.01}
\newcommand{\NMCLTwentySixHThirtyThreeOnProsePOne}{0.12}
\newcommand{\NMCLTwentySixHThirtyThreeOnSinkPOne}{0.00}
\newcommand{\NMCLTwentySixHThirtyThreeLiftPOne}{1.6}
\newcommand{\NMCLTwentySixHThirtyThreeNeffPOne}{227}
\newcommand{\NMCLTwentySixHThirtyThreeRPOne}{\ensuremath{-}0.09}
\newcommand{\NMCLThirtyNineHTwelveOnPrefixPTwo}{0.28}
\newcommand{\NMCLThirtyNineHTwentyFiveOnPrefixPTwo}{0.74}
\newcommand{\NMCLThirtyEightHTwentyTwoOnCategoryPTwo}{0.10}
\newcommand{\NMCLTwentyEightHTwentyOneOnCategoryPTwo}{0.24}
\newcommand{\NMCLThirtyEightHTwentyTwoLiftPTwo}{9}
\newcommand{\NMCLThirtyEightHThirtyFiveLiftPTwo}{14}
\newcommand{\NMCLTwentyEightHTwentyOneLiftPTwo}{10}
\newcommand{\NMCLThirtyNineHTenOnPrefixPTwo}{0.61}
\newcommand{\NMCLThirtyEightHThirtyFiveOnCategoryPTwo}{0.20}
\newcommand{\NMCLThirtyNineHTwentyNineOnPrefixPTwo}{0.60}
\newcommand{\NMCLThirtyHThirteenOnCategoryPTwo}{0.17}
\newcommand{\NMCLThirtyHThirteenLiftPTwo}{13}
\newcommand{\NMCLThirtySixHTenOnPrefixPTwo}{0.55}
\newcommand{\NMCLThirtyNineHThirtySixOnPrefixPTwo}{0.39}
\newcommand{\NMCLThirtyNineHThirtySevenOnPrefixPTwo}{0.35}
\newcommand{\NMCLThirtyHElevenOnSiblingPTen}{0.20}
\newcommand{\NMCLThirtyHElevenOnSiblingPNinety}{0.45}
\newcommand{\NMCLThirtyHElevenAboveRandomNotes}{92}
\newcommand{\NMCLThirtyEightHThirtyNineSiblingOverOwnNotes}{92}
\newcommand{\NMCQkSMinusNMax}{0.02}
\newcommand{\NMCLThirtyEightHTwentyTwoSiblingOverOwnNotes}{72}
\newcommand{\NMCIcPTwoCLThirtyEightHThirtyFive}{0.12}
\newcommand{\NMCLThirtyEightHThirtyFiveOnSiblingPTwoC}{0.11}
\newcommand{\NMCLThirtyEightHThirtyFiveOnOwnPTwoC}{0.10}
\newcommand{\NMCLThirtyEightHThirtyFiveSiblingOverOwnNotes}{49}
\newcommand{\NMCPTwoCRandomSixteenOnOwn}{0.01}
\newcommand{\NMCPTwoCBottomSixteenOnSibling}{0.05}
\newcommand{\NMCPTwoCBottomSixteenOnOwn}{0.04}
\newcommand{\NMCRegionReadersLiftMinOverNotes}{4}

\newcommand{\NMCProbeAgreementLowest}{64\%}
\newcommand{\NMCProbeDAgreement}{73\%}
\newcommand{\NMCProbeBAgreement}{68\%}
\newcommand{\NMCLocusCensoredPct}{0.3\%}
\newcommand{\NMCLocusAtQPctGold}{2\%}
\newcommand{\NMCLocusAtQPctWrong}{2\%}
\newcommand{\NMCLocusAtQPctPrevSame}{1\%}
\newcommand{\NMCLocusAtQPctPrevOther}{4\%}
\newcommand{\NMCLocusAtQPctFirst}{1\%}
\newcommand{\NMCLocusWindowPctGold}{59\%}
\newcommand{\NMCLocusWindowPctWrong}{34\%}
\newcommand{\NMCLocusWindowPctPrevSame}{12\%}
\newcommand{\NMCLocusWindowPctPrevOther}{54\%}
\newcommand{\NMCLocusWindowPctFirst}{71\%}
\newcommand{\NMCLocusInheritedPctGold}{23\%}
\newcommand{\NMCLocusInheritedPctWrong}{30\%}
\newcommand{\NMCLocusInheritedPctPrevSame}{67\%}
\newcommand{\NMCLocusInheritedPctPrevOther}{0\%}
\newcommand{\NMCLocusInheritedPctFirst}{0\%}
\newcommand{\NMCLocusProbeFailPctGold}{15\%}
\newcommand{\NMCLocusProbeFailPctWrong}{34\%}
\newcommand{\NMCLocusProbeFailPctPrevSame}{20\%}
\newcommand{\NMCLocusProbeFailPctPrevOther}{42\%}
\newcommand{\NMCLocusProbeFailPctFirst}{27\%}
\newcommand{\NMCLocusDPTen}{1}
\newcommand{\NMCLocusDPNinety}{55}
\newcommand{\NMCLocusWindowSharePerNoteMedian}{37\%}
\newcommand{\NMCLocusWindowSharePerNotePTen}{21\%}
\newcommand{\NMCLocusWindowSharePerNotePNinety}{55\%}
\newcommand{\NMCLocusInheritedSharePerNoteMedian}{26\%}
\newcommand{\NMCLocusInheritedSharePerNotePTen}{0\%}
\newcommand{\NMCLocusInheritedSharePerNotePNinety}{42\%}
\newcommand{\NMCLocusFlickerPct}{21\%}
\newcommand{\NMCLocusThetaOneDecisiveAtQ}{77\%}
\newcommand{\NMCLocusThetaThreeDecisiveAtQ}{60\%}
\newcommand{\NMCLocusInheritedPctSibling}{63\%}
\newcommand{\NMCNoteAloneDecisivePct}{3\%}
\newcommand{\NMCNoteAloneDecisiveFirstCodesPct}{12\%}
\newcommand{\NMCNoteAloneDecisiveGoldFirstCodesPct}{13\%}
\newcommand{\NMCNoteAloneMedianMargin}{\ensuremath{-}3.6}
\newcommand{\NMCLocusClassOwnDescBase}{2.3\%}
\newcommand{\NMCLocusClassChapterDescBase}{5.4\%}
\newcommand{\NMCLocusClassChapterDescShare}{22.4\%}
\newcommand{\NMCLocusClassChapterDescLift}{4.2}
\newcommand{\NMCLocusClassProseBase}{48.0\%}
\newcommand{\NMCLocusClassProseShare}{34.7\%}
\newcommand{\NMCLocusClassProseLift}{0.7}
\newcommand{\NMCLocusClassOtherDescBase}{15.8\%}
\newcommand{\NMCLocusClassOtherDescShare}{4.8\%}
\newcommand{\NMCLocusClassOtherDescLift}{0.3}
\newcommand{\NMCLocusClassPunctBase}{13.9\%}
\newcommand{\NMCLocusClassPunctShare}{3.7\%}
\newcommand{\NMCLocusClassPunctLift}{0.3}
\newcommand{\NMCLocusClassCodeBase}{14.5\%}
\newcommand{\NMCLocusClassCodeShare}{2.4\%}
\newcommand{\NMCLocusClassCodeLift}{0.2}
\newcommand{\NMCImpWindowLocus}{8\%}
\newcommand{\NMCImpWindowOneToThree}{9\%}
\newcommand{\NMCImpWindowFourToFifteen}{13\%}
\newcommand{\NMCImpWindowSixteenToSixtyThree}{21\%}
\newcommand{\NMCImpWindowSuffix}{48\%}
\newcommand{\NMCImpAtQLocus}{9\%}
\newcommand{\NMCImpAtQOneToThree}{13\%}
\newcommand{\NMCImpAtQFourToFifteen}{21\%}
\newcommand{\NMCImpAtQSixteenToSixtyThree}{26\%}
\newcommand{\NMCImpAtQSuffix}{32\%}
\newcommand{\NMCImpNullLocus}{3\%}
\newcommand{\NMCImpNullOneToThree}{5\%}
\newcommand{\NMCImpNullFourToFifteen}{14\%}
\newcommand{\NMCImpNullSixteenToSixtyThree}{31\%}
\newcommand{\NMCImpNullSuffix}{47\%}
\newcommand{\NMCProseEmissionSpearman}{+0.12}
\newcommand{\NMCProseTopFortyLTwelveToTwentyThreeImportanceShare}{75\%}
\newcommand{\NMCProseRankLTwentySixHThirtySeven}{18}
\newcommand{\NMCProseRankLTwentySixHThirtySix}{29}
\newcommand{\NMCLTwentyFourHNineteenSuffixShare}{52\%}
\newcommand{\NMCLTwentyFourHOneSuffixShare}{55\%}
\newcommand{\NMCLTwentyTwoHSixteenSuffixShare}{51\%}
\newcommand{\NMCProseActingHeadsSuffixShareMax}{27\%}
\newcommand{\NMCProseIcLTwentyFourHNineteen}{0.265}
\newcommand{\NMCProseCiLoLTwentyFourHNineteen}{0.244}
\newcommand{\NMCProseCiHiLTwentyFourHNineteen}{0.288}
\newcommand{\NMCRankPOneLTwentyFourHNineteen}{36}
\newcommand{\NMCIcPOneLTwentyFourHNineteen}{0.10}
\newcommand{\NMCProseIcLTwentyTwoHSixteen}{0.249}
\newcommand{\NMCProseCiLoLTwentyTwoHSixteen}{0.221}
\newcommand{\NMCProseCiHiLTwentyTwoHSixteen}{0.276}
\newcommand{\NMCRankPOneLTwentyTwoHSixteen}{49}
\newcommand{\NMCIcPOneLTwentyTwoHSixteen}{0.07}
\newcommand{\NMCProseIcLTwentyFourHOne}{0.228}
\newcommand{\NMCProseCiLoLTwentyFourHOne}{0.206}
\newcommand{\NMCProseCiHiLTwentyFourHOne}{0.251}
\newcommand{\NMCRankPOneLTwentyFourHOne}{14}
\newcommand{\NMCProseIcLEighteenHFour}{0.184}
\newcommand{\NMCProseCiLoLEighteenHFour}{0.111}
\newcommand{\NMCProseCiHiLEighteenHFour}{0.251}
\newcommand{\NMCRankPOneLEighteenHFour}{127}
\newcommand{\NMCIcPOneLEighteenHFour}{0.02}
\newcommand{\NMCProseIcLTwentyTwoHFifteen}{0.177}
\newcommand{\NMCProseCiLoLTwentyTwoHFifteen}{0.141}
\newcommand{\NMCProseCiHiLTwentyTwoHFifteen}{0.213}
\newcommand{\NMCRankPOneLTwentyTwoHFifteen}{43}
\newcommand{\NMCIcPOneLTwentyTwoHFifteen}{0.08}
\newcommand{\NMCProseIcLTwentyOneHEleven}{0.160}
\newcommand{\NMCProseCiLoLTwentyOneHEleven}{0.122}
\newcommand{\NMCProseCiHiLTwentyOneHEleven}{0.198}
\newcommand{\NMCRankPOneLTwentyOneHEleven}{67}
\newcommand{\NMCIcPOneLTwentyOneHEleven}{0.05}
\newcommand{\NMCProseIcLEighteenHThree}{0.158}
\newcommand{\NMCProseCiLoLEighteenHThree}{0.104}
\newcommand{\NMCProseCiHiLEighteenHThree}{0.216}
\newcommand{\NMCRankPOneLEighteenHThree}{332}
\newcommand{\NMCIcPOneLEighteenHThree}{0.00}
\newcommand{\NMCProseIcLTwentyTwoHSeventeen}{0.155}
\newcommand{\NMCProseCiLoLTwentyTwoHSeventeen}{0.123}
\newcommand{\NMCProseCiHiLTwentyTwoHSeventeen}{0.186}
\newcommand{\NMCRankPOneLTwentyTwoHSeventeen}{41}
\newcommand{\NMCIcPOneLTwentyTwoHSeventeen}{0.08}
\newcommand{\NMCProseIcLTwentyTwoHEighteen}{0.151}
\newcommand{\NMCProseCiLoLTwentyTwoHEighteen}{0.127}
\newcommand{\NMCProseCiHiLTwentyTwoHEighteen}{0.175}
\newcommand{\NMCRankPOneLTwentyTwoHEighteen}{169}
\newcommand{\NMCIcPOneLTwentyTwoHEighteen}{0.01}
\newcommand{\NMCProseIcLTwentyHNineteen}{0.143}
\newcommand{\NMCProseCiLoLTwentyHNineteen}{0.117}
\newcommand{\NMCProseCiHiLTwentyHNineteen}{0.170}
\newcommand{\NMCRankPOneLTwentyHNineteen}{1{,}557}
\newcommand{\NMCIcPOneLTwentyHNineteen}{\ensuremath{-}0.04}
\newcommand{\NMCProseIcLTwentyFourHSixteen}{0.128}
\newcommand{\NMCProseCiLoLTwentyFourHSixteen}{0.108}
\newcommand{\NMCProseCiHiLTwentyFourHSixteen}{0.149}
\newcommand{\NMCRankPOneLTwentyFourHSixteen}{37}
\newcommand{\NMCIcPOneLTwentyFourHSixteen}{0.09}
\newcommand{\NMCProseIcLNineteenHThirtyEight}{0.115}
\newcommand{\NMCProseCiLoLNineteenHThirtyEight}{0.075}
\newcommand{\NMCProseCiHiLNineteenHThirtyEight}{0.156}
\newcommand{\NMCRankPOneLNineteenHThirtyEight}{123}
\newcommand{\NMCIcPOneLNineteenHThirtyEight}{0.02}
\newcommand{\NMCProseIcLTwentyOneHOne}{0.113}
\newcommand{\NMCProseCiLoLTwentyOneHOne}{0.087}
\newcommand{\NMCProseCiHiLTwentyOneHOne}{0.141}
\newcommand{\NMCRankPOneLTwentyOneHOne}{40}
\newcommand{\NMCIcPOneLTwentyOneHOne}{0.09}
\newcommand{\NMCProseIcLNineteenHThirtySeven}{0.109}
\newcommand{\NMCProseCiLoLNineteenHThirtySeven}{0.081}
\newcommand{\NMCProseCiHiLNineteenHThirtySeven}{0.139}
\newcommand{\NMCRankPOneLNineteenHThirtySeven}{1{,}362}
\newcommand{\NMCIcPOneLNineteenHThirtySeven}{\ensuremath{-}0.01}
\newcommand{\NMCProseIcLSixteenHThirtyEight}{0.109}
\newcommand{\NMCProseCiLoLSixteenHThirtyEight}{0.081}
\newcommand{\NMCProseCiHiLSixteenHThirtyEight}{0.135}
\newcommand{\NMCRankPOneLSixteenHThirtyEight}{1{,}259}
\newcommand{\NMCIcPOneLSixteenHThirtyEight}{0.00}
\newcommand{\NMCProseIcLSeventeenHEighteen}{0.106}
\newcommand{\NMCProseCiLoLSeventeenHEighteen}{0.091}
\newcommand{\NMCProseCiHiLSeventeenHEighteen}{0.121}
\newcommand{\NMCRankPOneLSeventeenHEighteen}{1{,}224}
\newcommand{\NMCIcPOneLSeventeenHEighteen}{0.00}
\newcommand{\NMCTransportIcLTwentyOneHTwentyFive}{0.92}
\newcommand{\NMCTransportIcLTwentySixHThirtySeven}{0.77}
\newcommand{\NMCTransportIcLTwentyFourHOne}{0.50}
\newcommand{\NMCTransportRankLTwentyTwoHSixteen}{10}
\newcommand{\NMCTransportRankLTwentyFourHNineteen}{12}
\newcommand{\NMCTransportRankLThirtyNineHThirtyEight}{14}
\newcommand{\NMCTransportRankLThirtyNineHThirteen}{15}
\newcommand{\NMCTransportOverlapTopEight}{4}
\newcommand{\NMCTransportOverlapTopSixteen}{9}
\newcommand{\NMCTransportOverlapTopForty}{21}
\newcommand{\NMCTransportEmissionSpearman}{+0.34}
\newcommand{\NMCTransportEmissionSpearmanTopOneHundred}{+0.45}
\newcommand{\NMCTransportTopSixteenLTwelveToTwentyThreeCount}{5}
\newcommand{\NMCTransportTopSixteenLTwelveToTwentyThreeShare}{32\%}
\newcommand{\NMCTransportTopSixteenLTwentyFourToThirtyOneCount}{9}
\newcommand{\NMCTransportTopSixteenLTwentyFourToThirtyOneShare}{58\%}
\newcommand{\NMCTransportTopSixteenLThirtyTwoToThirtyNineCount}{2}
\newcommand{\NMCLocusAEmissionTopEight}{0.74}
\newcommand{\NMCLocusAEmissionTopEightPTen}{0.55}
\newcommand{\NMCLocusAEmissionTopEightPNinety}{0.98}
\newcommand{\NMCLocusAOwnTopEight}{1.62}
\newcommand{\NMCLocusARandomTopEight}{0.00}
\newcommand{\NMCLocusABottomTopEight}{\ensuremath{-}0.58}
\newcommand{\NMCLocusAEmissionWins}{100}
\newcommand{\NMCLocusAEmissionTopSixteen}{1.08}
\newcommand{\NMCLocusAEmissionTopSixteenPTen}{0.80}
\newcommand{\NMCLocusAEmissionTopSixteenPNinety}{1.34}
\newcommand{\NMCLocusAOwnTopSixteen}{1.95}
\newcommand{\NMCLocusARandomTopSixteen}{0.09}
\newcommand{\NMCLocusABottomTopSixteen}{\ensuremath{-}0.14}
\newcommand{\NMCLocusAEmissionTopForty}{1.58}
\newcommand{\NMCLocusAEmissionTopFortyPTen}{1.35}
\newcommand{\NMCLocusAEmissionTopFortyPNinety}{1.84}
\newcommand{\NMCLocusAOwnTopForty}{2.16}
\newcommand{\NMCLocusARandomTopForty}{0.20}
\newcommand{\NMCLocusABottomTopForty}{1.38}
\newcommand{\NMCLocusAOwnSuffixTopSixteen}{1.57}
\newcommand{\NMCLocusBProseSuffixTopEight}{0.45}
\newcommand{\NMCLocusBWindowDecidedAgreement}{0.04}
\newcommand{\NMCLocusBProseWindowTopEightPTen}{0.58}
\newcommand{\NMCLocusBProseWindowTopEightPNinety}{1.38}
\newcommand{\NMCLocusBProseBothTopEight}{1.26}
\newcommand{\NMCLocusBBottomWindowTopEight}{\ensuremath{-}0.11}
\newcommand{\NMCLocusBProseWindowTopSixteenPTen}{0.88}
\newcommand{\NMCLocusBProseWindowTopSixteenPNinety}{1.76}
\newcommand{\NMCLocusBProseBothTopSixteen}{1.48}
\newcommand{\NMCLocusBBottomWindowTopSixteen}{\ensuremath{-}0.01}
\newcommand{\NMCLocusBProseWindowTopFortyPTen}{1.18}
\newcommand{\NMCLocusBProseWindowTopFortyPNinety}{2.14}
\newcommand{\NMCLocusBProseSuffixTopForty}{0.77}
\newcommand{\NMCLocusBProseBothTopForty}{1.71}
\newcommand{\NMCLocusBBottomWindowTopForty}{0.09}
\newcommand{\NMCLocusReadRegionAboveRandomNotes}{84}
\newcommand{\NMCLocusReadRegionAboveRandomNotesTopPositions}{71}
\newcommand{\NMCLocusNoteReadersNoteMin}{30\%}
\newcommand{\NMCLocusNoteReadersNoteMax}{53\%}
\newcommand{\NMCLocusNoteReadersRegionMin}{7\%}
\newcommand{\NMCLocusNoteReadersRegionMax}{16\%}
\newcommand{\NMCLocusNoteReadersLiftMin}{3}
\newcommand{\NMCLocusNoteReadersLiftMax}{9}
\newcommand{\NMCLocusNoteReadersNeffMin}{27}
\newcommand{\NMCLocusNoteReadersNeffMax}{64}
\newcommand{\NMCLocusProseReadersProseMin}{60\%}
\newcommand{\NMCLocusProseReadersProseMax}{79\%}
\newcommand{\NMCLTwentyTwoHFifteenOnRegionLocus}{0.31}
\newcommand{\NMCLTwentyTwoHFifteenLiftLocus}{6}
\newcommand{\NMCLocusRRegionImportanceMax}{0.21}
\newcommand{\NMCLocusReadProseOtherChapters}{0.12}
\newcommand{\NMCLocusReadProseSink}{0.04}
\newcommand{\NMCLocusReadProseLift}{4.5}
\newcommand{\NMCLocusReadProseNeff}{328}
\newcommand{\NMCLocusReadEmissionOtherChapters}{0.21}
\newcommand{\NMCLocusReadEmissionSink}{0.11}
\newcommand{\NMCLocusReadEmissionLift}{3.1}
\newcommand{\NMCLocusReadEmissionNeff}{174}
\newcommand{\NMCLocusReadBottomRegion}{0.05}
\newcommand{\NMCLocusReadBottomOtherChapters}{0.14}
\newcommand{\NMCLocusReadBottomNote}{0.17}
\newcommand{\NMCLocusReadBottomProse}{0.56}
\newcommand{\NMCLocusReadBottomSink}{0.01}
\newcommand{\NMCLocusReadBottomLift}{2.9}
\newcommand{\NMCLocusReadBottomNeff}{287}
\newcommand{\NMCLocusReadRandomOtherChapters}{0.36}
\newcommand{\NMCLocusReadRandomSink}{0.21}
\newcommand{\NMCLocusReadRandomLift}{1.9}
\newcommand{\NMCLocusReadRandomNeff}{148}

\newcommand{\NMCShortLThirtyHElevenOnSibling}{0.263}
\newcommand{\NMCShortPTwoCRandomSixteenOnPrefix}{0.087}
\newcommand{\NMCShortPOneRandomSixteenLift}{1.25}

\newcommand{\NMDSelBaseMicroF}{0.076}
\newcommand{\NMDSelBaseValid}{0.99}
\newcommand{\NMDSelBaseMOnePOne}{0.815}
\newcommand{\NMDSelBaseMOnePTwo}{0.647}
\newcommand{\NMDSelBaseMOnePTwoC}{0.752}
\newcommand{\NMDSelBaseCompletedMicroF}{0.077}
\newcommand{\NMDSelTeacherMOnePOne}{0.887}
\newcommand{\NMDSelTeacherMOnePTwo}{0.750}
\newcommand{\NMDSelTeacherMOnePTwoC}{0.807}
\newcommand{\NMDSelVanillaSZeroMicroF}{0.101}
\newcommand{\NMDSelVanillaSOneMicroF}{0.092}
\newcommand{\NMDSelVanillaSTwoMicroF}{0.091}
\newcommand{\NMDSelVanillaMicroF}{0.095}
\newcommand{\NMDSelVanillaMicroFSd}{0.006}
\newcommand{\NMDSelVanillaSZeroValid}{0.90}
\newcommand{\NMDSelVanillaSOneValid}{0.95}
\newcommand{\NMDSelVanillaSTwoValid}{0.91}
\newcommand{\NMDSelVanillaSZeroMOnePTwoC}{0.752}
\newcommand{\NMDSelVanillaSOneMOnePTwoC}{0.751}
\newcommand{\NMDSelVanillaSTwoMOnePTwoC}{0.752}
\newcommand{\NMDSelVanillaSZeroMOnePOne}{0.839}
\newcommand{\NMDSelVanillaSOneMOnePOne}{0.837}
\newcommand{\NMDSelVanillaSTwoMOnePOne}{0.840}
\newcommand{\NMDSelVanillaMOnePOneGapPct}{33\%}
\newcommand{\NMDSelVanillaSZeroMOnePTwo}{0.665}
\newcommand{\NMDSelVanillaSOneMOnePTwo}{0.664}
\newcommand{\NMDSelVanillaSTwoMOnePTwo}{0.671}
\newcommand{\NMDSelVanillaMTwoLiftPOne}{2.01--2.09}
\newcommand{\NMDSelVanillaMTwoLiftPTwo}{3.29--3.31}
\newcommand{\NMDSelVanillaMTwoLiftPTwoC}{2.61--2.68}
\newcommand{\NMDSelMistillSZeroMicroF}{0.110}
\newcommand{\NMDSelMistillSOneMicroF}{0.099}
\newcommand{\NMDSelMistillSTwoMicroF}{0.103}
\newcommand{\NMDSelMistillMicroF}{0.104}
\newcommand{\NMDSelMistillMicroFSd}{0.006}
\newcommand{\NMDSelMistillSZeroValid}{0.92}
\newcommand{\NMDSelMistillSOneValid}{0.91}
\newcommand{\NMDSelMistillSTwoValid}{0.97}
\newcommand{\NMDSelMistillSZeroMOnePTwoC}{0.773}
\newcommand{\NMDSelMistillSOneMOnePTwoC}{0.772}
\newcommand{\NMDSelMistillSTwoMOnePTwoC}{0.772}
\newcommand{\NMDSelMistillMOnePTwoCGapPct}{37\%}
\newcommand{\NMDSelMistillSZeroMOnePOne}{0.832}
\newcommand{\NMDSelMistillSOneMOnePOne}{0.831}
\newcommand{\NMDSelMistillSTwoMOnePOne}{0.832}
\newcommand{\NMDSelMistillMOnePOneGapPct}{24\%}
\newcommand{\NMDSelMistillSZeroMOnePTwo}{0.663}
\newcommand{\NMDSelMistillSOneMOnePTwo}{0.663}
\newcommand{\NMDSelMistillSTwoMOnePTwo}{0.667}
\newcommand{\NMDSelMistillMTwoLiftPOne}{2.28--2.33}
\newcommand{\NMDSelMistillMTwoLiftPTwo}{3.56--3.58}
\newcommand{\NMDSelMistillMTwoLiftPTwoC}{2.84--2.89}

\newcommand{\NMDSelPairMTwoNotesRange}{78--99}
\newcommand{\NMDSelPairMTwoDeltaRange}{+0.20 to +0.32}
\newcommand{\NMDSelPairMTwoP}{\ensuremath{<10^{-4}}}

\newcommand{\NMDSelZnoteMicroFLambdaOne}{0.056}
\newcommand{\NMDSelZnoteMicroFLambdaPointThree}{0.071}
\newcommand{\NMDSelZnoteMicroFLambdaPointOne}{0.066}
\newcommand{\NMDSelZnoteValidLambdaOne}{0.71}
\newcommand{\NMDSelZnoteValidLambdaPointThree}{0.76}
\newcommand{\NMDSelZnoteValidLambdaPointOne}{0.89}
\newcommand{\NMDSelZnoteMOnePTwoCLambdaOne}{0.743}
\newcommand{\NMDSelZnoteMOnePTwoCLambdaPointThree}{0.743}
\newcommand{\NMDSelZnoteMOnePTwoCLambdaPointOne}{0.750}
\newcommand{\NMDSelAttnInterMicroFLambdaOne}{0.086}
\newcommand{\NMDSelAttnInterMicroFLambdaPointThree}{0.074}
\newcommand{\NMDSelAttnInterValidLambdaOne}{0.75}
\newcommand{\NMDSelAttnInterValidLambdaPointThree}{0.83}
\newcommand{\NMDSelAttnInterMOnePTwoCLambdaOne}{0.778}
\newcommand{\NMDSelAttnInterMOnePTwoCLambdaPointThree}{0.775}
\newcommand{\NMDSelGridMicroFMin}{0.033}
\newcommand{\NMDSelGridMicroFMax}{0.048}
\newcommand{\NMDSelPreregBestValid}{0.99}
\newcommand{\NMDSelPreregBestMOnePTwoC}{0.766}
\newcommand{\NMDSelOneSeedNoiseMicroF}{0.005}
\newcommand{\NMDSelMechToCERatioLambdaOne}{10}
\newcommand{\NMDSelMechToCERatioLambdaPointThree}{3}
\newcommand{\NMDSelCapRateLambdaOne}{27--29\%}
\newcommand{\NMDSelCapRateVanilla}{11\%}
\newcommand{\NMDSelCapRateBase}{1\%}
\newcommand{\NMDWriteRawPOneTermDrop}{93\%}
\newcommand{\NMDWriteRawCorrStart}{+0.11}
\newcommand{\NMDWriteRawCorrEnd}{\ensuremath{-}0.06}
\newcommand{\NMDWriteRawMOnePOneAfter}{0.300}
\newcommand{\NMDWriteZnoteCorr}{+0.38}
\newcommand{\NMDWriteZnoteCompletedMicroF}{0.073}

\newcommand{\NMDDoseLambdaPointOnePTwoCDelta}{+0.024}
\newcommand{\NMDDoseLambdaPointOnePTwoCHigher}{69}
\newcommand{\NMDDoseLambdaPointOnePTwoCLower}{11}
\newcommand{\NMDDoseLambdaPointOnePTwoCP}{\ensuremath{<10^{-4}}}
\newcommand{\NMDDoseLambdaPointOnePOneDelta}{\ensuremath{-}0.007}
\newcommand{\NMDDoseLambdaPointOnePOneHigher}{18}
\newcommand{\NMDDoseLambdaPointOnePOneLower}{33}
\newcommand{\NMDDoseLambdaPointOnePOneP}{0.05}
\newcommand{\NMDDoseLambdaPointOnePTwoDelta}{\ensuremath{-}0.002}
\newcommand{\NMDDoseLambdaPointOnePTwoP}{0.25}
\newcommand{\NMDDoseLambdaPointThreePTwoCDelta}{+0.034}
\newcommand{\NMDDoseLambdaPointThreePTwoCHigher}{73}
\newcommand{\NMDDoseLambdaPointThreePTwoCLower}{14}
\newcommand{\NMDDoseLambdaPointThreePTwoCP}{\ensuremath{<10^{-4}}}
\newcommand{\NMDDoseLambdaPointThreePOneDelta}{\ensuremath{-}0.015}
\newcommand{\NMDDoseLambdaPointThreePOneHigher}{15}
\newcommand{\NMDDoseLambdaPointThreePOneLower}{46}
\newcommand{\NMDDoseLambdaPointThreePOneP}{\ensuremath{10^{-4}}}
\newcommand{\NMDDoseLambdaPointThreePTwoDelta}{\ensuremath{-}0.011}
\newcommand{\NMDDoseLambdaPointThreePTwoHigher}{34}
\newcommand{\NMDDoseLambdaPointThreePTwoLower}{56}
\newcommand{\NMDDoseLambdaPointThreePTwoP}{0.03}
\newcommand{\NMDDoseLambdaOnePTwoCDelta}{+0.041}
\newcommand{\NMDDoseLambdaOnePTwoCHigher}{67}
\newcommand{\NMDDoseLambdaOnePTwoCLower}{16}
\newcommand{\NMDDoseLambdaOnePTwoCP}{\ensuremath{<10^{-4}}}
\newcommand{\NMDDoseLambdaOnePOneDelta}{\ensuremath{-}0.020}
\newcommand{\NMDDoseLambdaOnePOneHigher}{19}
\newcommand{\NMDDoseLambdaOnePOneLower}{48}
\newcommand{\NMDDoseLambdaOnePOneP}{\ensuremath{5{\times}10^{-4}}}
\newcommand{\NMDDoseLambdaOnePTwoDelta}{\ensuremath{-}0.016}
\newcommand{\NMDDoseLambdaOnePTwoHigher}{26}
\newcommand{\NMDDoseLambdaOnePTwoLower}{64}
\newcommand{\NMDDoseLambdaOnePTwoP}{\ensuremath{10^{-4}}}

\newcommand{\NMDRowsPerNoteMedian}{562}
\newcommand{\NMDRowsPerNoteMax}{2{,}903}
\newcommand{\NMDEventsPerNoteMedian}{255}
\newcommand{\NMDMistillFitPeakGiB}{37.5}
\newcommand{\NMDMistillFitSecPerNote}{27}
\newcommand{\NMDMistillTierOnePeakGiB}{52}
\newcommand{\NMDWallVanillaHundredNotesMin}{24--34}
\newcommand{\NMDWallMistillHundredNotesMin}{76--78}
\newcommand{\NMDWallMistillTierOne}{6\,h\,50\,min}
\newcommand{\NMDMistillMinPerStep}{3.1--3.3}
\newcommand{\NMDWallVanillaTierOneHours}{3.0}
\newcommand{\NMDMistillFpGiBOneGPU}{97}
\newcommand{\NMDQwenBasePoolTopWeightPOne}{0.742}
\newcommand{\NMDQwenBasePoolTopWeightPTwo}{0.569}
\newcommand{\NMDQwenBasePoolTopWeightPTwoC}{0.512}

\newcommand{\NMDCePilotFpStart}{0.536}
\newcommand{\NMDCePilotFpEnd}{0.419}
\newcommand{\NMDCePilotBfStart}{0.532}
\newcommand{\NMDCePilotBfEnd}{0.478}
\newcommand{\NMDTeacherTraceMedianTokens}{6.1k}
\newcommand{\NMDTeacherTraceNinetiethTokens}{9.7k}
\newcommand{\NMDGoldTruncatedTraces}{200}
\newcommand{\NMDComputeBaselineCampaignGPUHours}{245}
\newcommand{\NMDComputeMistillCampaignGPUHours}{570}

\newcommand{\NMDMOneNullPosMin}{0.011}
\newcommand{\NMDMOneNullPosMax}{0.019}
\newcommand{\NMDPowerMdeMicroF}{0.025}
\newcommand{\NMDQwenMThreeTopHeadArms}{17}

\newcommand{\NMDQwenVanillaGapClosed}{0.68}
\newcommand{\NMDGoldMOnePTwoMedianMargin}{\ensuremath{-}1.04}
\newcommand{\NMDSmolVanillaSZeroMOnePTwoMedianMargin}{+1.08}
\newcommand{\NMDCeQwenVanillaFpSZeroStepFive}{0.553}
\newcommand{\NMDCeQwenVanillaFpSZeroStepTwenty}{0.461}
\newcommand{\NMDCeQwenVanillaFpSZeroStepSixty}{0.439}
\newcommand{\NMDCeQwenVanillaFpSZeroStepOneHundredTwenty}{0.376}
\newcommand{\NMDCeQwenVanillaFpSOneStepFive}{0.543}
\newcommand{\NMDCeQwenVanillaFpSOneStepTwenty}{0.469}
\newcommand{\NMDCeQwenVanillaFpSOneStepSixty}{0.431}
\newcommand{\NMDCeQwenVanillaFpSOneStepOneHundredTwenty}{0.384}
\newcommand{\NMDCeQwenVanillaFpSTwoStepFive}{0.548}
\newcommand{\NMDCeQwenVanillaFpSTwoStepTwenty}{0.466}
\newcommand{\NMDCeQwenVanillaFpSTwoStepSixty}{0.430}
\newcommand{\NMDCeQwenVanillaFpSTwoStepOneHundredTwenty}{0.387}
\newcommand{\NMDCeQwenVanillaBfSZeroStepFive}{0.560}
\newcommand{\NMDCeQwenVanillaBfSZeroStepTwenty}{0.477}
\newcommand{\NMDCeQwenVanillaBfSZeroStepSixty}{0.469}
\newcommand{\NMDCeQwenVanillaBfSZeroStepOneHundredTwenty}{0.446}
\newcommand{\NMDCeQwenVanillaBfSOneStepFive}{0.550}
\newcommand{\NMDCeQwenVanillaBfSOneStepTwenty}{0.487}
\newcommand{\NMDCeQwenVanillaBfSOneStepSixty}{0.463}
\newcommand{\NMDCeQwenVanillaBfSOneStepOneHundredTwenty}{0.454}
\newcommand{\NMDCeQwenVanillaBfSTwoStepFive}{0.554}
\newcommand{\NMDCeQwenVanillaBfSTwoStepTwenty}{0.483}
\newcommand{\NMDCeQwenVanillaBfSTwoStepSixty}{0.460}
\newcommand{\NMDCeQwenVanillaBfSTwoStepOneHundredTwenty}{0.456}
\newcommand{\NMDCeQwenMistillSZeroStepFive}{0.560}
\newcommand{\NMDCeQwenMistillSZeroStepTwenty}{0.484}
\newcommand{\NMDCeQwenMistillSZeroStepSixty}{0.475}
\newcommand{\NMDCeQwenMistillSZeroStepOneHundredTwenty}{0.453}
\newcommand{\NMDCeSmolVanillaFpSZeroStepFive}{0.660}
\newcommand{\NMDCeSmolVanillaFpSZeroStepTwenty}{0.541}
\newcommand{\NMDCeSmolVanillaFpSZeroStepSixty}{0.506}
\newcommand{\NMDCeSmolVanillaFpSZeroStepOneHundredTwenty}{0.424}
\newcommand{\NMDCeSmolVanillaFpSOneStepFive}{0.645}
\newcommand{\NMDCeSmolVanillaFpSOneStepTwenty}{0.555}
\newcommand{\NMDCeSmolVanillaFpSOneStepSixty}{0.501}
\newcommand{\NMDCeSmolVanillaFpSOneStepOneHundredTwenty}{0.431}
\newcommand{\NMDCeSmolVanillaBfSZeroStepFive}{0.667}
\newcommand{\NMDCeSmolVanillaBfSZeroStepTwenty}{0.561}
\newcommand{\NMDCeSmolVanillaBfSZeroStepSixty}{0.545}
\newcommand{\NMDCeSmolVanillaBfSZeroStepOneHundredTwenty}{0.517}
\newcommand{\NMDCeSmolMistillSZeroStepFive}{0.668}
\newcommand{\NMDCeSmolMistillSZeroStepTwenty}{0.565}
\newcommand{\NMDCeSmolMistillSZeroStepSixty}{0.548}
\newcommand{\NMDCeSmolMistillSZeroStepOneHundredTwenty}{0.516}
\newcommand{\NMDPoolQwenMistillVsVanillaBfTaskTies}{61}
\newcommand{\NMDPoolQwenMistillVsVanillaBfMOnePOneHigher}{65}
\newcommand{\NMDPoolQwenMistillVsVanillaBfMOnePOneLower}{131}
\newcommand{\NMDPoolQwenMistillVsVanillaBfMOnePOneMean}{\ensuremath{-}0.006}
\newcommand{\NMDPairSmolMistillVsVanillaBfTaskHigher}{32}
\newcommand{\NMDPairSmolMistillVsVanillaBfTaskLower}{26}
\newcommand{\NMDPairSmolMistillVsVanillaBfTaskP}{0.51}
\newcommand{\NMDPairSmolMistillVsVanillaBfMOnePTwoCHigher}{36}
\newcommand{\NMDPairSmolMistillVsVanillaBfMOnePTwoCLower}{82}
\newcommand{\NMDPairSmolMistillVsVanillaBfMOnePTwoCMean}{\ensuremath{-}0.013}
\newcommand{\NMDReplicateSTwoMicroF}{0.155}
\newcommand{\NMDReplicateTaskHigher}{60}
\newcommand{\NMDReplicateTaskLower}{64}
\newcommand{\NMDReplicateTaskP}{0.79}
\newcommand{\NMDReplicateSTwoMOnePTwoC}{0.802}
\newcommand{\NMDReplicateNoiseMicroF}{0.005}
\newcommand{\NMDReplicateNoiseMOnePTwoC}{0.003}
\newcommand{\NMDReplicateNoiseLAS}{0.02}
\newcommand{\NMDSmolInterLossPlateau}{1.5}
\newcommand{\NMDSmolAttnVsBothTaskHigher}{25}
\newcommand{\NMDSmolAttnVsBothTaskLower}{32}
\newcommand{\NMDSmolAttnVsBothTaskP}{0.43}
\newcommand{\NMDSmolAttnPTwoAttnLossStart}{0.99}
\newcommand{\NMDSmolAttnPTwoAttnLossEnd}{0.24}
\newcommand{\NMDNoAttrGateMaxDiff}{0.014}
\newcommand{\NMDNoAttrCleanMarginDiff}{0.045}
\newcommand{\NMDCrossHwPTwoCDelta}{+0.002}
\newcommand{\NMDCrossHwPTwoCP}{0.57}
\newcommand{\NMDCrossHwPTwoDelta}{+0.000}
\newcommand{\NMDCrossHwPTwoP}{0.93}
\newcommand{\NMDCrossHwPOneDelta}{+0.001}
\newcommand{\NMDCrossHwPOneP}{0.16}
\newcommand{\NMDCrossHwMedianDiff}{0.006}
\newcommand{\NMDSignTestPowerTwentyNotes}{25\%}
\newcommand{\NMDTwentyNoteSmolVanillaPTwoC}{0.685}
\newcommand{\NMDTwentyNoteSmolFeatKDPTwoC}{0.677}
\newcommand{\NMDSmolFeatKDVsVanillaPTwoCHigher}{101}
\newcommand{\NMDTwentyNoteQwenVanillaSZeroPTwoC}{0.881}
\newcommand{\NMDTwentyNoteQwenFeatKDPTwoC}{0.814}
\newcommand{\NMDQwenVanillaSZeroVsFeatKDPTwoCDelta}{+0.001}
\newcommand{\NMDQwenVanillaSZeroVsFeatKDPTwoCHigher}{76}
\newcommand{\NMDQwenVanillaSZeroVsFeatKDPTwoCLower}{74}
\newcommand{\NMDQwenVanillaSZeroVsFeatKDPTwoCP}{0.93}
\newcommand{\NMDQwenVanillaSTwoVsFeatKDPTwoCDelta}{+0.044}
\newcommand{\NMDQwenVanillaSTwoVsFeatKDPTwoCHigher}{98}
\newcommand{\NMDQwenVanillaSTwoVsFeatKDPTwoCLower}{52}
\newcommand{\NMDQwenVanillaSTwoVsFeatKDPTwoCP}{\ensuremath{2{\times}10^{-4}}}

\newcommand{\NMDQwenMistillVsBasePTwoCMin}{+0.075}
\newcommand{\NMDQwenMistillVsBasePTwoCMax}{+0.083}
\newcommand{\NMDBottomSmolPTwoCImportanceLeastNeg}{\ensuremath{-}0.035}
\newcommand{\NMDBottomSmolPTwoCImportanceMostNeg}{\ensuremath{-}0.334}
\newcommand{\NMDBottomSmolPTwoCImportanceSum}{\ensuremath{-}3.66}
\newcommand{\NMDTopSmolPTwoCImportanceMin}{+0.048}
\newcommand{\NMDTopSmolPTwoCImportanceMax}{+0.319}
\newcommand{\NMDTopSmolPTwoCImportanceSum}{+4.49}
\newcommand{\NMDBottomQwenPTwoCImportanceSum}{\ensuremath{-}2.66}
\newcommand{\NMDTopQwenPTwoCImportanceSum}{+3.71}
\newcommand{\NMDBottomSmolSZeroSupervisedInOwnTopFortyPOne}{0}
\newcommand{\NMDBottomSmolSZeroSupervisedInOwnTopFortyPTwo}{0}
\newcommand{\NMDBottomSmolSZeroSupervisedInOwnTopFortyPTwoC}{1}
\newcommand{\NMDBottomSmolSOneSupervisedInOwnTopForty}{0}
\newcommand{\NMDBottomQwenSupervisedInOwnTopForty}{0}
\newcommand{\NMDBottomSmolSZeroBaseOverlapPOne}{38}
\newcommand{\NMDBottomSmolSZeroBaseOverlapPTwo}{28}
\newcommand{\NMDBottomSmolSZeroBaseOverlapPTwoC}{30}
\newcommand{\NMDBottomSmolSOneBaseOverlapPOne}{38}
\newcommand{\NMDBottomSmolSOneBaseOverlapPTwo}{28}
\newcommand{\NMDBottomSmolSOneBaseOverlapPTwoC}{31}
\newcommand{\NMDMistillSmolSZeroBaseOverlapPOne}{37}
\newcommand{\NMDMistillSmolSZeroBaseOverlapPTwo}{30}
\newcommand{\NMDMistillSmolSZeroBaseOverlapPTwoC}{32}
\newcommand{\NMDBottomQwenBaseOverlapPOne}{35}
\newcommand{\NMDBottomQwenBaseOverlapPTwo}{29}
\newcommand{\NMDBottomQwenBaseOverlapPTwoC}{32}
\newcommand{\NMDMistillQwenSZeroBaseOverlapPOne}{36}
\newcommand{\NMDMistillQwenSZeroBaseOverlapPTwo}{32}
\newcommand{\NMDMistillQwenSZeroBaseOverlapPTwoC}{31}
\newcommand{\NMDKEightSmolBaseOverlapPOne}{38}
\newcommand{\NMDKEightSmolBaseOverlapPTwo}{31}
\newcommand{\NMDKEightSmolBaseOverlapPTwoC}{33}
\newcommand{\NMDBottomSmolOwnTopEightBaseRanks}{0--4, 6, 9 and 18}

\newcommand{\NMDEvalHoursPerArm}{1--2}
\newcommand{\NMDOwnHeadsChainHoursSmol}{3.5}
\newcommand{\NMDOwnHeadsChainHoursQwen}{7}
\newcommand{\NMDTeacherValMicroF}{0.199}
\newcommand{\NMDRetrieverRecallCorpus}{0.939}
\newcommand{\NMDPoolQwenMistillVsVanillaMTwoPOneDelta}{+0.170}
\newcommand{\NMDPoolQwenMistillVsVanillaMTwoPOneHigher}{348}
\newcommand{\NMDPoolQwenMistillVsVanillaMTwoPOneLower}{102}
\newcommand{\NMDPoolQwenMistillVsVanillaMTwoPOneP}{\ensuremath{2{\times}10^{-32}}}

\iclrfinalcopy

\begin{document}

\maketitle
\lhead{Preprint}

\begin{abstract}
Reasoning-trained language models can choose a document's few labels from thousands of candidates, zero-shot, by reasoning in text. We ask which internal components make these choices, and whether a student can be taught where to look as well as what to write, so that its decisions can be traced to a few components and to what they read. We study clinical coding of hospital discharge summaries (MIMIC-IV) with all 5,651 candidate diagnosis codes in the prompt. The model writes each code one token at a time, so every choice is a single token: one picks a region of the code list, the next ones a code within it, and one separates the code from a similar code named earlier in the reasoning. At each such token we measure each attention head's causal effect, by removing it and by transplanting it into another patient's summary. On essentially every summary, a small set of heads, the same across summaries and partly specific to each kind of choice, is necessary and sufficient for the choice, and blocking what the tested heads read (the chosen region of the code list; the similar code named earlier) removes their effect. \mistill{} builds on these findings: besides the teacher's text, it supervises where the student attends at the tokens where the teacher decides. In a student from another model family, its own heads, found after training and transplanted into another summary, restore nearly twice as much of the choice between similar codes as before training, where chain-of-thought distillation alone raises it by less than a third. A student that already carried most of it gains a little, consistently across seeds; the gain is also seen on the students' own completed traces. On two corpora where human coders marked each code's evidence, a \mistill{} student's decision heads attend to that evidence more, and more specifically, than a chain-of-thought-distilled student's. Controls locate the gain in supervising attention at the decision tokens; in the cross-family student, a target built from unselected teacher heads loses it. We detect no task difference from chain-of-thought distillation trained at the same numerical precision. Beyond this teacher and task, the structure recurs in a second reasoning model; on legal-document labelling and code generation, the heads this procedure ranks restore more of the model's choices than random heads; and without further training the students keep their gain on Spanish case reports and a question-answering benchmark.\\
Code: \href{\anoncodeurl}{\texttt{\anoncodehost}}
\end{abstract}

\section{Introduction}
\label{sec:intro}

Many practical labelling tasks ask for a small, unordered set of labels chosen from a fixed universe of thousands to
hundreds of thousands, scored by \microf, \macrof{} and propensity-scored metrics that up-weight rare labels
\citep{schultheis2023generalized,ullah2024navigating,zhang2025elmo,ullah2026haste}. Clinical coding is the canonical case: a discharge summary receives a handful of
the tens of thousands of ICD-10-CM codes \citep{mullenbach2018explainable,edin2023automated,soroush2024large,motzfeldt2025code}. Reasoning-trained
language models now attempt such tasks zero-shot with the candidates in context, writing long traces that narrow to a
region of the label space and then argue between close alternatives \citep{guo2025deepseek}. Which internal components
make those decisions, and can a smaller model be taught where to look when it makes them, rather than only what to write?

\begin{figure*}[t]
  \centering
  \begin{tikzpicture}[
    font=\sffamily\scriptsize,
    box/.style={draw=black!60, rounded corners=4pt, thick, inner sep=4pt, fill=black!1.5},
    boxtitle/.style={anchor=north west, font=\sffamily\footnotesize, inner sep=0pt},
    chip/.style={draw=black!35, rounded corners=2pt, inner sep=2.2pt, minimum height=0.52cm,
                 font=\ttfamily\footnotesize, anchor=west, fill=white, text height=1.6ex, text depth=0.35ex},
    pone/.style={chip, draw=Teal, fill=Teal!15, very thick},
    ptwo/.style={chip, draw=NavyBlue, fill=NavyBlue!12, very thick},
    ptwoc/.style={chip, draw=BurntOrange, fill=BurntOrange!16, very thick},
    nullev/.style={chip, draw=black!55, dashed, fill=black!5},
    prose/.style={chip, draw=black!20, text=black!45},
    tag/.style={font=\sffamily\scriptsize\bfseries, inner sep=1pt},
    lab/.style={font=\sffamily\scriptsize, align=center, inner sep=1pt},
    brace/.style={decorate, decoration={brace, amplitude=4pt}, thick},
]
\node[box, anchor=north west, minimum width=13.9cm, minimum height=2.12cm] (prompt) at (0,0) {};
\node[boxtitle] at ([xshift=6pt,yshift=-6pt]prompt.north west)
  {\textbf{Prompt}\quad{\itshape\color{black!50}[discharge summary]}\quad
   {\color{black!75}+ the inventory: all 5{,}651 ICD-10-CM codes, one line each}};
\node[anchor=north west, inner sep=0pt] (inv) at ([xshift=12pt,yshift=-21pt]prompt.north west) {%
  \ttfamily\scriptsize\setlength{\tabcolsep}{3pt}\renewcommand{\arraystretch}{1.08}%
  \begin{tabular}{@{}l l l@{}}
    I10   & Essential (primary) hypertension & \\
    \rowcolor{Teal!13} N18.1 & Chronic kidney disease, stage 1 & \textcolor{GainRed}{\sffamily\bfseries near-miss}\\
    \rowcolor{Teal!13} N18.9 & Chronic kidney disease, unspecified & \textcolor{InteractionGreen}{\sffamily\bfseries\checkmark\ emitted}\\
    $\vdots$ & {\color{black!55}(5{,}648 more lines)} &
  \end{tabular}};
\draw[brace, Teal] ([xshift=4pt]inv.east |- 0,-1.05) -- ([xshift=4pt]inv.east |- 0,-1.66)
  node[midway, right=7pt, lab, align=left, text=Teal!80!black] (reglab)
  {\textbf{the region of the letter N:}\\every line whose code starts with N.\\\emph{Region readers} read here (Sec.~\ref{sec:char-read})};
\node[box, anchor=north west, minimum width=13.9cm, minimum height=3.12cm] (tracebox) at (0,-2.30) {};
\node[boxtitle] (ttl) at ([xshift=6pt,yshift=-6pt]tracebox.north west) {\textbf{Reasoning trace}};
\node[anchor=west, lab, inner sep=0pt] at ([xshift=7pt]ttl.east)
  {\textcolor{Teal!85!black}{\textbf{P1} chapter: which region}\;$\cdot$\;
   \textcolor{NavyBlue}{\textbf{P2} character: which code in it}\;$\cdot$\;
   \textcolor{BurntOrange!90!black}{\textbf{P2C} contrastive: against the near-miss}};
\def\ty{-4.10}
\node[prose] (c0) at (0.22,\ty) {\enspace$\cdots$\enspace};
\node[chip, right=2pt of c0] (nm) {N18.1};
\node[prose, right=2pt of nm] (c1) {\enspace$\cdots$\enspace};
\node[nullev, right=2pt of c1] (nul) {the};
\node[prose, right=2pt of nul] (c2) {\enspace$\cdots$\enspace};
\node[prose, right=2pt of c2] (c3) {\enspace$\cdots$\enspace};
\node[chip, right=2pt of c3, draw=Goldenrod!80!black, fill=Goldenrod!18] (loc) {kidney};
\node[chip, right=2pt of loc] (w2) {disease,};
\node[chip, right=2pt of w2] (w3) {unspecified};
\node[pone,  right=5pt of w3] (tN) {\textvisiblespace N};
\node[ptwo,  right=2pt of tN] (t1) {1};
\node[ptwo,  right=2pt of t1] (t8) {8};
\node[chip,  right=2pt of t8, text=black!55] (tdot) {.};
\node[ptwoc, right=2pt of tdot] (t9) {9};
\node[tag, text=Teal,        above=2pt of tN]   {P1};
\node[tag, text=NavyBlue,    above=2pt of t1]   {P2};
\node[tag, text=NavyBlue,    above=2pt of t8]   {P2};
\node[tag, text=BurntOrange, above=2pt of t9]   {P2C};
\draw[-{Stealth[length=5pt]}, thick, GainRed, densely dashed, rounded corners=6pt]
  ([yshift=13pt]t9.north) -- ++(0,0.26) -| ([yshift=1pt]nm.north)
  node[pos=0.25, fill=black!1.5, inner sep=1.5pt, lab, text=GainRed]
  {\texttt{N18.9} departs from its \emph{near-miss} \texttt{N18.1}, named earlier; a head re-reads it (\emph{loop-back}, Sec.~\ref{sec:char-read})};
\node[lab, below=4pt of nul, text=black!60, anchor=north] {\Pnull event: a prose\\position, random\\target (calibration)};
\node[lab, below=4pt of loc, text=Goldenrod!55!black, anchor=north] {\emph{locus}: the chapter can first\\be elicited here, by the\\\emph{prose-time set} (Sec.~\ref{sec:char-locus})};
\draw[brace, Teal] ([yshift=-3pt]t9.south east) -- ([yshift=-3pt]tN.south west)
  node[midway, below=6pt, anchor=north, lab, text=Teal!80!black]
  {the \emph{emission set} carries\\the chapter decision here};
\end{tikzpicture}

\vspace{2pt}
{\sffamily\scriptsize\setlength{\tabcolsep}{4pt}\renewcommand{\arraystretch}{1.12}%
\begin{tabular}{@{}p{0.485\textwidth}@{\hspace{0.02\textwidth}}p{0.485\textwidth}@{}}
\textbf{decision margin}: how strongly the model prefers the token it wrote over its competitors (Sec.~\ref{sec:setup-events}) &
\textbf{global set}: the same top-$k$ heads for every note, one ranking per kind of decision (Sec.~\ref{sec:setup-measure}) \\
\textbf{necessity} (mean-ablation): the share of the margin lost when a set of heads is replaced by its average output &
\textbf{span bucket}: a stretch of text (an inventory line, a note window, a trace sentence) over which attention is summed (Sec.~\ref{sec:distill-objective}) \\
\textbf{recovery} (knock-in; Fig.~\ref{fig:events-full}, bottom): the share of the decision restored when the set's outputs are copied into another note &
\textbf{own-heads recovery} (M4): the recovery of a student's own top heads, found after training; M1 and M4 are defined in Sec.~\ref{sec:distill-setup} \\
\end{tabular}}
  \caption{A coding decision is one token: writing \texttt{N18.9}, the model makes one decision per token of the code.
  Illustrative (public ICD-10-CM codes and descriptions; placeholder note and reasoning); full version:
  Figure~\ref{fig:events-full}. Tags such as \resid{R-C1} mark a result; Appendix~\ref{app:repro} has its
  reproduction entry.}
  \label{fig:events}
\end{figure*}

We answer the first question with an instrument built for large label spaces (Figure~\ref{fig:events}). With the full
inventory of 5{,}651 ICD-10-CM codes in the prompt, \teacher{} writes each code one token at a time, so every coding
decision is the choice of one token. At each such token we measure how much each attention head contributes to the
model's preference for what it wrote, by removing the head or by copying it into another patient's note
(Section~\ref{sec:setup}). A small set of heads, the same for every note and partly different for each kind of
decision, carries each decision: eight of 1{,}600 heads remove \NFaithPOneNecKEightGlobal{} of the
model's preference for a code's first letter (its chapter) when ablated and restore \NFaithPOneKiKEightGlobal{} of it
when knocked into another patient's note. Interventions on what these heads read show that some read the inventory lines of the chosen chapter and that one
re-reads a similar code named earlier. For \NLocusWindowPct{} of chapter decisions, the chapter can be elicited (the
model names it when forced to answer early) a median of \NLocusDMedian{} tokens before its letter is written,
from a different, mid-layer set of heads (Section~\ref{sec:char}). The picture is largely preserved on a retrieved shortlist, without the prompt's instruction to
narrow first, and in a second reasoner.

\mistill{} uses this map of where the teacher decides and what its heads read. To chain-of-thought distillation it adds
a loss that makes the student's attention at the teacher's decision tokens resemble the teacher's, each model pooled
over its own top-ranked heads, so that no head-to-head mapping is needed (Section~\ref{sec:distill}). After training we
find the student's own top heads again and measure how much of a decision they carry. For the contrastive decision
between similar codes, \smol's heads restore \NOwnSmolBasePTwoCKiForty{} of it before training,
\NOwnSmolVanillaBfSZeroPTwoCKiForty{} to \NOwnSmolVanillaSeedMeanPTwoCKiForty{} after chain-of-thought distillation
(16-bit weights, or a 32-bit master copy; Section~\ref{sec:distill-setup}) and
\NOwnSmolMistillSeedMeanPTwoCKiForty{} after \mistill. \qwenfour{} already carries most of it untrained; there
\mistill{} adds a small, seed-stable gain over chain-of-thought distillation trained at the same numerical precision
(\NPoolQwenMistillVsVanillaBfPTwoCClusterDelta), though not at the other character decisions. On the task, no difference
from chain-of-thought distillation is detected at matched precision. Controls locate the gain in supervising attention
at the teacher's decision tokens; on \smol{} a target built from unselected teacher heads loses it
(Section~\ref{sec:distill-ablation}).

Why want this? Our aim is not to beat chain-of-thought distillation on the task score but to keep the task from
suffering while gaining a property that the score does not measure: a student whose decisions can be examined.
Own-heads recovery says how much of a decision a few identifiable heads of the student carry by themselves; the more
they carry, the more of that decision can be examined by asking what those heads read. And what they read moves
towards what a human coder reads: on two corpora where coders marked each code's evidence (Spanish case reports read
zero-shot, discharge summaries from the students' own domain), a \mistill{} student's chapter-decision heads attend to
it more, and more specifically, than a chain-of-thought-distilled student's (on \NEaCxLiftMcAbove{} of
\NEaCxLiftMcN{} and \NEaMdLiftMcAbove{} of \NEaMdLiftMcN{} documents; Section~\ref{sec:distill-evidence}). We do not
claim that \mistill{} copies the teacher's algorithm into the student, nor a gain on the task.

\textbf{Main contributions.} (1) An instrument that turns decisions over a large label space into single-token events with a
causal, calibrated effect for every attention head (Section~\ref{sec:setup}). (2) A causal account of a
reasoning-trained clinical coder: which heads carry each decision, what they read, and where the chapter can first be
elicited (Section~\ref{sec:char}). (3) \mistill{}, and evidence of what it changes in the student's own heads, on the
teacher's traces and on the student's own completed ones (Section~\ref{sec:distill}). (4) The limits, as findings: the
gain is at the contrastive decision, not at every decision the loss supervises; three further mechanism readouts do not
separate the arms (Appendix~\ref{app:metrics}); and a pre-registered evidence-flip test detected no improvement in
evidence-dependent decisions (Section~\ref{sec:limitations}). The core study covers one clinical task and two reasoners of one
lineage; Table~\ref{tab:app-claims} (Appendix~\ref{app:repro}) lists each claim with its evidence and its boundary.

\section{Setting and instrument: decisions are tokens}
\label{sec:setup}

\subsection{Task instance, teacher, corpus}
\label{sec:setup-task}
To ask which heads carry a coding decision, we fix the task, locate the decision in the trace and
define a head's effect on it. The task is ICD-10-CM coding of \mimiciv{} discharge summaries
\citep{johnson2023mimic}. A code such as N18.9 begins with a letter that indicates its chapter, a broad clinical area;
we call it the chapter letter (the 25 letters map imperfectly onto the 22 ICD-10-CM chapters). The prompt holds the benchmark's full
inventory of 5{,}651 diagnosis codes and their descriptions (about 106k tokens with the note); retrieval enters only for
distillation (Sec.~\ref{sec:char-stability}). The teacher is \teacher{} \citep{qwen2025qwen3} in thinking mode, with 40
layers of 40 heads. The corpus is 100 traces over 94 notes spanning 21 of the 22 ICD-10-CM chapters,
with teacher \microf{} \NMTeacherMicroFFullInventory{}; statistics are per trace, and for brevity we call each trace a
note (Appendices~\ref{app:data} and~\ref{app:teacher}).

\subsection{Decisions are tokens}
\label{sec:setup-events}
The teacher's tokenizer writes a code as its chapter letter followed by one token per character
(N18.9 as \texttt{\textvisiblespace N}, \texttt{1}, \texttt{8}, \texttt{.}, \texttt{9}), so each
coding decision is the choice of one token (Figure~\ref{fig:events}). We distinguish three kinds of decision and call the
tokens at which they are made \emph{events}. The \emph{chapter decision} (a \Pone{} event) is the chapter
letter at a code's first mention; the inventory lines sharing that letter are the code's \emph{region}. A
\emph{character decision} (\Ptwo) is each later character (the separator is not scored). The \emph{contrastive decision} (\Ptwoc) is the character at
which a code departs from a sibling code named earlier in the trace (N18.9 after N18.1), its \emph{near-miss}; an audit
finds that mention a rejected alternative in \NAuditRejectedPct{} of events, an enumeration in most of the rest
(Appendix~\ref{app:instrument}). Phase~1 is \Pone{} and Phase~2 is \Ptwo{} with \Ptwoc{}: kinds of decision, not
stretches of the trace. \Pnull{} events, prose positions with a
random letter as target, are positions where nothing is decided and serve for calibration. The 100 traces hold
\NEvPOneCount{} \Pone, \NEvPTwoCount{} \Ptwo{} and \NEvPTwoCCount{} \Ptwoc{} events (a code can have several
near-misses; Appendix~\ref{app:instrument}).
Every event is scored at the position $q$ whose output is the decided token by the model's own \emph{decision margin},
$\margin(q)=\mathrm{logit}(\text{target})-\logsumexp(\text{competitors})$: how strongly the model prefers the token it
wrote over its competitors, a multi-way form of the usual patching metric \citep{zhang_towards_2024}.

\subsection{Measuring a head's effect on a decision}
\label{sec:setup-measure}
\emph{Necessity: remove the head.} A hook at the input of each layer's output projection intervenes on one head's
output at the event. Mean-ablation replaces it with its mean over the note; the head's importance is the margin this
removes,
$I_h(q)=\margin(q)-\margin(q\mid h\ \text{mean-ablated})$; a set's \emph{necessity} is the share of the clean margin its
ablation destroys (above 1, the decision flips). First-order attribution from one backward pass
\citep{syed_attribution_2024} screens all 1{,}600 heads and nominates the top-40 at each event, plus the running global
top-16, for exact ablation; the screen is only moderately faithful to exact effects (Appendix~\ref{app:instrument}), so
every set is judged by exact intervention. \emph{Sufficiency: copy the heads into another note.} Knock-in (denoising
activation patching; \citealp{heimersheim_how_2024}; drawn in Figure~\ref{fig:events-full}) tests a set in a context that favours a different decision. We
take the next note in the corpus, a different patient, at an event of the same kind where the model wrote a different
token. There we put in the clean note's token at $q$ and the set's outputs from the clean run, and read the margin of
the clean decision. \emph{Recovery} is the share of the clean decision this restores,
$(\margin_{\text{set}}-\margin_{\text{corr}})/(\margin_{\text{all}}-\margin_{\text{corr}})$: 0 if nothing returns, and 1
if the set does as well as all 1{,}600 heads, which return the clean margin (Appendix~\ref{app:instrument}).

\emph{Calibration: discount heads that matter everywhere.} A ranking of 1{,}600 heads always has a top. At \Pnull{}
events, where nothing is decided, the same procedure declares \NMNullRawBhPos{} heads positive and \NMNullRawBhNeg{}
negative (Benjamini--Hochberg, FDR 0.05; \citealp{benjamini1995controlling}). We therefore rank by the decision-specific
importance $I_c=I-I_{\Pnull}$, which makes \NMNullSplitBhDiscoveries{} discoveries when one half of the \Pnull{}
events is tested against the other. The top-$k$ heads by $I_c$ for a kind of decision, the same for every note, are its \emph{global set}, tested
against count-matched random heads, the note's own ranking and the global bottom-$k$. Importances are means over notes;
necessity and recovery are medians over notes of per-note means. Appendix~\ref{app:teacher} discloses a retired
instrument.

\section{Which heads carry the decisions?}
\label{sec:char}

\subsection{A small global set is necessary and sufficient on nearly every note}
\label{sec:char-sets}
Does one fixed set of heads per kind of decision carry it on every note? Nearly (Table~\ref{tab:faithfulness}; every
$k$ in Appendix~\ref{app:heads}). Ablating the global top-8 of
1{,}600 heads destroys \NFaithPOneNecKEightGlobal{}, \NFaithPTwoNecKEightGlobal{} and
\NFaithPTwoCNecKEightGlobal{} of the chapter, character and contrastive margins, and the top-40 flips
chapter and character decisions; random sets of the same size remove almost nothing. Sufficiency is tested in another
patient's context, which favours a different chapter by a median of \NFaithPOneCorrMargin{} nats: there the chapter
top-8's clean outputs recover \NFaithPOneKiKEightGlobal{} of the clean decision and the top-40
\NFaithPOneKiKFortyGlobal{}, while random sets recover at most \NFaithPOneKiKOneHundredRandom{}. The global set beats
random on nearly every note (the ``notes'' columns). Its bottom-$k$ oppose the decision at small $k$; at $k=40$ and
100 the chapter bottom set is large enough to include heads with large generic effects, and removing it destroys the
margin too (Appendix~\ref{app:heads}; the character bottom sets behave alike). A note's own ranking is sometimes the better carrier
(Table~\ref{tab:app-own}, Appendix~\ref{app:moved}); we keep global sets because per-note sets are not identifiable at
this event count (Appendix~\ref{app:teacher}) and the global top-40 is within \NMKiOwnGapTopForty{} of the note's own
on knock-in.

\begin{table}[t]
  \centering\small
  \caption{Global per-phase sets: necessity (share of the clean margin destroyed by mean-ablation;
  $>1$: flipped) and knock-in recovery in another patient's context, against random and bottom-$k$ sets. Medians over
  notes; ``notes'': where the global set beats random (of 100; \Ptwoc: of 92). Every $k$: Table~\ref{tab:faithfulness-app}.
  \resid{R-C3}}
  \label{tab:faithfulness}
  \providecommand{\gensd}[2]{\begin{tabular}[t]{@{}c@{}}#1\\[-2pt]{\scriptsize$\pm$#2}\end{tabular}}
\providecommand{\genstack}[2]{\begin{tabular}[t]{@{}c@{}}#1\\[-2pt]{\scriptsize #2}\end{tabular}}
\providecommand{\genstackl}[2]{\begin{tabular}[t]{@{}l@{}}#1\\[-2pt]{\scriptsize #2}\end{tabular}}
\providecommand{\genhead}[2]{\begin{tabular}[b]{@{}c@{}}#1\\#2\end{tabular}}
\providecommand{\genstackiii}[3]{\begin{tabular}[t]{@{}c@{}}#1\\[-2pt]{\scriptsize #2}\\[-3pt]{\scriptsize #3}\end{tabular}}
\setlength{\tabcolsep}{3.5pt}
\begin{tabular}{@{}r cccc cccc cccc@{}}
\toprule
 & \multicolumn{4}{c}{\Pone{} (chapter)} & \multicolumn{4}{c}{\Ptwo{} (character)} & \multicolumn{4}{c}{\Ptwoc{} (contrastive)} \\
\cmidrule(lr){2-5}\cmidrule(lr){6-9}\cmidrule(lr){10-13}
$k$ & global & random & bottom & notes & global & random & bottom & notes & global & random & bottom & notes \\
\midrule
\multicolumn{13}{@{}l}{\textit{Necessity: share of the clean margin destroyed by mean-ablating the set ($>1$: decision flipped)}} \\
3 & 0.24 & 0.00 & \ensuremath{-}0.15 & 92 & 0.41 & 0.01 & \ensuremath{-}0.04 & 95 & 0.26 & 0.00 & \ensuremath{-}0.06 & 90 \\
8 & 0.62 & 0.00 & \ensuremath{-}0.36 & 95 & 0.78 & 0.00 & \ensuremath{-}0.18 & 98 & 0.63 & 0.00 & \ensuremath{-}0.06 & 90 \\
40 & 1.24 & 0.04 & 0.89 & 100 & 1.61 & 0.06 & 0.76 & 97 & 0.86 & 0.04 & \ensuremath{-}0.50 & 90 \\
100 & 1.60 & 0.09 & 1.57 & 100 & 2.21 & 0.26 & 1.41 & 99 & 0.89 & 0.07 & \ensuremath{-}0.16 & 90 \\
\midrule
\multicolumn{13}{@{}l}{\textit{Sufficiency by knock-in: share of the clean decision restored in another patient's context}} \\
3 & 0.16 & 0.00 & \ensuremath{-}0.16 & 100 & 0.24 & 0.00 & \ensuremath{-}0.45 & 100 & 0.05 & 0.00 & \ensuremath{-}0.33 & 80 \\
8 & 0.67 & 0.00 & \ensuremath{-}0.28 & 100 & 0.48 & 0.01 & \ensuremath{-}0.54 & 100 & 0.69 & 0.00 & \ensuremath{-}0.54 & 92 \\
40 & 0.86 & 0.04 & \ensuremath{-}0.30 & 100 & 0.79 & 0.02 & \ensuremath{-}0.81 & 100 & 0.85 & 0.02 & \ensuremath{-}0.63 & 92 \\
100 & 0.96 & 0.09 & \ensuremath{-}0.33 & 100 & 0.89 & 0.05 & \ensuremath{-}0.80 & 100 & 0.94 & 0.05 & \ensuremath{-}0.78 & 92 \\
\bottomrule
\end{tabular}

\end{table}

\subsection{The set is phase-structured}
\label{sec:char-phases}
Do the kinds of decision share heads? Partly (Appendix~\ref{app:heads}, Table~\ref{tab:heads-app}). One head,
\head{39}{38}, is first by $I_c$ for all three (first-order at \Ptwoc). Around it each kind has its own heads, for
example \head{26}{36} for the chapter decision, \head{39}{37} for character decisions and \head{30}{11} for the
contrastive one; the contrastive-specific heads rank near the bottom for chapter and character decisions. The chapter
top-40 has \NHeadPOneTopFortyLayersZeroToElevenCount{} heads in layers 0--11 and
\NHeadPOneTopFortyLayersTwentyFourToThirtyOneCount{} in layers 24--31: at the code token the chapter decision is made
in middle and late layers. The character and contrastive rankings are first-order
(Appendix~\ref{app:instrument}): a ranking only chooses a set, and the set's effects in Table~\ref{tab:faithfulness}
are measured exactly. Exact ablation of the contrastive ranking's 56 nominated heads reorders its top
(Appendix~\ref{app:heads}).

\subsection{What the heads read}
\label{sec:char-read}
What do these heads attend to? Attention patterns only suggest roles, so each role is then tested by intervention. We
split each attention row at the event by the kind of text it lands on (Appendix~\ref{app:readside},
Figure~\ref{fig:readside}). In the chapter set, \emph{region readers}, led by \head{26}{36} and
\head{28}{21}, put up to \NReadLTwentySixHThirtySixRegion{} of their attention on the emitted chapter's inventory
lines, a \emph{region lift} (attention share over token share) of up to \NReadLTwentySixHThirtySixLift{}. \emph{Prose readers} at
the code token, led by \head{39}{38}, put \NReadLThirtyNineHThirteenProse{} to \NReadLThirtyNineHThirtyEightProse{} of
theirs on the reasoning prose. At contrastive events \head{30}{11} puts \NReadLThirtyHElevenSibling{} of its attention
on the earlier mentions of the near-miss (random heads \NReadPTwoCRandSibling{}; the code's own
\NReadLThirtyHElevenOwn{}): a \emph{near-miss loop-back}. These reads are causal:
blocking a head's attention to the keys it is named for removes its effect on the decision (all of \head{30}{11}'s,
\NKeyAblLTwentyEightHTwentyOneShare{} of \head{28}{21}'s; matched other keys remove none;
Appendix~\ref{app:readside}). Editing the earlier mention out of the trace
lowers the contrastive margin, but no difference was detected between the tested replacements (a filler or another code
in that slot): sibling-specific identity processing is not established by these edits, and generic disruption of the
stored trace remains an alternative.

\subsection{Where the chapter can be elicited, and by which heads}
\label{sec:char-locus}
At the code token the chapter margin is usually already decisive (median \Pone{} margin \NEvPOneMargin{} nats). Is
it settled earlier, in the reasoning? We cut the context at an earlier position and append a short suffix that makes the model name a code
(a \emph{forcing probe}); the chapter is \emph{elicited} if it names the right one. The \emph{locus} is the earliest
position from which the forced chapter margin stays at least 2 nats (Appendix~\ref{app:locus}). For
\NLocusWindowPct{} of the 2{,}064 chapter mentions the locus lies in the reasoning written since the previous code, a
median of \NLocusDMedian{} tokens before the letter, reached in one step, and in \NLocusOwnDescPct{} of these cases at a word of the code's
own inventory description (Figure~\ref{fig:locus}a, in Appendix~\ref{app:moved}).

Which heads act there? Attribution over the 64 tokens ending at the locus (the attribution window) yields the
\emph{prose-time set}, led by \head{24}{19}. We call the chapter global set of Section~\ref{sec:char-sets}, which acts
at the code token, the \emph{emission set}. The prose-time set is a different, mid-layer set:
\NLocusTopFortyLayersTwelveToTwentyThree{} of its top-40 are in layers 12--23, and it shares
\NLocusOverlapEmissionKEight{} of its top-8 and \NLocusOverlapEmissionKForty{} of its top-40 heads with the emission
set. The two sets matter at different positions (Figure~\ref{fig:locus}b). Ablated only over the window, the prose-time top-16
removes \NLocusProseWinKSixteen{} of the forced margin and the emission top-16 \NLocusEmissionWinKSixteen{}; ablated
only at the probe's suffix, the order reverses (emission top-16 \NMLocusEmissionSufKSixteen{}, prose-time top-16
\NMLocusProseSufKSixteen{}). At the locus the prose-time heads read the discharge summary (\NLocusReadProseNote{} of their
attention; random \NLocusReadRandomNote{}; Figure~\ref{fig:locus}c).

Elicitable is not decided. With the prose-time top-16 ablated over the window and the original continuation left in
place, the margin where the letter is written loses only \NFollowProseSixteenRatio{} of what the forced margin
loses (a median \NFollowProseSixteenDNWin{} nats, above random heads on \NFollowProseVsRandomNoteHigher{} of 100
notes); and when another note's residual stream is swapped in layer by layer, the code token acquires the chapter only
from layer \NResidFlipLayer{} (Appendix~\ref{app:locus}). A mid-layer set that reads the discharge summary makes
the chapter elicitable a few tokens before the letter, and the emission set decides it at the code, reading that prose.
\resid{R-C6}

\subsection{The mechanism on a retrieved shortlist, and what distillation takes from it}
\label{sec:char-stability}
Does this hold on the shortlist the students see (Sec.~\ref{sec:distill-setup})? Largely. Rerun unchanged on
new teacher traces for the same notes with the shortlist in the prompt (Appendix~\ref{app:hd002}),
the $I_c$ rankings agree at Spearman
\NShortPOneSpearman{}/\NShortPTwoSpearman{}/\NShortPTwoCSpearman{}
(chapter/character/contrastive), with
\NShortPOneOverlapForty{}/\NShortPTwoOverlapForty{}/\NShortPTwoCOverlapForty{}
of the top-40 shared; top-40 knock-in recovers
\NShortPOneKiKFortyShort{}/\NShortPTwoKiKFortyShort{}/\NShortPTwoCKiKFortyShort{}; and
the chapter can be elicited a median of \NShortLocusDMedianShort{} tokens before the letter from the same
prose-time set (Spearman \NShortLocusProseSpearman{}). What shifts is the chapter top-8:
\head{39}{38} leaves it, and it recovers less (\NShortPOneKiKEightShort{} against
\NFaithPOneKiKEightGlobal{}). The picture also holds without the prompt's narrowing instruction
(Appendix~\ref{app:teacher}) and, with different heads, in a second reasoner, \texttt{QwQ-32B}
(Appendix~\ref{app:second}). On fresh EURLEX57K acts and BigCodeBench tasks, development-ranked teacher heads showed
higher normalized recovery at structural first-mention events than count-matched random heads (paired mean differences
\pci{\NXTEurConfDeltaMean}{\NXTEurConfDeltaLo}{\NXTEurConfDeltaHi} and
\pci{\NXTBcbConfDeltaMean}{\NXTBcbConfDeltaLo}{\NXTBcbConfDeltaHi}); a programming distillation pilot did not reproduce
the clinical recovery advantage (Appendix~\ref{app:tasks2}).

\textbf{What distillation takes from the characterization.} From the shortlist run it takes \emph{where} to
supervise (the events and the 8 tokens before each chapter letter, where the chapter becomes elicitable), \emph{which
heads} to pool (each kind of decision's global top-40, plus the prose-time set on those 8 tokens: the teacher's pooled
weights $\wT$) and \emph{what they read} (the read-side categories, as span buckets).
Sec.~\ref{sec:distill-ablation} tests the first two: random positions, or unselected teacher heads, lose \smol's gain.

\newpage
\section{Distilling the mechanism}
\label{sec:distill}

\subsection{\mistill: supervising the student's pooled heads at the teacher's decision events}
\label{sec:distill-objective}
\mistill{} adds to chain-of-thought distillation a loss on \emph{where the student looks} at the positions where the
teacher makes its decisions. Both models are teacher-forced on the teacher's trace
$z^T(x)=[\mathrm{prompt}(x);\tau^T(x)]$, and the student minimises
\begin{equation}
  \label{eq:total}
  \mathcal{L}=\mathcal{L}_{\mathrm{CoT}}+\lambda_{\mathrm{P1}}\,\mathcal{L}^{\mathrm{P1}}+\lambda_{\mathrm{P2}}\,\mathcal{L}^{\mathrm{P2}},
\end{equation}
where $\mathcal{L}_{\mathrm{CoT}}$ is exactly the \vanillakd{} loss (Appendices~\ref{app:baselines}
and~\ref{app:moved}) and $\lambda_{\mathrm{P1}}=\lambda_{\mathrm{P2}}=0.1$, selected on the validation notes
(Appendix~\ref{app:obj-lambda}).

\paragraph{Pooled attention over span buckets.} The two models have different heads and, for \smol, different
tokens, so attention cannot be compared head by head or token by token: both sides are pooled. Over heads, each side
uses its own causally ranked top-40 with weights $w^M_h=\operatorname{softmax}(I_c/\tau)$: for the teacher ($M{=}T$)
its global sets on the shortlist (Sec.~\ref{sec:char-stability}); for the student ($M{=}S$) the same instrument run once
on the untrained student, fixed for the run. Over tokens, attention is summed within \emph{span buckets}, stretches of
text that exist in both tokenizations: one per shortlist line, per 32-word window of the note and per sentence of the
trace. The pooled attention at position $q$ on bucket $b$ is
\begin{equation}
  \label{eq:pool}
  \bar{\alpha}^{M}_{q,b}=\sum_{h}w^{M}_{h}\sum_{j\in b}\alpha^{M,h}_{q,j}.
\end{equation}
the share of the pooled heads' attention that lands on that stretch of text. At \Ptwo{} and \Ptwoc{} events the code's
own inventory line, its category siblings' lines, its earlier mentions ($S$), the near-miss's earlier mentions ($N$) and
the code's emitted prefix are separate buckets.

\paragraph{Two terms per phase.} At each supervised position $q$,
\begin{equation}
  \label{eq:terms}
  \ell_q=\mathrm{KL}\big(\bar{\alpha}^{T}_{q}\,\big\|\,\bar{\alpha}^{S}_{q}\big)
        +\big(m^{S}_{q}c^{S}_{q}-m^{T}_{q}c^{T}_{q}\big)^{2},
\end{equation}
The first term makes the student look where the teacher looks. The second, the \emph{interaction term}, ties that
look to the decision: $m_q$ is the pooled mass on the decision's region (the emitted chapter's lines at \Pone; the
code's own line, $S$ and its prefix at \Ptwo{} and \Ptwoc) and $c_q$ is the pooled heads' direct contribution to the
event's margin, as a share of it (Appendices~\ref{app:obj-loss} and~\ref{app:moved}). $\mathcal{L}^{\mathrm{P1}}$ averages $\ell_q$ over every \Pone{} event and its KL
term over the $W{=}8$ tokens before it (Sec.~\ref{sec:char-stability}); there the
teacher is pooled over its prose-time set and the student over its own \Pone{} set, as no student prose-time set was
measured (removing the window leaves most of the gain in place, with a small paired decrease, Appendix~\ref{app:ablations}). $\mathcal{L}^{\mathrm{P2}}$ averages $\ell_q$ over
\Ptwo{} and \Ptwoc{} events, located by character offsets (Appendix~\ref{app:obj-loss}).

A pre-registered third term per phase, a squared error on $c$ alone, was harmful at pilot size and was removed,
and $\lambda$'s range was extended below its pre-registered floor --- both post-hoc (Appendix~\ref{app:obj-write});
$c$ is a direct-path proxy whose faithfulness to exact effects is measured in Appendix~\ref{app:instrument}.
Section~\ref{sec:distill-ablation} removes each of \mistill's choices in turn.

\subsection{Setup}
\label{sec:distill-setup}
\paragraph{Candidates, corpus and teacher.} Distillation runs on a retrieved shortlist instead of the full inventory:
1{,}000 codes plus the category siblings of the top 200 (the retriever: Appendix~\ref{app:data}). Its recall, \NMRetrieverRecallVal{} on the validation
notes and \NMRetrieverRecallTest{} on the test notes, is a shared ceiling; prompts shrink to about 30k tokens,
and $w^T$ comes from the teacher's replication on this shortlist (Sec.~\ref{sec:char-stability}). The corpus is 500
training notes (493 with a valid teacher trace) and 150 test notes, new and one per patient; the 100 characterization
notes are the validation set. The teacher scores \NMTeacherShortlistTestMicroF{} \microf{} on the test notes.

\paragraph{Students.} \qwenfour{} shares the teacher's family and tokenizer; \smol{}
\citep{bakouch2025smollm3} comes from another family, and its tokenizer merges digit runs, so it is supervised on
\NCovSmolPTwoKeptPct{} of the character and \NCovSmolPTwoCKeptPct{} of the contrastive events
(Appendix~\ref{app:objective}). Every arm samples in thinking mode with the teacher's settings; \qwenfour{} runs under
YaRN \citep{peng2024yarn} (Appendices~\ref{app:base-eval} and~\ref{app:moved}).

\paragraph{Baselines.} One strong method per family, from official code where it exists, at its published settings:
\vanillakd{} (cross-entropy and KL on the teacher trace), \distillm{} (off-policy
logit distillation; \qwenfour{} only), on-policy distillation (\opd; for \smol{} \gold, its cross-tokenizer form), and
\featkd{}, the generic attention and representation control. Every arm sees the same 493 notes for two epochs at batch 8,
with the same optimiser, targets and evaluation (\gold{}: one epoch, capped rollouts); tuning
budgets and wall-clock are not matched (Appendix~\ref{app:baselines}).

\paragraph{Numerical precision.} \mistill{}, its controls, \opd{} and \gold{} train 16-bit (bf16) weights with no
32-bit (fp32) master copy; \vanillakd{}, \featkd{} and \distillm{} keep fp32 master weights. A bf16 run can learn less
from the same data, because small updates round away (Appendix~\ref{app:baselines}), so we also trained \vanillakd{}
in bf16. A comparison is \emph{matched} when both arms were trained at the same precision.

\paragraph{Metrics and statistics.} Task: \microf, \macrof, P@8, P@15, \pspfifteen, tail \macrof. Fidelity:
trace \ppl, \bleu{} and \las{} (a frozen \simulator{} predicting the codes with and without the rationale). Mechanism:
\emph{agreement with the teacher's choice} (M1), the share of the teacher's events at which the student's own margin
favours the teacher's token, the margin the interaction term acts on; and \emph{own-heads recovery} (M4), the knock-in
recovery of each arm's own top-40 heads, re-selected after training (Sec.~\ref{sec:setup}), against a random set. M2
region lift, M3 near-miss loop-back and M5 locus distance are in Appendix~\ref{app:metrics}. Seeds per arm are in
Tables~\ref{tab:main} and~\ref{tab:mech}. Task and fidelity differences carry paired bootstrap 95\% intervals and
Wilcoxon tests per note (corpus-level: Appendix~\ref{app:results}); mechanism differences are seed-matched and tested
with the note as the unit.

\subsection{The task: no detected task difference at matched precision}
\label{sec:distill-main}
Does supervising the decision events change what the student gets right? Not detectably, in a matched comparison
(Table~\ref{tab:main}, within a precision group). In bf16, \vanillakd{} scores \NQwenVanillaBfMicroF{} \microf{} on
\qwenfour{} against \mistill's \NQwenMistillMicroF{} (per note \NQwenMistillMicroFVsVanillaBfDelta{}
[\NQwenMistillMicroFVsVanillaBfLo{}, \NQwenMistillMicroFVsVanillaBfHi{}]) and \NSmolVanillaBfMicroF{} on \smol{}
against \NSmolMistillMicroF{} (learning curve: Appendix~\ref{app:abl-curve}). The fp32 baselines score higher, and the gap is the precision's: with fp32 master weights \vanillakd{} reaches
\NQwenVanillaMicroF{}, \NQwenVanillaMicroFVsVanillaBfDelta{} above the same objective in bf16, a reading fixed before
that control ran (Appendix~\ref{app:res-precision}). \mistill{} needs its loss weight re-selected for fp32. At the
$\lambda$ selected in bf16 it does not share that gain (\NQwenMistillFpMicroF; \NQwenMistillFpMicroFVsVanillaDelta{}
[\NQwenMistillFpMicroFVsVanillaLo, \NQwenMistillFpMicroFVsVanillaHi] from fp32 \vanillakd). Re-running the
pre-registered $\lambda$ selection with fp32 master weights (one seed) picks $\lambda=\NSweepSelected$, at which the
fp32 \mistill{} student scores \NQwenMistillFpSelMicroF{} over three seeds with no detected task difference from fp32
\vanillakd{} (per note \NQwenMistillFpSelMicroFVsVanillaDelta{} [\NQwenMistillFpSelMicroFVsVanillaLo,
\NQwenMistillFpSelMicroFVsVanillaHi]; corpus-level \NCorpusQwenMistillFpSelVsVanillaDelta{}
[\NCorpusQwenMistillFpSelVsVanillaLo, \NCorpusQwenMistillFpSelVsVanillaHi] over the three seed pairs;
Appendix~\ref{app:obj-fp32}).

On \qwenfour{} in bf16, \mistill{} does change the readout its loss acts on: at the teacher's contrastive events the
student's margin favours the teacher's choice more often (M1 \Ptwoc{} \NQwenMistillMOnePTwoC{} against
\NQwenVanillaBfMOnePTwoC{}; higher on \NMPoolQwenMOnePTwoCVsVanillaBfHigher{} of \NMPoolQwenMOnePTwoCVsVanillaBfN{}
note-pairs over three seed pairs, descriptive); on \smol{} it does not (Table~\ref{tab:main}). \opd, trained in bf16
on-policy, reaches \NQwenOPDMicroF{} (one seed), level with fp32 \vanillakd{}
(\NQwenOPDMicroFVsVanillaDelta{} [\NQwenOPDMicroFVsVanillaLo{}, \NQwenOPDMicroFVsVanillaHi{}]), so the bf16 cost is
claimed for off-policy objectives only (\gold: Appendix~\ref{app:res-dissoc}). The
absolute level is low: the retriever's top-20 alone scores \NRetrieverTopTwentyMicroF, above every student and just
below the teacher (Appendix~\ref{app:results}); the claims that follow rest on the mechanism, not on the task score.

\begin{table*}[t]
  \centering\small
  \caption{Distillation on \mimiciv{} (150 test notes; mean $\pm$ sd over seeds; single-seed rows marked). Rows are
  grouped by training precision (Sec.~\ref{sec:distill-setup}); \mistill's matched reference is the \vanillakd{} row of
  its group. The fp32 \mistill{} rows are at
  the $\lambda$ selected in bf16 (0.1) and for fp32 (0.01). Valid: share of parseable answers; M1: agreement with the
  teacher's choice at contrastive events. Paired 95\% intervals: Appendix~\ref{app:results}. Reproduction entries:
  \resid{R-D1}\resid{R-D2}\resid{R-D3}\resid{R-D4}\resid{R-D5}\resid{R-D17}\resid{R-D22}\resid{R-D30}}
  \label{tab:main}
  \providecommand{\gensd}[2]{\begin{tabular}[t]{@{}c@{}}#1\\[-2pt]{\scriptsize$\pm$#2}\end{tabular}}
\providecommand{\genstack}[2]{\begin{tabular}[t]{@{}c@{}}#1\\[-2pt]{\scriptsize #2}\end{tabular}}
\providecommand{\genstackl}[2]{\begin{tabular}[t]{@{}l@{}}#1\\[-2pt]{\scriptsize #2}\end{tabular}}
\providecommand{\genhead}[2]{\begin{tabular}[b]{@{}c@{}}#1\\#2\end{tabular}}
\providecommand{\genstackiii}[3]{\begin{tabular}[t]{@{}c@{}}#1\\[-2pt]{\scriptsize #2}\\[-3pt]{\scriptsize #3}\end{tabular}}
\footnotesize\setlength{\tabcolsep}{3.5pt}
\begin{adjustbox}{max width=\linewidth}
\begin{tabular}{@{}l c cccc c cc c@{}}
\toprule
 & & \multicolumn{4}{c}{external (task)} & & \multicolumn{2}{c}{fidelity} & mech. \\
\cmidrule(lr){3-6}\cmidrule(lr){8-9}\cmidrule(lr){10-10}
 & & \multicolumn{3}{c}{\textsf{F1}} & & & & & \\
\cmidrule(lr){3-5}
Arm & Seeds & \textsf{micro} & \textsf{macro} & \textsf{tail} & \pspfifteen & valid & \ppl & \las & \genhead{M1}{\Ptwoc} \\
\midrule
\multicolumn{10}{@{}l}{\qwenfour} \\
Base\seedmark{1} & 1 & 0.104 & 0.036 & 0.029 & 0.074 & 0.987 & 1.749 & 0.093 & 0.759 \\
\vanillakd{} (bf16) & 3 & 0.143{\scriptsize$\pm$.007} & 0.051{\scriptsize$\pm$.002} & 0.041{\scriptsize$\pm$.001} & 0.101{\scriptsize$\pm$.003} & 0.898{\scriptsize$\pm$.010} & 1.581{\scriptsize$\pm$.002} & 0.183{\scriptsize$\pm$.008} & 0.773{\scriptsize$\pm$.000} \\
\mistill{} (bf16) & 3 & 0.149{\scriptsize$\pm$.002} & 0.054{\scriptsize$\pm$.002} & 0.043{\scriptsize$\pm$.002} & 0.108{\scriptsize$\pm$.006} & 0.893{\scriptsize$\pm$.012} & 1.590{\scriptsize$\pm$.001} & 0.192{\scriptsize$\pm$.008} & 0.803{\scriptsize$\pm$.001} \\
\opd{} (bf16)\seedmark{1} & 1 & 0.179 & 0.068 & 0.063 & 0.131 & 0.973 & 1.638 & 0.229 & 0.770 \\
\vanillakd{} (fp32) & 3 & 0.187{\scriptsize$\pm$.007} & 0.072{\scriptsize$\pm$.005} & 0.059{\scriptsize$\pm$.010} & 0.138{\scriptsize$\pm$.005} & 0.824{\scriptsize$\pm$.028} & 1.516{\scriptsize$\pm$.002} & 0.267{\scriptsize$\pm$.004} & 0.791{\scriptsize$\pm$.002} \\
\featkd{} (fp32) & 3 & 0.153{\scriptsize$\pm$.010} & 0.054{\scriptsize$\pm$.006} & 0.042{\scriptsize$\pm$.003} & 0.104{\scriptsize$\pm$.008} & 0.780{\scriptsize$\pm$.046} & 1.586{\scriptsize$\pm$.004} & 0.213{\scriptsize$\pm$.017} & 0.771{\scriptsize$\pm$.003} \\
\distillm{} (fp32)\seedmark{1} & 1 & 0.187 & 0.067 & 0.050 & 0.139 & 0.880 & 1.556 & 0.271 & 0.788 \\
\mistill{} (fp32, $\lambda$ 0.1) & 3 & 0.165{\scriptsize$\pm$.004} & 0.059{\scriptsize$\pm$.005} & 0.046{\scriptsize$\pm$.005} & 0.115{\scriptsize$\pm$.003} & 0.824{\scriptsize$\pm$.054} & 1.524{\scriptsize$\pm$.001} & 0.228{\scriptsize$\pm$.014} & 0.779{\scriptsize$\pm$.001} \\
\mistill{} (fp32, $\lambda$ 0.01) & 3 & 0.176{\scriptsize$\pm$.009} & 0.062{\scriptsize$\pm$.006} & 0.047{\scriptsize$\pm$.006} & 0.127{\scriptsize$\pm$.009} & 0.833{\scriptsize$\pm$.013} & 1.516{\scriptsize$\pm$.001} & 0.251{\scriptsize$\pm$.022} & 0.797{\scriptsize$\pm$.002} \\
\midrule
\multicolumn{10}{@{}l}{\smol} \\
Base\seedmark{1} & 1 & 0.003 & 0.002 & 0.000 & 0.002 & 0.153 & 1.947 & \ensuremath{-}0.019 & 0.721 \\
\vanillakd{} (bf16)\seedmark{1} & 1 & 0.055 & 0.016 & 0.009 & 0.031 & 0.380 & 1.703 & 0.081 & 0.756 \\
\mistill{} (bf16) & 3 & 0.046{\scriptsize$\pm$.013} & 0.015{\scriptsize$\pm$.005} & 0.007{\scriptsize$\pm$.004} & 0.027{\scriptsize$\pm$.008} & 0.289{\scriptsize$\pm$.027} & 1.708{\scriptsize$\pm$.001} & 0.075{\scriptsize$\pm$.010} & 0.740{\scriptsize$\pm$.001} \\
\gold{} (bf16; 6{,}144 cap)\seedmark{1} & 1 & 0.000 & 0.000 & 0.000 & 0.000 & 0.613 & 2.132 & \ensuremath{-}0.051 & 0.732 \\
\vanillakd{} (fp32) & 2 & 0.130{\scriptsize$\pm$.025} & 0.048{\scriptsize$\pm$.002} & 0.027{\scriptsize$\pm$.012} & 0.085{\scriptsize$\pm$.022} & 0.547{\scriptsize$\pm$.066} & 1.621{\scriptsize$\pm$.001} & 0.226{\scriptsize$\pm$.016} & 0.787{\scriptsize$\pm$.002} \\
\featkd{} (fp32)\seedmark{1} & 1 & 0.062 & 0.023 & 0.012 & 0.040 & 0.360 & 1.760 & 0.117 & 0.746 \\
\bottomrule
\end{tabular}
\end{adjustbox}

\end{table*}

\subsection{The student's own heads recover more of the contrastive decision}
\label{sec:distill-mech}
Does the student come to carry the contrastive decision in its own heads? We rerun the instrument of
Section~\ref{sec:setup} on every trained student, re-select its top-40 heads and knock them into another note's context:
its own-heads recovery (M4; Table~\ref{tab:mech}; per seed, Appendix~\ref{app:res-paired}). The heads are chosen after
training and the readout is a knock-in effect, not the pooled attention the loss matches. Where the student starts low, \mistill{} raises recovery most. Untrained, \smol{}'s own heads restore
\NOwnSmolBasePTwoCKiForty{} of the contrastive decision. \vanillakd{} brings this to
\NOwnSmolVanillaBfSZeroPTwoCKiForty{} in bf16 and to \NOwnSmolVanillaSZeroPTwoCKiForty{} and
\NOwnSmolVanillaSOnePTwoCKiForty{} in fp32, and \mistill{} to \NOwnSmolMistillSeedMeanPTwoCKiForty{}
(\NOwnSmolMistillSeedMeanRatioBase$\times$ the untrained value), higher than bf16 \vanillakd{} on
\NPoolSmolMistillVsVanillaBfPTwoCClusterHigher{} of 150 notes and than fp32 on
\NPoolSmolMistillVsVanillaPTwoCClusterHigher{}. Where the student starts high, the gain is small but consistent:
untrained \qwenfour{} already restores \NMQwenBasePctTeacher{} of the teacher's \NOwnTeacherPTwoCKiForty{}, and
\mistill{} adds \NPoolQwenMistillVsVanillaBfPTwoCClusterDelta{} over bf16 \vanillakd{}, paired per note (higher on
\NPoolQwenMistillVsVanillaBfPTwoCClusterHigher{} of 150 notes; every seed pair agrees), the size of the gain on M1.
Over fp32 \vanillakd{} it is \NPoolQwenMistillVsVanillaPTwoCClusterDelta{} and not significant with the note as the
unit.

The gain is at the contrastive decision, not at every decision the loss supervises. At the character-level \Ptwo{}
decisions the matched comparison runs the other way (\NPoolQwenMistillVsVanillaBfPTwoClusterDelta, lower on
\NPoolQwenMistillVsVanillaBfPTwoClusterLower{} of 150), although \mistill{} is above fp32 \vanillakd{} there
(\NPoolQwenMistillVsVanillaPTwoClusterDelta). With fp32 master weights the gain follows the loss weight: at the selected
fp32 setting ($\lambda=0.01$) it holds on average over three seed pairs at $k=40$ (mean $\Delta$
\NRepLPointZeroOnePTwoCClusterMean{} [\NRepLPointZeroOnePTwoCClusterLo, \NRepLPointZeroOnePTwoCClusterHi], with no
detected task difference), not at the $\lambda$ transferred from bf16 and not at $k=8$ (Appendix~\ref{app:obj-fp32}).
These readouts are teacher-forced on all 150 notes. On the students' own generated traces (the notes whose reasoning closes), every trained \qwenfour{} arm's top-40 recovery stays close to its
teacher-forced value and the gap to bf16 \vanillakd{} persists; \smol{} closes its reasoning on a minority of the notes in bf16
(valid \NSmolVanillaBfValid{} for \vanillakd, \NSmolMistillValid{} for \mistill; Table~\ref{tab:main}), and on those the advantage persists (Appendix~\ref{app:res-owntrace}). Zero-shot and teacher-forced, with no further
training, the gain also holds for these clinically trained students on Spanish case reports and their English translation, for
\smol{} and at MuSiQue-Ans answer events (Appendix~\ref{app:zeroshot}). Three further readouts do not separate the arms
(Appendix~\ref{app:metrics}).

\begin{table}[t]
  \centering\small
  \caption{Own-heads recovery, and what the heads read, on each arm's own causally ranked heads, found after training
  (150 test notes; medians over notes, then mean $\pm$ sd over seeds; single-seed rows marked; in parentheses, the
  floor from 40 random heads). The text discusses the \Ptwoc{} column; the differences it quotes are paired per note, not differences of cells.
  Knock-in recovery of the own top-40 per kind of decision; region lift and emitted-prefix share of the own
  top-16 \Ptwoc{} heads. Per seed, with the note-clustered comparisons: Table~\ref{tab:app-mech-seeds}.
  \resid{R-D6}\resid{R-D12}\resid{R-D15}\resid{R-D17}\resid{R-D22}\resid{R-D27}\resid{R-D30}}
  \label{tab:mech}
  \providecommand{\gensd}[2]{\begin{tabular}[t]{@{}c@{}}#1\\[-2pt]{\scriptsize$\pm$#2}\end{tabular}}
\providecommand{\genstack}[2]{\begin{tabular}[t]{@{}c@{}}#1\\[-2pt]{\scriptsize #2}\end{tabular}}
\providecommand{\genstackl}[2]{\begin{tabular}[t]{@{}l@{}}#1\\[-2pt]{\scriptsize #2}\end{tabular}}
\providecommand{\genhead}[2]{\begin{tabular}[b]{@{}c@{}}#1\\#2\end{tabular}}
\providecommand{\genstackiii}[3]{\begin{tabular}[t]{@{}c@{}}#1\\[-2pt]{\scriptsize #2}\\[-3pt]{\scriptsize #3}\end{tabular}}
\footnotesize\setlength{\tabcolsep}{4pt}
\begin{adjustbox}{max width=\linewidth}
\begin{tabular}{@{}l c ccc cc@{}}
\toprule
 & & \multicolumn{3}{c}{own-heads recovery (M4), top-40} & \multicolumn{2}{c}{read side (\Ptwoc)} \\
\cmidrule(lr){3-5}\cmidrule(lr){6-7}
Arm & Seeds & \genhead{\textbf{\Ptwoc}}{(floor)} & \Ptwo & \Pone & \genhead{M2 lift}{(floor)} & \genhead{prefix}{share} \\
\midrule
Teacher (\teacher; shortlist run HD-002) &  & 0.847\,{\scriptsize(0.015)} & 0.821 & 0.915 & 3.38\,{\scriptsize(1.95)} & 0.260 \\
\midrule
\multicolumn{7}{@{}l}{\qwenfour} \\
Base\seedmark{1} & 1 & 0.788\,{\scriptsize(0.030)} & 0.815 & 0.946 & 2.56\,{\scriptsize(1.17)} & 0.185 \\
\vanillakd{} (bf16) & 3 & 0.839{\scriptsize$\pm$.007}\,{\scriptsize(0.037)} & 0.758{\scriptsize$\pm$.007} & 0.952{\scriptsize$\pm$.000} & 3.48{\scriptsize$\pm$.12}\,{\scriptsize(1.58)} & 0.188{\scriptsize$\pm$.001} \\
\mistill{} (bf16) & 3 & 0.872{\scriptsize$\pm$.002}\,{\scriptsize(0.025)} & 0.735{\scriptsize$\pm$.001} & 0.956{\scriptsize$\pm$.002} & 3.63{\scriptsize$\pm$.06}\,{\scriptsize(1.60)} & 0.269{\scriptsize$\pm$.007} \\
\vanillakd{} (fp32) & 3 & 0.859{\scriptsize$\pm$.028}\,{\scriptsize(0.036)} & 0.725{\scriptsize$\pm$.053} & 0.946{\scriptsize$\pm$.002} & 3.44{\scriptsize$\pm$.32}\,{\scriptsize(1.33)} & 0.208{\scriptsize$\pm$.024} \\
\mistill{} (fp32, $\lambda$ 0.1) & 3 & 0.852{\scriptsize$\pm$.017}\,{\scriptsize(0.024)} & 0.782{\scriptsize$\pm$.023} & 0.949{\scriptsize$\pm$.000} & 3.01{\scriptsize$\pm$.14}\,{\scriptsize(1.89)} & 0.349{\scriptsize$\pm$.012} \\
\mistill{} (fp32, $\lambda$ 0.01) & 3 & 0.894{\scriptsize$\pm$.003}\,{\scriptsize(0.032)} & 0.680{\scriptsize$\pm$.009} & 0.949{\scriptsize$\pm$.002} & 3.41{\scriptsize$\pm$.21}\,{\scriptsize(1.30)} & 0.341{\scriptsize$\pm$.003} \\
\midrule
\multicolumn{7}{@{}l}{\smol} \\
Base\seedmark{1} & 1 & 0.416\,{\scriptsize(0.107)} & 0.809 & 0.985 & 3.57\,{\scriptsize(1.51)} & 0.121 \\
\vanillakd{} (bf16)\seedmark{1} & 1 & 0.476\,{\scriptsize(0.098)} & 0.867 & 0.985 & 3.29\,{\scriptsize(1.39)} & 0.151 \\
\mistill{} (bf16) & 3 & 0.771{\scriptsize$\pm$.020}\,{\scriptsize(0.123)} & 0.864{\scriptsize$\pm$.006} & 1.005{\scriptsize$\pm$.001} & 2.65{\scriptsize$\pm$.00}\,{\scriptsize(1.33)} & 0.441{\scriptsize$\pm$.002} \\
\vanillakd{} (fp32) & 2 & 0.532{\scriptsize$\pm$.012}\,{\scriptsize(0.101)} & 0.724{\scriptsize$\pm$.015} & 0.976{\scriptsize$\pm$.003} & 3.81{\scriptsize$\pm$.08}\,{\scriptsize(1.49)} & 0.172{\scriptsize$\pm$.011} \\
\bottomrule
\end{tabular}
\end{adjustbox}

\end{table}

\subsection{What the gain depends on: positions, teacher heads, student heads}
\label{sec:distill-ablation}
Each control replaces one thing the loss builds on (the teacher's positions; its ranked heads, the source of
the target; the student's ranked heads; Sec.~\ref{sec:char-stability}) and otherwise trains like \mistill{} (bf16, 493
notes, two epochs); an effect counts only if it holds per note against every \mistill{} seed (Table~\ref{tab:app-link}).

\paragraph{The ranked heads need the teacher's positions.} Supervising the student's own top heads at random
positions (\randpos) pushes own-heads \Ptwoc{} recovery \emph{below the untrained student} on both students
(\NOwnQwenRandPosSZeroPTwoCKiForty{} against \NOwnQwenBasePTwoCKiForty{}; \NOwnSmolRandPosSZeroPTwoCKiForty{} against
\NOwnSmolBasePTwoCKiForty{}; Figure~\ref{fig:ablation}, Appendix~\ref{app:ablations}). Random heads there do no such harm: they match \mistill{} on \qwenfour{} (\NOwnQwenRandBothSZeroPTwoCKiForty) but gain nothing on \smol{} (\NTpRbSZeroMed, \NTpRbSOneMed; Appendix~\ref{app:teacherpool}). Nor is the gain event-local loss pressure:
\vanillakd{} with token loss up-weighted at the same positions, matching \mistill's extra loss,
recovers what plain \vanillakd{} does (\NOwnQwenEventWeightedSZeroPTwoCKiForty{} against
\NOwnQwenVanillaBfSZeroPTwoCKiForty, same seed; below each \mistill{} seed on $\geq$
\NPairQwenEventWeightedSZeroVsMistillSOnePTwoCLower{} of 150).

\paragraph{On \smol, the teacher's ranked heads cannot be replaced by unselected ones.} With the student's side of
the loss unchanged, we rebuilt the targets without the ranking's selection: from every head of the ranked pools'
layers, or from forty random ones. \smol{} then stays at \vanillakd's level (\NTpCtlMin{} to \NTpCtlMax{}
against \NOwnSmolVanillaBfSZeroPTwoCKiForty), below every \mistill{} seed, for each of three control targets, while
\mistill{} retrained through the same code keeps its gain: the identified teacher heads contribute as the source of
the target. Which of their properties matters is not separated, and \qwenfour{} is untested
(Appendix~\ref{app:teacherpool}). \resid{R-D59}\resid{R-D60}

\paragraph{The gain lands in the student's own top heads, whichever heads are supervised.} On \smol, which student
heads are supervised sets how much is gained: the base's 40 most
negatively attributed heads, which oppose the decision, yield at least as much as its top-40
(\NOwnSmolBottomSZeroPTwoCKiForty{} and \NOwnSmolBottomSOnePTwoCKiForty{} against \NOwnSmolMistillSZeroPTwoCKiForty{} and
\NOwnSmolMistillSOnePTwoCKiForty{}), its top-8 less (\NOwnSmolKEightSZeroPTwoCKiForty{}), and 40 random heads least
(\NOwnSmolRandHeadsSZeroPTwoCKiForty{}). Yet the supervised heads are not where the mechanism ends up: in every arm the own top-40 after training shares
\NMCarrierOverlapRange{} heads with the untrained student's, and the bottom-40 arm's holds at most one head it
supervised (Appendix~\ref{app:abl-carriers}).
On \qwenfour, whose untrained heads already restore \NMQwenBasePctTeacher{} as much as the teacher's, the choice
barely matters (forty random heads reproduce \mistill{} on one of four runs), except that the most negatively
attributed heads damage it (\NOwnQwenBottomSZeroPTwoCKiForty{}, below the untrained student).

\paragraph{What the student-side ranking buys is not isolated.} The attention term applied to every head of the pool
layers, uniform and unranked, makes the own heads recover more than \mistill's on
the same seed (\NOwnQwenUniformSZeroPTwoCKiForty{} against \NOwnQwenMistillSZeroPTwoCKiForty) but pays for it on the
task (\NQwenUniformMicroF{} against \NQwenMistillSZeroMicroF{} \microf) and on M1. So event-local attention
supervision is what increases
the student's own-heads recovery. The ranked configuration
has the better observed task--recovery trade-off, but the control also differs in term
composition, head count and weight, so what the ranking itself buys is not isolated here
(Appendix~\ref{app:ablations}). Under matched validation selection against one frozen random pool, the ranked pick is
higher
on both endpoints (\Ptwoc{} recovery \NMatchedPTwoCClusterMean, \microf{} \NMatchedTaskDelta;
Appendix~\ref{app:abl-matched}). \resid{R-D31}\resid{R-D32}

\paragraph{Seeds, phases, terms and dose.} \mistill's three \qwenfour{} seeds agree within
\NMQwenMistillSeedSpreadPTwoC{} and do not separate per note, while random-head (\randheads) and \vanillakd{} seeds
do. The \Ptwoc{} gain is the Phase-2 supervision's: supervising only \Ptwo{}/\Ptwoc{}
events keeps about nine-tenths of it and only \Pone{} events loses most of it, while a third of the dose or either term
alone keeps most of it (Figure~\ref{fig:ablation-c}, Appendix~\ref{app:ablations}).

\subsection{The student's decision heads attend to the annotated evidence}
\label{sec:distill-evidence}
Do the heads that carry the student's chapter decision look at the evidence a human coder would cite? CodiEsp-X marks it on
the Spanish case reports of Appendix~\ref{app:zeroshot} (\NEaCxDocs{} test documents, zero-shot), MDACE~\citep{cheng2023mdace}
on \NEmInputs{} \texttt{MIMIC-III} discharge summaries in the training format. Teacher-forced, we read each student's own top-40
chapter heads, ranked after training, over the $W{=}8$ tokens before each chapter letter. We score the \emph{lift} of the
note windows that hold the code's evidence: their share of the heads' within-note attention over their token share (1: no
preference). The teacher's prose-time heads, the window's target, reach \NEaCxTeach{} and \NEaMdTeach{}
(Appendix~\ref{app:evidence}). \mistill{}
students' heads prefer it more than \vanillakd{} students', by \pci{\NEaCxLiftMcMean}{\NEaCxLiftMcLo}{\NEaCxLiftMcHi}
(higher on \NEaCxLiftMcAbove{} of the \NEaCxLiftMcN{} documents where the teacher writes a code with marked evidence) and \pci{\NEaMdLiftMcMean}{\NEaMdLiftMcLo}{\NEaMdLiftMcHi}
(\NEaMdLiftMcAbove{} of \NEaMdLiftMcN), and spend \NEaCxAbsLvlMistillAvg{} against \NEaCxAbsLvlCotKdAvg{} and
\NEaMdAbsLvlMistillAvg{} against \NEaMdAbsLvlCotKdAvg{} of all their attention on it. \randheads{}
reaches about the same lift (\mistill{} $-$ \randheads{} \NEaCxLiftMrMean{} and \NEaMdLiftMrMean) but not this coverage
(\pci{\NEaCxAbsMrMean}{\NEaCxAbsMrLo}{\NEaCxAbsMrHi} and \pci{\NEaMdAbsMrMean}{\NEaMdAbsMrLo}{\NEaMdAbsMrHi}), and dropping the
window term lowers both (one seed; Table~\ref{tab:app-evattn}). The student's decision heads thus look at the evidence a coder
would cite. \resid{R-D52}\resid{R-D54}

\section{Related work}
\label{sec:related}

\paragraph{Distillation.}
Training on a teacher's rationales transfers multi-step reasoning \citep{hsieh2023distilling, ho2023large,
magister2023teaching, shridhar_distilling_2023}; students match its hidden states, attention maps or
relations \citep{romero2015fitnets, zagoruyko2017paying, jiao2020tinybert, wang2020minilm}, and LLM logit
distillation has contrastive, on-policy and cross-tokenizer forms \citep{ko2025distillm, agarwal2024onpolicy,
patino2025unlocking}: all transfer text, re-weight its tokens \citep{feng2024keypoint}, or act everywhere on fixed layers. \mistill{} keeps the sequence-level
loss \citep{kim2016sequence} and adds a target at the teacher's decision positions on each model's causally ranked
heads, so unlike circuit distillation \citep{wadhwa2025circuit} it needs no component correspondence
(Appendix~\ref{app:related}).

\paragraph{Interpretability and coding.}
Activation patching and circuit analyses localize behavior to components \citep{vig2020investigating, meng2022locating,
wang2023interpretability, conmy2023towards}, and attribution patching screens components by gradient
\citep{syed_attribution_2024, kramar_atp_2024}. ICD coders are supervised classifiers with label-wise attention \citep{mullenbach2018explainable,
edin2023automated}, LLMs generate or retrieve candidate labels in context \citep{zhu_icxml_2024,
doosterlinck_-context_2024}, and interpretability in coding has targeted trained classifiers \citep{wu_dila_2025, edin2024unsupervised}. New
here: decisions over the full inventory as token-level events scored by the model's margin, with per-note necessity
and sufficiency against a null calibration.

\section{Limitations and conclusion}
\label{sec:limitations}
\textbf{Limitations.} The core study covers one clinical task and two reasoners of one lineage. Read-side roles are causal for three
tested heads and descriptive elsewhere; character ranking is first-order. Reasoning loci cover only \NLocusWindowPct{} of chapter
mentions, which the code largely re-derives. A \Ptwoc{} event follows mention order, not necessarily rejection of an alternative. Sibling-specific identity processing is not
established by the near-miss edits, and the students' edit sensitivity is inconsistent across precision and family. On \NEFCases{} evidence-flip cases no improvement over \vanillakd{} in directional sensitivity was detected, and decode-time masking
of the evidence keys through nominated heads detected no selective effect (Appendix~\ref{app:evflip}); these tests leave evidence
propagated during prefill accessible. The evidence-attention result (Section~\ref{sec:distill-evidence}) is read-side: deleting the
annotated evidence weakens the chapter decision little or no more under \mistill{} than under \vanillakd{}
(\pci{\NEdCxDeltaMcMean}{\NEdCxDeltaMcLo}{\NEdCxDeltaMcHi} and \pci{\NEdMdDeltaMcMean}{\NEdMdDeltaMcLo}{\NEdMdDeltaMcHi} of its margin;
Appendix~\ref{app:evidence}). Distillation shares a retriever
ceiling across arms (Section~\ref{sec:distill-main}) and needs white-box access to both models. With fp32
master weights the mechanism gain and the task cost follow the loss weight, which must be selected for the precision. Most controls and ablations use one seed; $\tau$, span-bucket and symmetric-window ablations remain untested
(Appendices~\ref{app:ablations} and~\ref{app:teacherpool}); \mistill{} trains slower than \vanillakd{} (Appendix~\ref{app:objective}). Two declared extensions did not detect the recovery advantage over \vanillakd{}, and in both the selected random-head
configuration recovered more: distillation on MuSiQue-Ans itself, where no student causal claim is made (Appendix~\ref{app:musique}), and
a programming development pilot, where \mistill{} recovered less and task noninferiority was not established (Appendix~\ref{app:tasks2}).

\textbf{Conclusion.} In the clinical coder, small global, phase-structured head sets carry coding decisions; where a reasoning
locus is found, a distinct mid-layer set makes the chapter elicitable before emission. Event-local attention supervision raises
students' own-heads contrastive recovery on teacher and completed student traces across supervised student pools, with no detected
task difference at matched precision, and makes their decision heads attend more specifically to human-annotated evidence. The pre-registered evidence-flip test detected no improvement in evidence-dependent
decisions through the tested routes.

\subsection*{AI use statement}
In this work, we used generative AI tools (a large language model coding assistant) for implementing and debugging
the experiment code (event extraction, attribution and patching, training and evaluation scripts, readers and
figure scripts), for launching and monitoring runs on our clusters and rented GPUs, for analysing results against
pre-registered readings, for maintaining the experiment logbook and planning documents, and for drafting and editing
parts of the paper, including its tables, figures and bibliography entries. Generative AI also assisted in drafting the
evidence-flip counterfactuals of Appendix~\ref{app:evflip}, which were retained after human clinical review by the author,
whose approval ledger is the adjudication of record, and the public-task evidence pairs of Appendix~\ref{app:musique}, adjudicated
by the author under written conventions and frozen before any student evaluation or flipped, control or intervention output (original-only
teacher traces for eligibility were the declared exception; the ledger is released), and the event-extraction audit sheets of
Appendix~\ref{app:tasks2}, AI-drafted under written conventions and adopted or amended through the author's adjudication
(\NXTEurAuditRows{} and \NXTBcbAuditRows{} sampled events from \NXTEurAuditInputs{} acts and \NXTBcbAuditInputs{} tasks; earlier draft
sheets and example pairs from that screen were archived unapproved and support no reported result). We have not used generative AI tools to generate any other data, labels
or results: reported evaluation statistics were computed by the analysis scripts from model outputs and experiment
artifacts, and the tables and macros are generated rather than typed. All AI-assisted work was reviewed by the author: code was
tested (unit tests, hook validation, instrument self-tests) and its outputs cross-checked against independent reads
of the artifacts, every citation was checked against its primary source, and every claim was checked against the
artifacts it cites. We take responsibility for the final content of this work.

\subsection*{Ethics statement}
\mimiciv{}, and \texttt{MIMIC-III} for the MDACE evidence annotations of Appendix~\ref{app:evidence}, are used under the
PhysioNet credentialed-access data use agreement. No note text, reasoning-trace text or
other protected content appears in the paper, its figures, the supplementary material or the released code: the
paper reports only public ICD-10-CM codes, attention-head names and statistics, and the worked example (Figure~\ref{fig:events}) uses
public codes, their inventory descriptions and placeholder text. Automated clinical coding is a decision-support setting; the models here are research artifacts,
not validated for clinical use, and their errors (Section~\ref{sec:distill-main}) would require human review in any
deployment.

\subsection*{Reproducibility statement}
Appendix~\ref{app:repro} holds one reproduction entry per reported result (identifiers
\resid{R-C1}--\resid{R-C15} and \resid{R-D0}--\resid{R-D60}; R-C8 onwards and R-D18 onwards are the results added in
revision, R-D30 onwards those of the pre-registered transfer tests, R-D37 onwards the public-task extension of Appendix~\ref{app:musique},
whose data are public, R-D41 onwards the cross-task screen and programming pilot of Appendix~\ref{app:tasks2}, on the public
EURLEX57K, BigCodeBench and KodCode datasets, R-D47 onwards the zero-shot transfer evaluation of Appendix~\ref{app:zeroshot}, on
the public CodiEsp-D and MuSiQue-Ans datasets, R-D52 onwards the evidence reads of Appendix~\ref{app:evidence}, on CodiEsp-X
and on MDACE, whose annotations are public and whose notes are \texttt{MIMIC-III}, and R-D58 onwards the further controls on \smol{} of Appendix~\ref{app:teacherpool}), each with its data, code, configuration, seeds, compute, command, expected artifact and tolerance. Every number in the paper is generated from the experiment artifacts by the
readers named there (Appendix~\ref{app:repro}). The data are public under credentialed access: any researcher can
obtain \mimiciv{} and \texttt{MIMIC-III} from PhysioNet, and public code rebuilds the splits we use (Appendix~\ref{app:data}). Reruns reproduce the reported statistics within the stated intervals but not bit-exactly:
decoding is stochastic and some kernels are nondeterministic (Appendix~\ref{app:repro}). An anonymised snapshot of the
code repository at this version is at \url{\anoncodeurl}: the code, configurations and tests, the planning and
walkthrough documents that record every run and result, and an index mapping each result identifier to the
script that produces it. Its \texttt{results/} folder lists every reported number with the statistic it is and the
artifact it was computed from, so each can be traced without access to the data. The run-by-run experiment logbook is withheld, since it records per-note details that
fall under the data use agreement.

\bibliographystyle{styles/iclr2027_conference}
\bibliography{bib/references,bib/references_custom}

\appendix

\section{Task, data and candidate presentation}
\label{app:data}
\supports{Sec.~\ref{sec:setup-task}, Sec.~\ref{sec:distill-setup}; \resid{R-C1}, \resid{R-D0}}

\paragraph{Data and access.} The data are \mimiciv{} discharge summaries, from its \texttt{MIMIC-IV-Note} module,
with their ICD-10 codes \citep{johnson2023mimicnote, johnson2023mimic}. Both are public under credentialed access:
any researcher can obtain them from PhysioNet after completing its credentialing, which includes a course in
human-subjects research, and signing the data use agreement; we use them under that agreement. ICD coding on
\texttt{MIMIC-III} and \mimiciv{} is a standard clinical NLP benchmark for multi-label classification over a large
label space \citep{mullenbach2018explainable, vu2020label, huang2022plm, edin2023automated}, and ICD-10 coding on
\mimiciv{} has been proposed as a benchmark for extreme multi-label classification \citep{nguyen2023mimic}. We use
the preprocessed ICD-10 release and splits of \citet{edin2023automated}, which their public code rebuilds from the
PhysioNet files (\mimiciv{} and \texttt{MIMIC-IV-Note} v2.2), so anyone with access can recreate them: 122{,}279
summaries from 65{,}659 patients, with predefined train, validation and test splits of 89{,}099, 13{,}378 and
19{,}802 summaries and text that is lower-cased and stripped of punctuation. Its label universe of 7{,}942 codes mixes
5{,}651 ICD-10-CM diagnosis codes and 2{,}291 ICD-10-PCS procedure codes. We assign diagnoses only, so the inventory
is the 5{,}651 diagnosis codes with their descriptions. No note text, reasoning-trace text or other protected content
appears in the paper, its figures, the supplementary material or the code snapshot: we report public ICD-10-CM codes,
attention-head names and statistics, and the one example code pair (N18.9 named after N18.1) is public.

\paragraph{Characterization corpus (TR-006).} Notes come from the test split. A note is eligible if it has
800--2{,}500 words and 5--25 diagnosis codes in the inventory. Selection is stratified over the four quartiles of that
code count and, within a stratum, greedy on ICD-10 blocks not yet covered, with at most one note per patient. The 94
notes cover 235 ICD-10 blocks and 21 of the 22 ICD-10-CM chapters, with 5--25 gold codes each (median 13). A note
enters the prompt without its discharge-diagnosis section, which names diagnoses in words; six traces are pilot
variants of six of the notes with that section kept, hence 100 traces over 94 notes. The trace is the unit of
analysis, and we call each trace a note; the characterization on these traces is HD-001, and its replication on the
retrieved shortlist is HD-002 (Appendix~\ref{app:hd002}). On these traces \teacher{} scores \microf{} \NMTeacherMicroFFullInventory{};
all 100 are valid (Appendix~\ref{app:teacher}), each several thousand tokens of reasoning followed by a JSON answer.

\paragraph{The full inventory in context.} The prompt lists all 5{,}651 diagnosis codes, one line per code with its
description, in a single order shuffled once with a fixed seed and shared by every note: the shared order makes the
inventory a cacheable prefix, and the shuffle removes ICD order as a positional cue. The note and the instruction
follow (Appendix~\ref{app:teacher}). A prompt is about 106k tokens (median over the 100 prompts), of which the inventory is
about 105k, about 18.5 tokens per line.
\emph{Chapter} in this paper means a code's first character, the chapter letter; the inventory lines sharing
the emitted letter are the emitted chapter, or region. The inventory uses 25 letters, which map imperfectly onto the
22 ICD-10-CM chapters: A--B, S--T and V--Y each form one chapter, and D and H are each split across two. Where a true
ICD-10-CM chapter is meant, as in ``21 of the 22 chapters'', we say so.

\paragraph{The retrieved shortlist (distillation half).} The students are distilled on a retrieved shortlist rather
than the full inventory (Sec.~\ref{sec:distill-setup}), and the teacher is re-characterized on it (HD-002,
Appendix~\ref{app:hd002}). The retriever (RT-003) embeds note chunks with a frozen gte-large encoder
\citep{li2023towards}. A supervised multi-label linear head, logistic regression on
[max;\,mean]-pooled chunk embeddings trained on the \mimiciv{} train split with the patients of both corpora excluded,
scores every code, and its ranking is fused with the zero-shot similarity between chunks and code descriptions. The
top $K=1{,}000$ codes are kept, and the category siblings (same three-character category) of the top 200 are added,
so that the near-misses \Ptwoc{} events need stay in the list. The construction is deterministic and identical at
training, validation and test. A pre-registered rule fixed $K$: the smallest $K$ whose completed recall reaches 0.9 on
both the validation notes and the distillation corpus. The corpus condition never bound: at every $K$ of the sweep the
corpus recalls more than the validation notes (at $K=1{,}000$, \NMCRetrieverRecallKThousandCorpus{} against
\NMCRetrieverRecallKThousandVal{} without completion, \NMDRetrieverRecallCorpus{} against \NMRetrieverRecallVal{} with
it), so the validation notes alone select the same $K$ and the test notes' labels did not affect it. The sweep
(code-level macro recall of the fused retriever on the validation notes and the 650-note corpus):
\begin{center}\small
\begin{tabular}{@{}l cc@{}}
\toprule
$K$ / completion & validation & corpus \\
\midrule
500 / none & \NMCRetrieverRecallKFiveHundredVal{} & \NMCRetrieverRecallKFiveHundredCorpus{} \\
1{,}000 / none & \NMCRetrieverRecallKThousandVal{} & \NMCRetrieverRecallKThousandCorpus{} \\
1{,}000 / siblings of the top 200 (chosen) & \NMRetrieverRecallVal{} & \NMDRetrieverRecallCorpus{} \\
\bottomrule
\end{tabular}
\end{center} A zero-shot encoder alone recalls
\NMCZeroShotEncoderRecallAtFiveHundredVal{} of the gold codes at $K=500$, hence the supervised stage. Macro
recall over notes is \NMRetrieverRecallVal{} on the 100 validation notes, which are the characterization notes, and
\NMRetrieverRecallTest{} on the 150 test notes; it is a ceiling every distilled arm shares. A shortlist has about
1{,}550 lines, and its prompts are 27--34k tokens (median 30k), about 3.5 times shorter than the full inventory's;
header, line format, note and instruction are otherwise byte-identical. The teacher's traces on these prompts (TR-007
for the validation notes, TR-008 for the 650-note distillation corpus) score \microf{}
\NMCTeacherMicroFShortlistVal{} on the validation notes (the 99 with a valid trace) and
\NMTeacherShortlistTestMicroF{} on the test notes.

\section{Teacher, prompts, decoding; the retired instrument}
\label{app:teacher}
\supports{Sec.~\ref{sec:setup-task}, Sec.~\ref{sec:setup-measure}; \resid{R-C1}, \resid{R-C2}}

\paragraph{Teacher and decoding.} The teacher is \teacher{} \citep{qwen2025qwen3}
(\texttt{Qwen/Qwen3-14B}; revision in Appendix~\ref{app:base-models}), 40 layers of 40 attention heads, run in bf16 in
its native thinking mode. Static YaRN rope scaling \citep{peng2024yarn} with factor 4 extends its
native context of 40{,}960 tokens to 163{,}840; the positional scaling is part of the characterized system. Traces are
sampled at temperature 0.6, top-$p$ 0.95 and top-$k$ 20, with a fixed seed and a cap of 16{,}384 new tokens. The
shortlist traces of Appendix~\ref{app:hd002} use the same settings.

\paragraph{Prompt.} We describe the prompt rather than print it. It consists of a header naming the list as the
assignable ICD-10-CM diagnosis codes, the candidate lines (the full inventory, or the shortlist), a header
introducing the discharge summary, the note, and a short instruction, all in one user turn of the model's chat
template with thinking enabled. The instruction gives the role of a clinical coder and the task of coding the summary;
it points to the complete list above and asks the model to narrow first to the clinically plausible region and then
decide among the specific codes in it; and it asks for the final answer as a JSON list of code strings. The
instruction therefore invites a coarse-to-fine procedure. The phases we measure are defined by the type of token-level
decision (Section~\ref{sec:setup-events}), not by this wording; the control below removes the wording and finds the same
structure.

\paragraph{Instruction control.} To test whether the phase structure is an artefact of the narrow-then-decide sentence,
we regenerated the traces of \NNeutralNotes{} of the 100 notes (\NNeutralPatients{} patients, drawn with a fixed seed) with
that one sentence removed and nothing else changed (teacher, decoding, inventory, layout and the answer instruction), and
ran the characterization pipeline on them unchanged (Table~\ref{tab:app-neutral}). Four readings were fixed before the
run. (1)~HD-001's \emph{original} global sets, knocked into the new traces, beat random on at least 90\% of the notes for
the chapter and contrastive decisions: they do on \NNeutralOrigPOneKiKEightWins{} and \NNeutralOrigPTwoCKiKEightWins{} notes
at $k=8$, with recoveries at $k=40$ that match HD-001's own on the same notes under the original prompt. (2)~The sets
re-ranked on the new traces share at least 20 of 40 heads with HD-001's: they share \NNeutralOverlapPOneForty{} (chapter)
and \NNeutralOverlapPTwoCForty{} (contrastive), and the contrastive top-8 is the same \NNeutralOverlapPTwoCEight{} heads.
(3)~A window locus exists for a comparable share of chapter mentions: \NNeutralNeutralLocusWindowPct{} against
\NNeutralViewLocusWindowPct{} on the same notes with the instruction, with fewer inherited loci. (4)~The prose-time top-8
shares at most two heads with the emission top-8: it shares \NNeutralProseEmissionShared, and
\NNeutralProseTopEightSharedHDOne{} of its eight heads are HD-001's prose-time heads. Coverage is unchanged
(\NNeutralNeutralThinkClosed{} of 30 traces close the reasoning block and are valid, median \NNeutralNeutralGenTokens{}
generated tokens against \NNeutralOrigGenTokens). The structure is the model's, not the prompt's.

\begin{table}[!htbp]
  \centering\small
  \caption{Instruction control: the teacher's traces on \NNeutralNotes{} notes with the ``narrow first, then decide''
  sentence removed, read with the HD-001 pipeline. Knock-in recovery of HD-001's original global sets and of the sets
  re-ranked on the new traces (medians over notes; random-$k$ in parentheses; notes where the global set beats random),
  against HD-001's own cells on the same notes under the original prompt; the prose locus on both. \resid{R-C14}}
  \label{tab:app-neutral}
  \providecommand{\gensd}[2]{\begin{tabular}[t]{@{}c@{}}#1\\[-2pt]{\scriptsize$\pm$#2}\end{tabular}}
\providecommand{\genstack}[2]{\begin{tabular}[t]{@{}c@{}}#1\\[-2pt]{\scriptsize #2}\end{tabular}}
\providecommand{\genstackl}[2]{\begin{tabular}[t]{@{}l@{}}#1\\[-2pt]{\scriptsize #2}\end{tabular}}
\providecommand{\genhead}[2]{\begin{tabular}[b]{@{}c@{}}#1\\#2\end{tabular}}
\providecommand{\genstackiii}[3]{\begin{tabular}[t]{@{}c@{}}#1\\[-2pt]{\scriptsize #2}\\[-3pt]{\scriptsize #3}\end{tabular}}
\footnotesize\setlength{\tabcolsep}{3.5pt}
\begin{adjustbox}{max width=\linewidth}
\begin{tabular}{@{}l r cc cc c@{}}
\toprule
 & & \multicolumn{2}{c}{$k=8$} & \multicolumn{2}{c}{$k=40$} & original prompt, \\
\cmidrule(lr){3-4}\cmidrule(lr){5-6}
Event & events & recovery (random) & notes & recovery (random) & notes & same notes, $k=40$ \\
\midrule
\multicolumn{7}{@{}l}{\textit{HD-001's original global sets knocked into the neutral-prompt traces (30 notes): recovery, median over notes (random-$k$); notes where global $>$ random}} \\
\Pone & 599 & 0.66\,{\scriptsize(0.01)} & 30/30 & 0.89\,{\scriptsize(0.04)} & 30/30 & 0.88 \\
\Ptwo & 1{,}366 & 0.48\,{\scriptsize(0.01)} & 30/30 & 0.80\,{\scriptsize(0.02)} & 30/30 & 0.78 \\
\Ptwoc & 433 & 0.72\,{\scriptsize(0.00)} & 28/28 & 0.83\,{\scriptsize(0.01)} & 28/28 & 0.86 \\
\midrule
\multicolumn{7}{@{}l}{\textit{Sets re-ranked on the neutral traces: overlap with HD-001's rankings, and their own knock-in at $k=40$}} \\
\Pone &  & \multicolumn{2}{l}{top-40 shared 33/40; top-8 shared 7/8} & \multicolumn{2}{l}{knock-in $k=40$ 0.89\,{\scriptsize(0.04)}} & 0.88 \\
\Ptwoc &  & \multicolumn{2}{l}{top-40 shared 31/40; top-8 shared 8/8} & \multicolumn{2}{l}{knock-in $k=40$ 0.90\,{\scriptsize(0.01)}} & 0.86 \\
\midrule
\multicolumn{7}{@{}l}{\textit{Prose locus on the same 30 notes}} \\
window share / inherited / probe failure &  & \multicolumn{2}{l}{neutral: 47\% / 17\% / 32\% (median $d$ 6)} & \multicolumn{3}{l}{original prompt: 38\% / 26\% / 33\% (median $d$ 7)} \\
prose-time top-8 $\cap$ emission top-8 &  & \multicolumn{5}{l}{0 of 8 (HD-001: 0 of 8); 6 of the 8 prose-time heads are HD-001's} \\
\bottomrule
\end{tabular}
\end{adjustbox}

\end{table}

\paragraph{Validity and parsing.} A generation is valid when it closes its reasoning block before the cap. Its answer
is the last JSON list after that block, restricted to inventory codes and de-duplicated; a truncated generation is
invalid, is never scored, and its reasoning is never scraped for codes. A guard also flags degenerate generations:
at least 10 codes of which more than 90\% follow in ascending code order (nothing in the shuffled list suggests
sorting), more than 50 codes, or a generation that reaches the cap. All 100 characterization traces are valid and
none is degenerate. Sampling is seeded, but forward passes are not bit-exact on our hardware
(Appendix~\ref{app:instrument}), so a regenerated trace can differ; every measurement re-scores the stored traces by
teacher forcing.

\paragraph{The retired anchor-cosine instrument.} Our first Phase-1 instrument asked whether a head's write points
toward a direction fixed in advance. It scored a head by the cosine between its output at an event and an anchor, the
centroid of the embeddings of the note's gold codes, weighted by how concentrated the head's attention is and
averaged over events. Four runs on the same 100 traces (CS-001 to CS-004) retired it.
(i)~The anchor is not a well-formed object on real notes. Every note's gold codes span more than one ICD-10-CM
chapter (median \NMCGoldChaptersPerNoteMedian{}), and the score passes
\NMCAnchorCentroidNotesPassing{} of 100 notes against 5 expected by chance. A per-concept maximum over the
gold codes passes \NMCAnchorPerConceptNotesPassing{} of 100 notes, but diffusely:
\NMCAnchorPerConceptShiftedHeads{} heads shift together and none recurs across notes.
(ii)~Every head has a stable, idiosyncratic alignment with any fixed direction, so tests on the cosine cannot be
calibrated. Under a complete null, the per-note test (Benjamini--Hochberg at FDR 0.10 over heads) returned a
non-empty set in \NMCAnchorNullNonemptyTrials{} of 800 trials, and a head's null statistic kept its sign
across a note's mentions \NMCAnchorNullSignConsistency{} of the time.
(iii)~The positive control, an anchor on the code being emitted, passes
\NMCAnchorPositiveControlNotesTwoHundredDraws{} of 100 notes against a 200-draw null but
\NMCAnchorPositiveControlNotesTwentyDraws{} against the earlier 20-draw null: the size of the null matters.
(iv)~The one robust readout, recurrence of a head in the notes' top-20, found two heads: \head{27}{17}, in
\NMCLTwentySevenHSeventeenRecurrenceNotes{} of 100 notes, and \head{11}{10}. By chapter $I_c$ they rank
\NMCLTwentySevenHSeventeenRankPOne{} and \NMCLElevenHTenRankPOne{} of 1{,}600. Their recurrence is real as a correlation,
but it is not about the decision.
The replacement measures each head's effect on the model's own decision margin and validates every set by
intervention (Section~\ref{sec:setup-measure}, Appendix~\ref{app:instrument}).

\paragraph{Identifiability of sets fitted to one note.} A note has a median of 19 chapter decisions, one per first
mention. CS-004 asked whether a head set can be fitted to a single note from that many events. Under the
anchor-cosine statistic, sets fitted to one note (Benjamini--Hochberg at FDR 0.10 over 1{,}600 heads) were empty on
the median note and did not reproduce between the two halves of a note's mentions (median Jaccard index
\NMCPerNoteSplitHalfJaccard{}), and the most recurrent head cleared the first Benjamini--Hochberg
threshold on \NMCPerNoteBHPassNotesLTwentySevenHSeventeen{} of 100 notes: with about 19 events, a head must lie several
standard errors from zero to survive a search over 1{,}600 heads. The causal instrument agrees where it can be
checked. A ranking fitted to a note's own chapter events carries that note's chapter decision into another context
less well than the global ranking at $k\ge 8$, whereas for character decisions, with about 70 events per note, the
note's own ranking is the better carrier (Appendix~\ref{app:heads}). We therefore use global sets and validate them
on every note.

\section{The instrument in detail}
\label{app:instrument}
\supports{Sec.~\ref{sec:setup-events}, Sec.~\ref{sec:setup-measure}; \resid{R-C1}, \resid{R-C2}}

Figure~\ref{fig:events-full} shows, on the code pair of Section~\ref{sec:setup-events}, how a coding decision becomes a
token-level event, what each event type is, and where the margin that scores it is read.

\begin{figure*}[!htbp]
  \centering
  \input{sections/fig_events.tex}
  \caption{How a coding decision becomes a token-level event (illustrative: public ICD-10-CM codes and their inventory
  descriptions; the note and the reasoning prose are placeholders). The prompt holds the note and all 5{,}651 inventory
  lines; the region of a chapter letter is every line whose code starts with it. Writing \texttt{N18.9} after naming
  \texttt{N18.1}, the teacher decides one token at a time: the chapter letter (\Pone, Phase~1), each later character but the separator
  (\Ptwo, Phase~2), and the character at which the code diverges from its near-miss (\Ptwoc). \Pnull events, prose
  positions with a random target letter, calibrate the head rankings. Each event is scored by the model's own margin against its competitors (at \Ptwoc{}, the
  near-miss's character alone) at the position $q$ that emits the token; a head's importance is the margin it removes when mean-ablated there
  (Sec.~\ref{sec:setup-measure}). The locus (Appendix~\ref{app:locus}) is the earliest position from which the forced
  chapter margin stays at least 2 nats. \emph{Bottom:} testing a set of heads. For necessity the set's outputs at $q$
  are replaced by their means over the note. For the knock-in we take an event of the same kind in another patient's
  note, where the model wrote a different token, put in the clean note's token at $q$ and the set's outputs from the
  clean run, and read the margin of the clean decision. The two bars show the chapter top-8
  (Table~\ref{tab:faithfulness}). \resid{R-C1}}
  \label{fig:events-full}
\end{figure*}

\paragraph{Event extraction.} Codes are located in a trace by pattern and kept if they are inventory codes.
\teacher{}'s tokenizer writes a code as a chapter-letter token followed by one token per character; each of the 25
letters and 36 characters that inventory codes use is a single token both with a leading space and without, and a
letter token may carry the space. A mention is scored only if each of its characters is its own token in context,
which excludes 67 of the 2{,}131 first mentions (median 19 per note). At every event, $q$ is the input position whose
output is the decided token, the target is the emitted token, and the competitors depend on the kind
(Table~\ref{tab:events-app}):
\begin{itemize}[leftmargin=*,itemsep=1pt,topsep=2pt]
  \item \Pone{}: the chapter letter at the first mention of each inventory code; the competitors are the other
  inventory letters in the same spacing variant. \NMCFirstMentionsSiblingPct{} of first mentions are
  siblings of an earlier mention.
  \item \Ptwo{}: each later character of a first mention (the separator ``.'' is not scored); the competitors are the
  characters that continue the prefix to another inventory code, and a character without any is skipped.
  \item \Ptwoc{}: at a mention of a code whose sibling (same three-character category) the trace named earlier, the
  first character at which the two codes differ. The single competitor is the sibling's character, so the margin is
  the log-odds against the near-miss. The event records the positions of the earlier mentions of the code ($S$) and
  of the near-miss ($N$). Events are capped at two per (code, near-miss) pair, the first mention and the first repeat
  that has earlier mentions of the code: uncapped, repeats gave 9{,}474 events (1{,}098 in one note), repeats of the
  same decision rather than new ones; capped, 3{,}855, of which 2{,}208 have earlier mentions of the code.
  \item \Pnull{}: 20 positions per trace drawn at random from the prose at least 8 tokens from any code token, each
  with a random letter as the target and the other letters as competitors.
\end{itemize}
The faithfulness pass uses every \Pone{} event and a seeded subset of at most 50 \Ptwo{} and 40 \Ptwoc{} events per
note; 92 notes have a \Ptwoc{} event.

\paragraph{What the earlier sibling mentions are.} A \Ptwoc{} event is defined by mention order: the sibling was named
earlier in the trace, for whatever reason. To say what those earlier mentions are, we audited a stratified random sample
of \NAuditN{} contrastive events of the teacher's traces on the 150 test notes (eight strata by tokenizer alignment, gold
status and whether the emitted code reaches the final answer), reading the trace context around the earlier mention and
around the decision; the author, who holds the data credential, labeled each with a written rule, and only aggregate
shares leave the enclave (Table~\ref{tab:app-p2caudit}). In \NAuditRejectedPct{} of the sample (\NAuditRejectedWeighted{}
population-weighted) the earlier mention is a \emph{verified rejection}: the model considered the sibling for the same
condition and set it aside in favour of the emitted code. In \NAuditEnumerationPct{} (\NAuditEnumerationWeighted) it is an
\emph{enumeration}: the sibling is a neighbouring line in a scan or reproduction of the candidate block, not a considered
alternative. \NAuditBothPct{} endorse both codes and \NAuditOtherPct{} are other cases (the sibling endorsed and the emitted
code only enumerated, hedged pairs). The rejection share rises to \NAuditRejectedInAnswerPct{} among events whose emitted
code reaches the final answer and falls to \NAuditRejectedNotInAnswerPct{} among those whose code does not. The contrast is
clinically meaningful in \NAuditMeaningfulYesPct{} of the items (the exceptions are BMI bands, place-of-occurrence codes and
``other'' against ``unspecified'' residuals). We therefore call the event by what it is, the character at which the emitted
code diverges from a sibling mentioned earlier, and report the mechanism measurements on the verified rejections and on the
rest separately, for M1 and the own-heads top-40 recovery of the \qwenfour{} arms (verified rejections: $n$ \NAuditMOneRejectedN{}
events for M1 and \NAuditMFourRejectedN{} patched events for the knock-in; other labeled events, which are not only
enumerations: \NAuditMOneOtherN{} and \NAuditMFourOtherN; Appendix~\ref{app:readside}, Table~\ref{tab:app-audit-reads}).

\begin{table}[!htbp]
  \centering\small
  \caption{The audit of \NAuditN{} contrastive events: what the earlier mention of the sibling is. Raw shares with
  binomial 95\% intervals, and shares weighted by the eight strata's population sizes. \resid{R-C9}}
  \label{tab:app-p2caudit}
  \providecommand{\gensd}[2]{\begin{tabular}[t]{@{}c@{}}#1\\[-2pt]{\scriptsize$\pm$#2}\end{tabular}}
\providecommand{\genstack}[2]{\begin{tabular}[t]{@{}c@{}}#1\\[-2pt]{\scriptsize #2}\end{tabular}}
\providecommand{\genstackl}[2]{\begin{tabular}[t]{@{}l@{}}#1\\[-2pt]{\scriptsize #2}\end{tabular}}
\providecommand{\genhead}[2]{\begin{tabular}[b]{@{}c@{}}#1\\#2\end{tabular}}
\providecommand{\genstackiii}[3]{\begin{tabular}[t]{@{}c@{}}#1\\[-2pt]{\scriptsize #2}\\[-3pt]{\scriptsize #3}\end{tabular}}
\footnotesize
\begin{tabular}{@{}l c c c@{}}
\toprule
Type of the earlier sibling mention & items & raw share [95\% CI] & population-weighted \\
\midrule
rejected alternative & 64/181 & 35\% [28\%, 43\%] & 30\% \\
enumeration / list scan & 99/181 & 55\% [47\%, 62\%] & 62\% \\
both endorsed & 5/181 & 3\% [1\%, 6\%] & 3\% \\
other & 13/181 & 7\% [4\%, 12\%] & 6\% \\
\bottomrule
\end{tabular}

\end{table}

\begin{table}[!htbp]
  \centering\small
  \caption{Decision events over the 100 characterization traces: all events, events in the faithfulness pass (with
  the number of notes that have them), and the median decision margin (nats). \Pone{} and \Ptwo{} events are taken at
  the first mention of each inventory code, \Ptwoc{} events at the first mention and the first repeat of each (code,
  near-miss) pair. \resid{R-C1}}
  \label{tab:events-app}
  \providecommand{\gensd}[2]{\begin{tabular}[t]{@{}c@{}}#1\\[-2pt]{\scriptsize$\pm$#2}\end{tabular}}
\providecommand{\genstack}[2]{\begin{tabular}[t]{@{}c@{}}#1\\[-2pt]{\scriptsize #2}\end{tabular}}
\providecommand{\genstackl}[2]{\begin{tabular}[t]{@{}l@{}}#1\\[-2pt]{\scriptsize #2}\end{tabular}}
\providecommand{\genhead}[2]{\begin{tabular}[b]{@{}c@{}}#1\\#2\end{tabular}}
\providecommand{\genstackiii}[3]{\begin{tabular}[t]{@{}c@{}}#1\\[-2pt]{\scriptsize #2}\\[-3pt]{\scriptsize #3}\end{tabular}}
\begin{tabular}{@{}llrrr@{}}
\toprule
Event & Decision & \genhead{Events}{(100 traces)} & \genhead{Faithfulness}{pass (notes)} & \genhead{Median}{margin} \\
\midrule
\Pone & chapter letter (region of the label space) & 2{,}064 & 2{,}064 (100) & +6.02 \\
\Ptwo & each character (code within the prefix) & 7{,}688 & 4{,}587 (100) & +2.76 \\
\Ptwoc & character diverging from a named sibling & 3{,}855 & 1{,}959 (92) & +3.38 \\
\Pnull & prose position, random target (calibration) & 2{,}000 & --- & \ensuremath{-}4.95 \\
\bottomrule
\end{tabular}

\end{table}

\paragraph{Decision margin, and the audit at the code token.}
$\margin(q)=\mathrm{logit}(\text{target})-\logsumexp(\text{competitors})$ is the log-odds the model gives the emitted
token against its alternatives, a multi-way form of the usual patching metric \citep{wang2023interpretability,
zhang_towards_2024}; taken over all other letters, the \Pone{} margin is the log-odds of this chapter against any
other. The entropy over the target and its competitors says how close the call was. Before any attribution, an audit
read the margins at every event (Table~\ref{tab:app-audit}). At the code token the chapter margin is usually already decisive;
character decisions are far more often live, and some are sampled against the model's favourite; at
\Pnull{} events a random letter sits far below zero at high entropy, as it should.

\begin{table}[!htbp]
  \centering\small
  \caption{Decision margins at the events of all 100 traces (nats): 10th percentile, median and 90th percentile;
  median entropy over the target and its competitors; share of events with a margin below 5 nats and below zero.
  \resid{R-C1}}
  \label{tab:app-audit}
  \begin{adjustbox}{max width=\linewidth}
  \begin{tabular}{@{}lcccccc@{}}
    \toprule
    Event & p10 & median & p90 & entropy & margin $<5$ & margin $<0$ \\
    \midrule
    \Pone & \NMCPOneMarginPTen{} & \NEvPOneMargin & \NMCPOneMarginPNinety{} & \NMCPOneEntropyMedian{}
      & \NMCPOneMarginBelowFivePct{} & \NMCPOneMarginNegativePct{} \\
    \Ptwo & \NMCPTwoMarginPTen{} & \NEvPTwoMargin & \NMCPTwoMarginPNinety{} & \NMCPTwoEntropyMedian{}
      & \NMCPTwoMarginBelowFivePct{} & \NMCPTwoMarginNegativePct{} \\
    \Ptwoc & \NMCPTwoCMarginPTen{} & \NEvPTwoCMargin & \NMCPTwoCMarginPNinety{}
      & \NMCPTwoCEntropyMedian{} & \NMCPTwoCMarginBelowFivePct{} & \NMCPTwoCMarginNegativePct{} \\
    \Pnull & \NMCNullMarginPTen{} & \NEvNullMargin & \NMCNullMarginPNinety{}
      & \NMCNullEntropyMedian{} & \NMCNullMarginBelowFivePct{} & \NMCNullMarginNegativePct{} \\
    \bottomrule
  \end{tabular}
  \end{adjustbox}
\end{table}

\paragraph{The hook and mean-ablation.} Each attention layer's output projection takes the concatenation of its 40
heads' outputs. One forward pre-hook on that input can (i) accumulate each head's mean output, (ii) expose every
head's output as a gradient leaf, (iii) replace one head's slice at chosen positions by its mean (mean-ablation), or
(iv) replace it by values recorded in another run (knock-in). The unit of intervention is one head's output at one
position. The mean $\mu_h$ is taken over every position of the note's own prompt and trace, during the single prefill
that also builds the key--value cache; replacing a head's output by its mean removes the head's information while
keeping its typical contribution to the residual stream, whereas zeroing would move the model off its distribution
\citep{wang2023interpretability, heimersheim_how_2024}. Events are visited in decreasing position, so the cache is
only ever cropped, and each measurement is a single-token forward, or a forward and a backward at one position.
Before any run the hook was validated: with no intervention active the hooked forward is byte-identical to a plain
one; zeroing all heads of a layer through the hook equals zeroing that layer's attention output; on a small randomly
initialized model with the same code path, knocking in all heads reproduces the clean margin at every event; and in
the locus pass the gradient-enabled forward reproduced the plain forward's margins exactly.
$I_h(q)=\margin(q)-\margin(q\mid h\ \text{mean-ablated})$ is positive when the head supports the emitted choice and
negative when it opposes it. A set's necessity is the share of the clean margin destroyed when the set is
mean-ablated at $q$ (above 1, the decision flips).

\paragraph{Attribution as a screen, and its fidelity.} First-order attribution estimates the effect of mean-ablating
head $h$ as $A_h=(\mu_h-z_h)\cdot\partial\margin/\partial z_h$ for all 1{,}600 heads from one backward pass
\citep{syed_attribution_2024, kramar_atp_2024}. We measured its fidelity on the task before relying on it: at ten
chapter decisions of one note, every head was also patched exactly (Table~\ref{tab:app-selftest}). The hook of the
attribution pass detached the gradient leaf from the graph at every layer. That cut every path through later
attention layers (a head's effect through later queries, keys and values) and kept only each head's direct path.
Under that hook, attribution correlates \NMAttrPearsonAllDetached{} with the exact effect over all heads and
\NMAttrPearsonAboveNoiseDetached{} over heads above the noise floor, and its top-20 recovers
\NMCAttrTopTwentyRecoveredDetached{} of the exact top-20's effect: a screen, not a measurement. The defect
surfaced when the window attribution of Appendix~\ref{app:locus} came back zero away from the last position, and it
was fixed before that pass. The same self-test with the graph intact gives \NMAttrPearsonAllIntact{}
(\NMAttrPearsonAboveNoiseIntact{} above noise), Spearman \NMCAttrSpearmanIntact{},
\NMCAttrTopTwentyOverlapIntact{} of the exact top-20, and \NMCAttrTopTwentyRecoveredIntact{} of its
effect. Integrated gradients is worse under both hooks (\NMCIGPearsonIntact{} with the graph intact),
presumably because interpolating every layer at once moves the model off-distribution. Direct-logit attribution
\citep{elhage2021mathematical}, the projection of a head's write onto the margin direction in the unembedding, is
uncorrelated with the exact effect in the self-test and anticorrelated with first-order attribution in the full run
(median per-event Pearson \NMCDLAVsFirstOrderPearsonMedian{}). Because the distillation objective's interaction term uses
exactly this projection as its contribution quantity $c$ (Section~\ref{sec:distill-objective}), we measured its
faithfulness on the \NProxyEvents{} contrastive events with exact single-head patches: across the patched heads of an
event, $c$ correlates with the exact importance at Spearman \NProxyCSpearmanEvent{} (median over events; Pearson
\NProxyCPearsonEvent) and, pooled over events, at \NProxyCSpearmanPooled{} across \NProxyHeads{} heads, against
\NProxyASpearmanEvent{} and \NProxyASpearmanPooled{} for first-order attribution; the margin-weighted form $m\,c$ is
slightly lower (\NProxyMCSpearmanEvent{} / \NProxyMCSpearmanPooled), and its magnitude is about a quarter of the exact
effect (\NProxyAbsRatio). The proxy is non-zero only for heads whose effect runs through their direct path: it ranks the
final-layer heads \head{39}{12} and \head{39}{10} correctly ($c$ \NProxyHeadLThirtyNineHTwelveC{} and
\NProxyHeadLThirtyNineHTenC{} against exact importances \NProxyHeadLThirtyNineHTwelveI{} and \NProxyHeadLThirtyNineHTenI)
and assigns almost nothing to mid-layer heads such as \head{30}{11} (\NProxyHeadLThirtyHElevenC{} against
\NProxyHeadLThirtyHElevenI). $c$ is therefore a training-time proxy for the late direct path, not a causal estimate; the
empirical justification of the interaction term is the interaction-only ablation of Appendix~\ref{app:ablations}.

What each ranking rests on: \Pone{} and \Pnull{} importances are exact for the nominated heads (next paragraph) and
first-order elsewhere; \Ptwo{} importances are first-order under the detached hook, and a re-run of the attribution with the
intact hook and exact patches of the \Ptwo{} nominees on 50 notes keeps the ranking in substance (below,
Table~\ref{tab:app-p2intact}); the \Ptwoc{} global set is first-order under the detached hook and was refined by exact patching of 56
nominated heads at every \Ptwoc{} event of the faithfulness pass, where first-order and exact effects correlate at a
per-event median of \NMCPTwoCFirstOrderExactPearsonMedian{} (Appendix~\ref{app:heads}); the prose-time
ranking is first-order under the intact hook (Appendix~\ref{app:locus}). Every set the paper reports is judged by
exact intervention.

\begin{table}[!htbp]
  \centering\small
  \caption{Fidelity of attribution against exact patching: one note, ten chapter decisions, all 1{,}600 heads patched
  exactly at each. \emph{Detached}: the hook of the attribution pass, which kept only each head's direct path;
  \emph{intact}: the corrected hook, used at the locus. Integrated gradients: 5 steps. \resid{R-C2}}
  \label{tab:app-selftest}
  \begin{adjustbox}{max width=\linewidth}
  \begin{tabular}{@{}lcccc@{}}
    \toprule
     & \multicolumn{2}{c}{First-order} & Integrated gradients & Direct-logit \\
    \cmidrule(lr){2-3}
     & detached & intact & detached & attribution \\
    \midrule
    Pearson with the exact effect, all heads & \NMAttrPearsonAllDetached & \NMAttrPearsonAllIntact
      & \NMCIGPearsonAllDetached{} & $\approx 0$ (\NMCDLAPearsonSelftestMin{} to \NMCDLAPearsonSelftestMax{}) \\
    Pearson, heads above the noise floor & \NMAttrPearsonAboveNoiseDetached & \NMAttrPearsonAboveNoiseIntact
      & \NMCIGPearsonAboveNoiseDetached{} & --- \\
    Exact top-20 effect recovered by the top-20 & \NMCAttrTopTwentyRecoveredDetached{}
      & \NMCAttrTopTwentyRecoveredIntact{} & \NMCIGTopTwentyRecoveredDetached{} & --- \\
    \bottomrule
  \end{tabular}
  \end{adjustbox}
\end{table}

\paragraph{Exact patching of the nominated heads.} At every \Pone{} event and at a seeded half of the \Pnull{} events,
the attribution pass mean-ablates exactly the top-40 heads by $|A_h|$ together with the running global top-16 (the
heads with the largest mean first-order importance over the chapter events of the notes already processed): 48
exact patches per chapter event on average, and 26 per \Pnull{} event averaged over all of them. A head's importance
is its exact effect where measured and the first-order estimate elsewhere. \Ptwo{} and \Ptwoc{} events received no
exact patches in this pass; the faithfulness pass later patched the same 56 nominated heads exactly at every \Ptwoc{}
event of its subset.

\paragraph{The character ranking under the intact hook.} The \Ptwo{} ranking of Table~\ref{tab:heads-app} was computed
under the detached hook and never refined. We re-ran the attribution pass with the corrected hook on the first
\NPTwoIntactNotes{} characterization notes, this time patching the \Ptwo{} nominees exactly (the event's top-24 by $|A_h|$ and
the running global top-8), and compared it with the paper's cut-graph ranking recomputed on the same notes
(Table~\ref{tab:app-p2intact}). The two rankings share \NPTwoIntactPTwoTopForty{} of their top-40 and
\NPTwoIntactPTwoTopEight{} of their top-8 (Spearman \NPTwoIntactPTwoSpearman{} over 1{,}600 heads), and the heads the paper
names persist: \head{39}{10} at rank \NPTwoIntactHeadLThirtyNineHTenIntactRank{} (cut graph
\NPTwoIntactHeadLThirtyNineHTenCutRank), \head{39}{29} at \NPTwoIntactHeadLThirtyNineHTwentyNineIntactRank{}
(\NPTwoIntactHeadLThirtyNineHTwentyNineCutRank), \head{39}{12} at \NPTwoIntactHeadLThirtyNineHTwelveIntactRank{}
(\NPTwoIntactHeadLThirtyNineHTwelveCutRank) and \head{37}{6} at \NPTwoIntactHeadLThirtySevenHSixIntactRank{}
(\NPTwoIntactHeadLThirtySevenHSixCutRank). The intact graph promotes two mid-layer heads, \head{27}{11} and \head{27}{13},
whose character-level effect runs through later layers, the path the detached hook could not see; both carry positive
exact importance. Within the intact run the first-order and the exact-refined rankings agree (Spearman
\NPTwoIntactFirstOrderVsExactSpearman, \NPTwoIntactFirstOrderVsExactTopForty{} of 40). The chapter and contrastive rankings
move by similar amounts on the same 50 notes (\NPTwoIntactPOneTopForty{} and \NPTwoIntactPTwoCTopForty{} of 40 shared), and a
split of the full corpus into two halves moves a ranking by about as much (Appendix~\ref{app:heads}), so the difference is
the hook plus the 50-note sample, not a defect specific to the character ranking. No student used the teacher's \Ptwo{}
ranking directly (the students' pools are their own, Appendix~\ref{app:objective}).

\begin{table}[!htbp]
  \centering\small
  \caption{The attribution pass with the intact hook on \NPTwoIntactNotes{} notes against the paper's detached-hook
  rankings on the same notes: rank correlation and top-$k$ overlaps per phase, and the intact \Ptwo{} top-8 with each
  head's mean exact importance over the character events where it was patched. \resid{R-C8}}
  \label{tab:app-p2intact}
  \providecommand{\gensd}[2]{\begin{tabular}[t]{@{}c@{}}#1\\[-2pt]{\scriptsize$\pm$#2}\end{tabular}}
\providecommand{\genstack}[2]{\begin{tabular}[t]{@{}c@{}}#1\\[-2pt]{\scriptsize #2}\end{tabular}}
\providecommand{\genstackl}[2]{\begin{tabular}[t]{@{}l@{}}#1\\[-2pt]{\scriptsize #2}\end{tabular}}
\providecommand{\genhead}[2]{\begin{tabular}[b]{@{}c@{}}#1\\#2\end{tabular}}
\providecommand{\genstackiii}[3]{\begin{tabular}[t]{@{}c@{}}#1\\[-2pt]{\scriptsize #2}\\[-3pt]{\scriptsize #3}\end{tabular}}
\footnotesize
\begin{tabular}{@{}l c ccc@{}}
\toprule
Phase & Spearman & top-8 & top-16 & top-40 \\
\midrule
\Ptwo & 0.68 & 5/8 & 10/16 & 30/40 \\
\Pone & 0.65 & 6/8 & 12/16 & 33/40 \\
\Ptwoc & 0.53 & 7/8 & 14/16 & 31/40 \\
\midrule
\multicolumn{5}{@{}l}{\textit{\Ptwo{} top-8 under the intact hook, with exact standing: head (mean exact importance, nats; share of \Ptwo{} events patched)}} \\
\multicolumn{5}{@{}p{0.95\linewidth}}{\head{39}{38} (+0.44; 98\%), \head{39}{10} (+0.77; 54\%), \head{39}{12} (+0.57; 35\%), \head{39}{29} (+0.41; 19\%), \head{39}{37} (+0.74; 17\%), \head{27}{11} (+0.21; 98\%), \head{38}{24} (+0.48; 0\%), \head{27}{13} (+0.18; 97\%)} \\
\bottomrule
\end{tabular}

\end{table}

\paragraph{Noise floor.} The forward pass is not deterministic on this hardware: six identical single-token forwards
give a margin standard deviation of \NMNoiseFloorNats{} nats (range \NMCNoiseFloorRange{}). Every exact
effect is a difference of two such forwards, so single-head effects below about three standard deviations
(\NMCNoiseFloorThreeSd{} nats) cannot be told apart from zero. In the self-test, between
\NMCHeadsAboveFloorPerEventMin{} and \NMCHeadsAboveFloorPerEventMax{} heads per event exceed
that floor: a decision is influenced by many heads, most of them weakly. The paper therefore reports ranked sets and
means over notes, never single-head effects near the floor.

\paragraph{Knock-in (sufficiency by denoising).} Ablation shows that a set is necessary; sufficiency needs a context in
which the model makes a different decision \citep{wang2023interpretability, heimersheim_how_2024}. For each note, the
corrupted run is the next note in corpus order (cyclically), a different patient, at a matched event: the same kind
of event with a different decision (a different chapter letter at \Pone, a different character at \Ptwo{} and
\Ptwoc), preferring the same input token at $q$, then the same character depth, then the later position. The donor's
input token at the matched position is replaced by the clean note's token at $q$, the clean run's per-head outputs at
$q$ are written in for the heads of the set, and the clean decision's margin is read. The donor event shared the input
token at $q$ for \NMCDonorTokenMatchPOne{}, \NMCDonorTokenMatchPTwo{} and
\NMCDonorTokenMatchPTwoC{} of \Pone, \Ptwo{} and \Ptwoc{} events, and the character depth for
\NMCDonorDepthMatchPOne{}, \NMCDonorDepthMatchPTwo{} and \NMCDonorDepthMatchPTwoC{}; since the
clean token is always the one run, the ceiling below is exact either way. Writing in all 1{,}600 heads reproduces the
clean residual stream at $q$, so the MLPs see exactly their clean inputs and the all-heads knock-in must equal the
clean margin. It does, to the noise floor, on every event of every kind (median difference \NMKnockinCeilingDiff{}
nats; 10th--90th percentile \NMCKnockinCeilingPTen{} to \NMCKnockinCeilingPNinety{}). This checks the wiring and sets the scale,
\[
  \text{recovery}(\mathcal{S})=\frac{\margin_{\mathcal{S}}-\margin_{\text{corr}}}{\margin_{\text{all}}-\margin_{\text{corr}}},
\]
which is 0 when the other patient's decision stands and 1 when the set alone recovers the clean one. Events whose
range $\margin_{\text{all}}-\margin_{\text{corr}}$ is under 1 nat are excluded, because the corruption produced no
different decision there: \NMCKiExcludedPOne{} of chapter, \NMCKiExcludedPTwo{} of character and
\NMCKiExcludedPTwoC{} of contrastive events. The corrupted context favours another chapter by a median of
\NFaithPOneCorrMargin{} nats (it is against the clean chapter on \NMCCorruptedNegativePctPOne{} of events) and
another character by \NFaithPTwoCorrMargin{} nats (\NMCCorruptedNegativePctPTwo{}). At \Ptwoc{} events it is
close to indifferent (\NFaithPTwoCCorrMargin{} nats; against the clean character on
\NMCCorruptedNegativePctPTwoC{}), so contrastive knock-in measures how far a set moves an undecided context,
over a median range of \NMCKiRangePTwoCMedian{} nats.

\paragraph{\Pnull{} events and calibration.} A ranking of 1{,}600 heads always has a top; the question is whether the
top reflects the decision or the procedure. \Pnull{} events run the identical procedure where nothing is decided.
Raw importance fails this check: Benjamini--Hochberg at FDR 0.05 declares \NMNullRawBhPos{} heads positive and
\NMNullRawBhNeg{} negative at \Pnull{} events, and the null top-16 effects average
\NMCNullTopSixteenMeanEffect{} nats against \NMCPOneTopSixteenMeanEffect{} at chapter events. Some heads have
a generic effect on the letter logits, sharpening or flattening the distribution, and a generic effect moves any
letter margin at any position, consistently. Decision-specific importance $I_c=I-I_{\Pnull}$ subtracts it per note:
a head's mean importance over the note's events of a kind minus its mean over the note's \Pnull{} events, then
averaged over notes. The same check on $I_c$, one half of the \Pnull{} events against the other, gives
\NMNullSplitBhDiscoveries{} discoveries in either direction. The generic heads and the decision heads are different
heads: the raw and null rankings are unrelated (Spearman \NMCRawNullSpearman{}; they share
\NMCRawNullOverlapTopSixteen{} of their top-16 and \NMCRawNullOverlapTopForty{} of their top-40), and the
raw top-10 keep their ranks under $I_c$. For the chapter decision, \NHeadPOneBhPos{} heads have $I_c>0$ and
\NMCPOneBHNeg{} have $I_c<0$ at FDR 0.05: consistent effects in both directions, not a sparse set; sparsity is
a question of magnitude (Appendix~\ref{app:heads}). A phase's ranking by $I_c$ is its global set, and its pooled
weights are $w_h=\operatorname{softmax}(I_c/\tau)$ over the top-$k$, with $\tau$ the standard deviation of the top-$k$
importances.

\paragraph{Statistics.} The unit of every test is the note, because events within a note are not independent. Per
note, a head's importance for a kind is its mean over the note's events of that kind; its global importance is the
mean over notes.
\begin{itemize}[leftmargin=*,itemsep=1pt,topsep=2pt]
  \item \emph{Bootstrap intervals} resample notes 1{,}000 times and report percentile 95\% intervals
  \citep{efron1979bootstrap}; the intervals of the $I_c$ values tabulated in
  Appendices~\ref{app:heads} and~\ref{app:locus} exclude zero.
  \item A \emph{one-sample $t$-test} across the 100 per-note means gives each head's p-value. Bootstrap percentile
  p-values were too coarse to feed a multiple-comparison correction: in a synthetic check on pure noise they gave
  \NMCBootstrapPvalueSyntheticDiscoveries{} discoveries.
  \item \emph{Benjamini--Hochberg} \citep{benjamini1995controlling} controls the expected share of false
  discoveries among the heads it declares; under a complete null it should declare essentially none, which is the
  calibration check above.
  \item \emph{Sign tests} across notes compare paired sets (global against random, global against the note's own
  ranking). They count wins and assume nothing about effect sizes; every ``notes'' count in the paper is such a
  count.
  \item \emph{Recurrence} uses a binomial test per head with Bonferroni correction over the 1{,}600 heads
  (Appendix~\ref{app:heads}).
  \item \emph{Medians over notes of per-note means} are reported for necessity, a ratio to the clean margin that is
  unstable where that margin is near zero. Recovery is normalised per event by its ceiling and has no such problem.
  \item \emph{Patients.} The 100 traces come from \NFaithPatients{} patients: \NFaithPairedPatients{} patients contribute a
  second prompt variant (Appendix~\ref{app:data}). Re-aggregating with the patient as the unit (the two variants'
  per-note means averaged first) moves every knock-in median by at most \NFaithPatientsMaxKiMove{} and every \Pone{} and
  \Ptwoc{} necessity median by at most \NFaithPatientsMaxNecMoveNotPTwo; the \Ptwo{} necessity medians rise by
  \NFaithPatientsPTwoNecMoveMin{} to \NFaithPatientsPTwoNecMoveMax{} (the top-8 cell from \NFaithPTwoNecKEightGlobal{} to
  \NFaithPTwoNecKEightGlobalPatients), the read-side lifts move by at most \NFaithPatientsMaxMTwoMove{} nats, and no win
  count changes its character. The per-trace values are the conservative ones and are the ones reported.
\end{itemize}

\paragraph{Compute.} Generating the 100 traces took 6.4 GPU-hours on H200s. A note's key--value cache is about
17\,GiB at 110k tokens, and a single-token forward against it takes 72\,ms on an H200 and about 100\,ms on an RTX Pro
6000. The five passes over the corpus took: attribution, 6.3 GPU-hours (seven RTX Pro 6000 lanes, a median of 176\,s
per note); faithfulness, about 12 GPU-hours on B200 and RTX Pro 6000 GPUs (about 7{,}000 single-token forwards per
note); read side, about 1 GPU-hour (one prefill with query and key capture per note); prose locus, about 9.3
GPU-hours on B200s only, because the multi-position backward keeps every layer's expanded keys and values (130\,GiB
at 111k tokens); and the fixed prose-time set, about 3 GPU-hours. Each pass ran in under five hours of wall time on
two to seven lanes; HD-002 repeated all five on the shortlist prompts (Appendix~\ref{app:hd002}).

\section{Full head tables}
\label{app:heads}
\supports{Sec.~\ref{sec:char-sets}, Sec.~\ref{sec:char-phases}; \resid{R-C3}, \resid{R-C4}}

\paragraph{The global sets at every $k$.} Table~\ref{tab:faithfulness-app} is Table~\ref{tab:faithfulness} at every set
size tested, $k\in\{3,8,16,40,100\}$; the main text prints four of the five.

\begin{table}[!htbp]
  \centering\small
  \caption{Global per-phase sets at every $k$ (Table~\ref{tab:faithfulness} with the $k=16$ rows): necessity (share of the
  clean margin destroyed by mean-ablation; $>1$: flipped) and knock-in recovery in another patient's context, against
  count-matched random and bottom-$k$ sets. Medians over notes of per-note means; ``notes'': notes where the global set
  beats random (of 100; \Ptwoc: of 92). \resid{R-C3}}
  \label{tab:faithfulness-app}
  \providecommand{\gensd}[2]{\begin{tabular}[t]{@{}c@{}}#1\\[-2pt]{\scriptsize$\pm$#2}\end{tabular}}
\providecommand{\genstack}[2]{\begin{tabular}[t]{@{}c@{}}#1\\[-2pt]{\scriptsize #2}\end{tabular}}
\providecommand{\genstackl}[2]{\begin{tabular}[t]{@{}l@{}}#1\\[-2pt]{\scriptsize #2}\end{tabular}}
\providecommand{\genhead}[2]{\begin{tabular}[b]{@{}c@{}}#1\\#2\end{tabular}}
\providecommand{\genstackiii}[3]{\begin{tabular}[t]{@{}c@{}}#1\\[-2pt]{\scriptsize #2}\\[-3pt]{\scriptsize #3}\end{tabular}}
\setlength{\tabcolsep}{3.5pt}
\begin{tabular}{@{}r cccc cccc cccc@{}}
\toprule
 & \multicolumn{4}{c}{\Pone{} (chapter)} & \multicolumn{4}{c}{\Ptwo{} (character)} & \multicolumn{4}{c}{\Ptwoc{} (contrastive)} \\
\cmidrule(lr){2-5}\cmidrule(lr){6-9}\cmidrule(lr){10-13}
$k$ & global & random & bottom & notes & global & random & bottom & notes & global & random & bottom & notes \\
\midrule
\multicolumn{13}{@{}l}{\textit{Necessity: share of the clean margin destroyed by mean-ablating the set ($>1$: decision flipped)}} \\
3 & 0.24 & 0.00 & \ensuremath{-}0.15 & 92 & 0.41 & 0.01 & \ensuremath{-}0.04 & 95 & 0.26 & 0.00 & \ensuremath{-}0.06 & 90 \\
8 & 0.62 & 0.00 & \ensuremath{-}0.36 & 95 & 0.78 & 0.00 & \ensuremath{-}0.18 & 98 & 0.63 & 0.00 & \ensuremath{-}0.06 & 90 \\
16 & 0.88 & 0.02 & \ensuremath{-}0.05 & 97 & 1.10 & 0.06 & 0.11 & 97 & 0.72 & \ensuremath{-}0.01 & \ensuremath{-}0.12 & 90 \\
40 & 1.24 & 0.04 & 0.89 & 100 & 1.61 & 0.06 & 0.76 & 97 & 0.86 & 0.04 & \ensuremath{-}0.50 & 90 \\
100 & 1.60 & 0.09 & 1.57 & 100 & 2.21 & 0.26 & 1.41 & 99 & 0.89 & 0.07 & \ensuremath{-}0.16 & 90 \\
\midrule
\multicolumn{13}{@{}l}{\textit{Sufficiency by knock-in: share of the clean decision restored in another patient's context}} \\
3 & 0.16 & 0.00 & \ensuremath{-}0.16 & 100 & 0.24 & 0.00 & \ensuremath{-}0.45 & 100 & 0.05 & 0.00 & \ensuremath{-}0.33 & 80 \\
8 & 0.67 & 0.00 & \ensuremath{-}0.28 & 100 & 0.48 & 0.01 & \ensuremath{-}0.54 & 100 & 0.69 & 0.00 & \ensuremath{-}0.54 & 92 \\
16 & 0.76 & 0.02 & \ensuremath{-}0.48 & 100 & 0.64 & 0.01 & \ensuremath{-}0.63 & 100 & 0.73 & 0.00 & \ensuremath{-}0.56 & 92 \\
40 & 0.86 & 0.04 & \ensuremath{-}0.30 & 100 & 0.79 & 0.02 & \ensuremath{-}0.81 & 100 & 0.85 & 0.02 & \ensuremath{-}0.63 & 92 \\
100 & 0.96 & 0.09 & \ensuremath{-}0.33 & 100 & 0.89 & 0.05 & \ensuremath{-}0.80 & 100 & 0.94 & 0.05 & \ensuremath{-}0.78 & 92 \\
\bottomrule
\end{tabular}

\end{table}

\paragraph{Per-phase importances.} Table~\ref{tab:heads-app} gives decision-specific importance for representative
heads, and Table~\ref{tab:app-heads-more} completes it to every head in the chapter top-13, the character top-10 or
the contrastive top-9. Importances are means over notes of per-note $I_c$ in nats, with ranks among 1{,}600 heads;
every listed $I_c$ has a 95\% bootstrap interval that excludes zero. Chapter importances are exact for the nominated
heads, and character and contrastive importances are first-order (Appendix~\ref{app:instrument}). Profiles describe
the write side: \emph{shared} heads rank high in all three phases, while \emph{chapter-}, \emph{character-} and
\emph{contrastive-specific} heads rank high in one. Over all 1{,}600 heads, $I_c$ correlates \NHeadCorrPOnePTwo{}
between chapter and character decisions, \NHeadCorrPOnePTwoC{} between chapter and contrastive decisions and
\NHeadCorrPTwoPTwoC{} between character and contrastive decisions, and the top-40s share \NHeadOverlapFortyPOnePTwo{},
\NHeadOverlapFortyPOnePTwoC{} and \NHeadOverlapFortyPTwoPTwoC{} heads. The structure is a shared core that writes the
code string (\head{39}{38}) with phase-specific heads around it: heads in layers 26--28 for the chapter, heads in
layers 36--39 for the characters, and \head{30}{11}, ranked \NHeadLThirtyHElevenPOneRank{} on the chapter decision
but \NHeadLThirtyHElevenPTwoCRank{} on the contrastive one, where the decision is against a previously named
near-miss. Under
Benjamini--Hochberg at FDR 0.05, \NHeadPOneBhPos{}, \NHeadPTwoBhPos{} and \NHeadPTwoCBhPos{} heads have $I_c>0$ for
chapter, character and contrastive decisions: consistent small effects are widespread, and what makes a set is
magnitude (see Sparsity below).

\begin{table}[!htbp]
  \centering\small
  \caption{Decision-specific importance $I_c$ (nats; mean over notes; rank of 1{,}600) of representative heads, with
  write-side profiles and the layers of the chapter top-40. \Pone{} importances are exact for the nominated heads;
  the \Ptwo{} and \Ptwoc{} importances are the historical first-order rankings under the detached hook (the intact-hook
  sensitivity re-run of the \Ptwo{} ranking is Table~\ref{tab:app-p2intact}); the last column is the exact \Ptwoc{} refinement in raw importance
  $I$, not comparable with $I_c$. \resid{R-C4}}
  \label{tab:heads-app}
  \providecommand{\gensd}[2]{\begin{tabular}[t]{@{}c@{}}#1\\[-2pt]{\scriptsize$\pm$#2}\end{tabular}}
\providecommand{\genstack}[2]{\begin{tabular}[t]{@{}c@{}}#1\\[-2pt]{\scriptsize #2}\end{tabular}}
\providecommand{\genstackl}[2]{\begin{tabular}[t]{@{}l@{}}#1\\[-2pt]{\scriptsize #2}\end{tabular}}
\providecommand{\genhead}[2]{\begin{tabular}[b]{@{}c@{}}#1\\#2\end{tabular}}
\providecommand{\genstackiii}[3]{\begin{tabular}[t]{@{}c@{}}#1\\[-2pt]{\scriptsize #2}\\[-3pt]{\scriptsize #3}\end{tabular}}
\begin{tabular}{@{}l ccc c l@{}}
\toprule
 & \multicolumn{3}{c}{$I_c$ (nats; rank of 1{,}600)} & raw $I$ & \\
\cmidrule(lr){2-4}\cmidrule(lr){5-5}
Head & \Pone & \Ptwo & \Ptwoc & \Ptwoc{} exact & profile \\
\midrule
\head{39}{38} & 0.91 (1) & 0.97 (1) & 0.62 (1) & 0.22 (4) & shared (emission) \\
\head{26}{36} & 0.43 (3) & 0.08 (29) & 0.04 (28) & 0.03 (14) & chapter \\
\head{26}{37} & 0.29 (6) & 0.10 (21) & 0.05 (22) & 0.04 (9) & chapter \\
\head{27}{11} & 0.24 (7) & 0.06 (44) & 0.04 (32) & 0.03 (17) & chapter \\
\head{27}{13} & 0.21 (8) & 0.08 (34) & 0.03 (37) & 0.01 (42) & chapter \\
\head{39}{10} & 0.18 (13) & 0.42 (2) & 0.24 (7) & 0.36 (2) & shared, character-heavy \\
\head{39}{37} & 0.05 (72) & 0.29 (4) & 0.07 (18) & 0.02 (29) & character \\
\head{39}{12} & \ensuremath{-}0.07 (1{,}580) & 0.27 (8) & 0.24 (6) & 0.64 (1) & character \\
\head{38}{22} & \ensuremath{-}0.06 (1{,}572) & 0.23 (10) & 0.22 (8) & 0.00 (1{,}517) & character \\
\head{36}{10} & 0.06 (59) & 0.24 (9) & 0.02 (64) & 0.02 (35) & character \\
\head{30}{11} & \ensuremath{-}0.02 (1{,}500) & \ensuremath{-}0.08 (1{,}588) & 0.51 (2) & 0.28 (3) & contrastive \\
\head{38}{39} & \ensuremath{-}0.02 (1{,}485) & \ensuremath{-}0.04 (1{,}565) & 0.19 (9) & 0.12 (7) & contrastive \\
\midrule
\multicolumn{6}{@{}l}{\Pone{} top-40 by layer: 0--11: 0 (0\%) \quad 12--23: 2 (2\%) \quad 24--31: 24 (53\%) \quad 32--39: 14 (45\%)} \\
\bottomrule
\end{tabular}

\end{table}

\begin{table}[!htbp]
  \centering\small
  \caption{Heads in the chapter top-13, the character top-10 or the contrastive top-9 that are not in
  Table~\ref{tab:heads-app}: decision-specific importance $I_c$ (nats; mean over notes; rank of 1{,}600) and
  write-side profile. \resid{R-C4}}
  \label{tab:app-heads-more}
  \begin{adjustbox}{max width=\linewidth}
  \begin{tabular}{@{}lcccl@{}}
    \toprule
    Head & \Pone & \Ptwo & \Ptwoc & profile \\
    \midrule
    \head{39}{13} & \NMCIcPOneLThirtyNineHThirteen{} (\NMCRankPOneLThirtyNineHThirteen{})
      & \NMCIcPTwoLThirtyNineHThirteen{} (\NMCRankPTwoLThirtyNineHThirteen{})
      & \NMCIcPTwoCLThirtyNineHThirteen{} (\NMCRankPTwoCLThirtyNineHThirteen{}) & chapter, some character \\
    \head{38}{24} & \NMCIcPOneLThirtyEightHTwentyFour{} (\NMCRankPOneLThirtyEightHTwentyFour{})
      & \NMCIcPTwoLThirtyEightHTwentyFour{} (\NMCRankPTwoLThirtyEightHTwentyFour{})
      & \NMCIcPTwoCLThirtyEightHTwentyFour{} (\NMCRankPTwoCLThirtyEightHTwentyFour{}) & shared \\
    \head{28}{21} & \NMCIcPOneLTwentyEightHTwentyOne{} (\NMCRankPOneLTwentyEightHTwentyOne{})
      & \NMCIcPTwoLTwentyEightHTwentyOne{} (\NMCRankPTwoLTwentyEightHTwentyOne{})
      & \NMCIcPTwoCLTwentyEightHTwentyOne{} (\NMCRankPTwoCLTwentyEightHTwentyOne{}) & shared \\
    \head{35}{18} & \NMCIcPOneLThirtyFiveHEighteen{} (\NMCRankPOneLThirtyFiveHEighteen{})
      & \NMCIcPTwoLThirtyFiveHEighteen{} (\NMCRankPTwoLThirtyFiveHEighteen{})
      & \NMCIcPTwoCLThirtyFiveHEighteen{} (\NMCRankPTwoCLThirtyFiveHEighteen{}) & shared \\
    \head{39}{29} & \NMCIcPOneLThirtyNineHTwentyNine{} (\NMCRankPOneLThirtyNineHTwentyNine{})
      & \NMCIcPTwoLThirtyNineHTwentyNine{} (\NMCRankPTwoLThirtyNineHTwentyNine{})
      & \NMCIcPTwoCLThirtyNineHTwentyNine{} (\NMCRankPTwoCLThirtyNineHTwentyNine{}) & shared, character-heavy \\
    \head{37}{6} & \NMCIcPOneLThirtySevenHSix{} (\NMCRankPOneLThirtySevenHSix{})
      & \NMCIcPTwoLThirtySevenHSix{} (\NMCRankPTwoLThirtySevenHSix{})
      & \NMCIcPTwoCLThirtySevenHSix{} (\NMCRankPTwoCLThirtySevenHSix{}) & shared \\
    \head{26}{18} & \NMCIcPOneLTwentySixHEighteen{} (\NMCRankPOneLTwentySixHEighteen{})
      & \NMCIcPTwoLTwentySixHEighteen{} (\NMCRankPTwoLTwentySixHEighteen{})
      & \NMCIcPTwoCLTwentySixHEighteen{} (\NMCRankPTwoCLTwentySixHEighteen{}) & chapter \\
    \bottomrule
  \end{tabular}
  \end{adjustbox}
\end{table}

\paragraph{The exact contrastive refinement.} The contrastive global set, the one tested in
Table~\ref{tab:faithfulness-app} (and, measured on the shortlist, the one pooled for distillation), is the first-order
ranking. Exact mean-ablation of the 56 heads nominated at every \Ptwoc{} event of the faithfulness pass (the event's
top-40 by $|A_h|$ and the global top-16) gives the exact-merged ranking, exact where measured and first-order
elsewhere (Table~\ref{tab:app-p2c-exact}). It is in raw importance $I$, not \Pnull{}-calibrated, so
it is not comparable with $I_c$. First-order attribution had under-estimated \head{39}{12} and over-estimated
\head{30}{11}. The exact-merged and first-order top-40s share \NMCPTwoCExactFirstOrderOverlapTopForty{} heads,
every head of the exact top-8 is in both, and \NMCPTwoCExactBHPos{} heads have a positive exact-merged
importance under Benjamini--Hochberg at FDR 0.05. The contrastive reading stands: \head{30}{11} and \head{38}{39}
remain contrastive-specific, and two final-layer character heads, \head{39}{12} and \head{39}{10}, carry the largest
exact effects at the divergence character. This is why the first-order contrastive top-3 recovers only
\NFaithPTwoCKiKThreeGlobal{} on knock-in while the top-8 recovers \NFaithPTwoCKiKEightGlobal{}: the top of the
contrastive set is broader than the three heads that first-order attribution put first.

\begin{table}[!htbp]
  \centering\small
  \caption{The contrastive ranking after exact refinement, in exact-merged order: raw importance $I$ (nats; mean over
  the 92 notes with contrastive events; exact where measured) with its 95\% bootstrap interval, the first-order raw
  importance, and the head's rank in the first-order $I_c$ ranking that defines the contrastive global set. Raw $I$ is
  not \Pnull{}-calibrated and not comparable with $I_c$. \resid{R-C4}}
  \label{tab:app-p2c-exact}
  \begin{adjustbox}{max width=\linewidth}
  \begin{tabular}{@{}lcccc@{}}
    \toprule
    Head & $I$ exact-merged & 95\% interval & $I$ first-order & rank by first-order $I_c$ \\
    \midrule
    \head{39}{12} & \NHeadLThirtyNineHTwelvePTwoCExact
      & [\NMCPTwoCExactCiLoLThirtyNineHTwelve{}, \NMCPTwoCExactCiHiLThirtyNineHTwelve{}]
      & \NMCPTwoCFirstOrderILThirtyNineHTwelve{} & \NHeadLThirtyNineHTwelvePTwoCRank \\
    \head{39}{10} & \NHeadLThirtyNineHTenPTwoCExact
      & [\NMCPTwoCExactCiLoLThirtyNineHTen{}, \NMCPTwoCExactCiHiLThirtyNineHTen{}]
      & \NMCPTwoCFirstOrderILThirtyNineHTen{} & \NHeadLThirtyNineHTenPTwoCRank \\
    \head{30}{11} & \NHeadLThirtyHElevenPTwoCExact
      & [\NMCPTwoCExactCiLoLThirtyHEleven{}, \NMCPTwoCExactCiHiLThirtyHEleven{}]
      & \NMCPTwoCFirstOrderILThirtyHEleven{} & \NHeadLThirtyHElevenPTwoCRank \\
    \head{39}{38} & \NHeadLThirtyNineHThirtyEightPTwoCExact
      & [\NMCPTwoCExactCiLoLThirtyNineHThirtyEight{}, \NMCPTwoCExactCiHiLThirtyNineHThirtyEight{}]
      & \NMCPTwoCFirstOrderILThirtyNineHThirtyEight{} & \NHeadLThirtyNineHThirtyEightPTwoCRank \\
    \head{39}{29} & \NMCPTwoCExactLThirtyNineHTwentyNine{}
      & [\NMCPTwoCExactCiLoLThirtyNineHTwentyNine{}, \NMCPTwoCExactCiHiLThirtyNineHTwentyNine{}]
      & \NMCPTwoCFirstOrderILThirtyNineHTwentyNine{} & \NMCRankPTwoCLThirtyNineHTwentyNine{} \\
    \head{39}{25} & \NMCPTwoCExactLThirtyNineHTwentyFive{}
      & [\NMCPTwoCExactCiLoLThirtyNineHTwentyFive{}, \NMCPTwoCExactCiHiLThirtyNineHTwentyFive{}]
      & \NMCPTwoCFirstOrderILThirtyNineHTwentyFive{} & \NMCRankPTwoCLThirtyNineHTwentyFive{} \\
    \head{38}{39} & \NHeadLThirtyEightHThirtyNinePTwoCExact
      & [\NMCPTwoCExactCiLoLThirtyEightHThirtyNine{}, \NMCPTwoCExactCiHiLThirtyEightHThirtyNine{}]
      & \NMCPTwoCFirstOrderILThirtyEightHThirtyNine{} & \NHeadLThirtyEightHThirtyNinePTwoCRank \\
    \head{28}{21} & \NMCPTwoCExactLTwentyEightHTwentyOne{}
      & [\NMCPTwoCExactCiLoLTwentyEightHTwentyOne{}, \NMCPTwoCExactCiHiLTwentyEightHTwentyOne{}]
      & \NMCPTwoCFirstOrderILTwentyEightHTwentyOne{} & \NMCRankPTwoCLTwentyEightHTwentyOne{} \\
    \bottomrule
  \end{tabular}
  \end{adjustbox}
\end{table}

\paragraph{Layers.} At the code token the chapter top-40 has \NHeadPOneTopFortyLayersZeroToElevenCount{} heads in
layers 0--11 and \NHeadPOneTopFortyLayersTwelveToTwentyThreeCount{} in layers 12--23; it sits in the middle and late
layers, with \NHeadPOneTopFortyLayersTwentyFourToThirtyOneCount{} heads in layers 24--31 carrying
\NHeadPOneTopFortyLayersTwentyFourToThirtyOneShare{} of its importance and
\NHeadPOneTopFortyLayersThirtyTwoToThirtyNineCount{} in layers 32--39 carrying
\NHeadPOneTopFortyLayersThirtyTwoToThirtyNineShare{} (Table~\ref{tab:heads-app}, last row). At the prose locus the
set that decides the chapter sits one block earlier (Appendix~\ref{app:locus}).

\paragraph{Pooled weights at the operating point.} The pooled weights $w_h=\operatorname{softmax}(I_c/\tau)$ over a
phase's top-$k$ concentrate on the leader. At $k=3$ the leader gets \NMCWeightLeaderKThreePOne{} of the weight
for chapter decisions and \NMCWeightLeaderKThreePTwoC{} for contrastive ones, and at the top-40 operating point
\head{39}{38} still holds \NMCWeightLThirtyNineHThirtyEightTopFortyPOne{}, \NMCWeightLThirtyNineHThirtyEightTopFortyPTwo{} and
\NMCWeightLThirtyNineHThirtyEightTopFortyPTwoC{} of it (\Pone, \Ptwo, \Ptwoc): HD-001's pooled top-40 is close to one head. The
distillation half pools the shortlist's sets (HD-002, Appendix~\ref{app:hd002}), whose weights are spread; their
leaders are \head{27}{11} (\NMCShortWeightLeaderTopFortyPOne{}), \head{39}{10}
(\NMCShortWeightLeaderTopFortyPTwo{}) and \head{39}{29} (\NMCShortWeightLeaderTopFortyPTwoC{}).

\paragraph{Necessity and knock-in in full: the global set against the note's own ranking.}
Table~\ref{tab:faithfulness-app} tests the global set against random and bottom-$k$ controls; Table~\ref{tab:app-own} adds
each note's own ranking, fitted to that note's own events. For the chapter decision the two are level on necessity
(\NFaithPOneNecKEightGlobal{} against \NFaithPOneNecKEightOwn{} at $k=8$; the global set destroys at least as much as
the note's own on between \NMCPOneNecGlobalGeOwnNotesMin{} and \NMCPOneNecGlobalGeOwnNotesMax{}
of 100 notes for $k\le 40$, and on \NMCPOneNecGlobalGeOwnNotesKOneHundred{} at $k=100$), and the global set is the
better carrier on knock-in at every $k\ge 8$ (\NFaithPOneKiKEightGlobal{} against \NFaithPOneKiKEightOwn{} at $k=8$).
A ranking fitted to a note's own chapter events (a median of 19) plausibly picks up heads that were necessary on
those events by happenstance, whereas the cross-note ranking picks the heads that carry the decision as such. For
characters the note's own ranking is the better carrier at every $k$ and removes more up to $k=40$ (at $k=8$,
\NFaithPTwoKiKEightOwn{} against \NFaithPTwoKiKEightGlobal{} on knock-in and \NFaithPTwoNecKEightOwn{} against
\NFaithPTwoNecKEightGlobal{} on necessity), as expected of a ranking fitted to about 70 events and tested on them; for
contrastive events it is better on necessity at every $k$ and on knock-in at every $k$ except 8. At the top-40 operating point the global set's
knock-in is within \NMKiOwnGapTopForty{} of the note's own for characters and for contrastive events.

\begin{table}[!htbp]
  \centering\small
  \caption{The global set against each note's own ranking (fitted to that note's events), at every $k$: necessity
  (share of the clean margin destroyed; $>1$: flipped) and knock-in recovery in another patient's context. Medians over
  notes of per-note means (\Ptwoc: 92 notes). Random and bottom-$k$ controls are in Table~\ref{tab:faithfulness-app}.
  \resid{R-C3}}
  \label{tab:app-own}
  \begin{adjustbox}{max width=\linewidth}
  \setlength{\tabcolsep}{3.5pt}
  \begin{tabular}{@{}r cccccc @{\hspace{12pt}} cccccc@{}}
    \toprule
     & \multicolumn{6}{c}{Necessity} & \multicolumn{6}{c@{}}{Knock-in recovery} \\
    \cmidrule(lr){2-7}\cmidrule(l){8-13}
     & \multicolumn{2}{c}{\Pone} & \multicolumn{2}{c}{\Ptwo} & \multicolumn{2}{c}{\Ptwoc}
     & \multicolumn{2}{c}{\Pone} & \multicolumn{2}{c}{\Ptwo} & \multicolumn{2}{c@{}}{\Ptwoc} \\
    \cmidrule(lr){2-3}\cmidrule(lr){4-5}\cmidrule(lr){6-7}\cmidrule(lr){8-9}\cmidrule(lr){10-11}\cmidrule(l){12-13}
    $k$ & global & own & global & own & global & own & global & own & global & own & global & own \\
    \midrule
    3 & \NFaithPOneNecKThreeGlobal & \NFaithPOneNecKThreeOwn & \NFaithPTwoNecKThreeGlobal & \NFaithPTwoNecKThreeOwn
      & \NFaithPTwoCNecKThreeGlobal & \NFaithPTwoCNecKThreeOwn & \NFaithPOneKiKThreeGlobal & \NFaithPOneKiKThreeOwn
      & \NFaithPTwoKiKThreeGlobal & \NFaithPTwoKiKThreeOwn & \NFaithPTwoCKiKThreeGlobal & \NFaithPTwoCKiKThreeOwn \\
    8 & \NFaithPOneNecKEightGlobal & \NFaithPOneNecKEightOwn & \NFaithPTwoNecKEightGlobal & \NFaithPTwoNecKEightOwn
      & \NFaithPTwoCNecKEightGlobal & \NFaithPTwoCNecKEightOwn & \NFaithPOneKiKEightGlobal & \NFaithPOneKiKEightOwn
      & \NFaithPTwoKiKEightGlobal & \NFaithPTwoKiKEightOwn & \NFaithPTwoCKiKEightGlobal & \NFaithPTwoCKiKEightOwn \\
    16 & \NFaithPOneNecKSixteenGlobal & \NFaithPOneNecKSixteenOwn & \NFaithPTwoNecKSixteenGlobal
      & \NFaithPTwoNecKSixteenOwn & \NFaithPTwoCNecKSixteenGlobal & \NFaithPTwoCNecKSixteenOwn
      & \NFaithPOneKiKSixteenGlobal & \NFaithPOneKiKSixteenOwn & \NFaithPTwoKiKSixteenGlobal
      & \NFaithPTwoKiKSixteenOwn & \NFaithPTwoCKiKSixteenGlobal & \NFaithPTwoCKiKSixteenOwn \\
    40 & \NFaithPOneNecKFortyGlobal & \NFaithPOneNecKFortyOwn & \NFaithPTwoNecKFortyGlobal & \NFaithPTwoNecKFortyOwn
      & \NFaithPTwoCNecKFortyGlobal & \NFaithPTwoCNecKFortyOwn & \NFaithPOneKiKFortyGlobal & \NFaithPOneKiKFortyOwn
      & \NFaithPTwoKiKFortyGlobal & \NFaithPTwoKiKFortyOwn & \NFaithPTwoCKiKFortyGlobal & \NFaithPTwoCKiKFortyOwn \\
    100 & \NFaithPOneNecKOneHundredGlobal & \NFaithPOneNecKOneHundredOwn & \NFaithPTwoNecKOneHundredGlobal
      & \NFaithPTwoNecKOneHundredOwn & \NFaithPTwoCNecKOneHundredGlobal & \NFaithPTwoCNecKOneHundredOwn
      & \NFaithPOneKiKOneHundredGlobal & \NFaithPOneKiKOneHundredOwn & \NFaithPTwoKiKOneHundredGlobal
      & \NFaithPTwoKiKOneHundredOwn & \NFaithPTwoCKiKOneHundredGlobal & \NFaithPTwoCKiKOneHundredOwn \\
    \bottomrule
  \end{tabular}
  \end{adjustbox}
\end{table}

\paragraph{Spread, and the jump from three to eight heads.} Across notes the chapter top-8 destroys between
\NMCPOneNecTopEightPTen{} (10th percentile) and \NMCPOneNecTopEightPNinety{} (90th) of the margin. The ratio is
unstable on the few notes whose clean margins sit near zero, which is why medians are reported, and the tail is
heavier for characters, whose margins are more often near or below zero (Table~\ref{tab:app-audit}). Knock-in,
normalised by each event's ceiling, is tighter: across notes the chapter top-8 recovers between
\NMCPOneKiTopEightPTen{} and \NMCPOneKiTopEightPNinety{} (10th to 90th percentile), and more than half on
\NMCPOneKiTopEightAboveHalfNotes{} of 100 notes. Recovery jumps between three and eight heads, from
\NFaithPOneKiKThreeGlobal{} to \NFaithPOneKiKEightGlobal{}, and rises to \NFaithPOneKiKFortyGlobal{} at 40 and
\NFaithPOneKiKOneHundredGlobal{} at 100: heads four to eight of the chapter ranking (\head{38}{24}, \head{28}{21},
\head{26}{37}, \head{27}{11}, \head{27}{13}) act jointly with the top three. For characters and contrastive decisions
the top-8 recovers \NFaithPTwoKiKEightGlobal{} and \NFaithPTwoCKiKEightGlobal{}.

\paragraph{The bottom-$k$ oppose the decision.} Removing the eight most negatively ranked chapter heads raises the
chapter margin (necessity \NFaithPOneNecKEightBottom{}), and writing the bottom sets' clean outputs into the corrupted
run moves it further from the clean decision (recovery between \NFaithPOneKiKThreeBottom{} and
\NFaithPTwoKiKFortyBottom{}; Table~\ref{tab:faithfulness-app}). These heads argue for the alternative, which is why the
negative discoveries of the calibration (Appendix~\ref{app:instrument}) are effects rather than noise. At $k=40$ and
100 the chapter bottom set is large enough to include heads with large generic effects, and removing it destroys
the margin too (\NFaithPOneNecKFortyBottom{} and \NFaithPOneNecKOneHundredBottom{}).

\paragraph{Isolation is not a sufficiency test.} The attribution pass first tested sufficiency by isolation: keep only
the top-$k$ of each note's own ranking and mean-ablate every other head at $q$. The margin that remains is about
\NMCIsolationMarginRetained{} of the clean margin at every $k$, so the decision flips, while the same pass's
necessity bins read \NMCAttrPassOwnNecKThree{}, \NMCAttrPassOwnNecKEight{},
\NMCAttrPassOwnNecKSixteen{}, \NMCAttrPassOwnNecKForty{} and \NMCAttrPassOwnNecKOneHundred{} at
$k=3$ to 100, with the top-$k$ above random on \NMCAttrPassOwnNecWinsMin{} to
\NMCAttrPassOwnNecWinsMax{} of 100 notes, the same picture as Table~\ref{tab:app-own}. Mean-ablating some
1{,}560 heads removes the residual stream's base, not only the heads that argue for the competitors, so isolation
asks whether the set works without the rest of the model, which is not the question. The set is necessary on top of
a broad base. Knock-in, which holds everything outside the set at the
corrupted run's values, is the sufficiency test the paper uses.

\paragraph{Recurrence.} If each note's top-20 were a random draw from 1{,}600 heads, a given head would appear in it
with probability 20/1{,}600, so the number of notes in which it appears would be binomial over 100 notes; a head is
called recurrent if its count survives Bonferroni correction over all 1{,}600 heads. Using each note's top-20 by raw
importance $I$ (not \Pnull{}-calibrated), \NMCRecurrentHeadsCount{} heads are recurrent, against 1.25 notes
expected per head by chance. \head{39}{38} and \head{39}{13} are in the top-20 of every note
(\NMCRecurrenceLThirtyNineHThirtyEight{} and \NMCRecurrenceLThirtyNineHThirteen{} of 100), followed by \head{26}{36}
(\NMCRecurrenceLTwentySixHThirtySix{}), \head{26}{37} (\NMCRecurrenceLTwentySixHThirtySeven{}), \head{28}{21}
(\NMCRecurrenceLTwentyEightHTwentyOne{}), \head{26}{18} (\NMCRecurrenceLTwentySixHEighteen{}), \head{27}{13}
(\NMCRecurrenceLTwentySevenHThirteen{}), \head{27}{11} (\NMCRecurrenceLTwentySevenHEleven{}), \head{24}{1}
(\NMCRecurrenceLTwentyFourHOne{}) and \head{39}{10} (\NMCRecurrenceLThirtyNineHTen{}).

\paragraph{Sparsity.} Counting chapter heads above the calibration noise floor (the 95th percentile of the absolute
split-half \Pnull{} difference, \NMCNullFloorNats{} nats), \NMCPOneHeadsAboveFloor{} heads have
$I_c$ above the floor and \NMCPOneHeadsForEightyPct{} of them carry 80\% of the decision-specific importance.
Per note, counting raw $I$ above the note's own \Pnull{} floor, a median of
\NMCPerNoteHeadsForEightyPctMedian{} heads carry 80\%. The set is neither three heads nor diffuse. Importance
is not additive under ablation, so forty heads that flip the decision when removed, forty heads that recover
\NFaithPOneKiKFortyGlobal{} of it when knocked in, and \NMCPOneHeadsForEightyPct{} heads that carry 80\% of the
summed importance are one picture seen from three sides.

\paragraph{Stability of the discovery across notes.} The global sets are ranked on the 100 characterization notes and, in
Table~\ref{tab:faithfulness-app}, tested on them; a reader may ask whether the ranking is fitted to those notes. We split the
94 patients into two halves at random \NSplitHalfHDOneSplits{} times and ranked on each half with the same calibrated
contrast ranking (Table~\ref{tab:app-splithalf}). Across disjoint halves the top-8 is identical at every phase in both
the full-inventory and the shortlist run (median overlap \NSplitHalfHDTwoPTwoCTopEight{} of 8), the top-40 shares
\NSplitHalfHDTwoPTwoCTopForty{} to \NSplitHalfHDOnePTwoTopForty{} of 40, and the rankings over all 1{,}600 heads correlate at
Spearman \NSplitHalfHDOnePTwoSpearman{} to \NSplitHalfHDOnePTwoCSpearman{} on the full inventory and down to
\NSplitHalfHDTwoPTwoCSpearman{} for the shortlist's contrastive ranking, whose tail is noise-floor heads in arbitrary
order. On the exact contrastive effects, the top-8 chosen on one half carries \NSplitHalfExactCrossfitContrastEight{} (chosen
by the contrast ranking) to \NSplitHalfExactCrossfitExactEight{} (chosen by the exact ranking) of the importance of the
other half's own best eight. Each note is one per cent of the mean that ranks the heads, and the ranking is the same on
notes it never saw. The students' own-heads rows are cross-fit by construction: ranked on the 100 validation traces and
read on the 150 held-out test notes (Appendix~\ref{app:metrics}).

\begin{table}[!htbp]
  \centering\small
  \caption{Split-half stability of the head discovery: the contrast ranking on one random half of the patients against
  the other, \NSplitHalfHDOneSplits{} splits; medians over splits of the Spearman correlation over 1{,}600 heads and of the
  top-$k$ overlaps, and of each half's top-40 overlap with the full ranking. \resid{R-C13}}
  \label{tab:app-splithalf}
  \providecommand{\gensd}[2]{\begin{tabular}[t]{@{}c@{}}#1\\[-2pt]{\scriptsize$\pm$#2}\end{tabular}}
\providecommand{\genstack}[2]{\begin{tabular}[t]{@{}c@{}}#1\\[-2pt]{\scriptsize #2}\end{tabular}}
\providecommand{\genstackl}[2]{\begin{tabular}[t]{@{}l@{}}#1\\[-2pt]{\scriptsize #2}\end{tabular}}
\providecommand{\genhead}[2]{\begin{tabular}[b]{@{}c@{}}#1\\#2\end{tabular}}
\providecommand{\genstackiii}[3]{\begin{tabular}[t]{@{}c@{}}#1\\[-2pt]{\scriptsize #2}\\[-3pt]{\scriptsize #3}\end{tabular}}
\footnotesize\setlength{\tabcolsep}{4pt}
\begin{adjustbox}{max width=\linewidth}
\begin{tabular}{@{}l c ccc c@{}}
\toprule
Phase & Spearman (1{,}600 heads) & top-8 & top-16 & top-40 & half vs full top-40 \\
\midrule
\multicolumn{6}{@{}l}{\textit{shortlist (HD-002; the sets the students were supervised with)}} \\
\Pone & 0.81 & 8 & 15 & 37 & 38.5 \\
\Ptwo & 0.79 & 8 & 15 & 38 & 39.0 \\
\Ptwoc & 0.63 & 8 & 14 & 36 & 38.0 \\
\multicolumn{6}{@{}l}{\textit{full inventory (HD-001)}} \\
\Pone & 0.90 & 7 & 14 & 38 & 39.0 \\
\Ptwo & 0.94 & 8 & 16 & 37 & 38.5 \\
\Ptwoc & 0.82 & 8 & 16 & 35 & 37.0 \\
\bottomrule
\end{tabular}
\end{adjustbox}

\end{table}

\section{Read side}
\label{app:readside}
\supports{Sec.~\ref{sec:char-read}; \resid{R-C5}}

\begin{figure*}[!htbp]
  \centering
  \includegraphics[width=\linewidth]{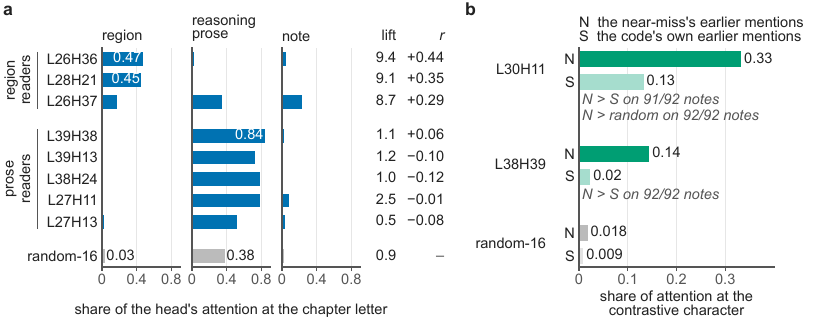}
  \caption{What the causal heads read (\teacher; attention at the event, medians over notes). (a) At
  chapter-letter events, the share of attention the global \Pone{} top-8 put on the emitted chapter's inventory
  lines (region), on the reasoning so far (prose) and on the note, with 16 random heads as reference; right, each
  head's region lift and $r$, the correlation across events between its region share and its importance. Region
  readers read the region, prose readers the reasoning. (b) At contrastive events, the share on the earlier
  mentions of the near-miss (N) and of the code itself (S) for the two contrastive-specific heads and 16 random
  heads. Descriptive: attention was not intervened on. \resid{R-C5}}
  \label{fig:readside}
\end{figure*}

\paragraph{Panel, events and position categories.} The read side reads the query--key half of the same heads whose
outputs the write side measured \citep{elhage2021mathematical}. A fixed panel of 68 heads (the global top-16 of each
phase, the global bottom-16 of the chapter and contrastive rankings, and 16 random heads drawn with a fixed seed) is
read at every \Pone{} event and at the \Ptwo{} and \Ptwoc{} events of the faithfulness pass. One prefill per note
with query and key capture gives each panel head's causal attention row at $q$, which is split over 15 categories of
key position: the attention sink (the first 8 tokens); the chat template; in the inventory, the emitted code's own
line, the other lines of its three-character category, the other lines of its chapter, other chapters' lines, the
near-miss's line, and inventory tokens outside any line; the note; the instruction; and in the trace, earlier mentions
of the code ($S$), earlier mentions of the near-miss ($N$), mentions of other codes, the characters of the current
mention written so far (the code prefix), and the remaining prose.

\paragraph{Definitions.} The region is the emitted chapter's inventory lines. \emph{On region} is the share of a
head's total attention that falls on the region. \emph{Region lift} divides the share of the head's inventory
attention that falls on the region by the region's share of inventory tokens, so 1 means that the head reads the
inventory indifferently. $n_{\mathrm{eff}}$ is the effective number of inventory tokens the head attends, the inverse
participation ratio $1/\sum_j p_j^2$ of its attention renormalised over inventory tokens; an inventory line is about
18 tokens, so an $n_{\mathrm{eff}}$ of a few dozen means a few lines. $r$ is the Pearson correlation, across a head's
events, between its share on the region (on $N$ at \Ptwoc) and its measured importance at that event. Values are
medians over notes of per-note means, and group comparisons are sign tests across notes.

\paragraph{Group level.} The global top-16 of every phase reads the emitted chapter's lines selectively, far above the
random panel, on every note (Table~\ref{tab:app-read-group}). The bottom-16 read them almost as selectively: region
selectivity is a property of heads that matter, with either sign, so what a head reads does not say which way it
writes; the write side (Appendix~\ref{app:heads}) does. Across the chapter top-16, Spearman's correlation between
lift and $I_c$ is \NMCPOneTopSixteenSpearmanLiftIc{}, because the most important head, \head{39}{38}, is a prose
reader.

\begin{table}[!htbp]
  \centering\small
  \caption{Read side by phase: region lift of the global top-16, 16 random heads and the global bottom-16 (\Pone{}
  top-16 with its 10th--90th percentile over notes); notes on which the top-16's lift beats the random panel's; and
  the top-16's attention on the sink, the inventory's code lines, the note, the trace prose and the code prefix (random
  panel in brackets). Medians over notes. \resid{R-C5}}
  \label{tab:app-read-group}
  \begin{adjustbox}{max width=\linewidth}
  \begin{tabular}{@{}lccccccccc@{}}
    \toprule
     & \multicolumn{3}{c}{Region lift} & & \multicolumn{5}{c@{}}{Top-16 attention on} \\
    \cmidrule(lr){2-4}\cmidrule(l){6-10}
    Kind & top-16 & random & bottom & top $>$ random & sink & inventory lines & note & prose & code prefix \\
    \midrule
    \Pone & \NReadPOneTopLift{} (\NMCPOneTopSixteenLiftPTen{}--\NMCPOneTopSixteenLiftPNinety{})
      & \NReadPOneRandLift & \NReadPOneBottomLift & \NReadPOneWins{}/100 & \NMCPOneTopSixteenOnSink{}
      & \NMCPOneTopSixteenOnInventoryLines{} & \NMCPOneTopSixteenOnNote{} & \NMCPOneTopSixteenOnProse{} & --- \\
    \Ptwo & \NReadPTwoTopLift & \NReadPTwoRandLift & --- & \NReadPTwoWins{}/100 & \NMCPTwoTopSixteenOnSink{}
      & \NMCPTwoTopSixteenOnInventoryLines{} & \NMCPTwoTopSixteenOnNote{} & \NMCPTwoTopSixteenOnProse{}
      & \NMCPTwoTopSixteenOnPrefix{} (\NMCPTwoRandomSixteenOnPrefix{}) \\
    \Ptwoc & \NReadPTwoCTopLift & \NReadPTwoCRandLift & \NReadPTwoCBottomLift & \NReadPTwoCWins{}/92
      & \NMCPTwoCTopSixteenOnSink{} & \NMCPTwoCTopSixteenOnInventoryLines{} & \NMCPTwoCTopSixteenOnNote{}
      & \NMCPTwoCTopSixteenOnProse{} & \NMCPTwoCTopSixteenOnPrefix{} (\NMCPTwoCRandomSixteenOnPrefix{}) \\
    \bottomrule
  \end{tabular}
  \end{adjustbox}
\end{table}

\paragraph{Chapter decisions: two families.} Table~\ref{tab:app-read-p1} gives the chapter top-16 head by head; the
random panel puts \NMCPOneRandomSixteenOnRegion{} of its attention on the region. Five heads in layers 24--28 are
\emph{region readers} (\head{26}{36}, \head{28}{21}, \head{26}{37}, \head{26}{18}, \head{24}{1}).
\head{26}{36} and \head{28}{21} put \NReadLTwentySixHThirtySixRegion{} and
\NReadLTwentyEightHTwentyOneRegion{} of their attention on the emitted chapter's lines (lift
\NReadLTwentySixHThirtySixLift{} and \NReadLTwentyEightHTwentyOneLift{}) over an effective few lines, and
\head{26}{37} and \head{26}{18} split theirs between the region and the note. For the region readers, reading the
region is what makes them matter: the more a region reader reads the region at an event, the larger its effect on that
decision ($r=$ \NReadLTwentySixHThirtySixR{} for \head{26}{36}, \NReadLTwentyEightHTwentyOneR{} for \head{28}{21},
\NReadLTwentySixHThirtySevenR{} for \head{26}{37} and \NMCLTwentySixHEighteenRPOne{} for \head{26}{18}). The
\emph{prose readers} (the late-layer \head{39}{38}, \head{39}{13} and \head{38}{24}, and \head{27}{11}, which is
chapter-specific on the write side) put \NReadLThirtyNineHThirteenProse{} to \NReadLThirtyNineHThirtyEightProse{} of
their attention on the trace prose, and their effect does not depend on where in the prose they look ($r\approx 0$).
For the mentions decided in the reasoning, they carry into the output a chapter that is already elicitable from the prose
(Appendix~\ref{app:locus}). The top three heads already contain
both families (the prose readers \head{39}{38} and \head{39}{13}, the region reader \head{26}{36}); heads four to
eight, which make the jump in knock-in recovery from three to eight heads (Appendix~\ref{app:heads}), add two region
readers (\head{28}{21}, \head{26}{37}) and three prose readers (\head{38}{24}, \head{27}{11}, \head{27}{13}). The
boundary between the families and the ``mixed'' heads is a judgement call; the table gives the numbers.

\begin{table}[!htbp]
  \centering\small
  \caption{Chapter decisions: the global \Pone{} top-16 at \Pone{} events. $I_c$ (nats); share of attention on the
  region (the emitted chapter's inventory lines), on the trace prose and on the sink; region lift; $n_{\mathrm{eff}}$
  over inventory tokens; $r$ between region share and importance across the head's events; family. Medians over notes.
  Descriptive: attention was not intervened on. \resid{R-C5}}
  \label{tab:app-read-p1}
  \begin{adjustbox}{max width=\linewidth}
  \begin{tabular}{@{}lccccccc l@{}}
    \toprule
    Head & $I_c$ & region & prose & sink & lift & $n_{\mathrm{eff}}$ & $r$ & family \\
    \midrule
    \head{26}{36} & \NHeadLTwentySixHThirtySixPOneIc & \NReadLTwentySixHThirtySixRegion
      & \NReadLTwentySixHThirtySixProse & \NMCLTwentySixHThirtySixOnSinkPOne{} & \NReadLTwentySixHThirtySixLift
      & \NMCLTwentySixHThirtySixNeffPOne{} & \NReadLTwentySixHThirtySixR & region reader \\
    \head{28}{21} & \NMCIcPOneLTwentyEightHTwentyOne{} & \NReadLTwentyEightHTwentyOneRegion
      & \NReadLTwentyEightHTwentyOneProse & \NMCLTwentyEightHTwentyOneOnSinkPOne{} & \NReadLTwentyEightHTwentyOneLift
      & \NMCLTwentyEightHTwentyOneNeffPOne{} & \NReadLTwentyEightHTwentyOneR & region reader \\
    \head{26}{37} & \NHeadLTwentySixHThirtySevenPOneIc & \NReadLTwentySixHThirtySevenRegion
      & \NReadLTwentySixHThirtySevenProse & \NMCLTwentySixHThirtySevenOnSinkPOne{} & \NReadLTwentySixHThirtySevenLift
      & \NMCLTwentySixHThirtySevenNeffPOne{} & \NReadLTwentySixHThirtySevenR
      & region reader; note \NMCLTwentySixHThirtySevenOnNotePOne{} \\
    \head{26}{18} & \NMCIcPOneLTwentySixHEighteen{} & \NMCLTwentySixHEighteenOnRegionPOne{} & \NMCLTwentySixHEighteenOnProsePOne{}
      & \NMCLTwentySixHEighteenOnSinkPOne{} & \NMCLTwentySixHEighteenLiftPOne{} & \NMCLTwentySixHEighteenNeffPOne{}
      & \NMCLTwentySixHEighteenRPOne{} & region reader; note \NMCLTwentySixHEighteenOnNotePOne{} \\
    \head{24}{1} & \NMCIcPOneLTwentyFourHOne{} & \NMCLTwentyFourHOneOnRegionPOne{} & \NMCLTwentyFourHOneOnProsePOne{}
      & \NMCLTwentyFourHOneOnSinkPOne{} & \NMCLTwentyFourHOneLiftPOne{} & \NMCLTwentyFourHOneNeffPOne{}
      & \NMCLTwentyFourHOneRPOne{} & region reader \\
    \head{25}{28} & \NMCIcPOneLTwentyFiveHTwentyEight{} & \NMCLTwentyFiveHTwentyEightOnRegionPOne{} & \NMCLTwentyFiveHTwentyEightOnProsePOne{}
      & \NMCLTwentyFiveHTwentyEightOnSinkPOne{} & \NMCLTwentyFiveHTwentyEightLiftPOne{} & \NMCLTwentyFiveHTwentyEightNeffPOne{}
      & \NMCLTwentyFiveHTwentyEightRPOne{} & mixed \\
    \head{39}{38} & \NHeadLThirtyNineHThirtyEightPOneIc & \NReadLThirtyNineHThirtyEightRegion
      & \NReadLThirtyNineHThirtyEightProse & \NMCLThirtyNineHThirtyEightOnSinkPOne{} & \NReadLThirtyNineHThirtyEightLift
      & \NMCLThirtyNineHThirtyEightNeffPOne{} & \NReadLThirtyNineHThirtyEightR & prose reader \\
    \head{39}{13} & \NMCIcPOneLThirtyNineHThirteen{} & \NReadLThirtyNineHThirteenRegion & \NReadLThirtyNineHThirteenProse
      & \NMCLThirtyNineHThirteenOnSinkPOne{} & \NReadLThirtyNineHThirteenLift & \NMCLThirtyNineHThirteenNeffPOne{}
      & \NReadLThirtyNineHThirteenR & prose reader \\
    \head{38}{24} & \NMCIcPOneLThirtyEightHTwentyFour{} & \NReadLThirtyEightHTwentyFourRegion
      & \NReadLThirtyEightHTwentyFourProse & \NMCLThirtyEightHTwentyFourOnSinkPOne{} & \NReadLThirtyEightHTwentyFourLift
      & \NMCLThirtyEightHTwentyFourNeffPOne{} & \NReadLThirtyEightHTwentyFourR & prose reader \\
    \head{27}{11} & \NHeadLTwentySevenHElevenPOneIc & \NReadLTwentySevenHElevenRegion & \NReadLTwentySevenHElevenProse
      & \NMCLTwentySevenHElevenOnSinkPOne{} & \NReadLTwentySevenHElevenLift & \NMCLTwentySevenHElevenNeffPOne{}
      & \NReadLTwentySevenHElevenR & prose reader \\
    \head{27}{13} & \NHeadLTwentySevenHThirteenPOneIc & \NMCLTwentySevenHThirteenOnRegionPOne{}
      & \NMCLTwentySevenHThirteenOnProsePOne{} & \NMCLTwentySevenHThirteenOnSinkPOne{} & \NMCLTwentySevenHThirteenLiftPOne{}
      & \NMCLTwentySevenHThirteenNeffPOne{} & \NMCLTwentySevenHThirteenRPOne{}
      & prose reader; other chapters \NMCLTwentySevenHThirteenOnOtherChaptersPOne{} \\
    \head{39}{29} & \NMCIcPOneLThirtyNineHTwentyNine{} & \NMCLThirtyNineHTwentyNineOnRegionPOne{} & \NMCLThirtyNineHTwentyNineOnProsePOne{}
      & \NMCLThirtyNineHTwentyNineOnSinkPOne{} & \NMCLThirtyNineHTwentyNineLiftPOne{} & \NMCLThirtyNineHTwentyNineNeffPOne{}
      & \NMCLThirtyNineHTwentyNineRPOne{} & prose reader \\
    \head{35}{18} & \NMCIcPOneLThirtyFiveHEighteen{} & \NMCLThirtyFiveHEighteenOnRegionPOne{} & \NMCLThirtyFiveHEighteenOnProsePOne{}
      & \NMCLThirtyFiveHEighteenOnSinkPOne{} & \NMCLThirtyFiveHEighteenLiftPOne{} & \NMCLThirtyFiveHEighteenNeffPOne{}
      & \NMCLThirtyFiveHEighteenRPOne{} & inventory, diffuse; other chapters \NMCLThirtyFiveHEighteenOnOtherChaptersPOne{} \\
    \head{37}{6} & \NMCIcPOneLThirtySevenHSix{} & \NMCLThirtySevenHSixOnRegionPOne{} & \NMCLThirtySevenHSixOnProsePOne{}
      & \NMCLThirtySevenHSixOnSinkPOne{} & \NMCLThirtySevenHSixLiftPOne{} & \NMCLThirtySevenHSixNeffPOne{}
      & \NMCLThirtySevenHSixRPOne{} & inventory, diffuse \\
    \head{39}{10} & \NHeadLThirtyNineHTenPOneIc & \NMCLThirtyNineHTenOnRegionPOne{} & \NMCLThirtyNineHTenOnProsePOne{}
      & \NMCLThirtyNineHTenOnSinkPOne{} & \NMCLThirtyNineHTenLiftPOne{} & \NMCLThirtyNineHTenNeffPOne{}
      & \NMCLThirtyNineHTenRPOne{} & sink \\
    \head{26}{33} & \NMCIcPOneLTwentySixHThirtyThree{} & \NMCLTwentySixHThirtyThreeOnRegionPOne{} & \NMCLTwentySixHThirtyThreeOnProsePOne{}
      & \NMCLTwentySixHThirtyThreeOnSinkPOne{} & \NMCLTwentySixHThirtyThreeLiftPOne{} & \NMCLTwentySixHThirtyThreeNeffPOne{}
      & \NMCLTwentySixHThirtyThreeRPOne{} & --- \\
    \bottomrule
  \end{tabular}
  \end{adjustbox}
\end{table}

\paragraph{Character decisions.} The character-specific heads read the code being written
(Table~\ref{tab:app-read-p2}): seven heads of the character top-16 put between \NMCLThirtyNineHTwelveOnPrefixPTwo{} and
\NMCLThirtyNineHTwentyFiveOnPrefixPTwo{} of their attention on the current mention's already-written characters, against
\NMCPTwoRandomSixteenOnPrefix{} for the random panel as a group, a copy-and-continue mechanism over the code
string. A second group reads the category's lines in the inventory, putting \NMCLThirtyEightHTwentyTwoOnCategoryPTwo{} to
\NMCLTwentyEightHTwentyOneOnCategoryPTwo{} of its attention on the emitted code's category siblings at lifts of
\NMCLThirtyEightHTwentyTwoLiftPTwo{} to \NMCLThirtyEightHThirtyFiveLiftPTwo{}: refinement within the category, on the read side.

\begin{table}[!htbp]
  \centering\small
  \caption{Character decisions: heads of the character global top-16 at \Ptwo{} events. Left, share of attention on
  the current mention's already-written characters (the code prefix), with the random and top-16 panels as groups;
  right, share on the emitted code's category siblings in the inventory, and region lift. Medians over notes.
  \resid{R-C5}}
  \label{tab:app-read-p2}
  \begin{adjustbox}{max width=\linewidth}
  \begin{tabular}{@{}lc@{\hspace{18pt}}lcc@{}}
    \toprule
    \multicolumn{2}{@{}c}{Code-prefix readers} & \multicolumn{3}{c@{}}{Category readers} \\
    \cmidrule(r){1-2}\cmidrule(l){3-5}
    Head & on prefix & Head & on category & lift \\
    \midrule
    \head{39}{25} & \NMCLThirtyNineHTwentyFiveOnPrefixPTwo{} & \head{28}{21} & \NMCLTwentyEightHTwentyOneOnCategoryPTwo{}
      & \NMCLTwentyEightHTwentyOneLiftPTwo{} \\
    \head{39}{10} & \NMCLThirtyNineHTenOnPrefixPTwo{} & \head{38}{35} & \NMCLThirtyEightHThirtyFiveOnCategoryPTwo{}
      & \NMCLThirtyEightHThirtyFiveLiftPTwo{} \\
    \head{39}{29} & \NMCLThirtyNineHTwentyNineOnPrefixPTwo{} & \head{30}{13} & \NMCLThirtyHThirteenOnCategoryPTwo{}
      & \NMCLThirtyHThirteenLiftPTwo{} \\
    \head{36}{10} & \NMCLThirtySixHTenOnPrefixPTwo{} & \head{38}{22} & \NMCLThirtyEightHTwentyTwoOnCategoryPTwo{}
      & \NMCLThirtyEightHTwentyTwoLiftPTwo{} \\
    \head{39}{36} & \NMCLThirtyNineHThirtySixOnPrefixPTwo{} & & & \\
    \head{39}{37} & \NMCLThirtyNineHThirtySevenOnPrefixPTwo{} & & & \\
    \head{39}{12} & \NMCLThirtyNineHTwelveOnPrefixPTwo{} & & & \\
    \midrule
    top-16 (group) & \NMCPTwoTopSixteenOnPrefix{} & & & \\
    random-16 (group) & \NMCPTwoRandomSixteenOnPrefix{} & & & \\
    \bottomrule
  \end{tabular}
  \end{adjustbox}
\end{table}

\paragraph{Contrastive decisions: the near-miss loop-back.} At the character where a code diverges from the
near-miss, the contrastive-specific head \head{30}{11} puts \NReadLThirtyHElevenSibling{} of its attention on the
earlier mentions of the near-miss ($N$; 10th--90th percentile over notes \NMCLThirtyHElevenOnSiblingPTen{} to
\NMCLThirtyHElevenOnSiblingPNinety{}), against \NReadPTwoCRandSibling{} for the random panel
(Table~\ref{tab:app-read-p2c}). It is above the random panel on \NMCLThirtyHElevenAboveRandomNotes{} of 92 notes and
above its own share on the code's earlier mentions ($S$) on \NMLThirtyHElevenSiblingOverOwnNotes{} of 92, and its
measured effect grows with its share on $N$ ($r=\NReadLThirtyHElevenPTwoCR$). \head{38}{39}, the other
contrastive-specific head, also reads $N$ more than $S$, on \NMCLThirtyEightHThirtyNineSiblingOverOwnNotes{} of 92 notes.
The head re-reads the sibling named earlier. As a group, the contrastive top-16 put \NShortReadPTwoCNearMissFull{}
of their attention on $N$ and \NMCPTwoCTopSixteenOnPrefix{} on the code prefix (random panel
\NMCPTwoCRandomSixteenOnPrefix{}). A query--key score, the similarity of a head's query at the event to its keys at $S$
minus that to its keys at $N$, computed at the 2{,}208 contrastive events with earlier mentions of the code, does not
separate heads:
it tops out at \NMCQkSMinusNMax{}, with a flat block of layer-34 and layer-35 heads at near-identical values,
and since \head{30}{11} attends to $N$ more than to $S$, such a score would rank the loop-back head negatively. The
near-miss loop-back is therefore measured as attention mass on $N$, as M3 is in the distillation half.

\begin{table}[!htbp]
  \centering\small
  \caption{Contrastive decisions: attention on the earlier mentions of the near-miss ($N$) and of the code itself
  ($S$) at \Ptwoc{} events (medians over the 92 notes; random and bottom panels: means over their 16 heads), and the
  notes on which a head puts more on $N$ than on $S$. $I_c$ is the head's contrastive importance (nats).
  \resid{R-C5}}
  \label{tab:app-read-p2c}
  \begin{adjustbox}{max width=\linewidth}
  \begin{tabular}{@{}lcccc@{}}
    \toprule
    Head & $I_c$ (\Ptwoc) & on $N$ & on $S$ & $N>S$ (notes of 92) \\
    \midrule
    \head{30}{11} & \NHeadLThirtyHElevenPTwoCIc & \NReadLThirtyHElevenSibling & \NReadLThirtyHElevenOwn
      & \NMLThirtyHElevenSiblingOverOwnNotes \\
    \head{38}{22} & \NHeadLThirtyEightHTwentyTwoPTwoCIc & \NReadLThirtyEightHTwentyTwoSibling
      & \NReadLThirtyEightHTwentyTwoOwn & \NMCLThirtyEightHTwentyTwoSiblingOverOwnNotes{} \\
    \head{38}{39} & \NHeadLThirtyEightHThirtyNinePTwoCIc & \NReadLThirtyEightHThirtyNineSibling
      & \NReadLThirtyEightHThirtyNineOwn & \NMCLThirtyEightHThirtyNineSiblingOverOwnNotes{} \\
    \head{38}{35} & \NMCIcPTwoCLThirtyEightHThirtyFive{} & \NMCLThirtyEightHThirtyFiveOnSiblingPTwoC{} & \NMCLThirtyEightHThirtyFiveOnOwnPTwoC{}
      & \NMCLThirtyEightHThirtyFiveSiblingOverOwnNotes{} \\
    random-16 & --- & \NReadPTwoCRandSibling & \NMCPTwoCRandomSixteenOnOwn{} & --- \\
    bottom-16 & --- & \NMCPTwoCBottomSixteenOnSibling{} & \NMCPTwoCBottomSixteenOnOwn{} & --- \\
    \bottomrule
  \end{tabular}
  \end{adjustbox}
\end{table}

\paragraph{Blocking what the heads read.} The tables above describe where the causal heads attend; whether a head's
effect on the decision runs \emph{through} what it attends to is a separate question, which we answered by intervening on
the keys. At the decision token, the named head is prevented from attending to the named key positions: they are removed
from its softmax and its attention renormalizes over the remaining keys, while every other head and layer runs
unchanged (bit-identical in the unit test; exact against eager attention). The key-side effect $D$ is the margin lost
against a reference that runs the same masked path with an empty key set, so that kernel noise cancels (the reference
differs from the clean margin by a median of a few hundredths of a nat); it is compared with the head's whole effect $E$, its mean-ablation at the same token, and with a matched control that blocks
the same number of tokens from other chapters' whole inventory lines (at \Pone) or other codes' whole earlier mentions
(at \Ptwoc). Table~\ref{tab:app-keyabl} gives the result on all 2{,}064 chapter events and on the \NKeyAblLThirtyHElevenEvents{}
contrastive events with earlier sibling mentions. For the near-miss loop-back, blocking \head{30}{11}'s attention to the
earlier mentions of the sibling removes its whole effect on the contrastive decision (median share
\NKeyAblLThirtyHElevenShare{} of $E$ over events where the head matters), blocking matched mentions of other codes removes
nothing (median \NKeyAblLThirtyHElevenDCtrlMedian{} nats against \NKeyAblLThirtyHElevenDNamedMedian), and the named keys
beat the control on \NKeyAblLThirtyHElevenNoteHigher{} of \NKeyAblLThirtyHElevenNoteN{} notes; the pair with \head{38}{39}
reads the same way, and the sibling's inventory line carries none of the pair's effect (share \NKeyAblSetLineShare). For
the region readers, blocking the emitted chapter's inventory lines removes \NKeyAblLTwentyEightHTwentyOneShare{} of
\head{28}{21}'s effect and \NKeyAblLTwentySixHThirtySixShare{} of \head{26}{36}'s, against nothing for matched lines of
other chapters, on \NKeyAblLTwentyEightHTwentyOneNoteHigher{} of 100 notes for both. The read-side names are therefore
causal: what these heads read is what their effect runs through.

\begin{table}[!htbp]
  \centering\small
  \caption{Key-side intervention on the named read-side heads at the decision token (\teacher, HD-001's notes). $E$: the
  head's mean-ablation effect on the margin (nats; share of events where $E>$ \NKeyAblMinEffect). Named keys: the margin
  lost when the head cannot attend to the positions it is named for; control: the same number of matched tokens of other
  chapters' lines or other codes' mentions. Share: median of $D/E$ over events where $E>$ \NKeyAblMinEffect. Right: the
  per-note mean of named minus control with its bootstrap interval, notes higher/lower, sign test. \resid{R-C10}}
  \label{tab:app-keyabl}
  \providecommand{\gensd}[2]{\begin{tabular}[t]{@{}c@{}}#1\\[-2pt]{\scriptsize$\pm$#2}\end{tabular}}
\providecommand{\genstack}[2]{\begin{tabular}[t]{@{}c@{}}#1\\[-2pt]{\scriptsize #2}\end{tabular}}
\providecommand{\genstackl}[2]{\begin{tabular}[t]{@{}l@{}}#1\\[-2pt]{\scriptsize #2}\end{tabular}}
\providecommand{\genhead}[2]{\begin{tabular}[b]{@{}c@{}}#1\\#2\end{tabular}}
\providecommand{\genstackiii}[3]{\begin{tabular}[t]{@{}c@{}}#1\\[-2pt]{\scriptsize #2}\\[-3pt]{\scriptsize #3}\end{tabular}}
\scriptsize\setlength{\tabcolsep}{3pt}
\begin{adjustbox}{max width=\linewidth}
\begin{tabular}{@{}l l c c c c c@{}}
\toprule
Head & event; named keys & \genhead{head effect $E$}{median / mean (share $> 0.1$)} & \genhead{named keys blocked}{median / mean} & \genhead{matched control}{median / mean} & \genhead{share of $E$}{median (events)} & \genhead{named $-$ control, per note}{mean [95\% CI]; higher/lower; $p$} \\
\midrule
region reader \head{26}{36} & \Pone; the emitted chapter's lines & +0.37 / +0.41 (74\%) & +0.12 / +0.25 & +0.00 / +0.00 & 0.33 (1{,}524) & \genstack{\pci{+0.28}{+0.25}{+0.31}}{100/0, $p$ \ensuremath{2{\times}10^{-30}}} \\
region reader \head{28}{21} & \Pone; the emitted chapter's lines & +0.22 / +0.28 (61\%) & +0.24 / +0.41 & +0.00 / \ensuremath{-}0.01 & 0.70 (1{,}269) & \genstack{\pci{+0.46}{+0.42}{+0.50}}{100/0, $p$ \ensuremath{2{\times}10^{-30}}} \\
near-miss loop-back \head{30}{11} & \Ptwoc; earlier mentions of the sibling & +0.12 / +0.27 (59\%) & +0.25 / +0.44 & +0.00 / \ensuremath{-}0.01 & 1.00 (777) & \genstack{\pci{+0.45}{+0.39}{+0.51}}{92/0, $p$ \ensuremath{4{\times}10^{-28}}} \\
\head{38}{39} & \Ptwoc; earlier mentions of the sibling & +0.00 / +0.12 (42\%) & +0.00 / +0.12 & +0.00 / \ensuremath{-}0.01 & 0.89 (558) & \genstack{\pci{+0.13}{+0.11}{+0.15}}{88/0, $p$ \ensuremath{6{\times}10^{-27}}} \\
\head{30}{11} $+$ \head{38}{39} & \Ptwoc; earlier mentions of the sibling & +0.12 / +0.40 (63\%) & +0.25 / +0.56 & +0.00 / \ensuremath{-}0.02 & 1.00 (831) & \genstack{\pci{+0.58}{+0.51}{+0.66}}{92/0, $p$ \ensuremath{4{\times}10^{-28}}} \\
\head{30}{11} $+$ \head{38}{39} & \Ptwoc; the sibling's inventory line (secondary) & +0.12 / +0.40 (63\%) & +0.00 / +0.00 & +0.00 / +0.00 & 0.00 (858) & \genstack{\pci{+0.00}{+0.00}{+0.01}}{47/27, $p$ 0.03} \\
\bottomrule
\end{tabular}
\end{adjustbox}

\end{table}

\paragraph{Editing the earlier mention.} The key-side intervention shows that \head{30}{11} acts through its read of
the earlier mention; it does not show that the decision depends on that mention's \emph{identity} as the sibling. We
edited the trace itself: at \NNearMissTeacherNeutralN{} contrastive events of the 100 notes (at most 20 per note, seeded),
the most recent earlier mention of the sibling was replaced by (i)~a non-code filler of the same token length,
(ii)~an unrelated code of the same shape from another chapter, or (iii)~a code of the same chapter but another category,
and the context was re-fed from the edit so that the margin at the original decision could be read; an identity edit
(the same string through the same path) is the reference, and the two named heads were also mean-ablated during each
read to measure how much of the change they mediate (Table~\ref{tab:app-nearmiss}). Removing the mention lowers the
contrastive margin (mean \NNearMissTeacherNeutralMean{} nats, median \NNearMissTeacherNeutralMedian; positive on
\NNearMissTeacherNeutralHigher{} of \NNearMissTeacherNeutralN{} events, sign test $p$ \NNearMissTeacherNeutralP); both
string-copy controls also lower it, and the paired comparisons detect no difference between the tested replacement classes (paired per event, neutral minus unrelated median
\NNearMissTeacherNeutralMinusUnrelatedMedian, $p$ \NNearMissTeacherNeutralMinusUnrelatedP; minus other-category
\NNearMissTeacherNeutralMinusOthercatMedian, $p$ \NNearMissTeacherNeutralMinusOthercatP), and \head{30}{11} with
\head{38}{39} mediate a median \NNearMissTeacherNeutralMediated{} of the change per event
(\NNearMissTeacherNeutralMediatedMeanFrac{} of the mean, carried by the large-effect tail; the two heads hold
\NNearMissTeacherHeadsShareClean{} of the clean margin). No difference was detected between the tested replacements
(a median paired difference of 0 is not equality): sibling-specific identity processing is not established by these
edits, generic disruption of the stored trace remains an alternative explanation, and most of the dependence runs
through other paths. The read-side claim is therefore exactly what Table~\ref{tab:app-keyabl} supports: the head
re-reads the earlier mention and acts through that read; whether it reads the near-miss's identity is not established.

\begin{table}[!htbp]
  \centering\small
  \caption{Editing the earlier mention of the sibling in the teacher's trace: the change of the contrastive margin at
  the original decision, per replacement (effect $=$ margin under the identity edit minus margin under the replacement;
  positive $=$ the edit lowered the margin). Mediated: share of the change removed when \head{30}{11} and \head{38}{39}
  are mean-ablated during both reads, median over events with an effect above half a nat. \resid{R-C11}}
  \label{tab:app-nearmiss}
  \providecommand{\gensd}[2]{\begin{tabular}[t]{@{}c@{}}#1\\[-2pt]{\scriptsize$\pm$#2}\end{tabular}}
\providecommand{\genstack}[2]{\begin{tabular}[t]{@{}c@{}}#1\\[-2pt]{\scriptsize #2}\end{tabular}}
\providecommand{\genstackl}[2]{\begin{tabular}[t]{@{}l@{}}#1\\[-2pt]{\scriptsize #2}\end{tabular}}
\providecommand{\genhead}[2]{\begin{tabular}[b]{@{}c@{}}#1\\#2\end{tabular}}
\providecommand{\genstackiii}[3]{\begin{tabular}[t]{@{}c@{}}#1\\[-2pt]{\scriptsize #2}\\[-3pt]{\scriptsize #3}\end{tabular}}
\footnotesize\setlength{\tabcolsep}{4pt}
\begin{adjustbox}{max width=\linewidth}
\begin{tabular}{@{}l r c c c c c@{}}
\toprule
Replacement of the earlier mention & events & \genhead{effect (nats)}{mean / median} & \genhead{events}{$>0$ / $<0$} & sign $p$ & \genhead{mediated by}{\head{30}{11}, \head{38}{39}} & \genhead{per-note mean,}{median over notes} \\
\midrule
identity (the same string; the reference) & 1{,}365 & +0.00 / +0.00 & 347 / 364 & 0.55 & --- & +0.00 \\
neutral filler, same token length & 1{,}365 & +1.02 / +0.31 & 796 / 502 & \ensuremath{3{\times}10^{-16}} & 0.09 & +0.96 \\
unrelated code, same shape, other chapter & 1{,}363 & +1.14 / +0.38 & 784 / 522 & \ensuremath{4{\times}10^{-13}} & 0.08 & +1.17 \\
other-category code, same chapter & 1{,}328 & +1.04 / +0.25 & 751 / 529 & \ensuremath{6{\times}10^{-10}} & 0.07 & +1.08 \\
\bottomrule
\end{tabular}
\end{adjustbox}

\end{table}

The same edits applied to the distilled students (teacher-forced on the teacher's test traces, the arm's own top-2
contrastive heads as the mediation set) give a result we report in full and do not turn into a claim
(Table~\ref{tab:app-nearmiss-students}). In the paper's bf16 \qwenfour{} pair, the \mistill{} student loses more
contrastive margin under the edit than the \vanillakd{} student on every seed (per note, pooled over the three seed pairs
with the note as the cluster: \NNearMissQwenMistillVsVanillaBfPooledDelta{} nats
[\NNearMissQwenMistillVsVanillaBfPooledLo, \NNearMissQwenMistillVsVanillaBfPooledHi], higher on
\NNearMissQwenMistillVsVanillaBfPooledHigher{} of 150 notes), and the \vanillakd{} student sits below the untrained
student. With fp32 master weights the ordering reverses on every seed at the $\lambda$ transferred from bf16
(\NNearMissQwenMistillFpVsVanillaPooledDelta) and returns, at a fifth of the size and on one seed, at the $\lambda$
selected for fp32 (\NNearMissQwenMistillFpSelVsVanillaPooledDelta, \NNearMissQwenMistillFpSelVsVanillaPooledHigher{} of
148 notes; Appendix~\ref{app:objective}); in \smol{} the \vanillakd{} students are the more context-dependent ones against
both precisions (\NNearMissSmolMistillVsVanillaBfPooledDelta{} and \NNearMissSmolMistillVsVanillaPooledDelta). In every
arm the string-copy controls match the neutral edit and the students' own top-2 heads mediate nothing
(\NNearMissQwenMistillSZeroMediated{} for \mistill{} seed 0). How much a student's contrastive margin depends on an
earlier code string is a property of the trained pair, not of the objective; the transfer claim of
Section~\ref{sec:distill-mech} stays at causal recovery on the student's own heads, which does replicate across seeds,
precisions and families.

\begin{table}[!htbp]
  \centering\small
  \caption{The same edits on the students (150 test notes, teacher-forced on the teacher's traces): per note, the mean
  edit effect over the three replacements on the events both arms share; treated minus control per seed pair, and pooled
  with the note as the cluster (bootstrap-over-notes 95\% interval; notes higher/lower; sign test). The ordering is
  pair-specific. \resid{R-D28}}
  \label{tab:app-nearmiss-students}
  \providecommand{\gensd}[2]{\begin{tabular}[t]{@{}c@{}}#1\\[-2pt]{\scriptsize$\pm$#2}\end{tabular}}
\providecommand{\genstack}[2]{\begin{tabular}[t]{@{}c@{}}#1\\[-2pt]{\scriptsize #2}\end{tabular}}
\providecommand{\genstackl}[2]{\begin{tabular}[t]{@{}l@{}}#1\\[-2pt]{\scriptsize #2}\end{tabular}}
\providecommand{\genhead}[2]{\begin{tabular}[b]{@{}c@{}}#1\\#2\end{tabular}}
\providecommand{\genstackiii}[3]{\begin{tabular}[t]{@{}c@{}}#1\\[-2pt]{\scriptsize #2}\\[-3pt]{\scriptsize #3}\end{tabular}}
\footnotesize\setlength{\tabcolsep}{4pt}
\begin{adjustbox}{max width=\linewidth}
\begin{tabular}{@{}l c l c@{}}
\toprule
Pair (treated $-$ control) & \genhead{seed}{pairs} & per seed pair: per-note $\Delta$ (notes higher / $n$) & \genhead{note-clustered}{$\Delta$ [95\% CI]; higher/lower; $p$} \\
\midrule
\multicolumn{4}{@{}l}{\qwenfour} \\
\qwenfour, bf16 pair: \mistill{} $-$ \vanillakd{} (bf16) & 3 & +0.30 (146/150); +0.28 (147/150); +0.28 (144/150) & \genstackiii{\pci{+0.29}{+0.26}{+0.31}}{147/3}{$p$ \ensuremath{8{\times}10^{-40}}} \\
\quad \mistill{} $-$ untrained & 3 & +0.22 (115/149); +0.22 (114/150); +0.22 (114/150) & \genstackiii{\pci{+0.22}{+0.16}{+0.28}}{116/34}{$p$ \ensuremath{10^{-11}}} \\
\quad \vanillakd{} (bf16) $-$ untrained & 3 & \ensuremath{-}0.08 (58/149); \ensuremath{-}0.06 (57/148); \ensuremath{-}0.07 (58/150) & \genstackiii{\pci{\ensuremath{-}0.07}{\ensuremath{-}0.12}{\ensuremath{-}0.02}}{57/93}{$p$ 0.004} \\
\qwenfour, fp32 pair at $\lambda$ 0.1: \mistill{} $-$ \vanillakd & 3 & \ensuremath{-}0.24 (24/150); \ensuremath{-}0.21 (24/148); \ensuremath{-}0.20 (29/149) & \genstackiii{\pci{\ensuremath{-}0.22}{\ensuremath{-}0.25}{\ensuremath{-}0.18}}{26/124}{$p$ \ensuremath{2{\times}10^{-16}}} \\
\quad \mistill{} (fp32) $-$ untrained & 3 & \ensuremath{-}0.10 (59/149); \ensuremath{-}0.07 (67/150); \ensuremath{-}0.07 (71/150) & \genstackiii{\pci{\ensuremath{-}0.08}{\ensuremath{-}0.15}{\ensuremath{-}0.01}}{67/83}{$p$ 0.22} \\
\quad \vanillakd{} (fp32) $-$ untrained & 3 & +0.13 (95/148); +0.14 (97/150); +0.13 (93/150) & \genstackiii{\pci{+0.13}{+0.08}{+0.20}}{94/56}{$p$ 0.002} \\
\qwenfour, fp32 pair at $\lambda$ 0.01 (selected for fp32): \mistill{} $-$ \vanillakd & 1 & +0.06 (115/148) & \genstackiii{\pci{+0.06}{+0.05}{+0.07}}{115/33}{$p$ \ensuremath{8{\times}10^{-12}}} \\
\quad \mistill{} (fp32, $\lambda$ 0.01) $-$ \mistill{} (fp32, $\lambda$ 0.1) & 1 & +0.30 (135/150) & \genstackiii{\pci{+0.30}{+0.26}{+0.34}}{135/15}{$p$ \ensuremath{3{\times}10^{-25}}} \\
\midrule
\multicolumn{4}{@{}l}{\smol} \\
\smol: \mistill{} $-$ \vanillakd{} (bf16) & 1 & \ensuremath{-}0.16 (37/148) & \genstackiii{\pci{\ensuremath{-}0.16}{\ensuremath{-}0.20}{\ensuremath{-}0.12}}{37/111}{$p$ \ensuremath{9{\times}10^{-10}}} \\
\smol: \mistill{} $-$ \vanillakd{} (fp32) & 2 & \ensuremath{-}0.54 (7/150); \ensuremath{-}0.48 (11/150) & \genstackiii{\pci{\ensuremath{-}0.51}{\ensuremath{-}0.56}{\ensuremath{-}0.46}}{10/140}{$p$ \ensuremath{2{\times}10^{-30}}} \\
\quad \mistill{} $-$ untrained & 2 & +0.34 (118/150); +0.33 (115/150) & \genstackiii{\pci{+0.34}{+0.27}{+0.40}}{116/34}{$p$ \ensuremath{10^{-11}}} \\
\quad \vanillakd{} (fp32) $-$ untrained & 2 & +0.88 (147/150); +0.81 (147/150) & \genstackiii{\pci{+0.85}{+0.77}{+0.92}}{147/3}{$p$ \ensuremath{8{\times}10^{-40}}} \\
\quad \vanillakd{} (bf16) $-$ untrained & 1 & +0.50 (139/150) & \genstack{+0.50}{139/11, $p$ \ensuremath{2{\times}10^{-29}}} \\
\bottomrule
\end{tabular}
\end{adjustbox}

\end{table}

\paragraph{Editing the note's distinguishing words.} A second edit asks whether the contrastive decision reads the
discharge summary's evidence for the emitted code over its sibling. The notes are lower-cased running text without
sentence punctuation, so the unit is the word: for each event whose emitted code has a content word in its inventory
description that the sibling's description lacks and that occurs in the note (\NEvidenceOk{} of the
\NNearMissTeacherNeutralN{} events above have one, \NEvidenceOkPct; the rest, mostly ``other'' against ``unspecified''
siblings, have none), every occurrence of such words in the note was replaced by a token-length-matched filler, and
likewise the sibling's distinguishing words, the same number of clinical words belonging only to other chapters'
descriptions, and the same number of words found in no description (Table~\ref{tab:app-evidence}). Removing the emitted
code's distinguishing words lowers the contrastive margin no more than removing the matched control words (per note
\NEvidenceEvidenceMinusControlNoteMean{} nats [\NEvidenceEvidenceMinusControlNoteLo, \NEvidenceEvidenceMinusControlNoteHi],
\NEvidenceEvidenceMinusControlNoteHigher{} of \NEvidenceEvidenceMinusControlNoteN{} notes higher, sign test $p$
\NEvidenceEvidenceMinusControlNoteP; against the no-description control
\NEvidenceEvidenceMinusControlNdNoteMean{} [\NEvidenceEvidenceMinusControlNdNoteLo, \NEvidenceEvidenceMinusControlNdNoteHi]),
and removing the sibling's words does not move it the other way. Given the teacher-forced reasoning, which usually restates
the evidence, the contrastive decision does not read the note's distinguishing words directly; the note is read earlier,
when the chapter becomes elicitable (Appendix~\ref{app:locus}).

\begin{table}[!htbp]
  \centering\small
  \caption{Editing the note: the contrastive margin's change when words are replaced in the discharge summary (one
  re-feed from the earliest edited token; effect $=$ margin under the identity re-feed minus margin under the variant),
  for the \NEvidenceOk{} events whose emitted code has a distinguishing description word in the note. \resid{R-C11}}
  \label{tab:app-evidence}
  \providecommand{\gensd}[2]{\begin{tabular}[t]{@{}c@{}}#1\\[-2pt]{\scriptsize$\pm$#2}\end{tabular}}
\providecommand{\genstack}[2]{\begin{tabular}[t]{@{}c@{}}#1\\[-2pt]{\scriptsize #2}\end{tabular}}
\providecommand{\genstackl}[2]{\begin{tabular}[t]{@{}l@{}}#1\\[-2pt]{\scriptsize #2}\end{tabular}}
\providecommand{\genhead}[2]{\begin{tabular}[b]{@{}c@{}}#1\\#2\end{tabular}}
\providecommand{\genstackiii}[3]{\begin{tabular}[t]{@{}c@{}}#1\\[-2pt]{\scriptsize #2}\\[-3pt]{\scriptsize #3}\end{tabular}}
\footnotesize
\begin{tabular}{@{}l r c c@{}}
\toprule
Words removed from the note & events & \genhead{effect (nats)}{mean / median} & \genhead{events}{$>0$ / $<0$} \\
\midrule
the emitted code's distinguishing words & 428 & +0.054 / +0.000 & 38\% / 32\% \\
the sibling's distinguishing words & 191 & \ensuremath{-}0.021 / +0.000 & 29\% / 37\% \\
matched other-chapter clinical words & 428 & \ensuremath{-}0.001 / +0.000 & 29\% / 30\% \\
matched words in no description & 424 & +0.000 / +0.000 & 30\% / 31\% \\
\bottomrule
\end{tabular}

\end{table}

\paragraph{The reads on verified rejections.} Restricting the student readouts to the audited events
(Appendix~\ref{app:instrument}, Table~\ref{tab:app-p2caudit}) splits them into verified rejections and the rest
(Table~\ref{tab:app-audit-reads}). Verified rejections are the harder events for every arm on the teacher-forced read (M1
positive on \NAuditQwenMistillSZeroMOneRejected{} of them for \mistill{} seed 0 against
\NAuditQwenMistillSZeroMOneOther{} on the rest; \NAuditMOneRejectedN{} and \NAuditMOneOtherN{} events), and on the knock-in
read the \mistill{} arms recover more on verified rejections than on the rest while the bf16 \vanillakd{} arms do not, so
the gap between the two objectives is, if anything, larger on the events that are rejections in the strict sense
(\NAuditMFourRejectedN{} patched rejection events per arm: a direction, not a test).

\begin{table}[!htbp]
  \centering\small
  \caption{The student readouts restricted to the audited contrastive events: M1 (share of events where the student's
  margin favours the teacher's choice) and own-heads knock-in recovery at $k=40$, on the verified rejections ($n$ = \NAuditMOneRejectedN{} events for M1, \NAuditMFourRejectedN{} patched events for the
  knock-in) and on the other labeled events (\NAuditMOneOtherN{} and \NAuditMFourOtherN; enumerations and the remaining
  classes). \resid{R-C9}}
  \label{tab:app-audit-reads}
  \providecommand{\gensd}[2]{\begin{tabular}[t]{@{}c@{}}#1\\[-2pt]{\scriptsize$\pm$#2}\end{tabular}}
\providecommand{\genstack}[2]{\begin{tabular}[t]{@{}c@{}}#1\\[-2pt]{\scriptsize #2}\end{tabular}}
\providecommand{\genstackl}[2]{\begin{tabular}[t]{@{}l@{}}#1\\[-2pt]{\scriptsize #2}\end{tabular}}
\providecommand{\genhead}[2]{\begin{tabular}[b]{@{}c@{}}#1\\#2\end{tabular}}
\providecommand{\genstackiii}[3]{\begin{tabular}[t]{@{}c@{}}#1\\[-2pt]{\scriptsize #2}\\[-3pt]{\scriptsize #3}\end{tabular}}
\footnotesize
\begin{tabular}{@{}l cc cc@{}}
\toprule
 & \multicolumn{2}{c}{M1 \Ptwoc{} ($P(\margin>0)$)} & \multicolumn{2}{c}{own-heads knock-in, $k=40$} \\
\cmidrule(lr){2-3}\cmidrule(lr){4-5}
Arm & rejections & other & rejections & other \\
\midrule
\mistill{} s0 & 0.794 & 0.914 & 0.867 & 0.850 \\
\quad s1 & 0.825 & 0.914 & 0.898 & 0.853 \\
\quad s2 & 0.825 & 0.914 & 0.935 & 0.894 \\
\vanillakd{} (bf16) s0 & 0.810 & 0.886 & 0.813 & 0.823 \\
\quad s1 & 0.810 & 0.895 & 0.816 & 0.826 \\
\quad s2 & 0.810 & 0.895 & 0.845 & 0.848 \\
\vanillakd{} (fp32) s0 & 0.825 & 0.895 & 0.876 & 1.002 \\
\mistill{} (fp32, $\lambda$ 0.1) s0 & 0.825 & 0.857 & 0.904 & 0.952 \\
untrained & 0.810 & 0.886 & 0.791 & 0.825 \\
\bottomrule
\end{tabular}

\end{table}

\paragraph{Caveat.} The read side of Tables~\ref{tab:app-read-group}--\ref{tab:app-read-p2c} is descriptive, and its
link to the write side is a correlation across events; the key-side intervention above makes the named roles causal for
the three heads it tested, and the trace edits leave open what the loop-back reads: no difference was detected between
the tested replacements, so a sibling-specific read of the near-miss's identity is not established. The families are read off medians over notes and hold across notes: the region readers' lift
exceeds \NMCRegionReadersLiftMinOverNotes{} on every note.

\section{Prose locus}
\label{app:locus}
\supports{Sec.~\ref{sec:char-locus}; \resid{R-C6}}

The events of Appendix~\ref{app:instrument} measure heads at the code token, where the chapter margin is usually already
decisive (Table~\ref{tab:app-audit}). Two additions locate where it can be elicited and by which heads.

\paragraph{Forcing probe.} Crop the cache to an earlier position $p$, append a short suffix that makes the model name a
code, and read the chapter margin (target: the letter eventually emitted; competitors: the other inventory letters) at
the suffix's last token. This asks which region the residual stream at $p$ supports. The suffix is
``\texttt{\textbackslash nSo the code here is}'' (suffix B). Four wordings were compared at the code token on a
two-note pilot, and B was best there. On the full corpus the four wordings make the emitted letter decisive at the
code token (margin at least 2 nats) for between \NMCProbeAgreementLowest{} and \NMCProbeDAgreement{}
of mentions, the chosen one for \NMCProbeBAgreement{}; the best is suffix D
(``\texttt{\ The corresponding ICD-10-CM code is}''). To check that the locus does not depend on the wording, the
walk-back was re-run with suffix D and with suffix A, the pilot's weakest, on \NLocusSuffixNotes{} of the 100 notes
(\NLocusSuffixBMentions{} chapter mentions; Table~\ref{tab:app-suffix}). The median distance is \NLocusSuffixBDMedian,
\NLocusSuffixDDMedian{} and \NLocusSuffixADMedian{} tokens under B, D and A; the two alternatives place a few more mentions
in the window (\NLocusSuffixDWindowPct{} and \NLocusSuffixAWindowPct{} against \NLocusSuffixBWindowPct), D with fewer probe
failures and A with fewer inherited loci; the category agrees per mention on \NLocusSuffixDAgree{} (D) and
\NLocusSuffixAAgree{} (A); and where two wordings both place the decision in the window their distances differ by a
median of \NLocusSuffixDDDiffMedian{} token (\NLocusSuffixDDWithinEight{} within one 8-token step). The locus is a property
of the trace, not of the suffix. A suffix
asks for \emph{the} code, and the model answers with its
best-supported one, so a mention that the model lists without endorsing it (a differential, a rejected candidate)
does not register. This is the probe's known limit.

\paragraph{Locus.} Walk back from the code token in steps of 8 tokens, up to 512 tokens or the start of the trace,
until the forced margin first drops below $\theta=2$ nats; record four more steps for the shape of the curve; then
refine the last crossing to the token. The locus is the earliest position from which the forced margin stays at least
$\theta$ all the way to the code token. Each mention takes 4--9 probes, each a short forward against the cropped cache
(about 0.1\,s); the cache is never copied but cropped and, when a later probe needs more of it, regrown from the token
ids. Every chapter mention receives one of five outcomes: \emph{at the code token} (decisive only there, $d=0$);
\emph{in the reasoning window} (the locus lies $d\ge 1$ tokens before the letter and after the end of the previous
mention's code); \emph{inherited} (the locus falls at or before the end of the previous mention's code, so the region
was settled with an earlier code; such mentions are counted, not attributed); \emph{probe failure} (not decisive even
with the cache cut one token before the letter); and \emph{censored} (still decisive 512 tokens back).
Table~\ref{tab:app-locus-outcomes} gives the outcomes for all 2{,}064 chapter mentions (18{,}400 probes) by kind of
mention; \NMCLocusCensoredPct{} of mentions are censored.

\begin{table}[!htbp]
  \centering\small
  \caption{The locus under three forcing suffixes on the same \NLocusSuffixNotes{} notes and \NLocusSuffixBMentions{}
  chapter mentions: outcome shares, the median distance of the window loci, and the per-mention agreement with the
  paper's suffix B. \resid{R-C13}}
  \label{tab:app-suffix}
  \providecommand{\gensd}[2]{\begin{tabular}[t]{@{}c@{}}#1\\[-2pt]{\scriptsize$\pm$#2}\end{tabular}}
\providecommand{\genstack}[2]{\begin{tabular}[t]{@{}c@{}}#1\\[-2pt]{\scriptsize #2}\end{tabular}}
\providecommand{\genstackl}[2]{\begin{tabular}[t]{@{}l@{}}#1\\[-2pt]{\scriptsize #2}\end{tabular}}
\providecommand{\genhead}[2]{\begin{tabular}[b]{@{}c@{}}#1\\#2\end{tabular}}
\providecommand{\genstackiii}[3]{\begin{tabular}[t]{@{}c@{}}#1\\[-2pt]{\scriptsize #2}\\[-3pt]{\scriptsize #3}\end{tabular}}
\footnotesize\setlength{\tabcolsep}{4pt}
\begin{adjustbox}{max width=\linewidth}
\begin{tabular}{@{}l cccc c c@{}}
\toprule
Forcing suffix & window & inherited & probe fails & at $q$ & \genhead{window: median $d$}{(IQR; $n$)} & \genhead{vs B: category agrees;}{median $|\Delta d|$ (within 8)} \\
\midrule
B ``\texttt{\textbackslash nSo the code here is}'' (the paper's) & 36.1\% & 29.7\% & 31.7\% & 2.3\% & 7 (3--21; $n$ 671) & --- \\
D ``\texttt{\ The corresponding ICD-10-CM code is}'' & 43.2\% & 28.7\% & 27.1\% & 0.6\% & 8 (4--20; $n$ 804) & 80.1\%; 1 (85\%) \\
A (the pilot's weakest) & 43.6\% & 22.5\% & 31.1\% & 2.3\% & 7 (3--18; $n$ 812) & 79.4\%; 1 (85\%) \\
\bottomrule
\end{tabular}
\end{adjustbox}

\end{table}

\begin{table}[!htbp]
  \centering\small
  \caption{Where the chapter can be elicited: the forcing probe's outcome for every chapter mention, as a share of each
  column's mentions. ``Previous code'' is the code mentioned just before, in the same or in another chapter.
  \resid{R-C6}}
  \label{tab:app-locus-outcomes}
  \begin{adjustbox}{max width=\linewidth}
  \begin{tabular}{@{}lcccccc@{}}
    \toprule
     & all & gold codes & wrong codes & \multicolumn{2}{c}{previous code} & first code \\
    \cmidrule(lr){5-6}
     & & & & same chapter & other chapter & of the note \\
    \midrule
    mentions & 2{,}064 & 226 & 1{,}838 & 906 & 1{,}058 & 100 \\
    at the code token & \NLocusAtQPct & \NMCLocusAtQPctGold{} & \NMCLocusAtQPctWrong{}
      & \NMCLocusAtQPctPrevSame{} & \NMCLocusAtQPctPrevOther{} & \NMCLocusAtQPctFirst{} \\
    in the reasoning window & \NLocusWindowPct & \NMCLocusWindowPctGold{} & \NMCLocusWindowPctWrong{}
      & \NMCLocusWindowPctPrevSame{} & \NMCLocusWindowPctPrevOther{}
      & \NMCLocusWindowPctFirst{} \\
    inherited from an earlier code & \NLocusInheritedPct & \NMCLocusInheritedPctGold{}
      & \NMCLocusInheritedPctWrong{} & \NMCLocusInheritedPctPrevSame{}
      & \NMCLocusInheritedPctPrevOther{} & \NMCLocusInheritedPctFirst{} \\
    probe failure & \NLocusProbeFailPct & \NMCLocusProbeFailPctGold{} & \NMCLocusProbeFailPctWrong{}
      & \NMCLocusProbeFailPctPrevSame{} & \NMCLocusProbeFailPctPrevOther{}
      & \NMCLocusProbeFailPctFirst{} \\
    \bottomrule
  \end{tabular}
  \end{adjustbox}
\end{table}

\paragraph{Distance, step size and robustness.} For the 750 mentions decided in the reasoning window, the locus lies a
median of \NLocusDMedian{} tokens before the letter (10th--90th percentile \NMCLocusDPTen{} to
\NMCLocusDPNinety{}); the first code of a note is decided further back (median \NLocusFirstCodeD{}). Per note, the
window share is \NMCLocusWindowSharePerNoteMedian{} (10th--90th percentile
\NMCLocusWindowSharePerNotePTen{} to \NMCLocusWindowSharePerNotePNinety{}) and the inherited share
\NMCLocusInheritedSharePerNoteMedian{} (\NMCLocusInheritedSharePerNotePTen{} to
\NMCLocusInheritedSharePerNotePNinety{}). Across the 8-token step in which the forced margin crosses $\theta$
it jumps by a median of \NLocusJump{} nats. The curve returns above $\theta$ at some earlier grid point for
\NMCLocusFlickerPct{} of mentions, which is why the locus is the earliest position from which the margin
\emph{stays} decisive; the 8-token grid is fine enough. Lowering $\theta$ to 1 or raising it to 3 moves the share
decisive at the code token from \NMCProbeBAgreement{} to \NMCLocusThetaOneDecisiveAtQ{} or
\NMCLocusThetaThreeDecisiveAtQ{}. The inherited mass is inheritance within a chapter:
\NMCLocusInheritedPctPrevSame{} of the mentions whose previous code shares their chapter, and
\NMCLocusInheritedPctSibling{} of the mentions of a sibling of an earlier code, are decided with the earlier
code and never re-decided. It is not evidence of narrowing far upstream.

\paragraph{The note alone.} With the cache cropped at the end of the prompt, before anything is written, the forced
margin is decisive for the emitted letter for \NMCNoteAloneDecisivePct{} of mentions,
\NMCNoteAloneDecisiveFirstCodesPct{} of first codes and \NMCNoteAloneDecisiveGoldFirstCodesPct{}
of gold first codes (median \NMCNoteAloneMedianMargin{} nats). The region is not read off the note; it is
written down in the reasoning.

\paragraph{What the locus token is.} Trace tokens were classed by the word they belong to: a word of the emitted code's
own inventory description (at least four letters, with structural words such as ``unspecified'' removed), a word of
another description in the same chapter, a word of a description in another chapter, part of a code string,
punctuation, or other prose (Table~\ref{tab:app-locus-classes}). \NLocusOwnDescPct{} of the loci are a word of the
emitted code's own description, \NLocusOwnDescLift{} times its base rate, and together with the words of other
descriptions in the emitted chapter they make up more than half of the loci.

\begin{table}[!htbp]
  \centering\small
  \caption{The locus token, for the 750 mentions decided in the reasoning window: share of each class among all
  attribution-window tokens (base rate), share among loci, and lift. \resid{R-C6}}
  \label{tab:app-locus-classes}
  \begin{adjustbox}{max width=\linewidth}
  \begin{tabular}{@{}lccc@{}}
    \toprule
    Class of token & base rate & at the locus & lift \\
    \midrule
    word of the emitted code's own description & \NMCLocusClassOwnDescBase{} & \NLocusOwnDescPct
      & \NLocusOwnDescLift \\
    word of another description in the same chapter & \NMCLocusClassChapterDescBase{}
      & \NMCLocusClassChapterDescShare{} & \NMCLocusClassChapterDescLift{} \\
    other prose & \NMCLocusClassProseBase{} & \NMCLocusClassProseShare{}
      & \NMCLocusClassProseLift{} \\
    word of a description in another chapter & \NMCLocusClassOtherDescBase{}
      & \NMCLocusClassOtherDescShare{} & \NMCLocusClassOtherDescLift{} \\
    punctuation & \NMCLocusClassPunctBase{} & \NMCLocusClassPunctShare{}
      & \NMCLocusClassPunctLift{} \\
    inside a code string & \NMCLocusClassCodeBase{} & \NMCLocusClassCodeShare{}
      & \NMCLocusClassCodeLift{} \\
    \bottomrule
  \end{tabular}
  \end{adjustbox}
\end{table}

\paragraph{Window attribution.} One forward of the 64 tokens that end at the locus, plus the suffix, against the
cropped cache, with every head's output at every one of those positions exposed as a gradient leaf (graph intact), and
one backward from the forced margin give $A[p,h]=(\mu_h-z_h(p))\cdot\partial\margin/\partial z_h(p)$: the first-order
effect of mean-ablating head $h$ at position $p$ on the decision made there. Heads at the window positions are the
prose-time computation; heads at the suffix positions are its transport into an emission. \Pnull{} events (the same
20 prose positions per trace, with a random letter as target) get the same window and set the floor: prose-time
importance is $I_c=I-I_{\Pnull}$ summed over the window positions, and transport importance the same over the suffix
positions. The pass ran on the 804 attributed mentions (750 decided in the window, 48 at the code token, 6 censored)
and on all 2{,}000 \Pnull{} positions. About half of a chapter decision's positive importance is at the suffix, the
transport; the rest is in the prose, concentrated at the locus token and the three tokens before it, more so than at
\Pnull{} positions, where nothing was decided (Table~\ref{tab:app-locus-position}). Importance falls off with distance
rather than piling up at the window's far edge, so 64 tokens are enough.

\begin{table}[!htbp]
  \centering\small
  \caption{Where the importance sits: share of each event's positive importance (window plus suffix, normalised to
  one) at the locus token, 1--3 tokens before it, 4--15 before, 16--63 before, and at the suffix. \resid{R-C6}}
  \label{tab:app-locus-position}
  \begin{adjustbox}{max width=\linewidth}
  \begin{tabular}{@{}lccccc@{}}
    \toprule
    Events & locus token & 1--3 before & 4--15 before & 16--63 before & suffix \\
    \midrule
    chapter, decided in the window (750) & \NMCImpWindowLocus{} & \NMCImpWindowOneToThree{}
      & \NMCImpWindowFourToFifteen{} & \NMCImpWindowSixteenToSixtyThree{} & \NMCImpWindowSuffix{} \\
    chapter, decided at the code token (48) & \NMCImpAtQLocus{} & \NMCImpAtQOneToThree{}
      & \NMCImpAtQFourToFifteen{} & \NMCImpAtQSixteenToSixtyThree{} & \NMCImpAtQSuffix{} \\
    \Pnull, random letter (2{,}000) & \NMCImpNullLocus{} & \NMCImpNullOneToThree{}
      & \NMCImpNullFourToFifteen{} & \NMCImpNullSixteenToSixtyThree{} & \NMCImpNullSuffix{} \\
    \bottomrule
  \end{tabular}
  \end{adjustbox}
\end{table}

\paragraph{The prose-time ranking.} Table~\ref{tab:app-prose-ranking} gives the prose-time set's top-16; every interval
excludes zero. The set shares \NLocusOverlapEmissionKEight{} of its top-8, \NLocusOverlapEmissionKSixteen{} of its
top-16 and \NLocusOverlapEmissionKForty{} of its top-40 heads with the emission set (the chapter global set), and the
two importances correlate at Spearman \NMCProseEmissionSpearman{} over all 1{,}600 heads. Its top-40 has
\NLocusTopFortyLayersZeroToEleven{} heads in layers 0--11, \NLocusTopFortyLayersTwelveToTwentyThree{} in layers 12--23
(\NMCProseTopFortyLTwelveToTwentyThreeImportanceShare{} of its importance), \NLocusTopFortyLayersTwentyFourToThirtyOne{} in
layers 24--31 and \NLocusTopFortyLayersThirtyTwoToThirtyNine{} in layers 32--39: one block earlier than the emission
set (Table~\ref{tab:heads-app}). The one emission top-16 head in the prose-time top-16 is the region reader
\head{24}{1}, third; the region readers \head{26}{37} and \head{26}{36} rank \NMCProseRankLTwentySixHThirtySeven{} and
\NMCProseRankLTwentySixHThirtySix{}; and the emission set's final-layer prose readers (\head{39}{38}, \head{39}{13},
\head{38}{24}) are not in the prose-time top-40. Ranked on the 750 window-decided mentions alone, the top-16 has the same heads with small reorderings.
\head{24}{19}, \head{24}{1} and \head{22}{16} split their importance between the prose and the transport
(\NMCLTwentyFourHNineteenSuffixShare{}, \NMCLTwentyFourHOneSuffixShare{} and \NMCLTwentyTwoHSixteenSuffixShare{} of it at the
suffix), whereas \head{18}{4}, \head{22}{15} and \head{18}{3} act mostly in the prose, at and before the locus token
(at most \NMCProseActingHeadsSuffixShareMax{} at the suffix).

\begin{table}[!htbp]
  \centering\small
  \caption{The prose-time ranking: decision-specific importance summed over the attribution-window positions (nats;
  mean over notes) with its 95\% bootstrap interval, and the head's rank and $I_c$ in the emission-time chapter
  ranking. \resid{R-C6}}
  \label{tab:app-prose-ranking}
  \begin{adjustbox}{max width=\linewidth}
  \begin{tabular}{@{}lcccc@{}}
    \toprule
    Head & $I_c$ (prose-time) & 95\% interval & emission rank & emission $I_c$ \\
    \midrule
    \head{24}{19} & \NMCProseIcLTwentyFourHNineteen{} & [\NMCProseCiLoLTwentyFourHNineteen{}, \NMCProseCiHiLTwentyFourHNineteen{}]
      & \NMCRankPOneLTwentyFourHNineteen{} & \NMCIcPOneLTwentyFourHNineteen{} \\
    \head{22}{16} & \NMCProseIcLTwentyTwoHSixteen{} & [\NMCProseCiLoLTwentyTwoHSixteen{}, \NMCProseCiHiLTwentyTwoHSixteen{}]
      & \NMCRankPOneLTwentyTwoHSixteen{} & \NMCIcPOneLTwentyTwoHSixteen{} \\
    \head{24}{1} & \NMCProseIcLTwentyFourHOne{} & [\NMCProseCiLoLTwentyFourHOne{}, \NMCProseCiHiLTwentyFourHOne{}]
      & \NMCRankPOneLTwentyFourHOne{} & \NMCIcPOneLTwentyFourHOne{} \\
    \head{18}{4} & \NMCProseIcLEighteenHFour{} & [\NMCProseCiLoLEighteenHFour{}, \NMCProseCiHiLEighteenHFour{}]
      & \NMCRankPOneLEighteenHFour{} & \NMCIcPOneLEighteenHFour{} \\
    \head{22}{15} & \NMCProseIcLTwentyTwoHFifteen{} & [\NMCProseCiLoLTwentyTwoHFifteen{}, \NMCProseCiHiLTwentyTwoHFifteen{}]
      & \NMCRankPOneLTwentyTwoHFifteen{} & \NMCIcPOneLTwentyTwoHFifteen{} \\
    \head{21}{11} & \NMCProseIcLTwentyOneHEleven{} & [\NMCProseCiLoLTwentyOneHEleven{}, \NMCProseCiHiLTwentyOneHEleven{}]
      & \NMCRankPOneLTwentyOneHEleven{} & \NMCIcPOneLTwentyOneHEleven{} \\
    \head{18}{3} & \NMCProseIcLEighteenHThree{} & [\NMCProseCiLoLEighteenHThree{}, \NMCProseCiHiLEighteenHThree{}]
      & \NMCRankPOneLEighteenHThree{} & \NMCIcPOneLEighteenHThree{} \\
    \head{22}{17} & \NMCProseIcLTwentyTwoHSeventeen{} & [\NMCProseCiLoLTwentyTwoHSeventeen{}, \NMCProseCiHiLTwentyTwoHSeventeen{}]
      & \NMCRankPOneLTwentyTwoHSeventeen{} & \NMCIcPOneLTwentyTwoHSeventeen{} \\
    \head{22}{18} & \NMCProseIcLTwentyTwoHEighteen{} & [\NMCProseCiLoLTwentyTwoHEighteen{}, \NMCProseCiHiLTwentyTwoHEighteen{}]
      & \NMCRankPOneLTwentyTwoHEighteen{} & \NMCIcPOneLTwentyTwoHEighteen{} \\
    \head{20}{19} & \NMCProseIcLTwentyHNineteen{} & [\NMCProseCiLoLTwentyHNineteen{}, \NMCProseCiHiLTwentyHNineteen{}]
      & \NMCRankPOneLTwentyHNineteen{} & \NMCIcPOneLTwentyHNineteen{} \\
    \head{24}{16} & \NMCProseIcLTwentyFourHSixteen{} & [\NMCProseCiLoLTwentyFourHSixteen{}, \NMCProseCiHiLTwentyFourHSixteen{}]
      & \NMCRankPOneLTwentyFourHSixteen{} & \NMCIcPOneLTwentyFourHSixteen{} \\
    \head{19}{38} & \NMCProseIcLNineteenHThirtyEight{} & [\NMCProseCiLoLNineteenHThirtyEight{}, \NMCProseCiHiLNineteenHThirtyEight{}]
      & \NMCRankPOneLNineteenHThirtyEight{} & \NMCIcPOneLNineteenHThirtyEight{} \\
    \head{21}{1} & \NMCProseIcLTwentyOneHOne{} & [\NMCProseCiLoLTwentyOneHOne{}, \NMCProseCiHiLTwentyOneHOne{}]
      & \NMCRankPOneLTwentyOneHOne{} & \NMCIcPOneLTwentyOneHOne{} \\
    \head{19}{37} & \NMCProseIcLNineteenHThirtySeven{} & [\NMCProseCiLoLNineteenHThirtySeven{}, \NMCProseCiHiLNineteenHThirtySeven{}]
      & \NMCRankPOneLNineteenHThirtySeven{} & \NMCIcPOneLNineteenHThirtySeven{} \\
    \head{16}{38} & \NMCProseIcLSixteenHThirtyEight{} & [\NMCProseCiLoLSixteenHThirtyEight{}, \NMCProseCiHiLSixteenHThirtyEight{}]
      & \NMCRankPOneLSixteenHThirtyEight{} & \NMCIcPOneLSixteenHThirtyEight{} \\
    \head{17}{18} & \NMCProseIcLSeventeenHEighteen{} & [\NMCProseCiLoLSeventeenHEighteen{}, \NMCProseCiHiLSeventeenHEighteen{}]
      & \NMCRankPOneLSeventeenHEighteen{} & \NMCIcPOneLSeventeenHEighteen{} \\
    \bottomrule
  \end{tabular}
  \end{adjustbox}
\end{table}

\paragraph{Suffix-position importance.} Summed over the suffix positions instead, $I_c$ ranks \head{21}{25} first
(\NMCTransportIcLTwentyOneHTwentyFive{} nats), followed by \head{26}{37}, \head{27}{13}, \head{25}{23}, \head{26}{36},
\head{20}{20}, \head{26}{18}, \head{27}{11} and \head{24}{1} (\NMCTransportIcLTwentySixHThirtySeven{} down to
\NMCTransportIcLTwentyFourHOne{}); the prose-time heads \head{22}{16} and \head{24}{19} rank
\NMCTransportRankLTwentyTwoHSixteen{} and \NMCTransportRankLTwentyFourHNineteen{}, and \head{39}{38} and \head{39}{13} rank
\NMCTransportRankLThirtyNineHThirtyEight{} and \NMCTransportRankLThirtyNineHThirteen{}. The transport ranking overlaps the emission
set (\NMCTransportOverlapTopEight{} of the top-8, \NMCTransportOverlapTopSixteen{} of the top-16 and
\NMCTransportOverlapTopForty{} of the top-40; Spearman \NMCTransportEmissionSpearman{} over all heads and
\NMCTransportEmissionSpearmanTopOneHundred{} over the emission top-100) and also contains the layer-22 and
layer-24 prose-time heads. Its top-16 has \NMCTransportTopSixteenLTwelveToTwentyThreeCount{} heads in layers 12--23
(\NMCTransportTopSixteenLTwelveToTwentyThreeShare{} of its importance), \NMCTransportTopSixteenLTwentyFourToThirtyOneCount{} in layers
24--31 (\NMCTransportTopSixteenLTwentyFourToThirtyOneShare{}) and \NMCTransportTopSixteenLThirtyTwoToThirtyNineCount{} in layers 32--39. One
suffix and one decision give two rankings: what the forced probe elicits in the prose is elicited mostly by heads one
block earlier than the ones that act at the suffix, and the dissociation is by position (window-only against
suffix-only ablation, below), not by disjoint heads. We called this second ranking a transport ranking in an earlier
version; the follow-through and residual-swap measurements below show that the emission layers do more than transport,
so we now name it by what it measures.

\paragraph{Exact faithfulness at the locus (pass A).} Over the window and suffix positions, the locus pass also
mean-ablated exactly the emission set's top-$k$, the event's own top-$k$ by window attribution, random-$k$ and the
emission bottom-$k$, and, at $k=16$, the two sets restricted to the window positions only and to the suffix positions
only (Table~\ref{tab:app-locus-passA}). The emission set flips the forced decision, but through the suffix: restricted
to the prose positions it removes \NLocusEmissionWinKSixteen{} of the forced margin, while the heads ranked at the
prose positions remove all of it from there (\NLocusOwnSetWinKSixteen{}). A per-event set is fitted to the event it is
tested on; the fair test is a fixed set (pass B).

\begin{table}[!htbp]
  \centering\small
  \caption{Pass A: share of the forced chapter margin destroyed by mean-ablating a set at the locus (804 mentions, 100
  notes; medians over notes of per-note means, emission set with its 10th--90th percentile). Top: over window and
  suffix positions; bottom: at $k=16$, restricted by position. \resid{R-C6}}
  \label{tab:app-locus-passA}
  \begin{adjustbox}{max width=\linewidth}
  \begin{tabular}{@{}lccccc@{}}
    \toprule
    $k$ & emission top-$k$ & event's own top-$k$ & random-$k$ & emission bottom-$k$ & emission $>$ random \\
    \midrule
    8 & \NMCLocusAEmissionTopEight{} (\NMCLocusAEmissionTopEightPTen{}--\NMCLocusAEmissionTopEightPNinety{})
      & \NMCLocusAOwnTopEight{} & \NMCLocusARandomTopEight{} & \NMCLocusABottomTopEight{}
      & \NMCLocusAEmissionWins{}/100 \\
    16 & \NMCLocusAEmissionTopSixteen{} (\NMCLocusAEmissionTopSixteenPTen{}--\NMCLocusAEmissionTopSixteenPNinety{})
      & \NMCLocusAOwnTopSixteen{} & \NMCLocusARandomTopSixteen{} & \NMCLocusABottomTopSixteen{}
      & \NMCLocusAEmissionWins{}/100 \\
    40 & \NMCLocusAEmissionTopForty{} (\NMCLocusAEmissionTopFortyPTen{}--\NMCLocusAEmissionTopFortyPNinety{})
      & \NMCLocusAOwnTopForty{} & \NMCLocusARandomTopForty{} & \NMCLocusABottomTopForty{}
      & \NMCLocusAEmissionWins{}/100 \\
    \midrule
    Set, $k=16$ & window positions only & suffix positions only & both & & \\
    \midrule
    emission top-16 & \NLocusEmissionWinKSixteen & \NMLocusEmissionSufKSixteen & \NMCLocusAEmissionTopSixteen{} & & \\
    event's own top-16 & \NLocusOwnSetWinKSixteen & \NMCLocusAOwnSuffixTopSixteen{} & \NMCLocusAOwnTopSixteen{}
      & & \\
    \bottomrule
  \end{tabular}
  \end{adjustbox}
\end{table}

\paragraph{The fixed prose-time set (pass B).} The global prose-time set, the same top-$k$ on every note, was
mean-ablated at the locus against count-matched random heads and the bottom of the prose-time ranking
(Table~\ref{tab:app-locus-passB}). Eight heads removed only at the prose positions remove the whole forced decision,
where the emission set's top-16 removed \NLocusEmissionWinKSixteen{} from the same positions; the same eight heads at
the suffix remove \NMCLocusBProseSuffixTopEight{}, the mirror image of the emission set
(\NMLocusEmissionSufKSixteen{} at the suffix). On the 750 window-decided mentions alone the numbers agree within
\NMCLocusBWindowDecidedAgreement{}. The chapter is elicitable in the prose through one set of heads and written at the
code token by another.

\paragraph{From elicitable to decided: the follow-through.} Passes A and B read the \emph{forced} margin at the locus.
Whether the chapter computed there is the one the code carries is a different question: the original continuation could
re-derive the chapter at the code token. We therefore ablated each set over the attribution window as in pass B and then
let the original continuation run, teacher-forced, to the chapter letter, and read the natural margin there
(Table~\ref{tab:app-follow}; the \NFollowMentions{} window-decided mentions of the 100 notes). Ablating the prose-time
top-16 over the window removes \NFollowProseSixteenDF{} nats of the forced margin at the locus but a median of
\NFollowProseSixteenDNWin{} nats of the natural margin at the code (mean \NFollowProseSixteenDNWinMean), a ratio of
\NFollowProseSixteenRatio{} over the \NFollowProseSixteenNBig{} mentions whose forced drop exceeds one nat; the natural
drop is above random heads' on \NFollowProseVsRandomNoteHigher{} of 100 notes (per-note mean difference
\NFollowProseVsRandomNoteMean{} nats [\NFollowProseVsRandomNoteLo, \NFollowProseVsRandomNoteHi], sign test $p$
\NFollowProseVsRandomNoteP) and flips \NFollowProseSixteenFlipped{} of the decisions. Ablating the same heads from the
window all the way to the code removes \NFollowProseSixteenDNAll{} nats, as much as at the locus, and ablating the
emission top-16 at the code token alone removes \NFollowEmissionAtCodeDrop. The prose-time computation feeds the
decision, but the code largely re-derives the chapter: ``elicitable in the reasoning'' is the right description, and
``decided in the reasoning'' is not. The pre-registered reading placed the boundary at a ratio of 0.25 for ``elicited,
not needed'' against 0.25--0.5 for ``partly decided''; the measured \NFollowProseSixteenRatio{} sits at that boundary, and
we use the weaker word throughout.

\begin{table}[!htbp]
  \centering\small
  \caption{Follow-through of the window ablation to the code: for each set mean-ablated over the attribution window,
  the forced margin removed at the locus and the natural margin removed at the chapter letter after the original
  continuation runs; the same with the ablation sustained from the window to the code; the ratio natural / forced on
  mentions whose forced drop exceeds one nat; and the share of natural decisions flipped. Medians over the
  \NFollowMentions{} window-decided mentions (means in parentheses). \resid{R-C12}}
  \label{tab:app-follow}
  \providecommand{\gensd}[2]{\begin{tabular}[t]{@{}c@{}}#1\\[-2pt]{\scriptsize$\pm$#2}\end{tabular}}
\providecommand{\genstack}[2]{\begin{tabular}[t]{@{}c@{}}#1\\[-2pt]{\scriptsize #2}\end{tabular}}
\providecommand{\genstackl}[2]{\begin{tabular}[t]{@{}l@{}}#1\\[-2pt]{\scriptsize #2}\end{tabular}}
\providecommand{\genhead}[2]{\begin{tabular}[b]{@{}c@{}}#1\\#2\end{tabular}}
\providecommand{\genstackiii}[3]{\begin{tabular}[t]{@{}c@{}}#1\\[-2pt]{\scriptsize #2}\\[-3pt]{\scriptsize #3}\end{tabular}}
\footnotesize\setlength{\tabcolsep}{4pt}
\begin{adjustbox}{max width=\linewidth}
\begin{tabular}{@{}l c c c c c@{}}
\toprule
Set mean-ablated & \genhead{forced drop at the locus}{window only} & \genhead{natural drop at the code}{window only: median (mean)} & \genhead{natural drop}{window $\to$ code} & \genhead{ratio natural / forced}{(mentions with $\ge 1$ nat)} & flipped \\
\midrule
prose-time top-8 & +3.27 & +0.65 (+1.11) & +3.45 & 0.22 (648) & 2.9\% \\
prose-time top-16 & +4.46 & +0.84 (+1.37) & +4.50 & 0.23 (686) & 4.0\% \\
emission top-16 & +0.45 & +0.27 (+0.37) & +5.40 & 0.40 (134) & 0.3\% \\
random-16 & \ensuremath{-}0.01 & \ensuremath{-}0.01 (\ensuremath{-}0.03) & +0.10 & 0.08 (5) & 0.0\% \\
prose-time bottom-16 & \ensuremath{-}0.17 & +0.41 (+0.54) & \ensuremath{-}0.19 & 0.50 (117) & 1.5\% \\
\bottomrule
\end{tabular}
\end{adjustbox}

\end{table}

\begin{figure}[!htbp]
  \centering
  \includegraphics[width=\linewidth]{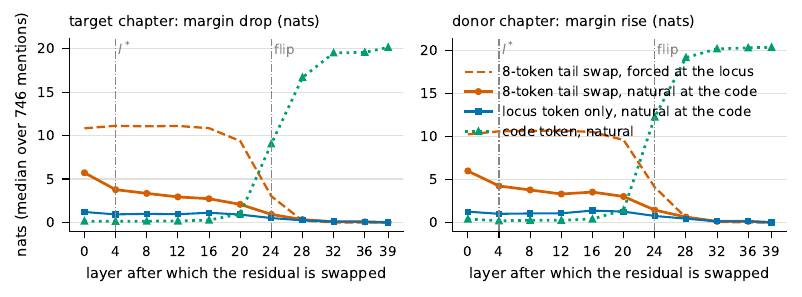}
  \caption{Residual swap by layer: the residual stream after layer $l$ is replaced by a different-chapter mention's at
  the 8-token tail before the locus (forced read at the locus, dashed; natural read at the code, solid), at the locus
  token only, or at the code token itself (dotted). Left: the target chapter's margin drop; right: the donor chapter's
  rise; medians over \NResidMentions{} mentions. Guides: $l^*$ and the layer from which the code-token swap flips the
  decision. \resid{R-C12}}
  \label{fig:resid}
\end{figure}

\paragraph{Swapping the residual stream.} The same question at the level of the residual stream, with one figure
(Figure~\ref{fig:resid}): for each window-decided mention with a different-chapter window mention in the same note as
donor (\NResidMentions{} mentions in \NResidNotes{} notes), the residual stream after layer $l$ was captured from the
donor's clean forward and written into the target's forward at one of three places: the locus token only, the last 8
window positions (the tail), or the code token; the forced margin at the locus and the natural margin at the code were
read for the target's and for the donor's chapter letter. The analysis layer $l^*$, fixed from the forced data alone as
the grid layer with the largest median forced drop under the tail swap, is $l^*=$ L\NResidLStar: the forced profile is
flat from the first layers to L16, so a swap there replaces little more than the eight words. At $l^*$ the tail swap
removes \NResidForcedDropLStar{} nats of the forced margin and \NResidNaturalDropLStar{} of the natural margin at the code,
and raises the donor letter's natural margin by \NResidDonorRiseLStar{}, on \NResidNaturalDropNoteHigher{} of
\NResidNotes{} notes (per-note mean drop \NResidNaturalDropNoteMean{} nats [\NResidNaturalDropNoteLo,
\NResidNaturalDropNoteHi]); the transport ratio, natural over forced, is \NResidRatio{} in median over the \NResidNBig{}
mentions with a forced drop above one nat, against the pre-registered 0.5 for ``transport''. Deeper, where the prose-time
set acts (L12--L20), the ratio of the median drops falls to \NResidRatioMediansLTwelve{}--\NResidRatioMediansLTwenty{},
which is the follow-through ratio above. Swapping the code token's own residual has no effect before L16 and hands the
natural decision to the donor from L\NResidFlipLayer{} on (median target drop \NResidCodeLTwentyFourDrop{} nats at L24 and
\NResidCodeLTwentyEightDrop{} at L28): the chapter enters the code token in layers 20--28, the emission layers, by
attention to the context. The prose words before the locus carry about half of what forces the chapter there and the
mid-layer state computed at the locus about a quarter; the emission layers decide the chapter at the code, reading the
prose the reasoning has just written. Together with the follow-through this fixes the wording of
Section~\ref{sec:char-locus}: the prose makes the chapter elicitable, and the emission set decides it at the code rather
than transporting a decision already made.

\begin{table}[!htbp]
  \centering\small
  \caption{Pass B: share of the forced chapter margin destroyed by mean-ablating the fixed global prose-time top-$k$
  at the window positions only (with its 10th--90th percentile over notes), at the suffix positions only, and at
  both, against random-$k$ and the prose-time bottom-$k$ at the window positions (804 mentions, 100 notes; medians over
  notes of per-note means), and the notes on which the prose-time set beats random. \resid{R-C6}}
  \label{tab:app-locus-passB}
  \begin{adjustbox}{max width=\linewidth}
  \begin{tabular}{@{}lcccccc@{}}
    \toprule
     & \multicolumn{3}{c}{Prose-time top-$k$} & random-$k$ & bottom-$k$ & prose-time $>$ random \\
    \cmidrule(lr){2-4}
    $k$ & window only & suffix only & both & window & window & (notes of 100) \\
    \midrule
    8 & \NLocusProseWinKEight{} (\NMCLocusBProseWindowTopEightPTen{}--\NMCLocusBProseWindowTopEightPNinety{})
      & \NMCLocusBProseSuffixTopEight{} & \NMCLocusBProseBothTopEight{} & \NLocusRandWinKEight
      & \NMCLocusBBottomWindowTopEight{} & \NLocusProseWinKEightWins \\
    16 & \NLocusProseWinKSixteen{} (\NMCLocusBProseWindowTopSixteenPTen{}--\NMCLocusBProseWindowTopSixteenPNinety{})
      & \NMLocusProseSufKSixteen & \NMCLocusBProseBothTopSixteen{} & \NLocusRandWinKSixteen
      & \NMCLocusBBottomWindowTopSixteen{} & \NLocusProseWinKSixteenWins \\
    40 & \NLocusProseWinKForty{} (\NMCLocusBProseWindowTopFortyPTen{}--\NMCLocusBProseWindowTopFortyPNinety{})
      & \NMCLocusBProseSuffixTopForty{} & \NMCLocusBProseBothTopForty{} & \NLocusRandWinKForty
      & \NMCLocusBBottomWindowTopForty{} & \NLocusProseWinKFortyWins \\
    \bottomrule
  \end{tabular}
  \end{adjustbox}
\end{table}

\paragraph{What the prose-time heads read.} Attention rows were read at the locus token in the natural context (no
suffix) for four 16-head groups: the prose-time top-16, the emission top-16, the prose-time bottom-16 and 16 random
heads (Table~\ref{tab:app-locus-read}). The prose-time set's share on the emitted chapter's lines beats the random
panel's on \NMCLocusReadRegionAboveRandomNotes{} of 100 notes
(\NMCLocusReadRegionAboveRandomNotesTopPositions{} at the three window positions where the set's
importance is largest), but its distinctive reading is the note: \NLocusReadProseNote{} of its attention, against
\NLocusReadRandomNote{} for the random heads and \NMCPOneTopSixteenOnNote{} for the emission top-16 at the code
token (Table~\ref{tab:app-read-group}). Per head there are two families, different from those at the code token.
\emph{Note readers} in layers 18--24 (\head{24}{19}, \head{24}{1}, \head{22}{17}, \head{21}{1}, \head{21}{11},
\head{18}{4}, \head{18}{3}, \head{24}{16}) put \NMCLocusNoteReadersNoteMin{} to
\NMCLocusNoteReadersNoteMax{} of their attention on the discharge summary, with a sharp
\NMCLocusNoteReadersRegionMin{} to \NMCLocusNoteReadersRegionMax{} on the emitted chapter's
lines (lifts \NMCLocusNoteReadersLiftMin{} to \NMCLocusNoteReadersLiftMax{}, over
\NMCLocusNoteReadersNeffMin{} to \NMCLocusNoteReadersNeffMax{} effective inventory tokens).
\emph{Prose readers} in layers 17--22 (\head{22}{16}, \head{22}{18}, \head{19}{38}, \head{19}{37}, \head{17}{18},
\head{20}{19}) put \NMCLocusProseReadersProseMin{} to \NMCLocusProseReadersProseMax{} on the
reasoning so far. One head, \head{22}{15}, reads the region at the locus (\NMCLTwentyTwoHFifteenOnRegionLocus{}, lift
\NMCLTwentyTwoHFifteenLiftLocus{}). The link between reading the region and mattering is weak here ($r$ at most
\NMCLocusRRegionImportanceMax{}): at the locus the prose-time heads' effect does not depend on how much of
the inventory they read. This is consistent with a narrowing driven by the note's evidence and the reasoning, which
the emission set's region readers then match against the inventory lines at the code token
(Appendix~\ref{app:readside}).

\begin{table}[!htbp]
  \centering\small
  \caption{What the prose-time heads read: attention at the locus token in the natural context (no suffix), mean over
  the 804 loci, for four 16-head groups: share on the region (the emitted chapter's lines), on other chapters' lines,
  on the note, on the reasoning prose and on the sink; region lift; $n_{\mathrm{eff}}$ over inventory tokens.
  Descriptive: attention was not intervened on. \resid{R-C6}}
  \label{tab:app-locus-read}
  \begin{adjustbox}{max width=\linewidth}
  \begin{tabular}{@{}lccccccc@{}}
    \toprule
    Group & region & other chapters & note & prose & sink & lift & $n_{\mathrm{eff}}$ \\
    \midrule
    prose-time top-16 & \NLocusReadProseRegion & \NMCLocusReadProseOtherChapters{} & \NLocusReadProseNote
      & \NLocusReadProseProseText & \NMCLocusReadProseSink{} & \NMCLocusReadProseLift{}
      & \NMCLocusReadProseNeff{} \\
    emission top-16 & \NLocusReadEmissionRegion & \NMCLocusReadEmissionOtherChapters{}
      & \NLocusReadEmissionNote & \NLocusReadEmissionProseText & \NMCLocusReadEmissionSink{}
      & \NMCLocusReadEmissionLift{} & \NMCLocusReadEmissionNeff{} \\
    prose-time bottom-16 & \NMCLocusReadBottomRegion{} & \NMCLocusReadBottomOtherChapters{}
      & \NMCLocusReadBottomNote{} & \NMCLocusReadBottomProse{} & \NMCLocusReadBottomSink{}
      & \NMCLocusReadBottomLift{} & \NMCLocusReadBottomNeff{} \\
    random-16 & \NLocusReadRandomRegion & \NMCLocusReadRandomOtherChapters{} & \NLocusReadRandomNote
      & \NLocusReadRandomProseText & \NMCLocusReadRandomSink{} & \NMCLocusReadRandomLift{}
      & \NMCLocusReadRandomNeff{} \\
    \bottomrule
  \end{tabular}
  \end{adjustbox}
\end{table}

\paragraph{Scope.} The locus results concern endorsed mentions whose chapter becomes elicitable in the reasoning window
(\NLocusWindowPct{} of chapter mentions). Another \NLocusInheritedPct{} inherit the chapter from an earlier code and are
counted, not attributed; for \NLocusProbeFailPct{} (\NMCLocusProbeFailPctWrong{} of wrong codes and
\NMCLocusProbeFailPctGold{} of gold ones) the probe never becomes decisive, and they are excluded. We do not
claim a locus for every mention. The prose-time ranking is first-order (fidelity \NMAttrPearsonAllIntact{} under the
intact graph, Appendix~\ref{app:instrument}); its standing as a set comes from the exact ablations of passes A and B and
from the follow-through, and what it reads at the locus (Table~\ref{tab:app-locus-read}) was not intervened on.

\section{Replication on the retrieved shortlist}
\label{app:hd002}
\supports{Sec.~\ref{sec:char-stability}; \resid{R-C7}}

HD-002 ran the five passes of Appendices~\ref{app:instrument}--\ref{app:locus}, with unchanged scripts and the intact
attribution hook (its chain started on 2026-09-14 on a checkout containing the 2026-09-13 hook fix, as did the student
rankings), on new
teacher traces for the same 100 characterization notes (TR-007; decoding as in Appendix~\ref{app:teacher}), with the
full inventory replaced by the retrieved shortlist of the distillation half (Appendix~\ref{app:data}: $K=1{,}000$ plus
the siblings of the top 200, about 1{,}550 lines). On these prompts the teacher scores \microf{}
\NMCTeacherMicroFShortlistVal{} (over the 99 of 100 traces that are valid), against
\NMTeacherMicroFFullInventory{} with the full inventory (all 100 valid). Table~\ref{tab:hdcompare} compares the two
runs.

\begin{table}[!htbp]
  \centering\small
  \caption{Full inventory (HD-001) vs retrieved shortlist (HD-002) on the same 100 notes.
  \resid{R-C7}}
  \label{tab:hdcompare}
  \providecommand{\gensd}[2]{\begin{tabular}[t]{@{}c@{}}#1\\[-2pt]{\scriptsize$\pm$#2}\end{tabular}}
\providecommand{\genstack}[2]{\begin{tabular}[t]{@{}c@{}}#1\\[-2pt]{\scriptsize #2}\end{tabular}}
\providecommand{\genstackl}[2]{\begin{tabular}[t]{@{}l@{}}#1\\[-2pt]{\scriptsize #2}\end{tabular}}
\providecommand{\genhead}[2]{\begin{tabular}[b]{@{}c@{}}#1\\#2\end{tabular}}
\providecommand{\genstackiii}[3]{\begin{tabular}[t]{@{}c@{}}#1\\[-2pt]{\scriptsize #2}\\[-3pt]{\scriptsize #3}\end{tabular}}
\begin{tabular}{@{}lcc@{}}
\toprule
Quantity & Full inventory (HD-001) & Shortlist (HD-002) \\
\midrule
\multicolumn{3}{@{}l}{\textit{Contrast rankings over 1{,}600 heads (agreement between the two; \Pone{} / \Ptwo{} / \Ptwoc)}} \\
Spearman & \multicolumn{2}{c}{0.55 / 0.63 / 0.55} \\
top-40 shared (of 40) & \multicolumn{2}{c}{31 / 31 / 32} \\
top-8 shared (of 8) & \multicolumn{2}{c}{5 / 6 / 7} \\
\midrule
\multicolumn{3}{@{}l}{\textit{Global per-phase sets (median over notes; \Pone{} / \Ptwo{} / \Ptwoc)}} \\
knock-in, $k=8$ & 0.67 / 0.48 / 0.69 & 0.47 / 0.46 / 0.71 \\
knock-in, $k=40$ & 0.86 / 0.79 / 0.85 & 0.92 / 0.82 / 0.85 \\
knock-in, random-40 & 0.04 / 0.02 / 0.02 & 0.04 / 0.03 / 0.02 \\
necessity, $k=40$ & 1.24 / 1.61 / 0.86 & 1.35 / 1.57 / 0.93 \\
\midrule
\multicolumn{3}{@{}l}{\textit{Read side (top-16 heads)}} \\
\Pone{} share on the emitted chapter's lines & 0.10 & 0.14 \\
\Pone{} region lift & 3.54 & 3.04 \\
\Ptwoc{} share on the near-miss's mentions & 0.074 & 0.054 \\
\midrule
\multicolumn{3}{@{}l}{\textit{Prose locus (chapter mentions)}} \\
median distance (tokens) & 7 & 9 \\
decided in the window & 0.36 & 0.27 \\
inherited from an earlier code & 0.29 & 0.35 \\
never decisive under the probe & 0.32 & 0.37 \\
prose-time ranking: Spearman / top-40 shared & \multicolumn{2}{c}{0.79 / 31} \\
window-only ablation, locus top-16 (pass A) & 1.08 & 1.04 \\
window-only ablation, emission top-16 (pass A) & 0.16 & 0.19 \\
window-only ablation, global prose-time top-8 (pass B) & 1.00 & 0.73 \\
\bottomrule
\end{tabular}

\end{table}

\paragraph{What is preserved.} The $I_c$ rankings agree at Spearman \NShortPOneSpearman{}, \NShortPTwoSpearman{} and
\NShortPTwoCSpearman{} (\Pone, \Ptwo, \Ptwoc) over all 1{,}600 heads, with \NShortPOneOverlapForty{},
\NShortPTwoOverlapForty{} and \NShortPTwoCOverlapForty{} of the top-40 and \NShortPOneOverlapEight{},
\NShortPTwoOverlapEight{} and \NShortPTwoCOverlapEight{} of the top-8 shared. The heads persist: the chapter top-8 keeps
\head{26}{36}, \head{27}{11}, \head{27}{13} and \head{28}{21}, the character top-8 keeps \head{39}{10}, \head{39}{29},
\head{39}{12} and \head{37}{6}, and the contrastive top-8 keeps \head{30}{11}; the anchor-cosine heads of
Appendix~\ref{app:teacher} stay non-causal. The global sets keep their causal standing: top-40 knock-in
recovers \NShortPOneKiKFortyShort{}, \NShortPTwoKiKFortyShort{} and \NShortPTwoCKiKFortyShort{} against
\NShortPOneKiKFortyRandShort{}, \NShortPTwoKiKFortyRandShort{} and \NShortPTwoCKiKFortyRandShort{} for random sets,
and top-40 necessity flips the chapter and character decisions in both runs. With fewer lines to read, the chapter
top-16 put more of their attention on the emitted chapter's lines (\NShortReadPOneRegionShort{} against
\NShortReadPOneRegionFull{}) at a lower lift (\NShortReadPOneLiftShort{} against \NShortReadPOneLiftFull{}). The
chapter can be elicited a median of \NShortLocusDMedianShort{} tokens before the letter (\NShortLocusDMedianFull{} on
the full inventory) from the same prose-time set (Spearman \NShortLocusProseSpearman{}, \NShortLocusProseOverlapForty{} of
the top-40 shared; top three \head{22}{16}, \head{24}{19} and \head{22}{17}), and the dissociation reproduces:
window-only ablation of the event's own locus top-16 removes \NShortLocusLocusWinSixteenShort{} of the forced margin,
that of the emission top-16 \NShortLocusEmissionWinSixteenShort{}.

\paragraph{What shifts.} \head{39}{38} leaves the chapter top-8 on the shortlist, and the chapter top-8's knock-in is
lower (\NShortPOneKiKEightShort{} against \NShortPOneKiKEightFull{}), although the top-40's is higher. Fewer chapter
mentions are decided in the reasoning window (\NShortLocusWindowShort{} against \NShortLocusWindowFull{}), more
inherit the chapter (\NShortLocusInheritedShort{} against \NShortLocusInheritedFull{}), and more never become decisive
under the probe (\NShortLocusProbeFailShort{} against \NShortLocusProbeFailFull{}). The fixed prose-time top-8 removes
less of the forced margin from the window positions (\NShortLocusProseWinEightShort{} against
\NShortLocusProseWinEightFull{}), and \head{30}{11}'s share on the near-miss's earlier mentions is lower
(\NMCShortLThirtyHElevenOnSibling{} against \NReadLThirtyHElevenSibling{}). The pooled weights spread out: the
shortlist's top-40 is not dominated by one head (Appendix~\ref{app:heads}).

\paragraph{What the distillation half takes.} The distillation half took its targets from HD-002: the global sets and their weights
for $\wT$, and the prose-time set for the teacher's pool on the 8-token supervision window before each chapter letter
(Sec.~\ref{sec:distill-objective}). Its teacher reference values are therefore HD-002's: the knock-in rows of
Table~\ref{tab:hdcompare} for M4, the locus distance for M5, and the values in Table~\ref{tab:app-teacher-values}.
HD-001 remains the paper's primary characterization.

\begin{table}[!htbp]
  \centering\small
  \caption{Further teacher reference values used by the distillation half, on the full inventory (HD-001) and on the
  shortlist (HD-002, the values used), with the readout that uses each. Random-16 panel in brackets. \resid{R-C7}}
  \label{tab:app-teacher-values}
  \begin{adjustbox}{max width=\linewidth}
  \begin{tabular}{@{}lccl@{}}
    \toprule
    Quantity & HD-001 & HD-002 & used by \\
    \midrule
    \head{30}{11} share on the near-miss's earlier mentions & \NReadLThirtyHElevenSibling
      & \NMCShortLThirtyHElevenOnSibling{} & M3 reference \\
    \Ptwoc{} top-16 share on the code prefix & \NMCPTwoCTopSixteenOnPrefix{} (\NMCPTwoCRandomSixteenOnPrefix{})
      & \NOwnTeacherPrefix{} (\NMCShortPTwoCRandomSixteenOnPrefix{}) & prefix-share readout \\
    \Pone{} top-16 region lift & \NReadPOneTopLift{} (\NReadPOneRandLift)
      & \NShortReadPOneLiftShort{} (\NMCShortPOneRandomSixteenLift{}) & M2 at \Pone \\
    \Ptwoc{} top-16 region lift & \NReadPTwoCTopLift{} (\NReadPTwoCRandLift)
      & \NOwnTeacherMTwoPTwoCLift{} (\NOwnTeacherMTwoPTwoCLiftRand) & M2 at \Ptwoc \\
    prose-time top three & \NLocusTopHeadOne, \NLocusTopHeadTwo, \NLocusTopHeadThree
      & \head{22}{16}, \head{24}{19}, \head{22}{17} & teacher's window pool \\
    \bottomrule
  \end{tabular}
  \end{adjustbox}
\end{table}

\section{A second teacher}
\label{app:second}
\supports{Secs.~\ref{sec:char-stability} and~\ref{sec:limitations}; \resid{R-C15}}

The characterization is of one reasoning model. To ask whether its structure is that model's or the task's, we re-ran the
whole pipeline (traces, attribution, knock-in, the prose locus) on a second reasoner, \texttt{QwQ-32B} \citep{qwen2025qwq},
with the same 100 notes and all 5{,}651 candidates in context. \texttt{QwQ-32B} has \NSecondLayers{} layers of \NSecondHeads{}
heads and a different training lineage (a Qwen2.5 base trained for reasoning with reinforcement learning, against the
Qwen3 family's pipeline); its tokenizer writes codes one character per token as \teacher's does, so every event
definition carries over, and it runs the 107k-token prompts under the same YaRN factor. The reading was fixed before the
run: recurrence of the structure, not of the heads. Two candidates were set aside at a two-note smoke test because they do
not walk the candidate list at length: \texttt{DeepSeek-R1-Distill-Qwen-14B} names a few codes from memory and produces
almost no contrastive events on the full inventory, and \texttt{Qwen3-32B} at the paper's decoding settings produces a
handful; \texttt{QwQ-32B} reasons at length over the candidates (median \NSecondTraceGenTokensMedian{} generated tokens
against the teacher's \NSecondTraceTeacherGenTokensMedian{} on the same examples; \NSecondTraceHitCap{} of
\NSecondTraceN{} traces reached the cap) and produces about twice the teacher's decision material per note
(\NSecondAllEventsPTwoC{} contrastive events against \NSecondAllEventsPTwoCTeacher).

\paragraph{Small global sets carry the decisions.} Ranked with the same calibrated contrast ranking, \texttt{QwQ-32B}'s
own global top-40 recovers \NSecondPOneKiKForty, \NSecondPTwoKiKForty{} and \NSecondPTwoCKiKForty{} of the chapter,
within-region and contrastive decisions in another patient's context (Table~\ref{tab:app-second}), against random sets'
\NSecondPTwoCKiKFortyRand{} at most, above random on \NSecondPTwoCKiKFortyWins{} notes in every phase at $k=8$ and $k=40$,
and the sets are necessary as well. The exact contrastive refinement puts one late head first by a wide margin
(\NSecondExactTopHead, \NSecondExactTopI{} nats [\NSecondExactTopLo, \NSecondExactTopHi]; second
\NSecondExactSecondHead{} at \NSecondExactSecondI), and first-order attribution correlates with the exact effects at a
median per-event Pearson of \NSecondExactPearsonMedian. The phase sets overlap as they do in \teacher{} (chapter and
contrastive top-40 share \NSecondOverlapPOnePTwoCForty{} heads against \NTeacherOverlapPOnePTwoCForty{}; at $k=8$,
\NSecondOverlapPOnePTwoCEight{} against \NTeacherOverlapPOnePTwoCEight), and the contrast top-40 sits in the last third of the
network (layers \NSecondPTwoCTopFortyLayerMin--\NSecondPTwoCTopFortyLayerMax{} for the contrastive set, with
\NSecondPTwoCTopFortyLastTwoLayers{} of 40 in the last two layers). Where it differs: it concentrates less at $k=8$ (its
within-region top-8 recovers \NSecondPTwoKiKEight), and its contrastive decision leans on one head.

\begin{table}[!htbp]
  \centering\small
  \caption{A second teacher: knock-in recovery and necessity of \texttt{QwQ-32B}'s own global sets beside \teacher's
  (HD-001), same 100 notes, full inventory, same instrument; medians over notes of per-note means, random-$k$ in
  parentheses, notes where the global set beats random. \resid{R-C15}}
  \label{tab:app-second}
  \providecommand{\gensd}[2]{\begin{tabular}[t]{@{}c@{}}#1\\[-2pt]{\scriptsize$\pm$#2}\end{tabular}}
\providecommand{\genstack}[2]{\begin{tabular}[t]{@{}c@{}}#1\\[-2pt]{\scriptsize #2}\end{tabular}}
\providecommand{\genstackl}[2]{\begin{tabular}[t]{@{}l@{}}#1\\[-2pt]{\scriptsize #2}\end{tabular}}
\providecommand{\genhead}[2]{\begin{tabular}[b]{@{}c@{}}#1\\#2\end{tabular}}
\providecommand{\genstackiii}[3]{\begin{tabular}[t]{@{}c@{}}#1\\[-2pt]{\scriptsize #2}\\[-3pt]{\scriptsize #3}\end{tabular}}
\footnotesize\setlength{\tabcolsep}{3.5pt}
\begin{adjustbox}{max width=\linewidth}
\begin{tabular}{@{}l c cc cc c@{}}
\toprule
 & & \multicolumn{2}{c}{knock-in, $k=8$} & \multicolumn{2}{c}{knock-in, $k=40$} & necessity, $k=40$ \\
\cmidrule(lr){3-4}\cmidrule(lr){5-6}\cmidrule(lr){7-7}
Event & events (QwQ / Qwen3-14B) & QwQ (random) / Qwen3-14B & notes & QwQ (random) / Qwen3-14B & notes & QwQ / Qwen3-14B \\
\midrule
\Pone & 4{,}407 / 2{,}064 & 0.60\,{\scriptsize(0.01)} / 0.67 & 100/100 & 0.90\,{\scriptsize(0.03)} / 0.86 & 100/100 & 1.64 / 1.24 \\
\Ptwo & 4{,}987 / 4{,}587 & 0.17\,{\scriptsize(0.00)} / 0.48 & 100/100 & 0.70\,{\scriptsize(0.01)} / 0.79 & 100/100 & 1.71 / 1.61 \\
\Ptwoc & 3{,}606 / 1{,}959 & 0.51\,{\scriptsize(0.00)} / 0.69 & 100/100 & 0.72\,{\scriptsize(0.00)} / 0.85 & 100/100 & 0.81 / 0.86 \\
\bottomrule
\end{tabular}
\end{adjustbox}

\end{table}

\paragraph{The chapter becomes elicitable in the reasoning, through a different set.} The forcing probe finds a window
locus for \NSecondLocusWindowPct{} of \texttt{QwQ-32B}'s chapter mentions (\teacher: \NLocusWindowPct), a median of
\NSecondLocusDMedian{} tokens before the letter (Table~\ref{tab:app-second-locus}); fewer mentions inherit the chapter and
more are never decisive under the probe. Its prose-time set is mid-to-late (median layer \NSecondLocusLayerMedian{} of
\NSecondLayers, against \NLocusTopFortyLayerMedian{} of 40 for \teacher), shares \NSecondLocusOverlapEmissionKEight{} of its
top-8 and \NSecondLocusOverlapEmissionKForty{} of its top-40 with the emission set, and reads the note at the locus
(\NSecondReadProseNote{} of its attention against \NSecondReadRandomNote{} for random heads). The window/suffix dissociation
of the emission set reproduces: its top-16 removes \NSecondLocusAEmisTopSixteenWin{} of the forced margin from the window
positions and \NSecondLocusAEmisTopSixteenSuf{} from the suffix. The fixed global prose-time set is necessary over the
window (top-16 window-only \NSecondLocusBKSixteenWin{} against random-16's \NSecondLocusBKSixteenRandWin), but, unlike
\teacher's (window \NLocusProseWinKSixteen{} against suffix \NLocusProseSufKSixteen), it acts about as much at the suffix
(\NSecondLocusBKSixteenSuf): in \texttt{QwQ-32B} the emission set stays out of the prose-time computation while the
prose-time set also takes part in writing the chapter out. The structure recurs, the specific heads do not, and the
second reasoner is still of Qwen lineage, which is the scope we claim.

\begin{table}[!htbp]
  \centering\small
  \caption{A second teacher: the prose locus of \texttt{QwQ-32B} beside \teacher's (HD-001): outcome shares and
  distances, the prose-time set's layers and its overlap with the emission set, the exact ablations at the locus (pass
  A: the emission set and the mention's own set; pass B: the fixed global prose-time set at window / suffix / both
  positions), and what the prose-time heads read at the locus token. \resid{R-C15}}
  \label{tab:app-second-locus}
  \providecommand{\gensd}[2]{\begin{tabular}[t]{@{}c@{}}#1\\[-2pt]{\scriptsize$\pm$#2}\end{tabular}}
\providecommand{\genstack}[2]{\begin{tabular}[t]{@{}c@{}}#1\\[-2pt]{\scriptsize #2}\end{tabular}}
\providecommand{\genstackl}[2]{\begin{tabular}[t]{@{}l@{}}#1\\[-2pt]{\scriptsize #2}\end{tabular}}
\providecommand{\genhead}[2]{\begin{tabular}[b]{@{}c@{}}#1\\#2\end{tabular}}
\providecommand{\genstackiii}[3]{\begin{tabular}[t]{@{}c@{}}#1\\[-2pt]{\scriptsize #2}\\[-3pt]{\scriptsize #3}\end{tabular}}
\footnotesize\setlength{\tabcolsep}{4pt}
\begin{adjustbox}{max width=\linewidth}
\begin{tabular}{@{}l c c@{}}
\toprule
Quantity & QwQ-32B & \teacher{} (HD-001) \\
\midrule
chapter mentions & 4{,}407 & 2{,}064 \\
locus in the reasoning window & 34.8\% & 36.3\% \\
at the code token / inherited / probe never decisive & 2.4\% / 18.2\% / 44.7\% & 2.3\% / 29.5\% / 31.6\% \\
gold mentions with a window locus & 50\% (347) & 59\% (226) \\
median locus distance (tokens; p10--p90) & 10 (2--48) & 7 (1--55) \\
prose-time top-40: median layer (of layers) & 41 (64) & 21 (40) \\
prose-time top-8 / top-40 shared with the emission set & 1 / 3 & 0 / 9 \\
\midrule
\multicolumn{3}{@{}l}{\textit{Exact ablation at the locus (share of the forced margin destroyed; medians over notes)}} \\
emission top-16, window and suffix & +0.90 & +1.08 \\
emission top-16, window positions only & +0.08 & +0.16 \\
emission top-16, suffix only & +0.90 & +1.01 \\
the mention's own locus top-16, window only & +0.60 & +1.08 \\
random-16 & +0.02 & +0.09 \\
\midrule
\multicolumn{3}{@{}l}{\textit{The fixed global prose-time set (pass B): window only / suffix only / both}} \\
prose-time top-8 & +0.26 / +0.49 / +0.70 & +1.00 / +0.45 / +1.26 \\
prose-time top-16 & +0.55 / +0.67 / +1.02 & +1.32 / +0.48 / +1.48 \\
prose-time top-40 & +0.88 / +0.91 / +1.29 & +1.69 / +0.77 / +1.71 \\
\midrule
\multicolumn{3}{@{}l}{\textit{Attention at the locus token (prose-time top-16; random-16 in parentheses)}} \\
share on the note & 0.20 (0.03) & 0.25 (0.08) \\
share on the emitted chapter's lines & 0.07 (0.02) & 0.08 (0.05) \\
\bottomrule
\end{tabular}
\end{adjustbox}

\end{table}

\providecommand{\appok}{\ding{51}}
\providecommand{\appno}{\ding{55}}
\providecommand{\appflag}[1]{\texttt{-{}-#1}}

\section{\mistill{} objective details}
\label{app:objective}
\supports{Sec.~\ref{sec:distill-objective}; \resid{R-D10}, \resid{R-D9}}

This appendix states the objective as it was trained, the pre-registered term that was measured and removed, how
$\lambda$ was selected, and what the objective costs to train. Everything in
Sections~\ref{app:obj-write} and~\ref{app:obj-lambda} is the selection stage: \qwenfour{} trained on 100 notes for
two epochs and evaluated on the 100 validation notes (the characterization notes, regenerated on the shortlist,
which never enter training).

\subsection{The loss as run}
\label{app:obj-loss}
\mistill{} minimises Eq.~(\ref{eq:total}) with $\lambda_{\mathrm{P1}}=\lambda_{\mathrm{P2}}=0.1$, teacher and student
teacher-forced on the teacher's trace. $\mathcal{L}_{\mathrm{CoT}}$ is the \vanillakd{} loss unchanged
(Appendix~\ref{app:baselines}): the mean cross-entropy over the trace tokens plus, with weight 1, the KL divergence
$\mathrm{KL}(p^{T}\|p^{S})$ between the teacher's stored distribution over its top-32 tokens and a tail bin and the
student's distribution over the same bins; for \smol, whose vocabulary differs, universal logit distillation (ULD; \citealp{boizard2025towards}) over
the same number of bins replaces the KL. Each phase adds two terms. With $R_P$ the supervised rows of phase $P$ and
$E_P\subseteq R_P$ its event rows,
\begin{equation}
  \label{eq:app-phase}
  \mathcal{L}^{P}=\frac{1}{|R_P|}\sum_{q\in R_P}\mathrm{KL}\big(\bar{\alpha}^{T}_{q}\,\big\|\,\bar{\alpha}^{S}_{q}\big)
  +\frac{1}{|E_P|}\sum_{e\in E_P}\big(m^{S}_{e}c^{S}_{e}-m^{T}_{e}c^{T}_{e}\big)^{2},
\end{equation}
which is Eq.~(\ref{eq:terms}) with each term averaged over the rows where it is defined: the attention term on every
supervised row, the interaction term only where a decision, and so a margin, exists. $R_{\mathrm{P1}}$ holds every
\Pone{} event and the $W{=}8$ positions before it, and $E_{\mathrm{P1}}$ the events alone; $R_{\mathrm{P2}}=E_{\mathrm{P2}}$
holds the \Ptwo{} and \Ptwoc{} events. Each row is read through one head pool per model: the \Pone{} set at a \Pone{}
event, the \Ptwo{} set at a \Ptwo{} event, the \Ptwoc{} set at a \Ptwoc{} event, and, on the window, the teacher's
prose-time set and the student's \Pone{} set (below).

\paragraph{Pooled attention, region mass and contribution.} $\bar{\alpha}^{M}_{q}$ is Eq.~(\ref{eq:pool}),
renormalised over buckets. At an event, $m_e$ is the pooled mass on the decision's region: every shortlist line whose
code starts with the emitted chapter letter at a \Pone{} event; the code's own line, its earlier mentions ($S$) and its
emitted prefix at a \Ptwo{} or \Ptwoc{} event. The pooled heads' direct contribution to the event's margin is
\begin{equation}
  \label{eq:app-c}
  c_e=\frac{\sum_h w_h\,\big(W^{h}_{O}z_{h,e}\big)^{\top}u_e}{\max\big(|\margin_e|,\,1\big)},\qquad
  u_e=U_{t}-\sum_{j\in C}\pi_j U_{j},\quad \pi=\operatorname{softmax}\big(\mathrm{logit}_{C}\big),
\end{equation}
where $z_{h,e}$ is head $h$'s output at the event (its share of the input to the layer's output projection),
$W^{h}_{O}$ that head's slice of the projection, $U$ the unembedding, and $t$ and $C$ the event's target and
competitors in the model's own vocabulary. $u_e$ is the gradient of $\margin_e$ with respect to the final residual
stream, with the scale of the final normalisation held fixed. The denominator is the model's own margin, detached and
floored at one nat so that a near-zero margin cannot inflate the term; $c$ is a scalar per event, free of width, head
count and tokenizer. It is a direct-path proxy, not a causal estimate: measured against exact single-head effects on the
teacher it correlates at Spearman \NProxyCSpearmanPooled{} pooled over events (\NProxyCSpearmanEvent{} per event) and sees
only heads whose effect runs through their direct path (Appendix~\ref{app:instrument}); the empirical case for the term
is the interaction-only ablation of Appendix~\ref{app:ablations}.

\paragraph{Span buckets.} Every span is a character range in the untemplated text [prompt; trace], so one bucket list
serves both models; chat-template tokens and the first-position attention sink fall into a separate bucket. The
buckets are the shortlist header, one bucket per shortlist line (code and description), the note in windows of 32
words (the notes lack reliable sentence punctuation, so fixed windows stand in for sentences), the instruction, and
the trace split at sentence boundaries, with pieces capped at 400 characters. At a \Ptwo{} or \Ptwoc{} event five
buckets are carved out of these: the code's own line, its category siblings' lines, the code's earlier mentions ($S$),
the near-miss's earlier mentions ($N$) and the code's emitted prefix. Each token's attention goes to the bucket that
contains its span, so the two models' pooled rows are distributions over one axis.

\paragraph{Head pools.} Each model pools a causally ranked top-40 per phase with $w_h=\operatorname{softmax}(I_c/\tau)$
over the 40 heads, $\tau$ the standard deviation of their importances. The teacher's pools are HD-002's global sets
on the shortlist (Appendix~\ref{app:hd002}) and, on the \Pone{} window, its prose-time set, weighted the same way. The
student's pools come from the same attribution pass (the calibrated ranking of Sec.~\ref{sec:setup-measure}) run once
on the untrained student, teacher-forced on the 100 validation traces. On \qwenfour{} one late head, \head{35}{25},
tops all three phases and carries most of each pool's weight (\NMDQwenBasePoolTopWeightPOne, \NMDQwenBasePoolTopWeightPTwo{} and \NMDQwenBasePoolTopWeightPTwoC{} at \Pone, \Ptwo{} and \Ptwoc); \smol's
pools are flatter and mid-layer, led at \Pone{} by \head{22}{4}, \head{19}{0} and \head{29}{11}. The pools are fixed
for the run, so the objective is stationary; re-ranking the student's pool during training was planned as an
ablation and not run.

\paragraph{The window, and its asymmetry.} The $W{=}8$ positions before each \Pone{} event lie between the
characterization's median locus distances, \NShortLocusDMedianFull{} tokens on the full inventory and
\NShortLocusDMedianShort{} on the shortlist (Sec.~\ref{sec:char-locus}; Appendix~\ref{app:hd002}). On them the teacher is
pooled over its prose-time set and the student over its own \Pone{} set, because no prose-time set had been measured
for either student when training began: on the window, the heads of the student's \Pone{} set are asked to reproduce
the attention of the teacher's prose-time heads. The no-window arm of Appendix~\ref{app:ablations}, which supervises
the chapter events alone, shows that the window is unnecessary for most of the gain: the contrastive gain over
\vanillakd{} is fully present without it, with a small paired decrease in own-heads \Ptwoc{} recovery
(\NPairQwenNoWindowSZeroVsMistillSZeroPTwoCDelta{} per note at $k=40$, $p$ \NPairQwenNoWindowSZeroVsMistillSZeroPTwoCP) and no
detected task difference. The window is kept as part of the objective as run; it is a small part of the \Pone{} term,
not the source of the transfer. A prose-time set for the untrained
\qwenfour{} was later produced as a by-product of its M5 locus pass, which makes a symmetric-window arm possible; it was
not run.

\paragraph{Events across tokenizers.} Each event carries an anchor, the first non-blank character its decision emits.
A model's position for the event is the last token that ends at or before the anchor, and the event is kept only if
that model's next token begins at the anchor (up to leading whitespace), that is, if the model makes the decision
there. A tokenizer that merges the decided character into a longer token, as \smol's does with digit runs, has no such
position. The event is then dropped rather than moved to a neighbouring token, which would match the teacher's
attention at one character against the student's at another. A cross-tokenizer student is therefore supervised on the
subset of the teacher's rows at which its own tokenization decides, and its target and competitor ids, hence its
margin and $c$, are its own. Window rows are paired by their distance in tokens before the event: the same text for
\qwenfour, whose tokenizer is the teacher's, and an approximation for \smol. Table~\ref{tab:app-coverage} counts what
survives. Over the 493 training notes \qwenfour{} realizes every teacher event; \smol{} realizes
\NCovSmolPOneKeptPct{} of the chapter events, \NCovSmolPTwoKeptPct{} of the character events and \NCovSmolPTwoCKeptPct{}
of the contrastive events (\NCovSmolPTwoCKept{} of \NCovQwenPTwoCTeacher; per-note median \NCovSmolPTwoCPerNoteMedian; one
note keeps no contrastive event), the dropped ones being multi-token digit runs. The cross-family comparison does not
depend on the dropped events: restricted to the contrastive events both tokenizers realize, \smol's own-heads recovery
gain over \vanillakd{} is unchanged (note-clustered \NAlignedSmolMistillVsVanillaPTwoCClusterDelta{} against fp32 and
\NAlignedSmolMistillVsVanillaBfPTwoCClusterDelta{} against bf16, Appendix~\ref{app:results}), while \qwenfour's gain on the
same subset is larger than on all events (\NAlignedQwenMistillVsVanillaBfPTwoCClusterDelta{} against
\NPoolQwenMistillVsVanillaBfPTwoCClusterDelta), so the common subset does not favour the cross-family student. The
table also gives the knock-in denominator: the share of events above the 1-nat range gate of
Appendix~\ref{app:instrument}, \NGateTeacherFullPTwoC{} of the teacher's contrastive events and
\NGateSmolBasePTwoC--\NGateQwenMistillPTwoC{} of the students'.

\begin{table}[!htbp]
  \centering\small
  \caption{Event coverage per tokenizer over the 493 training notes (teacher events, and the share each student's
  tokenizer realizes as one decision), and the knock-in denominator per cell family (share of events above the 1-nat
  range gate). \resid{R-D19}}
  \label{tab:app-coverage}
  \providecommand{\gensd}[2]{\begin{tabular}[t]{@{}c@{}}#1\\[-2pt]{\scriptsize$\pm$#2}\end{tabular}}
\providecommand{\genstack}[2]{\begin{tabular}[t]{@{}c@{}}#1\\[-2pt]{\scriptsize #2}\end{tabular}}
\providecommand{\genstackl}[2]{\begin{tabular}[t]{@{}l@{}}#1\\[-2pt]{\scriptsize #2}\end{tabular}}
\providecommand{\genhead}[2]{\begin{tabular}[b]{@{}c@{}}#1\\#2\end{tabular}}
\providecommand{\genstackiii}[3]{\begin{tabular}[t]{@{}c@{}}#1\\[-2pt]{\scriptsize #2}\\[-3pt]{\scriptsize #3}\end{tabular}}
\footnotesize\setlength{\tabcolsep}{4pt}
\begin{adjustbox}{max width=\linewidth}
\begin{tabular}{@{}l c cc c@{}}
\toprule
 & teacher & \qwenfour{} kept & \smol{} kept & \\
Event & events & (share) & (share) & \\
\midrule
\multicolumn{5}{@{}l}{\textit{Teacher events over the 493 training notes, and the share each student's tokenizer realizes as one decision}} \\
\Pone & 20{,}501 & 20{,}501 (100\%) & 20{,}500 (100\%) \\
\Ptwo & 77{,}027 & 77{,}027 (100\%) & 41{,}067 (53\%) \\
\Ptwoc & 57{,}584 & 57{,}584 (100\%) & 41{,}824 (73\%) \\
\midrule
\multicolumn{5}{@{}l}{\textit{Knock-in denominator: share of events above the 1-nat range gate (all-heads minus corrupted margin), per cell family}} \\
teacher, full inventory (HD-001) & \multicolumn{2}{c}{\Pone{} 1.00} & \Ptwo{} 0.91 & \Ptwoc{} 0.71 \\
teacher, shortlist (HD-002) & \multicolumn{2}{c}{\Pone{} 0.99} & \Ptwo{} 0.92 & \Ptwoc{} 0.73 \\
\qwenfour{} base & \multicolumn{2}{c}{\Pone{} 0.95} & \Ptwo{} 0.81 & \Ptwoc{} 0.64 \\
\qwenfour{} \mistill{} s0 & \multicolumn{2}{c}{\Pone{} 0.98} & \Ptwo{} 0.85 & \Ptwoc{} 0.68 \\
\qwenfour{} \vanillakd{} (bf16) s0 & \multicolumn{2}{c}{\Pone{} 0.98} & \Ptwo{} 0.85 & \Ptwoc{} 0.64 \\
\qwenfour{} \vanillakd{} (fp32) s0 & \multicolumn{2}{c}{\Pone{} 0.99} & \Ptwo{} 0.87 & \Ptwoc{} 0.66 \\
\smol{} base & \multicolumn{2}{c}{\Pone{} 0.93} & \Ptwo{} 0.73 & \Ptwoc{} 0.56 \\
\smol{} \mistill{} s0 & \multicolumn{2}{c}{\Pone{} 0.97} & \Ptwo{} 0.81 & \Ptwoc{} 0.60 \\
\smol{} \vanillakd{} (bf16) s0 & \multicolumn{2}{c}{\Pone{} 0.97} & \Ptwo{} 0.81 & \Ptwoc{} 0.61 \\
\smol{} \vanillakd{} (fp32) s0 & \multicolumn{2}{c}{\Pone{} 0.98} & \Ptwo{} 0.85 & \Ptwoc{} 0.65 \\
\bottomrule
\end{tabular}
\end{adjustbox}

\end{table}

\subsection{The write term: pre-registered, measured, removed}
\label{app:obj-write}
The pre-registered objective had a third term per phase, $\frac{1}{|E_P|}\sum_{e}(c^{S}_{e}-c^{T}_{e})^{2}$, which
matches the pooled heads' normalised contribution directly. At pilot size it was harmful in both forms tried
(Table~\ref{tab:app-write}). As written, a squared error against a target the student does not yet track is minimised
by a constant. Trained at $\lambda=(1,1)$, the \Pone{} write term fell by \NMDWriteRawPOneTermDrop{} while the
student--teacher correlation of $c$ went from \NMDWriteRawCorrStart{} to \NMDWriteRawCorrEnd; the
pooled heads stopped carrying the chapter margin (M1 at \Pone{} fell from \NMDSelBaseMOnePOne{} to
\NMDWriteRawMOnePOneAfter), and every cell of the pre-registered $3\times3$ grid scored below the untrained
student. Standardising $c$ over each note's events before the squared error removes the constant minimum, and the term
then learns the teacher's pattern (correlation \NMDWriteZnoteCorr). The arm still scores below \vanillakd{}
trained on the same 100 notes at every $\lambda$ tried, below the untrained student on completed answers even at
$\lambda=0.1$ (\NMDWriteZnoteCompletedMicroF{} against \NMDSelBaseCompletedMicroF), and its M1 falls in
every phase. The term was removed. Both forms remain in the code (\appflag{terms}, \appflag{write-norm}), and $c$
remains inside the interaction term, where it is the pooled heads' share of the margin that M1 reads
(Appendix~\ref{app:metrics}). The removal is a post-hoc change to the pre-registered objective and is labelled as one
wherever the objective is described.

\begin{table}[!htbp]
  \centering\small
  \caption{Selection stage on the 100 validation notes (\qwenfour, 100 training notes, two epochs, seed 0;
  $\lambda_{\mathrm{P1}}=\lambda_{\mathrm{P2}}=\lambda$). The pre-registered objective falls below the untrained student
  and the standardised write term below \vanillakd{} trained on the same notes; without the write term the objective
  reaches \vanillakd{} at $\lambda=0.1$. Multi-valued cells list $\lambda=1\,/\,0.3\,/\,0.1$. \resid{R-D10}}
  \label{tab:app-write}
  \begin{adjustbox}{max width=\linewidth}
  \begin{tabular}{@{}l ccc@{}}
    \toprule
    Arm & validation \microf & valid & M1 \Ptwoc \\
    \midrule
    untrained student & \NMDSelBaseMicroF{} & \NMDSelBaseValid{} & \NMDSelBaseMOnePTwoC{} \\
    \vanillakd{} (100 notes) & \NMDSelVanillaSZeroMicroF{} & \NMDSelVanillaSZeroValid{} & \NMDSelVanillaSZeroMOnePTwoC{} \\
    pre-registered (attention + write + interaction), best of $3\times3$ & \NMDSelGridMicroFMax{} & \NMDSelPreregBestValid{} & \NMDSelPreregBestMOnePTwoC{} \\
    standardised write term, $\lambda=1\,/\,0.3\,/\,0.1$ & \NMDSelZnoteMicroFLambdaOne\,/\,\NMDSelZnoteMicroFLambdaPointThree\,/\,\NMDSelZnoteMicroFLambdaPointOne{} & \NMDSelZnoteValidLambdaOne\,/\,\NMDSelZnoteValidLambdaPointThree\,/\,\NMDSelZnoteValidLambdaPointOne{} & \NMDSelZnoteMOnePTwoCLambdaOne\,/\,\NMDSelZnoteMOnePTwoCLambdaPointThree\,/\,\NMDSelZnoteMOnePTwoCLambdaPointOne{} \\
    attention + interaction (as run), $\lambda=1\,/\,0.3\,/\,0.1$ & \NMDSelAttnInterMicroFLambdaOne\,/\,\NMDSelAttnInterMicroFLambdaPointThree\,/\,\NMDSelMistillSZeroMicroF{} & \NMDSelAttnInterValidLambdaOne\,/\,\NMDSelAttnInterValidLambdaPointThree\,/\,\NMDSelMistillSZeroValid{} & \NMDSelAttnInterMOnePTwoCLambdaOne\,/\,\NMDSelAttnInterMOnePTwoCLambdaPointThree\,/\,\NMDSelMistillSZeroMOnePTwoC{} \\
    \bottomrule
  \end{tabular}
  \end{adjustbox}
\end{table}

\subsection{Selecting $\lambda$ on the validation notes}
\label{app:obj-lambda}
The pre-registered rule was: $\lambda_{\mathrm{P1}},\lambda_{\mathrm{P2}}\in\{0.3,1,3\}$, nine runs at 100 training
notes, evaluated on the validation notes; select by validation \microf{} with ties broken by M1 at \Ptwoc; use the
chosen pair for every \mistill{} and ablation run. On the pre-registered objective the grid is flat and below the
untrained student (\microf{} from \NMDSelGridMicroFMin{} to \NMDSelGridMicroFMax, against a one-seed
noise of \NMDSelOneSeedNoiseMicroF). It cannot be ranked,
and it is reported as the write term's negative rather than as a sensitivity to $\lambda$. At $\lambda=1$ the
mechanism terms also outweighed the task loss about \NMDSelMechToCERatioLambdaOne-fold (about
\NMDSelMechToCERatioLambdaPointThree-fold at 0.3), and the arms over-generated. Every generation that reached the
16{,}384-token cap on the validation notes was on one of the notes from which the discharge diagnosis had been
removed, which are most of them; at $\lambda=1$, \NMDSelCapRateLambdaOne{} of those notes' generations reached
the cap, against \NMDSelCapRateVanilla{} for \vanillakd{} and \NMDSelCapRateBase{} for the
untrained student. The
range was therefore extended below its pre-registered floor to $0.1$, disclosed as an extension.

On the objective without the write term only $\lambda=0.1$ reaches \vanillakd{} trained on the same notes
(Table~\ref{tab:app-write}), and the objective's effect on the decisions scales with $\lambda$
(Table~\ref{tab:app-dose}): the contrastive decision moves toward the teacher in proportion to the weight, the chapter
and character decisions move slightly away, and the task holds only at the smallest weight. The selected pair,
$\lambda_{\mathrm{P1}}=\lambda_{\mathrm{P2}}=0.1$, was confirmed against \vanillakd{} on three seeds each
(Table~\ref{tab:app-select}). On the task the two are not distinguished with the note as the unit (mean per-note difference
\NSelPairClusterTaskMean{} [\NSelPairClusterTaskLo, \NSelPairClusterTaskHi], higher on \NSelPairClusterTaskHigher{} of \NSelPairClusterTaskN{} validation
notes, sign test $p$ \NSelPairClusterTaskP; the pooled note-pair test that suggested an edge counted the 300 seed-pair observations as
independent and is reported as descriptive only). On the mechanism the two differ clearly: \mistill's supervised
heads read the emitted code's region more, and it carries the contrastive decision about a third of the way from the
untrained student to the teacher (M1 \Ptwoc{} \NSelPairClusterMOnePTwoCMean{} per note [\NSelPairClusterMOnePTwoCLo, \NSelPairClusterMOnePTwoCHi], higher on
\NSelPairClusterMOnePTwoCHigher{} of \NSelPairClusterMOnePTwoCN{} notes), where \vanillakd{} leaves it at the untrained student's level on every seed, at a
small, significant cost at the chapter decision (\Pone{} \NSelPairClusterMOnePOneMean{} [\NSelPairClusterMOnePOneLo, \NSelPairClusterMOnePOneHi], lower on
\NSelPairClusterMOnePOneLower{} of \NSelPairClusterMOnePOneN), the phase that is nearly saturated already; the character decision does not
move detectably (\Ptwo{} \NSelPairClusterMOnePTwoMean{} [\NSelPairClusterMOnePTwoLo, \NSelPairClusterMOnePTwoHi], $p$ \NSelPairClusterMOnePTwoP)
(Appendix~\ref{app:metrics}). The second axis was exercised once, at tier-1 size, as $\lambda_{\mathrm{P1}}=0.03$ with
$\lambda_{\mathrm{P2}}=0.1$ (Appendix~\ref{app:results}). No tuning budget was spent on the baselines' own
hyper-parameters (Appendix~\ref{app:baselines}).

\begin{table}[!htbp]
  \centering\small
  \caption{Dose-response of the objective without the write term on the validation notes: M1 differences from
  \vanillakd{} trained on the same 100 notes (seed 0), paired per note over the notes with events in every phase
  (mean per-note difference; notes higher\,/\,lower; sign-test $p$). \resid{R-D10}}
  \label{tab:app-dose}
  \begin{adjustbox}{max width=\linewidth}
  \begin{tabular}{@{}c ccc@{}}
    \toprule
    $\lambda$ & $\Delta$ M1 \Ptwoc & $\Delta$ M1 \Pone & $\Delta$ M1 \Ptwo \\
    \midrule
    0.1 & \NMDDoseLambdaPointOnePTwoCDelta{} (\NMDDoseLambdaPointOnePTwoCHigher/\NMDDoseLambdaPointOnePTwoCLower, $p$ \NMDDoseLambdaPointOnePTwoCP) & \NMDDoseLambdaPointOnePOneDelta{} (\NMDDoseLambdaPointOnePOneHigher/\NMDDoseLambdaPointOnePOneLower, $p$ \NMDDoseLambdaPointOnePOneP) & \NMDDoseLambdaPointOnePTwoDelta{} ($p$ \NMDDoseLambdaPointOnePTwoP) \\
    0.3 & \NMDDoseLambdaPointThreePTwoCDelta{} (\NMDDoseLambdaPointThreePTwoCHigher/\NMDDoseLambdaPointThreePTwoCLower, $p$ \NMDDoseLambdaPointThreePTwoCP) & \NMDDoseLambdaPointThreePOneDelta{} (\NMDDoseLambdaPointThreePOneHigher/\NMDDoseLambdaPointThreePOneLower, $p$ \NMDDoseLambdaPointThreePOneP) & \NMDDoseLambdaPointThreePTwoDelta{} (\NMDDoseLambdaPointThreePTwoHigher/\NMDDoseLambdaPointThreePTwoLower, $p$ \NMDDoseLambdaPointThreePTwoP) \\
    1 & \NMDDoseLambdaOnePTwoCDelta{} (\NMDDoseLambdaOnePTwoCHigher/\NMDDoseLambdaOnePTwoCLower, $p$ \NMDDoseLambdaOnePTwoCP) & \NMDDoseLambdaOnePOneDelta{} (\NMDDoseLambdaOnePOneHigher/\NMDDoseLambdaOnePOneLower, $p$ \NMDDoseLambdaOnePOneP) & \NMDDoseLambdaOnePTwoDelta{} (\NMDDoseLambdaOnePTwoHigher/\NMDDoseLambdaOnePTwoLower, $p$ \NMDDoseLambdaOnePTwoP) \\
    \bottomrule
  \end{tabular}
  \end{adjustbox}
\end{table}

\begin{table}[!htbp]
  \centering\small
  \caption{The selected objective ($\lambda=0.1$) against \vanillakd{} trained on the same 100 notes, three seeds each,
  on the validation notes. Anchors in the first column: the untrained student and the teacher on its own validation
  traces. Paired (pilot, validation-selection context): seed-matched, the three seed-pair differences averaged within each note so the note is the
  unit --- mean per-note difference [percentile bootstrap 95\% CI over notes], notes higher\,/\,lower\,/\,tied (100 task notes; 99 with M1 events), exact sign-test $p$ over notes
  (the earlier pooled note-pair counts, which treated the 300 seed-pair observations as independent, are kept in Appendix~\ref{app:res-paired}'s
  sources as descriptive only). M2: region lift of the
  supervised heads, top-16 median, at \Pone{}\,/\,\Ptwo{}\,/\,\Ptwoc; its paired entry gives ranges over the nine seed pairings (notes higher of 100, median difference, largest $p$). \resid{R-D10}}
  \label{tab:app-select}
  \begin{adjustbox}{max width=\linewidth}
  \begin{tabular}{@{}l ccc@{}}
    \toprule
    Readout (anchors) & \vanillakd{}, seeds 0\,/\,1\,/\,2 & \mistill, seeds 0\,/\,1\,/\,2 & paired \\
    \midrule
    \microf & \NMDSelVanillaSZeroMicroF\,/\,\NMDSelVanillaSOneMicroF\,/\,\NMDSelVanillaSTwoMicroF{} $=$ \pmsd{\NMDSelVanillaMicroF}{\NMDSelVanillaMicroFSd} & \NMDSelMistillSZeroMicroF\,/\,\NMDSelMistillSOneMicroF\,/\,\NMDSelMistillSTwoMicroF{} $=$ \pmsd{\NMDSelMistillMicroF}{\NMDSelMistillMicroFSd} & \NSelPairClusterTaskMean{} [\NSelPairClusterTaskLo, \NSelPairClusterTaskHi]; \NSelPairClusterTaskHigher/\NSelPairClusterTaskLower/\NSelPairClusterTaskTies, $p$ \NSelPairClusterTaskP{} \\
    valid & \NMDSelVanillaSZeroValid\,/\,\NMDSelVanillaSOneValid\,/\,\NMDSelVanillaSTwoValid{} & \NMDSelMistillSZeroValid\,/\,\NMDSelMistillSOneValid\,/\,\NMDSelMistillSTwoValid{} & --- \\
    M1 \Ptwoc{} (untrained \NMDSelBaseMOnePTwoC, teacher \NMDSelTeacherMOnePTwoC) & \NMDSelVanillaSZeroMOnePTwoC\,/\,\NMDSelVanillaSOneMOnePTwoC\,/\,\NMDSelVanillaSTwoMOnePTwoC{} & \NMDSelMistillSZeroMOnePTwoC\,/\,\NMDSelMistillSOneMOnePTwoC\,/\,\NMDSelMistillSTwoMOnePTwoC{} (\NMDSelMistillMOnePTwoCGapPct{} of the gap) & \NSelPairClusterMOnePTwoCMean{} [\NSelPairClusterMOnePTwoCLo, \NSelPairClusterMOnePTwoCHi]; \NSelPairClusterMOnePTwoCHigher/\NSelPairClusterMOnePTwoCLower/\NSelPairClusterMOnePTwoCTies, $p$ \NSelPairClusterMOnePTwoCP{} \\
    M1 \Pone{} (untrained \NMDSelBaseMOnePOne, teacher \NMDSelTeacherMOnePOne) & \NMDSelVanillaSZeroMOnePOne\,/\,\NMDSelVanillaSOneMOnePOne\,/\,\NMDSelVanillaSTwoMOnePOne{} (\NMDSelVanillaMOnePOneGapPct{} of the gap) & \NMDSelMistillSZeroMOnePOne\,/\,\NMDSelMistillSOneMOnePOne\,/\,\NMDSelMistillSTwoMOnePOne{} (\NMDSelMistillMOnePOneGapPct) & \NSelPairClusterMOnePOneMean{} [\NSelPairClusterMOnePOneLo, \NSelPairClusterMOnePOneHi]; \NSelPairClusterMOnePOneHigher/\NSelPairClusterMOnePOneLower/\NSelPairClusterMOnePOneTies, $p$ \NSelPairClusterMOnePOneP{} \\
    M1 \Ptwo{} (untrained \NMDSelBaseMOnePTwo, teacher \NMDSelTeacherMOnePTwo) & \NMDSelVanillaSZeroMOnePTwo\,/\,\NMDSelVanillaSOneMOnePTwo\,/\,\NMDSelVanillaSTwoMOnePTwo{} & \NMDSelMistillSZeroMOnePTwo\,/\,\NMDSelMistillSOneMOnePTwo\,/\,\NMDSelMistillSTwoMOnePTwo{} & \NSelPairClusterMOnePTwoMean{} [\NSelPairClusterMOnePTwoLo, \NSelPairClusterMOnePTwoHi]; \NSelPairClusterMOnePTwoHigher/\NSelPairClusterMOnePTwoLower/\NSelPairClusterMOnePTwoTies, $p$ \NSelPairClusterMOnePTwoP{} \\
    M2 lift, supervised heads & \NMDSelVanillaMTwoLiftPOne\,/\,\NMDSelVanillaMTwoLiftPTwo\,/\,\NMDSelVanillaMTwoLiftPTwoC{} & \NMDSelMistillMTwoLiftPOne\,/\,\NMDSelMistillMTwoLiftPTwo\,/\,\NMDSelMistillMTwoLiftPTwoC{} & \NMDSelPairMTwoNotesRange{} of 100 higher, \NMDSelPairMTwoDeltaRange, $p$ \NMDSelPairMTwoP{} \\
    \bottomrule
  \end{tabular}
  \end{adjustbox}
\end{table}

\subsection{Re-selecting $\lambda$ with fp32 master weights}
\label{app:obj-fp32}
The submission transferred the bf16-selected $\lambda=0.1$ to the fp32 \mistill{} arms and found a task cost against fp32
\vanillakd{} (Appendix~\ref{app:res-precision}). Whether that cost was the objective's or the dose's is answered by
re-running the pre-registered selection with fp32 master weights: the same rule, one seed at tier-1 size (493 notes, two
epochs), $\lambda_{\mathrm{P1}}=\lambda_{\mathrm{P2}}\in\{0.03,0.1,0.3,1\}$, the highest validation \microf{} with ties
within the measured 0.004 band broken by M1 at \Ptwoc. The sweep is monotone (Table~\ref{tab:app-fpsweep}): under fp32
master weights more mechanism weight costs both the task and the supervised contrastive readout at every step, the
opposite of the flat bf16 dose response of Table~\ref{tab:app-dose}, and the winner, 0.03, sits on the edge of the
range, so the range was extended downward by one grid step to 0.01, disclosed as an extension; the rule then selects
$\lambda=\NSweepSelected$ (a tie with 0.03 on validation \microf, \NSweepLPointZeroOneValMicroF{} against
\NSweepLPointZeroThreeValMicroF, broken by M1 at \Ptwoc), and the second-extension condition, 0.01 beating 0.03 by more than
the band, is not met. On the 150 test notes the selected arm's seed 0 scores \NSweepLPointZeroOneTestMicroF{}
(0.03: \NSweepLPointZeroThreeTestMicroF; the submitted $\lambda=0.1$: \NSweepLPointOneTestMicroF); against the three
\vanillakd{} seeds separately its per-note differences are \NSweepLPointZeroOneTaskVsVanillaSZeroNoteDelta,
\NSweepLPointZeroOneTaskVsVanillaSOneNoteDelta{} and \NSweepLPointZeroOneTaskVsVanillaSTwoNoteDelta, every interval through
zero, where the transferred $\lambda$ cost \NQwenMistillFpMicroFVsVanillaDelta{} (this paragraph is the one-seed selection record;
the three-seed task comparison is the next paragraph's). On its own heads the selected arm recovers \NOwnQwenMistillFpSelSZeroPTwoCKiForty{} of
the contrastive decision at $k=40$, more than fp32 \vanillakd's three seeds (note-clustered
\NSweepLPointZeroOnePTwoCVsVanillaClusterDelta, higher on \NSweepLPointZeroOnePTwoCVsVanillaClusterHigher{} of 150 notes,
$p$ \NSweepLPointZeroOnePTwoCVsVanillaClusterP) and than bf16 \mistill{} seed 0
(\NSweepLPointZeroOnePTwoCVsMistillSZeroDelta), where the $\lambda=0.1$ arms were at \vanillakd's level
(\NPoolQwenMistillFpVsVanillaPTwoCClusterDelta{} against fp32 \vanillakd, $p$ \NPoolQwenMistillFpVsVanillaPTwoCClusterP;
\NPoolQwenMistillFpVsVanillaBfPTwoCClusterDelta{} against bf16, $p$ \NPoolQwenMistillFpVsVanillaBfPTwoCClusterP;
Appendix~\ref{app:res-precision}); larger weights trade the contrastive heads for the chapter heads
(\Ptwoc{} recovery falls to \NOwnQwenMistillFpLPointThreeSZeroPTwoCKiForty{} at 0.3 while \Pone{} recovery rises) and cost
task score. The one cell that stays low is $k=8$ (\NOwnQwenMistillFpSelSZeroPTwoCKiEight), for a diagnosed reason: the
first-order ranking of this arm places one causally active head, \head{35}{26} (rank
\NSweepMistillSZeroRankLThirtyFiveHTwentySix{} in every bf16 arm), at rank
\NSweepLPointZeroOneRankLThirtyFiveHTwentySix. Moved into rank 8 of the arm's own ranking with nothing else changed, the
top-8 recover \NSweepLPointZeroOneProbeKiEight{}, above fp32 \vanillakd's (note-clustered
\NSweepLPointZeroOneProbeKiEightVsVanillaClusterDelta, \NSweepLPointZeroOneProbeKiEightVsVanillaClusterHigher{} of 150
notes); the probe set is post hoc and is reported as a diagnosis, the own-ranking cell as the result. The submitted ``cost
in fp32'' was therefore largely a $\lambda$-transfer effect: re-selection improves the measured operating point, although the
three-seed interval below does not establish the absence of a remaining task cost. The paper's main table reports the fp32
arm at the $\lambda$ selected for fp32 beside the three seeds at the transferred one, both labelled.

\paragraph{Three seeds at the selected setting.} The selection above is a one-seed procedure; two further seeds at
$\lambda=0.01$ read its result on average across three seed-matched comparisons with fp32 \vanillakd{} (R-D30). Own-heads
normalized \Ptwoc{} recovery at $k=40$ is higher by a mean $\Delta$ of \NRepLPointZeroOnePTwoCClusterMean{}
[\NRepLPointZeroOnePTwoCClusterLo, \NRepLPointZeroOnePTwoCClusterHi] (note-clustered over 150 test notes, higher on
\NRepLPointZeroOnePTwoCClusterHigher), the seed-pair effects being \NRepLPointZeroOnePTwoCSZeroMean{}
[\NRepLPointZeroOnePTwoCSZeroLo, \NRepLPointZeroOnePTwoCSZeroHi], \NRepLPointZeroOnePTwoCSOneMean{} [\NRepLPointZeroOnePTwoCSOneLo,
\NRepLPointZeroOnePTwoCSOneHi] and \NRepLPointZeroOnePTwoCSTwoMean{} [\NRepLPointZeroOnePTwoCSTwoLo, \NRepLPointZeroOnePTwoCSTwoHi];
the third pair's interval includes zero. The per-seed medians, \NOwnQwenMistillFpSelSZeroPTwoCKiForty{} /
\NOwnQwenMistillFpSelSOnePTwoCKiForty{} / \NOwnQwenMistillFpSelSTwoPTwoCKiForty{} against fp32 \vanillakd's
\NOwnQwenVanillaSZeroPTwoCKiForty{} / \NOwnQwenVanillaSOnePTwoCKiForty{} / \NOwnQwenVanillaSTwoPTwoCKiForty, are
Table~\ref{tab:app-mech-seeds}'s cells and are stated as levels; the paired mean is not their difference. The corpus \microf{}
difference over the three seed pairs is \NCorpusQwenMistillFpSelVsVanillaDelta{} [\NCorpusQwenMistillFpSelVsVanillaLo,
\NCorpusQwenMistillFpSelVsVanillaHi] (three-seed means \NQwenMistillFpSelMicroF{} against \NQwenVanillaMicroF), so no task
difference is detected and none is claimed as preserved. The gain is specific to the tested $k=40$ setting: at $k=8$, with each
arm's original own ranking, the pooled comparison reverses (\NRepLPointZeroOnePTwoCKiEightClusterMean{}
[\NRepLPointZeroOnePTwoCKiEightClusterLo, \NRepLPointZeroOnePTwoCKiEightClusterHi]; pairs \NRepLPointZeroOnePTwoCKiEightSZeroMean,
\NRepLPointZeroOnePTwoCKiEightSOneMean{} and \NRepLPointZeroOnePTwoCKiEightSTwoMean), and the association with whether \head{35}{26}
enters the own top-8 remains a diagnostic rather than an explanation. The M2 shortlist-region lift of the own top-16 heads, a
dimensionless quantity (attention to the emitted code's chapter region within the shortlist, normalized by its token share), is
lower under \mistill{} (\NRepLPointZeroOneMTwoPTwoCClusterMean{} [\NRepLPointZeroOneMTwoPTwoCClusterLo,
\NRepLPointZeroOneMTwoPTwoCClusterHi]; negative point estimates in all three pairs, seed 1's interval including zero). The same
direction appears on the students' own completed traces (Appendix~\ref{app:res-owntrace}) and in the separate prefix-edit
experiment (\NRepLPointZeroOneEditPooledMean{} [\NRepLPointZeroOneEditPooledLo, \NRepLPointZeroOneEditPooledHi] over
\NRepLPointZeroOneEditPooledNotes{} notes, the edited object being the teacher-event prefix), neither of which is an evidence-use
result (Appendix~\ref{app:evflip}). \resid{R-D30}

\begin{table}[!htbp]
  \centering\small
  \caption{Re-selecting $\lambda$ with fp32 master weights (\qwenfour, one seed per cell; 0.01 is the disclosed
  extension, 0.1 the submission's three-seed arm, here its seed 0): the selection statistics on the 100 validation notes,
  the 150 test notes, and the arm's own heads. Paired lines are note-clustered against the three fp32 \vanillakd{} seeds.
  The three-seed replication at 0.01 is R-D30 (Tables~\ref{tab:main} and~\ref{tab:app-mech-seeds}).
  \resid{R-D22}}
  \label{tab:app-fpsweep}
  \providecommand{\gensd}[2]{\begin{tabular}[t]{@{}c@{}}#1\\[-2pt]{\scriptsize$\pm$#2}\end{tabular}}
\providecommand{\genstack}[2]{\begin{tabular}[t]{@{}c@{}}#1\\[-2pt]{\scriptsize #2}\end{tabular}}
\providecommand{\genstackl}[2]{\begin{tabular}[t]{@{}l@{}}#1\\[-2pt]{\scriptsize #2}\end{tabular}}
\providecommand{\genhead}[2]{\begin{tabular}[b]{@{}c@{}}#1\\#2\end{tabular}}
\providecommand{\genstackiii}[3]{\begin{tabular}[t]{@{}c@{}}#1\\[-2pt]{\scriptsize #2}\\[-3pt]{\scriptsize #3}\end{tabular}}
\footnotesize\setlength{\tabcolsep}{4pt}
\begin{adjustbox}{max width=\linewidth}
\begin{tabular}{@{}l ccccc@{}}
\toprule
$\lambda_{\mathrm{P1}}=\lambda_{\mathrm{P2}}$ & 0.01 (ext.) & 0.03 & 0.1 (submitted) & 0.3 & 1 \\
\midrule
\multicolumn{6}{@{}l}{\textit{Selection on the 100 validation notes (one seed)}} \\
\microf & 0.183 & 0.181 & 0.146 & 0.131 & 0.061 \\
M1 \Ptwoc & 0.801 & 0.795 & 0.779 & 0.762 & 0.728 \\
valid / cap & 0.85 / 0.15 & 0.85 / 0.15 & 0.85 / 0.15 & 0.82 / 0.18 & 0.66 / 0.34 \\
\midrule
\multicolumn{6}{@{}l}{\textit{150 test notes (not used for selection)}} \\
\microf & 0.177 & 0.203 & 0.161 & 0.144 & 0.073 \\
$\Delta$ \microf{} vs fp32 \vanillakd{} [95\% CI] & \pci{\ensuremath{-}0.013}{\ensuremath{-}0.028}{+0.002} & \pci{+0.016}{\ensuremath{-}0.002}{+0.035} & \pci{\ensuremath{-}0.024}{\ensuremath{-}0.039}{\ensuremath{-}0.009} & \pci{\ensuremath{-}0.046}{\ensuremath{-}0.065}{\ensuremath{-}0.027} & \pci{\ensuremath{-}0.107}{\ensuremath{-}0.125}{\ensuremath{-}0.088} \\
M1 \Ptwoc & 0.800 & 0.797 & 0.778 & 0.756 & 0.721 \\
valid / cap & 0.82 / 0.18 & 0.83 / 0.17 & 0.89 / 0.11 & 0.80 / 0.20 & 0.65 / 0.35 \\
\midrule
\multicolumn{6}{@{}l}{\textit{Own heads (150 test notes; median over notes, random floor in parentheses)}} \\
\Ptwoc{} knock-in, $k=40$ & 0.892\,{\scriptsize(0.032)} & 0.888\,{\scriptsize(0.029)} & 0.870\,{\scriptsize(0.022)} & 0.643\,{\scriptsize(0.027)} & 0.632\,{\scriptsize(0.020)} \\
\quad vs fp32 \vanillakd{} s0--2, note-clustered & \genstackiii{+0.032}{113/37}{$p$ \ensuremath{4{\times}10^{-10}}} & \genstackiii{+0.028}{104/46}{$p$ \ensuremath{2{\times}10^{-6}}} & \genstackiii{+0.001}{77/73}{$p$ 0.81} & \genstackiii{\ensuremath{-}0.214}{7/143}{$p$ \ensuremath{4{\times}10^{-34}}} & \genstackiii{\ensuremath{-}0.219}{8/141}{$p$ \ensuremath{10^{-32}}} \\
\Ptwoc{} knock-in, $k=8$ & 0.133\,{\scriptsize(0.009)} & 0.528\,{\scriptsize(0.010)} & 0.186\,{\scriptsize(0.011)} & 0.187\,{\scriptsize(0.010)} & 0.155\,{\scriptsize(0.009)} \\
\quad vs fp32 \vanillakd{} s0--2, note-clustered & \genstackiii{\ensuremath{-}0.398}{0/150}{$p$ \ensuremath{10^{-45}}} & \genstackiii{+0.003}{81/69}{$p$ 0.37} & \genstackiii{\ensuremath{-}0.350}{0/150}{$p$ \ensuremath{10^{-45}}} & \genstackiii{\ensuremath{-}0.337}{0/150}{$p$ \ensuremath{10^{-45}}} & \genstackiii{\ensuremath{-}0.370}{0/149}{$p$ \ensuremath{3{\times}10^{-45}}} \\
\quad rank of \head{35}{26} in the own ranking & 12 & 7 & 11 & 1{,}128 & 1{,}131 \\
\Pone{} knock-in, $k=40$ / $k=8$ & 0.949 / 0.725 & 0.951 / 0.728 & 0.949 / 0.730 & 0.958 / 0.779 & 0.969 / 0.809 \\
M2 lift, \Ptwoc{} / \Pone & 3.48\,{\scriptsize(1.41)} / 2.40\,{\scriptsize(0.71)} & 2.70\,{\scriptsize(1.44)} / 2.28\,{\scriptsize(0.78)} & 3.00\,{\scriptsize(1.76)} / 2.63\,{\scriptsize(1.12)} & 3.12\,{\scriptsize(1.60)} / 3.01\,{\scriptsize(1.13)} & 3.61\,{\scriptsize(1.37)} / 3.73\,{\scriptsize(1.06)} \\
\bottomrule
\end{tabular}
\end{adjustbox}

\end{table}

\subsection{Implementation and training cost}
\label{app:obj-cost}
The teacher's side is computed once: for each of the 493 training notes with a valid teacher trace, the pooled bucket
distribution at every supervised row and $m^{T}$, $c^{T}$ at every event. Per note the objective supervises a median of
\NMDRowsPerNoteMedian{} rows (at most \NMDRowsPerNoteMax) at a median of
\NMDEventsPerNoteMedian{} events, 40 heads per row, where \featkd{} matches every head of four layer pairs at
64 sampled positions and hidden states at every position. On the student side, attention is computed only at the
supervised positions: the pooled layers' attention rows are recomputed from their checkpointed input hidden states
with the model's own projections, rotary embeddings and query/key norms, inside checkpointed chunks, and the heads'
outputs are rebuilt from the same rows. \mistill{} therefore needs no online teacher and trains on one GPU. Unit tests
check that every term is zero when the student's pooled quantities equal the teacher's, that the attention term with a
uniform pool over one layer's heads equals \featkd's bucket-pooled KL, and that the stored teacher quantities are
reproduced by an independent route (full attention outputs and per-head output-projection effects) on a small model;
a smoke run checks that each ablation flag changes only what it names.

Cost (\resid{R-D9}). On the corpus's longest note the \qwenfour{} step peaks at \NMDMistillFitPeakGiB\,GiB with
bf16 weights, flat in note length, at about \NMDMistillFitSecPerNote\,s per note; the tier-1 runs logged peaks of
\NMDMistillTierOnePeakGiB\,GiB. At 100 training notes \vanillakd{} trains in \NMDWallVanillaHundredNotesMin\,min
and \mistill{} in \NMDWallMistillHundredNotesMin\,min. A 493-note \mistill{} run takes \NMDWallMistillTierOne{}
on an H200 (\NMDMistillMinPerStep\,min per eight-note step, and the same on an RTX PRO 6000: the step is largely
CPU-bound), against \NMDWallVanillaTierOneHours\,h for a 493-note \vanillakd{} run with fp32 master weights on
different hardware, so the comparison is indicative only. With fp32 master weights \mistill{} needs about
\NMDMistillFpGiBOneGPU\,GiB on one GPU: seed 0 was split over two 80\,GB GPUs, seeds 1 and 2 ran on one 180\,GB GPU each.

\section{Baselines and training details}
\label{app:baselines}
\supports{Sec.~\ref{sec:distill-setup}; \resid{R-D0}, \resid{R-D2}--\resid{R-D4}}

The comparison uses one strong representative per distillation family, run from its official code where that code
exists and at its published settings. Two rules fixed the set before any run: a third-party method enters only if its
official code can be used directly, without re-implementing someone else's system, and the set covers the families
generically, so that it cannot be read as a few hand-picked weak baselines.

\subsection{Models and software}
\label{app:base-models}
The teacher is \teacher{} (\texttt{Qwen/Qwen3-14B}) in thinking mode, in bf16 with static YaRN $\times4$; its traces
(one per note: temperature 0.6, top-$p$ 0.95, top-$k$ 20, seed 0, a 16{,}384-token cap) are the targets of every arm.
\qwenfour{} (\texttt{Qwen/Qwen3-4B}; 36 layers of 32 heads, 8 key-value heads) shares the teacher's tokenizer and is
reasoning-trained; it runs in thinking mode under static YaRN $\times2$ (an 81{,}920-token window) in training,
generation and every teacher-forced metric, the teacher's positional scaling with one factor for the whole run. \smol{}
\citep{bakouch2025smollm3} (\texttt{HuggingFaceTB/SmolLM3-3B}; 36 layers of 16 heads, 4 key-value heads; a 65{,}536-token native window, no YaRN)
comes from another family and runs in its native thinking mode; its tokenizer writes digit runs as single tokens, so
some of its \Ptwo{} decisions span several characters (Appendix~\ref{app:objective}). The \las{} simulator is
\simulator{} without thinking (Appendix~\ref{app:metrics}); the retriever encoder is gte-large
(Appendix~\ref{app:data}). Checkpoints are the public Hugging Face releases, loaded offline from a local cache so that
a missing artifact fails instead of resolving to another revision; their revisions, the Hugging Face commit hashes in
the download records of the local copies, are
\begin{center}\scriptsize
\begin{tabular}{@{}lll@{}}
\texttt{Qwen/Qwen3-14B} & \texttt{40c069824f4251a91eefaf281ebe4c544efd3e18} & teacher \\
\texttt{Qwen/Qwen3-4B} & \texttt{1cfa9a7208912126459214e8b04321603b3df60c} & student \\
\texttt{HuggingFaceTB/SmolLM3-3B} & \texttt{a07cc9a04f16550a088caea529712d1d335b0ac1} & student \\
\texttt{Qwen/Qwen3-1.7B} & \texttt{70d244cc86ccca08cf5af4e1e306ecf908b1ad5e} & \las{} simulator \\
\texttt{thenlper/gte-large} & \texttt{4bef63f39fcc5e2d6b0aae83089f307af4970164} & retriever encoder \\
\end{tabular}
\end{center}
\noindent Software: PyTorch 2.11, Transformers 5.16, vLLM 0.29 for generation and TRL 1.13.0
for the on-policy arms.

\subsection{The shared training recipe}
\label{app:base-recipe}
The off-policy arms (\vanillakd, \featkd, \distillm, and \mistill{} with its controls and ablations) share one trainer
and one recipe: the 493 training notes with a valid teacher trace (the seven notes whose teacher trace is invalid or
degenerate are excluded from every arm), two epochs (124 optimiser steps), one trace per micro-step with gradient
accumulation to a global batch of 8, AdamW ($\beta_1=0.9$, $\beta_2=0.95$, no weight decay) at learning rate
$10^{-5}$ with a 5\% linear warm-up and cosine decay, gradient-norm clipping at 1, gradient checkpointing, and the loss
on the trace tokens only (the vocabulary logits are computed chunk by chunk and never held in full). The epoch order
is a function of the seed and the epoch, and a resumed run restores the optimiser, scheduler and random-number states,
so it continues the same data order and random draws. The teacher's per-position top-32 distributions are precomputed once. Each run writes its full
configuration (every argument, the example ids and the PyTorch version) to \texttt{config.json}.

\subsection{The arms}
\label{app:base-arms}
\paragraph{\vanillakd{}} (sequence-level; \citealp{kim2016sequence, hinton_distilling_2015}): $\mathcal{L}_{\mathrm{CoT}}$
alone, the cross-entropy on the trace tokens plus the top-32 KL with weight 1 (Appendix~\ref{app:objective}). On \smol{}
every arm replaces the KL by ULD \citep{boizard2025towards}, which compares the teacher's and the student's sorted
probabilities over the same number of bins and needs no shared vocabulary, so that the \smol{} arms differ only in
their own terms.

\paragraph{\distillm{}} \citep{ko2025distillm}: its contrastive loss, vendored verbatim from the official repository
(commit \texttt{fe4cf9b}, Apache-2.0) with a unit test, at its published settings, with the teacher's trace as the
preferred response and the untrained student's own generation for the same prompt as the dispreferred one; the teacher
runs online. The loss is defined only for a shared vocabulary and the official code has no cross-tokenizer path, so the
arm runs on \qwenfour{} only; \gold{} is the cross-family logit-distillation row.

\paragraph{\opd{}} (on-policy distillation in the GKD lineage; \citealp{agarwal2024onpolicy}): TRL's
\texttt{DistillationTrainer} \citep{vonwerra2020trl}. The student samples its own completions, capped at 12{,}288
tokens (no rollout reached the cap at any logged step), and matches the teacher's next-token distributions on them
under the generalised Jensen--Shannon divergence at $\beta=1$ (the reverse KL, TRL's default), for two epochs at
learning rate $10^{-5}$, with the student loaded in bf16 and no fp32 master copy. Trainer and vLLM rollout engine share
one 141\,GB H200, because the cluster's idle-GPU policy rules out a separate rollout server.

\paragraph{\gold{}} \citep{patino2025unlocking}, the cross-tokenizer on-policy arm for \smol: TRL's experimental
\texttt{GOLDTrainer} with the ULD loss and its text-span token alignment. One override computes the logits only over the
completion window, because full-sequence logits of 30k-token prompts do not fit; the loss, the alignment and the
on-policy machinery are upstream's. Like \opd, it trains bf16 weights without an fp32 master copy. The arm ran under deviations recorded before its final run: one epoch (a compute cut); rollouts capped at
6{,}144 tokens (16{,}384 and then 8{,}192 ran out of memory in the hybrid ULD loss on a 141\,GB H200); a per-device batch
of one; an on-policy fraction of $0.5$, because a fully on-policy first run collapsed to degenerate output with no
valid answer and was discarded; and the teacher's completion, which the off-policy half trains on, truncated to the
same cap to fit in memory, which cut the end of \NMDGoldTruncatedTraces{} of the 500 training traces. The resulting null, and why it is reported, are in
Appendix~\ref{app:res-dissoc}.

\paragraph{\featkd{}} (the MiniLM\,/\,TinyBERT lineage; \citealp{jiao2020tinybert, wang2020minilm,
zagoruyko2017paying, romero2015fitnets}): a KL divergence between the teacher's and the student's head-averaged,
bucket-pooled attention (the span buckets of Appendix~\ref{app:objective}) in four uniformly mapped layer pairs, at 64
trace positions sampled per note, plus a mean squared error between hidden states at those layers (the student's
through a linear adapter to the teacher's width) at every trace position, with weights 1 and 1, no phases and no
events; the teacher runs online. The hidden states are RMS-normalised per position before the error, because a raw
error against the teacher's massive activations swamped the loss (that run was discarded). It is the generic attention
and representation control, at matched data and schedule.

\paragraph{The adapter-seed caveat.} When the four \featkd{} runs were trained, the trainer seeded its random-number
generators after constructing the adapters, so the seed did not control their initialisation. The order was fixed
afterwards, and two runs at one seed then agreed to every printed digit; every \mistill{} run was trained after the fix.
The \featkd{} runs are therefore not bit-reproducible, and their seed spread (\pmsd{\NQwenFeatKDMicroF}{\NQwenFeatKDMicroFSd}
on \qwenfour) is an upper bound on the true seed variance. The effect the row is used for, \featkd{} below \vanillakd{}
by \NQwenFeatKDMicroFVsVanillaDelta{} [\NQwenFeatKDMicroFVsVanillaLo, \NQwenFeatKDMicroFVsVanillaHi], is larger than
that spread. No other arm has adapters.

\paragraph{Methods not run.} UniCoTT \citep{zhuang_unicott_2025} distils chain, tree and graph explanation structure
into small NLU students, and on chain traces its structural loss is degenerate; MoRSD \citep{yan_towards_2025} has an empty
public repository; CWT\,/\,Skip-Thinking \citep{chen_skip-thinking_2025} and Circuit Distillation
\citep{wadhwa2025circuit} have no public code; SemCoT \citep{he_semcot_2025} makes the student emit implicit tokens, so
it has no rationale to score for fidelity or mechanism.

\subsection{What is matched, seeds and precision}
\label{app:base-matched}
\paragraph{Matched and not matched.} Every arm sees the same shortlist prompts, the same 493 notes and teacher traces,
the same teacher targets, global batch, learning rate and optimiser, and the same evaluation script, decoding and
metrics, for two epochs (\gold{} one). Not matched: (i) hyper-parameter tuning: \mistill's $\lambda$ was selected on the
validation notes, with the range extended below its pre-registered floor (Appendix~\ref{app:obj-lambda}), while the
baselines run at their published defaults (\featkd's weights, \distillm's $\beta$, \opd's $\beta$, and \gold's
on-policy fraction as forced by its collapse); the matching budget for each baseline's own hyper-parameter was not
spent; (ii) wall-clock time and hardware (Appendix~\ref{app:obj-cost}); (iii) the on-policy arms' rollout budgets;
(iv) master-weight precision, below.

\paragraph{Seeds.} Three seeds: \qwenfour{} \vanillakd{} (fp32 and bf16), \featkd{} and \mistill{} (bf16 and fp32). Two: \smol{}
\vanillakd{} (fp32) and \mistill, and, among the controls and ablations, \qwenfour{} \randheads{} and \smol{} bottom-40.
One: every other cell (\distillm, \opd, \gold, \smol{} \featkd{} and \vanillakd{} (bf16), every other control and
ablation), marked in every table. A seed fixes the data order and the arm's
own random draws (\featkd's sampled positions, \randheads' heads, \randpos' positions); evaluation sampling is seeded
per note, independently of the training seed.

\paragraph{Precision.} \mistill, its controls and ablations, the \vanillakd{} runs of the learning curve and the
precision controls train with bf16 weights and no fp32 master copy, updated by plain AdamW at learning rate $10^{-5}$.
The off-policy baselines (\vanillakd, \featkd, \distillm) keep fp32 master weights under bf16 autocast; the on-policy
baselines (\opd, \gold), trained through TRL, load the student in bf16 and keep no fp32 master copy either. An update
smaller than half of bf16's spacing at a weight's magnitude rounds away, so a bf16 run can learn less from the same
data. A matched pair shows it: the fp32 \vanillakd{} pilot and the bf16 \vanillakd{} run of the selection stage share
their 100 notes, recipe, seed and optimiser steps, and their training cross-entropy falls from
\NMDCePilotFpStart{} to \NMDCePilotFpEnd{} and from \NMDCePilotBfStart{} to
\NMDCePilotBfEnd. The difference was found by comparing the dumped configurations of compared arms. It
touches every comparison between \mistill{} (or the on-policy arms) and the fp32 baselines, and none inside the \mistill{}
family or against the untrained student. It was controlled directly (Appendix~\ref{app:res-precision}): \vanillakd{} trained with bf16
weights, three \qwenfour{} seeds and one \smol{} seed, with the reading fixed before the result (at \mistill's task
level, the tier-1 task gap is the precision's; at fp32 \vanillakd's, the objective's; between, both). The task answer
with bf16 weights is the precision's. \mistill{} trained with fp32 master weights at the $\lambda$ selected in bf16
(\qwenfour; seed 0 with the student split over two H100s, seeds 1 and 2 on one B200 each) scores \NQwenMistillFpMicroF{}
and trails fp32 \vanillakd{} (\NQwenMistillFpMicroFVsVanillaDelta{} [\NQwenMistillFpMicroFVsVanillaLo,
\NQwenMistillFpMicroFVsVanillaHi]); with $\lambda$ re-selected for fp32 by the same rule the difference is not detected
(\NQwenMistillFpSelMicroFVsVanillaDelta{} [\NQwenMistillFpSelMicroFVsVanillaLo, \NQwenMistillFpSelMicroFVsVanillaHi];
Appendix~\ref{app:obj-fp32}). The \smol{} \vanillakd{} configurations lack the thinking flag, but for
this model the chat template's default renders the identical thinking-mode prompt, so precision is the one real
difference between those runs and \mistill's.

\subsection{Evaluation protocol and hardware}
\label{app:base-eval}
Every arm, and each untrained student, generates on the 150 test notes in thinking mode under the teacher's decoding:
temperature 0.6, top-$p$ 0.95, top-$k$ 20, sampling seeded per note, a 16{,}384-token cap, and the student's YaRN
setting, in vLLM. Greedy decoding is not used: Qwen3's documentation warns that it loops in thinking mode, and the
first pilot under greedy decoding looped to the cap on every note without an answer; the cap is the teacher's, because
the teacher's own shortlist traces are long (median about \NMDTeacherTraceMedianTokens{} tokens, 90th percentile about \NMDTeacherTraceNinetiethTokens). A mismatch between
\smol's training and evaluation templates cost one round of its evaluations and was fixed before any reported row.
Answers are parsed and scored as in Appendix~\ref{app:metrics}; each evaluation writes its decoding settings to
\texttt{metrics.json}.

Training and evaluation ran on two university GPU clusters (NVIDIA RTX PRO 6000 with 96\,GB, H200 with 141\,GB, B200)
and on rented cloud GPUs (H100, GH200, H200, RTX PRO 6000). \mistill-family runs train on one GPU (fp32 \mistill{} on
two). \featkd{} and \distillm{} need the teacher online: on a 96\,GB card the teacher sits on its own card (for \featkd,
beside a student split over two), while on a 141\,GB H200 everything fits one card. \opd{} and \gold{} colocate trainer
and rollout engine on one H200. Evaluations ran in vLLM on RTX PRO 6000 or H200 cards (the ablation exceptions are listed in
Appendix~\ref{app:abl-runs}). Rented instances held the data only while a run lasted and were terminated once their
outputs had been copied back and verified. The baseline campaign used about \NMDComputeBaselineCampaignGPUHours{} cluster GPU-hours, plus rented GPUs, and the
\mistill{} campaign about \NMDComputeMistillCampaignGPUHours{} GPU-hours, rented ones included (training, evaluation and the own-heads passes).

\section{Metrics}
\label{app:metrics}
\supports{Sec.~\ref{sec:distill-setup}, Sec.~\ref{sec:distill-mech}; \resid{R-D1}--\resid{R-D6}, \resid{R-D15}}

The metrics were fixed before any run, in three tiers, with the headline claim placed on the external tier. As
measured, the external tier shows no detected difference when both objectives train bf16 weights and a cost with fp32 master weights
(Sec.~\ref{sec:distill-main}), and the claims rest on causal
recovery on the students' own heads (Sec.~\ref{sec:distill-mech}), stated beside the task result. Every metric is
computed on the 150 test notes for every arm by one evaluation script.

\paragraph{Scoring an answer.} An arm's answer is the last JSON list after its closing \texttt{</think>}, restricted
to codes of the inventory. A generation that reaches the 16{,}384-token cap, or opens a reasoning block and never
closes it, is invalid and scores as an empty answer; codes are never scraped from unfinished reasoning. The valid rate
is reported beside the task metrics (Table~\ref{tab:main}), and the share of generations that reached the cap in
Table~\ref{tab:app-seeds}.

\paragraph{External task metrics.} \microf{} pools true and false positives over the notes, and \macrof{} averages
per-label F1 over the labels that occur in the gold sets or the predictions. Head and tail \macrof{} split the labels
present in the gold sets at the median training-split frequency among them (tail: at most the median). P@$k$ and
R@$k$ ($k=8,15$) rank the emitted codes in emission order and compute precision over $k$, so short answers are
penalised. \pspk{} and PSnDCG@$k$ \citep{jain2016extreme} weight each correct label by its inverse propensity,
estimated from \mimiciv{} training-split label frequencies with the usual $A=0.55$ and $B=1.5$, and are normalised by
the best value achievable on the note. The gap to the teacher recovered is
$(\text{arm}-\text{base})/(\text{teacher}-\text{base})$, with the teacher's test \microf{}
\NMTeacherShortlistTestMicroF{} on the same prompts. These follow the ICD-coding and extreme-classification
conventions \citep{mullenbach2018explainable, edin2023automated, Bhatia16}.

\paragraph{Fidelity to the teacher.} Teacher-trace \ppl{} is the student's perplexity on the teacher's trace of each test
note under teacher forcing, median over notes. \bleu{} \citep{papineni2002bleu} compares the arm's trace with the
teacher's at sentence level (whitespace tokens, brevity penalty, no smoothing), mean over notes. \las{} is the
leakage-adjusted simulatability used by \citet{wang_scott_2023}, adapted to multi-label coding. One simulator,
\simulator{} without thinking, is fine-tuned once on the teacher's training traces to predict the teacher's final
answer from the note with the rationale and, mixed in at equal weight, from the note alone. Per note,
$\las=\mathrm{F1}\big(\mathrm{sim}(\text{note}\oplus r),a_T\big)-\mathrm{F1}\big(\mathrm{sim}(\text{note}),a_T\big)$,
with $r$ the text inside the arm's reasoning block and $a_T$ the teacher's answer. The simulator does not see the
shortlist, which would add 30k tokens, so its absolute F1 is low and only the difference is read; the note-only term is
shared by every arm.

\paragraph{M1, the supervised readout.} M1 (decision-margin agreement) is read at the teacher's \Pone{}, \Ptwo{} and
\Ptwoc{} events on the teacher's test traces under teacher forcing: the student's margin for the event's target
against its competitors, with the events re-derived in the student's tokenizer (chunk-level decisions for \smol, whose
tokenizer merges digit runs), reported as $P(\margin>0)$ and as the median margin per kind. \Pnull{} events are a control and stay near zero for every arm
(\NMDMOneNullPosMin{} to \NMDMOneNullPosMax). Tables report the share over all events of a kind; paired tests use each note's share,
because events cluster by note. Every arm's cross-entropy acts on these tokens, and \mistill's interaction term
matches the pooled heads' share $c$ of exactly this margin (Eq.~\ref{eq:app-c}), so M1 is the readout the objective
supervises.

\paragraph{M2--M5 and the prefix share, on each arm's own heads.} The attribution pass of
Sec.~\ref{sec:setup-measure} is run on each trained checkpoint over the 100 validation traces and gives that arm's own
calibrated ranking per phase; every read below is then made on the 150 test notes, teacher-forced on the teacher's
traces.
\begin{itemize}[leftmargin=1.2em,itemsep=1pt,topsep=2pt]
  \item \emph{M4, causal recovery.} The arm's own top-$k$ ($k=8,40$) is knocked into a donor note's context at matched
  events, and the recovery of Sec.~\ref{sec:setup-measure},
  $(\margin_{\text{set}}-\margin_{\text{corr}})/(\margin_{\text{all}}-\margin_{\text{corr}})$, is averaged over the
  note's events; tables report the median over notes, beside the same measurement for a count-matched random set.
  Necessity at $k=40$ is the margin drop in nats when the set is mean-ablated; it scales with the clean margin, so it
  is reported, not interpreted, and never compared across hardware.
  \item \emph{M2, region lift.} For the arm's own top-16 heads at a phase: the share of their attention on the
  shortlist lines that falls on the emitted code's chapter (its own line, its category and sibling lines, the rest of
  its chapter), divided by that region's share of the shortlist tokens, so that 1 is chance. Per head it is the median
  over the note's events, averaged over the 16 heads; tables report the median over notes, beside 16 random heads, at
  \Pone{} and \Ptwoc.
  \item \emph{M3, near-miss loop-back.} The share of the own top-16 \Ptwoc{} heads' attention on the earlier mentions
  of the near-miss, against their share on the code's own earlier mentions, for the group and its top head.
  \item \emph{M5, locus distance.} The forcing probe of Sec.~\ref{sec:char-locus} run on the student: the median
  distance in tokens from the locus to the chapter letter, and the shares decided at the event, in the window and
  inherited. Pre-registered for three arms: the untrained \qwenfour, its fp32 \vanillakd{} seed 0 and its \mistill{} seed 0.
  \item \emph{Prefix share} (not pre-registered; added while reading the controls). The share of the own top-16
  \Ptwoc{} heads' attention on the code's already-emitted prefix in the trace.
\end{itemize}
These heads are found after training, by the trained model's own attribution with the corrected hook of
Appendix~\ref{app:instrument} (every student ranking was computed after the fix), and the quantities are not the ones the
loss matches; the loss names only the pools fixed before training. The students' rows are cross-fit: ranked on the 100
validation traces, read on the 150 held-out test notes. The teacher's reference row (HD-002) is read on the 100
characterization notes its sets were ranked on; the discovery is stable across disjoint halves of those notes
(Appendix~\ref{app:heads}, Table~\ref{tab:app-splithalf}), so that row's in-sample reading is a matter of leverage, each
note being one per cent of the ranking mean, not of selection. The two overlap: in every arm the trained
student's own top-40 shares most of its heads with the untrained student's top-40 (Appendix~\ref{app:abl-carriers}),
which is \mistill's supervised pool. What makes M4 non-circular is therefore the re-selection and the measurement ---
restoring a decision by writing the heads' outputs into another note's context, on notes the loss never saw --- not
disjoint heads.

\paragraph{Floors, medians and means.} The random floors of M4 differ by student: at $k=40$ the untrained \qwenfour's
are \NOwnQwenBasePOneKiFortyRand{} (\Pone) and \NOwnQwenBasePTwoCKiFortyRand{} (\Ptwoc), the untrained \smol's
\NOwnSmolBasePOneKiFortyRand{} and \NOwnSmolBasePTwoCKiFortyRand. Absolute recovery is therefore not comparable across
students, and every M4 value is printed beside its floor. The random sets are redrawn for every bin; every own-heads
cell ran the bins $k=8$ and $40$, the teacher's and the 20-note cells of the baseline campaign five bins, so floors
differ by draw while the reported cells do not. M4 values in the tables are medians over notes of per-note means.
Paired comparisons use the per-note means and report the median of the per-note differences; where the text quotes a
mean over notes (the \smol{} seed spreads, the concentration at $k=8$), it says so. Task and fidelity values are means
over seeds of the corpus-level metric, with the between-seed standard deviation.

\paragraph{Statistics.} Task and fidelity differences are paired on the 150 notes, with the seed-averaged per-note
value as the unit: a percentile bootstrap 95\% interval of the mean difference (1{,}000 resamples) and a two-sided
Wilcoxon signed-rank $p$ (zero differences dropped). For \microf{} the per-note value is the note's own F1, so these
intervals are for the mean per-note F1 difference, a different estimand from the difference of the corpus-level
\microf{} values in Table~\ref{tab:main}; the corpus-level difference with notes resampled is reported beside them (Table~\ref{tab:app-paired}). 150 paired notes resolve a per-note \microf{} difference of about
\NMDPowerMdeMicroF{} at 80\% power. Mechanism differences (M4, M2, and M1 per note) use two-sided sign tests on
the per-note values: within a seed pair, and across seed-matched pairs (seed $i$ of one arm against seed $i$ of the
other; a one-seed reference pairs with every seed). Because a note's seed pairs are not independent trials, the test
reported in the tables is \emph{note-clustered}: per note, the mean of its seed pairs' differences, then a two-sided sign
test and a Wilcoxon signed-rank test over the 150 notes, with a bootstrap-over-notes interval of the mean; the pooled
note-pair counts (450 for three pairs) are kept in the released numbers but are not the test. Clustering moves the
medians by less than 0.01 and the $p$-values by orders of magnitude where the pairs disagree: \mistill's gain over bf16
\vanillakd{} at \Ptwoc{} on \qwenfour{} is \NPoolQwenMistillVsVanillaBfPTwoCClusterDelta{} with
\NPoolQwenMistillVsVanillaBfPTwoCClusterHigher{} of 150 notes higher (sign $p$ \NPoolQwenMistillVsVanillaBfPTwoCClusterP,
Wilcoxon \NPoolQwenMistillVsVanillaBfPTwoCClusterWilcoxonP; every seed pair separately $p<10^{-3}$), its smaller gain over
fp32 \vanillakd{} (\NPoolQwenMistillVsVanillaPTwoCClusterDelta) is not significant with the note as the unit
(\NPoolQwenMistillVsVanillaPTwoCClusterHigher{} of 150, $p$ \NPoolQwenMistillVsVanillaPTwoCClusterP; the three seed pairs
disagree in sign) and is reported as descriptive, and every \smol{} gain keeps its sign and significance. On
\qwenfour{} a single-seed comparison is uninterpretable: two fp32 \vanillakd{} seeds separate with a highly significant
sign test (seed~2 minus seed~1, \NPairQwenVanillaSTwoVsVanillaSOnePTwoCDelta{} at \Ptwoc, higher on
\NPairQwenVanillaSTwoVsVanillaSOnePTwoCHigher{} of 150 notes, $p$ \NPairQwenVanillaSTwoVsVanillaSOnePTwoCP), hence the
seed matching. For the task, the per-note intervals are for the mean per-note F1 difference; the corpus-level
(pooled) \microf{} difference, with notes resampled, is reported beside them in Table~\ref{tab:app-paired}; its interval ends lie
within \NCorpusMaxGapToPerNote{} of the per-note interval's, and the two give the same zero-exclusion verdict for every pair but one
(fp32 \mistill{} minus bf16 \mistill, whose corpus interval excludes zero while its per-note interval does not; both are reported).

\paragraph{Readouts that do not discriminate (\resid{R-D15}).} Four readouts are reported for completeness and not used
to rank arms. \emph{M4 at \Pone} is at ceiling in every arm, the untrained students included (\NOwnQwenBasePOneKiForty{}
and \NOwnSmolBasePOneKiForty, against the teacher's \NOwnTeacherPOneKiForty): chapter selection is not what distillation
changes. \emph{M3} is present before distillation and barely moves on \qwenfour{} (the group's near-miss share lies
between \NOwnQwenRandBothSZeroMThreeNearMiss{} and \NOwnQwenBottomSZeroMThreeNearMiss{} across the twenty-three arms read, and
one head, \head{24}{21}, is the top loop-back head in \NMDQwenMThreeTopHeadArms{} of them), while on \smol{} it falls with
distillation (\NOwnSmolBaseMThreeNearMiss{} untrained, \NOwnSmolVanillaSZeroMThreeNearMiss{} for \vanillakd{} seed 0,
\NOwnSmolMistillSZeroMThreeNearMiss{} for \mistill) as \microf{} rises from \NSmolBaseMicroF{} to
\NSmolVanillaSZeroMicroF: a quantity that moves the wrong way across that improvement does not measure what the task
depends on. \emph{M5} places the chapter decision \NOwnQwenVanillaSZeroMFiveLocus--\NOwnQwenBaseMFiveLocus{} tokens
before the letter in all three students read (untrained \NOwnQwenBaseMFiveLocus, \vanillakd{}
\NOwnQwenVanillaSZeroMFiveLocus, \mistill{} \NOwnQwenMistillSZeroMFiveLocus), against the teacher's
\NShortLocusDMedianShort. \emph{M2 on \smol} follows which heads are supervised rather than recovery: \mistill{} and
\randpos{} share a \Ptwoc{} lift (\NOwnSmolMistillSZeroMTwoPTwoCLift{} and \NOwnSmolRandPosSZeroMTwoPTwoCLift) at
recoveries of \NOwnSmolMistillSZeroPTwoCKiForty{} and \NOwnSmolRandPosSZeroPTwoCKiForty. Across the ablations neither M2
nor the prefix share follows recovery on either student (Appendix~\ref{app:abl-readside}), so the claims rest on M4 at
\Ptwoc.

\section{Extended distillation results}
\label{app:results}
\supports{Secs.~\ref{sec:distill-main} and~\ref{sec:distill-mech}; \resid{R-D1}--\resid{R-D6}, \resid{R-D11}, \resid{R-D12}, \resid{R-D15}--\resid{R-D17}}

\subsection{Paired comparisons and per-seed rows}
\label{app:res-paired}
Table~\ref{tab:app-paired} gives the paired differences behind Table~\ref{tab:main} against both references, fp32
\vanillakd{} and \vanillakd{} trained with bf16 weights like \mistill; Table~\ref{tab:app-seeds} gives every seed with
its generation behaviour; and Table~\ref{tab:app-mech-seeds} gives the own-heads readouts of Table~\ref{tab:mech} for
every seed, with the seed-matched comparisons Section~\ref{sec:distill-mech} cites.
For calibration: supervised classifiers trained on the release's full 89{,}099-note training split reach
\microf{} 0.585 on its combined diagnosis-and-procedure task (PLM-ICD, ten seeds; \citealp{edin2023automated}) ---
a different regime (full supervision, classification heads, all 7{,}942 labels) from the zero-shot teacher and
few-hundred-note distillation studied here, quoted so the absolute \microf{} level can be placed. A second
calibration is the retriever alone (Table~\ref{tab:app-retriever}): the first $k$ codes of the shortlist that every arm
sees, scored as the prediction. Its top-20 reaches \microf{} \NRetrieverTopTwentyMicroF{} (precision
\NRetrieverTopTwentyMicroP, recall \NRetrieverTopTwentyMicroR), just below the teacher's \NMTeacherShortlistTestMicroF{} and
above every student, and its top-10 (\NRetrieverTopTenMicroF) matches fp32 \vanillakd; the students answer with short,
precision-heavy lists (\mistill{} seed 0: precision \NQwenMistillSZeroMicroP, recall \NQwenMistillSZeroMicroR), so the
generator adds selection beyond a fixed retrieval cut-off only for the teacher. The task scores in this paper are low in
absolute terms, and the mechanism claims do not rest on them. Against fp32 \vanillakd, \distillm{} (fp32) is not distinguished from it on the task and \featkd{}
(fp32) is lower; \opd, which trains bf16 weights, is also not distinguished from it on the task, while \mistill{} and bf16
\vanillakd, both off-policy with bf16 weights, are lower by similar amounts, and \mistill{} trained with fp32 master
weights is lower too (\NQwenMistillFpMicroFVsVanillaDelta{} [\NQwenMistillFpMicroFVsVanillaLo, \NQwenMistillFpMicroFVsVanillaHi]). Against bf16 \vanillakd, the precision-matched reference, \mistill{} is level on \microf,
\pspfifteen{} and \las{} on both students. Plain \vanillakd{} recovers \NMDQwenVanillaGapClosed{}
of the gap between the untrained \qwenfour{} and the teacher from 493 notes, and turns a cross-family student
that could not finish a trace into a working one. Its cost shows in validity: at fp32 a sixth of its \qwenfour{}
generations run to the cap in a repetition loop (Table~\ref{tab:app-seeds}), where the untrained student almost never
does; bf16 \vanillakd{} runs to the cap less often, and \mistill{} is at its level. The \mistill{} deficit against fp32
\vanillakd{} lies in the completed answers, at both precision (\NQwenMistillMicroP{} against \NQwenVanillaMicroP) and
recall (\NQwenMistillMicroR{} against \NQwenVanillaMicroR), and bf16 \vanillakd{} shows the same deficit
(\NQwenVanillaBfMicroP, \NQwenVanillaBfMicroR). On \smol{} the untrained student opens a reasoning block on every note
but rarely closes it (valid \NSmolBaseValid); fp32 \vanillakd{} teaches it to stop on about half the notes (\NSmolVanillaValid), while bf16
\vanillakd{} (\NSmolVanillaBfValid), \mistill{} (\NSmolMistillValid) and \featkd{} (\NSmolFeatKDValid) leave most
generations at the cap.

\begin{table}[!htbp]
  \centering\small
  \caption{Paired differences behind Table~\ref{tab:main}: arm minus reference on the same 150 test notes,
  seed-averaged per note (for \microf, the mean per-note F1 difference and, beside it, the difference of the corpus-level
  \microf{} with notes resampled 10{,}000 times, for the headline pairs), with percentile bootstrap 95\% intervals and
  Wilcoxon signed-rank or bootstrap $p$; references fp32 \vanillakd{} and bf16 \vanillakd; \opd{} and \gold{} train bf16
  weights. \resid{R-D1}\resid{R-D2}\resid{R-D3}\resid{R-D4}\resid{R-D5}\resid{R-D17}\resid{R-D18}}
  \label{tab:app-paired}
  \providecommand{\gensd}[2]{\begin{tabular}[t]{@{}c@{}}#1\\[-2pt]{\scriptsize$\pm$#2}\end{tabular}}
\providecommand{\genstack}[2]{\begin{tabular}[t]{@{}c@{}}#1\\[-2pt]{\scriptsize #2}\end{tabular}}
\providecommand{\genstackl}[2]{\begin{tabular}[t]{@{}l@{}}#1\\[-2pt]{\scriptsize #2}\end{tabular}}
\providecommand{\genhead}[2]{\begin{tabular}[b]{@{}c@{}}#1\\#2\end{tabular}}
\providecommand{\genstackiii}[3]{\begin{tabular}[t]{@{}c@{}}#1\\[-2pt]{\scriptsize #2}\\[-3pt]{\scriptsize #3}\end{tabular}}
\footnotesize\setlength{\tabcolsep}{4pt}
\begin{adjustbox}{max width=\linewidth}
\begin{tabular}{@{}l cccc@{}}
\toprule
Arm & \genhead{$\Delta$ per-note F1}{\relax[95\% CI]} & \genhead{$\Delta$ corpus \microf}{\relax[95\% CI], notes resampled} & $\Delta$ \pspfifteen{} [95\% CI] & $\Delta$ \las{} [95\% CI] \\
\midrule
\multicolumn{5}{@{}l}{\qwenfour} \\
\multicolumn{5}{@{}l}{\quad\textit{against \vanillakd}} \\
\quad Base & \genstack{\pci{\ensuremath{-}0.081}{\ensuremath{-}0.101}{\ensuremath{-}0.063}}{$p$ \ensuremath{3{\times}10^{-11}}} & --- & \genstack{\pci{\ensuremath{-}0.064}{\ensuremath{-}0.080}{\ensuremath{-}0.050}}{$p$ \ensuremath{5{\times}10^{-12}}} & \genstack{\pci{\ensuremath{-}0.174}{\ensuremath{-}0.206}{\ensuremath{-}0.142}}{$p$ \ensuremath{3{\times}10^{-18}}} \\
\quad \vanillakd{} (bf16) & \genstack{\pci{\ensuremath{-}0.041}{\ensuremath{-}0.056}{\ensuremath{-}0.026}}{$p$ \ensuremath{4{\times}10^{-6}}} & --- & \genstack{\pci{\ensuremath{-}0.037}{\ensuremath{-}0.050}{\ensuremath{-}0.025}}{$p$ \ensuremath{8{\times}10^{-7}}} & \genstack{\pci{\ensuremath{-}0.084}{\ensuremath{-}0.105}{\ensuremath{-}0.065}}{$p$ \ensuremath{10^{-12}}} \\
\quad \featkd & \genstack{\pci{\ensuremath{-}0.035}{\ensuremath{-}0.049}{\ensuremath{-}0.021}}{$p$ \ensuremath{2{\times}10^{-5}}} & --- & \genstack{\pci{\ensuremath{-}0.034}{\ensuremath{-}0.045}{\ensuremath{-}0.022}}{$p$ \ensuremath{3{\times}10^{-7}}} & \genstack{\pci{\ensuremath{-}0.054}{\ensuremath{-}0.074}{\ensuremath{-}0.035}}{$p$ \ensuremath{3{\times}10^{-8}}} \\
\quad \distillm & \genstack{\pci{+0.003}{\ensuremath{-}0.016}{+0.021}}{$p$ 0.57} & --- & \genstack{\pci{+0.001}{\ensuremath{-}0.015}{+0.017}}{$p$ 0.80} & \genstack{\pci{+0.004}{\ensuremath{-}0.027}{+0.032}}{$p$ 0.37} \\
\quad \opd & \genstack{\pci{\ensuremath{-}0.005}{\ensuremath{-}0.025}{+0.015}}{$p$ 0.67} & --- & \genstack{\pci{\ensuremath{-}0.007}{\ensuremath{-}0.023}{+0.010}}{$p$ 0.37} & \genstack{\pci{\ensuremath{-}0.038}{\ensuremath{-}0.065}{\ensuremath{-}0.011}}{$p$ 0.006} \\
\quad \mistill & \genstack{\pci{\ensuremath{-}0.034}{\ensuremath{-}0.050}{\ensuremath{-}0.018}}{$p$ \ensuremath{4{\times}10^{-5}}} & \genstack{\pci{\ensuremath{-}0.038}{\ensuremath{-}0.054}{\ensuremath{-}0.021}}{$p$ \ensuremath{<10^{-4}}} & \genstack{\pci{\ensuremath{-}0.030}{\ensuremath{-}0.044}{\ensuremath{-}0.016}}{$p$ \ensuremath{3{\times}10^{-5}}} & \genstack{\pci{\ensuremath{-}0.075}{\ensuremath{-}0.095}{\ensuremath{-}0.056}}{$p$ \ensuremath{10^{-11}}} \\
\quad \mistill{} (fp32, $\lambda$ 0.1) & \genstack{\pci{\ensuremath{-}0.024}{\ensuremath{-}0.039}{\ensuremath{-}0.009}}{$p$ 0.001} & \genstack{\pci{\ensuremath{-}0.022}{\ensuremath{-}0.037}{\ensuremath{-}0.008}}{$p$ 0.003} & \genstack{\pci{\ensuremath{-}0.023}{\ensuremath{-}0.035}{\ensuremath{-}0.010}}{$p$ 0.001} & \genstack{\pci{\ensuremath{-}0.039}{\ensuremath{-}0.058}{\ensuremath{-}0.021}}{$p$ \ensuremath{4{\times}10^{-5}}} \\
\quad \mistill{} (fp32, $\lambda$ 0.01) & \genstack{\pci{\ensuremath{-}0.013}{\ensuremath{-}0.028}{+0.002}}{$p$ 0.05} & \genstack{\pci{\ensuremath{-}0.012}{\ensuremath{-}0.026}{+0.003}}{$p$ 0.10} & \genstack{\pci{\ensuremath{-}0.010}{\ensuremath{-}0.023}{+0.003}}{$p$ 0.04} & \genstack{\pci{\ensuremath{-}0.016}{\ensuremath{-}0.035}{+0.003}}{$p$ 0.15} \\
\multicolumn{5}{@{}l}{\quad\textit{against \vanillakd{} (bf16)}} \\
\quad Base & \genstack{\pci{\ensuremath{-}0.040}{\ensuremath{-}0.056}{\ensuremath{-}0.023}}{$p$ \ensuremath{5{\times}10^{-6}}} & --- & \genstack{\pci{\ensuremath{-}0.027}{\ensuremath{-}0.039}{\ensuremath{-}0.015}}{$p$ \ensuremath{2{\times}10^{-5}}} & \genstack{\pci{\ensuremath{-}0.090}{\ensuremath{-}0.115}{\ensuremath{-}0.062}}{$p$ \ensuremath{5{\times}10^{-11}}} \\
\quad \vanillakd & \genstack{\pci{+0.041}{+0.026}{+0.056}}{$p$ \ensuremath{4{\times}10^{-6}}} & --- & \genstack{\pci{+0.037}{+0.025}{+0.050}}{$p$ \ensuremath{8{\times}10^{-7}}} & \genstack{\pci{+0.084}{+0.065}{+0.105}}{$p$ \ensuremath{10^{-12}}} \\
\quad \featkd & \genstack{\pci{+0.006}{\ensuremath{-}0.007}{+0.021}}{$p$ 0.57} & --- & \genstack{\pci{+0.004}{\ensuremath{-}0.008}{+0.015}}{$p$ 0.57} & \genstack{\pci{+0.030}{+0.012}{+0.048}}{$p$ 0.001} \\
\quad \distillm & \genstack{\pci{+0.044}{+0.024}{+0.063}}{$p$ \ensuremath{4{\times}10^{-5}}} & --- & \genstack{\pci{+0.038}{+0.021}{+0.054}}{$p$ \ensuremath{3{\times}10^{-5}}} & \genstack{\pci{+0.088}{+0.060}{+0.116}}{$p$ \ensuremath{10^{-8}}} \\
\quad \opd & \genstack{\pci{+0.037}{+0.018}{+0.055}}{$p$ \ensuremath{4{\times}10^{-4}}} & --- & \genstack{\pci{+0.030}{+0.014}{+0.046}}{$p$ \ensuremath{4{\times}10^{-4}}} & \genstack{\pci{+0.046}{+0.021}{+0.073}}{$p$ 0.001} \\
\quad \mistill & \genstack{\pci{+0.007}{\ensuremath{-}0.005}{+0.019}}{$p$ 0.27} & \genstack{\pci{+0.006}{\ensuremath{-}0.007}{+0.019}}{$p$ 0.35} & \genstack{\pci{+0.007}{\ensuremath{-}0.004}{+0.019}}{$p$ 0.13} & \genstack{\pci{+0.009}{\ensuremath{-}0.007}{+0.025}}{$p$ 0.30} \\
\quad \mistill{} (fp32, $\lambda$ 0.1) & \genstack{\pci{+0.017}{+0.004}{+0.030}}{$p$ 0.02} & \genstack{\pci{+0.022}{+0.008}{+0.035}}{$p$ 0.002} & \genstack{\pci{+0.015}{+0.004}{+0.027}}{$p$ 0.01} & \genstack{\pci{+0.045}{+0.025}{+0.066}}{$p$ \ensuremath{6{\times}10^{-5}}} \\
\quad \mistill{} (fp32, $\lambda$ 0.01) & \genstack{\pci{+0.028}{+0.012}{+0.044}}{$p$ 0.004} & \genstack{\pci{+0.032}{+0.017}{+0.047}}{$p$ \ensuremath{<10^{-4}}} & \genstack{\pci{+0.027}{+0.014}{+0.040}}{$p$ \ensuremath{5{\times}10^{-4}}} & \genstack{\pci{+0.068}{+0.048}{+0.089}}{$p$ \ensuremath{2{\times}10^{-9}}} \\
\midrule
\multicolumn{5}{@{}l}{\smol} \\
\multicolumn{5}{@{}l}{\quad\textit{against \vanillakd}} \\
\quad Base & \genstack{\pci{\ensuremath{-}0.112}{\ensuremath{-}0.130}{\ensuremath{-}0.094}}{$p$ \ensuremath{3{\times}10^{-19}}} & --- & \genstack{\pci{\ensuremath{-}0.083}{\ensuremath{-}0.097}{\ensuremath{-}0.069}}{$p$ \ensuremath{3{\times}10^{-19}}} & \genstack{\pci{\ensuremath{-}0.244}{\ensuremath{-}0.273}{\ensuremath{-}0.217}}{$p$ \ensuremath{8{\times}10^{-23}}} \\
\quad \vanillakd{} (bf16) & \genstack{\pci{\ensuremath{-}0.069}{\ensuremath{-}0.092}{\ensuremath{-}0.047}}{$p$ \ensuremath{2{\times}10^{-8}}} & --- & \genstack{\pci{\ensuremath{-}0.054}{\ensuremath{-}0.072}{\ensuremath{-}0.038}}{$p$ \ensuremath{4{\times}10^{-9}}} & \genstack{\pci{\ensuremath{-}0.145}{\ensuremath{-}0.179}{\ensuremath{-}0.112}}{$p$ \ensuremath{5{\times}10^{-12}}} \\
\quad \featkd & \genstack{\pci{\ensuremath{-}0.060}{\ensuremath{-}0.080}{\ensuremath{-}0.038}}{$p$ \ensuremath{2{\times}10^{-7}}} & --- & \genstack{\pci{\ensuremath{-}0.046}{\ensuremath{-}0.062}{\ensuremath{-}0.028}}{$p$ \ensuremath{3{\times}10^{-7}}} & \genstack{\pci{\ensuremath{-}0.109}{\ensuremath{-}0.142}{\ensuremath{-}0.077}}{$p$ \ensuremath{2{\times}10^{-9}}} \\
\quad \gold{} (6{,}144 cap) & \genstack{\pci{\ensuremath{-}0.115}{\ensuremath{-}0.133}{\ensuremath{-}0.097}}{$p$ \ensuremath{3{\times}10^{-19}}} & --- & \genstack{\pci{\ensuremath{-}0.085}{\ensuremath{-}0.100}{\ensuremath{-}0.072}}{$p$ \ensuremath{3{\times}10^{-19}}} & \genstack{\pci{\ensuremath{-}0.277}{\ensuremath{-}0.305}{\ensuremath{-}0.251}}{$p$ \ensuremath{7{\times}10^{-24}}} \\
\quad \mistill & \genstack{\pci{\ensuremath{-}0.075}{\ensuremath{-}0.092}{\ensuremath{-}0.058}}{$p$ \ensuremath{8{\times}10^{-14}}} & \genstack{\pci{\ensuremath{-}0.084}{\ensuremath{-}0.104}{\ensuremath{-}0.065}}{$p$ \ensuremath{<10^{-4}}} & \genstack{\pci{\ensuremath{-}0.058}{\ensuremath{-}0.071}{\ensuremath{-}0.045}}{$p$ \ensuremath{10^{-13}}} & \genstack{\pci{\ensuremath{-}0.151}{\ensuremath{-}0.179}{\ensuremath{-}0.126}}{$p$ \ensuremath{10^{-17}}} \\
\multicolumn{5}{@{}l}{\quad\textit{against \vanillakd{} (bf16)}} \\
\quad Base & \genstack{\pci{\ensuremath{-}0.043}{\ensuremath{-}0.058}{\ensuremath{-}0.028}}{$p$ \ensuremath{2{\times}10^{-6}}} & --- & \genstack{\pci{\ensuremath{-}0.029}{\ensuremath{-}0.040}{\ensuremath{-}0.019}}{$p$ \ensuremath{6{\times}10^{-8}}} & \genstack{\pci{\ensuremath{-}0.100}{\ensuremath{-}0.128}{\ensuremath{-}0.068}}{$p$ \ensuremath{3{\times}10^{-9}}} \\
\quad \vanillakd & \genstack{\pci{+0.069}{+0.047}{+0.092}}{$p$ \ensuremath{2{\times}10^{-8}}} & --- & \genstack{\pci{+0.054}{+0.038}{+0.072}}{$p$ \ensuremath{4{\times}10^{-9}}} & \genstack{\pci{+0.145}{+0.112}{+0.179}}{$p$ \ensuremath{5{\times}10^{-12}}} \\
\quad \featkd & \genstack{\pci{+0.009}{\ensuremath{-}0.011}{+0.029}}{$p$ 0.37} & --- & \genstack{\pci{+0.009}{\ensuremath{-}0.006}{+0.023}}{$p$ 0.22} & \genstack{\pci{+0.036}{+0.001}{+0.071}}{$p$ 0.05} \\
\quad \gold{} (6{,}144 cap) & \genstack{\pci{\ensuremath{-}0.046}{\ensuremath{-}0.060}{\ensuremath{-}0.032}}{$p$ \ensuremath{10^{-7}}} & --- & \genstack{\pci{\ensuremath{-}0.031}{\ensuremath{-}0.041}{\ensuremath{-}0.021}}{$p$ \ensuremath{10^{-11}}} & \genstack{\pci{\ensuremath{-}0.132}{\ensuremath{-}0.163}{\ensuremath{-}0.104}}{$p$ \ensuremath{2{\times}10^{-13}}} \\
\quad \mistill & \genstack{\pci{\ensuremath{-}0.006}{\ensuremath{-}0.021}{+0.011}}{$p$ 0.78} & \genstack{\pci{\ensuremath{-}0.009}{\ensuremath{-}0.027}{+0.009}}{$p$ 0.33} & \genstack{\pci{\ensuremath{-}0.004}{\ensuremath{-}0.014}{+0.008}}{$p$ 0.93} & \genstack{\pci{\ensuremath{-}0.007}{\ensuremath{-}0.037}{+0.026}}{$p$ 0.90} \\
\bottomrule
\end{tabular}
\end{adjustbox}

\end{table}

\begin{table}[!htbp]
  \centering\small
  \caption{The retriever alone as the prediction on the 150 test notes: the first $k$ codes of the RT-003 shortlist, and
  the whole category-completed shortlist, scored with the paper's scorer; recall is the pooled recall of the gold codes.
  \resid{R-D21}}
  \label{tab:app-retriever}
  \providecommand{\gensd}[2]{\begin{tabular}[t]{@{}c@{}}#1\\[-2pt]{\scriptsize$\pm$#2}\end{tabular}}
\providecommand{\genstack}[2]{\begin{tabular}[t]{@{}c@{}}#1\\[-2pt]{\scriptsize #2}\end{tabular}}
\providecommand{\genstackl}[2]{\begin{tabular}[t]{@{}l@{}}#1\\[-2pt]{\scriptsize #2}\end{tabular}}
\providecommand{\genhead}[2]{\begin{tabular}[b]{@{}c@{}}#1\\#2\end{tabular}}
\providecommand{\genstackiii}[3]{\begin{tabular}[t]{@{}c@{}}#1\\[-2pt]{\scriptsize #2}\\[-3pt]{\scriptsize #3}\end{tabular}}
\footnotesize\setlength{\tabcolsep}{4pt}
\begin{adjustbox}{max width=\linewidth}
\begin{tabular}{@{}l r cccc ccc c@{}}
\toprule
Prediction & length & \microf & micro-P & micro-R & \macrof & P@8 & P@15 & \pspfifteen & recall \\
\midrule
retriever top-5 & 5 & 0.126 & 0.256 & 0.083 & 0.044 & 0.160 & 0.085 & 0.080 & 0.083 \\
retriever top-10 & 10 & 0.179 & 0.227 & 0.148 & 0.060 & 0.243 & 0.151 & 0.144 & 0.148 \\
retriever top-20 & 20 & 0.217 & 0.191 & 0.249 & 0.071 & 0.243 & 0.205 & 0.199 & 0.249 \\
retriever top-1{,}000 & 1{,}000 & 0.028 & 0.014 & 0.917 & 0.012 & 0.243 & 0.205 & 0.199 & 0.917 \\
the shortlist (the prompt's candidate block) & 1{,}537 & 0.018 & 0.009 & 0.933 & 0.010 & 0.243 & 0.205 & 0.199 & 0.933 \\
\bottomrule
\end{tabular}
\end{adjustbox}

\end{table}

\begin{table}[!htbp]
  \centering\small
  \caption{Per-seed rows and seed-matched comparisons behind Table~\ref{tab:mech} (150 test notes; medians over notes;
  random-40 floors in parentheses; task: that seed's \microf). Seed-matched lines: per-note differences note-clustered over
  the seed pairs (per-note mean over the pairs; median; notes higher\,/\,lower of 150; sign test over notes), M2 the same
  over per-note lifts, and task the paired mean difference with its 95\% interval. \resid{R-D6}\resid{R-D12}\resid{R-D15}\resid{R-D17}}
  \label{tab:app-mech-seeds}
  \providecommand{\gensd}[2]{\begin{tabular}[t]{@{}c@{}}#1\\[-2pt]{\scriptsize$\pm$#2}\end{tabular}}
\providecommand{\genstack}[2]{\begin{tabular}[t]{@{}c@{}}#1\\[-2pt]{\scriptsize #2}\end{tabular}}
\providecommand{\genstackl}[2]{\begin{tabular}[t]{@{}l@{}}#1\\[-2pt]{\scriptsize #2}\end{tabular}}
\providecommand{\genhead}[2]{\begin{tabular}[b]{@{}c@{}}#1\\#2\end{tabular}}
\providecommand{\genstackiii}[3]{\begin{tabular}[t]{@{}c@{}}#1\\[-2pt]{\scriptsize #2}\\[-3pt]{\scriptsize #3}\end{tabular}}
\footnotesize\setlength{\tabcolsep}{3.5pt}
\begin{adjustbox}{max width=\linewidth}
\begin{tabular}{@{}l cccccc@{}}
\toprule
 & \multicolumn{3}{c}{M4 knock-in, own top-40} & \multicolumn{2}{c}{read side (\Ptwoc)} & task \\
\cmidrule(lr){2-4}\cmidrule(lr){5-6}\cmidrule(lr){7-7}
Arm & \genhead{\Ptwoc}{(random)} & \Ptwo & \Pone & \genhead{M2 lift}{(random)} & \genhead{prefix}{share} & \microf \\
\midrule
\genstackl{Teacher (\teacher)}{HD-002, 100 val.\ notes} & 0.847\,{\scriptsize(0.015)} & 0.821 & 0.915 & 3.38\,{\scriptsize(1.95)} & 0.260 & --- \\
\midrule
\multicolumn{7}{@{}l}{\qwenfour} \\
Base & 0.788\,{\scriptsize(0.030)} & 0.815 & 0.946 & 2.56\,{\scriptsize(1.17)} & 0.185 & 0.104 \\
\vanillakd s0 & 0.840\,{\scriptsize(0.035)} & 0.749 & 0.945 & 3.23\,{\scriptsize(1.46)} & 0.196 & 0.193 \\
\quad s1 & 0.848\,{\scriptsize(0.038)} & 0.763 & 0.948 & 3.30\,{\scriptsize(1.28)} & 0.193 & 0.189 \\
\quad s2 & 0.891\,{\scriptsize(0.035)} & 0.664 & 0.943 & 3.81\,{\scriptsize(1.26)} & 0.236 & 0.180 \\
\vanillakd{} (bf16) s0 & 0.834\,{\scriptsize(0.037)} & 0.751 & 0.952 & 3.36\,{\scriptsize(1.53)} & 0.187 & 0.136 \\
\quad s1 & 0.835\,{\scriptsize(0.036)} & 0.758 & 0.951 & 3.60\,{\scriptsize(1.60)} & 0.188 & 0.148 \\
\quad s2 & 0.847\,{\scriptsize(0.037)} & 0.764 & 0.952 & 3.47\,{\scriptsize(1.62)} & 0.188 & 0.147 \\
\featkd & 0.844\,{\scriptsize(0.048)} & 0.888 & 0.962 & 3.05\,{\scriptsize(1.57)} & 0.228 & 0.142 \\
\mistill s0 & 0.873\,{\scriptsize(0.025)} & 0.736 & 0.956 & 3.65\,{\scriptsize(1.56)} & 0.264 & 0.151 \\
\quad s1 & 0.870\,{\scriptsize(0.025)} & 0.736 & 0.955 & 3.67\,{\scriptsize(1.55)} & 0.265 & 0.147 \\
\quad s2 & 0.874\,{\scriptsize(0.026)} & 0.734 & 0.958 & 3.57\,{\scriptsize(1.70)} & 0.277 & 0.150 \\
\mistill{} (fp32, $\lambda$ 0.1) s0 & 0.870\,{\scriptsize(0.022)} & 0.791 & 0.949 & 3.00\,{\scriptsize(1.76)} & 0.362 & 0.161 \\
\quad s1 & 0.836\,{\scriptsize(0.028)} & 0.755 & 0.949 & 3.14\,{\scriptsize(2.25)} & 0.339 & 0.165 \\
\quad s2 & 0.850\,{\scriptsize(0.023)} & 0.799 & 0.949 & 2.87\,{\scriptsize(1.66)} & 0.346 & 0.169 \\
\mistill{} (fp32, $\lambda$ 0.01) s0 & 0.892\,{\scriptsize(0.032)} & 0.670 & 0.949 & 3.48\,{\scriptsize(1.41)} & 0.344 & 0.177 \\
\quad s1 & 0.897\,{\scriptsize(0.032)} & 0.686 & 0.951 & 3.59\,{\scriptsize(1.52)} & 0.339 & 0.184 \\
\quad s2 & 0.892\,{\scriptsize(0.031)} & 0.685 & 0.948 & 3.17\,{\scriptsize(0.96)} & 0.341 & 0.166 \\
\randheads & 0.869\,{\scriptsize(0.032)} & 0.880 & 0.958 & 3.66\,{\scriptsize(2.21)} & 0.290 & 0.133 \\
\randpos & 0.631\,{\scriptsize(0.042)} & 0.771 & 0.907 & 2.93\,{\scriptsize(1.45)} & 0.076 & 0.129 \\
\genstackl{\mistill $-$ \vanillakd}{seed-matched, 3 pairs; note-clustered} & \genstackiii{+0.011}{84/66}{$p$ 0.16} & \genstackiii{+0.019}{107/43}{$p$ \ensuremath{2{\times}10^{-7}}} & \genstackiii{+0.013}{105/45}{$p$ \ensuremath{10^{-6}}} & \genstackiii{+0.113}{100/50}{$p$ \ensuremath{5{\times}10^{-5}}} & --- & \genstack{\ensuremath{-}0.034}{[\ensuremath{-}0.050, \ensuremath{-}0.018]} \\
\genstackl{\mistill $-$ \vanillakd{} (bf16)}{seed-matched, 3 pairs; note-clustered} & \genstackiii{+0.027}{107/43}{$p$ \ensuremath{2{\times}10^{-7}}} & \genstackiii{\ensuremath{-}0.020}{42/108}{$p$ \ensuremath{7{\times}10^{-8}}} & \genstackiii{+0.006}{96/54}{$p$ \ensuremath{8{\times}10^{-4}}} & \genstackiii{+0.253}{118/32}{$p$ \ensuremath{9{\times}10^{-13}}} & --- & \genstack{+0.007}{[\ensuremath{-}0.005, +0.019]} \\
\genstackl{\mistill{} (fp32, $\lambda$ 0.1) $-$ \vanillakd}{seed-matched, 3 pairs; note-clustered} & \genstackiii{\ensuremath{-}0.005}{71/79}{$p$ 0.57} & \genstackiii{+0.057}{143/7}{$p$ \ensuremath{4{\times}10^{-34}}} & \genstackiii{+0.005}{86/64}{$p$ 0.09} & \genstackiii{\ensuremath{-}0.423}{13/137}{$p$ \ensuremath{3{\times}10^{-27}}} & --- & \genstack{\ensuremath{-}0.024}{[\ensuremath{-}0.039, \ensuremath{-}0.009]} \\
\genstackl{\mistill{} (fp32, $\lambda$ 0.1) $-$ \vanillakd{} (bf16)}{seed-matched, 3 pairs; note-clustered} & \genstackiii{+0.004}{80/70}{$p$ 0.46} & \genstackiii{+0.022}{99/51}{$p$ \ensuremath{10^{-4}}} & \genstackiii{\ensuremath{-}0.002}{70/80}{$p$ 0.46} & \genstackiii{\ensuremath{-}0.308}{33/117}{$p$ \ensuremath{3{\times}10^{-12}}} & --- & \genstack{+0.017}{[+0.004, +0.030]} \\
\genstackl{\mistill{} (fp32, $\lambda$ 0.1) $-$ \mistill{} (bf16)}{seed-matched, 3 pairs; note-clustered} & \genstackiii{\ensuremath{-}0.014}{58/92}{$p$ 0.007} & \genstackiii{+0.037}{123/27}{$p$ \ensuremath{8{\times}10^{-16}}} & \genstackiii{\ensuremath{-}0.006}{57/93}{$p$ 0.004} & \genstackiii{\ensuremath{-}0.562}{4/146}{$p$ \ensuremath{3{\times}10^{-38}}} & --- & \genstack{+0.010}{[\ensuremath{-}0.005, +0.025]} \\
\genstackl{\mistill{} (fp32, $\lambda$ 0.01) $-$ \vanillakd}{seed-matched, 3 pairs; note-clustered} & \genstackiii{+0.032}{118/32}{$p$ \ensuremath{9{\times}10^{-13}}} & \genstackiii{\ensuremath{-}0.041}{5/145}{$p$ \ensuremath{9{\times}10^{-37}}} & \genstackiii{+0.006}{96/54}{$p$ \ensuremath{8{\times}10^{-4}}} & \genstackiii{\ensuremath{-}0.199}{47/103}{$p$ \ensuremath{6{\times}10^{-6}}} & --- & \genstack{\ensuremath{-}0.013}{[\ensuremath{-}0.028, +0.002]} \\
\midrule
\multicolumn{7}{@{}l}{\smol} \\
Base & 0.416\,{\scriptsize(0.107)} & 0.809 & 0.985 & 3.57\,{\scriptsize(1.51)} & 0.121 & 0.003 \\
\vanillakd s0 & 0.541\,{\scriptsize(0.105)} & 0.735 & 0.975 & 3.75\,{\scriptsize(1.64)} & 0.165 & 0.148 \\
\quad s1 & 0.523\,{\scriptsize(0.098)} & 0.714 & 0.978 & 3.87\,{\scriptsize(1.34)} & 0.180 & 0.113 \\
\vanillakd{} (bf16) & 0.476\,{\scriptsize(0.098)} & 0.867 & 0.985 & 3.29\,{\scriptsize(1.39)} & 0.151 & 0.055 \\
\featkd & 0.628\,{\scriptsize(0.099)} & 0.933 & 0.988 & 2.74\,{\scriptsize(1.37)} & 0.097 & 0.062 \\
\mistill s0 & 0.789\,{\scriptsize(0.121)} & 0.866 & 1.004 & 2.65\,{\scriptsize(1.31)} & 0.439 & 0.060 \\
\quad s1 & 0.774\,{\scriptsize(0.127)} & 0.868 & 1.006 & 2.65\,{\scriptsize(1.37)} & 0.441 & 0.042 \\
\quad s2 & 0.750\,{\scriptsize(0.121)} & 0.857 & 1.005 & 2.65\,{\scriptsize(1.31)} & 0.442 & 0.036 \\
\randheads & 0.585\,{\scriptsize(0.116)} & 0.725 & 0.999 & 3.93\,{\scriptsize(2.01)} & 0.284 & 0.061 \\
\randpos & 0.304\,{\scriptsize(0.095)} & 0.891 & 0.993 & 2.52\,{\scriptsize(1.36)} & 0.050 & 0.063 \\
\genstackl{\mistill $-$ \vanillakd}{seed-matched, 3 pairs; note-clustered} & \genstackiii{+0.233}{132/18}{$p$ \ensuremath{10^{-22}}} & \genstackiii{+0.129}{136/14}{$p$ \ensuremath{3{\times}10^{-26}}} & \genstackiii{+0.028}{143/7}{$p$ \ensuremath{4{\times}10^{-34}}} & \genstackiii{\ensuremath{-}1.186}{0/150}{$p$ \ensuremath{10^{-45}}} & --- & \genstack{\ensuremath{-}0.075}{[\ensuremath{-}0.092, \ensuremath{-}0.058]} \\
\genstackl{\mistill $-$ \vanillakd{} (bf16)}{seed-matched, 3 pairs; note-clustered} & \genstackiii{+0.310}{130/20}{$p$ \ensuremath{6{\times}10^{-21}}} & \genstackiii{+0.013}{87/63}{$p$ 0.06} & \genstackiii{+0.016}{121/29}{$p$ \ensuremath{10^{-14}}} & \genstackiii{\ensuremath{-}0.659}{5/145}{$p$ \ensuremath{9{\times}10^{-37}}} & --- & \genstack{\ensuremath{-}0.006}{[\ensuremath{-}0.021, +0.011]} \\
\bottomrule
\end{tabular}
\end{adjustbox}

\end{table}

\begin{table}[!htbp]
  \centering\small
  \caption{Per-seed rows behind Table~\ref{tab:main}, with generation behaviour (150 test notes). Valid: share of
  generations with a parsed final answer; cap: share that reached the 16{,}384-token cap; tokens: mean generated tokens.
  (bf16) rows train bf16 weights without an fp32 master copy, as \mistill{} does; so do \opd{} and \gold{} (TRL).
  \gold{} ran with a 6{,}144-token rollout cap. \resid{R-D1}\resid{R-D2}\resid{R-D3}\resid{R-D4}\resid{R-D5}\resid{R-D17}}
  \label{tab:app-seeds}
  \scriptsize\setlength{\tabcolsep}{3pt}
  \begin{adjustbox}{max width=\linewidth}
  \begin{tabular}{@{}l c ccc ccc c ccc@{}}
    \toprule
    Arm & seed & \microf & micro-P & micro-R & valid & cap & tokens & \las & M1 \Pone & M1 \Ptwo & M1 \Ptwoc \\
    \midrule
    \multicolumn{12}{@{}l}{\qwenfour} \\
    Base & --- & \NQwenBaseSZeroMicroF{} & \NQwenBaseSZeroMicroP{} & \NQwenBaseSZeroMicroR{} & \NQwenBaseSZeroValid{} & \NQwenBaseSZeroHitCap{} & \NQwenBaseSZeroGenTokens{} & \NQwenBaseSZeroLAS{} & \NQwenBaseSZeroMOnePOne{} & \NQwenBaseSZeroMOnePTwo{} & \NQwenBaseSZeroMOnePTwoC{} \\
    \vanillakd & 0 & \NQwenVanillaSZeroMicroF{} & \NQwenVanillaSZeroMicroP{} & \NQwenVanillaSZeroMicroR{} & \NQwenVanillaSZeroValid{} & \NQwenVanillaSZeroHitCap{} & \NQwenVanillaSZeroGenTokens{} & \NQwenVanillaSZeroLAS{} & \NQwenVanillaSZeroMOnePOne{} & \NQwenVanillaSZeroMOnePTwo{} & \NQwenVanillaSZeroMOnePTwoC{} \\
     & 1 & \NQwenVanillaSOneMicroF{} & \NQwenVanillaSOneMicroP{} & \NQwenVanillaSOneMicroR{} & \NQwenVanillaSOneValid{} & \NQwenVanillaSOneHitCap{} & \NQwenVanillaSOneGenTokens{} & \NQwenVanillaSOneLAS{} & \NQwenVanillaSOneMOnePOne{} & \NQwenVanillaSOneMOnePTwo{} & \NQwenVanillaSOneMOnePTwoC{} \\
     & 2 & \NQwenVanillaSTwoMicroF{} & \NQwenVanillaSTwoMicroP{} & \NQwenVanillaSTwoMicroR{} & \NQwenVanillaSTwoValid{} & \NQwenVanillaSTwoHitCap{} & \NQwenVanillaSTwoGenTokens{} & \NQwenVanillaSTwoLAS{} & \NQwenVanillaSTwoMOnePOne{} & \NQwenVanillaSTwoMOnePTwo{} & \NQwenVanillaSTwoMOnePTwoC{} \\
    \vanillakd{} (bf16) & 0 & \NQwenVanillaBfSZeroMicroF{} & \NQwenVanillaBfSZeroMicroP{} & \NQwenVanillaBfSZeroMicroR{} & \NQwenVanillaBfSZeroValid{} & \NQwenVanillaBfSZeroHitCap{} & \NQwenVanillaBfSZeroGenTokens{} & \NQwenVanillaBfSZeroLAS{} & \NQwenVanillaBfSZeroMOnePOne{} & \NQwenVanillaBfSZeroMOnePTwo{} & \NQwenVanillaBfSZeroMOnePTwoC{} \\
     & 1 & \NQwenVanillaBfSOneMicroF{} & \NQwenVanillaBfSOneMicroP{} & \NQwenVanillaBfSOneMicroR{} & \NQwenVanillaBfSOneValid{} & \NQwenVanillaBfSOneHitCap{} & \NQwenVanillaBfSOneGenTokens{} & \NQwenVanillaBfSOneLAS{} & \NQwenVanillaBfSOneMOnePOne{} & \NQwenVanillaBfSOneMOnePTwo{} & \NQwenVanillaBfSOneMOnePTwoC{} \\
     & 2 & \NQwenVanillaBfSTwoMicroF{} & \NQwenVanillaBfSTwoMicroP{} & \NQwenVanillaBfSTwoMicroR{} & \NQwenVanillaBfSTwoValid{} & \NQwenVanillaBfSTwoHitCap{} & \NQwenVanillaBfSTwoGenTokens{} & \NQwenVanillaBfSTwoLAS{} & \NQwenVanillaBfSTwoMOnePOne{} & \NQwenVanillaBfSTwoMOnePTwo{} & \NQwenVanillaBfSTwoMOnePTwoC{} \\
    \featkd & 0 & \NQwenFeatKDSZeroMicroF{} & \NQwenFeatKDSZeroMicroP{} & \NQwenFeatKDSZeroMicroR{} & \NQwenFeatKDSZeroValid{} & \NQwenFeatKDSZeroHitCap{} & \NQwenFeatKDSZeroGenTokens{} & \NQwenFeatKDSZeroLAS{} & \NQwenFeatKDSZeroMOnePOne{} & \NQwenFeatKDSZeroMOnePTwo{} & \NQwenFeatKDSZeroMOnePTwoC{} \\
     & 1 & \NQwenFeatKDSOneMicroF{} & \NQwenFeatKDSOneMicroP{} & \NQwenFeatKDSOneMicroR{} & \NQwenFeatKDSOneValid{} & \NQwenFeatKDSOneHitCap{} & \NQwenFeatKDSOneGenTokens{} & \NQwenFeatKDSOneLAS{} & \NQwenFeatKDSOneMOnePOne{} & \NQwenFeatKDSOneMOnePTwo{} & \NQwenFeatKDSOneMOnePTwoC{} \\
     & 2 & \NQwenFeatKDSTwoMicroF{} & \NQwenFeatKDSTwoMicroP{} & \NQwenFeatKDSTwoMicroR{} & \NQwenFeatKDSTwoValid{} & \NQwenFeatKDSTwoHitCap{} & \NQwenFeatKDSTwoGenTokens{} & \NQwenFeatKDSTwoLAS{} & \NQwenFeatKDSTwoMOnePOne{} & \NQwenFeatKDSTwoMOnePTwo{} & \NQwenFeatKDSTwoMOnePTwoC{} \\
    \distillm & 0 & \NQwenDistillmSZeroMicroF{} & \NQwenDistillmSZeroMicroP{} & \NQwenDistillmSZeroMicroR{} & \NQwenDistillmSZeroValid{} & \NQwenDistillmSZeroHitCap{} & \NQwenDistillmSZeroGenTokens{} & \NQwenDistillmSZeroLAS{} & \NQwenDistillmSZeroMOnePOne{} & \NQwenDistillmSZeroMOnePTwo{} & \NQwenDistillmSZeroMOnePTwoC{} \\
    \opd & 0 & \NQwenOPDSZeroMicroF{} & \NQwenOPDSZeroMicroP{} & \NQwenOPDSZeroMicroR{} & \NQwenOPDSZeroValid{} & \NQwenOPDSZeroHitCap{} & \NQwenOPDSZeroGenTokens{} & \NQwenOPDSZeroLAS{} & \NQwenOPDSZeroMOnePOne{} & \NQwenOPDSZeroMOnePTwo{} & \NQwenOPDSZeroMOnePTwoC{} \\
    \mistill & 0 & \NQwenMistillSZeroMicroF{} & \NQwenMistillSZeroMicroP{} & \NQwenMistillSZeroMicroR{} & \NQwenMistillSZeroValid{} & \NQwenMistillSZeroHitCap{} & \NQwenMistillSZeroGenTokens{} & \NQwenMistillSZeroLAS{} & \NQwenMistillSZeroMOnePOne{} & \NQwenMistillSZeroMOnePTwo{} & \NQwenMistillSZeroMOnePTwoC{} \\
     & 1 & \NQwenMistillSOneMicroF{} & \NQwenMistillSOneMicroP{} & \NQwenMistillSOneMicroR{} & \NQwenMistillSOneValid{} & \NQwenMistillSOneHitCap{} & \NQwenMistillSOneGenTokens{} & \NQwenMistillSOneLAS{} & \NQwenMistillSOneMOnePOne{} & \NQwenMistillSOneMOnePTwo{} & \NQwenMistillSOneMOnePTwoC{} \\
     & 2 & \NQwenMistillSTwoMicroF{} & \NQwenMistillSTwoMicroP{} & \NQwenMistillSTwoMicroR{} & \NQwenMistillSTwoValid{} & \NQwenMistillSTwoHitCap{} & \NQwenMistillSTwoGenTokens{} & \NQwenMistillSTwoLAS{} & \NQwenMistillSTwoMOnePOne{} & \NQwenMistillSTwoMOnePTwo{} & \NQwenMistillSTwoMOnePTwoC{} \\
    \mistill{} (fp32, $\lambda$ 0.1) & 0 & \NQwenMistillFpSZeroMicroF{} & \NQwenMistillFpSZeroMicroP{} & \NQwenMistillFpSZeroMicroR{} & \NQwenMistillFpSZeroValid{} & \NQwenMistillFpSZeroHitCap{} & \NQwenMistillFpSZeroGenTokens{} & \NQwenMistillFpSZeroLAS{} & \NQwenMistillFpSZeroMOnePOne{} & \NQwenMistillFpSZeroMOnePTwo{} & \NQwenMistillFpSZeroMOnePTwoC{} \\
     & 1 & \NQwenMistillFpSOneMicroF{} & \NQwenMistillFpSOneMicroP{} & \NQwenMistillFpSOneMicroR{} & \NQwenMistillFpSOneValid{} & \NQwenMistillFpSOneHitCap{} & \NQwenMistillFpSOneGenTokens{} & \NQwenMistillFpSOneLAS{} & \NQwenMistillFpSOneMOnePOne{} & \NQwenMistillFpSOneMOnePTwo{} & \NQwenMistillFpSOneMOnePTwoC{} \\
     & 2 & \NQwenMistillFpSTwoMicroF{} & \NQwenMistillFpSTwoMicroP{} & \NQwenMistillFpSTwoMicroR{} & \NQwenMistillFpSTwoValid{} & \NQwenMistillFpSTwoHitCap{} & \NQwenMistillFpSTwoGenTokens{} & \NQwenMistillFpSTwoLAS{} & \NQwenMistillFpSTwoMOnePOne{} & \NQwenMistillFpSTwoMOnePTwo{} & \NQwenMistillFpSTwoMOnePTwoC{} \\
    \mistill{} (fp32, $\lambda$ 0.01) & 0 & \NQwenMistillFpSelSZeroMicroF{} & \NQwenMistillFpSelSZeroMicroP{} & \NQwenMistillFpSelSZeroMicroR{} & \NQwenMistillFpSelSZeroValid{} & \NQwenMistillFpSelSZeroHitCap{} & \NQwenMistillFpSelSZeroGenTokens{} & \NQwenMistillFpSelSZeroLAS{} & \NQwenMistillFpSelSZeroMOnePOne{} & \NQwenMistillFpSelSZeroMOnePTwo{} & \NQwenMistillFpSelSZeroMOnePTwoC{} \\
    \midrule
    \multicolumn{12}{@{}l}{\smol} \\
    Base & --- & \NSmolBaseSZeroMicroF{} & \NSmolBaseSZeroMicroP{} & \NSmolBaseSZeroMicroR{} & \NSmolBaseSZeroValid{} & \NSmolBaseSZeroHitCap{} & \NSmolBaseSZeroGenTokens{} & \NSmolBaseSZeroLAS{} & \NSmolBaseSZeroMOnePOne{} & \NSmolBaseSZeroMOnePTwo{} & \NSmolBaseSZeroMOnePTwoC{} \\
    \vanillakd & 0 & \NSmolVanillaSZeroMicroF{} & \NSmolVanillaSZeroMicroP{} & \NSmolVanillaSZeroMicroR{} & \NSmolVanillaSZeroValid{} & \NSmolVanillaSZeroHitCap{} & \NSmolVanillaSZeroGenTokens{} & \NSmolVanillaSZeroLAS{} & \NSmolVanillaSZeroMOnePOne{} & \NSmolVanillaSZeroMOnePTwo{} & \NSmolVanillaSZeroMOnePTwoC{} \\
     & 1 & \NSmolVanillaSOneMicroF{} & \NSmolVanillaSOneMicroP{} & \NSmolVanillaSOneMicroR{} & \NSmolVanillaSOneValid{} & \NSmolVanillaSOneHitCap{} & \NSmolVanillaSOneGenTokens{} & \NSmolVanillaSOneLAS{} & \NSmolVanillaSOneMOnePOne{} & \NSmolVanillaSOneMOnePTwo{} & \NSmolVanillaSOneMOnePTwoC{} \\
    \vanillakd{} (bf16) & 0 & \NSmolVanillaBfSZeroMicroF{} & \NSmolVanillaBfSZeroMicroP{} & \NSmolVanillaBfSZeroMicroR{} & \NSmolVanillaBfSZeroValid{} & \NSmolVanillaBfSZeroHitCap{} & \NSmolVanillaBfSZeroGenTokens{} & \NSmolVanillaBfSZeroLAS{} & \NSmolVanillaBfSZeroMOnePOne{} & \NSmolVanillaBfSZeroMOnePTwo{} & \NSmolVanillaBfSZeroMOnePTwoC{} \\
    \featkd & 0 & \NSmolFeatKDSZeroMicroF{} & \NSmolFeatKDSZeroMicroP{} & \NSmolFeatKDSZeroMicroR{} & \NSmolFeatKDSZeroValid{} & \NSmolFeatKDSZeroHitCap{} & \NSmolFeatKDSZeroGenTokens{} & \NSmolFeatKDSZeroLAS{} & \NSmolFeatKDSZeroMOnePOne{} & \NSmolFeatKDSZeroMOnePTwo{} & \NSmolFeatKDSZeroMOnePTwoC{} \\
    \gold & 0 & \NSmolGOLDSZeroMicroF{} & \NSmolGOLDSZeroMicroP{} & \NSmolGOLDSZeroMicroR{} & \NSmolGOLDSZeroValid{} & \NSmolGOLDSZeroHitCap{} & \NSmolGOLDSZeroGenTokens{} & \NSmolGOLDSZeroLAS{} & \NSmolGOLDSZeroMOnePOne{} & \NSmolGOLDSZeroMOnePTwo{} & \NSmolGOLDSZeroMOnePTwoC{} \\
    \mistill & 0 & \NSmolMistillSZeroMicroF{} & \NSmolMistillSZeroMicroP{} & \NSmolMistillSZeroMicroR{} & \NSmolMistillSZeroValid{} & \NSmolMistillSZeroHitCap{} & \NSmolMistillSZeroGenTokens{} & \NSmolMistillSZeroLAS{} & \NSmolMistillSZeroMOnePOne{} & \NSmolMistillSZeroMOnePTwo{} & \NSmolMistillSZeroMOnePTwoC{} \\
     & 1 & \NSmolMistillSOneMicroF{} & \NSmolMistillSOneMicroP{} & \NSmolMistillSOneMicroR{} & \NSmolMistillSOneValid{} & \NSmolMistillSOneHitCap{} & \NSmolMistillSOneGenTokens{} & \NSmolMistillSOneLAS{} & \NSmolMistillSOneMOnePOne{} & \NSmolMistillSOneMOnePTwo{} & \NSmolMistillSOneMOnePTwoC{} \\
     & 2 & \NSmolMistillSTwoMicroF{} & \NSmolMistillSTwoMicroP{} & \NSmolMistillSTwoMicroR{} & \NSmolMistillSTwoValid{} & \NSmolMistillSTwoHitCap{} & \NSmolMistillSTwoGenTokens{} & \NSmolMistillSTwoLAS{} & \NSmolMistillSTwoMOnePOne{} & \NSmolMistillSTwoMOnePTwo{} & \NSmolMistillSTwoMOnePTwoC{} \\
    \bottomrule
  \end{tabular}
  \end{adjustbox}
\end{table}

\subsection{Two dissociations: \opd{} and the \gold{} null}
\label{app:res-dissoc}
\paragraph{\opd{} matches fp32 \vanillakd{} on the answer and separates from it wherever the teacher is the reference.} Its
\microf{} is \NQwenOPDMicroF{} (against fp32 \vanillakd{} \NQwenOPDMicroFVsVanillaDelta{}
[\NQwenOPDMicroFVsVanillaLo, \NQwenOPDMicroFVsVanillaHi], $p$ \NQwenOPDMicroFVsVanillaP), but its \las{} is lower
(\NQwenOPDLAS, \NQwenOPDLASVsVanillaDelta{} [\NQwenOPDLASVsVanillaLo, \NQwenOPDLASVsVanillaHi], $p$
\NQwenOPDLASVsVanillaP), its teacher-trace \ppl{} higher (\NQwenOPDPPL{} against \NQwenVanillaPPL) and its M1 lower at
\Pone{} and \Ptwoc{} (\NQwenOPDMOnePOne{} and \NQwenOPDMOnePTwoC, against \NQwenVanillaMOnePOne{} and
\NQwenVanillaMOnePTwoC). Each of these is measured against the teacher, and each says the student has moved off the
teacher's trace distribution, which is what training on its own rollouts is for. Its \bleu{} is the table's highest
(\NQwenOPDBLEU{} against \NQwenVanillaBLEU), while its \las{} is significantly lower: the traces are more teacher-like
token for token and carry less of the decision. A comparison on surface fidelity alone would rank \opd{} first; \las{},
which asks whether a frozen simulator recovers the answer from the rationale, ranks it below fp32 \vanillakd{}. The
comparison mixes precisions, since \opd{} trains bf16 weights: against bf16 \vanillakd{} \opd{} is higher on \bleu{} and
on \las{} alike (\NQwenOPDBLEUVsVanillaBfDelta{} [\NQwenOPDBLEUVsVanillaBfLo, \NQwenOPDBLEUVsVanillaBfHi] and
\NQwenOPDLASVsVanillaBfDelta{} [\NQwenOPDLASVsVanillaBfLo, \NQwenOPDLASVsVanillaBfHi]; Table~\ref{tab:app-paired}), and its
M1 is level with bf16 \vanillakd's (\NQwenVanillaBfMOnePOne{} and \NQwenVanillaBfMOnePTwoC{} at \Pone{} and \Ptwoc), so
the dissociation holds against the fp32 reference only. It is still why \las{} is kept in the metric set: a trace can
be closer to the teacher's token for token and carry less of its decision. The compensation is validity: \opd{} is the most often valid distilled arm
(\NQwenOPDValid, cap \NQwenOPDHitCap), because the loops that cost the off-policy arms a sixth to a quarter of their
answers are, for an on-policy objective, training data. Its tail \macrof{} is nominally the table's best
(\NQwenOPDTailF{} against \NQwenVanillaTailF), at one seed; no weight is put on it.

\paragraph{\gold{} scores zero, and the mechanism metrics locate why.} \gold{} (bf16 weights, like \opd) scores \microf{} \NSmolGOLDMicroF: no
note yields a single scored code. The policy is not degenerate: \NSmolGOLDValid{} of its generations close a
well-formed answer block, the reasoning is fluent and on topic and names correct billable codes in prose, and its
teacher-trace \ppl{} (\NSmolGOLDPPL, the table's highest) and \bleu{} (\NSmolGOLDBLEU) are finite. Its final answers
list three-character ICD-10-CM categories (N18, say, where a billable code such as N18.9 is required), and every such
entry fails inventory validation. M1 places the failure at Phase~2. Against the untrained \smol{} (\Pone{}
\NSmolBaseMOnePOne, \Ptwo{} \NSmolBaseMOnePTwo, \Ptwoc{} \NSmolBaseMOnePTwoC), \gold{} improves the chapter decision
(\NSmolGOLDMOnePOne) and leaves the contrastive one near the untrained level (\NSmolGOLDMOnePTwoC), but its \Ptwo{}
agreement falls below the untrained student's, to \NSmolGOLDMOnePTwo{} with a negative median margin
(\NMDGoldMOnePTwoMedianMargin{} nats), where \vanillakd{} seed 0 on the same student reaches
\NSmolVanillaSZeroMOnePTwo{} (median \NMDSmolVanillaSZeroMOnePTwoMedianMargin{} nats). \Ptwo{} events are exactly the
choices of a billable code within a category, so the pre-registered phase-2 readout reports the failure that the
category-level answers make visible: task and mechanism fail together and at the same phase, the strongest internal
evidence available that the null is real rather than a scoring artefact. The cause is the run's cap: the teacher's
median trace is about \NMDTeacherTraceMedianTokens{} tokens with a 90th percentile near \NMDTeacherTraceNinetiethTokens, so truncating the teacher's completion to 6{,}144
tokens removed the end of the longer traces, where the answer block and its full-code emissions are, on \NMDGoldTruncatedTraces{} of the 500
training traces, and the arm learned the truncated behaviour. Its training losses stayed healthy to the last step, and
the loss-based collapse monitor that had caught the fully on-policy run reported health throughout; a generation-level
probe on a few notes during training is what would have caught it (it was added to the trainer afterwards and never
fired on a \mistill-family run). The cross-family on-policy cell is therefore unresolved rather than refuted: what is
measured is \gold{} under a 6{,}144-token rollout cap, and the row carries that cap wherever it is quoted. A clean rerun
needs more memory than a 141\,GB GPU offers at the teacher's trace lengths, or memory work in the hybrid loss; it was not
attempted. The same-family on-policy cell (\opd, no rollout ever capped) carries the on-policy comparison.

\subsection{The precision control}
\label{app:res-precision}
\mistill, its controls and the on-policy baselines (\opd, \gold) train bf16 weights without an fp32 master copy;
\vanillakd, \featkd{} and \distillm{} keep fp32 master weights (Appendix~\ref{app:base-matched}). The control trains \vanillakd{} with bf16 weights on the fp32 \vanillakd{} recipe
(493 notes, two epochs, thinking mode), three \qwenfour{} seeds (top-32 KL, YaRN $\times2$) and one \smol{} seed (ULD),
so that each control's dumped configuration differs from \mistill's in the objective only. Each is evaluated on the 150
test notes and read on its own heads with the instruments of Appendix~\ref{app:metrics}. The reading was fixed before
the result: bf16 \vanillakd{} at \mistill's task level means the tier-1 task gap to the fp32 baselines is the
precision's; at fp32 \vanillakd's level, the objective's; between, both.

\begin{table}[!htbp]
  \centering\small
  \caption{Precision control, task side (150 test notes; mean $\pm$ sd over seeds). $\Delta$: paired \microf{}
  difference, arm minus the reference's seed-averaged notes, with its 95\% bootstrap interval. \resid{R-D17}}
  \label{tab:app-precision}
  \scriptsize\setlength{\tabcolsep}{3pt}
  \begin{adjustbox}{max width=\linewidth}
  \begin{tabular}{@{}l l c c c c c c c@{}}
    \toprule
    Arm & weights & seeds & \microf & \las & valid & M1 \Ptwoc & $\Delta$ vs \mistill & $\Delta$ vs fp32 \vanillakd \\
    \midrule
    \multicolumn{9}{@{}l}{\qwenfour} \\
    \vanillakd & fp32 & 3 & \pmsd{\NQwenVanillaMicroF}{\NQwenVanillaMicroFSd} & \pmsd{\NQwenVanillaLAS}{\NQwenVanillaLASSd} & \NQwenVanillaValid{} & \NQwenVanillaMOnePTwoC{} & \pci{\NQwenVanillaMicroFVsMistillDelta}{\NQwenVanillaMicroFVsMistillLo}{\NQwenVanillaMicroFVsMistillHi} & --- \\
    \vanillakd & bf16 & 3 & \pmsd{\NQwenVanillaBfMicroF}{\NQwenVanillaBfMicroFSd} & \pmsd{\NQwenVanillaBfLAS}{\NQwenVanillaBfLASSd} & \NQwenVanillaBfValid{} & \NQwenVanillaBfMOnePTwoC{} & \pci{\NQwenVanillaBfMicroFVsMistillDelta}{\NQwenVanillaBfMicroFVsMistillLo}{\NQwenVanillaBfMicroFVsMistillHi} & \pci{\NQwenVanillaBfMicroFVsVanillaDelta}{\NQwenVanillaBfMicroFVsVanillaLo}{\NQwenVanillaBfMicroFVsVanillaHi} \\
    \mistill & bf16 & 3 & \pmsd{\NQwenMistillMicroF}{\NQwenMistillMicroFSd} & \pmsd{\NQwenMistillLAS}{\NQwenMistillLASSd} & \NQwenMistillValid{} & \NQwenMistillMOnePTwoC{} & --- & \pci{\NQwenMistillMicroFVsVanillaDelta}{\NQwenMistillMicroFVsVanillaLo}{\NQwenMistillMicroFVsVanillaHi} \\
    \mistill & fp32 & \NQwenMistillFpSeeds{} & \pmsd{\NQwenMistillFpMicroF}{\NQwenMistillFpMicroFSd} & \pmsd{\NQwenMistillFpLAS}{\NQwenMistillFpLASSd} & \NQwenMistillFpValid{} & \NQwenMistillFpMOnePTwoC{} & \pci{\NQwenMistillFpMicroFVsMistillDelta}{\NQwenMistillFpMicroFVsMistillLo}{\NQwenMistillFpMicroFVsMistillHi} & \pci{\NQwenMistillFpMicroFVsVanillaDelta}{\NQwenMistillFpMicroFVsVanillaLo}{\NQwenMistillFpMicroFVsVanillaHi} \\
    \midrule
    \multicolumn{9}{@{}l}{\smol} \\
    \vanillakd & fp32 & 2 & \pmsd{\NSmolVanillaMicroF}{\NSmolVanillaMicroFSd} & \pmsd{\NSmolVanillaLAS}{\NSmolVanillaLASSd} & \NSmolVanillaValid{} & \NSmolVanillaMOnePTwoC{} & \pci{\NSmolVanillaMicroFVsMistillDelta}{\NSmolVanillaMicroFVsMistillLo}{\NSmolVanillaMicroFVsMistillHi} & --- \\
    \vanillakd & bf16 & 1 & \NSmolVanillaBfMicroF{} & \NSmolVanillaBfLAS{} & \NSmolVanillaBfValid{} & \NSmolVanillaBfMOnePTwoC{} & \pci{\NSmolVanillaBfMicroFVsMistillDelta}{\NSmolVanillaBfMicroFVsMistillLo}{\NSmolVanillaBfMicroFVsMistillHi} & \pci{\NSmolVanillaBfMicroFVsVanillaDelta}{\NSmolVanillaBfMicroFVsVanillaLo}{\NSmolVanillaBfMicroFVsVanillaHi} \\
    \mistill & bf16 & 2 & \pmsd{\NSmolMistillMicroF}{\NSmolMistillMicroFSd} & \pmsd{\NSmolMistillLAS}{\NSmolMistillLASSd} & \NSmolMistillValid{} & \NSmolMistillMOnePTwoC{} & --- & \pci{\NSmolMistillMicroFVsVanillaDelta}{\NSmolMistillMicroFVsVanillaLo}{\NSmolMistillMicroFVsVanillaHi} \\
    \bottomrule
  \end{tabular}
  \end{adjustbox}
\end{table}

\begin{table}[!htbp]
  \centering\small
  \caption{Training cross-entropy on the trace tokens at optimiser steps 5, 20, 60 and 120 of 124, per seed. bf16
  \vanillakd{} tracks \mistill{} step for step, and fp32 \vanillakd{} falls below both from about step 20. \resid{R-D17}}
  \label{tab:app-ce}
  \begin{adjustbox}{max width=\linewidth}
  \begin{tabular}{@{}l l l cccc@{}}
    \toprule
    Student & Run (seeds) & weights & step 5 & step 20 & step 60 & step 120 \\
    \midrule
    \qwenfour & \vanillakd{} (0\,/\,1\,/\,2) & fp32 & \NMDCeQwenVanillaFpSZeroStepFive\,/\,\NMDCeQwenVanillaFpSOneStepFive\,/\,\NMDCeQwenVanillaFpSTwoStepFive{} & \NMDCeQwenVanillaFpSZeroStepTwenty\,/\,\NMDCeQwenVanillaFpSOneStepTwenty\,/\,\NMDCeQwenVanillaFpSTwoStepTwenty{} & \NMDCeQwenVanillaFpSZeroStepSixty\,/\,\NMDCeQwenVanillaFpSOneStepSixty\,/\,\NMDCeQwenVanillaFpSTwoStepSixty{} & \NMDCeQwenVanillaFpSZeroStepOneHundredTwenty\,/\,\NMDCeQwenVanillaFpSOneStepOneHundredTwenty\,/\,\NMDCeQwenVanillaFpSTwoStepOneHundredTwenty{} \\
     & \vanillakd{} (0\,/\,1\,/\,2) & bf16 & \NMDCeQwenVanillaBfSZeroStepFive\,/\,\NMDCeQwenVanillaBfSOneStepFive\,/\,\NMDCeQwenVanillaBfSTwoStepFive{} & \NMDCeQwenVanillaBfSZeroStepTwenty\,/\,\NMDCeQwenVanillaBfSOneStepTwenty\,/\,\NMDCeQwenVanillaBfSTwoStepTwenty{} & \NMDCeQwenVanillaBfSZeroStepSixty\,/\,\NMDCeQwenVanillaBfSOneStepSixty\,/\,\NMDCeQwenVanillaBfSTwoStepSixty{} & \NMDCeQwenVanillaBfSZeroStepOneHundredTwenty\,/\,\NMDCeQwenVanillaBfSOneStepOneHundredTwenty\,/\,\NMDCeQwenVanillaBfSTwoStepOneHundredTwenty{} \\
     & \mistill{} (0) & bf16 & \NMDCeQwenMistillSZeroStepFive{} & \NMDCeQwenMistillSZeroStepTwenty{} & \NMDCeQwenMistillSZeroStepSixty{} & \NMDCeQwenMistillSZeroStepOneHundredTwenty{} \\
    \midrule
    \smol & \vanillakd{} (0\,/\,1) & fp32 & \NMDCeSmolVanillaFpSZeroStepFive\,/\,\NMDCeSmolVanillaFpSOneStepFive{} & \NMDCeSmolVanillaFpSZeroStepTwenty\,/\,\NMDCeSmolVanillaFpSOneStepTwenty{} & \NMDCeSmolVanillaFpSZeroStepSixty\,/\,\NMDCeSmolVanillaFpSOneStepSixty{} & \NMDCeSmolVanillaFpSZeroStepOneHundredTwenty\,/\,\NMDCeSmolVanillaFpSOneStepOneHundredTwenty{} \\
     & \vanillakd{} (0) & bf16 & \NMDCeSmolVanillaBfSZeroStepFive{} & \NMDCeSmolVanillaBfSZeroStepTwenty{} & \NMDCeSmolVanillaBfSZeroStepSixty{} & \NMDCeSmolVanillaBfSZeroStepOneHundredTwenty{} \\
     & \mistill{} (0) & bf16 & \NMDCeSmolMistillSZeroStepFive{} & \NMDCeSmolMistillSZeroStepTwenty{} & \NMDCeSmolMistillSZeroStepSixty{} & \NMDCeSmolMistillSZeroStepOneHundredTwenty{} \\
    \bottomrule
  \end{tabular}
  \end{adjustbox}
\end{table}

\paragraph{The task: the precision's, on both students.} bf16 \vanillakd{} lands at \mistill's task level, not fp32
\vanillakd's (Table~\ref{tab:app-precision}): its three \qwenfour{} seeds (\NQwenVanillaBfSZeroMicroF,
\NQwenVanillaBfSOneMicroF, \NQwenVanillaBfSTwoMicroF) all sit far below fp32 \vanillakd{} and around \mistill. Paired
per note and pooled over the three \qwenfour{} seed pairs (note-pairs counted as if independent, descriptive), \mistill{} minus bf16 \vanillakd{} splits the task
\NCurvePairFullMistillVsVanillaBfTaskHigher{} higher and \NCurvePairFullMistillVsVanillaBfTaskLower{} lower
(\NMDPoolQwenMistillVsVanillaBfTaskTies{} ties; $p$ \NCurvePairFullMistillVsVanillaBfTaskP), while M1 at \Ptwoc{} favours
\mistill{} on \NCurvePairFullMistillVsVanillaBfMOnePTwoCHigher{} note-pairs against
\NCurvePairFullMistillVsVanillaBfMOnePTwoCLower{} ($p$ \NCurvePairFullMistillVsVanillaBfMOnePTwoCP, mean
\NCurvePairFullMistillVsVanillaBfMOnePTwoCMean): the supervised readout's lead holds, and is larger, at matched
precision. At \Pone{} the pooled comparison leans the other way (\NMDPoolQwenMistillVsVanillaBfMOnePOneHigher{} higher, \NMDPoolQwenMistillVsVanillaBfMOnePOneLower{} lower, mean \NMDPoolQwenMistillVsVanillaBfMOnePOneMean), and at \Ptwo{} it is not
significant. On \smol, one seed pair, the task splits \NMDPairSmolMistillVsVanillaBfTaskHigher{} higher and \NMDPairSmolMistillVsVanillaBfTaskLower{} lower ($p$ \NMDPairSmolMistillVsVanillaBfTaskP) and M1 at \Ptwoc{}
\NMDPairSmolMistillVsVanillaBfMOnePTwoCHigher{} higher and \NMDPairSmolMistillVsVanillaBfMOnePTwoCLower{} lower (mean \NMDPairSmolMistillVsVanillaBfMOnePTwoCMean): as against fp32 \vanillakd, \mistill's advantage on \smol{} is not in M1; on its own heads it holds, and grows
(below). The training loss agrees (Table~\ref{tab:app-ce}). What bf16 training costs against fp32 master weights is
\NQwenVanillaBfMicroFVsVanillaDelta{} \microf{} on \qwenfour{} and \NSmolVanillaBfMicroFVsVanillaDelta{} on \smol{} for
\vanillakd, about the size of \mistill's gap to fp32 \vanillakd. At the $\lambda$ selected in bf16, \mistill{} does not
share that gain: trained with fp32 master weights it moves \NQwenMistillFpMicroFVsMistillDelta{}
[\NQwenMistillFpMicroFVsMistillLo, \NQwenMistillFpMicroFVsMistillHi] from its bf16 runs and trails fp32 \vanillakd{} by
\NQwenMistillFpMicroFVsVanillaDelta{} [\NQwenMistillFpMicroFVsVanillaLo, \NQwenMistillFpMicroFVsVanillaHi] on the task and
\NQwenMistillFpLASVsVanillaDelta{} on \las, with M1 at \Ptwoc{} \NQwenMistillFpMOnePTwoC{} against fp32 \vanillakd's
\NQwenVanillaMOnePTwoC. That cost is the dose's, not the objective's: fp32 master weights have a different dose
response, and at the $\lambda$ the paper's own rule selects for fp32 the difference is not detected
(Appendix~\ref{app:obj-fp32}).

\paragraph{The fp32 arms' own heads.} The three fp32 \mistill{} arms at the transferred $\lambda$ were read on their own
heads like every other arm (Table~\ref{tab:app-mech-seeds}). Their own top-40 recover
\NOwnQwenMistillFpSZeroPTwoCKiForty, \NOwnQwenMistillFpSOnePTwoCKiForty{} and \NOwnQwenMistillFpSTwoPTwoCKiForty{} of the
contrastive decision against a random floor of \NOwnQwenMistillFpSZeroPTwoCKiFortyRand, but the gain over \vanillakd{}
seen in bf16 is not detectable here: note-clustered, \NPoolQwenMistillFpVsVanillaPTwoCClusterDelta{} against fp32
\vanillakd{} ($p$ \NPoolQwenMistillFpVsVanillaPTwoCClusterP) and \NPoolQwenMistillFpVsVanillaBfPTwoCClusterDelta{} against
bf16 \vanillakd{} ($p$ \NPoolQwenMistillFpVsVanillaBfPTwoCClusterP), \NPoolQwenMistillFpVsMistillPTwoCClusterDelta{} below the
bf16 \mistill{} arms ($p$ \NPoolQwenMistillFpVsMistillPTwoCClusterP). At $k=8$ their own top-8 recover
\NOwnQwenMistillFpSZeroPTwoCKiEight{} against about half for every other arm
(\NPoolQwenMistillFpVsVanillaBfPTwoCKiEightClusterDelta{} per note), and on the read side their contrastive heads' lift is
below every reference (\NPoolQwenMistillFpVsVanillaBfMTwoPTwoCClusterDelta{} against bf16 \vanillakd; a dimensionless
shortlist-region lift) while their
chapter heads' lift is above every reference (\NPoolQwenMistillFpVsVanillaBfMTwoPOneClusterDelta): at this dose the fp32
arms' specialization sits in the chapter phase. Both findings are the dose's. At the $\lambda$ selected for fp32 the
$k=40$ gain over fp32 \vanillakd{} returns (Appendix~\ref{app:obj-fp32}), and most of the $k=8$ deficit is one causally
active head placed at rank \NSweepLPointOneRankLThirtyFiveHTwentySix{} by these arms' first-order ranking (rank
\NSweepMistillSZeroRankLThirtyFiveHTwentySix{} in every bf16 arm); moved into the top-8 of seed 0's own ranking it raises
the cell to \NSweepLPointOneProbeKiEight.

\paragraph{The mechanism side.} On the students' own heads (Table~\ref{tab:mech}), \mistill's gain over bf16 \vanillakd{}
at \Ptwoc{} is \NPoolQwenMistillVsVanillaBfPTwoCClusterDelta{} on \qwenfour{} (note-clustered over three seed pairs: higher on
\NPoolQwenMistillVsVanillaBfPTwoCClusterHigher{} of 150 notes, lower on \NPoolQwenMistillVsVanillaBfPTwoCClusterLower), against
\NPoolQwenMistillVsVanillaPTwoCClusterDelta{} over fp32 \vanillakd, and \NPoolSmolMistillVsVanillaBfPTwoCClusterDelta{} on \smol{}
(three seed pairs; higher on \NPoolSmolMistillVsVanillaBfPTwoCClusterHigher{} of 150), against
\NPoolSmolMistillVsVanillaPTwoCClusterDelta: at matched precision the contrastive gain is larger. At \Ptwo{} on \qwenfour{}
it reverses (\NPoolQwenMistillVsVanillaBfPTwoClusterDelta{} per note, lower on \NPoolQwenMistillVsVanillaBfPTwoClusterLower{} of
\NPoolQwenMistillVsVanillaBfPTwoClusterN{} notes, note-clustered). bf16 \vanillakd's three seeds recover \NOwnQwenVanillaBfSZeroPTwoCKiForty,
\NOwnQwenVanillaBfSOnePTwoCKiForty{} and \NOwnQwenVanillaBfSTwoPTwoCKiForty{} at \Ptwoc{} (spread
\NOwnQwenVanillaBfSeedSpreadPTwoC, against \mistill's \NOwnQwenMistillSeedSpreadPTwoC{} and fp32 \vanillakd's
\NOwnQwenVanillaSeedSpreadPTwoC), and two of their three pairs separate per note where none of \mistill's do.

\subsection{On the students' own traces}
\label{app:res-owntrace}
Every mechanism readout above is teacher-forced on the teacher's traces. The question here is whether the
mechanism shows when the student reasons on its own, with its own events and its own near-misses. For each arm we built a
run from its own test-note generations (the traces whose reasoning block closes: \NOwnTraceQwenVanillaSOneNotes{} to
\NOwnTraceQwenBaseNotes{} of 150 for the \qwenfour{} arms, \NOwnTraceSmolMistillSTwoNotes{} to
\NOwnTraceSmolVanillaSOneNotes{} for \smol, whose generations often run to the cap; the count is the notes with a measurable
own-trace recovery, the completed traces being quoted beside it where they differ), extracted the events with the student's
tokenizer, and ran the knock-in with the arm's own global sets, the donor being the next own trace
(Table~\ref{tab:app-owntrace}). Every trained \qwenfour{} arm's top-40 recovery on its own traces stays close to its teacher-forced value on
the same notes (\mistill{} seed 0: \NOwnTraceQwenMistillSZeroKiForty{} against \NOwnTraceQwenMistillSZeroKiFortyTF; the largest
change at $k=40$ is \NOwnTraceQwenVanillaSOneKiFortyDelta{} per note, fp32 \vanillakd{} seed 1); the $k=8$ readout moves more (\mistill{} seed 1
\NOwnTraceQwenMistillSOneKiEightDelta, the fp32 arm's anomalous cell included); the untrained student (\NOwnTraceQwenBaseKiFortyDelta{} at $k=40$) and the \smol{} arms (\mistill{}
\NOwnTraceSmolMistillSZeroKiFortyDelta{} and \NOwnTraceSmolMistillSOneKiFortyDelta{} on \NOwnTraceSmolMistillSZeroNotes{} and
\NOwnTraceSmolMistillSOneNotes{} of 150 notes; fp32 \vanillakd{} \NOwnTraceSmolVanillaSOneKiFortyDelta{} on
\NOwnTraceSmolVanillaSOneNotes) recover more on their own, shorter traces. The fp32 $\lambda=0.01$ arm's three seeds recover
\NOwnTraceQwenMistillFpSelSZeroKiForty{} / \NOwnTraceQwenMistillFpSelSOneKiForty{} / \NOwnTraceQwenMistillFpSelSTwoKiForty{} on
their own traces against \NOwnTraceQwenMistillFpSelSZeroKiFortyTF{} / \NOwnTraceQwenMistillFpSelSOneKiFortyTF{} /
\NOwnTraceQwenMistillFpSelSTwoKiFortyTF{} teacher-forced on the same notes (\NOwnTraceQwenMistillFpSelSZeroNotes{} /
\NOwnTraceQwenMistillFpSelSOneNotes{} / \NOwnTraceQwenMistillFpSelSTwoNotes{} of 150; R-D30). The gap between the objectives persists on the students'
own reasoning: \mistill{} minus bf16 \vanillakd, on the notes both arms have events for and note-clustered over three seed
pairs, is \NOwnTraceQwenMistillVsVanillaBfPooledDelta{} at $k=40$ [\NOwnTraceQwenMistillVsVanillaBfPooledLo,
\NOwnTraceQwenMistillVsVanillaBfPooledHi] (higher on \NOwnTraceQwenMistillVsVanillaBfPooledHigher{} of
\NOwnTraceQwenMistillVsVanillaBfPooledNotes{} notes) and \NOwnTraceQwenMistillVsVanillaBfKiEightPooledDelta{} at $k=8$;
against fp32 \vanillakd{} it is not detected at $k=40$ (\NOwnTraceQwenMistillVsVanillaPooledDelta{}
[\NOwnTraceQwenMistillVsVanillaPooledLo, \NOwnTraceQwenMistillVsVanillaPooledHi]) and
\NOwnTraceQwenMistillVsVanillaKiEightPooledDelta{} at $k=8$, as teacher-forced; the fp32 $\lambda=0.01$ arm minus fp32 \vanillakd{}
is \NOwnTraceQwenMistillFpSelVsVanillaPooledDelta{} [\NOwnTraceQwenMistillFpSelVsVanillaPooledLo,
\NOwnTraceQwenMistillFpSelVsVanillaPooledHi] at $k=40$ (the per-note mean of the available pair differences over
\NOwnTraceQwenMistillFpSelVsVanillaPooledNotes{} distinct notes with at least one pair --- common notes per pair
\NOwnTraceQwenMistillFpSelVsVanillaSZeroN{} / \NOwnTraceQwenMistillFpSelVsVanillaSOneN{} / \NOwnTraceQwenMistillFpSelVsVanillaSTwoN{} ---
the direction of its teacher-forced result, conditional on completion and measurable events) and
\NOwnTraceQwenMistillFpSelVsVanillaKiEightPooledDelta{} at $k=8$; and the \smol{} recovery advantage persists on the completed traces
(\NOwnTraceSmolMistillSZeroKiForty{} and \NOwnTraceSmolMistillSOneKiForty{} against \NOwnTraceSmolVanillaBfSZeroKiForty{} and
\NOwnTraceSmolVanillaSOneKiForty{} for \vanillakd, on \NOwnTraceSmolMistillSTwoNotes--\NOwnTraceSmolVanillaSOneNotes{} notes; the paired
common-note comparisons rest on the notes both arms of a pair have events for). \smol{} \mistill{} seed 2 has
\NOwnTraceSmolMistillSTwoComplete{} completed traces out of 150, of which \NOwnTraceSmolMistillSTwoNotes{} support the \Ptwoc{}
recovery measurement; on those notes mean own-trace $k=40$ recovery is \NOwnTraceSmolMistillSTwoKiForty{} against
\NOwnTraceSmolMistillSTwoKiFortyTF{} under teacher forcing on the same notes (paired mean change \NOwnTraceSmolMistillSTwoKiFortyDelta,
higher on \NOwnTraceSmolMistillSTwoKiFortyHigher{} of \NOwnTraceSmolMistillSTwoKiFortyN; at $k=8$
\NOwnTraceSmolMistillSTwoKiEightDelta). This descriptive within-arm comparison has no seed-matched \vanillakd{} counterpart and is
conditional on completion and measurable events (R-D36). The
distilled student's own traces also contain more contrastive events than the \vanillakd{} student's at the same length,
so the comparison is read on the notes both cover. For the \qwenfour{} arms the teacher-forced readout is a close proxy for
the mechanism in the student's own reasoning; for \smol{} the completed-trace values move by arm (\mistill{} \NOwnTraceSmolMistillSZeroKiFortyDelta{} / \NOwnTraceSmolMistillSOneKiFortyDelta{} / \NOwnTraceSmolMistillSTwoKiFortyDelta, bf16 \vanillakd{} \NOwnTraceSmolVanillaBfSZeroKiFortyDelta, fp32 \vanillakd{} \NOwnTraceSmolVanillaSOneKiFortyDelta) on completed traces that are a minority of the notes
(coverage in the table); levels across arms on different completion sets are not a controlled cross-arm effect.

\begin{table}[!htbp]
  \centering\small
  \caption{Own-heads knock-in on each student's own generated test traces (events extracted from its own reasoning;
  coverage = notes with a measurable own-trace recovery, the completed-trace counts in the text where they differ), beside
  the teacher-forced cell on the same notes; means over notes of the per-note recovery, random floors in parentheses.
  Bottom: \mistill{} minus \vanillakd{} on the notes both arms have events for, per seed pair and note-clustered over the
  distinct notes with at least one available pair. \resid{R-D29}\resid{R-D30}\resid{R-D36}}
  \label{tab:app-owntrace}
  \providecommand{\gensd}[2]{\begin{tabular}[t]{@{}c@{}}#1\\[-2pt]{\scriptsize$\pm$#2}\end{tabular}}
\providecommand{\genstack}[2]{\begin{tabular}[t]{@{}c@{}}#1\\[-2pt]{\scriptsize #2}\end{tabular}}
\providecommand{\genstackl}[2]{\begin{tabular}[t]{@{}l@{}}#1\\[-2pt]{\scriptsize #2}\end{tabular}}
\providecommand{\genhead}[2]{\begin{tabular}[b]{@{}c@{}}#1\\#2\end{tabular}}
\providecommand{\genstackiii}[3]{\begin{tabular}[t]{@{}c@{}}#1\\[-2pt]{\scriptsize #2}\\[-3pt]{\scriptsize #3}\end{tabular}}
\footnotesize\setlength{\tabcolsep}{3.5pt}
\begin{adjustbox}{max width=\linewidth}
\begin{tabular}{@{}l c c c c c c@{}}
\toprule
 & & \multicolumn{3}{c}{\Ptwoc{} knock-in, own top-40} & \multicolumn{2}{c}{own top-8} \\
\cmidrule(lr){3-5}\cmidrule(lr){6-7}
Arm & \genhead{own}{notes} & \genhead{own traces}{(random)} & \genhead{teacher-forced,}{same notes} & \genhead{own $-$ TF}{(notes $\uparrow$ / $n$)} & own & TF \\
\midrule
\multicolumn{7}{@{}l}{\qwenfour} \\
\mistill{} s0 & 136 & 0.863\,{\scriptsize(0.034)} & 0.863 & +0.000 (71/136) & 0.617 & 0.607 \\
\quad s1 & 131 & 0.880\,{\scriptsize(0.033)} & 0.859 & +0.021 (76/131) & 0.662 & 0.594 \\
\quad s2 & 132 & 0.876\,{\scriptsize(0.033)} & 0.867 & +0.010 (74/132) & 0.654 & 0.602 \\
\vanillakd{} (bf16) s0 & 132 & 0.826\,{\scriptsize(0.037)} & 0.831 & \ensuremath{-}0.005 (55/132) & 0.489 & 0.478 \\
\quad s1 & 133 & 0.827\,{\scriptsize(0.036)} & 0.832 & \ensuremath{-}0.005 (67/133) & 0.459 & 0.475 \\
\quad s2 & 136 & 0.827\,{\scriptsize(0.033)} & 0.840 & \ensuremath{-}0.013 (65/136) & 0.471 & 0.480 \\
\vanillakd{} (fp32) s0 & 124 & 0.844\,{\scriptsize(0.043)} & 0.848 & \ensuremath{-}0.004 (56/124) & 0.519 & 0.518 \\
\quad s1 & 119 & 0.870\,{\scriptsize(0.045)} & 0.839 & +0.031 (64/119) & 0.512 & 0.502 \\
\quad s2 & 124 & 0.881\,{\scriptsize(0.038)} & 0.885 & \ensuremath{-}0.004 (56/124) & 0.601 & 0.601 \\
\mistill{} (fp32, $\lambda$ 0.1) s0 & 133 & 0.859\,{\scriptsize(0.026)} & 0.860 & \ensuremath{-}0.001 (65/133) & 0.179 & 0.180 \\
\mistill{} (fp32, $\lambda$ 0.01) s0 & 123 & 0.914\,{\scriptsize(0.034)} & 0.885 & +0.028 (70/123) & 0.140 & 0.133 \\
\quad s1 & 125 & 0.902\,{\scriptsize(0.030)} & 0.888 & +0.014 (68/125) & 0.583 & 0.579 \\
\quad s2 & 124 & 0.885\,{\scriptsize(0.029)} & 0.876 & +0.009 (64/124) & 0.140 & 0.130 \\
\randheads{} s0 & 131 & 0.865\,{\scriptsize(0.036)} & 0.864 & +0.000 (63/131) & 0.559 & 0.558 \\
untrained & 138 & 0.862\,{\scriptsize(0.028)} & 0.782 & +0.080 (78/138) & 0.512 & 0.497 \\
\multicolumn{7}{@{}l}{\smol} \\
\mistill{} s0 & 48 & 0.880\,{\scriptsize(0.090)} & 0.776 & +0.105 (29/48) & 0.270 & 0.237 \\
\quad s1 & 41 & 0.897\,{\scriptsize(0.078)} & 0.771 & +0.126 (27/41) & 0.249 & 0.223 \\
\quad s2 & 40 & 0.904\,{\scriptsize(0.113)} & 0.724 & +0.181 (25/40) & 0.267 & 0.235 \\
\vanillakd{} (bf16) s0 & 59 & 0.499\,{\scriptsize(0.071)} & 0.524 & \ensuremath{-}0.026 (34/59) & 0.309 & 0.219 \\
\vanillakd{} (fp32) s1 & 75 & 0.579\,{\scriptsize(0.093)} & 0.526 & +0.053 (40/75) & 0.353 & 0.351 \\
\midrule
\multicolumn{7}{@{}l}{\textit{\qwenfour{} \mistill{} $-$ \vanillakd{} on own traces, notes where both arms have events: per seed pair; note-clustered}} \\
vs bf16, $k=40$ & 3 & \multicolumn{3}{l}{+0.038 (79/122); +0.055 (78/117); +0.048 (80/119)} & \multicolumn{2}{l}{\pci{+0.041}{+0.020}{+0.062}, 102/46 of 148 notes, $p$ \ensuremath{5{\times}10^{-6}}} \\
vs bf16, $k=8$ & 3 & \multicolumn{3}{l}{+0.135 (95/122); +0.206 (99/117); +0.192 (93/119)} & \multicolumn{2}{l}{\pci{+0.166}{+0.137}{+0.192}, 129/19 of 148 notes, $p$ \ensuremath{3{\times}10^{-21}}} \\
vs fp32, $k=40$ & 3 & \multicolumn{3}{l}{+0.025 (65/116); +0.009 (59/105); \ensuremath{-}0.001 (53/107)} & \multicolumn{2}{l}{\pci{\ensuremath{-}0.001}{\ensuremath{-}0.026}{+0.023}, 75/67 of 142 notes, $p$ 0.56} \\
vs fp32, $k=8$ & 3 & \multicolumn{3}{l}{+0.095 (80/116); +0.144 (81/105); +0.064 (65/107)} & \multicolumn{2}{l}{\pci{+0.094}{+0.066}{+0.123}, 107/35 of 142 notes, $p$ \ensuremath{10^{-9}}} \\
fp32 $\lambda$ 0.01 vs fp32, $k=40$ & 3 & \multicolumn{3}{l}{+0.076 (73/105); +0.027 (59/103); +0.003 (55/104)} & \multicolumn{2}{l}{\pci{+0.029}{+0.006}{+0.051}, 87/50 of 137 notes, $p$ 0.002} \\
fp32 $\lambda$ 0.01 vs fp32, $k=8$ & 3 & \multicolumn{3}{l}{\ensuremath{-}0.370 (0/105); +0.054 (64/103); \ensuremath{-}0.452 (3/104)} & \multicolumn{2}{l}{\pci{\ensuremath{-}0.258}{\ensuremath{-}0.288}{\ensuremath{-}0.227}, 10/127 of 137 notes, $p$ \ensuremath{6{\times}10^{-27}}} \\
\bottomrule
\end{tabular}
\end{adjustbox}

\end{table}

\subsection{Extra tier-1 cells}
\label{app:res-extra}
\paragraph{A lighter Phase-1 weight.} $\lambda_{\mathrm{P1}}=0.03$ with $\lambda_{\mathrm{P2}}=0.1$ (\qwenfour, one seed)
changes nothing: \microf{} \NQwenLowPOneDoseMicroF{} (paired against \mistill's seed-averaged notes
\NQwenLowPOneDoseMicroFVsMistillDelta{} [\NQwenLowPOneDoseMicroFVsMistillLo, \NQwenLowPOneDoseMicroFVsMistillHi]) and M1
\NQwenLowPOneDoseMOnePOne{}\,/\,\NQwenLowPOneDoseMOnePTwo{}\,/\,\NQwenLowPOneDoseMOnePTwoC{} against seed 0's
\NQwenMistillSZeroMOnePOne{}\,/\,\NQwenMistillSZeroMOnePTwo{}\,/\,\NQwenMistillSZeroMOnePTwoC; the Phase-1 term has no
measurable effect at this dose. The phase ablations of Appendix~\ref{app:ablations} locate the rest.

\paragraph{A same-seed replicate on other hardware.} \mistill{} seed 2 retrained with the same configuration on a
different GPU type gives \microf{} \NMDReplicateSTwoMicroF{} against \NQwenMistillSTwoMicroF{} (paired
\NMDReplicateTaskHigher{} higher, \NMDReplicateTaskLower{} lower, $p$ \NMDReplicateTaskP) and M1 at \Ptwoc{} \NMDReplicateSTwoMOnePTwoC{} against
\NQwenMistillSTwoMOnePTwoC. The one-seed noise at tier-1 size is therefore about \NMDReplicateNoiseMicroF{}
\microf, \NMDReplicateNoiseMOnePTwoC{} M1 at \Ptwoc{} and \NMDReplicateNoiseLAS{} \las. The replicate enters
no mean.

\paragraph{\smol{} with the attention term only.} The \smol{} \mistill{} row is at bf16 \vanillakd's task level
(Appendix~\ref{app:res-precision}); before the precision control, its low task score was suspected to come from the
interaction term, whose target carries a cross-tokenizer offset (\smol's \Ptwo{} interaction loss sat near
\NMDSmolInterLossPlateau{} throughout training). The arm was retrained with the attention term only, same
epochs and $\lambda$: \microf{} \NSmolAttnTermMicroF{} (valid \NSmolAttnTermValid, cap \NSmolAttnTermHitCap), no better than
the both-terms seed 0 (\NSmolMistillSZeroMicroF; paired \NMDSmolAttnVsBothTaskHigher{} higher, \NMDSmolAttnVsBothTaskLower{} lower, $p$ \NMDSmolAttnVsBothTaskP), with M1
\NSmolAttnTermMOnePOne{}\,/\,\NSmolAttnTermMOnePTwo{}\,/\,\NSmolAttnTermMOnePTwoC{} against
\NSmolMistillSZeroMOnePOne{}\,/\,\NSmolMistillSZeroMOnePTwo{}\,/\,\NSmolMistillSZeroMOnePTwoC{} and \las{}
\NSmolAttnTermLAS{} against \NSmolMistillSZeroLAS, while its own \Ptwo{} attention loss trained normally
(from \NMDSmolAttnPTwoAttnLossStart{} to \NMDSmolAttnPTwoAttnLossEnd). The offset is real in the loss but not the cause of the row.

\subsection{Instrument checks}
\label{app:res-checks}
\paragraph{Knock-in without the attribution pass.} On the test notes the own-heads knock-in runs without its first
pass (the attribution over all heads, which the knock-in tables do not read), gated by rerunning the baseline
campaign's 20-note \vanillakd{} seed-0 cell that way. Every reported cell was reproduced, the largest difference being
\NMDNoAttrGateMaxDiff{} (\Ptwoc{} at $k=8$, the noisiest cell at 20 notes), with the random floors close. Raw
per-event margins differed by a median of \NMDNoAttrCleanMarginDiff{} nats, but so did the unpatched clean
margins of the two runs, which ran on different GPU types: that is the forward pass's own cross-hardware floor, not
the mode's. The gate's cell was not retained; its readout-level numbers are in the experiment logbook.

\paragraph{The instrument's cross-hardware floor.} The \smol{} $k=8$ own-heads cell was computed twice from one
checkpoint, on an RTX PRO 6000 (the cell of record) and on an H200. The attribution ranking's top-40 is identical at
every phase and its top-8 in the same order, and M4 differs per note by \NMDCrossHwPTwoCDelta, \NMDCrossHwPTwoDelta{} and \NMDCrossHwPOneDelta{} at
\Ptwoc, \Ptwo{} and \Pone{} (sign-test $p$ \NMDCrossHwPTwoCP, \NMDCrossHwPTwoP{} and \NMDCrossHwPOneP); in median the two reads differ by
\NMDCrossHwMedianDiff. Every per-note difference read as an effect in Appendix~\ref{app:ablations} is several
times larger.

\subsection{Readings withdrawn}
\label{app:res-withdrawn}
The baseline campaign read M4 on 20 test notes. At that size a sign test has about \NMDSignTestPowerTwentyNotes{}
power against a 0.65 win rate, and two readings did not survive the rerun on all 150. It read \smol{} \vanillakd{} at
\NMDTwentyNoteSmolVanillaPTwoC{} and \featkd{} at \NMDTwentyNoteSmolFeatKDPTwoC{} at \Ptwoc{}
(\vanillakd{} at least \featkd); at 150 notes they are \NOwnSmolVanillaSZeroPTwoCKiForty{} and
\NOwnSmolFeatKDSZeroPTwoCKiForty, and the order inverts (\featkd{} higher on \NMDSmolFeatKDVsVanillaPTwoCHigher{} of 150 notes).
And it read ``the mechanism ordering matches the task ordering on \qwenfour'' from \vanillakd{} seed 0 above
\featkd{} (\NMDTwentyNoteQwenVanillaSZeroPTwoC{} against \NMDTwentyNoteQwenFeatKDPTwoC); at 150
notes the seed-matched pair does not separate (\NOwnQwenVanillaSZeroPTwoCKiForty{} against \NOwnQwenFeatKDSZeroPTwoCKiForty; per note
\NMDQwenVanillaSZeroVsFeatKDPTwoCDelta, \NMDQwenVanillaSZeroVsFeatKDPTwoCHigher/\NMDQwenVanillaSZeroVsFeatKDPTwoCLower, $p$ \NMDQwenVanillaSZeroVsFeatKDPTwoCP), though \vanillakd's seed 2 sits above \featkd{} (\NMDQwenVanillaSTwoVsFeatKDPTwoCDelta, \NMDQwenVanillaSTwoVsFeatKDPTwoCHigher/\NMDQwenVanillaSTwoVsFeatKDPTwoCLower, $p$ \NMDQwenVanillaSTwoVsFeatKDPTwoCP).
The 20-note cells were labelled descriptive; the 150-note cells replace them, and every mechanism
number in the paper is on 150 notes.

\section{Ablation details}
\label{app:ablations}
\supports{Secs.~\ref{sec:distill-main} (training-set size) and~\ref{sec:distill-ablation}; \resid{R-D7}, \resid{R-D8}, \resid{R-D13}, \resid{R-D14}}

The controls of Sec.~\ref{sec:distill-mech} left three findings, each with a caveat they could not close: the
ranked heads need the teacher's positions (one control seed per student); where the untrained student's recovery is
low, random heads at the teacher's events yield only part of \mistill's gain (was that the heads'
ranking, or their number?); and \mistill's seeds agree where \vanillakd's do not (a property of the heads \mistill{}
supervises, or of supervising positions at all?). The ablations were built to close these caveats and were
pre-registered before any result (Table~\ref{tab:app-prereg}); a tier on \smol{} (A$'$) was added the same day, also
before its results. Every arm is one flag away from \mistill{} and trained at tier-1 size (493 notes, two epochs,
$\lambda=0.1$ unless the arm varies it, seed 0 unless stated) with bf16 weights, as \mistill{} is, so no comparison in
this appendix depends on the precision question of Appendix~\ref{app:res-precision}. Each arm is read with the
instruments of Appendix~\ref{app:metrics} on the 150 test notes: M4 knock-in at the arm's own ranked heads, M2 and the
prefix share at \Ptwoc, and the task.

\paragraph{The reading rule.} Fixed in advance: an arm whose \Ptwoc{} effect lies inside \mistill's seed band and whose
task effect lies inside the tier-1 noise band is reported as ``no effect'', not re-run, and every \Ptwoc{} comparison is
paired per note against each \mistill{} seed. As measured, \mistill's three \qwenfour{} seeds give
\NOwnQwenMistillSZeroPTwoCKiForty, \NOwnQwenMistillSOnePTwoCKiForty{} and \NOwnQwenMistillSTwoPTwoCKiForty{} in median, a
band narrower than the instrument's own spread (one checkpoint read on two GPUs differs by
\NMDCrossHwMedianDiff{} in median; Appendix~\ref{app:res-checks}). The rule is therefore applied through the paired
tests, uniformly, as an operational reading rather than a pre-registered one: a difference is read as an effect only
when it holds against every \mistill{} seed (sign test $p<0.05$ against each).

\begin{figure}[!htbp]
  \centering
  \includegraphics[width=0.75\linewidth]{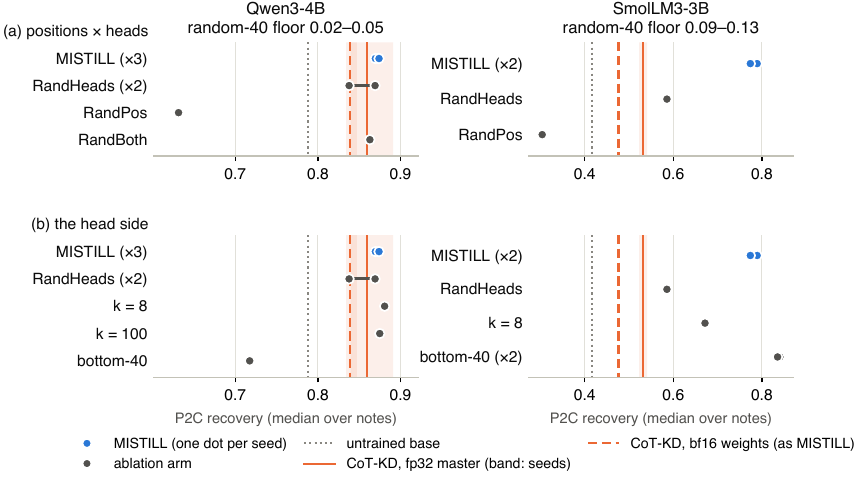}
  \caption{Where the transfer is located (Section~\ref{sec:distill-ablation}). (a) Own top heads supervised at random positions (\randpos) fall below the
  untrained student; random heads there do not (RandBoth; on \smol: Appendix~\ref{app:teacherpool}). (b) On \smol{} the supervised heads set how much is gained; on \qwenfour{} only the most negatively attributed heads matter,
  and they hurt. Phases, terms, dose: Figure~\ref{fig:ablation-c}. bf16; one seed unless marked. \resid{R-D7}\resid{R-D13}\resid{R-D14}}
  \label{fig:ablation}
\end{figure}

\begin{figure}[!htbp]
  \centering
  \includegraphics[width=0.92\linewidth]{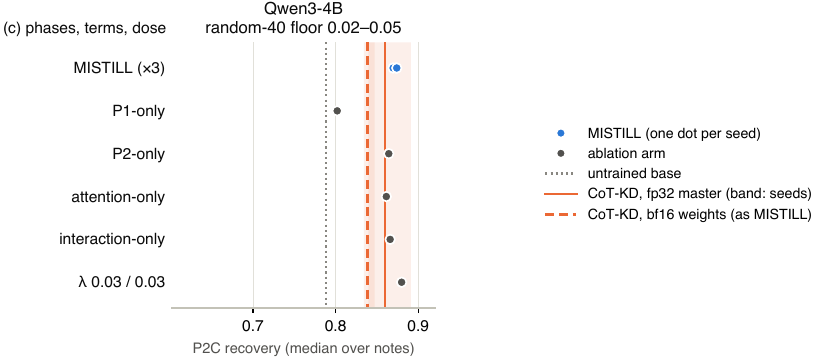}
  \caption{Figure~\ref{fig:ablation}, continued: (c) phases, terms and dose on \qwenfour{}. The \Ptwoc{} gain is the
  Phase-2 supervision's (\Ptwo{}-only keeps most of it, \Pone{}-only loses most of it); a third of the dose and either term
  alone leave it in place. All arms bf16, one seed unless marked. \resid{R-D7}\resid{R-D14}}
  \label{fig:ablation-c}
\end{figure}

\begin{table}[!htbp]
  \centering\footnotesize
  \caption{The ablations as pre-registered, before any result, with the outcome against each expectation
  (\appok: as expected; \appno: not). Wording follows the pre-registration, except that bottom-40 is described by
  what its heads are (the 40 most negatively attributed) and ``cost'' means the tier-1 task gap to fp32 \vanillakd{},
  which Appendix~\ref{app:res-precision} attributes to bf16 training. Tiers A to E are on \qwenfour; RandBoth on \smol{} and the teacher-side control: Appendix~\ref{app:teacherpool}. \resid{R-D7}\resid{R-D13}\resid{R-D14}}
  \label{tab:app-prereg}
  \setlength{\tabcolsep}{3pt}
  \begin{tabularx}{\linewidth}{@{}>{\raggedright\arraybackslash}p{0.14\linewidth} >{\raggedright\arraybackslash}p{0.19\linewidth} >{\raggedright\arraybackslash}X >{\raggedright\arraybackslash}X >{\raggedright\arraybackslash}X@{}}
    \toprule
    Tier & Arm (flag) & Question & Pre-registered expectation & Outcome \\
    \midrule
    A: positions, not heads & RandBoth (\appflag{heads random} \appflag{positions random}) & Is \randpos's collapse the positions, or the combination? & $\approx$ \randpos, below the untrained student & $=$ \mistill{} \appno \\
    A & bottom-40 (\appflag{heads bottom}) & The 40 most negatively attributed heads at the right events & $\approx$ \mistill{}\,/\,\randheads; if it falls, head identity has a floor & below the untrained student: the floor branch \\
    A & \Pone-only\,/\,\Ptwo-only ($\lambda_{\mathrm{P2}}=0$\,/\,$\lambda_{\mathrm{P1}}=0$) & Which events carry the \Ptwoc{} gain? & \Ptwo-only keeps the gain and most of the cost; \Pone-only neither & \Ptwo-only \appok; \Pone-only loses the gain \appok{} but keeps the cost \appno \\
    B: the trade-off & $\lambda$ 0.03\,/\,0.03 & Does a lighter dose cost less task and keep the gain? & the gain shrinks with the cost, or not & not: neither moves \\
    B & attention-only (\appflag{terms attn}), interaction-only (\appflag{terms inter}); both phases & Which term carries the gain, which the cost? & attention carries both; interaction adds the \Ptwoc{} margin & attention carries the cost; interaction alone keeps about four-fifths of the gain at no measurable cost \appno \\
    C: reproducibility & \randheads{} seed 1 & Is the seed stability the positions' or \mistill's heads'? & within 0.01 of \randheads{} seed 0 & fails (\NPairQwenRandHeadsSOneVsRandHeadsSZeroPTwoCDelta{} at \Ptwoc, \NPairQwenRandHeadsSOneVsRandHeadsSZeroPTwoDelta{} at \Ptwo) \\
    D: sensitivity & $k=8$, $k=100$ & Pool size & within seed noise of \mistill & \appok{} at \Ptwoc \\
    A$'$: heads where the recovery must be built, on \smol & bottom-40 (two seeds); $k=8$ & Which heads, and how many, on the student where head identity matters & bottom-40 $\le$ \randheads{} (\NOwnSmolRandHeadsSZeroPTwoCKiForty); $k=8$ near \mistill{} if the top of the ranking carries it, between \randheads{} and \mistill{} if the pool size does & bottom-40 at or above \mistill{} \appno; $k=8$ between \appok \\
    \midrule
    E (revision): the term, the ranking, the window & event-weighted \vanillakd{} (\appflag{event-weight 4}) & Does event-local loss pressure alone give the gain? & far below \mistill{} at \Ptwoc{} (A); matches it (B) & A: $=$ plain \vanillakd \appok \\
    E & uniform attention (\appflag{heads uniform} \appflag{terms attn}, $\lambda$ 0.135) & Does the ranking matter beyond event-local attention supervision? & below \mistill{} at \Ptwoc{} (A); matches it (B) & B, and above it on the causal readout, at a task cost \appno \\
    E & no window (\appflag{window 0}) & What does the \Pone{} window contribute? & \Pone{} recovery falls to \vanillakd's (A); nothing changes (B) & B \\
    C & \randheads{} pool 0, seeds 1 and 2 (\appflag{pool-seed 0}) & Is the control's seed spread the pool's or the training order's? & pool fixed, seeds within \mistill's spread (A); seeds separate (B); a seed reaches \mistill{} (C) & B at $k=40$, A at $k=8$; not C \\
    \bottomrule
  \end{tabularx}
\end{table}

\begin{table}[!htbp]
  \centering\small
  \caption{Ablations on each arm's own heads (150 test notes; one seed unless marked). \Ptwoc{} and \Ptwo: M4 knock-in
  of the own top-40, median over notes, random-40 floor in parentheses. $\Delta$ \Ptwoc: per-note paired median
  difference against each \mistill{} seed (\qwenfour{} seeds 0\,/\,1\,/\,2; \smol{} 0\,/\,1), sign-test $p$; vs base:
  against the untrained student, notes higher\,/\,lower. M2: \Ptwoc{} region lift (random). Prefix: share on the code's
  emitted prefix. \microf: with the paired difference from \mistill's seed-averaged notes [95\% interval]. All arms
  bf16. \resid{R-D7}\resid{R-D13}\resid{R-D14}}
  \label{tab:app-ablate}
  \providecommand{\gensd}[2]{\begin{tabular}[t]{@{}c@{}}#1\\[-2pt]{\scriptsize$\pm$#2}\end{tabular}}
\providecommand{\genstack}[2]{\begin{tabular}[t]{@{}c@{}}#1\\[-2pt]{\scriptsize #2}\end{tabular}}
\providecommand{\genstackl}[2]{\begin{tabular}[t]{@{}l@{}}#1\\[-2pt]{\scriptsize #2}\end{tabular}}
\providecommand{\genhead}[2]{\begin{tabular}[b]{@{}c@{}}#1\\#2\end{tabular}}
\providecommand{\genstackiii}[3]{\begin{tabular}[t]{@{}c@{}}#1\\[-2pt]{\scriptsize #2}\\[-3pt]{\scriptsize #3}\end{tabular}}
\scriptsize\setlength{\tabcolsep}{3pt}
\begin{adjustbox}{max width=\linewidth}
\begin{tabular}{@{}l c c c c c c c@{}}
\toprule
Arm (tier) & \genhead{\Ptwoc}{(random)} & \genhead{$\Delta$ \Ptwoc{} vs \mistill,}{per seed (sign $p$)} & \genhead{vs base}{(higher/lower)} & \Ptwo & \genhead{M2 \Ptwoc}{lift (random)} & prefix & \genhead{\microf}{($\Delta$ vs \mistill)} \\
\midrule
\multicolumn{8}{@{}l}{\qwenfour} \\
RandBoth (A) & 0.863\,{\scriptsize(0.035)} & \genstack{\ensuremath{-}0.005 / \ensuremath{-}0.007 / \ensuremath{-}0.006}{$p$ 0.16 / 0.37 / 0.22} & \genstack{+0.071}{119/31} & 0.738 & 2.69\,{\scriptsize(1.24)} & 0.139 & \genstack{0.138}{\pci{\ensuremath{-}0.010}{\ensuremath{-}0.028}{+0.007}} \\
bottom-40 (A) & 0.717\,{\scriptsize(0.049)} & \genstack{\ensuremath{-}0.173 / \ensuremath{-}0.175 / \ensuremath{-}0.180}{$p$ \ensuremath{2{\times}10^{-24}} / \ensuremath{3{\times}10^{-26}} / \ensuremath{2{\times}10^{-24}}} & \genstack{\ensuremath{-}0.088}{37/113} & 0.851 & 3.68\,{\scriptsize(0.80)} & 0.146 & \genstack{0.114}{\pci{\ensuremath{-}0.036}{\ensuremath{-}0.054}{\ensuremath{-}0.020}} \\
\Pone{}-only (A) & 0.802\,{\scriptsize(0.033)} & \genstack{\ensuremath{-}0.057 / \ensuremath{-}0.055 / \ensuremath{-}0.062}{$p$ \ensuremath{7{\times}10^{-8}} / \ensuremath{10^{-10}} / \ensuremath{3{\times}10^{-9}}} & \genstack{+0.028}{89/61} & 0.794 & 3.40\,{\scriptsize(1.59)} & 0.189 & \genstack{0.131}{\pci{\ensuremath{-}0.022}{\ensuremath{-}0.040}{\ensuremath{-}0.001}} \\
\Ptwo{}-only (A) & 0.864\,{\scriptsize(0.026)} & \genstack{\ensuremath{-}0.006 / \ensuremath{-}0.006 / \ensuremath{-}0.008}{$p$ \ensuremath{8{\times}10^{-4}} / 0.02 / \ensuremath{2{\times}10^{-4}}} & \genstack{+0.073}{121/29} & 0.732 & 3.53\,{\scriptsize(1.42)} & 0.283 & \genstack{0.154}{\pci{+0.004}{\ensuremath{-}0.011}{+0.020}} \\
$\lambda$ 0.03\,/\,0.03 (B) & 0.880\,{\scriptsize(0.026)} & \genstack{+0.009 / +0.009 / +0.004}{$p$ 0.22 / 0.03 / 0.81} & \genstack{+0.091}{130/20} & 0.748 & 3.75\,{\scriptsize(1.43)} & 0.233 & \genstack{0.150}{\pci{\ensuremath{-}0.002}{\ensuremath{-}0.021}{+0.017}} \\
attention-only (B) & 0.861\,{\scriptsize(0.039)} & \genstack{\ensuremath{-}0.008 / \ensuremath{-}0.009 / \ensuremath{-}0.012}{$p$ 0.29 / 0.09 / 0.02} & \genstack{+0.078}{117/33} & 0.766 & 3.67\,{\scriptsize(1.71)} & 0.219 & \genstack{0.130}{\pci{\ensuremath{-}0.019}{\ensuremath{-}0.038}{\ensuremath{-}0.002}} \\
interaction-only (B) & 0.866\,{\scriptsize(0.023)} & \genstack{\ensuremath{-}0.010 / \ensuremath{-}0.010 / \ensuremath{-}0.015}{$p$ \ensuremath{10^{-5}} / \ensuremath{8{\times}10^{-4}} / \ensuremath{5{\times}10^{-5}}} & \genstack{+0.064}{119/31} & 0.712 & 3.24\,{\scriptsize(1.18)} & 0.285 & \genstack{0.157}{\pci{+0.010}{\ensuremath{-}0.008}{+0.027}} \\
\randheads s1 (C) & 0.838\,{\scriptsize(0.040)} & \genstack{\ensuremath{-}0.031 / \ensuremath{-}0.027 / \ensuremath{-}0.028}{$p$ 0.001 / \ensuremath{10^{-5}} / \ensuremath{10^{-6}}} & \genstack{+0.051}{110/40} & 0.791 & 4.10\,{\scriptsize(1.82)} & 0.244 & \genstack{0.125\seedmark{2}}{\pci{\ensuremath{-}0.023}{\ensuremath{-}0.037}{\ensuremath{-}0.010}} \\
\randheads{} (pool 0) s1 (C) & 0.821\,{\scriptsize(0.042)} & \genstack{\ensuremath{-}0.054 / \ensuremath{-}0.052 / \ensuremath{-}0.060}{$p$ \ensuremath{3{\times}10^{-8}} / \ensuremath{4{\times}10^{-10}} / \ensuremath{3{\times}10^{-9}}} & \genstack{+0.028}{89/61} & 0.783 & 4.03\,{\scriptsize(1.38)} & 0.241 & \genstack{0.142\seedmark{2}}{\pci{\ensuremath{-}0.007}{\ensuremath{-}0.023}{+0.007}} \\
\randheads{} (pool 0) s2 (C) & 0.833\,{\scriptsize(0.039)} & \genstack{\ensuremath{-}0.031 / \ensuremath{-}0.034 / \ensuremath{-}0.038}{$p$ \ensuremath{2{\times}10^{-4}} / \ensuremath{3{\times}10^{-5}} / \ensuremath{2{\times}10^{-6}}} & \genstack{+0.038}{102/48} & 0.774 & 4.07\,{\scriptsize(1.39)} & 0.244 & \genstack{0.142\seedmark{2}}{\pci{\ensuremath{-}0.007}{\ensuremath{-}0.023}{+0.007}} \\
$k=8$ (D) & 0.881\,{\scriptsize(0.029)} & \genstack{+0.007 / +0.008 / +0.007}{$p$ 0.12 / 0.01 / 0.29} & \genstack{+0.086}{124/26} & 0.753 & 3.35\,{\scriptsize(1.65)} & 0.248 & \genstack{0.153}{\pci{+0.005}{\ensuremath{-}0.013}{+0.023}} \\
$k=100$ (D) & 0.875\,{\scriptsize(0.030)} & \genstack{+0.006 / +0.006 / +0.001}{$p$ 0.37 / 0.09 / 0.81} & \genstack{+0.083}{121/29} & 0.861 & 3.31\,{\scriptsize(1.76)} & 0.268 & \genstack{0.141}{\pci{\ensuremath{-}0.008}{\ensuremath{-}0.025}{+0.009}} \\
event-weighted \vanillakd (E) & 0.828\,{\scriptsize(0.036)} & \genstack{\ensuremath{-}0.042 / \ensuremath{-}0.043 / \ensuremath{-}0.042}{$p$ \ensuremath{10^{-5}} / \ensuremath{5{\times}10^{-5}} / \ensuremath{2{\times}10^{-6}}} & \genstack{+0.042}{99/51} & 0.761 & 3.28\,{\scriptsize(1.45)} & 0.185 & \genstack{0.141}{\pci{\ensuremath{-}0.018}{\ensuremath{-}0.039}{+0.003}} \\
uniform attention, all heads (E) & 0.915\,{\scriptsize(0.035)} & \genstack{+0.047 / +0.052 / +0.045}{$p$ \ensuremath{8{\times}10^{-16}} / \ensuremath{2{\times}10^{-16}} / \ensuremath{10^{-14}}} & \genstack{+0.123}{136/14} & 0.889 & 3.68\,{\scriptsize(1.33)} & 0.422 & \genstack{0.121}{\pci{\ensuremath{-}0.028}{\ensuremath{-}0.046}{\ensuremath{-}0.011}} \\
no window (E) & 0.870\,{\scriptsize(0.025)} & \genstack{\ensuremath{-}0.008 / \ensuremath{-}0.010 / \ensuremath{-}0.011}{$p$ \ensuremath{10^{-9}} / \ensuremath{2{\times}10^{-4}} / \ensuremath{7{\times}10^{-8}}} & \genstack{+0.068}{119/31} & 0.745 & 3.71\,{\scriptsize(1.32)} & 0.281 & \genstack{0.147}{\pci{\ensuremath{-}0.001}{\ensuremath{-}0.018}{+0.014}} \\
\midrule
\multicolumn{8}{@{}l}{\smol} \\
bottom-40 (A$'$) & 0.841\,{\scriptsize(0.098)} & \genstack{+0.034 / +0.070 / +0.062}{$p$ 0.04 / 0.002 / 0.004} & \genstack{+0.409}{141/9} & 0.857 & 2.45\,{\scriptsize(1.93)} & 0.279 & \genstack{0.048\seedmark{2}}{\pci{\ensuremath{-}0.001}{\ensuremath{-}0.014}{+0.012}} \\
bottom-40 s1 (A$'$) & 0.835\,{\scriptsize(0.096)} & \genstack{+0.010 / +0.026 / +0.040}{$p$ 0.81 / 0.29 / 0.06} & \genstack{+0.407}{137/13} & 0.862 & 2.46\,{\scriptsize(1.70)} & 0.277 & \genstack{0.048\seedmark{2}}{\pci{\ensuremath{-}0.001}{\ensuremath{-}0.014}{+0.012}} \\
$k=8$ (A$'$) & 0.671\,{\scriptsize(0.115)} & \genstack{\ensuremath{-}0.085 / \ensuremath{-}0.089 / \ensuremath{-}0.069}{$p$ \ensuremath{3{\times}10^{-12}} / \ensuremath{10^{-10}} / \ensuremath{3{\times}10^{-9}}} & \genstack{+0.298}{130/20} & 0.860 & 3.11\,{\scriptsize(1.29)} & 0.388 & \genstack{0.047}{\pci{\ensuremath{-}0.001}{\ensuremath{-}0.016}{+0.015}} \\
\bottomrule
\end{tabular}
\end{adjustbox}

\end{table}

\subsection{Tier by tier}
\label{app:abl-tiers}
\paragraph{Tier A: the positions, and whose heads.} Both head-side expectations failed, and together they sharpen the
reading of Sec.~\ref{sec:distill-mech} rather than overturn it. RandBoth restores \NOwnQwenRandBothSZeroPTwoCKiForty, with
no paired difference from any \mistill{} seed (Table~\ref{tab:app-ablate}) or from \randheads{}
(\NPairQwenRandBothSZeroVsRandHeadsSZeroPTwoCDelta, $p$ \NPairQwenRandBothSZeroVsRandHeadsSZeroPTwoCP), and lies
\NPairQwenRandBothSZeroVsRandPosSZeroPTwoCDelta{} above \randpos{} (higher on
\NPairQwenRandBothSZeroVsRandPosSZeroPTwoCHigher{} of 150 notes). \randpos's collapse is therefore not ``wrong
positions'' as such but the student's own ranked heads pointed at the wrong positions; random heads at random positions
do no such harm (on \smol{} they gain nothing either: Appendix~\ref{app:teacherpool}). Bottom-40 falls below the untrained student: it restores \NOwnQwenBottomSZeroPTwoCKiForty{} against
\NOwnQwenBasePTwoCKiForty, lower on \NPairQwenBottomSZeroVsBasePTwoCLower{} of 150 notes and below every \mistill{} seed,
so head identity has a floor even on the student where random heads can stand in for its top-ranked ones. The phases
place the \Ptwoc{} gain in the Phase-2 supervision: \Ptwo-only restores \NOwnQwenPTwoOnlySZeroPTwoCKiForty{} and
\Pone-only \NOwnQwenPOneOnlySZeroPTwoCKiForty{} (\Ptwo-only minus \Pone-only
\NPairQwenPTwoOnlySZeroVsPOneOnlySZeroPTwoCDelta, higher on \NPairQwenPTwoOnlySZeroVsPOneOnlySZeroPTwoCHigher{} of 150).
\Ptwo-only sits slightly below every \mistill{} seed per note, outside \mistill's seed band, so ``keeps the gain'' means
about nine-tenths of it (\Ptwo-only minus the untrained student \NPairQwenPTwoOnlySZeroVsBasePTwoCDelta, against
\NMDQwenMistillVsBasePTwoCMin{} to \NMDQwenMistillVsBasePTwoCMax{} for \mistill's seeds). The task does not follow the \Ptwoc{} gain: \Pone-only
scores \NQwenPOneOnlyMicroF{} (\NQwenPOneOnlyMicroFVsMistillDelta{} [\NQwenPOneOnlyMicroFVsMistillLo,
\NQwenPOneOnlyMicroFVsMistillHi] against \mistill) and \Ptwo-only \NQwenPTwoOnlyMicroF{}
(\NQwenPTwoOnlyMicroFVsMistillDelta{} [\NQwenPTwoOnlyMicroFVsMistillLo, \NQwenPTwoOnlyMicroFVsMistillHi]).

\paragraph{Tier B: dose and terms.} At a third of the weight neither the gain (\NOwnQwenLowDoseSZeroPTwoCKiForty) nor the
task (\NQwenLowDoseMicroF, \NQwenLowDoseMicroFVsMistillDelta{} [\NQwenLowDoseMicroFVsMistillLo,
\NQwenLowDoseMicroFVsMistillHi] against \mistill) moves: $\lambda=0.1$ sits on a plateau, and a lighter dose does not buy
task score back in this range. The term arms act on both phases, because the terms switch is global in the trainer;
since the \Pone{} term moves nothing at a third of its weight (Appendix~\ref{app:res-extra}), this is close to switching
the term at \Ptwo{} alone. Attention-only restores \NOwnQwenAttnOnlySZeroPTwoCKiForty{} and interaction-only
\NOwnQwenInterOnlySZeroPTwoCKiForty, no different from each other per note ($p$
\NPairQwenAttnOnlySZeroVsInterOnlySZeroPTwoCP). Attention-only keeps \mistill's gain over the untrained student
(\NPairQwenAttnOnlySZeroVsBasePTwoCDelta; below \mistill{} only against seed 2) at a lower task score
(\NQwenAttnOnlyMicroF, \NQwenAttnOnlyMicroFVsMistillDelta{} [\NQwenAttnOnlyMicroFVsMistillLo,
\NQwenAttnOnlyMicroFVsMistillHi]); interaction-only sits below every \mistill{} seed per note, keeping about four-fifths
of the gain (\NPairQwenInterOnlySZeroVsBasePTwoCDelta{} over the untrained student), at \mistill's task score
(\NQwenInterOnlyMicroF, \NQwenInterOnlyMicroFVsMistillDelta{} [\NQwenInterOnlyMicroFVsMistillLo,
\NQwenInterOnlyMicroFVsMistillHi]).

\paragraph{Tier C: reproducibility.} The pre-registered test fails. \randheads' two seeds differ at \Ptwoc{} (seed 1 minus
seed 0 \NPairQwenRandHeadsSOneVsRandHeadsSZeroPTwoCDelta, lower on \NPairQwenRandHeadsSOneVsRandHeadsSZeroPTwoCLower{} of
150, $p$ \NPairQwenRandHeadsSOneVsRandHeadsSZeroPTwoCP), at \Ptwo{} (\NPairQwenRandHeadsSOneVsRandHeadsSZeroPTwoDelta, lower
on \NPairQwenRandHeadsSOneVsRandHeadsSZeroPTwoLower{} of 150) and on the task (\NQwenRandHeadsSZeroMicroF{} and
\NQwenRandHeadsSOneMicroF), where \mistill's seeds do not separate per note (seed 1 minus seed 0
\NPairQwenMistillSOneVsMistillSZeroPTwoCDelta, \NPairQwenMistillSOneVsMistillSZeroPTwoCHigher{} higher and
\NPairQwenMistillSOneVsMistillSZeroPTwoCLower{} lower). The seed stability tracks \mistill's heads, the student's own
ranked pool, rather than position supervision as such. Both \randheads{} seeds train bf16 weights, as \mistill's do, so
bf16 training alone does not make seeds agree; it does narrow \vanillakd's spread, from \NOwnQwenVanillaSeedSpreadPTwoC{} to
\NOwnQwenVanillaBfSeedSpreadPTwoC, but two of the three bf16 \vanillakd{} seed pairs still separate per note
(Appendix~\ref{app:res-precision}). The two \randheads{} seeds also draw two different random pools, so their spread mixes
the pool with the training order. Holding one pool fixed (two more seeds on a frozen draw, Table~\ref{tab:app-ablate})
separates the two: training order alone moves \Ptwoc{} recovery by
\NPairQwenRandHeadsPoolSTwoVsRandHeadsPoolSOnePTwoCDelta{} at $k=40$ (higher on
\NPairQwenRandHeadsPoolSTwoVsRandHeadsPoolSOnePTwoCHigher{} of 150 notes) and
\NPairQwenRandHeadsPoolSTwoVsRandHeadsPoolSOnePTwoCKiEightDelta{} at $k=8$, four to five times \mistill's seed spread, and the
pool moves it by up to \NPairQwenRandHeadsPoolSOneVsRandHeadsSZeroPTwoCDelta{} at $k=40$ and
\NPairQwenRandHeadsPoolSOneVsRandHeadsSZeroPTwoCKiEightDelta{} at $k=8$. Random heads are not seed-stable even when the
pool is held fixed. Across the four \randheads{} runs over three pools, \mistill{} is higher by
\NPoolQwenMistillVsRandHeadsAllPTwoCClusterDelta{} at $k=40$ (note-clustered over the four pairs;
\NPoolQwenMistillVsRandHeadsAllPTwoCClusterHigher{} of 150 notes, $p$ \NPoolQwenMistillVsRandHeadsAllPTwoCClusterP; one pool
matches it) and by \NPoolQwenMistillVsRandHeadsAllPTwoCKiEightClusterDelta{} at $k=8$
(\NPoolQwenMistillVsRandHeadsAllPTwoCKiEightClusterHigher{} of 150, every run below): the $k=8$ gap is the robust one, the
$k=40$ gap depends on which pool is drawn (against pool 0 under matched selection it is \NMatchedPTwoCClusterMean,
Appendix~\ref{app:abl-matched}).

\paragraph{Tier D: pool size (\qwenfour).} Eight, forty or a hundred supervised heads give the same \Ptwoc{} recovery
(\NOwnQwenKEightSZeroPTwoCKiForty, \NOwnQwenMistillSZeroPTwoCKiForty{} and \NOwnQwenKHundredSZeroPTwoCKiForty; no paired
difference from \mistill{} holds against every seed). Only the widest pool lifts \Ptwo{}
(\NOwnQwenKHundredSZeroPTwoKiForty, above \mistill{} seed 0 on \NPairQwenKHundredSZeroVsMistillSZeroPTwoHigher{} of 150
notes), as \randheads{} (\NOwnQwenRandHeadsSZeroPTwoKiForty) and bottom-40 (\NOwnQwenBottomSZeroPTwoKiForty) also do;
\mistill's own \Ptwo{} recovery (\NOwnQwenMistillSZeroPTwoKiForty) lies below the untrained student's
(\NOwnQwenBasePTwoKiForty).

\paragraph{Tier E: what the attention term does, and what the ranking buys.} Three controls added after the submission
close the two readings the controls above left open, each pre-registered with its branches before its first job and
each one flag away from \mistill{} at tier-1 size with bf16 weights (Table~\ref{tab:app-ablate}). \emph{Event-weighted
\vanillakd} asks whether event-local loss pressure alone explains the gain (a key-token weighting in the sense of KPOD,
\citealp{feng2024keypoint}, with the key tokens placed by the teacher's causal instrument rather than learned): plain \vanillakd{} with its cross-entropy and
KL multiplied by 4 at the teacher's event tokens and window positions (about a tenth of the trace tokens), which adds the
same extra loss \mistill's two terms add (0.27 against 0.25 nats at the start, both decaying to 0.16), with no attention
term. Its own-heads recovery is plain \vanillakd's (\NOwnQwenEventWeightedSZeroPTwoCKiForty{} against
\NOwnQwenVanillaBfSZeroPTwoCKiForty{} at $k=40$; \NOwnQwenEventWeightedSZeroPTwoCKiEight{} against
\NOwnQwenVanillaBfSZeroPTwoCKiEight{} at $k=8$; per note \NPairQwenEventWeightedSZeroVsVanillaBfSZeroPTwoCDelta, $p$
\NPairQwenEventWeightedSZeroVsVanillaBfSZeroPTwoCP), below \mistill{} on \NPairQwenEventWeightedSZeroVsMistillSZeroPTwoCLower{}
of 150 notes at $k=40$ and \NPairQwenEventWeightedSZeroVsMistillSZeroPTwoCKiEightLower{} at $k=8$, and below \randheads{}
(\NPairQwenEventWeightedSZeroVsRandHeadsSZeroPTwoCDelta); its supervised readout is \vanillakd's too (M1 at \Ptwoc{}
\NQwenEventWeightedMOnePTwoC{} against \NQwenVanillaBfMOnePTwoC), and the weighting raises the share of generations that run
to the cap to \NQwenEventWeightedHitCap. Event-specific loss pressure does not produce the mechanism gain; the attention
and interaction terms do. \emph{Uniform attention} asks whether the causal ranking matters beyond event-local attention
supervision: the attention term at \emph{every} head of the supervised layers with equal weights, no ranking and no
interaction term, $\lambda$ raised to 0.135 so that the extra loss matches \mistill's at the start. On the causal readout it
does not merely match \mistill, it exceeds it: own-heads \Ptwoc{} recovery \NOwnQwenUniformSZeroPTwoCKiForty{} at $k=40$
(higher than \mistill{} seed 0 on \NPairQwenUniformSZeroVsMistillSZeroPTwoCHigher{} of 150 notes,
\NPairQwenUniformSZeroVsMistillSZeroPTwoCDelta) and \NOwnQwenUniformSZeroPTwoCKiEight{} at $k=8$, above the ranked
attention-only arm (\NOwnQwenAttnOnlySZeroPTwoCKiForty{} / \NOwnQwenAttnOnlySZeroPTwoCKiEight) and \randheads, with its
contrastive heads reading the region as much as \mistill's (M2 lift \NOwnQwenUniformSZeroMTwoPTwoCLift{} against
\NOwnQwenMistillSZeroMTwoPTwoCLift). The two tested configurations differ on the other side of the ledger: the uniform control
pays on the task (\microf{} \NQwenUniformMicroF{} against \NQwenMistillSZeroMicroF; \NQwenUniformHitCap{} of generations run
to the cap) and loses the supervised readout (M1 at \Ptwoc{} \NQwenUniformMOnePTwoC, below \vanillakd's), where
\mistill's forty ranked heads keep both. With \randheads{} (the same terms at forty random heads of the same layers,
most of the $k=40$ gain) and the event-weighted control (no attention term, no gain), the four arms say that event-local
\emph{attention} supervision in those layers increases the student's own-heads causal recovery; the ranked
configuration has a better observed task--recovery trade-off than the uniform control, but the two also differ in term
composition (no interaction term), head count and weight, so this comparison does not isolate what the ranking itself
buys. \emph{No window} removes the $W{=}8$ positions before each chapter letter from the \Pone{} term (half its rows).
The main recovery gain remains, with a small paired decrease: \microf{} \NQwenNoWindowMicroF{} against \NQwenMistillSZeroMicroF{}
(no detected task difference), every teacher-forced M1
cell identical to three decimals, own-heads recovery at \Pone{} \NOwnQwenNoWindowSZeroPOneKiForty{} against
\NOwnQwenMistillSZeroPOneKiForty{} and at \Ptwoc{} \NOwnQwenNoWindowSZeroPTwoCKiForty{} against
\NOwnQwenMistillSZeroPTwoCKiForty{} (per note \NPairQwenNoWindowSZeroVsMistillSZeroPTwoCDelta, $p$
\NPairQwenNoWindowSZeroVsMistillSZeroPTwoCP; lower on every seed pair, Table~\ref{tab:app-ablate}), with the contrastive gain
over \vanillakd{} fully present (\NPairQwenNoWindowSZeroVsVanillaBfSZeroPTwoCKiEightDelta{} at $k=8$ on
\NPairQwenNoWindowSZeroVsVanillaBfSZeroPTwoCKiEightHigher{} of 150 notes); its chapter heads read the region slightly more
without the window (M2 lift \NOwnQwenNoWindowSZeroMTwoPOneLift{} against \NOwnQwenMistillSZeroMTwoPOneLift). The window is
unnecessary for most of the gain, not inert (Appendix~\ref{app:objective}).

\paragraph{Tier A$'$: which heads, and how many, where the recovery must be built.} The untrained \smol{} restores
\NOwnSmolBasePTwoCKiForty{} of the contrastive decision against a floor of \NOwnSmolBasePTwoCKiFortyRand, and on it both
the identity and the number of the supervised heads matter. The base's 40 most negatively attributed heads yield at
least as much as its top-40: \NOwnSmolBottomSZeroPTwoCKiForty{} and \NOwnSmolBottomSOnePTwoCKiForty{} on two seeds against
\mistill's \NOwnSmolMistillSZeroPTwoCKiForty{} and \NOwnSmolMistillSOnePTwoCKiForty, seed 0 above both \mistill{} seeds per
note and seed 1 level with them (Table~\ref{tab:app-ablate}), both far above \randheads{}
(\NPairSmolBottomSZeroVsRandHeadsSZeroPTwoCDelta{} and \NPairSmolBottomSOneVsRandHeadsSZeroPTwoCDelta). The base's top-8
yield less (\NOwnSmolKEightSZeroPTwoCKiForty: below both \mistill{} seeds; \NPairSmolKEightSZeroVsRandHeadsSZeroPTwoCDelta{}
over \randheads, higher on \NPairSmolKEightSZeroVsRandHeadsSZeroPTwoCHigher{} of 150; \NPairSmolKEightSZeroVsBottomSZeroPTwoCDelta{}
and \NPairSmolKEightSZeroVsBottomSOnePTwoCDelta{} against bottom-40's seeds), and 40 random heads least
(\NOwnSmolRandHeadsSZeroPTwoCKiForty). \Ptwo{} does not separate the structured pools (bottom-40
\NOwnSmolBottomSZeroPTwoKiForty{} and \NOwnSmolBottomSOnePTwoKiForty, $k=8$ \NOwnSmolKEightSZeroPTwoKiForty, \mistill{}
\NOwnSmolMistillSZeroPTwoKiForty{} and \NOwnSmolMistillSOnePTwoKiForty; $k=8$ minus bottom-40
\NPairSmolKEightSZeroVsBottomSZeroPTwoDelta, $p$ \NPairSmolKEightSZeroVsBottomSZeroPTwoP). Bottom-40's two seeds agree at
\Ptwo{} (\NPairSmolBottomSOneVsBottomSZeroPTwoDelta, $p$ \NPairSmolBottomSOneVsBottomSZeroPTwoP), on the read side and in
their carriers, but differ at \Ptwoc{} (\NPairSmolBottomSOneVsBottomSZeroPTwoCDelta, lower on
\NPairSmolBottomSOneVsBottomSZeroPTwoCLower{} of 150, $p$ \NPairSmolBottomSOneVsBottomSZeroPTwoCP), where \mistill's do not
($p$ \NPairSmolMistillSOneVsMistillSZeroPTwoCP): the pattern of tier C's \randheads{} seeds on \qwenfour. Against the
pre-registration, bottom-40 was expected at or below \randheads; it is level with or above \mistill, the highest cells on
this student. $k=8$ lands between \randheads{} and \mistill, the ``pool size matters'' branch, where on \qwenfour{} pool
size did not matter. On the task every A$'$ arm is at \mistill's level (bottom-40 \NSmolBottomMicroF{} over two seeds,
\NSmolBottomMicroFVsMistillDelta{} [\NSmolBottomMicroFVsMistillLo, \NSmolBottomMicroFVsMistillHi]; $k=8$ \NSmolKEightMicroF,
\NSmolKEightMicroFVsMistillDelta{} [\NSmolKEightMicroFVsMistillLo, \NSmolKEightMicroFVsMistillHi]), as is every
attention-supervising arm on this student.

\paragraph{Task effects, read against the bf16 band.} Every arm here trains bf16 weights, so task differences are read
against bf16 \vanillakd{} (\pmsd{\NQwenVanillaBfMicroF}{\NQwenVanillaBfMicroFSd} on \qwenfour, \NSmolVanillaBfMicroF{} on
\smol), not as a cost against the fp32 baselines. The single-seed \qwenfour{} ablations span \NQwenBottomMicroF{}
(bottom-40) to \NQwenInterOnlyMicroF{} (interaction-only) around it, and the attention-supervising \smol{} arms,
\mistill's seeds included, span \NSmolMistillSOneMicroF{} to \NSmolRandPosMicroF{} around \NSmolVanillaBfMicroF.

\subsection{Matched selection against a frozen random pool}
\label{app:abl-matched}
\begin{figure}[!htbp]
  \centering
  \includegraphics[width=\linewidth]{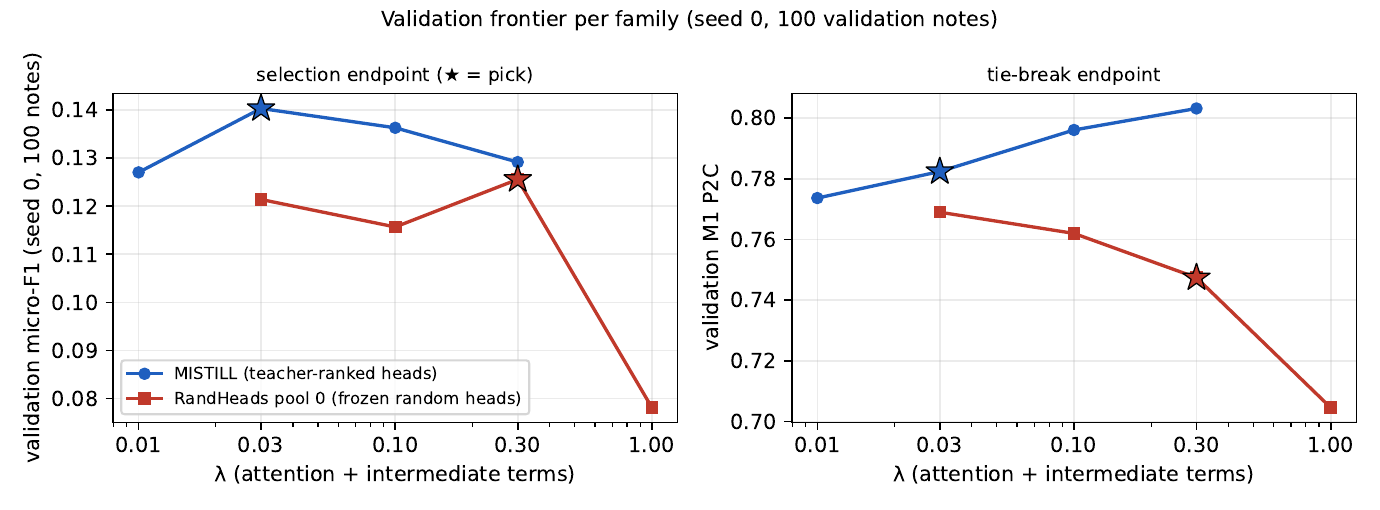}
  \caption{Matched selection: the validation frontier per family (seed 0, 100 validation notes; \microf{} left, M1 at
  \Ptwoc{} right) over the same initial grid plus the same one-step extension rule, boundary cells included; the picks are
  starred. Ties are cells within 0.004 of the best validation \microf{}, a selection tolerance rather than an uncertainty
  band. \resid{R-D31}}
  \label{fig:a2-frontier}
\end{figure}

The \randheads{} runs above train at one transferred $\lambda$ while \mistill's $\lambda$ had been selected. One selection
protocol for both families removes that asymmetry (R-D31, R-D32): the tier-1 bf16 line, the initial grid
$\lambda\in\{0.03,0.1,0.3\}$ on seed 0, selection by validation \microf{} on the 100 validation notes with ties within 0.004 of
the best cell broken by M1 at \Ptwoc, one boundary cell one step outward when the pick sits on the grid's edge, and three
test seeds at each pick; no test sweep anywhere. \mistill{} picks $\lambda=\NMatchedSelectMistillPick$ (its extension to 0.01
was lower) and \randheads{} pool 0 picks $\lambda=\NMatchedSelectRandHeadsPick$ (its extension to 1 collapsed).
Figure~\ref{fig:a2-frontier} shows both validation frontiers and Table~\ref{tab:app-matched} the cells, the six test cells at
the picks and the paired comparisons. At the picks \mistill{} exceeds the frozen pool in own-heads normalized \Ptwoc{} recovery
at $k=40$ (mean paired $\Delta$ \NMatchedPTwoCClusterMean{} [\NMatchedPTwoCClusterLo, \NMatchedPTwoCClusterHi], the three seed
pairs averaged within each of 150 test notes, higher on \NMatchedPTwoCClusterHigher; per pair \NMatchedPTwoCSZeroMean,
\NMatchedPTwoCSOneMean{} and \NMatchedPTwoCSTwoMean, every interval above zero; the per-seed medians are
\NMatchedMistillSeedMeanPTwoCKiForty{} against \NMatchedRandHeadsSeedMeanPTwoCKiForty{} on average) and in corpus \microf{}
(\NMatchedTaskDelta{} [\NMatchedTaskLo, \NMatchedTaskHi]; per pair \NMatchedTaskSZeroDelta, \NMatchedTaskSOneDelta{} and
\NMatchedTaskSTwoDelta); at $k=8$ the recovery gap is \NMatchedPTwoCKiEightClusterMean{} [\NMatchedPTwoCKiEightClusterLo,
\NMatchedPTwoCKiEightClusterHi]. The M2 shortlist-region lift of the own top-16 heads, a dimensionless quantity, is higher for
\randheads{} (\NMatchedMTwoPTwoCClusterMean{} [\NMatchedMTwoPTwoCClusterLo, \NMatchedMTwoPTwoCClusterHi]) despite its lower
own-heads recovery; \Pone{} recovery is saturated for both (\NMatchedPOneClusterMean).

\emph{Tie band.} Both primary picks were decided by full-precision validation gaps of \NMatchedSelectMistillGapFull{} (\mistill,
$\lambda$ 0.1 behind 0.03) and \NMatchedSelectRandHeadsGapFull{} (\randheads, 0.03 behind 0.3), each outside the inclusive
0.004 band. The locked 0.004 tie band selected $\lambda=\NMatchedSelectMistillPick$ for \mistill{} and
\NMatchedSelectRandHeadsPick{} for \randheads{} pool 0; a secondary 0.005-band check on the existing validation cells instead
selects \NMatchedSelectMistillPickFive{} and \NMatchedSelectRandHeadsPickFive; the latter is an unextended lower-boundary
choice. The reported test comparison therefore remains conditional on the locked selection rule.

\emph{Scope.} This is evidence for ranked supervision under matched tuning against one frozen random pool in the tested
\qwenfour{} bf16 setting (three training seeds are not three pools); because the selected $\lambda$ differ, it compares the two
configurations after the declared tuning procedure rather than head identity at a fixed $\lambda$; the note bootstrap
conditions on the tested seeds, settings and pool; and the task difference is against \randheads, not a task claim against
matched-precision \vanillakd. The single-$\lambda$ comparisons of Table~\ref{tab:app-ablate} and tier C keep their original
scope. \resid{R-D31}\resid{R-D32}

\begin{table}[!htbp]
  \centering\small
  \caption{Matched selection against frozen \randheads{} pool 0 (\qwenfour, bf16): the validation cells per $\lambda$ with the
  picks at the locked 0.004 band ($\star$) and at the 0.005 sensitivity band ($\dagger$); the six test cells at the picks (150
  test notes, three seeds; own-heads cells are medians over notes with random floors, as in Table~\ref{tab:app-mech-seeds});
  the paired comparisons as separately labelled rows (per-note means of the seed-pair differences, note-clustered).
  \resid{R-D31}\resid{R-D32}}
  \label{tab:app-matched}
  \providecommand{\gensd}[2]{\begin{tabular}[t]{@{}c@{}}#1\\[-2pt]{\scriptsize$\pm$#2}\end{tabular}}
\providecommand{\genstack}[2]{\begin{tabular}[t]{@{}c@{}}#1\\[-2pt]{\scriptsize #2}\end{tabular}}
\providecommand{\genstackl}[2]{\begin{tabular}[t]{@{}l@{}}#1\\[-2pt]{\scriptsize #2}\end{tabular}}
\providecommand{\genhead}[2]{\begin{tabular}[b]{@{}c@{}}#1\\#2\end{tabular}}
\providecommand{\genstackiii}[3]{\begin{tabular}[t]{@{}c@{}}#1\\[-2pt]{\scriptsize #2}\\[-3pt]{\scriptsize #3}\end{tabular}}
\scriptsize\setlength{\tabcolsep}{3pt}
\begin{adjustbox}{max width=\linewidth}
\begin{tabular}{@{}p{0.17\linewidth} p{0.22\linewidth} p{0.22\linewidth} p{0.37\linewidth}@{}}
\toprule
Quantity & \mistill{} ($\lambda$ 0.03, matched pick) & \randheads{} pool 0 ($\lambda$ 0.3, matched pick) & \mistill{} $-$ \randheads{} \\
\midrule
\multicolumn{4}{@{}p{0.98\linewidth}}{\textit{Selection on the 100 validation notes (seed 0): cells of the initial grid plus the one-step extension; $\star$ = the pick at the locked 0.004 band; $\dagger$ = the pick at the 0.005 sensitivity band}} \\
validation \microf{} per $\lambda$ & 0.01: 0.127; 0.03: 0.140$\star$; 0.1: 0.136$^{\dagger}$; 0.3: 0.129 & 0.03: 0.121$^{\dagger}$; 0.1: 0.116; 0.3: 0.126$\star$; 1: 0.078 & --- \\
validation M1 \Ptwoc{} per $\lambda$ & 0.01: 0.774; 0.03: 0.782$\star$; 0.1: 0.796$^{\dagger}$; 0.3: 0.803 & 0.03: 0.769$^{\dagger}$; 0.1: 0.762; 0.3: 0.747$\star$; 1: 0.705 & --- \\
full-precision gap of the runner-up to the best cell & 0.0040204002 & 0.0041750956 & --- \\
\midrule
\multicolumn{4}{@{}p{0.98\linewidth}}{\textit{The six test cells at the picks (150 test notes; three seeds; per seed s0 / s1 / s2, then mean$\pm$sd over seeds)}} \\
test \microf & 0.169 / 0.152 / 0.144\newline 0.155{\scriptsize$\pm$.013} & 0.126 / 0.124 / 0.121\newline 0.124{\scriptsize$\pm$.002} & corpus-level +0.032 [+0.019, +0.044]\newline per pair +0.043 [+0.020, +0.065]; +0.029 [+0.009, +0.049]; +0.023 [+0.001, +0.045] \\
M1 \Ptwoc{} (notes with events) & 0.789 / 0.790 / 0.790\newline 0.790{\scriptsize$\pm$.001} & 0.750 / 0.752 / 0.752\newline 0.751{\scriptsize$\pm$.001} & --- \\
own-heads \Ptwoc{} knock-in, $k=40$ (median; random floor) & 0.877 / 0.884 / 0.878\newline 0.880{\scriptsize$\pm$.004} (0.030) & 0.819 / 0.819 / 0.824\newline 0.821{\scriptsize$\pm$.003} (0.044) & --- \\
\quad paired mean $\Delta$, note-clustered (150 notes) & --- & --- & +0.065 [+0.053, +0.077] (125/25, $p$ \ensuremath{3{\times}10^{-17}})\newline per pair: +0.062 [+0.049, +0.075]; +0.070 [+0.057, +0.083]; +0.063 [+0.049, +0.076] \\
own-heads \Ptwoc{} knock-in, $k=8$ (median) & 0.569 / 0.576 / 0.562\newline 0.569{\scriptsize$\pm$.007} & 0.471 / 0.453 / 0.460\newline 0.461{\scriptsize$\pm$.009} & --- \\
\quad paired mean $\Delta$, note-clustered & --- & --- & +0.114 [+0.104, +0.124] (148/2, $p$ \ensuremath{2{\times}10^{-41}})\newline per pair: +0.108 [+0.098, +0.119]; +0.120 [+0.110, +0.131]; +0.114 [+0.104, +0.125] \\
own-heads \Pone{} knock-in, $k=40$ (median) & 0.952 / 0.954 / 0.955 & 0.961 / 0.965 / 0.964 & --- \\
\quad paired mean $\Delta$, note-clustered & --- & --- & \ensuremath{-}0.011 [\ensuremath{-}0.014, \ensuremath{-}0.007] (43/107, $p$ \ensuremath{2{\times}10^{-7}})\newline per pair: \ensuremath{-}0.010 [\ensuremath{-}0.014, \ensuremath{-}0.007]; \ensuremath{-}0.012 [\ensuremath{-}0.016, \ensuremath{-}0.009]; \ensuremath{-}0.010 [\ensuremath{-}0.013, \ensuremath{-}0.006] \\
M2 shortlist-region lift, own top-16 \Ptwoc{} heads (median; dimensionless) & 3.69 / 3.81 / 3.71\newline 3.73{\scriptsize$\pm$.06} & 3.79 / 3.83 / 3.78\newline 3.80{\scriptsize$\pm$.03} & --- \\
\quad paired mean $\Delta$, note-clustered & --- & --- & \ensuremath{-}0.426 [\ensuremath{-}0.650, \ensuremath{-}0.261] (50/100, $p$ \ensuremath{5{\times}10^{-5}})\newline per pair: \ensuremath{-}0.429 [\ensuremath{-}0.661, \ensuremath{-}0.257]; \ensuremath{-}0.411 [\ensuremath{-}0.638, \ensuremath{-}0.247]; \ensuremath{-}0.438 [\ensuremath{-}0.659, \ensuremath{-}0.269] \\
\bottomrule
\end{tabular}
\end{adjustbox}

\end{table}

\subsection{Which heads the objective reaches, and where the mechanism lands}
\label{app:abl-carriers}
\paragraph{What bottom-40 is.} \appflag{heads bottom} gives the pool's weights to the 40 heads that the untrained
student's calibrated ranking puts last, keeping the weight profile of the top-40 pool (as \randheads{} does), so the arm
differs from \mistill{} in the ranking alone. The attribution is signed, so these are the most negatively attributed
heads, whose contribution works against the decision, not heads without effect: at \Ptwoc{} on \smol{} their importances
run from \NMDBottomSmolPTwoCImportanceLeastNeg{} to \NMDBottomSmolPTwoCImportanceMostNeg{} (sum \NMDBottomSmolPTwoCImportanceSum) against \NMDTopSmolPTwoCImportanceMin{} to \NMDTopSmolPTwoCImportanceMax{} (sum \NMDTopSmolPTwoCImportanceSum) for the top-40,
with the median head at zero; on \qwenfour{} the sums are \NMDBottomQwenPTwoCImportanceSum{} against \NMDTopQwenPTwoCImportanceSum.

\paragraph{Where the gain lands.} In every arm, on both students, the causal carriers after training are
the untrained student's own top-ranked heads, whichever heads were supervised. Every one of the 29 arms read this way
(10 on \smol, 19 on \qwenfour, controls and \vanillakd{} at both precisions included) shares \NMCarrierOverlapRange{} of its own
top-40 with the untrained student's top-40 at every phase. Bottom-40's own top-40 holds almost none of the 40 heads it
supervised (\NMDBottomSmolSZeroSupervisedInOwnTopFortyPOne, \NMDBottomSmolSZeroSupervisedInOwnTopFortyPTwo{} and \NMDBottomSmolSZeroSupervisedInOwnTopFortyPTwoC{} at \Pone, \Ptwo{} and \Ptwoc{} on \smol{} seed 0, \NMDBottomSmolSOneSupervisedInOwnTopForty{} at every phase on seed 1 and \NMDBottomQwenSupervisedInOwnTopForty{} on \qwenfour) and shares as many heads with the untrained student's
top-40 as \mistill's does, whose supervised heads are that top-40 (at \Pone{}\,/\,\Ptwo{}\,/\,\Ptwoc: on \smol{} \NMDBottomSmolSZeroBaseOverlapPOne\,/\,\NMDBottomSmolSZeroBaseOverlapPTwo\,/\,\NMDBottomSmolSZeroBaseOverlapPTwoC{} for seed 0 and \NMDBottomSmolSOneBaseOverlapPOne\,/\,\NMDBottomSmolSOneBaseOverlapPTwo\,/\,\NMDBottomSmolSOneBaseOverlapPTwoC{} for seed 1, against \mistill{} seed 0's \NMDMistillSmolSZeroBaseOverlapPOne\,/\,\NMDMistillSmolSZeroBaseOverlapPTwo\,/\,\NMDMistillSmolSZeroBaseOverlapPTwoC; on \qwenfour{} \NMDBottomQwenBaseOverlapPOne\,/\,\NMDBottomQwenBaseOverlapPTwo\,/\,\NMDBottomQwenBaseOverlapPTwoC{} against \mistill{} seed 0's \NMDMistillQwenSZeroBaseOverlapPOne\,/\,\NMDMistillQwenSZeroBaseOverlapPTwo\,/\,\NMDMistillQwenSZeroBaseOverlapPTwoC); on
\smol{} its own top-8 at \Ptwoc{} are the same eight heads on both seeds (the untrained student's ranks \NMDBottomSmolOwnTopEightBaseRanks).
\smol{} $k=8$'s own top-40 holds all 8 of its supervised heads, which are the untrained student's top-8
(and \NMDKEightSmolBaseOverlapPOne\,/\,\NMDKEightSmolBaseOverlapPTwo\,/\,\NMDKEightSmolBaseOverlapPTwoC{} of the untrained student's top-40 at \Pone{}\,/\,\Ptwo{}\,/\,\Ptwoc). How strongly the recovery concentrates in those heads follows the size of the gain,
on both students: knock-in at the arm's own top-8 restores a mean of
\NPairSmolBottomSZeroVsMistillSZeroPTwoCKiEightMeanA{} and \NPairSmolBottomSOneVsMistillSZeroPTwoCKiEightMeanA{} for \smol{}
bottom-40's two seeds (recovery at $k=40$ \NOwnSmolBottomSZeroPTwoCKiForty{} and \NOwnSmolBottomSOnePTwoCKiForty), against
\NPairSmolBottomSZeroVsMistillSZeroPTwoCKiEightMeanB{} and \NPairSmolBottomSZeroVsMistillSOnePTwoCKiEightMeanB{} for
\mistill{} and \NPairSmolKEightSZeroVsMistillSZeroPTwoCKiEightMeanA{} for $k=8$; on \qwenfour, bottom-40 (recovery
\NOwnQwenBottomSZeroPTwoCKiForty) restores \NPairQwenBottomSZeroVsMistillSZeroPTwoCKiEightMeanA{} against \mistill's
\NPairQwenBottomSZeroVsMistillSZeroPTwoCKiEightMeanB, lower on \NPairQwenBottomSZeroVsMistillSZeroPTwoCKiEightLower{} of 150
notes. The supervised heads are an injection point, not the carriers.

\paragraph{The read side across the ablations.}
\label{app:abl-readside}
Across the arms of Sec.~\ref{sec:distill-mech}, M2 at \Ptwoc{} agreed with M4 on \qwenfour{} and the prefix share followed
recovery on both students; the ablations break both. On \qwenfour, RandBoth restores \NOwnQwenRandBothSZeroPTwoCKiForty{}
with an M2 lift of \NOwnQwenRandBothSZeroMTwoPTwoCLift{} (the untrained student \NOwnQwenBaseMTwoPTwoCLift, \mistill{}
\NOwnQwenMistillSZeroMTwoPTwoCLift) and a prefix share of \NOwnQwenRandBothSZeroPrefix, below the untrained student's
\NOwnQwenBasePrefix, and bottom-40 restores \NOwnQwenBottomSZeroPTwoCKiForty{} with \mistill's lift
(\NOwnQwenBottomSZeroMTwoPTwoCLift). On \smol, M2 runs opposite to recovery across the three structured pools ($k=8$
\NOwnSmolKEightSZeroMTwoPTwoCLift, \mistill{} \NOwnSmolMistillSZeroMTwoPTwoCLift, bottom-40 \NOwnSmolBottomSZeroMTwoPTwoCLift{}
at recoveries \NOwnSmolKEightSZeroPTwoCKiForty, \NOwnSmolMistillSZeroPTwoCKiForty, \NOwnSmolBottomSZeroPTwoCKiForty), and the
prefix share puts bottom-40 (\NOwnSmolBottomSZeroPrefix) at \randheads' level (\NOwnSmolRandHeadsSZeroPrefix), below $k=8$
(\NOwnSmolKEightSZeroPrefix) and \mistill{} (\NOwnSmolMistillSZeroPrefix). M4 at \Ptwoc{} is the readout the conclusions
rest on. M3 stays flat on \qwenfour{} and falls with distillation on \smol{} (bottom-40 \NOwnSmolBottomSZeroMThreeNearMiss,
$k=8$ \NOwnSmolKEightSZeroMThreeNearMiss, \mistill{} \NOwnSmolMistillSZeroMThreeNearMiss, \vanillakd{}
\NOwnSmolVanillaSZeroMThreeNearMiss{} and \NOwnSmolVanillaSOneMThreeNearMiss).

\paragraph{The instrument's floor.} The same checkpoint read on two GPU types gives the identical top-40 and per-note
differences of \NMDCrossHwPTwoCDelta, \NMDCrossHwPTwoDelta{} and \NMDCrossHwPOneDelta{} at \Ptwoc, \Ptwo{} and \Pone (Appendix~\ref{app:res-checks}), about a third of the smallest per-note
difference read as an effect here (\Ptwo-only against \mistill{} seed 0, \NPairQwenPTwoOnlySZeroVsMistillSZeroPTwoCDelta).

\subsection{How the runs were done, and what they do not control for}
\label{app:abl-runs}
The \qwenfour{} ablation arms trained on rented 4-GPU H100 nodes, one arm per GPU (the step is largely CPU-bound, so co-located arms
slow each other), except $k=8$ (a GH200) and $k=100$ (a B200); their own-heads passes ran on the same GPUs or on a
cluster. The \smol{} A$'$ arms trained on rented RTX PRO 6000 and H200 instances or on a cluster RTX PRO 6000. Task
evaluations ran on cluster RTX PRO 6000 cards, except \Pone-only, \Ptwo-only and attention-only (a rented evaluation
instance) and \smol{} $k=8$ (its H200, with a different vLLM build); generation is stochastic in any case, and the
instrument's cross-hardware floor is above. Each run's \texttt{config.json} records its one differing flag.

\Pone-only and \Ptwo-only are $\lambda$-zero arms ($\lambda_{\mathrm{P2}}=0$ and $\lambda_{\mathrm{P1}}=0$): the trainer's
literal single-phase path took two to three times the memory of the two-phase objective, for a cause not found, and a zero weight contributes exactly zero gradient. A literal \Ptwo-only run meant as
a cross-check was stopped early, so the equivalence rests on that argument, not on a run. The term arms act on both
phases. One seed per cell, except \qwenfour{} \randheads{} and \smol{} bottom-40 (seeds 0 and 1). Three faults are
disclosed rather than diagnosed: the attribution passes' host memory grew on a B200 node far beyond what the same code uses on an RTX PRO 6000
node (resumed per note, nothing lost); the single-phase path's memory growth; and one
evaluation (\qwenfour{} bottom-40) that skipped \las{} and M1 while exiting cleanly, resubmitted.

\paragraph{Not run.} The temperature of the student pool ($\tau\times0.5$ and $\times2$), the symmetric-window cell,
per-phase term arms, \smol{} \Pone-only and \Ptwo-only, a \smol{} attention-only cell at $\lambda=0.03$, re-ranking the
student's pool during training, and baseline tuning were planned and not run; the tier-E controls (event-weighted
\vanillakd, uniform attention, no window) and the fixed-pool \randheads{} seeds trained on cluster RTX PRO 6000 and H200
cards and their own-heads passes on the same or on rented H100s.

\subsection{Training-set size}
\label{app:abl-curve}
\begin{figure}[!htbp]
  \centering
  \includegraphics[width=\linewidth]{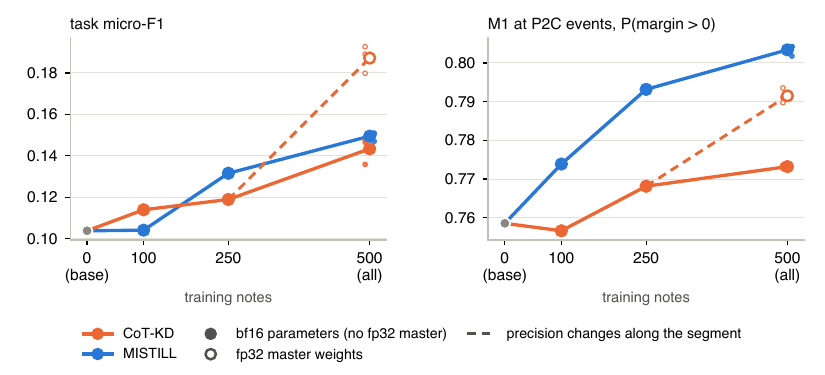}
  \caption{Learning curve on \qwenfour{} (seed 0 at 100 and 250 notes; seed means at all 493): task \microf{} (left)
  and M1 at \Ptwoc{} events (right). Filled markers train bf16 weights, open markers fp32 master weights; the dashed
  segment is where the precision changes. With bf16 weights the two objectives tie on the task at every size, and
  \mistill{} leads on its supervised readout. \resid{R-D8}\resid{R-D17}}
  \label{fig:curve}
\end{figure}

The learning curve (Figure~\ref{fig:curve}) trains \vanillakd{} and \mistill{} on 100 and 250 notes (seed 0, both with bf16
weights) and uses the tier-1 rows at 493 notes (Table~\ref{tab:app-curve}). With bf16 weights the two tie on the task
at every size: at 100 notes \mistill{} is higher on \NCurvePairOneHundredMistillVsVanillaTaskHigher{} notes and lower on
\NCurvePairOneHundredMistillVsVanillaTaskLower{} ($p$ \NCurvePairOneHundredMistillVsVanillaTaskP), at 250 notes
\NCurvePairTwoFiftyMistillVsVanillaTaskHigher{} and \NCurvePairTwoFiftyMistillVsVanillaTaskLower{} ($p$
\NCurvePairTwoFiftyMistillVsVanillaTaskP), and at 493 notes against bf16 \vanillakd{}
\NCurvePairFullMistillVsVanillaBfTaskHigher{} and \NCurvePairFullMistillVsVanillaBfTaskLower{} note-pairs over three seed
pairs ($p$ \NCurvePairFullMistillVsVanillaBfTaskP; pooled note-pairs, descriptive). The one point where the arms separate, fp32 \vanillakd{} at 493 notes
(seed 0: \NCurveVanillaFullSZeroMicroF{} against \mistill's \NCurveMistillFullSZeroMicroF), is the one that mixes
precisions. \mistill{} leads on M1 at \Ptwoc{} at every size, by a mean of
\NCurvePairOneHundredMistillVsVanillaMOnePTwoCMean, \NCurvePairTwoFiftyMistillVsVanillaMOnePTwoCMean{} and, at 493 notes,
\NCurvePairFullMistillVsVanillaBfMOnePTwoCMean{} against bf16 \vanillakd{} (\NCurvePairFullMistillVsVanillaMOnePTwoCMean{}
against fp32 \vanillakd, which reaches \NQwenVanillaMOnePTwoC{} on its own). The curve was run to decide whether to grow the
corpus beyond 500 notes; it was not grown.

\begin{table}[!htbp]
  \centering\small
  \caption{Learning curve on \qwenfour{} (150 test notes; seed 0 at 100 and 250 notes, three seeds on the 500-note split, whose 493 notes with a valid teacher trace are the tier-1 training set, with the
  between-seed sd below). Paired: notes (or note-pairs, pooled over seed pairs and counted as if independent: descriptive) on which \mistill{} is higher\,/\,lower,
  exact sign test. The second row at the full split is precision-matched. \resid{R-D8}\resid{R-D17}}
  \label{tab:app-curve}
  \providecommand{\gensd}[2]{\begin{tabular}[t]{@{}c@{}}#1\\[-2pt]{\scriptsize$\pm$#2}\end{tabular}}
\providecommand{\genstack}[2]{\begin{tabular}[t]{@{}c@{}}#1\\[-2pt]{\scriptsize #2}\end{tabular}}
\providecommand{\genstackl}[2]{\begin{tabular}[t]{@{}l@{}}#1\\[-2pt]{\scriptsize #2}\end{tabular}}
\providecommand{\genhead}[2]{\begin{tabular}[b]{@{}c@{}}#1\\#2\end{tabular}}
\providecommand{\genstackiii}[3]{\begin{tabular}[t]{@{}c@{}}#1\\[-2pt]{\scriptsize #2}\\[-3pt]{\scriptsize #3}\end{tabular}}
\footnotesize\setlength{\tabcolsep}{3.5pt}
\begin{adjustbox}{max width=\linewidth}
\begin{tabular}{@{}l l cc cc c c@{}}
\toprule
 & & \multicolumn{2}{c}{\microf} & \multicolumn{2}{c}{M1 \Ptwoc} & \multicolumn{2}{c}{paired, per note} \\
\cmidrule(lr){3-4}\cmidrule(lr){5-6}\cmidrule(lr){7-8}
\genhead{training}{notes} & \genhead{precision}{(CoT-KD / \mistill)} & CoT-KD & \mistill & CoT-KD & \mistill & task & M1 \Ptwoc \\
\midrule
0 (base) & --- & \multicolumn{2}{c}{0.104} & \multicolumn{2}{c}{0.759} & --- & --- \\
100 & bf16 / bf16 & 0.114 & 0.104 & 0.757 & 0.774 & \genstack{58/59}{$p$ 1.00} & \genstack{102/21}{$p$ \ensuremath{<10^{-4}}} \\
250 & bf16 / bf16 & 0.119 & 0.132 & 0.768 & 0.793 & \genstack{67/54}{$p$ 0.28} & \genstack{111/15}{$p$ \ensuremath{<10^{-4}}} \\
500, 3 seeds & fp32 / bf16 & \gensd{0.187}{.007} & \gensd{0.149}{.002} & \gensd{0.791}{.002} & \gensd{0.803}{.001} & \genstack{167/234}{$p$ 0.001} & \genstack{241/137}{$p$ \ensuremath{<10^{-4}}} \\
500, 3 seeds & bf16 / bf16 & \gensd{0.143}{.007} & \gensd{0.149}{.002} & \gensd{0.773}{.000} & \gensd{0.803}{.001} & \genstack{205/184}{$p$ 0.31} & \genstack{348/29}{$p$ \ensuremath{<10^{-4}}} \\
\bottomrule
\end{tabular}
\end{adjustbox}

\end{table}

\section{The evidence-flip test: is the recovered pathway used?}
\label{app:evflip}
\supports{Sec.~\ref{sec:limitations}; R-D33, R-D34, R-D35}
Whether the student's recovered heads are \emph{used} to read the clinical evidence is a different question from whether
they recover the teacher's decision, and the preceding appendices do not answer it. A pre-registered evidence-flip test
asks it directly; its rules were locked before the confirmation cohort was drawn, with dated pre-generation amendments,
and it is reported here whatever its outcome.

\subsection{Design and amendments}
\label{app:ef-design}
A case is a patient's note with an emitted gold code $A$ supported by evidence tokens in the note and an Excludes1
partner $B$ (the pairing is a proposal aid, not a claim of clinical exclusivity). The \emph{flip} rewrites the evidence so
that the note supports $B$ instead; the flips were drafted with generative-AI assistance and retained after the author's
clinical review, and the approval ledger (accept, or exclude with a reason code) is the author's adjudication, frozen before
any generation. Two controls accompany the flip, an irrelevant edit of matched length and an order permutation. Each checkpoint answered every case $N=4$ times under each condition with the evaluation's decoding. An
answer scores $v=+1$ if it emits $A$ and not $B$, $-1$ if $B$ and not $A$, and $0$ otherwise (an output that hits the
generation cap or leaves its reasoning block open is invalid, scored $0$ and counted); the directional sensitivity of a
checkpoint on a case is $D=[\bar v_{\mathrm{original}}-\bar v_{\mathrm{flipped}}]/2$. The case, one per patient, is the
unit: intervals are case-clustered percentile bootstraps and tests exact sign tests. $D$ is a directional-sensitivity
score, not counterfactual accuracy or a fraction of correct switches. The locked primary comparison is bf16 \mistill{}
seeds 0--2 minus bf16 \vanillakd{} seeds 0--2, the three seed-matched pair differences averaged within each case, with a
pre-registered minimum effect $\delta=\NEFDelta$ in $D$; the secondary is the fp32 pair at $\lambda=0.01$; the teacher and
the matched-selection \randheads{} pick (Appendix~\ref{app:abl-matched}) are descriptive. A development pilot on a disjoint
cohort fixed the procedure, the decode regime and the sample size; it was calibration and an operational check of the
pipeline, not a validation of a general semantic-use assay. The amendments after the lock, each dated before any
confirmatory output, were a scoring correction, the authoring budget $K=60\to80$ (adaptive on the authoring yield), and the
sensitivity analyses below.

\subsection{Population and accounting}
\label{app:ef-population}
\NEFCohortDrawn{} patients were drawn disjoint from every development cohort (one note each; the shortlist retriever was
retrained without them). The teacher's eligibility screen and the Excludes1 pairing made \NEFCohortCandidateAttempts{}
candidate $(A, B)$ attempts over their screen traces and retained \NEFCohortProposals{} proposals, one per patient;
\NEFCohortPropRejected{} attempts were rejected (\NEFCohortPropNoAEvidence{} with no evidence token for the emitted code,
\NEFCohortPropNoPartner{} with no Excludes1 partner, \NEFCohortPropNoBWords{} with no words for $B$ --- attempt counts, since a
failed attempt does not exclude its patient from a later valid proposal), and \NEFCohortNoProposal{} patients retained no
proposal. The first \NEFCohortReviewed{} proposals were reviewed in order: \NEFCohortAccepted{} flips were retained after
clinical review and \NEFCohortExcluded{} were excluded with reason codes (\NEFCohortRuleThree{} where no coherent flip could be
written, \NEFCohortRuleOne{} for the pair, \NEFCohortRuleTwo{} for the evidence);
\NEFCohortNotReached{} proposals were never reviewed. The screen traces existed and were withheld during authoring, and the
flips were frozen by hash before any generation. The population is therefore teacher-success-conditioned and
author-approved: \NEFCases{} cases from \NEFPatients{} patients. \resid{R-D35}

\subsection{The behavioural half}
\label{app:ef-behav}
Table~\ref{tab:app-ef-behav} gives every checkpoint. The teacher shows positive directional sensitivity with incomplete
switching: $D$ = \NEFTeacherD{} [\NEFTeacherDLo, \NEFTeacherDHi], it answers $B$ in \NEFTeacherFollowFlip{} of the flipped
outputs and succeeds on both sides in \NEFTeacherPaired{} of the sample pairs. Every student is lower (bf16 \mistill{}
\NEFMistillSZeroD{} / \NEFMistillSOneD{} / \NEFMistillSTwoD, bf16 \vanillakd{} \NEFVanillaBfSZeroD{} / \NEFVanillaBfSOneD{} /
\NEFVanillaBfSTwoD; the fp32 arms and the \randheads{} pick in the table), switches rarely (bf16 \mistill{} follows the
flipped evidence in \NEFMistillSZeroFollowFlip--\NEFMistillSOneFollowFlip{} of its flipped outputs, paired success
\NEFMistillSZeroPaired), and its $D$ comes mostly from losing $A$ rather than adopting $B$ (the decomposition column). The
locked primary is not detected: $\Delta D$ = \NEFPrimaryMean{} [\NEFPrimaryLo, \NEFPrimaryHi] over \NEFPrimaryN{} cases
(\NEFPrimaryHigher{} higher, \NEFPrimaryLower{} lower, \NEFPrimaryTies{} tied; sign $p$ \NEFPrimaryP), below the
pre-registered $\delta$; the secondary fp32 pair gives \NEFSecondaryMean{} [\NEFSecondaryLo, \NEFSecondaryHi]. The SD of the
pooled per-case difference is \NEFPrimarySD, so the approximate detectable $\delta$ at this $n$ is \NEFPrimaryDetectable{}
(80\% power, two-sided 0.05, a normal approximation on the observed variance): a precision statement beside the observed
interval, not an exact power guarantee for the locked bootstrap and sign-test decision. The controls are reported as two
statistics, the agreement of each control condition's $v$ with the original condition's $v$ at the same sample index
(matching wrong answers and matching zeros count) and the rate of the expected $A$-only answer under the control; on the
\NEFCtrlN{} cases without the one case whose irrelevant edit fell one word short of its matched length, the agreement moves
by at most \NEFCtrlMaxAgreeShift{} and the expected rate by at most \NEFCtrlMaxExpShift, and $D$ is untouched by
construction. A category-level secondary (the three-character category of $A$ against $B$, on the cases whose categories
differ) tells the same story as the exact-code primary. \resid{R-D33}

\begin{table}[!htbp]
  \centering\small
  \caption{The evidence-flip test, behavioural half: every locked checkpoint on the \NEFCases{} approved cases (four
  samples per condition; the generation path of the evaluations). $D$ and its case-clustered interval; $s$ = [Pr($v_{\mathrm{original}}=+1$) + Pr($v_{\mathrm{flipped}}=-1$)]/2, the mean of the two follow rates;
  the follow rates ($A$-only under the original note, $B$-only under the flip); paired success (both at the same sample
  index); the two control statistics (agreement with the original $v$; expected $A$-only rate); invalid outputs; the
  decomposition of $D$ and the category-level secondary. Bottom: the locked primary and the secondary seed-pair groups.
  \resid{R-D33}}
  \label{tab:app-ef-behav}
  \providecommand{\gensd}[2]{\begin{tabular}[t]{@{}c@{}}#1\\[-2pt]{\scriptsize$\pm$#2}\end{tabular}}
\providecommand{\genstack}[2]{\begin{tabular}[t]{@{}c@{}}#1\\[-2pt]{\scriptsize #2}\end{tabular}}
\providecommand{\genstackl}[2]{\begin{tabular}[t]{@{}l@{}}#1\\[-2pt]{\scriptsize #2}\end{tabular}}
\providecommand{\genhead}[2]{\begin{tabular}[b]{@{}c@{}}#1\\#2\end{tabular}}
\providecommand{\genstackiii}[3]{\begin{tabular}[t]{@{}c@{}}#1\\[-2pt]{\scriptsize #2}\\[-3pt]{\scriptsize #3}\end{tabular}}
\scriptsize\setlength{\tabcolsep}{2.5pt}
\begin{adjustbox}{max width=\linewidth}
\begin{tabular}{@{}l l c c c c c c l@{}}
\toprule
Arm & $D$ [95\% CI] & $s$ & \genhead{follows}{orig.\ / flip} & \genhead{paired}{success} & \genhead{agree with}{orig.\ $v$: irr.\ / order} & \genhead{expected A-only}{irr.\ / order} & \genhead{invalid}{orig.\ / flip} & \genhead{$D$ = loss of A-only + gain of B-only;}{category-level $D_{\mathrm{cat}}$} \\
\midrule
\multicolumn{9}{@{}l}{\textit{Per arm: 33 cases $\times$ 4 samples per condition; case-clustered 95\% intervals}} \\
Teacher (\teacher) & \pci{+0.511}{+0.398}{+0.621} & 0.568 & 0.689 / 0.447 & 0.333 & 0.667 / 0.652 & 0.644 / 0.674 & 0.030 / 0.023 & 0.292 + 0.220; $D_{\mathrm{cat}}$ \pci{+0.593}{+0.472}{+0.706} \\
\mistill{} (bf16) s0 & \pci{+0.242}{+0.178}{+0.307} & 0.322 & 0.561 / 0.083 & 0.053 & 0.561 / 0.629 & 0.508 / 0.485 & 0.091 / 0.114 & 0.201 + 0.042; $D_{\mathrm{cat}}$ \pci{+0.226}{+0.149}{+0.302} \\
\quad s1 & \pci{+0.223}{+0.152}{+0.299} & 0.326 & 0.553 / 0.098 & 0.053 & 0.780 / 0.697 & 0.606 / 0.515 & 0.091 / 0.098 & 0.189 + 0.034; $D_{\mathrm{cat}}$ \pci{+0.246}{+0.153}{+0.347} \\
\quad s2 & \pci{+0.178}{+0.091}{+0.261} & 0.288 & 0.492 / 0.083 & 0.053 & 0.674 / 0.644 & 0.477 / 0.477 & 0.076 / 0.061 & 0.148 + 0.030; $D_{\mathrm{cat}}$ \pci{+0.145}{+0.036}{+0.258} \\
\vanillakd{} (bf16) s0 & \pci{+0.205}{+0.129}{+0.280} & 0.303 & 0.500 / 0.106 & 0.045 & 0.727 / 0.629 & 0.455 / 0.341 & 0.068 / 0.106 & 0.155 + 0.049; $D_{\mathrm{cat}}$ \pci{+0.230}{+0.137}{+0.335} \\
\quad s1 & \pci{+0.174}{+0.091}{+0.258} & 0.273 & 0.455 / 0.091 & 0.045 & 0.629 / 0.720 & 0.538 / 0.379 & 0.136 / 0.144 & 0.133 + 0.042; $D_{\mathrm{cat}}$ \pci{+0.258}{+0.165}{+0.359} \\
\quad s2 & \pci{+0.208}{+0.121}{+0.292} & 0.303 & 0.500 / 0.106 & 0.061 & 0.667 / 0.636 & 0.439 / 0.356 & 0.091 / 0.106 & 0.163 + 0.045; $D_{\mathrm{cat}}$ \pci{+0.262}{+0.145}{+0.383} \\
\mistill{} (fp32, $\lambda$ 0.01) s0 & \pci{+0.269}{+0.197}{+0.341} & 0.348 & 0.568 / 0.129 & 0.091 & 0.659 / 0.652 & 0.591 / 0.485 & 0.144 / 0.152 & 0.205 + 0.064; $D_{\mathrm{cat}}$ \pci{+0.266}{+0.173}{+0.363} \\
\quad s1 & \pci{+0.273}{+0.201}{+0.345} & 0.360 & 0.606 / 0.114 & 0.045 & 0.636 / 0.591 & 0.545 / 0.530 & 0.121 / 0.159 & 0.231 + 0.042; $D_{\mathrm{cat}}$ \pci{+0.298}{+0.210}{+0.395} \\
\quad s2 & \pci{+0.265}{+0.182}{+0.345} & 0.371 & 0.576 / 0.167 & 0.106 & 0.629 / 0.583 & 0.553 / 0.462 & 0.159 / 0.144 & 0.193 + 0.072; $D_{\mathrm{cat}}$ \pci{+0.302}{+0.198}{+0.407} \\
\vanillakd{} (fp32) s0 & \pci{+0.258}{+0.167}{+0.348} & 0.356 & 0.591 / 0.121 & 0.091 & 0.682 / 0.644 & 0.606 / 0.485 & 0.167 / 0.212 & 0.197 + 0.061; $D_{\mathrm{cat}}$ \pci{+0.327}{+0.206}{+0.444} \\
\quad s1 & \pci{+0.280}{+0.193}{+0.375} & 0.364 & 0.591 / 0.136 & 0.091 & 0.667 / 0.644 & 0.561 / 0.492 & 0.152 / 0.182 & 0.216 + 0.064; $D_{\mathrm{cat}}$ \pci{+0.254}{+0.133}{+0.379} \\
\quad s2 & \pci{+0.288}{+0.208}{+0.364} & 0.371 & 0.576 / 0.167 & 0.106 & 0.667 / 0.538 & 0.583 / 0.523 & 0.144 / 0.258 & 0.216 + 0.072; $D_{\mathrm{cat}}$ \pci{+0.319}{+0.230}{+0.403} \\
\randheads{} (pool 0, $\lambda$ 0.3) s0 & \pci{+0.170}{+0.091}{+0.254} & 0.280 & 0.439 / 0.121 & 0.053 & 0.674 / 0.750 & 0.386 / 0.348 & 0.106 / 0.098 & 0.117 + 0.053; $D_{\mathrm{cat}}$ \pci{+0.226}{+0.125}{+0.335} \\
\midrule
\multicolumn{9}{@{}l}{\textit{Seed-matched pairs pooled per case (\mistill{} $-$ \vanillakd): mean $\Delta D$ [95\% CI]; cases higher / lower / tied; sign $p$; SD; approximate detectable $\delta$}} \\
locked primary: bf16 s0--s2 pairs & \multicolumn{8}{l}{+0.019 [\ensuremath{-}0.023, +0.061]; 15 / 13 / 5; $p$ 0.85; SD 0.125; detectable $\delta$ 0.061 (pre-registered $\delta$ 0.10); per pair +0.038 [\ensuremath{-}0.034, +0.114], +0.049 [\ensuremath{-}0.027, +0.114], \ensuremath{-}0.030 [\ensuremath{-}0.102, +0.045]} \\
secondary: fp32 ($\lambda$ 0.01) s0--s2 pairs & \multicolumn{8}{l}{\ensuremath{-}0.006 [\ensuremath{-}0.057, +0.042]; 14 / 15 / 4; $p$ 1.00; SD 0.146; detectable $\delta$ 0.071 (pre-registered $\delta$ 0.10); per pair +0.011 [\ensuremath{-}0.053, +0.076], \ensuremath{-}0.008 [\ensuremath{-}0.106, +0.083], \ensuremath{-}0.023 [\ensuremath{-}0.110, +0.064]} \\
\bottomrule
\end{tabular}
\end{adjustbox}

\end{table}

\subsection{The causal half}
\label{app:ef-causal}
\emph{What is intervened on.} Under the HF generation path the prompt is prefilled clean and, during generation, a key mask
removes the nominated heads' access to the evidence-key positions (\emph{target}) or to token-matched non-evidence positions
on the same heads (\emph{sham}). Evidence already propagated into other cached positions during the prefill remains
accessible, so the test covers the direct reading route of the nominated heads, not every use of the evidence or of those
heads. The nominated sets are the teacher's named pair \head{30}{11} + \head{38}{39} and its own top-40, and each bf16
seed-0 student's own top-40; the criterion is the loss of directional sensitivity $\Delta D = D_{\mathrm{clean}} -
D_{\mathrm{masked}}$, target against sham, and the training-benefit interaction $L = G_{\mathrm{sham}} - G_{\mathrm{target}}$
with $G = D_{\mistill} - D_{\vanillakd}$ on the own top-40. No random-head or component-necessity control ran in the
confirmation, so the sham tests position specificity on the same heads and does not establish the unique importance of the
ranked heads; the causal half covers the teacher and the bf16 seed-0 students only, with no fp32 or three-seed causal
replication.

\emph{Result} (Table~\ref{tab:app-ef-causal}). No selective effect is detected in any checkpoint: target minus sham is
\NEFCausalTeacherNamedSelMean{} [\NEFCausalTeacherNamedSelLo, \NEFCausalTeacherNamedSelHi] for the teacher's named pair,
\NEFCausalTeacherTopFortySelMean{} [\NEFCausalTeacherTopFortySelLo, \NEFCausalTeacherTopFortySelHi] for its own top-40,
\NEFCausalMistillSZeroTopFortySelMean{} [\NEFCausalMistillSZeroTopFortySelLo, \NEFCausalMistillSZeroTopFortySelHi] for
\mistill{} and \NEFCausalVanillaBfSZeroTopFortySelMean{} [\NEFCausalVanillaBfSZeroTopFortySelLo,
\NEFCausalVanillaBfSZeroTopFortySelHi] for \vanillakd, and $L$ = \NEFLMean{} [\NEFLLo, \NEFLHi] over \NEFLN{} cases. $L$ is
the difference of the two student contrasts and cannot by itself establish a null in either student or the teacher; the
per-arm rows stay beside it. Its SD is \NEFPrecisionLSD, an approximate detectable $L$ of \NEFPrecisionLDetectable; the
per-arm contrasts have detectable effects of \NEFPrecisionTargetShamCotkdDetectable{} to
\NEFPrecisionTargetShamTeacherDetectable. Observed, not tested: on \mistill's own top-40 both tested masks lower $D$
(sham \NEFCausalMistillSZeroTopFortyShamMean, target \NEFCausalMistillSZeroTopFortyTargetMean), a non-specific sensitivity
to either mask on those heads; on the teacher both masks slightly raise it. \resid{R-D34}

\begin{table}[!htbp]
  \centering\small
  \caption{The evidence-flip test, causal half: the HF generation path, four samples per condition, the decode-regime key
  mask on the nominated heads after a clean prefill; $\Delta D = D_{\mathrm{clean}} - D_{\mathrm{masked}}$ with
  case-clustered intervals; target minus sham on the \NEFCases{} cases and on the \NEFSensN{}-case sensitivity; the
  training-benefit interaction $L$ and its components; the execution-path offset of each checkpoint. \resid{R-D34}}
  \label{tab:app-ef-causal}
  \providecommand{\gensd}[2]{\begin{tabular}[t]{@{}c@{}}#1\\[-2pt]{\scriptsize$\pm$#2}\end{tabular}}
\providecommand{\genstack}[2]{\begin{tabular}[t]{@{}c@{}}#1\\[-2pt]{\scriptsize #2}\end{tabular}}
\providecommand{\genstackl}[2]{\begin{tabular}[t]{@{}l@{}}#1\\[-2pt]{\scriptsize #2}\end{tabular}}
\providecommand{\genhead}[2]{\begin{tabular}[b]{@{}c@{}}#1\\#2\end{tabular}}
\providecommand{\genstackiii}[3]{\begin{tabular}[t]{@{}c@{}}#1\\[-2pt]{\scriptsize #2}\\[-3pt]{\scriptsize #3}\end{tabular}}
\scriptsize\setlength{\tabcolsep}{3pt}
\begin{adjustbox}{max width=\linewidth}
\begin{tabular}{@{}l l c c c c c@{}}
\toprule
Checkpoint & nominated set & HF clean $D$ [95\% CI] & sham $\Delta D$ & target $\Delta D$ & target $-$ sham & \genhead{target $-$ sham,}{29 cases} \\
\midrule
\multicolumn{7}{@{}l}{\textit{HF \texttt{generate}, decode regime after a clean prefill; $\Delta D = D_{\mathrm{clean}} - D_{\mathrm{masked}}$; 33 cases (29 in the sensitivity); case-clustered 95\% intervals}} \\
Teacher (\teacher) & named pair \head{30}{11} + \head{38}{39} & \pci{+0.413}{+0.277}{+0.538} & \ensuremath{-}0.064 [\ensuremath{-}0.140, +0.008] & \ensuremath{-}0.087 [\ensuremath{-}0.163, \ensuremath{-}0.011] & \textbf{\ensuremath{-}0.023} [\ensuremath{-}0.091, +0.045] & \ensuremath{-}0.026 [\ensuremath{-}0.099, +0.048] \\
 & own top-40 (HD-002) & \pci{+0.413}{+0.277}{+0.538} & \ensuremath{-}0.049 [\ensuremath{-}0.114, +0.015] & \ensuremath{-}0.019 [\ensuremath{-}0.102, +0.061] & \textbf{+0.030} [\ensuremath{-}0.049, +0.106] & +0.047 [\ensuremath{-}0.034, +0.134] \\
\mistill{} (bf16) s0 & own top-40 (HD-S) & \pci{+0.254}{+0.167}{+0.341} & +0.095 [+0.019, +0.170] & +0.098 [+0.019, +0.182] & \textbf{+0.004} [\ensuremath{-}0.068, +0.076] & +0.000 [\ensuremath{-}0.078, +0.082] \\
\vanillakd{} (bf16) s0 & own top-40 (HD-S) & \pci{+0.201}{+0.140}{+0.261} & +0.011 [\ensuremath{-}0.045, +0.072] & +0.053 [\ensuremath{-}0.011, +0.114] & \textbf{+0.042} [\ensuremath{-}0.011, +0.095] & +0.034 [\ensuremath{-}0.026, +0.091] \\
\midrule
\multicolumn{7}{@{}l}{\textit{The training-benefit interaction on the own top-40 ($G = D_{\mistill} - D_{\vanillakd}$, bf16 seed 0)}} \\
$G$ under clean / sham / target &  & \multicolumn{5}{l}{+0.053 [\ensuremath{-}0.011, +0.117] / \ensuremath{-}0.030 [\ensuremath{-}0.102, +0.042] / +0.008 [\ensuremath{-}0.057, +0.072]} \\
$L = G_{\mathrm{sham}} - G_{\mathrm{target}}$ &  & \multicolumn{5}{l}{\textbf{\ensuremath{-}0.038} [\ensuremath{-}0.121, +0.045] (9 / 16 cases, $p$ 0.23; SD 0.251, detectable 0.122); 29-case sensitivity \ensuremath{-}0.034 [\ensuremath{-}0.129, +0.060]} \\
\midrule
\multicolumn{7}{@{}l}{\textit{Execution-path offset: $D$(vLLM path) $-$ $D$(HF clean), paired per case}} \\
teacher / \mistill{} s0 / \vanillakd{} s0 &  & \multicolumn{5}{l}{+0.098 [+0.019, +0.174] ($p$ 0.03) / \ensuremath{-}0.011 [\ensuremath{-}0.102, +0.083] / +0.004 [\ensuremath{-}0.068, +0.080]} \\
\bottomrule
\end{tabular}
\end{adjustbox}

\end{table}

\subsection{Sensitivities, the execution path, and the limits of the test}
\label{app:ef-limits}
\emph{Pre-generation sensitivity.} Three accepted cases are insertion-only edits whose original-condition target and sham
masks are empty (on that side those arms equal the clean condition by construction; the flipped-side masks act), and one
case's flipped-side target and sham masks differ in length (21 against 16 tokens), the one exception to ``token-matched'';
these facts are recorded in the frozen approval ledger. Over the \NEFSensN{} cases without them, $L$ = \NEFSensLMean{}
[\NEFSensLLo, \NEFSensLHi] and the per-arm contrasts are the last column of Table~\ref{tab:app-ef-causal}: the qualitative
non-detection conclusion is unchanged while the estimates and intervals move. One further case's irrelevant edit fell one
word short of its matched length; its control sensitivity is reported in Appendix~\ref{app:ef-behav} and does not enter
$D$. \emph{Execution path.} The teacher's $D$ is \NEFTeacherD{} under the generation path of the behavioural half and
\NEFCausalTeacherCleanMean{} under the clean HF path, a paired difference of \NEFBackendTeacherMean{}
[\NEFBackendTeacherLo, \NEFBackendTeacherHi]; no student offset is detected (\NEFBackendMistillSZeroMean{}
[\NEFBackendMistillSZeroLo, \NEFBackendMistillSZeroHi] and \NEFBackendVanillaBfSZeroMean{} [\NEFBackendVanillaBfSZeroLo,
\NEFBackendVanillaBfSZeroHi]). Every causal contrast is HF against HF, so this level is not subtracted into $\Delta D$ or
$L$; the causal non-detection is nevertheless specific to the HF execution setup, and because hardware and sampling paths
differ as well, the offset cannot be attributed to the software backend alone. The teacher ran with YaRN $\times4$ under
both paths and the students with $\times2$; the eligibility screen ran the teacher at $\times2$. \emph{Limits.} A
teacher-success-conditioned, author-approved population of \NEFCases{} cases; the decode-time direct reading route; the
teacher and the bf16 seed-0 students for the causal half; both tested masks only. These tests do not establish improved
evidence-dependent decisions through the tested routes, and ``not detected'' is never ``does not transfer'': the paper's
positive contribution remains the increased own-heads causal recoverability, not semantic pathway transfer.

\section{A public-task extension: answer-emission distillation on MuSiQue-Ans}
\label{app:musique}
\supports{Sec.~\ref{sec:limitations}; R-D37, R-D38, R-D39, R-D40}
This appendix reports a preregistered extension of the recipe to a public, passage-grounded multi-hop QA task, run under a
design locked before any reserved output (the lock, its artifact manifest and every reader are in
the released code). It tests whether the recovery advantage of Section~\ref{sec:distill-main} reappears on a second task under a declared
adaptation; it is reported whatever its outcome, with the comparison against the matched random pool beside the primary one.

\paragraph{Adaptation and what differs from the clinical setting.} The task is MuSiQue-Ans \citep{trivedi_musique_2022} (2--4-hop questions over 20 titled passages;
19,938 train, 2,417 labelled dev). The teacher is Qwen3-14B, the student Qwen3-4B, both reasoning in the same format as before. The event is
the \emph{final answer line}: a \Ptwo{} answer-emission event at every token of the emitted answer at which a candidate answer first
diverges from the alternatives named in the passages or the reasoning (the task's candidate inventory). The teacher's final-line events on
its own traces define the supervision targets; the reported recovery is evaluated teacher-forced on the teacher's ranking-set traces, as in
the clinical setting (not on the students' freely generated traces). The clinical headline concerns
contrastive \Ptwoc{} recovery; a contrastive event was specified for this task but validated in the development audit on only 6 of 25 sampled
instances (the answer-emission event: 24 of 25), so the locked design made answer emission primary and treated contrastive and
reasoning-site events as diagnostics only. This is therefore a declared extension of the recipe, not a matched replication that isolates the
task as the only difference. Teacher traces were generated once (sampling, seed = question index, cap 16,384, YaRN off); an abstaining final
answer contributes no mechanism event by rule.

\paragraph{Locked design.} Splits by base-question group (derivatives of one source question stay together): \NMQTrain{} training questions
(\NMQTrainGroups{} groups, a group-proportional prefix of the public train split), \NMQRankQ{} ranking questions (\NMQRankGroups{} groups,
disjoint), \NMQVal{} selection questions, and two reserved slices of the labelled dev split: \NMQEval{} evaluation questions (\NMQEvalGroups{}
groups) and the base questions of the evidence pairs. Head sets are located, not assumed: the teacher's and the untrained student's final-line
\Ptwo{} rankings on the ranking questions pass the attribution's null split-half calibration, and their top-40 sets are necessary (knock-out
$+0.72$ / $+0.90$ against $+0.02$ / $+0.03$ for random heads) and partly sufficient (knock-in $+0.35$ / $+0.28$ against $+0.01$) before any
training. Arms: \vanillakd{}, \mistill{} (pool = the untrained student's own final-line top-40 with the attribution's weight profile) and one
frozen \randheads{} pool (the same weights on heads redrawn with a fixed seed, unrestricted layers; identities serialised and enforced by the
trainer), on the bf16 recipe of Appendix~\ref{app:objective} (lr $10^{-5}$, global batch 8, two epochs, loss phase \Ptwo{} only), every arm
on the same \NMQTrain{} questions in the same order; the \NMQNoEvents{} questions without a final-line event receive the CoT loss only in the
mechanism arms. Selection on all \NMQVal{} selection questions by a rule fixed at the lock: among cells with a valid-answer rate of at least
0.95, the smallest $\lambda$ within 0.017 answer-F1 of the best (a noise-informed tie band fixed from five base-student decoding repeats), one outward
step if the pick sits on the grid's edge; $\lambda \in \{0.03, 0.1, 0.3\}$ at seed 0, then seeds 1 and 2 at the pick (\NMQCells{} cells in all,
the 13 planned and the two outward cells). Both families picked $\lambda = \NMQLambda$ (every \mistill{} cell within the band; the stronger
\randheads{} cells below the validity floor through runaway generations). The fixed sequence of readings: (1) own-heads final-line \Ptwo{}
recovery at $k=40$, \mistill{} $-$ \vanillakd{} on the questions in common support, lower 95\,\% bound $>0$; (2) answer macro-F1 on the
reserved questions, lower bound $>-$\NMQTaskMargin{} (an explicit tolerated loss); (3) paired correctness on the evidence pairs, lower bound
$>0$; (4) a causal route gated on the teacher. Intervals are percentile bootstraps over base-question groups (10,000 draws), conditional on the observed checkpoints; every reported
number carries both counts. Declared magnitude bounds, $\pm$0.03 (recovery) and $\pm$0.10 (behaviour), are reported only as ``within''.

\paragraph{Evidence pairs.} A pair edits one sentence of the bridge passage so that another passage already present supports a different
final answer of the same type; an irrelevant factual edit and a meaning-preserving paraphrase are the controls. Every eligible two-hop base of
the reserved pair slice was screened in a frozen order (385 bases; 195 with a correct teacher answer on the original question --- the
declared exception: original-only teacher traces determined eligibility before the freeze); 21 edits were AI-drafted under written
conventions, 17 pairs were built and submitted, and the author's adjudication (the ledger is the record) accepted \NMQPairs{} pairs in
\NMQPairGroups{} groups. The cohort was frozen before any student evaluation and before any flipped, control or intervention output (the
planning projection had been about 34 pairs; the attained precision governs the reading). Paired correctness $P$ is success on both the
original and the flipped input at the same sample index (A on the original, B on the flip), averaged over the four repeats.

\paragraph{Results.} Table~\ref{tab:app-mq-reads} gives the four readings; Table~\ref{tab:app-mq-arms} the per-checkpoint values.
(1) Every trained student's own final-line set is recoverable (per-arm means \NMQRecTrainedMin{}--\NMQRecTrainedMax{} on their
comparison-specific common support, Table~\ref{tab:app-mq-arms}, against the untrained student's \NMQRecBaseKi{} on the characterization
population), but \mistill{} $-$ \vanillakd{} is \pci{\NMQRecMean}{\NMQRecLo}{\NMQRecHi} on \NMQRecN{} questions in \NMQRecGroups{} groups
(seed pairs \NMQRecSZeroMean, \NMQRecSOneMean, \NMQRecSTwoMean): the criterion is not met and the interval lies within $\pm$0.03. The secondary
contrast is negative, \mistill{} $-$ \randheads{} \pci{\NMQRecRandMean}{\NMQRecRandLo}{\NMQRecRandHi} (\NMQRecRandN{} / \NMQRecGroups), every
seed pair below zero: the student trained against the frozen random pool has the more recoverable own set. One frozen pool was run; this is
not a distribution over random pools, and we offer no explanation (the random pool spans more layers than the located one, a hypothesis
the design did not test). (2) Answer F1 on the \NMQEval{} reserved questions: \mistill{} $-$ \vanillakd{} \pci{\NMQTaskMean}{\NMQTaskLo}{\NMQTaskHi},
lower bound above $-$\NMQTaskMargin{}: the numerical noninferiority criterion is met (a loss smaller than the margin is permitted; unchanged
performance is not established), and after step 1 the reading is descriptive; the same models evaluated on a second GPU type give
\pci{\NMQTaskRtxMean}{\NMQTaskRtxLo}{\NMQTaskRtxHi} (single-model F1 moves by up to 0.014 between cards while the paired contrast moves by
0.002, so every reported evaluation of a compared pair shares one card type). (3) Paired correctness on the \NMQPairs{} pairs:
$G_{P,\mathrm{clean}}$ = \pci{\NMQBehMean}{\NMQBehLo}{\NMQBehHi} (seed pairs \NMQBehSZeroDelta, \NMQBehSOneDelta, \NMQBehSTwoDelta), unresolved
and within $\pm$0.10; no invalid answers in any arm. (4) The teacher gate (Table~\ref{tab:app-mq-gate}): with the prefill-inclusive
evidence-key mask over its own frozen top-40 on the \NMQPairs{} pairs (clean / sham / target / three layer-matched random draws, $N=4$;
\NMQGateSlots{} generations), the teacher's clean paired correctness is \NMQGateCleanPMean{} (the recorded equal-group mean; equal-pair
\NMQGateCleanPPair) and $E_P = P_{\mathrm{sham}} - P_{\mathrm{target}}$ is
\pci{\NMQGateEPMean}{\NMQGateEPLo}{\NMQGateEPHi} with $E_P^{\mathrm{rand}}$ \pci{\NMQGateEPRandMean}{\NMQGateEPRandLo}{\NMQGateEPRandHi}
(the recorded equal-group analysis; the equal-pair sensitivity in the table reaches the same conclusion); the sham and validity bands are met,
both intervals reach zero, and the detectable effect at this cohort size (\NMQGateEPDetectDelta{} / \NMQGateEPRandDetectDelta) is about twice
the locked $\delta_0$ of \NMQGateDeltaZero: inconclusive, not calibrated, so under the locked gate rule no student causal run was made and no
student causal claim exists for this task.

\paragraph{What this does and does not show.} Under this declared answer-emission adaptation, the recipe's recovery advantage over
\vanillakd{} was tested and not detected, with the random-pool control recovering more; the native task met its noninferiority criterion; the
behavioural and causal questions are unresolved at the attained cohort size. It does not show that the clinical result is wrong, that the
mechanism ``fails'' on this task (the located sets are recoverable in every arm), or why the two tasks differ. The development record (the
adapter's audit, the selection views, the training-memory workaround that ran the mechanism arms as exact-resume segments, and the
per-evaluation acceptance checks) is in the released repository.

\begin{table}[!htbp]
  \centering\small
  \caption{The public-task extension, the four pre-registered readings: estimate with a 95\,\% percentile bootstrap over base-question
  groups, both unit counts, and the locked criterion. \resid{R-D37}, \resid{R-D38}, \resid{R-D39}, \resid{R-D40}}
  \label{tab:app-mq-reads}
  \providecommand{\gensd}[2]{\begin{tabular}[t]{@{}c@{}}#1\\[-2pt]{\scriptsize$\pm$#2}\end{tabular}}
\providecommand{\genstack}[2]{\begin{tabular}[t]{@{}c@{}}#1\\[-2pt]{\scriptsize #2}\end{tabular}}
\providecommand{\genstackl}[2]{\begin{tabular}[t]{@{}l@{}}#1\\[-2pt]{\scriptsize #2}\end{tabular}}
\providecommand{\genhead}[2]{\begin{tabular}[b]{@{}c@{}}#1\\#2\end{tabular}}
\providecommand{\genstackiii}[3]{\begin{tabular}[t]{@{}c@{}}#1\\[-2pt]{\scriptsize #2}\\[-3pt]{\scriptsize #3}\end{tabular}}
\scriptsize\setlength{\tabcolsep}{3pt}
\begin{adjustbox}{max width=\linewidth}
\begin{tabular}{@{}p{0.40\linewidth} p{0.20\linewidth} p{0.17\linewidth} p{0.23\linewidth}@{}}
\toprule
Pre-registered step and contrast & Estimate [95\% CI] & Units (both counts) & Criterion \\
\midrule
1 recovery: own-heads final-line \Ptwo{} knock-in, $k=40$, \mistill{} $-$ \vanillakd{} & \pci{\NMQRecMean}{\NMQRecLo}{\NMQRecHi} & \NMQRecN{} questions / \NMQRecGroups{} groups; 3 seed pairs & lower bound $>0$: not met; within $\pm$0.03 \\
\quad secondary: \mistill{} $-$ \randheads{} pool 0 & \pci{\NMQRecRandMean}{\NMQRecRandLo}{\NMQRecRandHi} & \NMQRecRandN{} / \NMQRecGroups{} & the random pool's own heads recover more \\
2 task: answer macro-F1 on \texttt{mq\_eval}, \mistill{} $-$ \vanillakd{} & \pci{\NMQTaskMean}{\NMQTaskLo}{\NMQTaskHi} & \NMQEval{} / \NMQEvalGroups{} & lower bound $>-$\NMQTaskMargin{}: met (noninferiority); descriptive after step 1 \\
\quad secondary: \mistill{} $-$ \randheads{} & \pci{\NMQTaskRandMean}{\NMQTaskRandLo}{\NMQTaskRandHi} & \NMQEval{} / \NMQEvalGroups{} & --- \\
\quad sensitivity: the same models on the RTX card & \pci{\NMQTaskRtxMean}{\NMQTaskRtxLo}{\NMQTaskRtxHi} & \NMQEval{} / \NMQEvalGroups{} & same conclusion \\
3 behaviour: $G_{P,\mathrm{clean}}$ on the frozen pairs, \mistill{} $-$ \vanillakd{} & \pci{\NMQBehMean}{\NMQBehLo}{\NMQBehHi} & \NMQPairs{} pairs / \NMQPairGroups{} groups & lower bound $>0$: not met; within $\pm$0.10; unresolved \\
\quad secondary: \mistill{} $-$ \randheads{} & \pci{\NMQBehRandMean}{\NMQBehRandLo}{\NMQBehRandHi} & \NMQPairs{} / \NMQPairGroups{} & --- \\
4 causal: teacher gate $E_P$ (sham $-$ target, recorded equal-group) & \pci{\NMQGateEPMean}{\NMQGateEPLo}{\NMQGateEPHi} & \NMQPairs{} / \NMQPairGroups{} & interval reaches 0; detectable $\delta$ \NMQGateEPDetectDelta{} $>$ \NMQGateDeltaZero: inconclusive \\
\quad $E_P^{\mathrm{rand}}$ (random $-$ target, recorded equal-group) & \pci{\NMQGateEPRandMean}{\NMQGateEPRandLo}{\NMQGateEPRandHi} & \NMQPairs{} / \NMQPairGroups{} & detectable $\delta$ \NMQGateEPRandDetectDelta{}: inconclusive $\Rightarrow$ no student causal run \\
\bottomrule
\end{tabular}
\end{adjustbox}

\end{table}

\begin{table}[!htbp]
  \centering\small
  \caption{The public-task extension per checkpoint: answer F1 and valid-answer rate on the \NMQEval{} reserved questions (one GPU type),
  paired correctness $P$ on the \NMQPairs{} evidence pairs, and each student's own-heads final-line \Ptwo{} recovery at $k=40$ as a
  comparison-specific common-support mean (\vanillakd{} and \mistill{} from the primary comparison's \NMQRecN{} questions, \randheads{} from
  the secondary comparison's \NMQRecRandN; the untrained student's value from the separate characterization population); descriptive
  context, not a paired base-to-trained estimate. \resid{R-D37}, \resid{R-D38}, \resid{R-D39}}
  \label{tab:app-mq-arms}
  \providecommand{\gensd}[2]{\begin{tabular}[t]{@{}c@{}}#1\\[-2pt]{\scriptsize$\pm$#2}\end{tabular}}
\providecommand{\genstack}[2]{\begin{tabular}[t]{@{}c@{}}#1\\[-2pt]{\scriptsize #2}\end{tabular}}
\providecommand{\genstackl}[2]{\begin{tabular}[t]{@{}l@{}}#1\\[-2pt]{\scriptsize #2}\end{tabular}}
\providecommand{\genhead}[2]{\begin{tabular}[b]{@{}c@{}}#1\\#2\end{tabular}}
\providecommand{\genstackiii}[3]{\begin{tabular}[t]{@{}c@{}}#1\\[-2pt]{\scriptsize #2}\\[-3pt]{\scriptsize #3}\end{tabular}}
\scriptsize\setlength{\tabcolsep}{3pt}
\begin{adjustbox}{max width=\linewidth}
\begin{tabular}{@{}p{0.30\linewidth} p{0.17\linewidth} p{0.14\linewidth} p{0.14\linewidth} p{0.21\linewidth}@{}}
\toprule
Checkpoint & answer F1 (\texttt{mq\_eval}) & valid & $P$ (16 pairs) & own-heads recovery (common support) \\
\midrule
untrained Qwen3-4B & \NMQFBase & \NMQValidBase & \NMQPBase & \NMQRecBaseKi{} (Stage 5) \\
\vanillakd{} s0 / s1 / s2 & \NMQFCotKdSZero / \NMQFCotKdSOne / \NMQFCotKdSTwo & \NMQValidCotKdSZero / \NMQValidCotKdSOne / \NMQValidCotKdSTwo & \NMQPCotKdSZero / \NMQPCotKdSOne / \NMQPCotKdSTwo & \NMQRecSZeroMeanB / \NMQRecSOneMeanB / \NMQRecSTwoMeanB \\
\mistill{} ($\lambda$ \NMQLambda) s0 / s1 / s2 & \NMQFMistillSZero / \NMQFMistillSOne / \NMQFMistillSTwo & \NMQValidMistillSZero / \NMQValidMistillSOne / \NMQValidMistillSTwo & \NMQPMistillSZero / \NMQPMistillSOne / \NMQPMistillSTwo & \NMQRecSZeroMeanA / \NMQRecSOneMeanA / \NMQRecSTwoMeanA \\
\randheads{} pool 0 ($\lambda$ \NMQLambda) s0 / s1 / s2 & \NMQFRandHeadsSZero / \NMQFRandHeadsSOne / \NMQFRandHeadsSTwo & \NMQValidRandHeadsSZero / \NMQValidRandHeadsSOne / \NMQValidRandHeadsSTwo & \NMQPRandHeadsSZero / \NMQPRandHeadsSOne / \NMQPRandHeadsSTwo & \NMQRecRandSZeroMeanB / \NMQRecRandSOneMeanB / \NMQRecRandSTwoMeanB \\
\bottomrule
\end{tabular}
\end{adjustbox}

\end{table}

\begin{table}[!htbp]
  \centering\small
  \caption{The teacher evidence-mask gate on the \NMQPairs{} pairs in two weightings: the recorded analysis (equal-group: the mean of
  per-group means, bootstrap over groups, as the pinned reader computes it) and the equal-pair sensitivity (mean over pairs, groups
  resampled) from the same per-pair values; clean $P$ in the same two conventions. \resid{R-D40}}
  \label{tab:app-mq-gate}
  \providecommand{\gensd}[2]{\begin{tabular}[t]{@{}c@{}}#1\\[-2pt]{\scriptsize$\pm$#2}\end{tabular}}
\providecommand{\genstack}[2]{\begin{tabular}[t]{@{}c@{}}#1\\[-2pt]{\scriptsize #2}\end{tabular}}
\providecommand{\genstackl}[2]{\begin{tabular}[t]{@{}l@{}}#1\\[-2pt]{\scriptsize #2}\end{tabular}}
\providecommand{\genhead}[2]{\begin{tabular}[b]{@{}c@{}}#1\\#2\end{tabular}}
\providecommand{\genstackiii}[3]{\begin{tabular}[t]{@{}c@{}}#1\\[-2pt]{\scriptsize #2}\\[-3pt]{\scriptsize #3}\end{tabular}}
\scriptsize\setlength{\tabcolsep}{3pt}
\begin{adjustbox}{max width=\linewidth}
\begin{tabular}{@{}p{0.34\linewidth} p{0.33\linewidth} p{0.33\linewidth}@{}}
\toprule
Teacher-gate contrast (\NMQPairs{} pairs / \NMQPairGroups{} groups) & recorded: equal-group mean, bootstrap over groups & sensitivity: equal-pair mean, groups resampled \\
\midrule
clean $P$ (prerequisite $\ge 0.30$) & \pci{\NMQGateCleanPMean}{\NMQGateCleanPLo}{\NMQGateCleanPHi} & \NMQGateCleanPPair \\
$E_P$ = $P_{\mathrm{sham}} - P_{\mathrm{target}}$ & \pci{\NMQGateEPMean}{\NMQGateEPLo}{\NMQGateEPHi} & \pci{\NMQGatePairEPMean}{\NMQGatePairEPLo}{\NMQGatePairEPHi} \\
$E_P^{\mathrm{rand}}$ = mean $P_{\mathrm{random}} - P_{\mathrm{target}}$ & \pci{\NMQGateEPRandMean}{\NMQGateEPRandLo}{\NMQGateEPRandHi} & \pci{\NMQGatePairEPRandMean}{\NMQGatePairEPRandLo}{\NMQGatePairEPRandHi} \\
\bottomrule
\end{tabular}
\end{adjustbox}

\end{table}

\section{Structural characterization across tasks and a programming distillation pilot}
\label{app:tasks2}
\supports{Sec.~\ref{sec:char}, Sec.~\ref{sec:limitations}; R-D41--R-D46}
This appendix reports a screen of public tasks for a second setting in which the paper's two halves could be tested, run after the
public-task extension of Appendix~\ref{app:musique}. Its stages were fixed in sequence: development data informed the event adapters and
the choice of tasks; the rankings and the characterization procedure were frozen before the fresh-input confirmation, whose operational
reading was declared before its outputs and confirmed by the author afterwards; the pilot's recipe was locked before comparative training,
and its advancement thresholds were fixed before any pilot outcome was inspected. Two tasks passed the screen's characterization criteria
and one of them, programming, received a locked distillation pilot; both outcomes are reported. A revised Wikipedia see-also screening
configuration fell below the prespecified event-coverage threshold (\NXTWikiCovRev{} of \NXTWikiCovRevN{} inputs with a contrastive event
against \NXTWikiCovMin{}; the original configuration \NXTWikiCovOrig{} of \NXTWikiCovOrigN{}) and was not advanced; the development record
is retained.

\paragraph{Tasks and events.} EURLEX57K \citep{chalkidis_eurlex_2019}: the teacher (Qwen3-14B) labels an act with EuroVoc concepts from
a shown candidate list. BigCodeBench \citep{zhuo_bigcodebench_2025}: library-based code generation, the prompt naming the libraries to use.
The adapters locate eligible first namings in the reasoning: exact candidate-concept namings on EURLEX and qualified API-attribute namings
on programming. \Ptwo{} events mark trie-divergent tokens; \Ptwoc{} events contrast a naming with an earlier related naming at their first
differing token, under the task's structural relation rule (a EuroVoc broader, narrower or sibling concept; the same module, a different
attribute). Neither eventual use in the final answer nor semantic rejection of the earlier item is required, and a \Ptwo{} and a
\Ptwoc{} event can coincide on one naming. The structural scope was adopted after the initial screen and before the fresh audit and
confirmation. The revised extractors were audited by the author on fresh inputs, \NXTEurAuditRows{} sampled events from
\NXTEurAuditInputs{} acts and \NXTBcbAuditRows{} from \NXTBcbAuditInputs{} tasks: \NXTEurAuditExtraction{} / \NXTEurAuditRows{} and
\NXTBcbAuditExtraction{} / \NXTBcbAuditRows{} extractions correct, with a verified rejection of the earlier item in
\NXTEurAuditVerifiedRej{} / \NXTEurAuditRows{} and \NXTBcbAuditVerifiedRej{} / \NXTBcbAuditRows{} cases. One residual of the programming
rule is disclosed rather than revised: first-mention events on a doctest line copied without its prompt marker,
\NXTBcbResidualAonrNum{} / \NXTBcbResidualAonrDen{} events on the audit inputs, \NXTBcbResidualBzeroNum{} / \NXTBcbResidualBzeroDen{} on
the ranking batch and \NXTBcbResidualBoneNum{} / \NXTBcbResidualBoneDen{} on the assessment batch.

\paragraph{Characterization on fresh inputs.} The teacher's heads were ranked once per task on a development batch, assessed on the
disjoint development assessment batch, and then, with the ranking frozen, on inputs never used in development (100 acts and 100 tasks).
Recovery is the knock-in reader's $R=(m_{\mathrm{patched}}-m_{\mathrm{corrupt}})/(m_{\mathrm{all}}-m_{\mathrm{corrupt}})$ per event,
with the all-heads-minus-corrupt range at least 1 nat (the all-heads / clean agreement is a separate check), averaged per input and then
over inputs, with 95\,\% percentile bootstraps over inputs; necessity is the normalized margin loss under ablation of the same set, a
different measure. Table~\ref{tab:app-xt-char} gives both stages with their coverage (screened, event-bearing and evaluable inputs;
evaluable events). At $k=40$ on the fresh inputs the ranked set recovers \pci{\NXTEurConfRecMean}{\NXTEurConfRecLo}{\NXTEurConfRecHi}
of the range on EURLEX and \pci{\NXTBcbConfRecMean}{\NXTBcbConfRecLo}{\NXTBcbConfRecHi} on programming, against
\NXTEurConfRecRandMean{} and \NXTBcbConfRecRandMean{} for count-matched random heads: ranked minus random
\pci{\NXTEurConfDeltaMean}{\NXTEurConfDeltaLo}{\NXTEurConfDeltaHi} across \NXTEurConfEvaluable{} acts and
\pci{\NXTBcbConfDeltaMean}{\NXTBcbConfDeltaLo}{\NXTBcbConfDeltaHi} across \NXTBcbConfEvaluable{} tasks, positive on
\NXTEurConfAbove{} / \NXTEurConfEvaluable{} and \NXTBcbConfAbove{} / \NXTBcbConfEvaluable{}, respectively, with both split halves
positive; necessity \NXTEurConfNecMean{} and \NXTBcbConfNecMean{} against \NXTEurConfNecRandMean{} and \NXTBcbConfNecRandMean{} for
random heads. The fresh-input point estimates were larger than the development estimates (\NXTEurDevDeltaMean{} and
\NXTBcbDevDeltaMean{}); no between-stage test is supplied. Recovery is partial at $k=40$ and reaches \NXTEurConfRecHundredMean{} and
\NXTBcbConfRecHundredMean{} at $k=100$. The earlier-naming read-back diagnostic (the attention of the ranked \Ptwoc{} heads to the
alternative's earlier namings, paired per input against each control panel; \NXTEurConfRbRandN{} acts and \NXTBcbConfRbRandN{} tasks)
finds the ranked heads above random heads on both tasks (\pci{\NXTEurConfRbRandMean}{\NXTEurConfRbRandLo}{\NXTEurConfRbRandHi};
\pci{\NXTBcbConfRbRandMean}{\NXTBcbConfRbRandLo}{\NXTBcbConfRbRandHi}) and above the bottom of their own ranking on EURLEX
(\pci{\NXTEurConfRbBottomMean}{\NXTEurConfRbBottomLo}{\NXTEurConfRbBottomHi}), while on programming no difference from the bottom-40
was detected (\pci{\NXTBcbConfRbBottomMean}{\NXTBcbConfRbBottomLo}{\NXTBcbConfRbBottomHi}); the diagnostic reuses the clinical one
without inheriting its semantic interpretation. Natively, the programming teacher passes \NXTBcbNativePass{}
[\NXTBcbNativePassLo, \NXTBcbNativePassHi] of the fresh tasks (sandboxed execution; \NXTBcbNativePassCount{} passed,
\NXTBcbNativeFailCount{} failed, \NXTBcbNativeNoCodeCount{} without code), while the EURLEX teacher's micro-F1 on the fresh acts is
\pci{\NXTEurNativeTeacherMean}{\NXTEurNativeTeacherLo}{\NXTEurNativeTeacherHi} against
\pci{\NXTEurNativeRefMean}{\NXTEurNativeRefLo}{\NXTEurNativeRefHi} for the top-\NXTEurNativeRefN{} of a retrieval ranking: the EURLEX
characterization does not establish competitive native task performance.

\paragraph{The programming pilot: locked design.} Student Qwen3-4B; \NXTPilotTrainUsable{} usable KodCode \citep{xu_kodcode_2025}
training problems (of a locked split of \NXTPilotTrainSplit) with teacher traces, every arm on the same problems in the same order;
arms \vanillakd{}, \mistill{} and \randheads{}. The mechanism arms supervise \Ptwo{} and \Ptwoc{} events with separate pools by event
kind, the teacher's attention targets and, for \mistill{}, the untrained student's own top-40 per kind ranked on the development
batches; the \randheads{} pool keeps the weight profile on heads redrawn once with a fixed seed and frozen across seeds (overlap with
the \mistill{} pools: \NXTPilotPoolOverlapPtwo{} of \NXTPilotPoolPtwo{} \Ptwo{} heads, \NXTPilotPoolOverlapPtwoc{} of
\NXTPilotPoolPtwoc{} \Ptwoc{} heads); the \NXTPilotNoEvents{} problems without a first-mention event receive the CoT loss only. The
bf16 recipe of Appendix~\ref{app:objective}, with $\lambda$ fixed at 0.01 from the preceding DT-004 recipe
(Appendix~\ref{app:musique}) without programming tuning; two paired training seeds. Trainings ran as exact-resume segments on one GPU
type (H200), on the cluster and on rented pods with identical environments and hash-verified inputs. Before the pilot, on the
development assessment inputs, no gap between the untrained student's own-heads recovery and the teacher's was detected at $k=40$
(student minus teacher \pci{\NXTPilotGapKfortyMean}{\NXTPilotGapKfortyLo}{\NXTPilotGapKfortyHi}; \NXTPilotGapKfortyN{} inputs), and
the student was below the teacher at $k=29$ (\pci{\NXTPilotGapKtwonineMean}{\NXTPilotGapKtwonineLo}{\NXTPilotGapKtwonineHi}). The
recipe was locked (\NXTPilotLockAt) before comparative training; the advancement thresholds were fixed by the author
(\NXTPilotThresholdsAt) before any pilot outcome was inspected (two \vanillakd{} evaluations had completed, unread): a seed-averaged
median gain in own-heads \Ptwoc{} recovery at $k=40$ of at least \NXTPilotPrimaryMin{} over \vanillakd{} and \NXTPilotVsRandomMin{}
over \randheads{}, both seed pairs positive for both contrasts, and a seed-averaged pass@1 change of at least \NXTPilotPassFloor{}
(``meaningful'' and ``acceptable''; no significance requirement). They are thresholds for advancing to an independent confirmation,
not efficacy or noninferiority criteria. One amendment preceded the outcome: the pass@1 gate requires a terminal sandbox result for
every task.

\paragraph{What was evaluated.} Mechanism recovery is teacher-forced on the teacher's traces: each trained student's own heads are
located on the development ranking batches and evaluated on the development assessment batch, with the paired contrasts on the common
support (inputs evaluable in both arms of both seed pairs: \NXTPilotPtwocKfortyMcN{} inputs in \NXTPilotRecGroups{} near-duplicate
groups). Intervals are 10,000-draw percentile bootstraps resampling those groups, each input carrying all its runs and both seeds,
with an input-weighted mean within each resample; they are conditional on the two trained seed pairs. Native pass@1 is a separate
readout on the \NXTPilotPassMcN{} development tasks of the three screening batches (\NXTPilotPassGroups{} groups), executed in a
sandbox; they were not training items, which does not make them an independent confirmation.

\paragraph{Result.} Table~\ref{tab:app-xt-pilot} gives the contrasts. In the locked programming development pilot, \mistill{} reduced
own-heads \Ptwoc{} recovery relative to \vanillakd{} at the primary $k=40$ budget (median paired difference
\NXTPilotPtwocKfortyMcMed; mean \pci{\NXTPilotPtwocKfortyMcMean}{\NXTPilotPtwocKfortyMcLo}{\NXTPilotPtwocKfortyMcHi};
\NXTPilotPtwocKfortyMcAbove{} of \NXTPilotPtwocKfortyMcN{} inputs above zero; both seed pairs negative,
\NXTPilotPtwocKfortyMcSZeroMed{} and \NXTPilotPtwocKfortyMcSOneMed). The selected random-head configuration also exceeded \mistill{}
(median \NXTPilotPtwocKfortyMrMed; mean \pci{\NXTPilotPtwocKfortyMrMean}{\NXTPilotPtwocKfortyMrLo}{\NXTPilotPtwocKfortyMrHi}), while
between \randheads{} and \vanillakd{} no difference was detected
(\pci{\NXTPilotPtwocKfortyRcMean}{\NXTPilotPtwocKfortyRcLo}{\NXTPilotPtwocKfortyRcHi}). The raw knock-in gain in nats agrees in sign
(\mistill{} $-$ \vanillakd{} median \NXTPilotRawKfortyMcMed; mean
\pci{\NXTPilotRawKfortyMcMean}{\NXTPilotRawKfortyMcLo}{\NXTPilotRawKfortyMcHi}), so a normalizing-denominator explanation is
insufficient. The pass@1 point estimate met the prespecified floor (\mistill{} $-$ \vanillakd{}
\pci{\NXTPilotPassMcMean}{\NXTPilotPassMcLo}{\NXTPilotPassMcHi}; per arm \NXTPilotPassOneCotKdSZero{} / \NXTPilotPassOneCotKdSOne,
\NXTPilotPassOneMistillSZero{} / \NXTPilotPassOneMistillSOne{} and \NXTPilotPassOneRandHeadsSZero{} / \NXTPilotPassOneRandHeadsSOne{}
for \vanillakd{}, \mistill{} and \randheads{}), but its interval does not establish noninferiority. At the prespecified $k=29$
sensitivity no advantage over \vanillakd{} was detected (median \NXTPilotPtwocKtwonineMcMed; mean
\pci{\NXTPilotPtwocKtwonineMcMean}{\NXTPilotPtwocKtwonineMcLo}{\NXTPilotPtwocKtwonineMcHi}); the three contrasts there are
\mistill{} $-$ \vanillakd{} \NXTPilotPtwocKtwonineMcMed, \mistill{} $-$ \randheads{} \NXTPilotPtwocKtwonineMrMed{} and
\randheads{} $-$ \vanillakd{} \NXTPilotPtwocKtwonineRcMed{} (medians; descriptive, and medians of paired differences need not add).
The pilot did not advance to independent confirmation, and none was run.

\paragraph{What this does and does not show.} The structural characterization reproduces on two tasks on fresh inputs, within the
scope above. Under the tested recipe, event-local attention supervision at the teacher's first-mention positions did not raise, and in
this pilot lowered, the student's own-heads recovery relative to \vanillakd{}; the paired pass@1 estimate is
\pci{\NXTPilotPassMcMean}{\NXTPilotPassMcLo}{\NXTPilotPassMcHi}, and its interval does not establish noninferiority. The estimates
are conditional on this recipe, student, $\lambda$ and two seed pairs, on development assessment inputs; they are not an
independently confirmed effect. The result does not show that the clinical measurements are wrong, that every programming
configuration must fail, or why the tasks differ: no gap detected at $k=40$ between the untrained student and the teacher (which is
not evidence of equal recovery or of no headroom), the structural rather than rejection semantics of the events, the combined
\Ptwo{}/\Ptwoc{} objective, the untuned $\lambda$ and differences in training source are hypotheses the pilot did not isolate. The
development record (the screen's criteria and their revisions, the audits, the locks and thresholds with timestamps, a cross-host
training check, one evaluation restart and the disclosure of one log line read before the comparison) is in the development
repository; the locks and thresholds (\path|PLAN/locks/DT-005-*|) and the readers (\path|scripts/dt005/|) are in the anonymous code.

\begin{table}[!htbp]
  \centering\small
  \caption{The two further tasks: the teacher's ranked \Ptwoc{} top-40 against count-matched random heads at $k=40$, on the development
  assessment batch and on fresh inputs. Coverage as screened / event-bearing / evaluable inputs and evaluable / all events; recovery as
  the fraction of the all-heads-minus-corrupt margin range (means with 95\,\% percentile bootstraps over inputs); necessity as the
  normalized margin loss under ablation; the read-back contrast as the paired difference in attention mass on the alternative's earlier
  namings, with the inputs on which the ranked heads are higher. \resid{R-D41}, \resid{R-D42}}
  \label{tab:app-xt-char}
  \providecommand{\gensd}[2]{\begin{tabular}[t]{@{}c@{}}#1\\[-2pt]{\scriptsize$\pm$#2}\end{tabular}}
\providecommand{\genstack}[2]{\begin{tabular}[t]{@{}c@{}}#1\\[-2pt]{\scriptsize #2}\end{tabular}}
\providecommand{\genstackl}[2]{\begin{tabular}[t]{@{}l@{}}#1\\[-2pt]{\scriptsize #2}\end{tabular}}
\providecommand{\genhead}[2]{\begin{tabular}[b]{@{}c@{}}#1\\#2\end{tabular}}
\providecommand{\genstackiii}[3]{\begin{tabular}[t]{@{}c@{}}#1\\[-2pt]{\scriptsize #2}\\[-3pt]{\scriptsize #3}\end{tabular}}
\footnotesize\setlength{\tabcolsep}{3pt}
\begin{adjustbox}{max width=\linewidth}
\begin{tabular}{@{}l c c c c@{}}
\toprule
 & \genhead{EURLEX57K}{development} & \genhead{EURLEX57K}{fresh} & \genhead{BigCodeBench}{development} & \genhead{BigCodeBench}{fresh} \\
\midrule
inputs: screened / with P2C / evaluable & \NXTEurDevScreened / \NXTEurDevWithEvents / \NXTEurDevEvaluable & \NXTEurConfScreened / \NXTEurConfWithEvents / \NXTEurConfEvaluable & \NXTBcbDevScreened / \NXTBcbDevWithEvents / \NXTBcbDevEvaluable & \NXTBcbConfScreened / \NXTBcbConfWithEvents / \NXTBcbConfEvaluable \\
P2C events: evaluable / all & \NXTEurDevEventsEvaluable / \NXTEurDevEvents & \NXTEurConfEventsEvaluable / \NXTEurConfEvents & \NXTBcbDevEventsEvaluable / \NXTBcbDevEvents & \NXTBcbConfEventsEvaluable / \NXTBcbConfEvents \\
ranked top-40 recovery & \pci{\NXTEurDevRecMean}{\NXTEurDevRecLo}{\NXTEurDevRecHi} & \pci{\NXTEurConfRecMean}{\NXTEurConfRecLo}{\NXTEurConfRecHi} & \pci{\NXTBcbDevRecMean}{\NXTBcbDevRecLo}{\NXTBcbDevRecHi} & \pci{\NXTBcbConfRecMean}{\NXTBcbConfRecLo}{\NXTBcbConfRecHi} \\
ranked $-$ random (paired per input) & \pci{\NXTEurDevDeltaMean}{\NXTEurDevDeltaLo}{\NXTEurDevDeltaHi} & \pci{\NXTEurConfDeltaMean}{\NXTEurConfDeltaLo}{\NXTEurConfDeltaHi} & \pci{\NXTBcbDevDeltaMean}{\NXTBcbDevDeltaLo}{\NXTBcbDevDeltaHi} & \pci{\NXTBcbConfDeltaMean}{\NXTBcbConfDeltaLo}{\NXTBcbConfDeltaHi} \\
inputs ranked $>$ random / evaluable & \NXTEurDevAbove / \NXTEurDevEvaluable & \NXTEurConfAbove / \NXTEurConfEvaluable & \NXTBcbDevAbove / \NXTBcbDevEvaluable & \NXTBcbConfAbove / \NXTBcbConfEvaluable \\
necessity: ranked / random & \NXTEurDevNecMean / \NXTEurDevNecRandMean & \NXTEurConfNecMean / \NXTEurConfNecRandMean & \NXTBcbDevNecMean / \NXTBcbDevNecRandMean & \NXTBcbConfNecMean / \NXTBcbConfNecRandMean \\
ranked top-100 recovery, $k=100$ & --- & \NXTEurConfRecHundredMean & --- & \NXTBcbConfRecHundredMean \\
read-back: top-40 $-$ random & \pci{\NXTEurDevRbRandMean}{\NXTEurDevRbRandLo}{\NXTEurDevRbRandHi} & \pci{\NXTEurConfRbRandMean}{\NXTEurConfRbRandLo}{\NXTEurConfRbRandHi} & \pci{\NXTBcbDevRbRandMean}{\NXTBcbDevRbRandLo}{\NXTBcbDevRbRandHi} & \pci{\NXTBcbConfRbRandMean}{\NXTBcbConfRbRandLo}{\NXTBcbConfRbRandHi} \\
\quad inputs higher / $n$ & \NXTEurDevRbRandHigher / \NXTEurDevRbRandN & \NXTEurConfRbRandHigher / \NXTEurConfRbRandN & \NXTBcbDevRbRandHigher / \NXTBcbDevRbRandN & \NXTBcbConfRbRandHigher / \NXTBcbConfRbRandN \\
read-back: top-40 $-$ bottom-40 & \pci{\NXTEurDevRbBottomMean}{\NXTEurDevRbBottomLo}{\NXTEurDevRbBottomHi} & \pci{\NXTEurConfRbBottomMean}{\NXTEurConfRbBottomLo}{\NXTEurConfRbBottomHi} & \pci{\NXTBcbDevRbBottomMean}{\NXTBcbDevRbBottomLo}{\NXTBcbDevRbBottomHi} & \pci{\NXTBcbConfRbBottomMean}{\NXTBcbConfRbBottomLo}{\NXTBcbConfRbBottomHi} \\
\quad inputs higher / $n$ & \NXTEurDevRbBottomHigher / \NXTEurDevRbBottomN & \NXTEurConfRbBottomHigher / \NXTEurConfRbBottomN & \NXTBcbDevRbBottomHigher / \NXTBcbDevRbBottomN & \NXTBcbConfRbBottomHigher / \NXTBcbConfRbBottomN \\
\bottomrule
\end{tabular}
\end{adjustbox}

\end{table}

\begin{table}[!htbp]
  \centering\small
  \caption{The programming pilot: paired differences in own-heads \Ptwoc{} recovery on the \NXTPilotPtwocKfortyMcN{} common-support
  development inputs (seed-averaged median; mean with a 95\,\% percentile bootstrap resampling near-duplicate groups; the two seed
  pairs' medians at four decimals), the raw knock-in gain, and, as a separate panel, pass@1 on the development tasks (the seed columns
  there are mean pass-rate changes). The author's thresholds, fixed before any pilot outcome was inspected, in the last column.
  \resid{R-D43}, \resid{R-D44}, \resid{R-D45}, \resid{R-D46}}
  \label{tab:app-xt-pilot}
  \providecommand{\gensd}[2]{\begin{tabular}[t]{@{}c@{}}#1\\[-2pt]{\scriptsize$\pm$#2}\end{tabular}}
\providecommand{\genstack}[2]{\begin{tabular}[t]{@{}c@{}}#1\\[-2pt]{\scriptsize #2}\end{tabular}}
\providecommand{\genstackl}[2]{\begin{tabular}[t]{@{}l@{}}#1\\[-2pt]{\scriptsize #2}\end{tabular}}
\providecommand{\genhead}[2]{\begin{tabular}[b]{@{}c@{}}#1\\#2\end{tabular}}
\providecommand{\genstackiii}[3]{\begin{tabular}[t]{@{}c@{}}#1\\[-2pt]{\scriptsize #2}\\[-3pt]{\scriptsize #3}\end{tabular}}
\footnotesize\setlength{\tabcolsep}{4pt}
\begin{adjustbox}{max width=\linewidth}
\begin{tabular}{@{}l l c c c c l@{}}
\toprule
Readout & Contrast & \genhead{seed-averaged}{median} & \genhead{mean}{95\,\% group bootstrap} & \genhead{seed 0 / seed 1}{medians} & \genhead{inputs}{above / $n$} & Threshold \\
\midrule
\Ptwoc{} $k=40$ & \mistill{} $-$ \vanillakd{} & \NXTPilotPtwocKfortyMcMed & \pci{\NXTPilotPtwocKfortyMcMean}{\NXTPilotPtwocKfortyMcLo}{\NXTPilotPtwocKfortyMcHi} & \NXTPilotPtwocKfortyMcSZeroMed / \NXTPilotPtwocKfortyMcSOneMed & \NXTPilotPtwocKfortyMcAbove / \NXTPilotPtwocKfortyMcN & $\ge$ \NXTPilotPrimaryMin: \NXTPilotMeaningful \\
\Ptwoc{} $k=40$ & \mistill{} $-$ \randheads{} & \NXTPilotPtwocKfortyMrMed & \pci{\NXTPilotPtwocKfortyMrMean}{\NXTPilotPtwocKfortyMrLo}{\NXTPilotPtwocKfortyMrHi} & \NXTPilotPtwocKfortyMrSZeroMed / \NXTPilotPtwocKfortyMrSOneMed & \NXTPilotPtwocKfortyMrAbove / \NXTPilotPtwocKfortyMrN & $\ge$ \NXTPilotVsRandomMin \\
\Ptwoc{} $k=40$ & \randheads{} $-$ \vanillakd{} & \NXTPilotPtwocKfortyRcMed & \pci{\NXTPilotPtwocKfortyRcMean}{\NXTPilotPtwocKfortyRcLo}{\NXTPilotPtwocKfortyRcHi} & \NXTPilotPtwocKfortyRcSZeroMed / \NXTPilotPtwocKfortyRcSOneMed & \NXTPilotPtwocKfortyRcAbove / \NXTPilotPtwocKfortyRcN & --- \\
\Ptwoc{} $k=29$ & \mistill{} $-$ \vanillakd{} & \NXTPilotPtwocKtwonineMcMed & \pci{\NXTPilotPtwocKtwonineMcMean}{\NXTPilotPtwocKtwonineMcLo}{\NXTPilotPtwocKtwonineMcHi} & \NXTPilotPtwocKtwonineMcSZeroMed / \NXTPilotPtwocKtwonineMcSOneMed & \NXTPilotPtwocKtwonineMcAbove / \NXTPilotPtwocKtwonineMcN & sensitivity \\
\Ptwoc{} $k=29$ & \mistill{} $-$ \randheads{} & \NXTPilotPtwocKtwonineMrMed & \pci{\NXTPilotPtwocKtwonineMrMean}{\NXTPilotPtwocKtwonineMrLo}{\NXTPilotPtwocKtwonineMrHi} & \NXTPilotPtwocKtwonineMrSZeroMed / \NXTPilotPtwocKtwonineMrSOneMed & \NXTPilotPtwocKtwonineMrAbove / \NXTPilotPtwocKtwonineMrN & sensitivity \\
\Ptwoc{} $k=29$ & \randheads{} $-$ \vanillakd{} & \NXTPilotPtwocKtwonineRcMed & \pci{\NXTPilotPtwocKtwonineRcMean}{\NXTPilotPtwocKtwonineRcLo}{\NXTPilotPtwocKtwonineRcHi} & \NXTPilotPtwocKtwonineRcSZeroMed / \NXTPilotPtwocKtwonineRcSOneMed & \NXTPilotPtwocKtwonineRcAbove / \NXTPilotPtwocKtwonineRcN & sensitivity \\
\midrule
raw gain (nats), $k=40$ & \mistill{} $-$ \vanillakd{} & \NXTPilotRawKfortyMcMed & \pci{\NXTPilotRawKfortyMcMean}{\NXTPilotRawKfortyMcLo}{\NXTPilotRawKfortyMcHi} & \NXTPilotRawKfortyMcSZeroMed / \NXTPilotRawKfortyMcSOneMed & --- & --- \\
\midrule
\multicolumn{7}{@{}l}{\emph{pass@1 on the \NXTPilotPassMcN{} development tasks (\NXTPilotPassGroups{} groups); seed columns: mean pass-rate changes}} \\
pass@1 & \mistill{} $-$ \vanillakd{} & --- & \pci{\NXTPilotPassMcMean}{\NXTPilotPassMcLo}{\NXTPilotPassMcHi} & \NXTPilotPassMcSZeroMean / \NXTPilotPassMcSOneMean & --- & $\ge$ \NXTPilotPassFloor: \NXTPilotAcceptable{} (point) \\
pass@1 & \mistill{} $-$ \randheads{} & --- & \pci{\NXTPilotPassMrMean}{\NXTPilotPassMrLo}{\NXTPilotPassMrHi} & \NXTPilotPassMrSZeroMean / \NXTPilotPassMrSOneMean & --- & --- \\
\bottomrule
\end{tabular}
\end{adjustbox}

\end{table}

\section{Zero-shot transfer of the installed mechanism}
\label{app:zeroshot}
\supports{Sec.~\ref{sec:distill-mech}; \resid{R-D47}\resid{R-D48}\resid{R-D49}\resid{R-D50}\resid{R-D51}}
Is the own-heads gain of Section~\ref{sec:distill-mech} a property of the student, or of its training distribution? We evaluate the
locked, \mimiciv{}-trained students without any further training on a public clinical corpus that changes the language, the code
inventory and the prompt, and on three further shifts. Each read below was pre-registered before its outputs existed; the programme's
full set of pre-registered reads is in the released repository.

\paragraph{Students and data.} The bf16 \mistill{}, bf16 \vanillakd{} and matched-selection \randheads{} students of the main study
(\qwenfour{}, three seeds each; \randheads{} is pool 0 at its selected $\lambda$, Appendix~\ref{app:abl-matched}) and the untrained
\qwenfour{}; for the change of student family, the \smol{} students (\mistill{}, three seeds; bf16 \vanillakd{}; fp32 \vanillakd{},
two seeds; \randheads{}, one run) and the untrained \smol{}. CodiEsp-D~\citep{mirandaescalada_codiesp_2020} is a corpus of Spanish
clinical case reports annotated with CIE-10-ES diagnosis codes (CC BY 4.0). From its development split we draw disjoint batches: b0
(\NZsDocsBzero{} documents) for each student's own ranking, b1 (\NZsDocsBone) for the knock-in, and conf (\NZsDocsConf{}, fresh) for
the replication. The organisers' English machine translation of the conf documents (\NZsDocsEn) gives the change of language with
identical candidates. Each prompt carries a retrieved shortlist of about 560 candidate codes, and the teacher's traces (\teacher{},
the main study's decoding) are generated on each batch.

\paragraph{Instrument.} Each student ranks its own heads on b0 with the procedure of Section~\ref{sec:setup}, and its own top-$k$
heads are knocked in at the teacher's first-mention \Ptwoc{} events of b1, conf and the translation, from a donor document. Recovery
is $(m_{\text{patched}}-m_{\text{corrupt}})/(m_{\text{all}}-m_{\text{corrupt}})$ per event, over events with a range of at least one
nat, averaged over a document's evaluable events. Contrasts are seed-paired for \qwenfour{} (the per-document difference averaged over
the three seed pairs) and unpaired for \smol{} (per document, the mean over the three \mistill{} seeds minus the comparison run, or minus the mean of the two
fp32 \vanillakd{} runs). We
report the mean over documents with a 10,000-draw percentile bootstrap over documents and the documents above and below zero; a
contrast is detected when its interval lies above zero. On MuSiQue-Ans the clinical \qwenfour{} students are ranked and knocked in on
the \NZsMqQuestions{} questions of Appendix~\ref{app:musique} at the answer line, with that appendix's final-site \Ptwo{} estimand.
All readouts are teacher-forced on the teacher's traces, as in Section~\ref{sec:distill-mech}.

\paragraph{Results.} On CodiEsp-D the bf16 \mistill{} students' own top-40 heads recover more of the contrastive decision than bf16
\vanillakd's on b1 (\pci{\NZsBoneMcMean}{\NZsBoneMcLo}{\NZsBoneMcHi}; \NZsBoneMcAbove{} of \NZsBoneMcN{} documents above zero) and on
the fresh documents (\pci{\NZsConfMcMean}{\NZsConfMcLo}{\NZsConfMcHi}; \NZsConfMcAbove{} of \NZsConfMcN), and more than the
matched-selection \randheads{} on both (\pci{\NZsBoneMrMean}{\NZsBoneMrLo}{\NZsBoneMrHi} and
\pci{\NZsConfMrMean}{\NZsConfMrLo}{\NZsConfMrHi}; Table~\ref{tab:zs}). The advantage sits in the top heads: at $k=8$ it is
\pci{\NZsBoneKeightMcMean}{\NZsBoneKeightMcLo}{\NZsBoneKeightMcHi} on b1 and \pci{\NZsConfKeightMcMean}{\NZsConfKeightMcLo}{\NZsConfKeightMcHi}
on conf. The advantage is also detected on the organisers' English machine translation of the same documents, with similar
estimates (\pci{\NZsEnMcMean}{\NZsEnMcLo}{\NZsEnMcHi} at $k=40$,
\pci{\NZsEnKeightMcMean}{\NZsEnKeightMcLo}{\NZsEnKeightMcHi} at $k=8$). For \smol{}, whose untrained recovery is lowest, the estimated
advantage is largest: \mistill's own top-40 recover \NZsSmolBoneMedMistillLo--\NZsSmolBoneMedMistillHi{} on b1 against \NZsSmolBoneMedCotKd{}
for bf16 \vanillakd{} (\pci{\NZsSmolBoneMcMean}{\NZsSmolBoneMcLo}{\NZsSmolBoneMcHi}; on conf
\pci{\NZsSmolConfMcMean}{\NZsSmolConfMcLo}{\NZsSmolConfMcHi}), and more than fp32 \vanillakd{}
(\pci{\NZsSmolBoneMfMean}{\NZsSmolBoneMfLo}{\NZsSmolBoneMfHi}) and \randheads{} (\pci{\NZsSmolBoneMrMean}{\NZsSmolBoneMrLo}{\NZsSmolBoneMrHi}).
At MuSiQue-Ans answer events the clinical \mistill{} students' own top-40 recover more than bf16 \vanillakd's
(\pci{\NZsMqMcMean}{\NZsMqMcLo}{\NZsMqMcHi}; \NZsMqMcAbove{} of \NZsMqMcN{} questions above zero) and than \randheads'
(\pci{\NZsMqMrMean}{\NZsMqMrLo}{\NZsMqMrHi}), at a lower level than on the clinical corpus.

\begin{table}[!htbp]
  \centering\footnotesize
  \caption{Zero-shot transfer of the installed mechanism (no further training): differences in own-heads \Ptwoc{} recovery at $k$
  (MuSiQue-Ans: final-site \Ptwo{}), teacher-forced on the teacher's traces. Each difference is the mean over documents
  (MuSiQue-Ans: questions) of the per-document difference, with a 95\% bootstrap interval over documents, on the documents evaluable
  in every arm of that comparison ($n$, given per comparison); ``docs $+$\,/\,$-$'' counts the documents above and below zero for
  \mistill{} $-$ \vanillakd{}. \qwenfour{} contrasts are seed-paired over three pairs; \smol{} contrasts are unpaired. The per-arm
  levels are in Table~\ref{tab:zs-levels}. \resid{R-D47}\resid{R-D48}\resid{R-D49}\resid{R-D50}\resid{R-D51}}
  \label{tab:zs}
  \providecommand{\gensd}[2]{\begin{tabular}[t]{@{}c@{}}#1\\[-2pt]{\scriptsize$\pm$#2}\end{tabular}}
\providecommand{\genstack}[2]{\begin{tabular}[t]{@{}c@{}}#1\\[-2pt]{\scriptsize #2}\end{tabular}}
\providecommand{\genstackl}[2]{\begin{tabular}[t]{@{}l@{}}#1\\[-2pt]{\scriptsize #2}\end{tabular}}
\providecommand{\genhead}[2]{\begin{tabular}[b]{@{}c@{}}#1\\#2\end{tabular}}
\providecommand{\genstackiii}[3]{\begin{tabular}[t]{@{}c@{}}#1\\[-2pt]{\scriptsize #2}\\[-3pt]{\scriptsize #3}\end{tabular}}
\providecommand{\genwrapl}[2]{\begin{tabular}[t]{@{}l@{}}#1\\#2\end{tabular}}
\small\setlength{\tabcolsep}{5pt}
\begin{tabular}{@{}l c r c c r c@{}}
\toprule
 & & \multicolumn{3}{c}{\mistill{} $-$ \vanillakd{}} & \multicolumn{2}{c}{\mistill{} $-$ \randheads{}} \\
\cmidrule(lr){3-5}\cmidrule(l){6-7}
setting & $k$ & $n$ & mean [95\% CI] & docs $+$\,/\,$-$ & $n$ & mean [95\% CI] \\
\midrule
\multicolumn{7}{@{}l}{\qwenfour{} students: three seeds per arm, seed-paired} \\
CodiEsp-D b1 & 40 & \NZsBoneMcN & \pci{\NZsBoneMcMean}{\NZsBoneMcLo}{\NZsBoneMcHi} & \NZsBoneMcAbove{} / \NZsBoneMcBelow & \NZsBoneMrN & \pci{\NZsBoneMrMean}{\NZsBoneMrLo}{\NZsBoneMrHi} \\
\genwrapl{CodiEsp-D conf}{(fresh)} & 40 & \NZsConfMcN & \pci{\NZsConfMcMean}{\NZsConfMcLo}{\NZsConfMcHi} & \NZsConfMcAbove{} / \NZsConfMcBelow & \NZsConfMrN & \pci{\NZsConfMrMean}{\NZsConfMrLo}{\NZsConfMrHi} \\
\genwrapl{conf, English}{machine translation} & 40 & \NZsEnMcN & \pci{\NZsEnMcMean}{\NZsEnMcLo}{\NZsEnMcHi} & \NZsEnMcAbove{} / \NZsEnMcBelow & \NZsEnMrN & \pci{\NZsEnMrMean}{\NZsEnMrLo}{\NZsEnMrHi} \\
\midrule
CodiEsp-D b1 & 8 & \NZsBoneKeightMcN & \pci{\NZsBoneKeightMcMean}{\NZsBoneKeightMcLo}{\NZsBoneKeightMcHi} & \NZsBoneKeightMcAbove{} / \NZsBoneKeightMcBelow & \NZsBoneKeightMrN & \pci{\NZsBoneKeightMrMean}{\NZsBoneKeightMrLo}{\NZsBoneKeightMrHi} \\
\genwrapl{CodiEsp-D conf}{(fresh)} & 8 & \NZsConfKeightMcN & \pci{\NZsConfKeightMcMean}{\NZsConfKeightMcLo}{\NZsConfKeightMcHi} & \NZsConfKeightMcAbove{} / \NZsConfKeightMcBelow & \NZsConfKeightMrN & \pci{\NZsConfKeightMrMean}{\NZsConfKeightMrLo}{\NZsConfKeightMrHi} \\
\genwrapl{conf, English}{machine translation} & 8 & \NZsEnKeightMcN & \pci{\NZsEnKeightMcMean}{\NZsEnKeightMcLo}{\NZsEnKeightMcHi} & \NZsEnKeightMcAbove{} / \NZsEnKeightMcBelow & \NZsEnKeightMrN & \pci{\NZsEnKeightMrMean}{\NZsEnKeightMrLo}{\NZsEnKeightMrHi} \\
\midrule
\genwrapl{MuSiQue-Ans}{answer events} & 40 & \NZsMqMcN & \pci{\NZsMqMcMean}{\NZsMqMcLo}{\NZsMqMcHi} & \NZsMqMcAbove{} / \NZsMqMcBelow & \NZsMqMrN & \pci{\NZsMqMrMean}{\NZsMqMrLo}{\NZsMqMrHi} \\
\midrule
\multicolumn{7}{@{}l}{\smol{} students: \mistill{} three seeds; \vanillakd{}, \randheads{} one run each; unpaired} \\
CodiEsp-D b1 & 40 & \NZsSmolBoneMcN & \pci{\NZsSmolBoneMcMean}{\NZsSmolBoneMcLo}{\NZsSmolBoneMcHi} & \NZsSmolBoneMcAbove{} / \NZsSmolBoneMcBelow & \NZsSmolBoneMrN & \pci{\NZsSmolBoneMrMean}{\NZsSmolBoneMrLo}{\NZsSmolBoneMrHi} \\
\genwrapl{CodiEsp-D conf}{(fresh)} & 40 & \NZsSmolConfMcN & \pci{\NZsSmolConfMcMean}{\NZsSmolConfMcLo}{\NZsSmolConfMcHi} & \NZsSmolConfMcAbove{} / \NZsSmolConfMcBelow & \NZsSmolConfMrN & \pci{\NZsSmolConfMrMean}{\NZsSmolConfMrLo}{\NZsSmolConfMrHi} \\
\bottomrule
\end{tabular}

\end{table}

\begin{table}[!htbp]
  \centering\footnotesize
  \caption{The own-heads recovery levels behind Table~\ref{tab:zs}: each arm's median over its own evaluable documents (MuSiQue-Ans:
  questions) at the row's $k$. For \qwenfour{}, the range of the three per-seed medians (one value when both ends round alike); for
  \smol{}, \mistill's range over its three seeds and one run for each other arm. \resid{R-D47}\resid{R-D48}\resid{R-D49}\resid{R-D50}\resid{R-D51}}
  \label{tab:zs-levels}
  \providecommand{\gensd}[2]{\begin{tabular}[t]{@{}c@{}}#1\\[-2pt]{\scriptsize$\pm$#2}\end{tabular}}
\providecommand{\genstack}[2]{\begin{tabular}[t]{@{}c@{}}#1\\[-2pt]{\scriptsize #2}\end{tabular}}
\providecommand{\genstackl}[2]{\begin{tabular}[t]{@{}l@{}}#1\\[-2pt]{\scriptsize #2}\end{tabular}}
\providecommand{\genhead}[2]{\begin{tabular}[b]{@{}c@{}}#1\\#2\end{tabular}}
\providecommand{\genstackiii}[3]{\begin{tabular}[t]{@{}c@{}}#1\\[-2pt]{\scriptsize #2}\\[-3pt]{\scriptsize #3}\end{tabular}}
\providecommand{\genwrapl}[2]{\begin{tabular}[t]{@{}l@{}}#1\\#2\end{tabular}}
\small\setlength{\tabcolsep}{5pt}
\begin{tabular}{@{}l c c c c c@{}}
\toprule
 & & \multicolumn{4}{c}{own top-$k$ recovery, median over each arm's documents} \\
\cmidrule(l){3-6}
setting & $k$ & \mistill{} & \vanillakd{} & \randheads{} & untrained \\
\midrule
\multicolumn{6}{@{}l}{\qwenfour{} students: ranges over the three seeds} \\
CodiEsp-D b1 & 40 & \NZsBoneMedMistillRange & \NZsBoneMedCotKdRange & \NZsBoneMedRandHeadsRange & \NZsBoneMedBase \\
\genwrapl{CodiEsp-D conf}{(fresh)} & 40 & \NZsConfMedMistillRange & \NZsConfMedCotKdRange & \NZsConfMedRandHeadsRange & \NZsConfMedBase \\
\genwrapl{conf, English}{machine translation} & 40 & \NZsEnMedMistillRange & \NZsEnMedCotKdRange & \NZsEnMedRandHeadsRange & \NZsEnMedBase \\
\midrule
CodiEsp-D b1 & 8 & \NZsBoneKeightMedMistillRange & \NZsBoneKeightMedCotKdRange & \NZsBoneKeightMedRandHeadsRange & \NZsBoneKeightMedBase \\
\genwrapl{CodiEsp-D conf}{(fresh)} & 8 & \NZsConfKeightMedMistillRange & \NZsConfKeightMedCotKdRange & \NZsConfKeightMedRandHeadsRange & \NZsConfKeightMedBase \\
\genwrapl{conf, English}{machine translation} & 8 & \NZsEnKeightMedMistillRange & \NZsEnKeightMedCotKdRange & \NZsEnKeightMedRandHeadsRange & \NZsEnKeightMedBase \\
\midrule
\genwrapl{MuSiQue-Ans}{answer events} & 40 & \NZsMqMedMistillRange & \NZsMqMedCotKdRange & \NZsMqMedRandHeadsRange & \NZsMqMedBase \\
\midrule
\multicolumn{6}{@{}l}{\smol{} students: \mistill{} range over three seeds; others one run} \\
CodiEsp-D b1 & 40 & \NZsSmolBoneMedMistillRange & \NZsSmolBoneMedCotKd & \NZsSmolBoneMedRandHeads & \NZsSmolBoneMedBase \\
\genwrapl{CodiEsp-D conf}{(fresh)} & 40 & \NZsSmolConfMedMistillRange & \NZsSmolConfMedCotKd & \NZsSmolConfMedRandHeads & \NZsSmolConfMedBase \\
\bottomrule
\end{tabular}

\end{table}

\section{Evidence attention on the students' own decision heads, and the deletion test}
\label{app:evidence}
\supports{Sec.~\ref{sec:distill-evidence}, Sec.~\ref{sec:limitations}; \resid{R-D52}\resid{R-D53}\resid{R-D54}\resid{R-D55}\resid{R-D56}\resid{R-D57}}
Do the heads that carry the student's chapter decision attend to the evidence a human coder marks for each code, and does the
decision depend on that evidence? Each read below was pre-registered before its outputs existed. Its code and every input were
hashed before the first output, and a frozen reader computed it. All of them are reported here, detected or not.

\paragraph{Corpora.} CodiEsp-X is CodiEsp's evidence annotation~\citep{mirandaescalada_codiesp_2020}: for every gold code of a
case report, the span that supports it. We use the \NEaCxDocs{} documents of the official test split, read zero-shot with the
prompts, retrieved shortlists and teacher's traces of Appendix~\ref{app:zeroshot}'s protocol. The first read (Stage B below) used
that appendix's conf (\NZsDocsConf) and b1 (\NZsDocsBone) development batches. MDACE~\citep{cheng2023mdace} annotates
\texttt{MIMIC-III}~\citep{johnson2016mimic} charts with the evidence of their ICD-10-CM codes; its inpatient ICD-10 subset has
\NEmCharts{} charts. We render each chart's discharge summary in the students' training format:
\begin{itemize}
  \item the de-identification placeholders in \mimiciv{}'s form;
  \item the text cut at the first ``discharge diagnosis'', as in training (Appendix~\ref{app:data});
  \item the retrieved shortlist of Section~\ref{sec:distill-setup}.
\end{itemize}
A code is gold when it has an annotated span inside the cut text and is in the students' 5,651-code inventory. Of the
\NEmSpans{} ICD-10-CM spans in the summaries, \NEmSpansKept{} are kept, \NEmSpansCut{} fall after the cut and \NEmSpansOut{}
belong to codes outside the inventory. That leaves \NEmInputs{} inputs (\NEmNoGold{} charts have no gold code) and \NEmGold{} gold
codes, a median of \NEmGoldMedian{} per input; the summaries have a median of \NEmWordsMedian{} words. The teacher's traces on
them (\teacher{}, the main study's decoding) are all closed and valid (\NEmTraces). \texttt{MIMIC-III} and \mimiciv{} overlap in
2008--2012 with no public crosswalk, so an MDACE admission could also be one of the 500 training notes (well under one chart is
expected). Every arm trained on the same notes, so this bears on no contrast. \texttt{MIMIC-III} is used under the PhysioNet data
use agreement, and only aggregates are reported.

\paragraph{Models and heads.} All 18 models are teacher-forced on the teacher's traces, so every arm reads identical sequences:
\begin{itemize}
  \item three seeds each of bf16 \mistill{}, bf16 \vanillakd{}, \randheads{} (pool 0 at its selected $\lambda$), fp32 \mistill{}
    ($\lambda$ 0.01) and fp32 \vanillakd{};
  \item the untrained \qwenfour{};
  \item \mistill{} trained without the window term (one seed; Appendix~\ref{app:ablations});
  \item the teacher.
\end{itemize}
A student's own heads are its top-40 chapter-decision (\Pone) set, ranked after training and pooled with the weights of
Eq.~\ref{eq:pool}. On CodiEsp-X the ranking is the one on b0 of Appendix~\ref{app:zeroshot}; on MDACE it is the main study's
own-head ranking (Section~\ref{sec:distill-mech}), and the CodiEsp ranking is also read there. The teacher's reference is its
prose-time set (Section~\ref{sec:char-locus}), the target of \mistill's window term. The supervised pool is the heads the loss
names, identical in every arm.

\paragraph{Instrument.} At each \Pone{} event of a code with evidence, the pooled attention is summed over the $W{=}8$ trace
tokens before the chapter letter. The evidence windows are the tokens of the 32-word note windows (the loss's note buckets) that
overlap the code's spans. Three quantities are read:
\begin{itemize}
  \item the lift: the windows' share of the pool's within-note attention over their share of the note's tokens (1 = attention in
    proportion to length);
  \item the absolute share: the windows' share of all the pool's attention;
  \item the note share: the note's share of all the pool's attention.
\end{itemize}
The lift depends on how much of the note the evidence windows cover. It therefore compares arms within a corpus, not across
corpora. Per document we average over its evidenced events; a document whose teacher trace has no \Pone{} event on a code with evidence is not scored (\NEaCxLiftMcN{} of \NEaCxDocs{} CodiEsp-X documents and \NEaMdLiftMcN{} of \NEmInputs{} MDACE inputs are). A contrast averages the per-document differences over the three seed
pairs; we report its mean over documents with a 10,000-draw percentile bootstrap over documents, and it is detected when its
interval lies above zero. The validity check is read first: the teacher's lift, with its interval above 1. Every annotated span
mapped exactly onto its prompt.

\paragraph{Registration order.}
\begin{itemize}
  \item Stage B was the first read. Its primary was the supervised pool on conf; the own heads were a secondary.
  \item The confirmatory read on the test documents (B2) made the own heads the primary. It was registered after Stage B's read
    and before any of its own outputs existed.
  \item The MDACE reads were registered after the deletion test's read on CodiEsp-X and before any MDACE output.
\end{itemize}

\paragraph{Results: attention.} The teacher's prose-time heads point at the evidence on both corpora: lift
\pci{\NEaCxTeach}{\NEaCxTeachLo}{\NEaCxTeachHi} and \pci{\NEaMdTeach}{\NEaMdTeachLo}{\NEaMdTeachHi}. The primary, \mistill{}
$-$ \vanillakd{} on the own heads' lift, is detected on both. On CodiEsp-X it is
\pci{\NEaCxLiftMcMean}{\NEaCxLiftMcLo}{\NEaCxLiftMcHi}, above zero on \NEaCxLiftMcAbove{} of \NEaCxLiftMcN{} documents; on MDACE it
is \pci{\NEaMdLiftMcMean}{\NEaMdLiftMcLo}{\NEaMdLiftMcHi}, on \NEaMdLiftMcAbove{} of \NEaMdLiftMcN{}
(Tables~\ref{tab:app-evattn} and~\ref{tab:app-evattn-levels}). The difference also holds:
\begin{itemize}
  \item at the contrastive decision;
  \item in both precisions;
  \item on MDACE, for the CodiEsp-ranked heads.
\end{itemize}
Against \randheads{} the two halves differ. The specificity is shared: \randheads{} is slightly higher on CodiEsp-X and level on
MDACE. The coverage is \mistill's: its heads put more of all their attention on the evidence and read more of the note. The
student trained without the window term is lower on both lift and coverage, on both corpora.

On CodiEsp-X's development batches (Stage B), the own-heads difference is the same size:
\pci{\NEbConfLiftMcMean}{\NEbConfLiftMcLo}{\NEbConfLiftMcHi} on conf and \pci{\NEbBoneLiftMcMean}{\NEbBoneLiftMcLo}{\NEbBoneLiftMcHi}
on b1. Stage B's primary, the supervised pool, went the other way and was not detected:
\pci{\NEbConfSupMcMean}{\NEbConfSupMcLo}{\NEbConfSupMcHi} on conf and \pci{\NEbBoneSupMcMean}{\NEbBoneSupMcLo}{\NEbBoneSupMcHi} on
b1. Its note share rose (\NEbConfSupNoteMcMean{} on conf) and, post hoc, so did its absolute share on the evidence
(\NEbConfSupAbsMcMean). A per-head read on the test documents locates the reversal in the pooling. Of the \NEaPhRead{} heads read
(the own chapter heads shared by every \mistill{} and \vanillakd{} seed, and the supervised pool), \NEaPhUp{} point more
specifically at the evidence under \mistill{}; \NEaPhUpCi{} of them have intervals above zero and \NEaPhDownCi{} have intervals
below. Weighted as the supervised pool, the per-head differences still average \NEaPhWsup. The pool's lift, however, is a ratio
over its summed attention, and one late head, \NEaPhTopHead{} with \NEaPhTopW{} of the pool's weight, dominates that sum. The
supervised pool's reversal recurs on MDACE (Table~\ref{tab:app-evattn}).

\paragraph{Results: the deletion test.} Does the decision depend on that evidence? Take each gold code with an evidenced \Pone{} or
\Ptwoc{} event in the trace. Its evidence tokens are replaced by a filler token. A control replaces runs of other note tokens,
matched to the evidence's own run lengths. Let $m$ be the code's own decision margin. For the evidence ($v{=}$ev) or the control
($v{=}$ct), $d_v=(m-m_v)/\max(|m|,1\,\mathrm{nat})$, and the read is $\Delta=d_{\mathrm{ev}}-d_{\mathrm{ct}}$ at the chapter
decision. The author fixed the bar before any output: detected when the interval lies above zero; meaningful at a mean of at least
\NEdThrMin{} with every seed pair's median positive. Before any real output the edit unit was amended twice, from text to words to
single tokens, because smoke runs refused retokenised edits; both amendments are recorded in the released plan. Every arm read the
same \NEdCxEvents{} events on CodiEsp-X and \NEdMdEvents{} on MDACE.

Deleting the evidence weakens the teacher's chapter decision on both corpora, by
\pci{\NEdCxTeach}{\NEdCxTeachLo}{\NEdCxTeachHi} and \pci{\NEdMdTeach}{\NEdMdTeachLo}{\NEdMdTeachHi} of its margin. Distillation
itself creates that dependence: the untrained student shows none, while every trained student carries part of the teacher's
(Table~\ref{tab:app-evdel}). \mistill{} adds little or nothing to it.
\begin{itemize}
  \item On CodiEsp-X: \pci{\NEdCxDeltaMcMean}{\NEdCxDeltaMcLo}{\NEdCxDeltaMcHi} over \vanillakd{}, detected but below the bar. The
    seed pairs' medians are \NEdCxDeltaMcPairZero, \NEdCxDeltaMcPairOne{} and \NEdCxDeltaMcPairTwo, and \mistill{} does not differ
    from \randheads{} or from the student without the window term.
  \item On MDACE: \pci{\NEdMdDeltaMcMean}{\NEdMdDeltaMcLo}{\NEdMdDeltaMcHi}, not detected.
  \item At the contrastive decision, \mistill{} does not differ from \vanillakd{}.
\end{itemize}
The mediation read is descriptive only. It ablates each student's own \Pone{} heads at the window and the decision, and those heads
also carry the decision itself.

\paragraph{Results: spontaneous evidence.} In the students' own generations on CodiEsp, the reasoning quotes the evidence of an
emitted gold code at about the same rate in every arm, the untrained student's included. On the test documents the rates are
\NEsTestQuoteLvlMistillRange{} for \mistill{}, \NEsTestQuoteLvlCotKdRange{} for \vanillakd{} and \NEsTestQuoteLvlBase{} for the
untrained student. \mistill{} $-$ \vanillakd{} is \pci{\NEsConfQuoteMcMean}{\NEsConfQuoteMcLo}{\NEsConfQuoteMcHi} on conf and
\pci{\NEsTestQuoteMcMean}{\NEsTestQuoteMcLo}{\NEsTestQuoteMcHi} on the test documents, not detected on either.

\paragraph{What this establishes.} The corpora are two, with independent human annotation: one read zero-shot, one in the students'
own domain. On both, \mistill{} students' own decision heads attend to the evidence for each code more, and more specifically,
than \vanillakd{} students'. The specificity is shared with random-head supervision; the coverage is \mistill's; the window term
contributes. This is an attention result, teacher-forced, on one student family. The decision's causal dependence on that evidence
rises little or not at all, and the students do not quote the evidence more.

\begin{table}[!htbp]
  \centering\footnotesize
  \caption{Evidence attention on each student's own chapter-decision heads, in the window before each chapter letter: \mistill{}
  against the comparison arms, on the CodiEsp-X test documents (zero-shot) and on MDACE (in domain). Each difference is the mean
  over documents of the per-document difference, with a 95\% bootstrap interval over documents. $n$ is the number of documents
  evaluable in every arm of that comparison; ``docs $+$\,/\,$-$'' counts the documents above and below zero. Contrasts use three
  seed pairs, except \mistill{} s0 $-$ no-window s0 (one pair). The per-arm levels are in Table~\ref{tab:app-evattn-levels}.
  \resid{R-D52}\resid{R-D53}\resid{R-D54}}
  \label{tab:app-evattn}
  \providecommand{\gensd}[2]{\begin{tabular}[t]{@{}c@{}}#1\\[-2pt]{\scriptsize$\pm$#2}\end{tabular}}
\providecommand{\genstack}[2]{\begin{tabular}[t]{@{}c@{}}#1\\[-2pt]{\scriptsize #2}\end{tabular}}
\providecommand{\genstackl}[2]{\begin{tabular}[t]{@{}l@{}}#1\\[-2pt]{\scriptsize #2}\end{tabular}}
\providecommand{\genhead}[2]{\begin{tabular}[b]{@{}c@{}}#1\\#2\end{tabular}}
\providecommand{\genstackiii}[3]{\begin{tabular}[t]{@{}c@{}}#1\\[-2pt]{\scriptsize #2}\\[-3pt]{\scriptsize #3}\end{tabular}}
\footnotesize\setlength{\tabcolsep}{4pt}
\begin{tabular}{@{}l r c c@{}}
\toprule
contrast & $n$ & mean [95\% CI] & docs $+$\,/\,$-$ \\
\midrule
\multicolumn{4}{@{}l}{\textbf{CodiEsp-X, the test documents (zero-shot; own heads ranked on b0)}} \\
\multicolumn{4}{@{}l}{own chapter-decision heads, the window: lift of the evidence windows} \\
\quad \mistill{} $-$ \vanillakd{} (primary) & \NEaCxLiftMcN & \pci{\NEaCxLiftMcMean}{\NEaCxLiftMcLo}{\NEaCxLiftMcHi} & \NEaCxLiftMcAbove{} / \NEaCxLiftMcBelow \\
\quad \mistill{} $-$ \randheads{} & \NEaCxLiftMrN & \pci{\NEaCxLiftMrMean}{\NEaCxLiftMrLo}{\NEaCxLiftMrHi} & \NEaCxLiftMrAbove{} / \NEaCxLiftMrBelow \\
\quad \mistill{} s0 $-$ no-window s0 & \NEaCxLiftMwN & \pci{\NEaCxLiftMwMean}{\NEaCxLiftMwLo}{\NEaCxLiftMwHi} & \NEaCxLiftMwAbove{} / \NEaCxLiftMwBelow \\
\quad fp32 \mistill{} $-$ fp32 \vanillakd{} & \NEaCxLiftMfN & \pci{\NEaCxLiftMfMean}{\NEaCxLiftMfLo}{\NEaCxLiftMfHi} & \NEaCxLiftMfAbove{} / \NEaCxLiftMfBelow \\
\multicolumn{4}{@{}l}{own chapter-decision heads, the window: absolute share on the evidence windows} \\
\quad \mistill{} $-$ \vanillakd{} & \NEaCxAbsMcN & \pci{\NEaCxAbsMcMean}{\NEaCxAbsMcLo}{\NEaCxAbsMcHi} & \NEaCxAbsMcAbove{} / \NEaCxAbsMcBelow \\
\quad \mistill{} $-$ \randheads{} & \NEaCxAbsMrN & \pci{\NEaCxAbsMrMean}{\NEaCxAbsMrLo}{\NEaCxAbsMrHi} & \NEaCxAbsMrAbove{} / \NEaCxAbsMrBelow \\
\quad \mistill{} s0 $-$ no-window s0 & \NEaCxAbsMwN & \pci{\NEaCxAbsMwMean}{\NEaCxAbsMwLo}{\NEaCxAbsMwHi} & \NEaCxAbsMwAbove{} / \NEaCxAbsMwBelow \\
\multicolumn{4}{@{}l}{own chapter-decision heads, the window: note share} \\
\quad \mistill{} $-$ \vanillakd{} & \NEaCxNoteMcN & \pci{\NEaCxNoteMcMean}{\NEaCxNoteMcLo}{\NEaCxNoteMcHi} & \NEaCxNoteMcAbove{} / \NEaCxNoteMcBelow \\
\quad \mistill{} $-$ \randheads{} & \NEaCxNoteMrN & \pci{\NEaCxNoteMrMean}{\NEaCxNoteMrLo}{\NEaCxNoteMrHi} & \NEaCxNoteMrAbove{} / \NEaCxNoteMrBelow \\
\multicolumn{4}{@{}l}{own heads at the contrastive decision (\Ptwoc{} events): lift} \\
\quad \mistill{} $-$ \vanillakd{} & \NEaCxPcMcN & \pci{\NEaCxPcMcMean}{\NEaCxPcMcLo}{\NEaCxPcMcHi} & \NEaCxPcMcAbove{} / \NEaCxPcMcBelow \\
\quad \mistill{} $-$ \randheads{} & \NEaCxPcMrN & \pci{\NEaCxPcMrMean}{\NEaCxPcMrLo}{\NEaCxPcMrHi} & \NEaCxPcMrAbove{} / \NEaCxPcMrBelow \\
\multicolumn{4}{@{}l}{other pools, the window} \\
\quad supervised pool, lift: \mistill{} $-$ \vanillakd{} & \NEaCxSupMcN & \pci{\NEaCxSupMcMean}{\NEaCxSupMcLo}{\NEaCxSupMcHi} & \NEaCxSupMcAbove{} / \NEaCxSupMcBelow \\
\quad supervised pool, note share: \mistill{} $-$ \vanillakd{} & \NEaCxSupNoteMcN & \pci{\NEaCxSupNoteMcMean}{\NEaCxSupNoteMcLo}{\NEaCxSupNoteMcHi} & \NEaCxSupNoteMcAbove{} / \NEaCxSupNoteMcBelow \\
\midrule
\multicolumn{4}{@{}l}{\textbf{MDACE (in domain; own heads from the main study's ranking)}} \\
\multicolumn{4}{@{}l}{own chapter-decision heads, the window: lift of the evidence windows} \\
\quad \mistill{} $-$ \vanillakd{} (primary) & \NEaMdLiftMcN & \pci{\NEaMdLiftMcMean}{\NEaMdLiftMcLo}{\NEaMdLiftMcHi} & \NEaMdLiftMcAbove{} / \NEaMdLiftMcBelow \\
\quad \mistill{} $-$ \randheads{} & \NEaMdLiftMrN & \pci{\NEaMdLiftMrMean}{\NEaMdLiftMrLo}{\NEaMdLiftMrHi} & \NEaMdLiftMrAbove{} / \NEaMdLiftMrBelow \\
\quad \mistill{} s0 $-$ no-window s0 & \NEaMdLiftMwN & \pci{\NEaMdLiftMwMean}{\NEaMdLiftMwLo}{\NEaMdLiftMwHi} & \NEaMdLiftMwAbove{} / \NEaMdLiftMwBelow \\
\quad fp32 \mistill{} $-$ fp32 \vanillakd{} & \NEaMdLiftMfN & \pci{\NEaMdLiftMfMean}{\NEaMdLiftMfLo}{\NEaMdLiftMfHi} & \NEaMdLiftMfAbove{} / \NEaMdLiftMfBelow \\
\multicolumn{4}{@{}l}{own chapter-decision heads, the window: absolute share on the evidence windows} \\
\quad \mistill{} $-$ \vanillakd{} & \NEaMdAbsMcN & \pci{\NEaMdAbsMcMean}{\NEaMdAbsMcLo}{\NEaMdAbsMcHi} & \NEaMdAbsMcAbove{} / \NEaMdAbsMcBelow \\
\quad \mistill{} $-$ \randheads{} & \NEaMdAbsMrN & \pci{\NEaMdAbsMrMean}{\NEaMdAbsMrLo}{\NEaMdAbsMrHi} & \NEaMdAbsMrAbove{} / \NEaMdAbsMrBelow \\
\quad \mistill{} s0 $-$ no-window s0 & \NEaMdAbsMwN & \pci{\NEaMdAbsMwMean}{\NEaMdAbsMwLo}{\NEaMdAbsMwHi} & \NEaMdAbsMwAbove{} / \NEaMdAbsMwBelow \\
\multicolumn{4}{@{}l}{own chapter-decision heads, the window: note share} \\
\quad \mistill{} $-$ \vanillakd{} & \NEaMdNoteMcN & \pci{\NEaMdNoteMcMean}{\NEaMdNoteMcLo}{\NEaMdNoteMcHi} & \NEaMdNoteMcAbove{} / \NEaMdNoteMcBelow \\
\quad \mistill{} $-$ \randheads{} & \NEaMdNoteMrN & \pci{\NEaMdNoteMrMean}{\NEaMdNoteMrLo}{\NEaMdNoteMrHi} & \NEaMdNoteMrAbove{} / \NEaMdNoteMrBelow \\
\multicolumn{4}{@{}l}{own heads at the contrastive decision (\Ptwoc{} events): lift} \\
\quad \mistill{} $-$ \vanillakd{} & \NEaMdPcMcN & \pci{\NEaMdPcMcMean}{\NEaMdPcMcLo}{\NEaMdPcMcHi} & \NEaMdPcMcAbove{} / \NEaMdPcMcBelow \\
\quad \mistill{} $-$ \randheads{} & \NEaMdPcMrN & \pci{\NEaMdPcMrMean}{\NEaMdPcMrLo}{\NEaMdPcMrHi} & \NEaMdPcMrAbove{} / \NEaMdPcMrBelow \\
\multicolumn{4}{@{}l}{other pools, the window} \\
\quad CodiEsp-ranked own heads, lift: \mistill{} $-$ \vanillakd{} & \NEaMdCxhMcN & \pci{\NEaMdCxhMcMean}{\NEaMdCxhMcLo}{\NEaMdCxhMcHi} & \NEaMdCxhMcAbove{} / \NEaMdCxhMcBelow \\
\quad supervised pool, lift: \mistill{} $-$ \vanillakd{} & \NEaMdSupMcN & \pci{\NEaMdSupMcMean}{\NEaMdSupMcLo}{\NEaMdSupMcHi} & \NEaMdSupMcAbove{} / \NEaMdSupMcBelow \\
\quad supervised pool, note share: \mistill{} $-$ \vanillakd{} & \NEaMdSupNoteMcN & \pci{\NEaMdSupNoteMcMean}{\NEaMdSupNoteMcLo}{\NEaMdSupNoteMcHi} & \NEaMdSupNoteMcAbove{} / \NEaMdSupNoteMcBelow \\
\bottomrule
\end{tabular}

\end{table}

\begin{table}[!htbp]
  \centering\footnotesize
  \caption{The levels behind Table~\ref{tab:app-evattn}. Each cell is the arm's mean over its evaluable documents of the evidence
  windows' lift and absolute share, on its own chapter-decision heads in the window; for the three-seed arms, the range over the
  seeds (one value when both ends round alike). $^{\dagger}$The teacher has no student-side ranking; its row is its prose-time set,
  the target of the window term. $^{\ddagger}$One seed. \resid{R-D52}\resid{R-D54}}
  \label{tab:app-evattn-levels}
  \providecommand{\gensd}[2]{\begin{tabular}[t]{@{}c@{}}#1\\[-2pt]{\scriptsize$\pm$#2}\end{tabular}}
\providecommand{\genstack}[2]{\begin{tabular}[t]{@{}c@{}}#1\\[-2pt]{\scriptsize #2}\end{tabular}}
\providecommand{\genstackl}[2]{\begin{tabular}[t]{@{}l@{}}#1\\[-2pt]{\scriptsize #2}\end{tabular}}
\providecommand{\genhead}[2]{\begin{tabular}[b]{@{}c@{}}#1\\#2\end{tabular}}
\providecommand{\genstackiii}[3]{\begin{tabular}[t]{@{}c@{}}#1\\[-2pt]{\scriptsize #2}\\[-3pt]{\scriptsize #3}\end{tabular}}
\footnotesize\setlength{\tabcolsep}{4pt}
\begin{tabular}{@{}l c c c c@{}}
\toprule
 & \multicolumn{2}{c}{CodiEsp-X (zero-shot)} & \multicolumn{2}{c}{MDACE (in domain)} \\
\cmidrule(lr){2-3}\cmidrule(l){4-5}
model (its own chapter-decision heads) & lift & absolute share & lift & absolute share \\
\midrule
teacher's prose-time heads$^{\dagger}$ & \NEaCxLvlTeacherLift & \NEaCxLvlTeacherAbs & \NEaMdLvlTeacherLift & \NEaMdLvlTeacherAbs \\
untrained \qwenfour{} & \NEaCxLiftLvlBase & \NEaCxAbsLvlBase & \NEaMdLiftLvlBase & \NEaMdAbsLvlBase \\
\mistill{} (bf16) & \NEaCxLiftLvlMistillRange & \NEaCxAbsLvlMistillRange & \NEaMdLiftLvlMistillRange & \NEaMdAbsLvlMistillRange \\
\vanillakd{} (bf16) & \NEaCxLiftLvlCotKdRange & \NEaCxAbsLvlCotKdRange & \NEaMdLiftLvlCotKdRange & \NEaMdAbsLvlCotKdRange \\
\randheads{} & \NEaCxLiftLvlRandHeadsRange & \NEaCxAbsLvlRandHeadsRange & \NEaMdLiftLvlRandHeadsRange & \NEaMdAbsLvlRandHeadsRange \\
\mistill{} without the window term$^{\ddagger}$ & \NEaCxLiftLvlNoWin & \NEaCxAbsLvlNoWin & \NEaMdLiftLvlNoWin & \NEaMdAbsLvlNoWin \\
\mistill{} (fp32, $\lambda$ 0.01) & \NEaCxLiftLvlMistillFpRange & \NEaCxAbsLvlMistillFpRange & \NEaMdLiftLvlMistillFpRange & \NEaMdAbsLvlMistillFpRange \\
\vanillakd{} (fp32) & \NEaCxLiftLvlCotKdFpRange & \NEaCxAbsLvlCotKdFpRange & \NEaMdLiftLvlCotKdFpRange & \NEaMdAbsLvlCotKdFpRange \\
\bottomrule
\end{tabular}

\end{table}

\begin{table}[!htbp]
  \centering\footnotesize
  \caption{The deletion test, levels. $\Delta$ is the drop of the code's own chapter margin when its annotated evidence is replaced
  by a filler, minus the drop for a matched control, as a share of the clean margin: each arm's mean over documents, with ranges
  over the three seeds. The lower block ablates each student's own \Pone{} heads at the window and the decision (the teacher: its
  prose-time set); it is descriptive, since those heads also carry the decision. $^{\ddagger}$One seed. \resid{R-D55}\resid{R-D56}}
  \label{tab:app-evdel-levels}
  \providecommand{\gensd}[2]{\begin{tabular}[t]{@{}c@{}}#1\\[-2pt]{\scriptsize$\pm$#2}\end{tabular}}
\providecommand{\genstack}[2]{\begin{tabular}[t]{@{}c@{}}#1\\[-2pt]{\scriptsize #2}\end{tabular}}
\providecommand{\genstackl}[2]{\begin{tabular}[t]{@{}l@{}}#1\\[-2pt]{\scriptsize #2}\end{tabular}}
\providecommand{\genhead}[2]{\begin{tabular}[b]{@{}c@{}}#1\\#2\end{tabular}}
\providecommand{\genstackiii}[3]{\begin{tabular}[t]{@{}c@{}}#1\\[-2pt]{\scriptsize #2}\\[-3pt]{\scriptsize #3}\end{tabular}}
\footnotesize\setlength{\tabcolsep}{4pt}
\begin{tabular}{@{}l c c@{}}
\toprule
 & CodiEsp-X (zero-shot) & MDACE (in domain) \\
\midrule
\multicolumn{3}{@{}l}{$\Delta$ at the chapter decision (share of the clean margin)} \\
\quad teacher & \NEdCxDeltaLvlTeacher & \NEdMdDeltaLvlTeacher \\
\quad untrained \qwenfour{} & \NEdCxDeltaLvlBase & \NEdMdDeltaLvlBase \\
\quad \mistill{} (bf16) & \NEdCxDeltaLvlMistillRange & \NEdMdDeltaLvlMistillRange \\
\quad \vanillakd{} (bf16) & \NEdCxDeltaLvlCotKdRange & \NEdMdDeltaLvlCotKdRange \\
\quad \randheads{} & \NEdCxDeltaLvlRandHeadsRange & \NEdMdDeltaLvlRandHeadsRange \\
\quad \mistill{} without the window term$^{\ddagger}$ & \NEdCxDeltaLvlNoWin & \NEdMdDeltaLvlNoWin \\
\quad \mistill{} (fp32, $\lambda$ 0.01) & \NEdCxDeltaLvlMistillFpRange & \NEdMdDeltaLvlMistillFpRange \\
\quad \vanillakd{} (fp32) & \NEdCxDeltaLvlCotKdFpRange & \NEdMdDeltaLvlCotKdFpRange \\
\midrule
\multicolumn{3}{@{}l}{$\Delta$ with the own \Pone{} heads ablated (teacher: its prose-time set); descriptive} \\
\quad teacher & \NEdCxAblLvlTeacher & \NEdMdAblLvlTeacher \\
\quad \mistill{} (bf16) & \NEdCxAblLvlMistillRange & \NEdMdAblLvlMistillRange \\
\quad \vanillakd{} (bf16) & \NEdCxAblLvlCotKdRange & \NEdMdAblLvlCotKdRange \\
\bottomrule
\end{tabular}

\end{table}

\begin{table}[!htbp]
  \centering\footnotesize
  \caption{The deletion test, contrasts: $\Delta$ (Table~\ref{tab:app-evdel-levels}), \mistill{} against the comparison arms.
  Each difference is the mean over documents of the per-document difference, with a 95\% bootstrap interval over documents; $n$
  is the documents evaluable in every arm of that comparison, and ``docs $+$\,/\,$-$'' counts those above and below zero. The bar
  was fixed by the author before any output: detected when the interval lies above zero; meaningful at a mean of at least
  \NEdThrMin{} with every seed pair's median positive. \resid{R-D55}\resid{R-D56}}
  \label{tab:app-evdel}
  \providecommand{\gensd}[2]{\begin{tabular}[t]{@{}c@{}}#1\\[-2pt]{\scriptsize$\pm$#2}\end{tabular}}
\providecommand{\genstack}[2]{\begin{tabular}[t]{@{}c@{}}#1\\[-2pt]{\scriptsize #2}\end{tabular}}
\providecommand{\genstackl}[2]{\begin{tabular}[t]{@{}l@{}}#1\\[-2pt]{\scriptsize #2}\end{tabular}}
\providecommand{\genhead}[2]{\begin{tabular}[b]{@{}c@{}}#1\\#2\end{tabular}}
\providecommand{\genstackiii}[3]{\begin{tabular}[t]{@{}c@{}}#1\\[-2pt]{\scriptsize #2}\\[-3pt]{\scriptsize #3}\end{tabular}}
\footnotesize\setlength{\tabcolsep}{4pt}
\begin{tabular}{@{}l r c c@{}}
\toprule
contrast & $n$ & mean [95\% CI] & docs $+$\,/\,$-$ \\
\midrule
\multicolumn{4}{@{}l}{\textbf{CodiEsp-X, the test documents}} \\
\quad \mistill{} $-$ \vanillakd{} (primary) & \NEdCxDeltaMcN & \pci{\NEdCxDeltaMcMean}{\NEdCxDeltaMcLo}{\NEdCxDeltaMcHi} & \NEdCxDeltaMcAbove{} / \NEdCxDeltaMcBelow \\
\quad\quad per seed pair, the median &  & \NEdCxDeltaMcPairZero{} / \NEdCxDeltaMcPairOne{} / \NEdCxDeltaMcPairTwo &  \\
\quad \mistill{} $-$ \randheads{} & \NEdCxDeltaMrN & \pci{\NEdCxDeltaMrMean}{\NEdCxDeltaMrLo}{\NEdCxDeltaMrHi} & \NEdCxDeltaMrAbove{} / \NEdCxDeltaMrBelow \\
\quad \mistill{} s0 $-$ no-window s0 & \NEdCxDeltaMwN & \pci{\NEdCxDeltaMwMean}{\NEdCxDeltaMwLo}{\NEdCxDeltaMwHi} & \NEdCxDeltaMwAbove{} / \NEdCxDeltaMwBelow \\
\quad fp32 \mistill{} $-$ fp32 \vanillakd{} & \NEdCxDeltaMfN & \pci{\NEdCxDeltaMfMean}{\NEdCxDeltaMfLo}{\NEdCxDeltaMfHi} & \NEdCxDeltaMfAbove{} / \NEdCxDeltaMfBelow \\
\quad \Ptwoc{} events: \mistill{} $-$ \vanillakd{} & \NEdCxPcDeltaMcN & \pci{\NEdCxPcDeltaMcMean}{\NEdCxPcDeltaMcLo}{\NEdCxPcDeltaMcHi} & \NEdCxPcDeltaMcAbove{} / \NEdCxPcDeltaMcBelow \\
\quad mediated part (descriptive): \mistill{} $-$ \vanillakd{} & \NEdCxMediatedMcN & \pci{\NEdCxMediatedMcMean}{\NEdCxMediatedMcLo}{\NEdCxMediatedMcHi} & \NEdCxMediatedMcAbove{} / \NEdCxMediatedMcBelow \\
\midrule
\multicolumn{4}{@{}l}{\textbf{MDACE}} \\
\quad \mistill{} $-$ \vanillakd{} (primary) & \NEdMdDeltaMcN & \pci{\NEdMdDeltaMcMean}{\NEdMdDeltaMcLo}{\NEdMdDeltaMcHi} & \NEdMdDeltaMcAbove{} / \NEdMdDeltaMcBelow \\
\quad\quad per seed pair, the median &  & \NEdMdDeltaMcPairZero{} / \NEdMdDeltaMcPairOne{} / \NEdMdDeltaMcPairTwo &  \\
\quad \mistill{} $-$ \randheads{} & \NEdMdDeltaMrN & \pci{\NEdMdDeltaMrMean}{\NEdMdDeltaMrLo}{\NEdMdDeltaMrHi} & \NEdMdDeltaMrAbove{} / \NEdMdDeltaMrBelow \\
\quad \mistill{} s0 $-$ no-window s0 & \NEdMdDeltaMwN & \pci{\NEdMdDeltaMwMean}{\NEdMdDeltaMwLo}{\NEdMdDeltaMwHi} & \NEdMdDeltaMwAbove{} / \NEdMdDeltaMwBelow \\
\quad fp32 \mistill{} $-$ fp32 \vanillakd{} & \NEdMdDeltaMfN & \pci{\NEdMdDeltaMfMean}{\NEdMdDeltaMfLo}{\NEdMdDeltaMfHi} & \NEdMdDeltaMfAbove{} / \NEdMdDeltaMfBelow \\
\quad \Ptwoc{} events: \mistill{} $-$ \vanillakd{} & \NEdMdPcDeltaMcN & \pci{\NEdMdPcDeltaMcMean}{\NEdMdPcDeltaMcLo}{\NEdMdPcDeltaMcHi} & \NEdMdPcDeltaMcAbove{} / \NEdMdPcDeltaMcBelow \\
\quad mediated part (descriptive): \mistill{} $-$ \vanillakd{} & \NEdMdMediatedMcN & \pci{\NEdMdMediatedMcMean}{\NEdMdMediatedMcLo}{\NEdMdMediatedMcHi} & \NEdMdMediatedMcAbove{} / \NEdMdMediatedMcBelow \\
\bottomrule
\end{tabular}

\end{table}

\section{Result index and reproduction entries}
\label{app:repro}
\supports{every \resid{} in the paper}

\begin{table}[!htbp]
  \centering\footnotesize
  \caption{Claim, evidence and boundary: what each result establishes and what it does not (the pointer is in Sec.~1).
  Evidence names the section and the table that carry it; the boundary is the paper's own limit on the claim.}
  \label{tab:app-claims}
  \setlength{\tabcolsep}{3pt}
  \begin{tabularx}{\linewidth}{@{}>{\raggedright\arraybackslash}p{0.30\linewidth} >{\raggedright\arraybackslash}X >{\raggedright\arraybackslash}X@{}}
    \toprule
    Claim & Evidence & Not established \\
    \midrule
    Chapter decision: a small, global head set is necessary and sufficient on nearly every note & Sec.~\ref{sec:char}; ablation and knock-in per note; Table~\ref{tab:heads-app}; replicated on the shortlist (Appendix~\ref{app:hd002}), in a second reasoner and without the narrowing instruction & that the global set is each note's best set; roles of MLPs; single-head effects near the noise floor \\
    Within-region (character) decision: carried by a phase-specific set & Sec.~\ref{sec:char}; the intact-hook sensitivity re-run, Table~\ref{tab:app-p2intact} & the ranking beyond first order (the historical rankings are detached-hook; the re-run covers 50 notes) \\
    Contrastive decision: a global set recovers it; \head{30}{11} acts through its read of the earlier mention & Sec.~\ref{sec:char}; key-side mask, Table~\ref{tab:app-keyabl}; the audit, Table~\ref{tab:app-audit-reads} & sibling-specific identity processing (no difference detected between the tested replacements, Table~\ref{tab:app-nearmiss}); that every \Ptwoc{} event is a rejected alternative (\NAuditRejectedPct{} are) \\
    Read paths: region readers and the near-miss loop-back & key-side intervention on three heads, Table~\ref{tab:app-keyabl}; descriptive elsewhere & that attention patterns are causal beyond the three heads tested \\
    Temporal locus: for mentions elicitable in the reasoning, the chapter is elicitable several tokens before the code, by a disjoint mid-layer set, and decided at the code & Sec.~\ref{sec:char-locus}; follow-through and residual swap, Appendix~\ref{app:locus} & a locus for the probe-failure mentions; ``decided in the prose'' (the window set removes a quarter of the natural margin) \\
    Recoverability transfer: \mistill{} raises the student's own-heads causal recovery of the contrastive decision, on the teacher's traces and on its own completed ones; at the selected fp32 setting the gain holds on average across three seed pairs at $k=40$ & Sec.~\ref{sec:distill-mech}, Table~\ref{tab:mech}; note-clustered, Table~\ref{tab:app-mech-seeds}; own traces, Table~\ref{tab:app-owntrace}; three fp32 seeds, Appendix~\ref{app:obj-fp32} & a character-level gain at matched precision (it reverses); a gain at $k=8$ for the fp32 setting (the pooled comparison reverses); evidence-use transfer (next row); edit sensitivity consistent across precision and family \\
    What the gain depends on: ranked student heads supervised away from the teacher's decision positions fall below the untrained student (both students); on \smol, targets built from unselected teacher heads leave recovery at \vanillakd's level, so the identified teacher heads contribute as the source of the target, while which student heads receive it matters less & Sec.~\ref{sec:distill-ablation}; Table~\ref{tab:app-ablate}; the teacher's pools replaced, with a positive control, Appendix~\ref{app:teacherpool}, Tables~\ref{tab:app-link} and~\ref{tab:app-teacherpool} & the teacher-side control on \qwenfour, where random heads at random positions match \mistill; which property of the ranked target is needed; that the order or weights of the ranking matter, or that another selection of teacher heads would fail; the span buckets; the student-side ranking's own value \\
    Task relation: no detected task difference from \vanillakd{} at matched precision; a better observed task--recovery trade-off than the uniform control; under matched selection the ranked pick is above a frozen random pool on both endpoints & Sec.~\ref{sec:distill-main}, Tables~\ref{tab:main} and~\ref{tab:app-paired}; Sec.~\ref{sec:distill-ablation}, Appendices~\ref{app:ablations} and~\ref{app:abl-matched} & equivalence (no margin was fixed); what the ranking buys against generic attention supervision in general (the uniform control also differs in term composition, head count and weight; the matched comparison covers one frozen pool, conditional on the selection rule) \\
    \bottomrule
  \end{tabularx}
\end{table}

\begin{table}[!htbp]
  \centering\footnotesize
  \caption{Claim, evidence and boundary, continued from Table~\ref{tab:app-claims}.}
  \label{tab:app-claims-two}
  \setlength{\tabcolsep}{3pt}
  \begin{tabularx}{\linewidth}{@{}>{\raggedright\arraybackslash}p{0.30\linewidth} >{\raggedright\arraybackslash}X >{\raggedright\arraybackslash}X@{}}
    \toprule
    Claim & Evidence & Not established \\
    \midrule
    Evidence use: tested, not detected --- on the approved evidence-flip cases \mistill{} shows no detected improvement in directional sensitivity over \vanillakd, and masking the evidence keys through the nominated heads at decode time shows no detected target-versus-sham effect & Sec.~\ref{sec:limitations}; Appendix~\ref{app:evflip}, Tables~\ref{tab:app-ef-behav} and~\ref{tab:app-ef-causal} & that the pathway is never used (``not detected'' on one population through the direct reading route, with the detectable effects stated); any route through evidence propagated during prefill; fp32 or three-seed causal replication \\
    Public-task extension: a preregistered answer-emission adaptation on MuSiQue-Ans did not detect the recovery advantage over \vanillakd{} (\pci{\NMQRecMean}{\NMQRecLo}{\NMQRecHi}), and the matched random pool recovered more; the native task met its noninferiority criterion & Appendix~\ref{app:musique}, Tables~\ref{tab:app-mq-reads}--\ref{tab:app-mq-gate} & that the clinical result transfers to other tasks; why the two tasks differ; evidence-dependent behaviour on that task (unresolved at \NMQPairs{} pairs); any student causal claim there (the teacher gate was inconclusive) \\
    Cross-task structural characterization: head sets ranked on development data recover the teacher's structural first-mention decisions above count-matched random sets on fresh EURLEX57K acts and BigCodeBench tasks; a locked programming distillation pilot did not reproduce the recovery advantage (\mistill{} $-$ \vanillakd{} \pci{\NXTPilotPtwocKfortyMcMean}{\NXTPilotPtwocKfortyMcLo}{\NXTPilotPtwocKfortyMcHi}) and the selected random-head configuration recovered more & Appendix~\ref{app:tasks2}, Tables~\ref{tab:app-xt-char} and~\ref{tab:app-xt-pilot} & semantic rejection at the events; full replication of the clinical mechanism (partial recovery; read-back not distinguishable from the bottom-ranked heads on programming); competitive native EURLEX tagging; noninferiority of task performance in the pilot; why the tasks differ \\
    Zero-shot transfer: without further training, \mistill's own-heads recovery advantage holds on a public Spanish clinical corpus, on its English machine translation, for the \smol{} students and at MuSiQue-Ans answer events (bf16 \mistill{} $-$ \vanillakd{} on CodiEsp-D \pci{\NZsBoneMcMean}{\NZsBoneMcLo}{\NZsBoneMcHi}) & Appendix~\ref{app:zeroshot}, Table~\ref{tab:zs} & a task-score benefit; transfer outside the settings of Table~\ref{tab:zs} \\
    \bottomrule
  \end{tabularx}
\end{table}
\begin{table}[!htbp]
  \centering\footnotesize
  \caption{Claim, evidence and boundary, continued from Table~\ref{tab:app-claims-two}.}
  \label{tab:app-claims-three}
  \setlength{\tabcolsep}{3pt}
  \begin{tabularx}{\linewidth}{@{}>{\raggedright\arraybackslash}p{0.30\linewidth} >{\raggedright\arraybackslash}X >{\raggedright\arraybackslash}X@{}}
    \toprule
    Claim & Evidence & Not established \\
    \midrule
    Evidence attention: \mistill{} students' own chapter-decision heads attend to the human-annotated evidence for each code more, and more specifically, than \vanillakd{} students', on CodiEsp-X (zero-shot) and MDACE (in domain); lift \pci{\NEaCxLiftMcMean}{\NEaCxLiftMcLo}{\NEaCxLiftMcHi} and \pci{\NEaMdLiftMcMean}{\NEaMdLiftMcLo}{\NEaMdLiftMcHi} & Sec.~\ref{sec:distill-evidence}; Appendix~\ref{app:evidence}, Tables~\ref{tab:app-evattn} and~\ref{tab:app-evattn-levels} & that the decision depends on that evidence more (the deletion test: \pci{\NEdCxDeltaMcMean}{\NEdCxDeltaMcLo}{\NEdCxDeltaMcHi}, below the pre-set bar, and \pci{\NEdMdDeltaMcMean}{\NEdMdDeltaMcLo}{\NEdMdDeltaMcHi}, Table~\ref{tab:app-evdel}); specificity beyond random-head supervision (shared); more quoting of the evidence in the students' own generations (not detected); the students' own traces; other student families \\
    Scope & two reasoners of one lineage; one tokenizer; the clinical characterization and distillation results on one clinical task, plus declared public-task extensions (Appendices~\ref{app:musique} and~\ref{app:tasks2}) a zero-shot transfer evaluation (Appendix~\ref{app:zeroshot}) and evidence reads on two annotated corpora (Appendix~\ref{app:evidence}); a retriever ceiling shared by every clinical arm & other label spaces, lineages and presentations of the codes; generality beyond the tested adaptations \\
    \bottomrule
  \end{tabularx}
\end{table}

One entry per result identifier; the code is at \url{\anoncodeurl}, whose
\texttt{results/} folder lists every reported number with the statistic it is and the artifact and reader it comes from. Data require PhysioNet credentialed
access to \mimiciv{} (Appendix~\ref{app:data}); no note or trace text is released. Reruns reproduce the reported statistics within the stated bands, not bit-exactly.

\paragraph{Layout.} Code is in \path|src/mistill/| (events, buckets, losses, the objective, targets, the intervention
hook), \path|scripts/| (training, evaluation, attribution and patching passes, readers) and \path|slurm/| (job files and
the per-cluster site settings). Artifacts are per-experiment directories under \path|artifacts/| (paths below are
relative to it); the experiment logbook \path|LOGBOOK.md| holds every run record with its job identifiers and every
incident. Each training run writes \path|config.json|; each evaluation directory holds \path|metrics.json| (with the
decoding settings), \path|predictions.jsonl|, \path|las.json|, \path|mechanism.json| (M1) and \path|generations/|; each
own-heads cell \path|M6-<arm>/| holds \path|faithfulness2.json| (M4), \path|readside.json| (M2, M3, prefix share) and 150
per-note files, beside the arm's ranking in \path|HD-S-<arm>/head_sets_contrast.json|. Every number and generated table
in the paper is written by \path|scripts/paper_numbers.py|, which wraps the readers named below and records its inputs
in \path|paper_numbers.log|.

\paragraph{Data and pipeline.} The distillation corpus (500 training and 150 test notes, one per patient) is built by
\path|scripts/build_dt_corpus.py| and presented on the retrieved shortlist by \path|scripts/build_mimic_shortlist_set.py|
(\path|data/mimic4/dtk|; the validation notes \path|data/mimic4/tr006k|). Teacher traces come from
\path|scripts/pilot_teacher_mimic.py| (\path|TR-007/| validation, \path|TR-008/| corpus), teacher targets from
\path|scripts/precompute_targets.py| (with \appflag{mech} for \mistill). Training: \path|slurm/dt_grid.sh| (baselines),
\path|slurm/dt_opd.sbatch| (\opd, \gold), \path|slurm/dt001_grid.sh| (\mistill, pilot, tier 1, and
\texttt{MODE=ablations} with the arm's one flag); evaluation: \path|slurm/dt_eval.sbatch|
(\path|scripts/eval_student.py|, \path|scripts/las_simulator.py|, \path|scripts/eval_mechanism.py|); own heads:
\path|slurm/dt001_s6_arm.sbatch| (attribution on the trained checkpoint, then \path|scripts/knockin_bins.py| with
\appflag{no-attr} at bins 8 and 40, then \path|scripts/read_side.py| with \appflag{no-attr}, and
\path|scripts/prose_locus.py| for the M5 arms).

\paragraph{Compute.} One tier-1 training run takes about \NMDWallVanillaTierOneHours\,h for fp32 \vanillakd{} and
\NMDWallMistillTierOne{} for \mistill{} on one GPU (Appendix~\ref{app:obj-cost}); an evaluation with M1 and \las{}
takes \NMDEvalHoursPerArm\,h on an RTX PRO 6000; an own-heads chain takes about \NMDOwnHeadsChainHoursSmol\,h per \smol{} arm and \NMDOwnHeadsChainHoursQwen\,h per
\qwenfour{} arm.

\paragraph{Nondeterminism classes, cited per entry.} (G) Generation samples at temperature 0.6 with a per-note seed in
vLLM; a fixed seed does not reproduce tokens across GPUs, library builds or batch composition, so task, fidelity and
validity reproduce within the seed and replicate bands, not token for token. (K) Teacher-forced passes (margins,
attribution, knock-in) vary slightly across GPUs: the teacher's forward-pass noise is \NMNoiseFloorNats{} nats
(s.d.), one clean margin differs by a median of \NMDNoAttrCleanMarginDiff{} nats between two GPU types, and one
own-heads cell read on two GPUs differs per note by \NMDCrossHwPTwoCDelta, \NMDCrossHwPTwoDelta{} and \NMDCrossHwPOneDelta{} at \Ptwoc, \Ptwo{} and \Pone with an identical top-40. (T)
Training: a same-seed rerun on other hardware differs by about \NMDReplicateNoiseMicroF{} \microf,
\NMDReplicateNoiseMOnePTwoC{} M1 at \Ptwoc{} and \NMDReplicateNoiseLAS{} \las; resumed runs restore every
random-number state. (F) The \featkd{} runs are not bit-reproducible (Appendix~\ref{app:base-arms}).

\begin{description}[style=unboxed,leftmargin=0pt,itemsep=3pt,parsep=1pt,topsep=4pt,font=\normalfont\small\sffamily\bfseries]
\item[\residtarget{R-C1}\quad Decisions are token-level events.] Sec.~\ref{sec:setup-events}, Appendix~\ref{app:instrument}, Table~\ref{tab:events-app};
Appendix~\ref{app:instrument}. \emph{Artifacts:} \path|HD-001/<note>/events.jsonl|, \path|HD-001/READOUT.log| (audit
block). \emph{Reader:} \path|scripts/count_events.py|, \path|scripts/paper_numbers.py|. \emph{Seeds:} one teacher trace
per note. \emph{Expected:} \NEvPOneCount{} \Pone, \NEvPTwoCount{} \Ptwo, \NEvPTwoCCount{} \Ptwoc{} and \NEvNullCount{}
\Pnull{} events; median \Pone{} margin \NEvPOneMargin{} nats. Counts are exact given the stored traces; margins within (K).

\item[\residtarget{R-C2}\quad The attribution screen and the hook.] Sec.~\ref{sec:setup-measure}; Appendices~\ref{app:instrument}
and~\ref{app:teacher}. \emph{Artifacts:} \path|HD-001/ANALYSIS.log|, per-note \path|attr.npz|; the detached-graph incident
in the logbook. \emph{Reader:} \path|scripts/attribute_heads.py| (self-test mode), \path|scripts/read_attribution.py|. \emph{Seeds:} one teacher trace per
note. \emph{Expected:} correlation with exact effects \NMAttrPearsonAllDetached{} under the detached hook and
\NMAttrPearsonAllIntact{} with the graph intact (\NMAttrPearsonAboveNoiseDetached{} and
\NMAttrPearsonAboveNoiseIntact{} above noise); noise floor \NMNoiseFloorNats{} nats; all-heads
knock-in equal to the clean margin (median difference \NMKnockinCeilingDiff{} nats). \emph{Nondeterminism:} K.

\item[\residtarget{R-C3}\quad Global sets are necessary and sufficient.] Sec.~\ref{sec:char-sets}, Table~\ref{tab:faithfulness} (every $k$: Table~\ref{tab:faithfulness-app});
Appendix~\ref{app:heads}. \emph{Artifacts:} \path|HD-001/faithfulness2.json|, \path|HD-001/KNOCKIN.log|, per-note
\path|patch2.jsonl|. \emph{Reader:} \path|scripts/knockin_bins.py|, \path|scripts/read_knockin.py|. \emph{Seeds:} one teacher trace per note;
random sets drawn per bin. \emph{Expected:}
necessity of the top-8 \NFaithPOneNecKEightGlobal, \NFaithPTwoNecKEightGlobal{} and \NFaithPTwoCNecKEightGlobal{}
(\Pone, \Ptwo, \Ptwoc); knock-in of the top-40 \NFaithPOneKiKFortyGlobal, \NFaithPTwoKiKFortyGlobal{} and
\NFaithPTwoCKiKFortyGlobal{} against random \NFaithPOneKiKFortyRandom, \NFaithPTwoKiKFortyRandom{} and
\NFaithPTwoCKiKFortyRandom; knock-in above random on \NFaithPOneKiKFortyWins, \NFaithPTwoKiKFortyWins{} and
\NFaithPTwoCKiKFortyWins{} notes. Medians over notes. \emph{Nondeterminism:} K.

\item[\residtarget{R-C4}\quad Phase structure.] Sec.~\ref{sec:char-phases}, Table~\ref{tab:heads-app}; Appendix~\ref{app:heads}.
\emph{Artifacts:} \path|HD-001/head_sets_contrast.json|, \path|HD-001/PHASE_TABLE.log|, \path|HD-001/faithfulness2.json|
(exact \Ptwoc{} refinement). \emph{Reader:} \path|scripts/read_attribution.py| with \appflag{contrast},
\path|scripts/read_knockin.py|. \emph{Seeds:} one teacher trace per note. \emph{Expected:} \head{39}{38} first in every phase ($I_c$
\NHeadLThirtyNineHThirtyEightPOneIc, \NHeadLThirtyNineHThirtyEightPTwoIc, \NHeadLThirtyNineHThirtyEightPTwoCIc);
\head{30}{11} rank \NHeadLThirtyHElevenPTwoCRank{} at \Ptwoc{} and \NHeadLThirtyHElevenPOneRank{} at \Pone;
\NHeadPOneTopFortyLayersTwentyFourToThirtyOneCount{} of the chapter top-40 in layers 24--31. \Ptwo{} and \Ptwoc{}
rankings are first-order. \emph{Nondeterminism:} K.

\item[\residtarget{R-C5}\quad What the heads read (descriptive).] Sec.~\ref{sec:char-read}, Figure~\ref{fig:readside};
Appendix~\ref{app:readside}. \emph{Artifacts:} \path|HD-001/readside.json|, \path|HD-001/read_panel.json|, per-note
\path|read.npz|. \emph{Reader:} \path|scripts/read_side.py|, \path|scripts/read_readout.py|. \emph{Seeds:} one teacher trace per note. \emph{Expected:}
\head{26}{36} region share \NReadLTwentySixHThirtySixRegion, lift \NReadLTwentySixHThirtySixLift, correlation with
importance \NReadLTwentySixHThirtySixR; \head{39}{38} prose share \NReadLThirtyNineHThirtyEightProse; \head{30}{11} on the
near-miss's mentions \NReadLThirtyHElevenSibling{} (random \NReadPTwoCRandSibling). Not intervened on.
\emph{Nondeterminism:} K.

\item[\residtarget{R-C6}\quad The prose locus.] Sec.~\ref{sec:char-locus}, Figure~\ref{fig:locus}; Appendix~\ref{app:locus}.
\emph{Artifacts:} \path|HD-001/locus.json|, \path|HD-001/locus_sets.json|, \path|HD-001/head_sets_locus.json|,
\path|HD-001/read_locus_panel.json|. \emph{Reader:} \path|scripts/prose_locus.py| and \path|scripts/read_locus.py| (pass
A), \path|scripts/locus_sets.py| and \path|scripts/read_locus_sets.py| (pass B). \emph{Seeds:} one teacher trace per note. \emph{Expected:} \NLocusWindowPct{} of
mentions decided in the window (\NLocusInheritedPct{} inherited, \NLocusProbeFailPct{} never decisive), median
\NLocusDMedian{} tokens before the letter; prose-time top-8 window-only \NLocusProseWinKEight{} (random
\NLocusRandWinKEight) on \NLocusProseWinKEightWins{} of 100 notes against the emission top-16's
\NLocusEmissionWinKSixteen; prose-time heads on the note \NLocusReadProseNote. \emph{Nondeterminism:} K.

\item[\residtarget{R-C7}\quad Largely preserved on the shortlist.] Sec.~\ref{sec:char-stability}; Appendix~\ref{app:hd002}.
\emph{Artifacts:} \path|HD-002/compare.json| and the HD-002 readouts. \emph{Reader:} the R-C3--R-C6 pipeline on
\path|tr006k| / \path|TR-007|, then \path|scripts/compare_hd.py|. \emph{Seeds:} one new teacher trace per note. \emph{Expected:} Spearman \NShortPOneSpearman,
\NShortPTwoSpearman, \NShortPTwoCSpearman; top-40 overlap \NShortPOneOverlapForty, \NShortPTwoOverlapForty,
\NShortPTwoCOverlapForty; top-40 knock-in \NShortPOneKiKFortyShort, \NShortPTwoKiKFortyShort, \NShortPTwoCKiKFortyShort;
locus \NShortLocusDMedianShort{} tokens (full inventory \NShortLocusDMedianFull); prose-time Spearman
\NShortLocusProseSpearman; the chapter top-8 knock-in shifts (\NShortPOneKiKEightShort{} against
\NShortPOneKiKEightFull). \emph{Nondeterminism:} G (new traces), K.

\item[\residtarget{R-D0}\quad Setup.] Sec.~\ref{sec:distill-setup}; Appendices~\ref{app:data} and~\ref{app:baselines}.
\emph{Artifacts:} \path|RT-003/bakeoff_*.json|, \path|DT-000/tokenizer_audit.json|, \path|TR-007/|, \path|TR-008/|.
\emph{Reader:} \path|scripts/build_mimic_shortlist_set.py| (recall sweep), \path|scripts/tokenizer_audit.py|. \emph{Seeds:} teacher sampling seed 0.
\emph{Expected:} retriever recall \NMRetrieverRecallVal{} (validation), \NMRetrieverRecallTest{}
(test split) and \NMDRetrieverRecallCorpus{} (the 650-note corpus); teacher \microf{} \NMDTeacherValMicroF{} (validation) and \NMTeacherShortlistTestMicroF{} (test); 493
of 500 training traces valid. \emph{Nondeterminism:} G (teacher traces).

\item[\residtarget{R-D1}\quad Untrained students.] Table~\ref{tab:main}. \emph{Artifacts:} \path|DT-000/eval/Qwen3-4B__base__s0/|,
\path|DT-000/eval/SmolLM3-3B__base__s0/|, linked with every main-table row in \path|DT-001/eval_main/|. \emph{Reader:}
\path|scripts/read_dt.py| over \path|DT-001/eval_main| with \appflag{reference vanilla} \appflag{markdown}. \emph{Seeds:} none (no training). \emph{Expected:}
\microf{} \NQwenBaseMicroF{} (valid \NQwenBaseValid) and \NSmolBaseMicroF{} (valid \NSmolBaseValid).
\emph{Nondeterminism:} G.

\item[\residtarget{R-D2}\quad \vanillakd{} (fp32).] Table~\ref{tab:main}; Appendix~\ref{app:res-paired}. \emph{Artifacts:}
\path|DT-000/{runs,eval}/Qwen3-4B__vanilla__s{0,1,2}/|, \path|DT-000/{runs,eval}/SmolLM3-3B__vanilla__s0/|,
\path|DT-001/{runs,eval}/SmolLM3-3B__vanilla__s1/|. \emph{Reader:} as R-D1. \emph{Seeds:} 3 and 2. \emph{Expected:}
\qwenfour{} \pmsd{\NQwenVanillaMicroF}{\NQwenVanillaMicroFSd}, against the untrained student
\NQwenVanillaMicroFVsBaseDelta{} [\NQwenVanillaMicroFVsBaseLo, \NQwenVanillaMicroFVsBaseHi], gap recovered
\NMDQwenVanillaGapClosed, \las{} \NQwenVanillaLAS; \smol{} \pmsd{\NSmolVanillaMicroF}{\NSmolVanillaMicroFSd}
(\NSmolVanillaSZeroMicroF{} and \NSmolVanillaSOneMicroF). \emph{Nondeterminism:} G, T.

\item[\residtarget{R-D3}\quad \featkd{} below \vanillakd.] Table~\ref{tab:main}; Appendix~\ref{app:base-arms}. \emph{Artifacts:}
\path|DT-000/{runs,eval}/Qwen3-4B__feature_kd__s{0,1,2}/|, \path|DT-000/{runs,eval}/SmolLM3-3B__feature_kd__s0/|.
\emph{Reader:} as R-D1. \emph{Seeds:} 3 and 1. \emph{Expected:} \pmsd{\NQwenFeatKDMicroF}{\NQwenFeatKDMicroFSd},
\NQwenFeatKDMicroFVsVanillaDelta{} [\NQwenFeatKDMicroFVsVanillaLo, \NQwenFeatKDMicroFVsVanillaHi] against fp32
\vanillakd, \las{} \NQwenFeatKDLASVsVanillaDelta; \smol{} \NSmolFeatKDMicroF, \NSmolFeatKDMicroFVsVanillaDelta{}
[\NSmolFeatKDMicroFVsVanillaLo, \NSmolFeatKDMicroFVsVanillaHi]. \emph{Nondeterminism:} G, T, F.

\item[\residtarget{R-D4}\quad \distillm, \opd{} and \gold.] Table~\ref{tab:main}; Appendix~\ref{app:res-dissoc}. \emph{Artifacts:}
\path|DT-000/{runs,eval}/Qwen3-4B__{distillm2,opd}__s0/|, \path|DT-000/eval/SmolLM3-3B__opd__s0/| (the \gold{} arm; its
checkpoint \path|DT-000/runs/SmolLM3-3B__opd__s0.podcopy/final/|). \emph{Reader:} as R-D1. \emph{Seeds:} 1 each; \opd{} and \gold{} train bf16 weights.
\emph{Expected:} \distillm{} \NQwenDistillmMicroF{} (\NQwenDistillmMicroFVsVanillaDelta, $p$
\NQwenDistillmMicroFVsVanillaP); \opd{} \NQwenOPDMicroF, \las{} \NQwenOPDLASVsVanillaDelta{} ($p$
\NQwenOPDLASVsVanillaP), \bleu{} \NQwenOPDBLEU, valid \NQwenOPDValid; \gold{} \NSmolGOLDMicroF{} with M1 at \Ptwo{}
\NSmolGOLDMOnePTwo. \emph{Nondeterminism:} G, T (on-policy rollouts are sampled).

\item[\residtarget{R-D5}\quad \mistill{} at tier 1.] Sec.~\ref{sec:distill-main}, Table~\ref{tab:main}; Appendix~\ref{app:results}.
\emph{Artifacts:} \path|DT-001/{runs,eval}/Qwen3-4B__mech__s{0,1,2}/|, \path|DT-001/{runs,eval}/SmolLM3-3B__mech__s{0,1,2}/|.
\emph{Reader:} as R-D1, and \path|scripts/pilot_compare.py| (\texttt{pooled}) for the seed-matched M1 counts. \emph{Seeds:} 3 and 3 (the third \smol{} seed: R-D27).
\emph{Expected:} \qwenfour{} \pmsd{\NQwenMistillMicroF}{\NQwenMistillMicroFSd} against bf16 \vanillakd's
\pmsd{\NQwenVanillaBfMicroF}{\NQwenVanillaBfMicroFSd} (per note \NQwenVanillaBfMicroFVsMistillDelta{}
[\NQwenVanillaBfMicroFVsMistillLo, \NQwenVanillaBfMicroFVsMistillHi]); \smol{}
\pmsd{\NSmolMistillMicroF}{\NSmolMistillMicroFSd} against \NSmolVanillaBfMicroF{} (\NSmolVanillaBfMicroFVsMistillDelta{}
[\NSmolVanillaBfMicroFVsMistillLo, \NSmolVanillaBfMicroFVsMistillHi]); M1 at \Ptwoc{} \NQwenMistillMOnePTwoC{} against
\NQwenVanillaBfMOnePTwoC{} (bf16) and \NQwenVanillaMOnePTwoC{} (fp32). \emph{Nondeterminism:} G, T.

\item[\residtarget{R-D6}\quad Mechanism on each arm's own heads.] Sec.~\ref{sec:distill-mech}, Table~\ref{tab:mech};
Appendix~\ref{app:metrics}. \emph{Artifacts:} \path|M6-<arm>/|, \path|HD-S-<arm>/|, \path|HD-002/| (teacher row).
\emph{Reader:} \path|scripts/read_m4.py| with \appflag{prefix} \texttt{M6-} \appflag{teacher} \path|artifacts/HD-002|
\appflag{only}; \path|scripts/read_mech.py| with \appflag{paired}; \path|scripts/m4_paired.py| with \appflag{cluster-notes} (the
pooled note-pair counts of the submission are superseded by R-D18);
\path|scripts/read_shares.py|. \emph{Seeds:} as the arms' (Table~\ref{tab:mech}); pooling pairs seed $i$ with seed $i$.
\emph{Expected:} untrained \qwenfour{} \NOwnQwenBasePTwoCKiForty{} (teacher
\NOwnTeacherPTwoCKiForty), untrained \smol{} \NOwnSmolBasePTwoCKiForty{} (floor \NOwnSmolBasePTwoCKiFortyRand); \mistill{}
minus fp32 \vanillakd, seed-matched and note-clustered: \qwenfour{} \NPoolQwenMistillVsVanillaPTwoCClusterDelta{}
(\NPoolQwenMistillVsVanillaPTwoCClusterHigher/\NPoolQwenMistillVsVanillaPTwoCClusterLower of 150 notes, $p$ \NPoolQwenMistillVsVanillaPTwoCClusterP, descriptive), M2
\NPoolQwenMistillVsVanillaMTwoPTwoCDelta{} at \Ptwoc{} and \NMDPoolQwenMistillVsVanillaMTwoPOneDelta{} (\NMDPoolQwenMistillVsVanillaMTwoPOneHigher/\NMDPoolQwenMistillVsVanillaMTwoPOneLower, $p$ \NMDPoolQwenMistillVsVanillaMTwoPOneP) at \Pone; \smol{}
\NPoolSmolMistillVsVanillaPTwoCClusterDelta{} (higher on \NPoolSmolMistillVsVanillaPTwoCClusterHigher{} of 150 notes).
Against bf16 \vanillakd: \NPoolQwenMistillVsVanillaBfPTwoCClusterDelta{} at \Ptwoc{} (\NPoolQwenMistillVsVanillaBfPTwoCClusterHigher/\NPoolQwenMistillVsVanillaBfPTwoCClusterLower), \NPoolQwenMistillVsVanillaBfPTwoClusterDelta{} at \Ptwo; \smol{}
\NPoolSmolMistillVsVanillaBfPTwoCClusterDelta. \emph{Nondeterminism:} K, T.

\item[\residtarget{R-D7}\quad Positions and heads: the controls.] Secs.~\ref{sec:distill-mech} and~\ref{sec:distill-ablation},
Figure~\ref{fig:ablation}; Appendix~\ref{app:ablations}. \emph{Artifacts:}
\path|DT-001/{runs,eval}/{Qwen3-4B,SmolLM3-3B}__mech_{randheads,randpos}_t1__s0/|,
\path|DT-001/{runs,eval}/Qwen3-4B__mech_randboth_t1__s0/|, and their \path|M6-| and \path|HD-S-| cells. \emph{Reader:} as
R-D6. \emph{Seeds:} one per control. \emph{Expected:} \randpos{} \NOwnQwenRandPosSZeroPTwoCKiForty{} and
\NOwnSmolRandPosSZeroPTwoCKiForty{} (untrained \NOwnQwenBasePTwoCKiForty{} and \NOwnSmolBasePTwoCKiForty); \randheads{}
\NOwnQwenRandHeadsSZeroPTwoCKiForty{} (\qwenfour) and \NOwnSmolRandHeadsSZeroPTwoCKiForty{} (\smol; \mistill{}
\NOwnSmolMistillSZeroPTwoCKiForty); RandBoth \NOwnQwenRandBothSZeroPTwoCKiForty. \emph{Nondeterminism:} K, T.

\item[\residtarget{R-D8}\quad Training-set size.] Sec.~\ref{sec:distill-main}, Figure~\ref{fig:curve};
Appendix~\ref{app:abl-curve}. \emph{Artifacts:} \path|DT-001/eval/Qwen3-4B__{mech,vanilla}_c100__s0/|,
\path|DT-001/eval/Qwen3-4B__{mech,vanilla}_250__s0/|, the tier-1 rows and \path|DT-001/eval/Qwen3-4B__vanilla_bf16__s{0,1,2}/|.
\emph{Reader:} \path|scripts/pilot_compare.py| (\texttt{paired}, \texttt{pooled}); the figure from
\path|scripts/fig_curve.py|. \emph{Seeds:} 1 at 100 and 250 notes, 3 at 493. \emph{Expected:} Table~\ref{tab:app-curve}.
\emph{Nondeterminism:} G, T.

\item[\residtarget{R-D9}\quad Training cost.] Sec.~\ref{sec:limitations}; Appendix~\ref{app:obj-cost}. \emph{Artifacts:} each run's
log and \path|metrics.jsonl| (logged peak memory and step times); the logbook's run records. \emph{Seeds:} not applicable. \emph{Expected:} peak
\NMDMistillFitPeakGiB\,GiB and about \NMDMistillFitSecPerNote\,s per note on one GPU without an online teacher;
at 100 notes \NMDWallVanillaHundredNotesMin\,min for \vanillakd{} against \NMDWallMistillHundredNotesMin\,min for
\mistill; \NMDWallMistillTierOne{} for a 493-note \mistill{} run. Hardware-dependent; indicative only.

\item[\residtarget{R-D10}\quad The objective's selection on the validation notes.] Sec.~\ref{sec:distill-objective};
Appendix~\ref{app:objective}. \emph{Artifacts:} \path|DT-001/eval/<pilot-arm>/| for the grid
(\path|Qwen3-4B__mech_l<a>_<b>__s0|), the write-term forms and the selected arm on three seeds, \path|vanilla_100__s{0,1,2}|,
\path|Qwen3-4B__base__val| and \path|Qwen3-14B__teacher__val| (the M1 anchor); \path|DT-001/M2val-<arm>/readside.json|.
\emph{Reader:} \path|scripts/pilot_compare.py| (\texttt{breakdown}, \texttt{paired}, \texttt{pooled}, \texttt{m2}),
\path|scripts/select_lambda.py|, \path|scripts/diag_mech_terms.py| (write-term diagnostics), \path|slurm/m2val.sh|.
\emph{Seeds:} 1, and 3 for the selected pair. \emph{Expected:} Tables~\ref{tab:app-write}--\ref{tab:app-select}; one-seed
noise \NMDSelOneSeedNoiseMicroF{} on validation \microf. \emph{Nondeterminism:} G, T.

\item[\residtarget{R-D11}\quad Extra tier-1 cells.] Appendix~\ref{app:res-extra}. \emph{Artifacts:}
\path|DT-001/eval/Qwen3-4B__mech_l0.03_0.1__s0/|, \path|DT-001/eval/Qwen3-4B__mech__s2_gh200/| (the same-seed replicate),
\path|DT-001/eval/SmolLM3-3B__mech_attn__s0/|. \emph{Seeds:} 1 each. \emph{Reader:} \path|scripts/read_dt.py|; \path|scripts/pilot_compare.py| (\texttt{paired}) for the replicate. \emph{Expected:} \NQwenLowPOneDoseMicroF; replicate \NMDReplicateSTwoMicroF{} against
\NQwenMistillSTwoMicroF; \smol{} attention only \NSmolAttnTermMicroF. The replicate's differences define (T).

\item[\residtarget{R-D12}\quad Reproducible recovery magnitude.] Secs.~\ref{sec:distill-mech} and~\ref{sec:distill-ablation},
Table~\ref{tab:mech}. \emph{Artifacts:} \path|M6-<arm>/| of every seed. \emph{Reader:} \path|scripts/m4_paired.py|.
\emph{Seeds:} 3 and 3 (\mistill), 3 and 2 (\vanillakd), 2 (\randheads, bottom-40; 4 \randheads{} runs with R-D26). \emph{Expected:} \mistill's \qwenfour{} seeds \NOwnQwenMistillSZeroPTwoCKiForty, \NOwnQwenMistillSOnePTwoCKiForty,
\NOwnQwenMistillSTwoPTwoCKiForty{} against fp32 \vanillakd's \NOwnQwenVanillaSZeroPTwoCKiForty,
\NOwnQwenVanillaSOnePTwoCKiForty, \NOwnQwenVanillaSTwoPTwoCKiForty; \smol{} means over notes
\NPairSmolMistillSOneVsMistillSZeroPTwoCMeanB{} and \NPairSmolMistillSOneVsMistillSZeroPTwoCMeanA{} (\mistill) against
\NPairSmolVanillaSOneVsVanillaSZeroPTwoCMeanB{} and \NPairSmolVanillaSOneVsVanillaSZeroPTwoCMeanA{} (\vanillakd); \randheads'
seeds differ by \NPairQwenRandHeadsSOneVsRandHeadsSZeroPTwoCDelta, \smol{} bottom-40's by
\NPairSmolBottomSOneVsBottomSZeroPTwoCDelta{} ($p$ \NPairSmolBottomSOneVsBottomSZeroPTwoCP). bf16 \vanillakd's three \qwenfour{} seeds
spread \NOwnQwenVanillaBfSeedSpreadPTwoC{} at \Ptwoc{} (\mistill's \NOwnQwenMistillSeedSpreadPTwoC). \emph{Nondeterminism:} K, T.

\item[\residtarget{R-D13}\quad Which heads, how many, and where the mechanism lands.] Sec.~\ref{sec:distill-ablation},
Figure~\ref{fig:ablation}; Appendix~\ref{app:abl-carriers}. \emph{Artifacts:}
\path|DT-001/{runs,eval}/Qwen3-4B__mech_{bottom,k8,k100}_t1__s0/|,
\path|DT-001/{runs,eval}/SmolLM3-3B__mech_{bottom_t1__s0,bottom_t1__s1,k8_t1__s0}/|, their \path|M6-| and \path|HD-S-|
cells, and \path|HD-S-{Qwen3-4B,SmolLM3-3B}/| for the overlaps. \emph{Reader:}
\texttt{STUDENT=<student> bash} \path|scripts/read_ablations.sh| (wrapping \path|scripts/read_m4.py|,
\path|scripts/read_mech.py|, \path|scripts/read_shares.py|, \path|scripts/m4_paired.py|, \path|scripts/read_dt.py| with
\appflag{compact}, and \path|scripts/head_overlap.py|). \emph{Seeds:} 1 (\smol{} bottom-40: 2). \emph{Expected:} \qwenfour{} bottom-40
\NOwnQwenBottomSZeroPTwoCKiForty; \smol{} bottom-40 \NOwnSmolBottomSZeroPTwoCKiForty{} and
\NOwnSmolBottomSOnePTwoCKiForty; $k=8$ \NOwnQwenKEightSZeroPTwoCKiForty{} (\qwenfour) and \NOwnSmolKEightSZeroPTwoCKiForty{}
(\smol); $k=100$ \NOwnQwenKHundredSZeroPTwoCKiForty; carriers \NMCarrierOverlapRange{} of 40.
\emph{Nondeterminism:} K, T.

\item[\residtarget{R-D14}\quad Phases, terms and dose.] Sec.~\ref{sec:distill-ablation}, Figure~\ref{fig:ablation};
Appendix~\ref{app:abl-tiers}. \emph{Artifacts:}
\path|DT-001/{runs,eval}/Qwen3-4B__mech_{p1only_lam0,p2only_lam0,attnonly,interonly}_t1__s0/|,
\path|DT-001/{runs,eval}/Qwen3-4B__mech_l0.03_0.03__s0/|, \path|DT-001/{runs,eval}/Qwen3-4B__mech_randheads_t1__s1/|,
their \path|M6-| and \path|HD-S-| cells. \emph{Reader:} as R-D13. \emph{Seeds:} 1 (\randheads: 2). \emph{Expected:} \Pone-only
\NOwnQwenPOneOnlySZeroPTwoCKiForty, \Ptwo-only \NOwnQwenPTwoOnlySZeroPTwoCKiForty, attention-only
\NOwnQwenAttnOnlySZeroPTwoCKiForty, interaction-only \NOwnQwenInterOnlySZeroPTwoCKiForty, $\lambda$ 0.03\,/\,0.03
\NOwnQwenLowDoseSZeroPTwoCKiForty; task in Table~\ref{tab:app-ablate}. \emph{Nondeterminism:} K, T.

\item[\residtarget{R-D15}\quad Readouts that do not discriminate.] Sec.~\ref{sec:distill-mech}; Appendix~\ref{app:metrics}.
\emph{Artifacts:} \path|M6-<arm>/readside.json|, \path|M6-<arm>/locus.json| (M5 arms). \emph{Reader:}
\path|scripts/read_mech.py|, \path|scripts/read_shares.py|. \emph{Seeds:} as R-D6; M5 one seed per arm. \emph{Expected:} \Pone{} recovery \NOwnQwenBasePOneKiForty{} and
\NOwnSmolBasePOneKiForty{} untrained; M5 \NOwnQwenBaseMFiveLocus, \NOwnQwenVanillaSZeroMFiveLocus{} and
\NOwnQwenMistillSZeroMFiveLocus{} against the teacher's \NShortLocusDMedianShort; M3 on \smol{}
\NOwnSmolBaseMThreeNearMiss{} to \NOwnSmolMistillSZeroMThreeNearMiss. \emph{Nondeterminism:} K.

\item[\residtarget{R-D16}\quad Instrument checks.] Appendix~\ref{app:res-checks}. \emph{Artifacts:}
\path|_race_h200pod.HD-S-SmolLM3-3B__mech_k8_t1__s0/|, \path|_race_h200pod.M6-SmolLM3-3B__mech_k8_t1__s0/|; the gate's
cell was not retained (its numbers are in the logbook). \emph{Reader:} link the two copies side by side and run
\path|scripts/read_m4.py| and \path|scripts/m4_paired.py| (pair \texttt{k8\_h200:k8\_rtx}); \path|scripts/gate_noattr.py|
for the gate. \emph{Seeds:} one cell each. \emph{Expected:} largest gate difference \NMDNoAttrGateMaxDiff; cross-hardware per-note
differences \NMDCrossHwPTwoCDelta, \NMDCrossHwPTwoDelta{} and \NMDCrossHwPOneDelta{} at \Ptwoc, \Ptwo{} and \Pone. \emph{Nondeterminism:} K.

\item[\residtarget{R-D17}\quad Precision control.] Secs.~\ref{sec:distill-setup} and~\ref{sec:distill-main}, Tables~\ref{tab:main}
and~\ref{tab:mech}; Appendix~\ref{app:res-precision}. \emph{Artifacts:}
\path|DT-001/{runs,eval}/Qwen3-4B__vanilla_bf16__s{0,1,2}/|, \path|DT-001/{runs,eval}/SmolLM3-3B__vanilla_bf16__s0/|,
their \path|HD-S-| and \path|M6-| cells, \path|DT-001/{runs,eval}/Qwen3-4B__mech_fp32__s{0,1,2}/|.
\emph{Reader:} \path|scripts/read_precision.sh|; \path|scripts/pilot_compare.py| (\texttt{pooled}). \emph{Seeds:} 3 and 1; fp32
\mistill{} 3 at the bf16-selected $\lambda$ (its own heads and the $\lambda$ re-selected for fp32: R-D22). \emph{Expected:} Table~\ref{tab:app-precision} (bf16 \vanillakd{} minus fp32
\NQwenVanillaBfMicroFVsVanillaDelta{} on \qwenfour, \NSmolVanillaBfMicroFVsVanillaDelta{} on \smol); training
cross-entropy as in Table~\ref{tab:app-ce}; own heads as in Table~\ref{tab:mech}; fp32 \mistill{} \NQwenMistillFpMicroF{}
(\NQwenMistillFpMicroFVsVanillaDelta{} against fp32 \vanillakd).
\emph{Nondeterminism:} G, T, K.

\item[\residtarget{R-C8}\quad The character ranking under the intact hook.] Appendix~\ref{app:instrument},
Table~\ref{tab:app-p2intact}. \emph{Artifacts:} \path|REBUTTAL/W3.4/HD-001-intact/| (50 notes, exact \Ptwo{} patches),
\path|REBUTTAL/W3.4/compare_intact_vs_cut_50notes.json|. \emph{Reader:} \path|scripts/attribute_heads.py| (mode full,
\appflag{exact-kinds P2}) via \path|slurm/hd001_multilane.sbatch|, \path|scripts/read_attribution.py|,
\path|scripts/rebuttal/w34_compare_intact_vs_cut.py|. \emph{Seeds:} one teacher trace per note. \emph{Expected:} Spearman
\NPTwoIntactPTwoSpearman{} and \NPTwoIntactPTwoTopForty{} of 40 shared at \Ptwo; first-order vs exact-refined
\NPTwoIntactFirstOrderVsExactSpearman. \emph{Nondeterminism:} K.

\item[\residtarget{R-C9}\quad The contrastive events, audited.] Sec.~\ref{sec:setup-events}, Appendix~\ref{app:instrument},
Table~\ref{tab:app-p2caudit}; Appendix~\ref{app:readside}, Table~\ref{tab:app-audit-reads}. \emph{Artifacts:}
\path|REBUTTAL/W3.1/audit/| (the blinded sheet, key and labels; enclave only) and \path|REBUTTAL/W3.1/audit_readout.json|.
\emph{Reader:} \path|scripts/audit_frame.py| (the stratified frame, seed 0), \path|scripts/read_audit.py|. \emph{Seeds:}
one draw of \NAuditN{} events. \emph{Expected:} \NAuditRejectedPct{} verified rejections (\NAuditRejectedWeighted{}
weighted), \NAuditEnumerationPct{} enumerations, \NAuditMeaningfulYesPct{} clinically meaningful; the restricted reads of
Table~\ref{tab:app-audit-reads}. Labels are the author's; exact given the stored labels.

\item[\residtarget{R-C10}\quad Blocking what the heads read.] Sec.~\ref{sec:char-read}; Appendix~\ref{app:readside},
Table~\ref{tab:app-keyabl}. \emph{Artifacts:} \path|REBUTTAL/S2-A/teacher/<note>/key_intervene.jsonl|,
\path|REBUTTAL/S2-A/teacher_readout.json|. \emph{Reader:} \path|scripts/key_intervene.py| (the pass; \path|src/mistill/keymask.py|),
\path|scripts/read_key_intervene.py|; \path|tests/test_keymask.py| is the exactness test. \emph{Seeds:} one teacher trace
per note; seeded matched controls. \emph{Expected:} shares \NKeyAblLThirtyHElevenShare{} (\head{30}{11}),
\NKeyAblLTwentyEightHTwentyOneShare{} (\head{28}{21}) and \NKeyAblLTwentySixHThirtySixShare{} (\head{26}{36}); named keys
beat the control on \NKeyAblLThirtyHElevenNoteHigher{} of \NKeyAblLThirtyHElevenNoteN{} and
\NKeyAblLTwentyEightHTwentyOneNoteHigher{} of 100 notes. \emph{Nondeterminism:} K (the empty-mask reference cancels the
kernel's own noise).

\item[\residtarget{R-C11}\quad Editing the earlier mention and the note's words.] Sec.~\ref{sec:char-read};
Appendix~\ref{app:readside}, Tables~\ref{tab:app-nearmiss} and~\ref{tab:app-evidence}. \emph{Artifacts:}
\path|REBUTTAL/W3.2/teacher/<note>/prefix_edit.jsonl|, \path|REBUTTAL/W3.2/teacher_readout.json|;
\path|REBUTTAL/W3.2/evidence_teacher/<note>/evidence_edit.jsonl|, \path|REBUTTAL/W3.2/evidence_readout.json|.
\emph{Reader:} \path|scripts/prefix_edit.py| and \path|scripts/read_prefix_edit.py|; \path|scripts/evidence_edit.py| and
\path|scripts/read_evidence_edit.py|. \emph{Seeds:} one teacher trace per note; the event subset and the replacement codes
are seeded. \emph{Expected:} neutral-edit mean effect \NNearMissTeacherNeutralMean{} nats on \NNearMissTeacherNeutralN{}
events, string-copy controls equal (paired median \NNearMissTeacherNeutralMinusUnrelatedMedian), mediated share
\NNearMissTeacherNeutralMediated; evidence minus control \NEvidenceEvidenceMinusControlNoteMean{} per note. Margins move on
the bf16 logit grid (0.125 nats), so medians sit on it. \emph{Nondeterminism:} K (the identity re-feed is the reference).

\item[\residtarget{R-C12}\quad From elicitable to decided.] Sec.~\ref{sec:char-locus}; Appendix~\ref{app:locus},
Table~\ref{tab:app-follow} and Figure~\ref{fig:resid}. \emph{Artifacts:} \path|REBUTTAL/W3.3/teacher/<note>/follow_through.jsonl|,
\path|REBUTTAL/W3.3/teacher_readout.json|; \path|REBUTTAL/S2-B/teacher/<note>/resid_swap.jsonl|,
\path|REBUTTAL/S2-B/teacher_readout.json|. \emph{Reader:} \path|scripts/follow_through.py| and
\path|scripts/read_follow_through.py|; \path|scripts/resid_swap.py| (\path|src/mistill/resid.py|) and
\path|scripts/read_resid_swap.py|; the figure from \path|scripts/fig_resid.py|. \emph{Seeds:} one teacher trace per note;
the sets are the paper's (pass B's panel and seed). \emph{Expected:} follow-through ratio \NFollowProseSixteenRatio{}
(natural drop \NFollowProseSixteenDNWin{} of forced \NFollowProseSixteenDF); $l^*=$ L\NResidLStar, transport ratio
\NResidRatio, flip layer L\NResidFlipLayer. \emph{Nondeterminism:} K (the clean forced margin drifts by a median of
\NFollowDriftMedian{} nats against pass A's stored value; every comparison is within one run).

\item[\residtarget{R-C13}\quad Robustness of the characterization.] Appendices~\ref{app:instrument} (patients),
\ref{app:heads} (Table~\ref{tab:app-splithalf}) and~\ref{app:locus} (Table~\ref{tab:app-suffix}). \emph{Artifacts:}
\path|REBUTTAL/W0.3/collapse.json|; \path|REBUTTAL/W0.9/split_half_HD-00{1,2}.json|;
\path|REBUTTAL/S2-E/HD-001-suffix{D,A}/| with the paper's \path|HD-001/<note>/locus.jsonl|. \emph{Reader:}
\path|scripts/hd001_collapse_patients.py| (fails if the 100-trace cells stop matching \path|faithfulness2.json|),
\path|scripts/split_half_ranking.py| (200 patient-level splits, seed 0), \path|scripts/prose_locus.py| with
\texttt{SUFFIX=D} / \texttt{A} via \path|slurm/hd001_locus.sbatch| and the per-mention comparison in
\path|scripts/paper_numbers_rebuttal.py|. \emph{Expected:} knock-in medians move by at most \NFaithPatientsMaxKiMove{} per
patient; split-half top-8 overlap \NSplitHalfHDTwoPTwoCTopEight{} of 8; median locus distance \NLocusSuffixBDMedian{}
tokens under every suffix, category agreement \NLocusSuffixDAgree. Exact given the stored records (K for the suffix re-runs).

\item[\residtarget{R-C14}\quad The instruction control.] Sec.~\ref{sec:char-stability}; Appendix~\ref{app:teacher},
Table~\ref{tab:app-neutral}. \emph{Artifacts:} \path|data/mimic4/tr006_w36a/| (the \NNeutralNotes{} prompts),
\path|REBUTTAL/W3.6/TR-A/| (traces), \path|REBUTTAL/W3.6/HD-A/|, \path|REBUTTAL/W3.6/KI-A-orig/|,
\path|REBUTTAL/W3.6/HD-001-view30/|, \path|REBUTTAL/W3.6/w36a_read.json|. \emph{Reader:}
\path|scripts/rebuttal/w36_build_neutral.py| (the prompts), \path|scripts/rebuttal/w36a_chain.sh| (the HD-001 pipeline),
\path|scripts/rebuttal/w36a_read.py|. \emph{Seeds:} the note draw and the teacher sampling, seed 0. \emph{Expected:} the
four criteria of Table~\ref{tab:app-neutral}: original sets beat random on \NNeutralOrigPOneKiKEightWins{} and
\NNeutralOrigPTwoCKiKEightWins{} notes; re-ranked top-40 overlaps \NNeutralOverlapPOneForty{} and \NNeutralOverlapPTwoCForty;
window share \NNeutralNeutralLocusWindowPct; prose-time top-8 shares \NNeutralProseEmissionShared{} with the emission top-8.
\emph{Nondeterminism:} G (new traces), K.

\item[\residtarget{R-C15}\quad A second teacher.] Sec.~\ref{sec:char-stability}; Appendix~\ref{app:second},
Tables~\ref{tab:app-second} and~\ref{tab:app-second-locus}. \emph{Artifacts:} \path|REBUTTAL/S2-C/TR-QWQ/| (100 traces of
\texttt{QwQ-32B}; \path|events_vs_teacher.json|), \path|REBUTTAL/S2-C/HD-QWQ/| (attribution, \path|head_sets*.json|,
\path|faithfulness2.json|, \path|locus.json|, \path|locus_sets.json|), the smoke tests of the two candidates set aside.
\emph{Reader:} the R-C1--R-C6 pipeline unchanged (\path|scripts/pilot_teacher_mimic.py|, \path|slurm/hd001_attribute.sbatch|,
\path|scripts/knockin_bins.py|, \path|scripts/prose_locus.py| with \appflag{gqa-attn}, \path|scripts/locus_sets.py| and
their readers; \path|src/mistill/gqa_attn.py| keeps the 64-layer gradient forwards in memory), launched by
\path|scripts/rebuttal/s2c_stage2.sh|. \emph{Seeds:} teacher sampling seed 0. \emph{Expected:} top-40 knock-in
\NSecondPOneKiKForty, \NSecondPTwoKiKForty, \NSecondPTwoCKiKForty{} against random \NSecondPTwoCKiKFortyRand; window share
\NSecondLocusWindowPct; prose-time top-8 shares \NSecondLocusOverlapEmissionKEight{} head with the emission top-8.
\emph{Nondeterminism:} G, K.

\item[\residtarget{R-D18}\quad Corrected inference.] Sec.~\ref{sec:distill-setup}; Appendices~\ref{app:metrics} and~\ref{app:results},
Tables~\ref{tab:app-paired} and~\ref{tab:app-mech-seeds}. \emph{Artifacts:} \path|REBUTTAL/W0.1/boot_corpus_f1.json|;
the \path|M6-<arm>/| cells. \emph{Reader:} \path|scripts/boot_corpus_f1.py| (10{,}000 note resamples, seed 0; each run's
pooled \microf{} is first reproduced from its predictions); \path|scripts/m4_paired.py| with \appflag{cluster-notes}
(mirrored in \path|scripts/paper_numbers.py| and checked against the CLI at every run). \emph{Expected:} corpus-level
\mistill{} $-$ bf16 \vanillakd{} \NCorpusQwenMistillVsVanillaBfDelta{} [\NCorpusQwenMistillVsVanillaBfLo,
\NCorpusQwenMistillVsVanillaBfHi]; note-clustered \Ptwoc{} gains \NPoolQwenMistillVsVanillaBfPTwoCClusterDelta{}
(\NPoolQwenMistillVsVanillaBfPTwoCClusterHigher{} of 150, $p$ \NPoolQwenMistillVsVanillaBfPTwoCClusterP) and
\NPoolSmolMistillVsVanillaBfPTwoCClusterDelta. Exact given the stored predictions and cells.

\item[\residtarget{R-D19}\quad Event coverage and the aligned subset.] Sec.~\ref{sec:distill-setup}; Appendix~\ref{app:objective},
Table~\ref{tab:app-coverage}. \emph{Artifacts:} \path|REBUTTAL/W0.4/coverage_{Qwen3-4B,SmolLM3-3B}.json|,
\path|DT-001/targets_mech_dtk/|, the \path|M6-<arm>/| cells' \path|events.jsonl| and \path|faithfulness2.json|.
\emph{Reader:} \path|scripts/event_coverage.py|; \path|scripts/m4_paired.py| with \appflag{aligned-with}. \emph{Expected:}
\smol{} keeps \NCovSmolPOneKeptPct{} / \NCovSmolPTwoKeptPct{} / \NCovSmolPTwoCKeptPct{} of the \Pone{} / \Ptwo{} / \Ptwoc{} events;
aligned-subset gain \NAlignedSmolMistillVsVanillaBfPTwoCClusterDelta. Exact given the stored targets and cells.

\item[\residtarget{R-D20}\quad Faithfulness of the contribution proxy.] Sec.~\ref{sec:distill-objective};
Appendices~\ref{app:instrument} and~\ref{app:objective}. \emph{Artifacts:} \path|HD-001/<note>/attr.npz| and
\path|exact2.npz|, \path|REBUTTAL/W0.7/proxy_faithfulness.json|. \emph{Reader:} \path|scripts/proxy_faithfulness.py|.
\emph{Expected:} Spearman of $c$ with the exact importance \NProxyCSpearmanPooled{} pooled over \NProxyHeads{} heads
(\NProxyCSpearmanEvent{} per event), against \NProxyASpearmanPooled{} for first-order attribution. Exact given the stored
patches.

\item[\residtarget{R-D21}\quad The retriever alone.] Sec.~\ref{sec:distill-main}; Appendix~\ref{app:results},
Table~\ref{tab:app-retriever}. \emph{Artifacts:} \path|REBUTTAL/W0.8/retriever_rows.json|; \path|RT-003/retriever|,
\path|data/mimic4/dtk|. \emph{Reader:} \path|scripts/retriever_rows.py| (the RT-003 retriever re-run on the 150 test notes;
the scorer of \path|scripts/eval_student.py|). \emph{Expected:} top-20 \microf{} \NRetrieverTopTwentyMicroF{} (precision
\NRetrieverTopTwentyMicroP, recall \NRetrieverTopTwentyMicroR); the re-run shortlist equals the stored one on
\NRetrieverShortlistReproduced{} of 150 notes exactly (median Jaccard \NRetrieverShortlistJaccardMedian; CPU against GPU
dense scores reorder the rank-1{,}000 margin).

\item[\residtarget{R-D22}\quad fp32 own heads and the re-selected $\lambda$.] Secs.~\ref{sec:distill-main}
and~\ref{sec:distill-mech}, Tables~\ref{tab:main} and~\ref{tab:mech}; Appendices~\ref{app:objective}
(Table~\ref{tab:app-fpsweep}) and~\ref{app:results}. \emph{Artifacts:} \path|M6-Qwen3-4B__mech_fp32__s{0,1,2}/| and
\path|HD-S-| cells (the transferred $\lambda$); \path|DT-001/{runs,eval}/Qwen3-4B__mech_fp32_l{0.01,0.03,0.3,1}__s0/| (val
cells), \path|DT-001/eval/<arm>_test/|, \path|DT-001/eval/Qwen3-4B__mech_fp32__s0_val/|, their \path|HD-S-| and \path|M6-|
cells, \path|M6probe-Qwen3-4B__mech_fp32{,_l0.01}__s0/| (the \head{35}{26} probe), \path|REBUTTAL/W1.2/|.
\emph{Reader:} \path|scripts/select_lambda.py| with \appflag{fp32}; \path|scripts/rebuttal/w12_paired_test.py|;
\path|scripts/m4_paired.py| (\appflag{cluster-notes}, \appflag{dump} / \appflag{from-dumps} across hosts);
\path|slurm/rebuttal_stage2.sh| (training), \path|scripts/rebuttal/w12_k8_probe.sh|. \emph{Seeds:} 3 at $\lambda=0.1$, 1 per
swept $\lambda$. \emph{Expected:} selected $\lambda$ \NSweepSelected{} (val \microf{} \NSweepLPointZeroOneValMicroF{} against
\NSweepLPointZeroThreeValMicroF); test \microf{} \NSweepLPointZeroOneTestMicroF; own-heads \Ptwoc{} knock-in
\NOwnQwenMistillFpSelSZeroPTwoCKiForty{} at $k=40$ (\NSweepLPointZeroOnePTwoCVsVanillaClusterDelta{} over fp32 \vanillakd) and
\NOwnQwenMistillFpSelSZeroPTwoCKiEight{} at $k=8$ (\NSweepLPointZeroOneProbeKiEight{} with \head{35}{26} in rank 8); the
transferred-$\lambda$ arms' \NPoolQwenMistillFpVsVanillaPTwoCClusterDelta{} against fp32 \vanillakd{} ($p$ \NPoolQwenMistillFpVsVanillaPTwoCClusterP;
\NPoolQwenMistillFpVsVanillaBfPTwoCClusterDelta{} against bf16, $p$ \NPoolQwenMistillFpVsVanillaBfPTwoCClusterP). \emph{Nondeterminism:} G, T, K.

\item[\residtarget{R-D23}\quad Event-weighted \vanillakd.] Sec.~\ref{sec:distill-ablation}; Appendix~\ref{app:ablations},
Table~\ref{tab:app-ablate}. \emph{Artifacts:} \path|DT-001/{runs,eval}/Qwen3-4B__vanilla_evw4_t1__s0/| and its
\path|HD-S-| / \path|M6-| cells. \emph{Reader:} as R-D13 (\path|slurm/rebuttal_stage2.sh| trains; \appflag{event-weight 4}).
\emph{Seeds:} 1. \emph{Expected:} own-heads \Ptwoc{} knock-in \NOwnQwenEventWeightedSZeroPTwoCKiForty{} /
\NOwnQwenEventWeightedSZeroPTwoCKiEight{} ($k=40$ / 8), against \mistill{} seed 0
\NPairQwenEventWeightedSZeroVsMistillSZeroPTwoCDelta; \microf{} \NQwenEventWeightedMicroF. \emph{Nondeterminism:} G, T, K.

\item[\residtarget{R-D24}\quad Uniform attention.] Sec.~\ref{sec:distill-ablation}; Appendix~\ref{app:ablations},
Table~\ref{tab:app-ablate}. \emph{Artifacts:} \path|DT-001/{runs,eval}/Qwen3-4B__mech_uniform_t1__s0/| and its \path|HD-S-| /
\path|M6-| cells. \emph{Reader:} as R-D13 (\appflag{heads uniform} \appflag{terms attn}, $\lambda$ 0.135). \emph{Seeds:} 1.
\emph{Expected:} \NOwnQwenUniformSZeroPTwoCKiForty{} / \NOwnQwenUniformSZeroPTwoCKiEight; against \mistill{} seed 0
\NPairQwenUniformSZeroVsMistillSZeroPTwoCDelta{} (\NPairQwenUniformSZeroVsMistillSZeroPTwoCHigher{} of 150); \microf{}
\NQwenUniformMicroF. \emph{Nondeterminism:} G, T, K.

\item[\residtarget{R-D25}\quad No window.] Secs.~\ref{sec:distill-objective} and~\ref{sec:distill-ablation};
Appendices~\ref{app:objective} and~\ref{app:ablations}, Table~\ref{tab:app-ablate}. \emph{Artifacts:}
\path|DT-001/{runs,eval}/Qwen3-4B__mech_nowindow_t1__s0/| and its \path|HD-S-| / \path|M6-| cells. \emph{Reader:} as R-D13
(\appflag{window 0}). \emph{Seeds:} 1. \emph{Expected:} \NOwnQwenNoWindowSZeroPOneKiForty{} at \Pone{} and
\NOwnQwenNoWindowSZeroPTwoCKiForty{} at \Ptwoc{} against \mistill{} seed 0's \NOwnQwenMistillSZeroPOneKiForty{} and
\NOwnQwenMistillSZeroPTwoCKiForty; \microf{} \NQwenNoWindowMicroF. \emph{Nondeterminism:} G, T, K.

\item[\residtarget{R-D26}\quad Fixed-pool \randheads{} seeds.] Sec.~\ref{sec:distill-ablation}; Appendix~\ref{app:ablations},
Table~\ref{tab:app-ablate}. \emph{Artifacts:} \path|DT-001/{runs,eval}/Qwen3-4B__mech_randheads_pool0_t1__s{1,2}/| and their
\path|HD-S-| / \path|M6-| cells; \path|REBUTTAL/W2.2/w22_read.log|. \emph{Reader:} as R-D13 (\appflag{pool-seed 0});
\path|scripts/m4_paired.py| with \appflag{cluster-notes} over the four \randheads{} runs. \emph{Seeds:} 2 on one frozen
pool. \emph{Expected:} pool fixed, seeds differ by \NPairQwenRandHeadsPoolSTwoVsRandHeadsPoolSOnePTwoCDelta{} at $k=40$;
\mistill{} $-$ \randheads{} over four runs \NPoolQwenMistillVsRandHeadsAllPTwoCClusterDelta{} ($k=40$) and
\NPoolQwenMistillVsRandHeadsAllPTwoCKiEightClusterDelta{} ($k=8$). \emph{Nondeterminism:} G, T, K.

\item[\residtarget{R-D27}\quad A third \smol{} seed.] Secs.~\ref{sec:distill-main} and~\ref{sec:distill-mech},
Tables~\ref{tab:main} and~\ref{tab:mech}; Appendix~\ref{app:results}. \emph{Artifacts:}
\path|DT-001/{runs,eval}/SmolLM3-3B__mech__s2/| and its \path|HD-S-| / \path|M6-| cells; \path|REBUTTAL/W2.4/|.
\emph{Reader:} as R-D5 and R-D6; the three-seed mean and its bootstrap-over-notes interval in
\path|scripts/paper_numbers_rebuttal.py| (\path|scripts/rebuttal/w24_three_seed_mean.py| is the stand-alone form).
\emph{Seeds:} 3. \emph{Expected:} seed 2's own-heads \Ptwoc{} knock-in \NOwnSmolMistillSTwoPTwoCKiForty; three-seed mean
\NOwnSmolMistillThreeSeedPTwoCKiForty{} [\NOwnSmolMistillThreeSeedPTwoCKiFortyLo, \NOwnSmolMistillThreeSeedPTwoCKiFortyHi],
\NOwnSmolMistillThreeSeedRatioBase$\times$ the untrained student's mean \NOwnSmolBasePTwoCKiFortyMean; as means of per-seed
medians (Table~\ref{tab:mech}), \NOwnSmolMistillSeedMeanPTwoCKiForty{} against the untrained median \NOwnSmolBasePTwoCKiForty{}
(\NOwnSmolMistillSeedMeanRatioBase$\times$). \emph{Nondeterminism:} G, T, K.

\item[\residtarget{R-D28}\quad The students under the near-miss edits.] Appendix~\ref{app:readside},
Table~\ref{tab:app-nearmiss-students}. \emph{Artifacts:} \path|REBUTTAL/W3.2/<student>__<arm>__s<seed>/| and
\path|REBUTTAL/W3.2/{base,SmolLM3-3B__base}/| (per-note \path|prefix_edit.jsonl|), the per-arm \path|*_readout.json| and
\path|seeded_transfer_*.json|. \emph{Reader:} \path|scripts/prefix_edit.py| via \path|slurm/rebuttal_stage6.sh|,
\path|scripts/read_prefix_edit.py|, \path|scripts/rebuttal/w32_seeded_transfer.py|. \emph{Seeds:} every seed of every arm.
\emph{Expected:} bf16 \qwenfour{} pair \NNearMissQwenMistillVsVanillaBfPooledDelta{} nats per note
(\NNearMissQwenMistillVsVanillaBfPooledHigher{} of 150); fp32 pair \NNearMissQwenMistillFpVsVanillaPooledDelta{} at
$\lambda$ 0.1 and \NNearMissQwenMistillFpSelVsVanillaPooledDelta{} at 0.01; \smol{} \NNearMissSmolMistillVsVanillaBfPooledDelta.
\emph{Nondeterminism:} K.

\item[\residtarget{R-D29}\quad The mechanism on the students' own traces.] Sec.~\ref{sec:distill-mech};
Appendix~\ref{app:results}, Table~\ref{tab:app-owntrace}. \emph{Artifacts:} \path|REBUTTAL/W3.5/own{,_dt000,_smol}/<arm>/|
(the own-trace runs built from each arm's generations, with coverage), \path|M6own-<arm>/|, \path|REBUTTAL/W3.5/w35_read.json|.
\emph{Reader:} \path|scripts/own_trace_run.py| (CPU), \path|scripts/knockin_bins.py| with \appflag{no-attr} and the arm's
own \path|HD-S-| sets via \path|scripts/rebuttal/lambda/run_w35.sh|, \path|scripts/rebuttal/w35_read.py|. \emph{Seeds:}
every seed of the 16 arms read. \emph{Expected:} \mistill{} seed 0 own-trace \Ptwoc{} knock-in
\NOwnTraceQwenMistillSZeroKiForty{} (teacher-forced \NOwnTraceQwenMistillSZeroKiFortyTF); \mistill{} $-$ bf16 \vanillakd{}
\NOwnTraceQwenMistillVsVanillaBfPooledDelta{} ($k=40$) and \NOwnTraceQwenMistillVsVanillaBfKiEightPooledDelta{} ($k=8$),
note-clustered. \emph{Nondeterminism:} K (the students' own traces are the stored generations).
\item[\residtarget{R-D30}\quad The fp32 operating point over three seeds.] Secs.~\ref{sec:distill-main}
and~\ref{sec:distill-mech}, Tables~\ref{tab:main} and~\ref{tab:mech}; Appendices~\ref{app:objective} (\ref{app:obj-fp32})
and~\ref{app:results} (Tables~\ref{tab:app-mech-seeds} and~\ref{tab:app-owntrace}). \emph{Artifacts:}
\path|DT-001/{runs,eval}/Qwen3-4B__mech_fp32_l0.01__s{1,2}/| (seed 0: R-D22), their \path|HD-S-| and \path|M6-| cells,
\path|M6own-Qwen3-4B__mech_fp32_l0.01__s{0,1,2}/|, \path|REBUTTAL/A3/| (\path|m4_paired_P2C_{k40,k8,m2}.md|,
\path|w35_read_l0.01.json|, \path|w32_l0.01.json|), \path|REBUTTAL/W0.1/boot_corpus_f1.json| (the three-seed pairs).
\emph{Reader:} \path|scripts/m4_paired.py| (\appflag{cluster-notes}), \path|scripts/boot_corpus_f1.py|,
\path|scripts/rebuttal/w35_read.py|, \path|scripts/rebuttal/w32_seeded_transfer.py|; the pipeline's note-clustered mirror is
checked against the reader's tables at every run. \emph{Seeds:} 3 and 3. \emph{Expected:} own-heads \Ptwoc{} knock-in at
$k=40$, \mistill{} $-$ fp32 \vanillakd{} \NRepLPointZeroOnePTwoCClusterMean{} [\NRepLPointZeroOnePTwoCClusterLo,
\NRepLPointZeroOnePTwoCClusterHi] note-clustered (pairs \NRepLPointZeroOnePTwoCSZeroMean, \NRepLPointZeroOnePTwoCSOneMean,
\NRepLPointZeroOnePTwoCSTwoMean); medians \NOwnQwenMistillFpSelSeedMeanPTwoCKiForty{} against
\NOwnQwenVanillaSeedMeanPTwoCKiForty; corpus \microf{} \NCorpusQwenMistillFpSelVsVanillaDelta{}
[\NCorpusQwenMistillFpSelVsVanillaLo, \NCorpusQwenMistillFpSelVsVanillaHi]; $k=8$ \NRepLPointZeroOnePTwoCKiEightClusterMean;
M2 \NRepLPointZeroOneMTwoPTwoCClusterMean; own traces \NOwnTraceQwenMistillFpSelVsVanillaPooledDelta{} over
\NOwnTraceQwenMistillFpSelVsVanillaPooledNotes{} notes; prefix edits \NRepLPointZeroOneEditPooledMean. \emph{Nondeterminism:} G, T, K.
\item[\residtarget{R-D31}\quad Matched selection: recovery.] Sec.~\ref{sec:distill-ablation}; Appendix~\ref{app:abl-matched},
Figure~\ref{fig:a2-frontier}, Table~\ref{tab:app-matched}. \emph{Artifacts:}
\path|DT-001/{runs,eval}/Qwen3-4B__mech_l{0.01,0.03,0.3}__s0/|, \path|…mech_l0.03__s{1,2}/|,
\path|…mech_randheads_pool0_l{0.03,0.1,0.3,1}__s0/|, \path|…pool0_l0.3__s{1,2}/| (validation cells unsuffixed, the seed-0
test cells under \path|_test|); \path|HD-S-| and \path|M6-| cells of the six pick arms; \path|REBUTTAL/A2/select_*.json|
(\path|.tie005.json| = the 0.005-band sensitivity), \path|m4_paired_M6_{P2C_k40,P2C_k8,P2C_m2,P1_k40}.md|,
\path|frontier_val.{png,pdf}|. \emph{Reader:} \path|scripts/select_lambda.py| (\appflag{family}, \appflag{tie}),
\path|scripts/m4_paired.py| (\appflag{cluster-notes}), \path|scripts/rebuttal/plan3/a2_frontier_plot.py|. \emph{Seeds:} 3 and
3 at the picks; the selection on seed 0. \emph{Expected:} picks \NMatchedSelectMistillPick{} / \NMatchedSelectRandHeadsPick{}
(0.005 band: \NMatchedSelectMistillPickFive{} / \NMatchedSelectRandHeadsPickFive); own-heads \Ptwoc{} knock-in at $k=40$
\NMatchedPTwoCClusterMean{} [\NMatchedPTwoCClusterLo, \NMatchedPTwoCClusterHi] note-clustered; $k=8$
\NMatchedPTwoCKiEightClusterMean; M2 \NMatchedMTwoPTwoCClusterMean; medians \NMatchedMistillSeedMeanPTwoCKiForty{} against
\NMatchedRandHeadsSeedMeanPTwoCKiForty. \emph{Nondeterminism:} G, T, K.
\item[\residtarget{R-D32}\quad Matched selection: task.] Sec.~\ref{sec:distill-ablation}; Appendix~\ref{app:abl-matched},
Table~\ref{tab:app-matched}. \emph{Artifacts:} the six test evaluations of R-D31; \path|REBUTTAL/A2/boot_corpus_f1_a2.json|.
\emph{Reader:} \path|scripts/boot_corpus_f1.py| (\appflag{pairs}). \emph{Seeds:} 3 and 3. \emph{Expected:} corpus \microf{}
\NMatchedTaskDelta{} [\NMatchedTaskLo, \NMatchedTaskHi] (pairs \NMatchedTaskSZeroDelta, \NMatchedTaskSOneDelta,
\NMatchedTaskSTwoDelta); test \microf{} \NMatchedMistillSeedMeanMicroF{} against \NMatchedRandHeadsSeedMeanMicroF.
\emph{Nondeterminism:} G, T.
\item[\residtarget{R-D33}\quad Evidence-flip test: behavioural half.] Sec.~\ref{sec:limitations};
Appendix~\ref{app:evflip}, Table~\ref{tab:app-ef-behav}. \emph{Artifacts:} \path|data/mimic4/ef_conf_{original,flipped,irrelevant,order}/|
(the enclave; the cases, inventory and manifest), \path|REBUTTAL/A1/efconf/gen/<arm>/<condition>/generations/| (14
checkpoints $\times$ 4 conditions $\times$ 132), \path|score_behav.json|, \path|read_ef.{md,json}|, \path|pooled_pairs.{md,json}|,
\path|read_ef_extra.{md,json}|. \emph{Reader:} \path|scripts/rebuttal/plan3/ef_read.sh| (\path|ef_score.py| + \path|read_ef.py|),
\path|scripts/rebuttal/ef_pooled_pairs.py|, \path|scripts/rebuttal/read_ef_extra.py|. \emph{Seeds:} 3 and 3 (bf16 and fp32
pairs); the teacher and the \randheads{} pick descriptive. \emph{Expected:} teacher $D$ \NEFTeacherD{} [\NEFTeacherDLo,
\NEFTeacherDHi]; primary $\Delta D$ \NEFPrimaryMean{} [\NEFPrimaryLo, \NEFPrimaryHi] (\NEFPrimaryHigher{} / \NEFPrimaryLower{}
/ \NEFPrimaryTies, $p$ \NEFPrimaryP; detectable $\delta$ \NEFPrimaryDetectable); secondary \NEFSecondaryMean{}
[\NEFSecondaryLo, \NEFSecondaryHi]. \emph{Nondeterminism:} G (sampled generations; the paired seed key is shared across
conditions).
\item[\residtarget{R-D34}\quad Evidence-flip test: causal half.] Sec.~\ref{sec:limitations}; Appendix~\ref{app:evflip},
Table~\ref{tab:app-ef-causal}. \emph{Artifacts:} \path|REBUTTAL/A1/efconf/hf/<arm>/{clean,named,top40}/{sham,target}/<condition>/generations/|
(merged from \path|hf_shards/| with coverage against the frozen ledger; per-shard host provenance in \path|config.json|),
\path|read_ef.{md,json}|, \path|read_ef_extra_hf.{md,json}|, \path|read_ef.sens29.{md,json}|,
\path|approval_ledger.sensitivity29.csv|. \emph{Reader:} \path|scripts/rebuttal/ef_intervene_generate.py|
(\path|MISTILL_KEYMASK_FAST=1|; \path|tests/test_ef_intervene.py|), \path|scripts/rebuttal/plan3/ef_merge_shards.py|,
\path|plan3/ef_read.sh|, \path|read_ef_extra.py| (\appflag{hf}). \emph{Seeds:} the teacher and the bf16 seed-0 students.
\emph{Expected:} target $-$ sham \NEFCausalTeacherNamedSelMean{} (teacher, named), \NEFCausalTeacherTopFortySelMean{}
(teacher, top-40), \NEFCausalMistillSZeroTopFortySelMean{} (\mistill), \NEFCausalVanillaBfSZeroTopFortySelMean{} (\vanillakd);
$L$ \NEFLMean{} [\NEFLLo, \NEFLHi] (29 cases \NEFSensLMean); teacher offset \NEFBackendTeacherMean{} [\NEFBackendTeacherLo,
\NEFBackendTeacherHi]. \emph{Nondeterminism:} G (HF sampling with the row-seeded sampler; the fast and general key-mask
paths are byte-identical on the tests).
\item[\residtarget{R-D35}\quad Evidence-flip cohort.] Appendix~\ref{app:ef-population}. \emph{Artifacts:}
\path|REBUTTAL/A1/LOCK.json| (the lock, pinned to the memo's locked rules), \path|efconf_prop/cohort_row.json| (the draw and
the proposals), \path|efconf_prop/selection_first_80.json| (the reviewed set, verified against the frozen proposals),
\path|efconf_prop/authored/FROZEN.txt| (the flips' hashes), \path|efconf/approval_ledger.frozen-20261018T2224.csv| (the
author's decisions and reason codes), \path|efconf/cohort_row.json| (the reconciliation). \emph{Reader:}
\path|scripts/rebuttal/ef_build_cohort.py| (\texttt{lock} $\to$ \texttt{draw} $\to$ \texttt{retriever} $\to$ \texttt{shortlist}
$\to$ \texttt{screen} $\to$ \texttt{cases} $\to$ \texttt{authored} $\to$ \texttt{freeze}; \path|tests/test_ef_freeze.py|).
\emph{Expected:} \NEFCohortDrawn{} drawn, \NEFCohortProposals{} proposals, \NEFCohortReviewed{} reviewed =
\NEFCohortAccepted{} accepted + \NEFCohortExcluded{} excluded, \NEFCohortNotReached{} not reached.
\emph{Nondeterminism:} none (the draw is seeded; every stage is hash-verified).
\item[\residtarget{R-D36}\quad \smol{} \mistill{} seed 2 on its own traces.] Appendix~\ref{app:results},
Table~\ref{tab:app-owntrace}. \emph{Artifacts:} \path|REBUTTAL/W3.5/own/SmolLM3-3B__mech__s2/| and
\path|summary_SmolLM3-3B__mech__s2.json| (completed traces), \path|M6own-SmolLM3-3B__mech__s2/|,
\path|REBUTTAL/A4/w35_read_smol_s2.json|. \emph{Reader:} as R-D29 (\path|scripts/rebuttal/plan3/a3_owntrace.sh|,
\path|w35_read.py|). \emph{Seeds:} the third \mistill{} seed; no \vanillakd{} seed-2 counterpart. \emph{Expected:}
\NOwnTraceSmolMistillSTwoComplete{} completed traces, \NOwnTraceSmolMistillSTwoNotes{} measurable; own-trace \Ptwoc{}
knock-in \NOwnTraceSmolMistillSTwoKiForty{} against \NOwnTraceSmolMistillSTwoKiFortyTF{} teacher-forced
(\NOwnTraceSmolMistillSTwoKiFortyDelta). \emph{Nondeterminism:} K.
\item[\residtarget{R-D37}\quad Public-task extension: recovery.] Sec.~\ref{sec:limitations}; Appendix~\ref{app:musique},
Tables~\ref{tab:app-mq-reads}--\ref{tab:app-mq-arms}. \emph{Artifacts:} \path|DT-004/LOCK.json|, \path|DT-004/ARTIFACTS.lock.json| (+
amendments), \path|DT-004/runs/<cell>/|, \path|HD-S-<cell>/| (own rankings and knock-in on the \NMQRankQ{} ranking questions),
\path|DT-004/mq_rank_groups.json|, \path|REBUTTAL/B2/reads/recovery_mistill_vs_cotkd.md| and \path|recovery_mistill_vs_randheads.md|.
\emph{Reader:} \path|scripts/m4_paired.py| (\texttt{--kind P2 --k 40 --cfg global --min-range 1.0 --cluster-notes --group-map}). \emph{Seeds:} 3 and 3;
one frozen random pool. \emph{Expected:} \mistill{} $-$ \vanillakd{} \pci{\NMQRecMean}{\NMQRecLo}{\NMQRecHi} on \NMQRecN{} questions /
\NMQRecGroups{} groups; \mistill{} $-$ \randheads{} \pci{\NMQRecRandMean}{\NMQRecRandLo}{\NMQRecRandHi} (\NMQRecRandN). \emph{Nondeterminism:} G, T, K.
\item[\residtarget{R-D38}\quad Public-task extension: task.] Appendix~\ref{app:musique}, Tables~\ref{tab:app-mq-reads}--\ref{tab:app-mq-arms}.
\emph{Artifacts:} \path|data/musique/mq_eval/| (public; rebuilt by \path|scripts/dt004/build_musique_sets.py|), \path|DT-004/eval_a100/<cell>_test/|
(the one-card set; \path|DT-004/eval/<cell>_test/| the second card), \path|DT-004/eval/_verify/|, \path|REBUTTAL/B2/reads/task_gate_*.json|.
\emph{Reader:} \path|scripts/dt004/task_gate_read.py| (official MuSiQue answer-F1 semantics, \path|mistill.metrics.qa_em_f1|); acceptance by
\path|scripts/dt004/verify_eval.py|. \emph{Seeds:} 3 and 3. \emph{Expected:} \pci{\NMQTaskMean}{\NMQTaskLo}{\NMQTaskHi} on \NMQEval{} /
\NMQEvalGroups{} (per-arm F1 in Table~\ref{tab:app-mq-arms}); second card \pci{\NMQTaskRtxMean}{\NMQTaskRtxLo}{\NMQTaskRtxHi}. \emph{Nondeterminism:} G.
\item[\residtarget{R-D39}\quad Public-task extension: behaviour.] Appendix~\ref{app:musique}, Table~\ref{tab:app-mq-reads}. \emph{Artifacts:}
\path|data/musique/ef_mq_{original,flipped,irrelevant,control}/| (public), \path|REBUTTAL/B2/ef_mq/| (cases, the author's ledger, \path|FROZEN.txt|),
\path|REBUTTAL/B2/ef_mq_behav/<cell>/|, \path|REBUTTAL/B2/reads/behav.json| and \path|gp_*.json|. \emph{Reader:} \path|scripts/rebuttal/read_ef.py| (\texttt{--task musique --behav}),
\path|scripts/dt004/gp_read.py|. \emph{Seeds:} 3 and 3, $N=4$ per (pair, condition), sample-index-paired. \emph{Expected:}
$G_{P,\mathrm{clean}}$ \pci{\NMQBehMean}{\NMQBehLo}{\NMQBehHi} on \NMQPairs{} pairs / \NMQPairGroups{} groups; vs \randheads{}
\pci{\NMQBehRandMean}{\NMQBehRandLo}{\NMQBehRandHi}. \emph{Nondeterminism:} G.
\item[\residtarget{R-D40}\quad Public-task extension: the teacher evidence-mask gate.] Appendix~\ref{app:musique}, Table~\ref{tab:app-mq-gate}.
\emph{Artifacts:} \path|REBUTTAL/B2/gate_teacher_prefill/| (\NMQGateSlots{} HF generations), \path|REBUTTAL/B2/gate_read/prefill.{json,md}|,
\path|prefill_weighting.{json,md}|, \path|HD-MQ/head_sets_contrast_final.json| (the frozen teacher set). \emph{Reader:}
\path|scripts/rebuttal/read_ef.py| (\texttt{--task musique --hf prefill=...}; the recorded equal-group analysis) and \path|scripts/dt004/gate_weighting_sensitivity.py|
(the equal-pair sensitivity). \emph{Seeds:} the teacher only; three random draws (draw protocol 2). \emph{Expected:} clean $P$ \NMQGateCleanPMean{} (equal-group, recorded; equal-pair \NMQGateCleanPPair);
$E_P$ \pci{\NMQGateEPMean}{\NMQGateEPLo}{\NMQGateEPHi}, $E_P^{\mathrm{rand}}$ \pci{\NMQGateEPRandMean}{\NMQGateEPRandLo}{\NMQGateEPRandHi}
(equal-pair: \pci{\NMQGatePairEPMean}{\NMQGatePairEPLo}{\NMQGatePairEPHi}, \pci{\NMQGatePairEPRandMean}{\NMQGatePairEPRandLo}{\NMQGatePairEPRandHi});
inconclusive, no student run. \emph{Nondeterminism:} K (the HF path is deterministic given the seed).
\item[\residtarget{R-D41}\quad Cross-task screen: EURLEX57K structural characterization.] Appendix~\ref{app:tasks2}, Table~\ref{tab:app-xt-char}.
\emph{Artifacts:} \path|DT-005/eurlex_to_r1/| (\path|HD-rank-b0/|, \path|HD-assess-{b1,conf}/|, \path|CAUSAL_{b1,conf}_P2C.json|, \path|CAUSAL_conf_P2C_allk.json|,
\path|CONF_VERDICT.json|, \path|READBACK_{b1,conf}_PAIRED.json|, \path|NATIVE_conf.json|, \path|audit_a1r_trace/| with the blank sheet and the completed
annotations \path|audit_sheet.codex_review.csv|, SHA-256 \texttt{2b952bfd\ldots{}f9f4b74c}), \path|DT-005/CHAR_LOCK.json|.
\emph{Reader:} \path|scripts/dt005/screen_causal_read.py| (recovery, necessity; bootstrap over inputs), \path|scripts/dt005/screen_readback.py|, the native reader;
\path|scripts/paper_numbers_dt005.py| registers the macros and reads the audit totals from the completed sheet, checked row by row against the blank one. \emph{Seeds:} the teacher only; one random draw per control panel (seed 0). \emph{Expected:}
ranked $-$ random at $k=40$ \pci{\NXTEurConfDeltaMean}{\NXTEurConfDeltaLo}{\NXTEurConfDeltaHi} on \NXTEurConfEvaluable{} fresh acts; verdict \NXTEurConfVerdict.
\emph{Nondeterminism:} K.
\item[\residtarget{R-D42}\quad Cross-task screen: BigCodeBench structural characterization.] Appendix~\ref{app:tasks2}, Table~\ref{tab:app-xt-char}.
\emph{Artifacts:} \path|DT-005/bigcodebench_tci_r1/| (the same layout; \path|HD-rank-b02/|; the completed annotations \path|audit_sheet.codex_review.csv|, SHA-256
\texttt{975df4c5\ldots{}2ecd50447}), \path|DT-005/bigcodebench/TR-conf/EXEC.json| (the teacher's sandboxed pass@1).
\emph{Reader:} as R-D41. \emph{Expected:} ranked $-$ random at $k=40$ \pci{\NXTBcbConfDeltaMean}{\NXTBcbConfDeltaLo}{\NXTBcbConfDeltaHi} on \NXTBcbConfEvaluable{} fresh tasks;
verdict \NXTBcbConfVerdict. \emph{Nondeterminism:} K.
\item[\residtarget{R-D43}\quad Programming pilot: training.] Appendix~\ref{app:tasks2}. \emph{Artifacts:} \path|DT-005/pilot/| (\path|PILOT_LOCK.md|, \path|ARTIFACTS.lock.json| with its amendment,
\path|THRESHOLDS.json|, \path|pools/|, \path|targets_train2k/|, \path|targets_mech_train2k/|, \path|runs/<run>/final/| for the six runs, \path|reliability/NVIDIA_H200/|).
\emph{Code:} \path|scripts/train_student.py| through \path|scripts/dt005/pilot_chain.sbatch| (exact-resume segments). \emph{Seeds:} 0 and 1 per arm. \emph{Compute:}
about 3 GPU-hours per attention arm and 1 per \vanillakd{} run on one H200. \emph{Nondeterminism:} the device's run-to-run floor (\path|RESUME_CHECK_*.md|).
\item[\residtarget{R-D44}\quad Programming pilot: mechanism evaluation.] Appendix~\ref{app:tasks2}, Table~\ref{tab:app-xt-pilot}. \emph{Artifacts:} \path|DT-005/pilot/eval/<run>/|
(\path|HD-rank-b02/|, \path|HD-assess-b1/|, \path|CAUSAL-b1_P2{,C}.json|, \path|M1-b1/|). \emph{Reader:} \path|scripts/dt005/pilot_eval.sbatch| (the locked stages), \path|scripts/dt005/screen_causal_read.py|.
\emph{Expected:} per-run own-heads \Ptwoc{} recovery on the \NXTPilotPtwocKfortyMcN{} common-support inputs as read by R-D46. \emph{Nondeterminism:} K.
\item[\residtarget{R-D45}\quad Programming pilot: native pass@1.] Appendix~\ref{app:tasks2}, Table~\ref{tab:app-xt-pilot}. \emph{Artifacts:} \path|DT-005/pilot/eval/<run>/GEN-b{0,1,2}/|
(generations with \path|record.json| per task, \path|gold.status| / \path|gold.passed| written by the sandbox). \emph{Reader:} \path|scripts/dt005/bcb/exec_runs.sh| (containment-probed sandbox, CPU),
\path|scripts/dt005/terminal_counts.py| (completeness). \emph{Expected:} per-arm pass@1 \NXTPilotPassOneCotKdSZero{} / \NXTPilotPassOneCotKdSOne{} (\vanillakd), \NXTPilotPassOneMistillSZero{} /
\NXTPilotPassOneMistillSOne{} (\mistill), \NXTPilotPassOneRandHeadsSZero{} / \NXTPilotPassOneRandHeadsSOne{} (\randheads). \emph{Nondeterminism:} none given the generations.
\item[\residtarget{R-D46}\quad Programming pilot: the frozen comparison.] Appendix~\ref{app:tasks2}, Table~\ref{tab:app-xt-pilot}. \emph{Artifacts:} \path|DT-005/pilot/PILOT_RESULT.{json,md}|.
\emph{Reader:} \path|scripts/dt005/pilot_compare.py| (locked; refuses without \path|THRESHOLDS.json| and the lock; the group bootstrap). \emph{Expected:} \mistill{} $-$ \vanillakd{} at $k=40$ median
\NXTPilotPtwocKfortyMcMed{}, mean \pci{\NXTPilotPtwocKfortyMcMean}{\NXTPilotPtwocKfortyMcLo}{\NXTPilotPtwocKfortyMcHi}; meaningful \NXTPilotMeaningful{}, acceptable \NXTPilotAcceptable{}.
\emph{Nondeterminism:} the bootstrap seed is fixed.
\item[\residtarget{R-D47}\quad Zero-shot transfer: CodiEsp-D b1.] Appendix~\ref{app:zeroshot}, Tables~\ref{tab:zs} and~\ref{tab:zs-levels}. \emph{Artifacts:}
\path|DT-006/codiesp/transfer/<arm>/| (\path|HD-rank-b0/|, \path|HD-assess-b1/|, \path|CAUSAL-b1_P2C.json|),
\path|DT-006/codiesp/TRANSFER_RESULT.json|; the teacher's traces \path|DT-005/codiesp/TR-{b0,b1}/|. \emph{Run:}
\path|scripts/dt006/codiesp_transfer.sbatch| (step \texttt{mech}). \emph{Reader:} \path|scripts/dt006/codiesp_transfer_read.py| (\texttt{-{}-mech};
seed-paired; the 10,000-draw bootstrap over documents). \emph{Expected:} \mistill{} $-$ \vanillakd{}
\pci{\NZsBoneMcMean}{\NZsBoneMcLo}{\NZsBoneMcHi} (\NZsBoneMcAbove{} / \NZsBoneMcBelow{} documents), $-$ \randheads{}
\pci{\NZsBoneMrMean}{\NZsBoneMrLo}{\NZsBoneMrHi}. \emph{Nondeterminism:} attention kernels; the bootstrap seed is fixed.
\item[\residtarget{R-D48}\quad Zero-shot transfer: the fresh documents and $k=8$.] Appendix~\ref{app:zeroshot}, Tables~\ref{tab:zs} and~\ref{tab:zs-levels}.
\emph{Artifacts:} \path|DT-006/codiesp/transfer/<arm>/| (\path|HD-assess-conf/|, \path|CAUSAL-conf_P2C.json|),
\path|DT-006/codiesp/TRANSFER_RESULT_{conf_k40,conf_k8,b1_k8}.json|; the teacher's traces \path|DT-005/codiesp/TR-conf/|. \emph{Run:}
\path|scripts/dt006/codiesp_transfer.sbatch| (step \texttt{mechconf}). \emph{Reader:} \path|scripts/dt006/codiesp_transfer_read.py| (\texttt{-{}-mech})
with \texttt{-{}-causal-tag} and \texttt{-{}-k} (guarded by \path|scripts/dt006/reads_6b_conf_fp32.sh|). \emph{Expected:} conf, $k=40$:
\pci{\NZsConfMcMean}{\NZsConfMcLo}{\NZsConfMcHi}, $-$ \randheads{} \pci{\NZsConfMrMean}{\NZsConfMrLo}{\NZsConfMrHi}; $k=8$:
\pci{\NZsBoneKeightMcMean}{\NZsBoneKeightMcLo}{\NZsBoneKeightMcHi} (b1), \pci{\NZsConfKeightMcMean}{\NZsConfKeightMcLo}{\NZsConfKeightMcHi}
(conf). \emph{Nondeterminism:} attention kernels; the bootstrap seed is fixed.
\item[\residtarget{R-D49}\quad Zero-shot transfer: the English machine translation.] Appendix~\ref{app:zeroshot}, Tables~\ref{tab:zs} and~\ref{tab:zs-levels}.
\emph{Artifacts:} \path|data/codiesp/screen/conf_en/| (rendered by \path|scripts/dt005/render_screen.py| with \texttt{-{}-lang en} and the conf
shortlists), the teacher's traces \path|DT-005/codiesp/TR-conf_en/|, \path|DT-006/codiesp/transfer/<arm>/CAUSAL-conf_en_P2C.json|,
\path|DT-006/codiesp/XTASK_conf_en{,_k8}.json|. \emph{Run:} \path|scripts/dt006/codiesp_transfer.sbatch| (step \texttt{mechconf} on
\texttt{conf\_en}). \emph{Reader:} \path|scripts/dt006/xtask_read.py| (\path|groups/qwen_bf16.json|; guarded by
\path|scripts/dt006/reads_6c_xtask.sh|). \emph{Expected:} \pci{\NZsEnMcMean}{\NZsEnMcLo}{\NZsEnMcHi} at $k=40$,
\pci{\NZsEnKeightMcMean}{\NZsEnKeightMcLo}{\NZsEnKeightMcHi} at $k=8$. \emph{Nondeterminism:} decoding of the teacher's traces is
stochastic (the frozen decoding, per-example seeds); attention kernels.
\item[\residtarget{R-D50}\quad Zero-shot transfer: the \smol{} students.] Appendix~\ref{app:zeroshot}, Tables~\ref{tab:zs} and~\ref{tab:zs-levels}.
\emph{Artifacts:} \path|DT-006/codiesp/transfer/SmolLM3-3B__*/|, \path|DT-006/codiesp/XTASK_smol_{b1,conf}.json|. \emph{Run:}
\path|scripts/dt006/codiesp_transfer.sbatch| (the \smol{} convention: YaRN 1.0, token-mode events). \emph{Reader:}
\path|scripts/dt006/xtask_read.py| (\path|groups/smol.json|, unpaired). \emph{Expected:} \mistill{} $-$ bf16 \vanillakd{}
\pci{\NZsSmolBoneMcMean}{\NZsSmolBoneMcLo}{\NZsSmolBoneMcHi} (b1), \pci{\NZsSmolConfMcMean}{\NZsSmolConfMcLo}{\NZsSmolConfMcHi} (conf).
\emph{Nondeterminism:} attention kernels; the bootstrap seed is fixed.
\item[\residtarget{R-D51}\quad Zero-shot transfer: MuSiQue-Ans answer events.] Appendix~\ref{app:zeroshot}, Tables~\ref{tab:zs} and~\ref{tab:zs-levels}.
\emph{Artifacts:} \path|DT-006/xtask/musique/<arm>/| (\path|HD-rank/|, \path|HD-assess/|, \path|CAUSAL-assess_P2.json|),
\path|DT-006/codiesp/XTASK_musique.json|; the teacher's traces of Appendix~\ref{app:musique}. \emph{Run:}
\path|scripts/dt006/xtask_mech.sbatch| (\texttt{TASK=musique}). \emph{Reader:} \path|scripts/dt006/xtask_read.py|
(\path|groups/qwen_bf16.json|). \emph{Expected:} \pci{\NZsMqMcMean}{\NZsMqMcLo}{\NZsMqMcHi}, $-$ \randheads{}
\pci{\NZsMqMrMean}{\NZsMqMrLo}{\NZsMqMrHi}. \emph{Nondeterminism:} attention kernels; the bootstrap seed is fixed.
\item[\residtarget{R-D52}\quad Evidence attention on own heads: CodiEsp-X.] Appendix~\ref{app:evidence}, Tables~\ref{tab:app-evattn}
and~\ref{tab:app-evattn-levels}. \emph{Artifacts:} \path|DT-007/B2/<arm>/test250/|, \path|DT-007/B2_RESULT_test250.json| (under
\path|DT-007/B2/LOCK.json|); the own-head rankings \path|DT-006/codiesp/transfer/<arm>/HD-rank-b0/| (the no-window student's in
\path|DT-007/B2/own_rank/|); the teacher's traces \path|DT-006/codiesp/transfer/teacher/GEN-test250/|. \emph{Run:}
\path|scripts/dt007/evidence_heads_b2.sbatch| (the no-window ranking: \path|scripts/dt007/rank_own.sbatch|). \emph{Reader:}
\path|scripts/dt007/evidence_heads_b2_read.py| (guarded by \path|scripts/dt007/read_b2.sbatch|). \emph{Expected:} \mistill{} $-$
\vanillakd{} lift \pci{\NEaCxLiftMcMean}{\NEaCxLiftMcLo}{\NEaCxLiftMcHi} (\NEaCxLiftMcAbove{} / \NEaCxLiftMcBelow{} documents); the
teacher's validity \NEaCxTeach. \emph{Nondeterminism:} attention kernels; the bootstrap seed is fixed.
\item[\residtarget{R-D53}\quad Evidence attention: the development batches and the per-head read.] Appendix~\ref{app:evidence},
Table~\ref{tab:app-evattn}. \emph{Artifacts:} \path|DT-007/B/<arm>/{conf,b1}/|, \path|DT-007/B_RESULT_{conf,b1}.json| (under
\path|DT-007/B/LOCK.json|), \path|DT-007/B_POSTHOC_{conf,b1}.json|; \path|DT-007/B2ph/<arm>/test250/|,
\path|DT-007/B2_PERHEAD_test250.json|; the teacher's traces \path|DT-005/codiesp/TR-{conf,b1}/|. \emph{Run:}
\path|scripts/dt007/evidence_heads.sbatch|; \path|scripts/dt007/evidence_heads_b2.sbatch| with \texttt{PERHEAD=1}. \emph{Reader:}
\path|scripts/dt007/evidence_heads_read.py| (guarded by \path|scripts/dt007/read_b.sbatch|), \path|scripts/dt007/evidence_heads_posthoc.py|
(registered after the read; descriptive), \path|scripts/dt007/perhead_read.py|. \emph{Expected:} the supervised pool
\pci{\NEbConfSupMcMean}{\NEbConfSupMcLo}{\NEbConfSupMcHi} and the own heads \pci{\NEbConfLiftMcMean}{\NEbConfLiftMcLo}{\NEbConfLiftMcHi}
(conf); \NEaPhUp{} of \NEaPhRead{} heads more specific. \emph{Nondeterminism:} attention kernels; the bootstrap seed is fixed.
\item[\residtarget{R-D54}\quad Evidence attention on own heads: MDACE.] Appendix~\ref{app:evidence}, Tables~\ref{tab:app-evattn}
and~\ref{tab:app-evattn-levels}. \emph{Data:} the MDACE annotations (\texttt{solventum-oss/MDACE}, MIT) and the \texttt{MIMIC-III}
notes (credentialed access), extracted by \path|scripts/dt007/mdace_extract.py| and built by \path|scripts/dt007/mdace_build.py|; the
shortlist by \path|scripts/dt007/mdace_prep.sbatch|; the teacher's traces by \path|scripts/dt007/mdace_gen.sbatch| into
\path|DT-007/E/teacher/GEN-mdace/|. \emph{Artifacts:} \path|DT-007/E/B2/<arm>/mdace/|, \path|DT-007/E/E_RESULT_mdace.json| (block
\texttt{b2}; under \path|DT-007/E/LOCK.json|); the own-head rankings \path|HD-S-<arm>/|. \emph{Run:} \path|scripts/dt007/mdace_e.sbatch|.
\emph{Reader:} \path|scripts/dt007/mdace_read.py| (the frozen B2 and deletion readers' rules, equality-tested against them in
\path|tests/test_dt007_e.py|; guarded by \path|scripts/dt007/read_e.sbatch|). \emph{Expected:}
\pci{\NEaMdLiftMcMean}{\NEaMdLiftMcLo}{\NEaMdLiftMcHi} (\NEaMdLiftMcAbove{} / \NEaMdLiftMcBelow{} documents); the teacher's validity
\NEaMdTeach. \emph{Nondeterminism:} decoding of the teacher's traces is stochastic (per-example seeds); attention kernels.
\item[\residtarget{R-D55}\quad The deletion test: CodiEsp-X.] Appendix~\ref{app:evidence}, Table~\ref{tab:app-evdel}. \emph{Artifacts:}
\path|DT-007/D1/<arm>/test250/|, \path|DT-007/D1_RESULT_test250.json| (under \path|DT-007/D1/LOCK.json|, with the author's
\path|DT-007/D1/THRESHOLDS.json|). \emph{Run:} \path|scripts/dt007/evidence_delete.sbatch|. \emph{Reader:}
\path|scripts/dt007/evidence_delete_read.py| (guarded by \path|scripts/dt007/read_d1.sbatch|). \emph{Expected:}
\pci{\NEdCxDeltaMcMean}{\NEdCxDeltaMcLo}{\NEdCxDeltaMcHi}, detected and below the bar; the teacher's validity \NEdCxTeach.
\emph{Nondeterminism:} attention kernels; the control runs and the bootstrap use fixed seeds.
\item[\residtarget{R-D56}\quad The deletion test: MDACE.] Appendix~\ref{app:evidence}, Table~\ref{tab:app-evdel}. \emph{Artifacts:}
\path|DT-007/E/D1/<arm>/mdace/|, \path|DT-007/E/E_RESULT_mdace.json| (block \texttt{d1}; under \path|DT-007/E/LOCK.json|, with the
same thresholds file). \emph{Run:} \path|scripts/dt007/mdace_e.sbatch|. \emph{Reader:} \path|scripts/dt007/mdace_read.py|.
\emph{Expected:} \pci{\NEdMdDeltaMcMean}{\NEdMdDeltaMcLo}{\NEdMdDeltaMcHi}, not detected; the teacher's validity \NEdMdTeach.
\emph{Nondeterminism:} as R-D54.
\item[\residtarget{R-D57}\quad Spontaneous evidence in the students' generations.] Appendix~\ref{app:evidence}. \emph{Artifacts:} the
students' own generations \path|DT-006/codiesp/transfer/<arm>/GEN-{conf,test250}/|; \path|DT-007/A/<arm>/|,
\path|DT-007/A_RESULT_{conf,test250}.json|. \emph{Run:} \path|scripts/dt007/a_run.sbatch|. \emph{Reader:} \path|scripts/dt007/a_read.py|
(its self-check against the frozen evidence-follow reader passed on every arm). \emph{Expected:}
\pci{\NEsConfQuoteMcMean}{\NEsConfQuoteMcLo}{\NEsConfQuoteMcHi} (conf) and \pci{\NEsTestQuoteMcMean}{\NEsTestQuoteMcLo}{\NEsTestQuoteMcHi}
(test), not detected. \emph{Nondeterminism:} none (the generations are fixed); the bootstrap seed is fixed.
\end{description}

\section{Extended related work}
\label{app:related}
\supports{Sec.~\ref{sec:related}}
\paragraph{Chain-of-thought distillation.}
Training small students on a teacher's rationales transfers multi-step reasoning
\citep{hsieh2023distilling, ho2023large, magister2023teaching, shridhar_distilling_2023}; later work
makes rationales and answers consistent \citep{wang_scott_2023, chen_mcc-kd_2023}, selects or chunks
rationales \citep{yan_towards_2025, chen_skip-thinking_2025}, distills agent trajectories
\citep{kang_distilling_2025}, and asks which rationale properties matter
\citep{wadhwa-etal-2024-investigating, chen_unveiling_2025}. All transfer text, not the mechanism
that uses it. \mistill{} keeps the sequence-level loss \citep{kim2016sequence} and adds a
target for that mechanism.

\paragraph{Attention and representation distillation.}
Beyond output distributions \citep{hinton_distilling_2015}, students can match a teacher's hidden
states \citep{romero2015fitnets}, attention maps \citep{zagoruyko2017paying}, layer-wise attention and
hidden states \citep{jiao2020tinybert}, or last-layer self-attention and value relations
\citep{wang2020minilm}. Logit distillation for LLMs now includes contrastive
(\distillm; \citealp{ko2025distillm}), on-policy \citep{agarwal2024onpolicy} and cross-tokenizer
objectives (ULD, DSKD, \gold; \citealp{boizard2025towards, zhang2024dual, patino2025unlocking}). All
act at every position, on the output distribution or on fixed layers. \mistill{} chooses the heads (causally ranked) and
positions (at and just before the teacher's decision events), and matches pooled attention plus its interaction with the
heads' contribution to the decision margin. Circuit distillation aligns the representations of functionally corresponding circuit
components in teacher and student \citep{wadhwa2025circuit}; \mistill{} instead pools each model's own
causally ranked heads, so it needs no correspondence between them. Circuit distillation is, by its authors' own
statement, contingent on the prior identification of a well-defined circuit in the teacher: its two tasks, entity
tracking and theory of mind, take their circuits from earlier work, and student heads are mapped to teacher heads by
an ablation-based matching. \mistill{} locates its targets with its own instrument on any task whose decisions are
token events, and supervises each model's own heads, so no external circuit and no mapping are needed. KPOD
\citep{feng2024keypoint} re-weights the tokens of a black-box teacher's rationale with a learned mask and a progressive
schedule, using the teacher's text alone; the event-weighted control of Appendix~\ref{app:ablations} is the fixed form of
that idea, with the key tokens placed by the teacher's causal instrument rather than learned, and it recovers what plain
\vanillakd{} does.

\paragraph{Mechanistic interpretability.}
Causal mediation and activation patching localize behavior to components
\citep{vig2020investigating, meng2022locating}; circuit analyses find the components that implement a
task \citep{elhage2021mathematical, wang2023interpretability, conmy2023towards}; attribution patching
screens every component with a gradient approximation \citep{syed_attribution_2024, kramar_atp_2024}; and metric and corruption choices matter
\citep{zhang_towards_2024, heimersheim_how_2024}. Retrieval heads fetch information from long
contexts \citep{wu_retrieval_2024}. New here are a full inventory of thousands of codes in context,
decisions as token-level events scored by the model's own margin, and per-note necessity and knock-in
sufficiency against a null calibration.

\paragraph{ICD coding and extreme classification.}
ICD coders are typically supervised multi-label classifiers, often with label-wise attention
\citep{mullenbach2018explainable}; \citet{edin2023automated} review and replicate them on MIMIC-III and
MIMIC-IV \citep{johnson2023mimic}. Large language models code poorly when asked for codes directly
\citep{soroush2024large}, and recent agentic systems implement the official coding guidelines over the full ICD-10
system \citep{motzfeldt2025code}. Coding is extreme multi-label classification, whose
benchmarks \citep{Bhatia16} report propensity-scored metrics that up-weight rare labels
\citep{jain2016extreme}; recent work optimizes long-tail and macro-at-$k$ metrics directly
\citep{schultheis2023generalized, schultheis2024consistent}, trains dual encoders over the label space
\citep{gupta2024dual}, and scales training in large output spaces with dynamic sparsity and low precision
\citep{ullah2024navigating, zhang2025elmo, ullah2026haste}. LLMs tackle it by generating or retrieving candidate labels
in context \citep{zhu_icxml_2024, doosterlinck_-context_2024}. Interpretability in coding has targeted trained
classifiers \citep{mullenbach2018explainable, wu_dila_2025, edin2024unsupervised}; we ask how a reasoning-trained model
selects codes from the full inventory in context, and whether that can be distilled.

\section{Material moved from the main text}
\label{app:moved}
\supports{Sec.~\ref{sec:setup-measure}, Sec.~\ref{sec:char}, Sec.~\ref{sec:distill}; \resid{R-C2}, \resid{R-C3}, \resid{R-C4}, \resid{R-C5}, \resid{R-C6}, \resid{R-D6}}
This version rewrites the abstract and Sections~\ref{sec:intro} to~\ref{sec:distill} of the earlier versions in plainer
words and brings a compact worked example into the main text (Figure~\ref{fig:events}); the rewrite added and removed
no result, number or claim. What left the main text, for the rewrite and to make room for the controls of
Appendix~\ref{app:teacherpool}, is given here in full, in the order of the sections it left; every number is also in
the appendix or the table named beside it.

\begin{figure*}[!htbp]
  \centering
  \includegraphics[width=\linewidth]{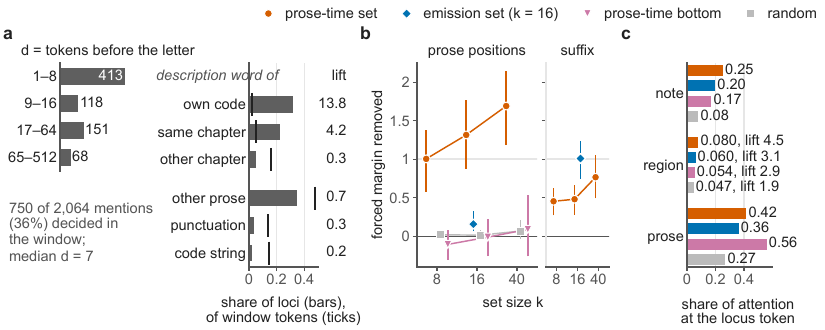}
  \caption{Where the chapter can be elicited, and by which heads. (a) Mentions with a locus in the reasoning: distance $d$ from
  locus to chapter letter, and the classes of locus token (bars: share of loci; ticks: share of window tokens; right,
  lift). (b) Share of the forced chapter margin removed when a fixed set is mean-ablated only at the window positions
  (left) or only at the probe suffix (right): prose-time top-$k$, emission top-16, prose-time bottom-$k$, random-$k$;
  medians, whiskers 10th--90th percentile. (c) Attention at the locus token (natural context), four 16-head panels;
  descriptive. \resid{R-C6}}
  \label{fig:locus}
\end{figure*}

\paragraph{The locus figure (from Section~\ref{sec:char-locus}).} Figure~\ref{fig:locus} stood in
Section~\ref{sec:char-locus}; the worked example took its place. The figure and its caption are unchanged, and
Section~\ref{sec:char-locus} still states the numbers it cited from the figure.

\paragraph{From Section~\ref{sec:setup-measure} (the instrument).} The correlation of first-order attribution with exact
effects is \NMAttrPearsonAllDetached{} under the hook used at the code token (which detached the graph at every layer)
and \NMAttrPearsonAllIntact{} with the graph intact; every set is therefore judged by exact intervention. Forward passes
vary by \NMNoiseFloorNats{} nats (s.d.), so we report means over notes. Knocking in all 1{,}600 heads returns the clean
margin (median difference \NMKnockinCeilingDiff{} nats). Details: Appendix~\ref{app:instrument}.

\paragraph{From Section~\ref{sec:char-sets} (the global set against random, note by note).} For chapter and character
decisions the global set beats random on \NFaithPOneNecKThreeWins{}--\NFaithPOneNecKFortyWins{} of 100 notes for
necessity and on \NFaithPOneKiKEightWins{} for knock-in (contrastive: \NFaithPTwoCNecKEightWins{} and
\NFaithPTwoCKiKEightWins{} of 92; \NFaithPTwoCKiKThreeWins{} at $k=3$). These are the ``notes'' columns of
Tables~\ref{tab:faithfulness} and~\ref{tab:faithfulness-app}. As a carrier the global set beats the note's own ranking
only for the chapter decision at $k\ge 8$ (\NFaithPOneKiKEightGlobal{} against \NFaithPOneKiKEightOwn{}); for characters
the note's own ranking, fitted to the events it is tested on, is better at every $k$ (Table~\ref{tab:app-own}).

\paragraph{From Section~\ref{sec:char-phases} (the named heads).} \head{39}{38} is a shared core, first by $I_c$ in every
phase (first-order at \Ptwoc). Around it are chapter-specific heads (\head{26}{36}, \head{26}{37}, \head{27}{11},
\head{27}{13}; the last two also gain character-level rank under the intact hook, Table~\ref{tab:app-p2intact}),
character-specific heads (\head{39}{37}, \head{39}{12}, \head{38}{22}, \head{36}{10}) and contrastive-specific heads
(\head{30}{11}, \head{38}{39}), which rank near the bottom for chapter and character decisions. Exact ablation of the
contrastive ranking's 56 nominated heads reorders its top (\head{39}{12}, \head{39}{10}, \head{30}{11}, \head{39}{38},
in raw importance). The full head tables are in Appendix~\ref{app:heads}.

\paragraph{From Section~\ref{sec:char-read} (what the heads read).} The region readers matter more on events where
they read more of the emitted chapter's inventory lines (correlation \NReadLTwentySixHThirtySixR{}). The prose readers
at the code token are
\head{39}{38}, \head{39}{13}, \head{38}{24} and \head{27}{11}. At contrastive events \head{30}{11} puts more of its
attention on the earlier mentions of the near-miss than on the code's own on \NMLThirtyHElevenSiblingOverOwnNotes{} of
92 notes. Details: Appendix~\ref{app:readside}.

\paragraph{From Section~\ref{sec:char-locus} (the locus).} In \NLocusOwnDescPct{} of the cases with a locus in the
reasoning, the locus is a word of the code's own inventory description (lift \NLocusOwnDescLift{}). The prose-time set
is led by \head{24}{19}, \head{22}{16} and \head{24}{1}. Ablated only at the window positions, its top-8 removes
\NLocusProseWinKEight{} of the forced margin, above random (\NLocusRandWinKEight{}) on \NLocusProseWinKEightWins{} of
100 notes. Mentions that inherit the chapter or never become decisive are not attributed. Details:
Appendix~\ref{app:locus}.

\paragraph{From Section~\ref{sec:distill-objective} (the objective).} $\mathcal{L}_{\mathrm{CoT}}$ is exactly the
\vanillakd{} loss: cross-entropy on the trace tokens plus a KL term to the teacher's stored top-32 logits for
\qwenfour; a universal logit distillation term for \smol, whose vocabulary differs. The teacher's side of the loss is
precomputed once. In the interaction term, $c_q$ is the pooled heads' direct contribution to the event's margin: their
writes projected on the margin's gradient direction in the unembedding, divided by $|\margin_q|$ (floored at one
nat). \featkd{} also matches bucket-pooled attention, over all heads of four mapped layer pairs at sampled positions.

\paragraph{From Section~\ref{sec:distill-setup} (the setup).} Every arm samples in thinking mode with the teacher's
settings and a 16{,}384-token cap; \qwenfour{} runs under YaRN \citep{peng2024yarn} $\times 2$ for up to 52k tokens
(Appendix~\ref{app:base-eval}). The shortlist comes from a linear head on frozen chunk embeddings fused with
zero-shot description similarity (Appendix~\ref{app:data}). \gold{} trains one epoch with rollouts capped at
6{,}144 tokens; its lists parse (valid rate \NSmolGOLDValid{}) yet score zero, naming categories rather than billable
codes (Appendix~\ref{app:res-dissoc}): the metric tiers must be kept apart. On the 150 paired test notes,
\qwenfour{}'s \vanillakd{} and \mistill{} (both precisions), \featkd{} and \smol{}'s \mistill{} have three seeds,
\smol{}'s \vanillakd{} two, every other cell one. Task and fidelity
differences are tested per note; the corpus-level interval, notes resampled, is reported beside it, its endpoints within
\NCorpusMaxGapToPerNote{} (Appendix~\ref{app:results}).

\paragraph{From Section~\ref{sec:distill-mech} and Table~\ref{tab:mech} (own-heads recovery).} \smol{}'s undistilled
\Ptwoc{} recovery, \NOwnSmolBasePTwoCKiForty, is a median over notes; \mistill's \NOwnSmolMistillSeedMeanPTwoCKiForty{}
is the mean of three seeds' medians (Table~\ref{tab:app-mech-seeds}). As means over notes, the per-note three-seed
recovery is \NOwnSmolMistillThreeSeedPTwoCKiForty{} [\NOwnSmolMistillThreeSeedPTwoCKiFortyLo,
\NOwnSmolMistillThreeSeedPTwoCKiFortyHi] against the untrained \NOwnSmolBasePTwoCKiFortyMean. The note-clustered tests
give $p$ \NPoolSmolMistillVsVanillaBfPTwoCClusterP{} for \smol{} against bf16 \vanillakd{} and $p$
\NPoolQwenMistillVsVanillaBfPTwoCClusterP{} for \qwenfour. On \smol{}'s completed own traces \mistill's recovery runs
above its teacher-forced value (Appendix~\ref{app:res-owntrace}).

\paragraph{From Section~\ref{sec:distill-ablation} (the controls).} With the student's own top heads supervised at
random positions (\randpos), the heads' reading of the code's prefix collapses together with the recovery
(Appendix~\ref{app:ablations}). On \qwenfour, random heads at random positions (RandBoth,
\NOwnQwenRandBothSZeroPTwoCKiForty) are level with every \mistill{} seed (Appendix~\ref{app:abl-tiers}), which places
\randpos's harm in the student's own ranked heads at random positions; on \smol{} they do not fall below the untrained
student either, but gain nothing (Appendix~\ref{app:teacherpool}). The uniform control applies the attention term to every head of the pool layers, with no
interaction term, at $\lambda$ 0.135. The event-weighted control up-weights the token loss at the same events and
window positions. \Ptwo{} and \Ptwoc{} events are located by character offsets and mapped to each tokenizer. On \qwenfour, whose
untrained heads already restore \NMQwenBasePctTeacher{} as much as the teacher's, there is little room, and the
choice of supervised heads barely matters. Neither M2 nor the prefix-reading share follows recovery across the
ablations of phases, terms and dose, so the claims rest on own-heads recovery (Appendix~\ref{app:ablations}). The
ranked configuration has a better observed task--recovery trade-off than the uniform control, which also differs from
it in term composition, head count and weight (Appendix~\ref{app:ablations}). When the teacher-side control took a paragraph of its own
in Section~\ref{sec:distill-ablation}, these went out to pay for its lines: on \smol{} the untrained student's \Ptwoc{}
recovery is \NOwnSmolBasePTwoCKiForty{} and its random floor \NOwnSmolBasePTwoCKiFortyRand{} (Table~\ref{tab:mech});
on \qwenfour{} eight, forty or a hundred top heads reproduce \mistill, and the four random-head runs cover three pools
(Appendix~\ref{app:abl-tiers}); the uniform control's M1 at \Ptwoc{} is \NQwenUniformMOnePTwoC{} against \mistill's
\NQwenMistillSZeroMOnePTwoC; the matched-selection differences are in own-heads \Ptwoc{} recovery and corpus \microf,
over three seed pairs, note-clustered, conditional on the selection rule and the pool
(Appendix~\ref{app:abl-matched}); and \randheads{} and \vanillakd{} seeds spread more than \mistill's as well as
separating per note, so seed stability tracks the student's ranked heads (Appendix~\ref{app:ablations}).

\section{Further controls on \smol: random placement, the teacher's pools, and a positive control}
\label{app:teacherpool}
\supports{Sec.~\ref{sec:distill-ablation}; \resid{R-D58}, \resid{R-D59}, \resid{R-D60}}
Section~\ref{sec:distill-ablation} and Appendix~\ref{app:ablations} vary the student's side of the loss: which of its
heads are supervised, and where. Two cells were missing. On \smol{}, the student whose recovery has to be built, random
heads had been tried only at the teacher's events; and on neither student had the \emph{teacher's} ranked pools, the
heads Section~\ref{sec:char} identifies, been replaced. Both were run after the controls of Appendix~\ref{app:ablations}, each
pre-registered with its readings before any job, together with a positive control. Every arm is \smol{} under
\mistill's recipe (bf16, 493 notes, two epochs, $\lambda=0.1$, the student's own ranked top-40 unless stated), read as
in Appendix~\ref{app:ablations}: own-heads \Ptwoc{} knock-in recovery of the arm's own top-40, re-ranked after training,
on the 150 test notes, paired per note against each \mistill{} seed; an arm is \emph{below} \mistill{} only if more
notes are lower against every seed, each with sign-test $p<0.05$. Table~\ref{tab:app-teacherpool} gives every cell. Table~\ref{tab:app-link}
sets them beside the earlier controls of both students, by what each replaces.

\begin{table}[!htbp]
  \centering\small
  \caption{What each control of Section~\ref{sec:distill-ablation} replaces, and the own-heads \Ptwoc{} recovery of
  the student trained that way (top-40, heads found after training; medians over the 150 test notes; bf16; one seed
  per value, ranges over seeds, pools or control targets; ---: not run). The first block replaces the
  teacher's side of the loss (its positions, its ranked heads) or the attention terms, the second the student's ranked heads. Cells on \smol{}
  from this appendix: Table~\ref{tab:app-teacherpool}; the others: Table~\ref{tab:app-ablate}.
  \resid{R-D7}\resid{R-D13}\resid{R-D23}\resid{R-D24}\resid{R-D26}\resid{R-D58}\resid{R-D59}}
  \label{tab:app-link}
  \resizebox{\linewidth}{!}{\begin{tabular}{@{}llcc@{}}
    \toprule
    Replaced & by & \smol & \qwenfour \\
    \midrule
    nothing: \mistill{} (three seeds) & & \NOwnSmolMistillSTwoPTwoCKiForty--\NOwnSmolMistillSZeroPTwoCKiForty & \NOwnQwenMistillSOnePTwoCKiForty--\NOwnQwenMistillSTwoPTwoCKiForty \\
    \midrule
    the teacher's positions & random positions (\randpos) & \NOwnSmolRandPosSZeroPTwoCKiForty & \NOwnQwenRandPosSZeroPTwoCKiForty \\
    the positions and the student's heads & both random (RandBoth) & \NTpRbSZeroMed, \NTpRbSOneMed & \NOwnQwenRandBothSZeroPTwoCKiForty \\
    the teacher's ranked heads & unselected heads of the same layers & \NTpCtlMin--\NTpCtlMax & untested \\
    the attention terms & token loss up-weighted at the positions & --- & \NOwnQwenEventWeightedSZeroPTwoCKiForty \\
    \midrule
    the student's ranked heads & forty random heads (\randheads) & \NOwnSmolRandHeadsSZeroPTwoCKiForty & \NOwnQwenRandHeadsPoolSOnePTwoCKiForty--\NOwnQwenRandHeadsSZeroPTwoCKiForty \\
     & its forty most negative heads & \NOwnSmolBottomSOnePTwoCKiForty, \NOwnSmolBottomSZeroPTwoCKiForty & \NOwnQwenBottomSZeroPTwoCKiForty \\
     & every head of its pool layers, attention term only & --- & \NOwnQwenUniformSZeroPTwoCKiForty \\
    \midrule
    no attention supervision: untrained; \vanillakd & & \NOwnSmolBasePTwoCKiForty; \NOwnSmolVanillaBfSZeroPTwoCKiForty & \NOwnQwenBasePTwoCKiForty; \NOwnQwenVanillaBfSZeroPTwoCKiForty--\NOwnQwenVanillaBfSTwoPTwoCKiForty \\
    \bottomrule
  \end{tabular}}
\end{table}

\paragraph{Random heads at random positions (RandBoth) gain nothing on \smol. \resid{R-D58}} Two runs, with different
random pools, positions and data order, restore \NTpRbSZeroMed{} and \NTpRbSOneMed: below every \mistill{} seed (at
least \NTpRbMinLower{} of 150 notes lower), below random heads at the teacher's events (\randheads{}
\NOwnSmolRandHeadsSZeroPTwoCKiForty; \NTpRbSZeroVsRandHeadsDelta{} and \NTpRbSOneVsRandHeadsDelta{} per note, lower
on \NTpRbSZeroVsRandHeadsLower{} and \NTpRbSOneVsRandHeadsLower{} notes; the first run shares \randheads's pool) and
below bf16 \vanillakd; neither differs from the untrained student ($p$ \NTpRbSZeroBaseP{} and \NTpRbSOneBaseP), nor
the two from each other ($p$ \NTpRbSeedsP). The stated expectation, a level near \randheads's, failed. So on this
student random heads gain only at the teacher's events. On \qwenfour{} RandBoth matches \mistill{}
(\NOwnQwenRandBothSZeroPTwoCKiForty; Appendix~\ref{app:abl-tiers}); that student's untrained heads already restore
\NOwnQwenBasePTwoCKiForty. The sentence ``random heads at random positions do no such harm'' therefore holds on both
students only in its literal sense: neither falls below its untrained level.

\paragraph{The teacher's ranked pools replaced by unselected heads. \resid{R-D59}} The student's side is \mistill's,
unchanged. On the teacher's side the pools of the three kinds of event are replaced, by every head of the ranked pool's
own layers with equal weights (\emph{uniform}), or by forty heads drawn at random from those layers, the ranked heads
excluded, carrying the ranked weight profile (\emph{random}; two pools); the prose-time pool of the window is kept.
Each control is run in two scopes: only the attention target is built from the control pools (\emph{attention}), or the
interaction target too (\emph{full}). Before any student was trained, the control targets were checked to differ from
the ranked one (median KL at contrastive events \NTpKlUni, \NTpKlRandZero{} and \NTpKlRandOne{} nats). All seven
cells are below every \mistill{} seed (\NTpCtlMin{} to \NTpCtlMax{} against \NOwnSmolMistillSTwoPTwoCKiForty{} to
\NOwnSmolMistillSZeroPTwoCKiForty; at least \NTpCtlMinLower{} of 150 notes lower) and none differs from bf16
\vanillakd{} (\NOwnSmolVanillaBfSZeroPTwoCKiForty) by the sign test; against the untrained student they are level or a
little higher (medians \NTpTrZeroABaseDelta{} to \NTpTuASZeroBaseDelta{} per note). The pre-registered readings that the ranked pools
matter to the loss hold; the reading that they do not, which we had stated as the likelier one, does not. With the
student's side as in \mistill, the gain over chain-of-thought distillation is therefore not obtained from unselected
heads of the same layers: on this student the heads the characterization identifies contribute as the \emph{source of
the target}, while which student heads receive it matters less (Section~\ref{sec:distill-ablation}).

\paragraph{What the control does not show.} (i) The seven cells are three control targets, not seven tests: the two
scopes of a pool do not differ per note (at most \NTpScopeMaxDiff; correlation \NTpScopeCorrMin{} to
\NTpScopeCorrMax), so the attention pattern is not separated from the interaction target. (ii) The control targets
put \NTpRegionUni{} (uniform), \NTpRegionRandZero{} and \NTpRegionRandOne{} (random) of their pooled attention on
the decision's region at contrastive events, against the ranked pools' \NTpRegionRanked, and carry almost no direct
contribution to the margin (median $|c|$ \NTpWriteCtlMin{} to \NTpWriteCtlMax{} against \NTpWriteRanked); the
uniform cells sit \NTpUniOverRandMin{} to \NTpUniOverRandMax{} above the random ones, in the order of that region
mass. Which property of the ranked target is needed (its attention pattern, its write target, or its mass on the
region) is not established, and a target with the ranked region mass built without the ranked heads was not tried.
(iii) The controls are \emph{unselected} heads. Nothing here shows that the order or the weights of the ranking matter,
or that another selection of teacher heads (ranks 41 to 80, the most negatively attributed forty, the ranked forty with
equal weights, heads chosen by region mass) would fail. (iv) The students did fit the control targets: the logged
attention loss at character events falls from \NTpFitUniStart{} to \NTpFitUniEnd{} (uniform),
\NTpFitRandZeroStart{} to \NTpFitRandZeroEnd{} and \NTpFitRandOneStart{} to \NTpFitRandOneEnd{} (random), against
\NTpFitMistillStart{} to \NTpFitMistillEnd{} for \mistill. (v) At top-8 the control cells fall below the untrained
student (\NTpNegShareMin{} to \NTpNegShareMax\% of notes with negative recovery, \NTpNegShareMistill\% for
\mistill); \mistill{} itself has no top-8 gain on this student, so the top-8 readout is not offered as support. (vi)
No task claim follows: \microf{} differs from \mistill's by \NTpTaskDeltaMin{} to \NTpTaskDeltaMax, every interval
includes zero, three cells are lower by the Wilcoxon test, and the task is near its floor on this student. (vii) Not
run: the control on \qwenfour, where random heads at random positions match \mistill{} at top-40; an arm unranked on
both sides at the events (\featkd{} restores \NOwnSmolFeatKDSZeroPTwoCKiForty); attention-only and interaction-only arms
with the ranked target on \smol; any control of the span buckets; a second \randheads{} pool. No read-side pass was
run for these cells. The decision events themselves are located by parsing the code string
(Section~\ref{sec:setup-events}), not by measuring heads.

\paragraph{Positive control: \mistill{} through the code that ran these cells. \resid{R-D60}} The nine cells above were
trained and read with a later version of the code than the \mistill{} seeds they are compared with. Two cells,
pre-registered with pass rules, exclude that change of code as the reason for their low levels. \mistill{} seed 0's stored
weights, read again through the later code, restore \NTpPcRMed{} (stored read
\NOwnSmolMistillSZeroPTwoCKiForty; \NTpPcRVsSZeroDelta{} per note). \mistill{} seed 0 retrained from the stored
targets (its logged loss terms within \NTpPcTrainMaxRel\% of the original run's through step 110 of 124) restores
\NTpPcTMed{} (\NTpPcTVsSZeroDelta{} per note against the stored seed, not detected; \microf{} \NTpPcTMicroF).
Both are level with \mistill{} by the reading rule, their per-note correlation with the stored seed is \NTpPcCorr, and
each of the nine cells is below both (\NTpPcLowerMin{} to \NTpPcLowerMax{} of 150 notes lower). This settles the
code question only; it adds nothing to what the nine cells show.

\begin{table}[!htbp]
  \centering\footnotesize
  \caption{The further controls on \smol{} (bf16; 150 test notes; one seed unless stated). \Ptwoc{} top-40: own-heads
  knock-in recovery of the arm's own top-40, median over notes. Lower than \mistill: notes (of 150) on which the cell is
  lower than \mistill{} seed 0\,/\,1\,/\,2, paired per note (the positive controls are level with \mistill{} by the
  reading rule). Lower than PC: notes on which it is lower than the positive controls PC-R\,/\,PC-T. $\Delta$
  \vanillakd: median per-note difference from bf16 \vanillakd{} (sign-test $p$). \Ptwoc{} top-8 and \Ptwo{} top-40:
  medians. $\Delta$ \microf: against \mistill's seed-averaged notes [95\% interval].
  \resid{R-D58}\resid{R-D59}\resid{R-D60}}
  \label{tab:app-teacherpool}
  \providecommand{\gensd}[2]{\begin{tabular}[t]{@{}c@{}}#1\\[-2pt]{\scriptsize$\pm$#2}\end{tabular}}
\providecommand{\genstack}[2]{\begin{tabular}[t]{@{}c@{}}#1\\[-2pt]{\scriptsize #2}\end{tabular}}
\providecommand{\genstackl}[2]{\begin{tabular}[t]{@{}l@{}}#1\\[-2pt]{\scriptsize #2}\end{tabular}}
\providecommand{\genhead}[2]{\begin{tabular}[b]{@{}c@{}}#1\\#2\end{tabular}}
\providecommand{\genstackiii}[3]{\begin{tabular}[t]{@{}c@{}}#1\\[-2pt]{\scriptsize #2}\\[-3pt]{\scriptsize #3}\end{tabular}}
\setlength{\tabcolsep}{4pt}
\begin{adjustbox}{max width=\linewidth}
\begin{tabular}{@{}l c c c c c c c@{}}
\toprule
 & \Ptwoc{} & lower than & lower than & $\Delta$ \vanillakd{} & \Ptwoc{} & \Ptwo{} & \\
Cell & top-40 & \mistill & PC & (bf16) ($p$) & top-8 & top-40 & $\Delta$ \microf \\
\midrule
\multicolumn{8}{@{}l}{\emph{References (stored reads)}} \\
\mistill, seed 0 & \NOwnSmolMistillSZeroPTwoCKiForty & --- & --- & --- & \NOwnSmolMistillSZeroPTwoCKiEight & \NOwnSmolMistillSZeroPTwoKiForty & --- \\
\mistill, seed 1 & \NOwnSmolMistillSOnePTwoCKiForty & --- & --- & --- & \NOwnSmolMistillSOnePTwoCKiEight & \NOwnSmolMistillSOnePTwoKiForty & --- \\
\mistill, seed 2 & \NOwnSmolMistillSTwoPTwoCKiForty & --- & --- & --- & \NOwnSmolMistillSTwoPTwoCKiEight & \NOwnSmolMistillSTwoPTwoKiForty & --- \\
\vanillakd{} (bf16) & \NOwnSmolVanillaBfSZeroPTwoCKiForty & --- & --- & --- & \NOwnSmolVanillaBfSZeroPTwoCKiEight & \NOwnSmolVanillaBfSZeroPTwoKiForty & --- \\
untrained & \NOwnSmolBasePTwoCKiForty & --- & --- & --- & \NOwnSmolBasePTwoCKiEight & \NOwnSmolBasePTwoKiForty & --- \\
\randheads{} (at the events) & \NOwnSmolRandHeadsSZeroPTwoCKiForty & --- & --- & --- & \NOwnSmolRandHeadsSZeroPTwoCKiEight & \NOwnSmolRandHeadsSZeroPTwoKiForty & --- \\
\midrule
\multicolumn{8}{@{}l}{\emph{Random student heads at random positions; the teacher's ranked target}} \\
RandBoth, seed 0 & \NTpRbSZeroMed & \NTpRbSZeroLowerSZero\,/\,\NTpRbSZeroLowerSOne\,/\,\NTpRbSZeroLowerSTwo & \NTpRbSZeroLowerPcR\,/\,\NTpRbSZeroLowerPcT & \NTpRbSZeroBfDelta{} (\NTpRbSZeroBfP) & \NTpRbSZeroKEight & \NTpRbSZeroPTwo & \pci{\NTpRbSZeroTaskDelta}{\NTpRbSZeroTaskLo}{\NTpRbSZeroTaskHi} \\
RandBoth, seed 1 & \NTpRbSOneMed & \NTpRbSOneLowerSZero\,/\,\NTpRbSOneLowerSOne\,/\,\NTpRbSOneLowerSTwo & \NTpRbSOneLowerPcR\,/\,\NTpRbSOneLowerPcT & \NTpRbSOneBfDelta{} (\NTpRbSOneBfP) & \NTpRbSOneKEight & \NTpRbSOnePTwo & \pci{\NTpRbSOneTaskDelta}{\NTpRbSOneTaskLo}{\NTpRbSOneTaskHi} \\
\midrule
\multicolumn{8}{@{}l}{\emph{The student's side as in \mistill; the teacher's event pools replaced (pools, scope)}} \\
uniform, attention, seed 0 & \NTpTuASZeroMed & \NTpTuASZeroLowerSZero\,/\,\NTpTuASZeroLowerSOne\,/\,\NTpTuASZeroLowerSTwo & \NTpTuASZeroLowerPcR\,/\,\NTpTuASZeroLowerPcT & \NTpTuASZeroBfDelta{} (\NTpTuASZeroBfP) & \NTpTuASZeroKEight & \NTpTuASZeroPTwo & \pci{\NTpTuASZeroTaskDelta}{\NTpTuASZeroTaskLo}{\NTpTuASZeroTaskHi} \\
uniform, attention, seed 1 & \NTpTuASOneMed & \NTpTuASOneLowerSZero\,/\,\NTpTuASOneLowerSOne\,/\,\NTpTuASOneLowerSTwo & \NTpTuASOneLowerPcR\,/\,\NTpTuASOneLowerPcT & \NTpTuASOneBfDelta{} (\NTpTuASOneBfP) & \NTpTuASOneKEight & \NTpTuASOnePTwo & \pci{\NTpTuASOneTaskDelta}{\NTpTuASOneTaskLo}{\NTpTuASOneTaskHi} \\
uniform, full & \NTpTuFMed & \NTpTuFLowerSZero\,/\,\NTpTuFLowerSOne\,/\,\NTpTuFLowerSTwo & \NTpTuFLowerPcR\,/\,\NTpTuFLowerPcT & \NTpTuFBfDelta{} (\NTpTuFBfP) & \NTpTuFKEight & \NTpTuFPTwo & \pci{\NTpTuFTaskDelta}{\NTpTuFTaskLo}{\NTpTuFTaskHi} \\
random pool 0, attention & \NTpTrZeroAMed & \NTpTrZeroALowerSZero\,/\,\NTpTrZeroALowerSOne\,/\,\NTpTrZeroALowerSTwo & \NTpTrZeroALowerPcR\,/\,\NTpTrZeroALowerPcT & \NTpTrZeroABfDelta{} (\NTpTrZeroABfP) & \NTpTrZeroAKEight & \NTpTrZeroAPTwo & \pci{\NTpTrZeroATaskDelta}{\NTpTrZeroATaskLo}{\NTpTrZeroATaskHi} \\
random pool 0, full & \NTpTrZeroFMed & \NTpTrZeroFLowerSZero\,/\,\NTpTrZeroFLowerSOne\,/\,\NTpTrZeroFLowerSTwo & \NTpTrZeroFLowerPcR\,/\,\NTpTrZeroFLowerPcT & \NTpTrZeroFBfDelta{} (\NTpTrZeroFBfP) & \NTpTrZeroFKEight & \NTpTrZeroFPTwo & \pci{\NTpTrZeroFTaskDelta}{\NTpTrZeroFTaskLo}{\NTpTrZeroFTaskHi} \\
random pool 1, attention & \NTpTrOneAMed & \NTpTrOneALowerSZero\,/\,\NTpTrOneALowerSOne\,/\,\NTpTrOneALowerSTwo & \NTpTrOneALowerPcR\,/\,\NTpTrOneALowerPcT & \NTpTrOneABfDelta{} (\NTpTrOneABfP) & \NTpTrOneAKEight & \NTpTrOneAPTwo & \pci{\NTpTrOneATaskDelta}{\NTpTrOneATaskLo}{\NTpTrOneATaskHi} \\
random pool 1, full & \NTpTrOneFMed & \NTpTrOneFLowerSZero\,/\,\NTpTrOneFLowerSOne\,/\,\NTpTrOneFLowerSTwo & \NTpTrOneFLowerPcR\,/\,\NTpTrOneFLowerPcT & \NTpTrOneFBfDelta{} (\NTpTrOneFBfP) & \NTpTrOneFKEight & \NTpTrOneFPTwo & \pci{\NTpTrOneFTaskDelta}{\NTpTrOneFTaskLo}{\NTpTrOneFTaskHi} \\
\midrule
\multicolumn{8}{@{}l}{\emph{Positive control: \mistill{} seed 0 through the later code}} \\
PC-R: stored weights, read again & \NTpPcRMed & \NTpPcRLowerSZero\,/\,\NTpPcRLowerSOne\,/\,\NTpPcRLowerSTwo & --- & --- & \NTpPcRKEight & \NTpPcRPTwo & --- \\
PC-T: retrained & \NTpPcTMed & \NTpPcTLowerSZero\,/\,\NTpPcTLowerSOne\,/\,\NTpPcTLowerSTwo & --- & --- & \NTpPcTKEight & \NTpPcTPTwo & \pci{\NTpPcTTaskDelta}{\NTpPcTTaskLo}{\NTpPcTTaskHi} \\
\bottomrule
\end{tabular}
\end{adjustbox}

\end{table}

\paragraph{Reproduction entries.} As in Appendix~\ref{app:repro}; the plans, with the pre-registered readings, their
amendments and the logs, are in the code release.
\begin{description}[style=unboxed,leftmargin=0pt,itemsep=3pt,parsep=1pt,topsep=4pt,font=\normalfont\small\sffamily\bfseries]
\item[\residtarget{R-D58}\quad Random heads at random positions on \smol.] Sec.~\ref{sec:distill-ablation},
Appendix~\ref{app:teacherpool}, Table~\ref{tab:app-teacherpool}. \emph{Artifacts:}
\path|DT-001/{runs,eval}/SmolLM3-3B__mech_randboth_t1__s{0,1}/| and their \path|M6-| and \path|HD-S-| cells;
\path|DT-008/reads/|. \emph{Run:} \path|slurm/dt001_grid.sh| (\texttt{MODE=tier1}), \path|slurm/dt001_s6_arm.sbatch|.
\emph{Reader:} \path|scripts/dt009/read_cell.sh| (\path|scripts/m4_paired.py|), \path|scripts/dt009/read_task.sh|.
\emph{Expected:} \NTpRbSZeroMed{} and \NTpRbSOneMed, below every \mistill{} seed and \randheads.
\emph{Nondeterminism:} as R-D7.
\item[\residtarget{R-D59}\quad The teacher's pools replaced.] Sec.~\ref{sec:distill-ablation},
Appendix~\ref{app:teacherpool}, Table~\ref{tab:app-teacherpool}. \emph{Artifacts:} the control targets
\path|DT-009/targets_*/| and \path|DT-009/divergence_*.json|;
\path|DT-001/{runs,eval}/SmolLM3-3B__mech_{tuniA,tuni,trand0,trand1,trand0A,trand1A}_t1__s*/| and their \path|M6-| and
\path|HD-S-| cells; \path|DT-009/reads/|. \emph{Run:} \path|scripts/precompute_targets.py| (\texttt{-{}-teacher-heads}),
\path|scripts/dt009/target_divergence.py|, \path|slurm/dt001_grid.sh|, \path|slurm/dt001_s6_arm.sbatch|.
\emph{Reader:} \path|scripts/dt009/read_cell.sh|, \path|scripts/dt009/read_task.sh|. \emph{Expected:}
\NTpCtlMin{} to \NTpCtlMax{} in the seven cells, each below every \mistill{} seed. \emph{Nondeterminism:} as R-D7;
the random pools are fixed by their seeds.
\item[\residtarget{R-D60}\quad Positive control.] Appendix~\ref{app:teacherpool}, Table~\ref{tab:app-teacherpool}.
\emph{Artifacts:} \path|DT-001/runs/SmolLM3-3B__mech_{pcread,pctrain_t1}__s0/| and their \path|M6-| and \path|HD-S-|
cells; \path|DT-010/reads/|. \emph{Run:} \path|scripts/dt010/prep_pc.sh|, \path|scripts/dt010/launch_pc.sh|.
\emph{Reader:} \path|scripts/dt010/read_pc.sh| (\path|scripts/dt010/pc_outcome.py|). \emph{Expected:} \NTpPcRMed{}
and \NTpPcTMed, each above the nine cells on at least \NTpPcLowerMin{} of 150 notes. \emph{Nondeterminism:}
training is not bitwise reproducible; attention kernels.
\end{description}

\end{document}